%% file: main.tex
\documentclass[11pt,a4paper,titlepage,twoside,openright, dvipsnames]{book}

\usepackage{style/uomthesis}

\input{preamble}
\input{macro}
\input{math_commands}

\begin{document}

\setlength{\abovedisplayskip}{6pt}
\setlength{\belowdisplayskip}{6pt}

% Set spacing above and below alignment environments
\setlength{\abovedisplayshortskip}{6pt}
\setlength{\belowdisplayshortskip}{6pt}

% \thispagestyle{empty}
% \begin{picture}(0,0)
% \put(-650,-450){\includegraphics[width=13in,height=9in]{figures/Portada, contraportada y lomo para imprimir Copia Digital (1).pdf}}
% \end{picture}

% Reset to standard page geometry for the rest of the document
% \restoregeometry

    %%
    %% Front matter
    %%
    \begin{frontmatter}

        \frontmatterheadings

        \input{frontmatter/metadata.tex}

        % the title page is centered horizontally
        \symmetricmargin
        \maketitle
        \bookmargin

        % \input{frontmatter/abstract.tex}
        % \makedeclaration
        
        \input{frontmatter/summary.tex}
        \input{frontmatter/acknowledgements.tex}

        \input{frontmatter/publications.tex}

        {\singlespacing
            \tableofcontents

            % \clearpage
            % \addcontentsline{toc}{chapter}{List of Figures}
            % \listoffigures

            % \clearpage
            % \addcontentsline{toc}{chapter}{List of Tables}
            % \listoftables
        }

    \end{frontmatter}

    %%
    %% Main matter
    %%
    \hypersetup{colorlinks,citecolor={MidnightBlue},urlcolor={MidnightBlue}, linkcolor={Maroon}} 

    \begin{mainmatter}

        \mainmatterheadings

        \include{chapters/1_introduction}
        \part{\scshape Efficiency Through Continuous Modeling}
        \include{chapters/2_ckconv}
        \include{chapters/4_ccnn}
        \include{chapters/X_gridifier}

        \include{chapters/3_flexconv}
        \include{chapters/5_dnarch}
        \part{\scshape Efficiency Through Symmetry Preservation}
        \include{chapters/5_1_coattentive}

        \include{chapters/6_attgconv}
        \include{chapters/7_gselfatt}
        \include{chapters/8_waveletnets}
        \include{chapters/9_partialgcnn}
        %\include{chapters/discussion}
        \include{chapters/conclusion}

        \addcontentsline{toc}{chapter}{Bibliography}
        \bibliographystyle{plainnat}
        \bibliography{bibliography}

    \end{mainmatter}

    %%
    %% Appendix
    %%
    \begin{appendices}

    \mainmatterheadings

        \part*{\scshape Appendix}
        \include{backmatter/1_appx_ckconv}

        \include{backmatter/appx_4_ccnn}
        \include{backmatter/appx_X_gridifier}
        \include{backmatter/appx_3_flexconv}
        \include{backmatter/appx_5_dnarch}
        \include{backmatter/appx_co_attentive}
        \include{backmatter/appx_6_attgconv}
        \include{backmatter/appx_7_gselfatt}
        \include{backmatter/appx_8_waveletnets}
        \include{backmatter/appx_9_partialgcnn}

    \end{appendices}

    %%
    %% Back matter
    %%
    \glsaddall
    \begin{backmatter}
        \frontmatterheadings

        %\printglossary[title=Definition of Symbols and Acronyms]

        \clearpage
        % \addcontentsline{toc}{chapter}{Index}
        % \printindex

\include{backmatter/siks_dissertatiereeks}

    \end{backmatter}

\end{document}

%% file: preamble.tex
\usepackage[breakable, xparse, skins]{tcolorbox}
\usepackage{siunitx}

\usepackage{amssymb}
\usepackage{amsmath}
\usepackage{amsthm}
\usepackage[mathscr]{eucal}
\usepackage{nicefrac}

\usepackage{makecell}

\newtheorem{theorem}{Theorem}[section]
\newtheorem{corollary}{Corollary}[theorem]

\newtheorem{proposition}{Proposition}[section]
\newtheorem{definition}{Definition}[theorem]

\newtheorem{subdefinition}{Example}[theorem]
\newtheorem{claim}{Proposition}[section]

\usepackage{mathtools}

\usepackage{color}
\usepackage{xcolor}

\usepackage{graphicx}
\usepackage{sidecap}
\usepackage{subcaption}
\usepackage{floatrow}

\usepackage{enumitem}

\usepackage{csquotes}

\usepackage{booktabs}       % professional-quality tables
\usepackage{xltabular} % for the cross-page table
\usepackage{multicol}
\usepackage{multirow}
\usepackage{wrapfig}

\usepackage{algorithm,algpseudocode}

\usepackage{float}
\floatstyle{plaintop}
\restylefloat{table}

\usepackage{pifont}% http://ctan.org/pkg/pifont
\newcommand{\cmark}{\ding{51}}%
\newcommand{\xmark}{\ding{55}}%

\usepackage{url}
\usepackage[linktoc=all]{hyperref}
\hypersetup{
    colorlinks,
    citecolor=black,
    filecolor=black,
    linkcolor=black,
    urlcolor=black
}

\usepackage[numbers,sort&compress,square]{natbib}

\usepackage{makeidx}
\makeindex

\usepackage[toc,nonumberlist,nopostdot]{glossaries}
\makeglossaries
\setglossarystyle{treegroup}

\newcommand*\markblankpages{}

\usepackage[outer=36.02988mm,inner=25.47697mm,top=36.0317mm,bottom=50.9565mm]{geometry}
\newcommand*\symmetricmargin{
    \newgeometry{outer=30.7534mm,inner=30.7534mm,top=36.0317mm,bottom=50.9565mm}
}
\newcommand*\bookmargin{
    \newgeometry{outer=36.02988mm,inner=25.47697mm,top=36.0317mm,bottom=50.9565mm}
}

\usepackage{lipsum}

\usepackage[labelfont={bf,singlespacing},
            textfont={singlespacing},
            justification={justified},
            singlelinecheck=false,
            margin=0pt,
            figurewithin=chapter,
            tablewithin=chapter]{caption}
\usepackage[titletoc]{appendix}

\usepackage[helvetica]{quotchap} 

\usepackage[contents={}]{background}
\usepackage[yyyymmdd,hhmmss]{datetime}

\font\btt=rm-lmtk10
\newcommand{\mlp}{{\tt MLP}}
\newcommand{\mlppsi}{\mlp$^{\bpsi}$}

%% file: macro.tex
\newcommand*\cleartoleftpage{
    \clearpage
    \ifodd\value{page}\hbox{}\newpage\fi
}

%% file: math_commands.tex
\usepackage{MnSymbol}%
\usepackage{wasysym}%

\usepackage[cal=dutchcal,
 calscaled=1,
 scr=euler]{mathalfa}
 
\usepackage{dsfont}

\newcommand{\mat}[1]{\ensuremath{{\mathbf{\MakeUppercase{{#1}}}}}}
\renewcommand{\vec}[1]{\ensuremath{\mathbf{\MakeLowercase{{#1}}}}}

\newcommand{\gr}[1]{\ensuremath{\mathcal{#1}}}

\renewcommand{\gg}{\gr{g}}
\newcommand{\gi}{\gr{i}}
\newcommand{\gj}{\gr{j}}
\newcommand{\gh}{\gr{h}}

\newcommand{\Wm}{\mat{W}}
\newcommand{\Xm}{\mat{X}}
\newcommand{\Ym}{\mat{Y}}
\newcommand{\Am}{\mat{A}}

\newcommand{\Vm}{\mat{V}}
\newcommand{\Um}{\mat{U}}
\newcommand{\Pm}{\mat{P}}

\def\gA{{\mathcal{A}}}

\def\gC{{\mathcal{C}}}
\def\gD{{\mathcal{D}}}
\def\gE{{\mathcal{E}}}
\def\gF{{\mathcal{F}}}
\def\gG{{\mathcal{G}}}
\def\gH{{\mathcal{H}}}

\def\gK{{\mathcal{K}}}
\def\gL{{\mathcal{L}}}

\def\gN{{\mathcal{N}}}
\def\gO{{\mathcal{O}}}
\def\gP{{\mathcal{P}}}

\def\gS{{\mathcal{S}}}
\def\gT{{\mathcal{T}}}
\def\gU{{\mathcal{U}}}
\def\gV{{\mathcal{V}}}
\def\gW{{\mathcal{W}}}
\def\gX{{\mathcal{X}}}
\def\gY{{\mathcal{Y}}}
\def\gZ{{\mathcal{Z}}}

\def\emA{{A}}

\def\sC{{\mathbb{C}}}

\def\sL{{\mathbb{L}}}

\def\sR{{\mathbb{R}}}
\def\sS{{\mathbb{S}}}

\def\sZ{{\mathbb{Z}}}

\newcommand{\xv}{\vec{x}}
\newcommand{\bv}{\vec{b}}

\newcommand{\cv}{\vec{c}}
\newcommand{\hv}{\vec{h}}
\newcommand{\yv}{\vec{y}}
\newcommand{\gv}{\vec{g}}
\newcommand{\wv}{\vec{w}}

\newcommand{\W}{\Wm}
\newcommand{\X}{\Xm}
\newcommand{\Y}{\Ym}
\newcommand{\A}{\Am}

\newcommand{\Nin}{\mathrm{N_{in}}}
\newcommand{\Nout}{\mathrm{N_{out}}}

\newcommand{\Lt}{\mathrm{L}}
\newcommand{\Nt}{\mathrm{N}}
\newcommand{\Dt}{\mathrm{D}}
\newcommand{\Kt}{\mathrm{K}}
\newcommand{\Xt}{\mathrm{X}}
\newcommand{\Yt}{\mathrm{Y}}
\newcommand{\ct}{\mathrm{c}}

\newcommand{\eu}{\mathrm{e}\mkern1mu}
\newcommand{\du}{\mathrm{d}\mkern1mu}
\newcommand{\iu}{{i\mkern1mu}}
\newcommand{\ltwo}{\mathrm{L}^{2}}

\def\bpsi{{\boldsymbol{\psi}}}

\newcommand{\SA}{\operatorname{\mathrm{SA}}}
\newcommand{\MHSA}{\operatorname{\mathrm{MHSA}}}
\newcommand{\cat}{\operatorname*{\mathrm{concat}}}

\newcommand{\WV}{\W_{\text{val}}}
\newcommand{\WQ}{\W_{\text{qry}}}
\newcommand{\WK}{\W_{\text{key}}}
\newcommand{\WO}{\W_{\text{out}}}

\newcommand{\rmfsa}{{\btt R4M\_SA}}

\newcommand{\rfsa}{{\btt R4\_SA}}
\newcommand{\resa}{{\btt R8\_SA}}
\newcommand{\rssa}{{\btt R16\_SA}}
\newcommand{\rtsa}{{\btt R12\_SA}}
\newcommand{\zdsa}{{\btt Z2\_SA}}
\newcommand{\msa}{{\btt Z2M\_SA}}

\newcommand{\rcnn}{{\btt R4\_CNN}}
\newcommand{\zcnn}{{\btt Z2\_CNN}}
\newcommand{\gsa}{{\btt GSA-Nets}}
\newcommand{\gcnn}{{\btt G-CNNs}}

\newcommand{\eerror}{\Delta_{\mathrm{equiv}}}
\newcommand{\relu}{$\mathrm{ReLU}$}
\newcommand{\leakyrelu}{$\mathrm{LeakyReLU}$}
\newcommand{\swish}{$\mathrm{Swish}$}
\newcommand{\siren}{$\mathrm{SIREN}$}

\newcommand{\Ft}{\mathrm{F}}
\newcommand{\Ht}{\mathrm{H}}

\newcommand{\ev}{\vec{e}}

%% file: frontmatter/metadata.tex
% Dissertation title
\title{The Good, the Efficient and the Inductive Biases \hspace{15cm}{\LARGE Exploring Efficiency in Deep Learning Through the Use of Inductive Biases}
}

% Author
\author{David W. Romero}

% Date of submission
\submissionmonth{September}
\submissionyear{2024}

% Department
\department{Department of Computer Science}

% University
\university{\scshape Vrije Universiteit Amsterdam}

%% file: frontmatter/summary.tex
\begin{summary}%
The emergence of Deep Learning has marked a profound shift in the paradigm of machine learning, a change driven by the numerous breakthroughs it has achieved in recent years. However, as the field evolves and Deep Learning becomes increasingly present in everyday tools and applications, there is an increasing need to address unresolved challenges related to its efficiency and sustainability. 

This dissertation delves into the role of inductive biases, particularly continuous modeling and symmetry preservation, to address these challenges and enhance the efficiency of Deep Learning. The dissertation is structured in two main parts.

 The first part investigates continuous modeling as a tool to improve the efficiency of Deep Learning algorithms. Continuous modeling involves the idea of parameterizing neural operations directly in a continuous space. The research presented in this part highlights the substantial benefits of continuous modeling for the (\textit{i}) computational efficiency --in time and memory--, (\textit{ii}) the parameter efficiency, and (\textit{iii}) the complexity of designing neural architectures for new datasets and tasks, coined \enquote{\textit{design efficiency}}. %in a reduced complexity in thethe  Our multiple contributions establish that \textit{continuous modeling}: the idea of parameterizing neural operations directly on a continuous space, yields significant benefits in computational efficiency --both in terms of time and memory--, parameter efficiency and the complexity of designing neural architectures for novel datasets and tasks.

In the second half, the focus shifts towards the influence of symmetry preservation on the efficiency of Deep Learning algorithms. Symmetry preservation involves designing neural operations that align with the inherent symmetries of data. The research presented in this part highlights significant gains both in data and parameter efficiency through the use of symmetry preservation. However, it also acknowledges a resulting trade-off of increased computational costs. %  and parameter efficiency of Deep Learning algorithms, albeit at the cost of increased computational demands.

The dissertation concludes with a thorough critical evaluation of the research findings, openly discussing their limitations and proposing strategies to address them, informed by literature and the author's insights. It ends by identifying promising future research avenues in the exploration of inductive biases for efficiency, and their wider implications for Deep Learning.
%Concluding the dissertation, is a critical evaluation of the research findings. Here, the limitations of the work are candidly discussed abnd suggestions for overcoming these limitations are provided, drawing from both the literature, and the viewpoints of the author. The dissertation concludes by highlighting promising future directions for the exploration of inductive biases and their broader implications.
%This dissertation concludes with a critical evaluation of our work, discussing its limitations and suggesting potential solutions grounded in both literature on which I have been involved as well as the broader academic spectrum. We finish by outlining the most promising avenues for future research in the realm of inductive biases in deep learning and their broader implications.
\end{summary}

%% file: frontmatter/acknowledgements.tex
\begin{acknowledgements}
   I would like to start by expressing my profound gratitude to my advisors Mark, Erik and Jakub for their continued confidence and support in all of my endeavors during my PhD. Mark for believing in me from the very beginning and always supporting and encouraging me to grow both as a scientist and as a person. Erik for helping me improve my mathematical knowledge, rigor and precision. Without your guidance, this dissertation would not have been what it is today. And Jakub, for always being there to listen and for making my PhD journey as fun as it was! I will never forget how fun our meetings were --and your technical contributions too, of course.

   Further, I would like to thank the many great researchers I had the honor to collaborate with during my PhD, many of whom --perhaps unknowingly-- have deeply impacted the way I conduct research: Anna Kuzina, Robert-Jan Bruintjes, Jan van Gemert, Jean-Baptiste Cordonnier, Gabriele Cesa, Guillaume Sauti\`{e}re, Auke Wiggers, Taco Cohen, Suhas Lohit, Tycho van der Ouderaa, Albert Gu, Efstratios Gavves, Neil Zeghidour, Stephano Massaroli, Michael Poli, Hermann Kumbong, Putri van der Linden and Sharvaree Vadgama. I am particularly grateful to Jean-Baptiste Cordonnier for teaching me great deal of good software engineering practices; to Neil Zeghidour for his great advice on so many different topics and for encouraging me to continue valuing the principles of signal processing, regardless of how far Deep Learning has come; and to Jan van Gemert for teaching me the value of asking the right questions. I am very thankful for all of our discussions and conversations. I learned a tremendous amount from each of you. 

    To the students I had the pleasure of supervising: Joep van Genderingen, Cees Kandoorp, Meena Alfons, Alonso Urbano and especially David Knigge, who became a close collaborator and friend afterward. I am extremely proud of all your achievements. 

    To my fellow QDA, CI and VU members: Ali el Hassouni, Alessandro Zonta, Luis Silvestrin, Jan Klein, Floris den Hengst, Anne Fischer, Tariq Dam, B\"{u}lent \"{U}ndes, Jacob Kooi, Arwin Gansekoele, Louk Smalbil, Olivier Moulin, Tariq Dam, Yvonne Blokland, Frank Bennis, Vincent François-Lavet, Shujian Yu, Yura Perugachi-Diaz,  Fuda van Diggelen, Jie Luo, Babak Kargar, Mateo de Carlo, Tugay Karag\"{u}zel, Karine Miras, Anil Yaman, Guszti Eiben, and those mentioned above. I really enjoyed those Friday-drinks evenings in the old building and playing real-life \textit{Among Us!} at De Jansakkerhoeve. 

    To the many members of the UvA who made me feel at home both at the UvA and all around the globe, although I was not formally part of the it: Riccardo Valperga, Samuele Papa, Miltos Kofinas, Alex Gabel, Floor Eijkelboom, Tin Veljkovi\'{c}, David Wessels, David Vos, David Zhang, Emiel Hoogeboom, Mohammad Derakhshani, Ivona Najdenkoska and those already mentioned above.

    To my TU Berlin, Universidad Nacional and high-school friends who have always believed in me, encouraged me to pursue my dreams, and have always cherished each of my successes, regardless of how small they might seem. I am particularly thankful to Maria Paula Luna, Sebastian Barragan, David Amezquita, Sebastian Medina, Eric Sanchez, Roldany Gutierrez, David Torres, Eric Liberra, Patricia Aponte, Ludwig Winkler, Sami Ede, Lorenz Vaitl and Helge Mohn. %, who lie very close to my hearth.

    To many other people who have been crucial in my journey to become the person I am today: Ana Cecilia Vargas, Samuel Triana, Yeimi Fajardo, Giselle van Dongen, Marianne van Boxtel, Marja Hortulanus, Gert van Dongen and Ronny H\"{a}nsch. I am especially grateful to Lola Chavarro, who always welcomed me into her home with love and affection. I hope you will be with me in spirit at my defense, wherever you may be now.
    
   To my aunts, uncles, cousins and grandparents from both the Romero and Guzman sides. Your unconditional love and support has profoundly marked me and has been a cornerstone throughout my life. I am particularly grateful to Amparo Gibson, David Gibson and Yesid Romero, who have always treated me as a son of their own and have supported me to the fullest extent of their ability. 
   
   To my parents, without whom I would not be writing this thesis today. My dad, Wilson, who nurtured my curiosity, love and admiration for science since an early age, and to my mom, Olga, who taught me to believe in myself, dream big, and that with hard work I could achieve any goal I set for myself. I admire and love you from the bottom of my heart, and I aspire to be as wonderful a parent as you have been to me. 

   To my brothers, Yecid and Martin for their love, support and admiration. For keeping me up to date with what the new generation is up to and getting me addicted to new stupid games all the time. You mean the world to me.

   To Jacqueline, for her unconditional love and support. Thank you for helping me become a better person and guiding me and supporting me in taking the best decisions, both professionally and personally. I am very excited about our new adventure on the other side of the globe and look forward to sharing many more amazing experiences with you. Finally, to Carla, Hans and Mark for welcoming me into your family and for cherishing each of my successes with such enthusiasm and joy. I am truly happy to be part of your family. 
\end{acknowledgements}

%% file: frontmatter/publications.tex
\begin{publications}
\renewcommand{\thefootnote}{\fnsymbol{footnote}}

\textbf{\sc{Part I: Efficiency Through Continuous Modeling}}
\begin{itemize}
    \item \textbf{David W. Romero}, Anna Kuzina, Erik J Bekkers, Jakub M Tomczak and Mark Hoogendoorn. CKConv: Continuous kernel convolution for sequential data. In \textit{International Conference on Learning Representations}, 2022.% URL: \url{https://openreview.net/forum?id=8FhxBtXSl0}.
    \item \textbf{David W. Romero}\footnote{Equal contribution.}, David M. Knigge$^*$, Albert Gu, Erik J. Bekkers, Efstratios Gavves, Jakub M. Tomczak and Mark Hoogendoorn. Towards a General Purpose CNN for Long Range Dependencies in $N$D. In \textit{ICML Workshop on Continuous Time Methods for Machine Learning}, 2022.
    \item David M. Knigge$^*$, \textbf{David W. Romero}$^*$, Albert Gu, Efstratios Gavves, Erik J. Bekkers, Jakub M. Tomczak, Mark Hoogendoorn and Jan-Jakob Sonke. Modelling long range dependencies in $N$D: From task-specific to a general purpose CNN. In \textit{International Conference on Learning Representations}, 2023.
    \item Putri A. van der Linden$^*$, \textbf{David W. Romero}$^*$ and Erik J. Bekkers. Learned Gridification for Efficient Point Cloud Processing. In \textit{Proceedings of the 2$^{nd}$ Annual Workshop on Topology, Algebra, and Geometry in Machine Learning (ICML)}, 2023.
    \item \textbf{David W. Romero}$^*$, Robert-Jan Bruintjes$^*$, Jakub M. Tomczak, Erik J. Bekkers, Mark Hoogendoorn and Jan C. van Gemert. Flexconv: Continuous kernel convolutions with differentiable kernel sizes. In \textit{International Conference on Learning Representations}, 2022.
    \item \textbf{David W. Romero} and Neil Zeghidour. DNArch: Learning convolutional neural architectures by backpropagation. \textit{Under review}, 2023.
\end{itemize}
\textbf{\sc{Part II: Efficiency Through Symmetry Preservation}}
\begin{itemize}
    \item \textbf{David W. Romero} and Mark Hoogendoorn. Co-attentive equivariant neural networks: Focusing equivariance on transformations co-occurring in data. In \textit{International Conference on Learning Representations}, 2020.
    \item \textbf{David W. Romero}, Erik J. Bekkers, Jakub M. Tomczak and Mark Hoogendoorn. Attentive group equivariant convolutional networks. In \textit{International Conference on Machine Learning}, pp. 8188-8199. PMLR, 2020.
    \item \textbf{David W. Romero} and Jean-Baptiste Cordonnier. Group equivariant stand-alone self-attention for vision. In \textit{International Conference on Learning Representations}, 2021.
    \item \textbf{David W. Romero}, Erik J. Bekkers, Jakub M. Tomczak and Mark Hoogendoorn. Wavelet networks: Scale-translation equivariant learning from raw time-series. In \textit{Transactions
on Machine Learning Research}, 2023.
    \item \textbf{David W. Romero} and Suhas Lohit. Learning equivariances and partial equivariances from data. In \textit{Advances in Neural Information Processing Systems 35}, pp. 36466-36478, 2022.
\end{itemize}
\textbf{\sc{Other publications}}
\begin{itemize}
    \item David M. Knigge, \textbf{David W. Romero} and Erik J. Bekkers. Exploiting redundancy: Separable group convolutional networks on lie groups. In \textit{International Conference on Machine Learning} pp. 11359-11386. PMLR, 2022.
    \item Tycho van der Ouderaa, \textbf{David W. Romero} and Mark van der Wilk. Relaxing equivariance constraints with non-stationary continuous filters. In \textit{Advances in Neural Information Processing Systems 35}, pp. 33818-33830, 2022.
    \item Stefano Massaroli, Michael Poli, Daniel Y. Fu, Hermann Kumbong, \textbf{David W. Romero}, Rom N. Parnichkun, Arman Timalsina, Quinn McIntyre, Beidi Chen, Atri Rudra, Ce Zhang, Christopher R\'{e}, Stefano Ermon and Yosua Bengio. Laughing hyena distillery: Extracting compact recurrences from convolutions. In \textit{Advances in Neural Information Processing Systems}, 2023.
    \item Erik J. Bekkers, Sharvaree Vadgama, Rob Hesselink, Putri A. van der Linden and \textbf{David W. Romero}. Fast, Expressive $\mathrm{SE}(n)$ Equivariant Networks trough Weight Sharing in Position-Orientation Space. In \textit{International Conference on Learning Representations}, 2024.
    \item Alonso Urbano and \textbf{David W. Romero}. Self-Supervised Detection of Perfect and Partial Input-Dependent Symmetries. In \textit{ICML Workshop on Geometry-grounded Representation Learning and Generative Modeling}, 2024.
    % \item Riccardo Valperga, Samuele Papa, \textbf{David W. Romero} and Efstratios Gavves. Contrastive Implicit Representation Learning. \textit{Under review}, 2023.
\end{itemize}
\end{publications}

%% file: chapters/1_introduction.tex
% path to figures directory
\graphicspath{{figures/chapter-1/}}

%=========================================================================

% \begin{savequote}[75mm]
% Nulla facilisi. In vel sem. Morbi id urna in diam dignissim feugiat. Proin molestie tortor eu velit. Aliquam erat volutpat. Nullam    ultrices, diam tempus vulputate egestas, eros pede varius leo.
% \qauthor{Quoteauthor Lastname}
% \end{savequote}

\chapter{Introduction}
	\label{chapter:introduction}
	%

%=========================================================================

Before the advent of Deep Learning, the landscape of machine learning was characterized by traditional algorithms that relied heavily on handcrafted features. These algorithms lacked the capacity to automatically extract meaningful patterns from raw information, and often struggled to unravel the intricate complexities of real-world data.

The emergence of Deep Learning marked a profound shift in the paradigm of machine learning. It bestowed machines with the ability to automatically extract patterns, features and abstract concepts directly from raw data. This paradigm shift empowered researchers with an unprecedented level of abstraction for the design of machine learning algorithms, allowing them to conceptualize machine learning models in terms of the functional families these models could and should represent.

An interesting observation is that even though neural networks --the foundation of Deep Learning--, have existed for over five decades \citep{mcculloch1943logical, rosenblatt1958perceptron}, it is only in recent years that we have started to witness the true potential of these methods. The turning point came with the ImageNet Large Scale Visual Recognition Challenge in 2012 when AlexNet \citep{krizhevsky2012imagenet} took center stage. Arguably, the recent success of Deep Learning can be largely attributed to the synergistic advancement of three critical pillars, two of which are (\textit{i}) the access to vast amounts of data, and (\textit{ii}) the exponential growth in compute power.\footnote{The third pillar is crucial for the motivation and relevance of this dissertation, and will be described in detail in the following subsection.}
These two pillars underscore an important property of Deep Learning methods, which still holds today: \textit{Deep Learning thrives in rich environments, both in terms of compute and data}.

Nevertheless, these ideal conditions are far from universally met. For example, in the medical domain. Gathering expensive expert-annotated datasets for computer-aided diagnosis poses several challenges due to factors such as the cost associated with data collection and labeling, privacy concerns surrounding sensitive medical information, and variations in protocols across different institution and devices \citep{bartoletti2019ai, manne2021application}. Furthermore, the deployment and training of neural networks on devices with limited resources introduces another layer of constraints. For instance, for the democratization of these methods to ensure that communities worldwide have equitable access to them. In several parts of the world, computational power remains scarce, making it challenging to use resource-intensive models \cite{van2003digital, jordan2015machine}. Moreover, several promising applications such as autonomous driving and Internet of Things (IoT), demand for neural networks to have the capability to operate on light, portable hardware.

At a global scale, the increasing size and widespread adoption of deep neural networks present compelling reasons to address efficiency from both environmental and financial considerations. From an environmental perspective, research has revealed that the training of a --by now small-- model with $\sim$350 million parameters produces carbon emissions comparable to the lifetime emissions of five cars \cite{strubell2019energy}. From a financial standpoint, the costs of training such models currently runs over the millions of dollars \cite{knight2023openai}. As the interest in deploying large neural networks across diverse modalities and applications increases, these observations raise questions about the environmental and financial sustainability of Deep Learning approaches.

These examples highlight the importance of efficiency in the neural models that we use. The development of more efficient neural architectures holds not only the potential to unlock a myriad of real-world applications where Deep Learning can have a substantial impact, but also to enhance its globally accessibility, reduce its environmental impact, and ensure its economic viability. In doing so, we can fully leverage the power of Deep Learning to drive positive change, both in science and society.

\vspace{-7mm}
\section{Inductive Biases and Deep Learning}
In addition to the two pillars responsible for the success of Deep Learning mentioned earlier, there is a third equally important pillar: \textit{the design of better neural architectures}. 

\begin{figure}
    \centering
    \hspace{20mm}
    \begin{subfigure}[b]{0.27\textwidth}
    \captionsetup{justification=centering}
        \centering
        \includegraphics[width=\textwidth]{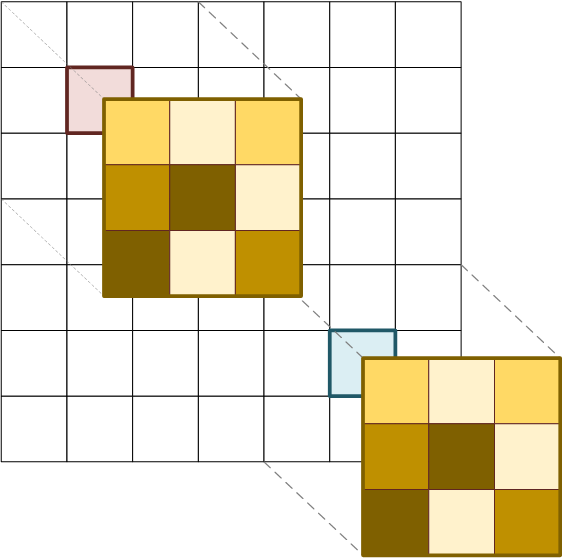}
        \caption{Convolutional layer.}
        \label{fig:0_conv}
    \end{subfigure}
    \hfill
    \begin{subfigure}[b]{0.27\textwidth}
    \captionsetup{justification=centering}
        \centering
        \includegraphics[width=\textwidth]{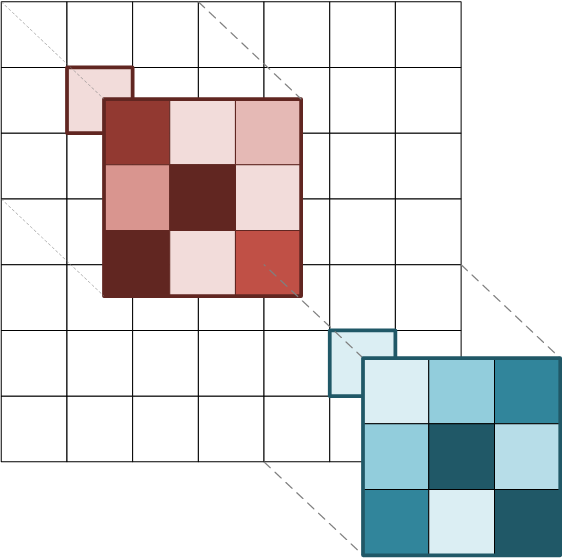}
        \caption{Fully connected layer.}
        \label{fig:0_fully}
    \end{subfigure}\hspace{20mm}
    \caption{Convolutional vs fully connected layers. Convolutional layers share the same weights across all positions. Hence, the same mapping is applied (and learned) at all positions --here depicted by weights of the same color in \ref{fig:0_conv}. In contrast, fully connected layers apply (and learn) a \textit{unique} set of weights to each position (\ref{fig:0_fully}). This increases the number of weights needed by $O(\mathrm{N^2})$ for an $\mathrm{N}{\times}{\mathrm{N}}$ image, and hampers generalization across positions. Since all features must be independently learned at each position, fully connected layers need $O(\mathrm{N^2})$ more data samples than convolutions.}
    \label{fig:0_conv_vs_fully_connected}
\end{figure}

Despite the substantial growth in data and computing power, the neural architectures that sparked the recent Deep Learning revolution differ from the original designs proposed in the early 1950s. Initial models were primarily composed of fully connected layers interleaved with point-wise nonlinearities. The turning point for the recent Deep Learning revolution came with AlexNet \cite{krizhevsky2012imagenet}: a prominent example of a \textit{Convolutional Neural Network} (CNN) \cite{fukushima1982neocognitron, lecun1989backpropagation}. The main defining characteristic of CNNs is the departure from the use of fully connected layers in favor of \textit{convolutions}: linear operations that respect the translation symmetry of visual data (Fig.~\ref{fig:0_conv}).  

Convolutional Neural Networks (CNNs) are a prime example of the use and effectiveness of considering \textit{inductive bias} in the design of neural architectures. Formally, an \textit{inductive bias} represents a set of assumptions, constraints, or prior knowledge considered in the design and training of a machine learning model such that its decision making process favors solutions that align better with the problem at hand. In the case of Convolutional Neural Networks (CNNs), the key inductive bias is \textit{translation preservation}. This inductive bias arises from the use of convolutions, which make the learned feature representations be shared across positions --instead of the independent feature representations per position learned by fully connected layers (Fig.~\ref{fig:0_fully}). As a result, CNNs learn mappings that inherently respect translation, resulting in vastly improved generalization and data efficiency.

This example underscores the potential that inductive biases may hold for the improvement of Deep Learning algorithms. In this dissertation, we focus on the role that inductive biases can play in improving various facets of Deep Learning efficiency.%we focus on their effect for multiple efficiency aspects of Deep Learning.

\vspace{-7mm}
\section{Efficiency through Inductive Biases}

When thinking about efficiency, probably the first concept that comes to mind is that of \textit{sparsity}. In the context of Deep Learning, sparsity can be seen as an inductive bias that favors small neural architectures over bigger ones, either in terms of the whole architecture \citep{wen2016learning,hoefler2021sparsity} or the subset of neurons that are activated for a particular input \citep{ghodrati2021frameexit, teerapittayanon2016branchynet, schuster2022confident}. The concept of sparsity in Deep Learning has been broadly studied in the literature, and whole dissertations have been written in the topic \citep{lee2020toward,gong2021exploiting,frankle2023lottery}. However, sparsity generally touches exclusively upon two aspects of efficiency: \textit{compute} and \textit{parameter efficiency}. In this dissertation we are interested in (complementary) inductive biases that touch upon Deep Learning efficiency in a broader sense. 

\vspace{-7mm}
\subsection{Efficiency Aspects of Deep Learning}\label{sec:0_efficiency_aspects}

Before discussing specific inductive biases, it is beneficial to define the various axes on which Deep Learning efficiency can be measured. We categorize efficiency into the following axes:
\begin{itemize}
    \item \textbf{Compute efficiency:} This refers to the computational resources, e.g., memory, execution time, needed to train and deploy deep neural networks. A compute-efficient model requires fewer computational resources to make predictions.
    \item \textbf{Data efficiency:} Data efficiency pertains to the amount of data required for a network to generalize effectively. Models that achieve high performance with less data are more data-efficient.
    \item \textbf{Parameter efficiency:} Parameter efficiency refers to the size --in terms of the number of parameters used-- of the neural network needed to reach a certain accuracy level. Models that require fewer parameters to reach a comparable validation metric are more parameter-efficient.%A parameter-efficient model requires less parameters to reach a given level in the validation metric.
    \item \textbf{Design efficiency:} Design efficiency refers to the resources in terms of compute, human hours, memory, experimentation, etc., needed to \textit{design} a high-performing architecture for new datasets, tasks or modalities. Architectures that require fewer resources to be adapted to new settings are more design-efficient.% A design-efficient architecture requires fewer resources to adapt to new tasks, datasets, or modalities.
    \item \textbf{Environmental efficiency:} Environmental efficiency relates to the ecological footprint of designing, training and deploying neural networks. Environmentally-efficient architectures incur in lower environmental costs for their design, training and deployment. As Deep Learning models continue to grow in size and complexity, and become increasingly present in everyday tools, the importance of environmental efficiency becomes increasingly critical.%, especially in the context of global climate change.
    \item \textbf{Financial efficiency:} Financial efficiency refers to the cost-effectiveness of designing, training, deploying and maintaining neural networks. Factors considered here include hardware investment, operational costs, and maintenance. Architectures that required lower lower overall inversions are more financially efficient. % in their lifecycle.% less inversion on its life cycle.
\end{itemize}

As can be seen from this classification, Deep Learning efficiency is a broad multifaceted concept that comprise several different aspects, which go beyond the usual characterization in terms of parameter and computational efficiency.%Note that, efficiency in Deep Learning is a broad multifaceted concept that entails several aspects beyond compute and parameter efficiency.

\vspace{-7mm}
\subsection{Continuous modeling and symmetry preservation}

This dissertation aims to investigate inductive biases that broadly improve the efficiency of Deep Learning algorithms. We focus on two less explored inductive biases which have significant impact on various efficiency facets and the learning, generalization and applicability of Deep Learning algorithms: \textit{continuous modeling} and \textit{symmetry preservation}. This section introduces and showcases the significance of these concepts.% outline their relevance and importance for Deep Learning methods.

\vspace{-7mm}
\subsubsection{Continuous modeling} 
Continuous modeling refers to the idea of modeling neural operations directly in a continuous space. This fundamental inductive bias offers significant advantages:

\begin{figure}
    \centering
    \includegraphics[width=0.75 \textwidth]{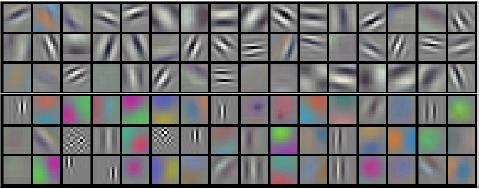}
    \caption{Kernels learned by AlexNet \citep{krizhevsky2012imagenet}. Despite their discrete parameterization, the convolutional kernels learn discrete approximations of continuous functions. Representing the kernels in a continuous domain facilitates the learning of such kernels.}
    \label{fig:0_alexnet_kernels}
\end{figure}

\begin{itemize}
    \item \textbf{Natural data representation.} Many real-world phenomena are continuous in nature. For example, images, audio signals, sensor readings, and time series are all inherently continuous. Nevertheless, well-established neural architectures such as CNNs approach the modeling of these signals in an inherently discrete manner. A key example lies in the representation of convolutional kernels within CNNs, typically structured as sets of discrete independent weights \citep{krizhevsky2012imagenet, he2016deep, huang2017densely}. For example, a $3{\times}3$ convolutional kernel is conventionally represented by a collection of $9$ independent weights. However, looking at the convolutional kernels that these networks learn, one finds that these kernels in fact represent continuous functions, such as color-gradient or edge detectors (Fig.~\ref{fig:0_alexnet_kernels}). This inherent continuity in the learned kernels suggests an opportunity for a more natural data representation by modeling these kernels as continuous functions.
    
    By harnessing continuous modeling, we can describe learnable mappings \textit{directly} in a continuous space. This allows for the construction of neural architectures that inherently align with the nature of the data, and might be able to capture more nuanced patterns and relationships in the data that may be overlooked or inadequately represented by discrete approximations.

    \item \textbf{Efficient long context.} A crucial challenge in machine learning is modeling long-term input dependencies, especially when dealing with large inputs \citep{van2016wavenet, van2016conditional}. This challenge is particularly hard for neural architectures that parameterize their learnable mappings in a discrete manner, e.g., CNNs, as the parameters needed to consider a certain context window grows linearly with its size. As a result, considering large contexts is impractical for these methods and structural modifications are needed to synthetically increase their context window \citep{yu2015multi}.
    
    A powerful alternative to address this challenge comes from the use of self-attention \citep{vaswani2017attention}: a versatile operation able to capture dependencies across long contexts with a fixed number of parameters, and the basis of the popular Transformer architecture. Unfortunately, self-attention induces a quadratic cost both in terms of memory and time relative to the size of the context window, which makes it very expensive for long context applications.
    
    Modeling neural operations directly in a continuous space makes it possible to consider long contexts under a fixed parameter budget and with better scaling properties than self-attention ($\mathrm{N\ logN}$ vs. $\mathrm{N^2}$) (Ch.~\ref{chapter:ckconv},~\ref{chapter:ccnn}). Extensions of such parameterizations have been shown to perform on par with Transformers in language modeling \citep{poli2023hyena} with superior scaling properties that enable them to scale to contexts involving millions of tokens \citep{nguyen2023hyenadna}.% A comprehensive discussion of these advancements is provided in Ch.~\ref{chapter:conclusion}.

    \item \textbf{Resolution-agnostic architectures.} A recurrent problem in Deep Learning is the need to redesign neural architectures to accommodate inputs of different resolution \citep{he2015spatial, lecun2015deep}. Common neural architectures are tied to a particular resolution, e.g., \citet{krizhevsky2012imagenet, he2016deep}. As a consequence, using a neural architecture designed for the processing of $32{\times}32$ images to process images of other size, e.g., $256{\times}256$, would lead to poor results. This effect is observed even if both images are re-scaled replicas of the exact same image. As discussed in Ch.~\ref{chapter:ccnn}, this problem results from modeling neural operations in a discrete manner. Doing so ties them to a particular resolution, making their output drastically different to inputs of different resolutions, e.g., for a $3{\times}3$ convolution.

    Continuous modeling allows for the scaling of feature maps and receptive fields in a way that maintains consistency across different resolutions. This capability enables the creation of neural networks that generalize across resolutions, even if they have not been seen during training. This eliminates the need for separate models for each resolution \citep{ronneberger2015u, he2016deep} or extensive data augmentation techniques to simulate different resolutions \citep{girshick2014rich, xie2017aggregated}, thereby simplifying the training process and reducing computational and design costs. Consequently, this approach paves the way for more versatile, efficient, and effective models better suited for the diverse nature of real-world data.%  This adaptability is obtained by adjusting the sampling resolution of continuous neural operations to align with the input's current resolution.
        
    \item \textbf{Natural handling of irregularly sampled data.} Many real-world scenarios involve data that is not collected uniformly at regular intervals. Prominent examples are (\textit{i}) healthcare time-series data \citep{reyna2019early, johnson2016mimic}, where metrics such as heart rate or blood pressure can be recorded at varying times, and (\textit{ii}) sparsely scanned volumetric data \citep{curless1996volumetric, wu20153d}, which generate point clouds that do not conform to a uniform grid.\break Due to their discrete nature, traditional neural architectures, e.g., (discrete) CNNs, are unable to handle irregular data. This stems from the fact that they parameterize kernels in a regular grid, and cannot be sampled at arbitrary positions \citep{simonovsky2017dynamic}. 
    
    Through the use of continuous modeling, it is possible to define neural components that operate in a continuous space \citep{schutt2017schnet, wu2019pointconv}. Since the resulting models are defined in a continuous space, they can be easily queried at arbitrary positions. This property gives them the ability to manage irregular data in a native manner, thereby broadening the scope of applications where Deep Learning can be used.%to applications where irregular data is prevalent.
    
    \item \textbf{Neural Architecture Search on a continuous space.} Traditionally, neural architectures are regarded as discrete entities, characterized by a discrete number of layers, channels, kernel sizes, patches, etc. \citep{lecun1989backpropagation, kipf2016semi, vaswani2017attention}. This perspective makes Neural Architecture Search (NAS) --the automated process of identifying high-performing neural architectures-- very computationally expensive and time-consuming due to the combinatorial nature of the optimization problem involved \citep{zoph17, pham2018efficient}.

    Through the use of continuous modeling, it is possible to frame neural architectures as entities in a continuous space (Ch.~\ref{chapter:flexconv},~\ref{chapter:dnarch}). This is turn can be used to dramatically reduce the complexity of designing neural architectures. This continuous perspective enables the use of gradient-optimization methods for Neural Architecture Search \citep{liu2018darts, liang2019darts+}, which remarkably increases the speed of the search process, and enables the exploration of a much broader set of candidate architectures.% As a result, the search explores many more combination while being orders of magnitude faster compared to traditional combinatorial approaches.
\end{itemize}
In summary, continuous modeling embodies neural networks with the inherent properties of continuous data, providing several advantages over discrete approaches. These benefits collectively enhance both the efficacy and versatility of deep learning methods across a very diverse set of applications.

\vspace{-7mm}
\subsubsection{Symmetry preservation} 
Symmetry preservation refers to the idea of constructing neural operations that consider the inherent symmetries of data. This fundamental inductive bias offers significant advantages for the learning, generalization and inference dynamics of neural networks:
\begin{itemize}
    \item \textbf{Improved learning and generalization.} Leveraging the inherent symmetries in data as prior knowledge for the design of neural architectures that respect them, leads to a substantial reduction in amount of data required to train them \citep{lecun1989backpropagation, cohen2016group}. That is, to neural architectures with enhanced data efficiency. Unlike unconstrained models, which must learn these symmetrical patterns from the data itself, architectures design to respect these symmetries inherently incorporate them. 

    Moreover, models that respect data symmetries inherently generalize better than those without such constraints. This is because symmetry-preserving models can recognize symmetric modifications of learned patterns, even if particular symmetric modifications of these patterns have not been not been seen during training \citep{weiler2019general,sosnovik2020scaleequivariant}. For example, a CNN is able to recognize patterns even if these have been translated to unseen positions through the use of convolutions (Fig.~\ref{fig:0_conv}). As symmetries are preserved by construction irrespective of the input, this approach is more effective than data augmentation, particularly when these symmetries are well understood, e.g., translation, rotation, scaling \citep{wang2022data}.
    
    \item \textbf{Reliable predictions under input symmetries.} A key advantage of symmetry-preserving models is their ability to provide consistent predictions under symmetrical transformations of the input, both during training and inference \citep{cohen2016group, weiler2018learning}. These guarantees carry immense significance, especially for tasks where maintaining prediction stability under symmetric transformations is essential.
    
    A core example is medical imaging. In tasks like identifying cancerous cells in tissue slice images \citep{veeling2018rotation}, it is evident that geometric transformations such as rotations, translations, and scaling should \textit{not} alter the predictions of the network. However, networks that do not consider symmetry preservation can yield unexpected inconsistent and potentially hazardous variations in their prediction, which is a major barrier to the use of neural networks in critical applications \citep{unay2007robustness,hoogendoorn2022waarom}.
   
    By ensuring prediction consistency under symmetrical conditions, symmetry preserving models bolster their utility and reliability in real-world scenarios, addressing a primary limitations of unconstrained neural networks.

    \item \textbf{Powerful neural networks with fewer parameters.} 
    A closer look at the parameterization of symmetry-preserving networks reveals that symmetry preservation is effectively encoded in the weights of the neural network by means of weight sharing \citep{ravanbakhsh2017equivariance, diaconu2019learning}. Symmetry preservation implicitly ties together several weights that would otherwise be independently parameterized in unconstrained networks.
    
    A great example of this can be seen in CNNs, when compared to fully-connected networks (Fig.~\ref{fig:0_conv_vs_fully_connected}). In CNNs, weights are tied to be identical across all spatial positions, unlike in fully connected networks where weights at each position can have completely different values. Symmetry preservation effectively leads to networks with fewer parameters, thereby making these models more parameter efficient.
\end{itemize}
In summary, symmetry preservation embodies neural networks with a better understanding of the data being processed. This ability grants them improved data and parameter efficiency, as well as enhanced explainability, robustness and generalization. These advancements render symmetry preserving networks highly valuable for a wide range of practical and critical applications.% making them valuable in various practical applications.

\vspace{-7mm}
\section{Research Objective}
The primary objective of this dissertation is the study of continuous modeling and symmetry preservation to improve the efficiency of Deep Learning techniques. Based on this objective, we define the following two research questions:
\begin{itemize}
    \item \textbf{RQ 1.} Can continuous modeling be used to improve Deep Learning efficiency? If so, which specific efficiency aspects as per Sec.~\ref{sec:0_efficiency_aspects} does it improve?
    \item \textbf{RQ 2.} Can symmetry preservation be used to improve Deep Learning efficiency? If so, which specific efficiency aspects as per Sec.~\ref{sec:0_efficiency_aspects} does it improve?
\end{itemize}
The body of this dissertation is divided in two parts devoted to the study of each of the inductive biases under consideration. The following section provides detailed insights into the dissertation structure and the contributions of each chapter. The efficiency contributions of the techniques proposed in each chapter are summarized in Tab.~\ref{tab:0_efficiency_contribution}.

\vspace{-7mm}
\section{Dissertation Structure and Contributions}
This dissertation is structured in two main parts, devoted to each of the inductive biases under study. In accordance with the guidelines set forth by the Doctorate Regulations of the Vrije Universiteit Amsterdam, this section lists the papers on which each chapter is based, along with a brief summary and an account of my contributions to each paper. Table~\ref{tab:0_efficiency_contribution} summarizes the efficiency contributions of each chapter.%  is crucial to keep in mind that these aspects are equally important and improvements along all efficiency axes indirectly contribute to environmental and financial efficiency.

\begin{table}[h]
\captionsetup{justification=centering}
\caption{Efficiency contributions as per Sec.~\protect\ref{sec:0_efficiency_aspects} of each chapter.\protect\footnote{Note that Tab.~\ref{tab:0_efficiency_contribution} does not depict the environmental and financial efficiency aspects defined in Sec.~\ref{sec:0_efficiency_aspects}. From a technical standpoint, enhancements in environmental and financial efficiently are implicitly driven by advancements across the remaining efficiency axes. Consequently, we do not explicitly add them here. However, we consider important to acknowledge the importance of these efficiency aspects, even more as Deep Learning continuous to permeate everyday tools and applications.}}
\vspace{-2.5mm}
    \centering
    \begin{tabular}{cccccc}
    \toprule
         & & Compute  & Data & Parameter & Design  \\
         & & Efficiency & Efficiency & Efficiency & Efficiency \\
         \midrule
       \multirow{5}{*}{\rotatebox{90}{\thead{\sc{Part I:} \\ \sc{Continuous} \\ \sc{Modelling}}}} &  Chapter~\ref{chapter:ckconv} & {\color{Green}\cmark }  & - & {\color{Green}\cmark }  & {\color{Green}\cmark }  \\ 
          & Chapter~\ref{chapter:ccnn} & {\color{Green}\cmark }  & - & {\color{Green}\cmark }  & {\color{Green}\cmark }  \\ 
          & Chapter~\ref{chapter:gridification} & {\color{Green}\cmark }  & - & {\color{Green}\cmark }  & - \\ 
          & Chapter~\ref{chapter:flexconv} & {\color{Green}\cmark } & - & - & {\color{Green}\cmark }  \\ 
          &  Chapter~\ref{chapter:dnarch} & {\color{Green}\cmark }  & - & {\color{Green}\cmark }  & {\color{Green}\cmark }  \\ 
         \midrule
         \multirow{5}{*}{\rotatebox{90}{\thead{\sc{Part II:} \\ \sc{Symmetry} \\ \sc{Preservation}}}} & Chapter~\ref{chapter:coattentive} & - & {\color{Green}\cmark }  & - & - \\ 
         & Chapter~\ref{chapter:attgconv}  & - & {\color{Green}\cmark }  & - & - \\ 
          & Chapter~\ref{chapter:g_selfatt}  & - & {\color{Green}\cmark }  & {\color{Green}\cmark }  & -  \\ 
           & Chapter~\ref{chapter:waveletnets} & - & {\color{Green}\cmark }  & {\color{Green}\cmark } & - \\ 
            & Chapter~\ref{chapter:partial_equiv} & {\color{Green}\cmark }  & {\color{Green}\cmark }  & {\color{Green}\cmark }  & {\color{Green}\cmark }  \\ 
        \bottomrule
    \end{tabular}
    \label{tab:0_efficiency_contribution}
\end{table}
%Note that Tab.~\ref{tab:0_efficiency_contribution} does not depict the environmental and financial efficiency aspects of Deep Learning outlined in Sec.~\ref{sec:0_efficiency_aspects}. From a technical standpoint, enhancements in environmental and financial efficiency are implicitly driven by advancements across the other efficiency dimensions. Consequently, we refrain from adding them explicitly in Tab.~\ref{tab:0_efficiency_contribution}. However, we consider \textit{crucial} to acknowledge the importance of these efficiency aspects, as these become increasingly important with Deep Learning continuously permeating everyday tools and applications.

\textbf{\sc{Part I: Efficiency Through Continuous Modeling}}
\begin{description}
    \item[Chapter~\ref{chapter:ckconv}.] \textbf{Continuous Kernel Convolutions for Sequential Data.} This chapter is based on the paper:

    David W. Romero, Anna Kuzina, Erik J Bekkers, Jakub M Tomczak and Mark Hoogendoorn. CKConv: Continuous kernel convolution for sequential data. In \textit{International Conference on Learning Representations}, 2022.

    \textit{Summary:} This chapter introduces a continuous parameterization of convolutional kernels through the use of MLPs that map coordinates to the value of convolutional kernels at those coordinates. This approach facilitates: (\textit{i}) the creation of large, parameter-efficient convolutional kernels able to capture global context at a much lower cost than Transformers; (\textit{ii}) the construction of resolution agnostic neural architectures; and (\textit{iii}) an effective way to process irregularly-sampled data. %{c}_i$ to the value of the kernel at that  convolutional kernels, with which it is possible to create global convolutional kernels in a parameter efficient manner. By doing so, we infuse convolutional neural networks with the ability to model global context, akin to Transformers, but at a much lower cost.
    % \textbf{Efficiency contribution:} Parameter Efficiency, Computational Efficiency, Design Efficiency, Environmental Efficiency.

    \textit{Personal contributions:} Original idea to use MLPs to model convolutional kernels, theory, implementation, ablations, experiments and writing. 

    \item[Chapter~\ref{chapter:ccnn}.] \textbf{Modeling Long Context in $N$D: From Task-Specific to a General-Purpose CNN.} This chapter is based on the papers:

    David W. Romero$^*$, David M. Knigge$^*$, Albert Gu, Erik J. Bekkers, Efstratios Gavves, Jakub M. Tomczak and Mark Hoogendoorn. Towards a General Purpose CNN for Long Range Dependencies in $N$D. In \textit{ICML Workshop on Continuous Time Methods for Machine Learning}, 2022.

    David M. Knigge$^*$, David W. Romero$^*$, Albert Gu, Efstratios Gavves, Erik J. Bekkers, Jakub M. Tomczak, Mark Hoogendoorn and Jan-Jakob Sonke. modeling long range dependencies in $N$D: From task-specific to a general purpose CNN. In \textit{International Conference on Learning Representations}, 2023.

    \textit{Summary:} This chapter presents a convolutional framework able to handle data of varying dimensionality, lengths and resolutions without architectural changes. Its core insight is that this can be achieved through the use of layers able to model global context regardless of these factors in a resolution agnostic way. The resulting architecture shows high versatility, delivering good results across diverse modalities and datasets without requiring customization for each use-case.

    \textit{Personal contribution:} This work was carried out in equal contribution with David M. Knigge. We co-developed the ideas and insights presented in these papers. We both contributed to theory, implementation, ablations, experiments and writing.

    \item[Chapter~\ref{chapter:gridification}.] \textbf{Learned Gridification for Efficient Point-Cloud Processing.} This chapter is based on the paper:

    Putri A. van der Linden$^*$, David W. Romero$^*$ and Erik J. Bekkers. Learned Gridification for Efficient Point Cloud Processing. In \textit{Proceedings of the 2$^{nd}$ Annual Workshop on Topology, Algebra, and Geometry in Machine Learning (ICML)}, 2023.

    \textit{Summary:} This chapter introduces a method to reduce the compute and memory complexity of point-cloud processing significantly. It involves converting point-clouds into regular grid representations through a learnable \enquote{gridification} step, which is then efficiency processed with grid-based operations, e.g., \texttt{Conv3D}. After processing the grid representation, it can be converted back to point-cloud form for tasks such as segmentation. Gridified networks match and even surpass state of the art methods while being remarkably faster and more scalable.

    \textit{Personal contribution:} This work was carried out in equal contribution with Putri A. van der Linden. We co-developed the idea and formulation of learnable gridification, contributed to theory, implementation, ablations, experiments and writing.

    \item [Chapter~\ref{chapter:flexconv}.] \textbf{Continuous Kernel Convolutions with Differentiable Kernel Sizes.} This chapter is based on the paper: 

    David W. Romero$^*$, Robert-Jan Bruintjes$^*$, Jakub M. Tomczak, Erik J. Bekkers, Mark Hoogendoorn and Jan C. van Gemert. Flexconv: Continuous kernel convolutions with differentiable kernel sizes. In \textit{International Conference on Learning Representations}, 2022.

    \textit{Summary:} This chapter improves the parameterization of Continuous Kernel Convolutions to make their kernel sizes learnable by backpropagation. This allows CNNs to autonomously find the optimal kernel sizes that leads to the best accuracy, therefore releasing users from the need to pre-specify kernel sizes, and the need to rely on simple network configurations, e.g. with $3{\times}3$ kernels at all layers.

    \textit{Personal contributions:} This work was carried out in equal contribution with Robert-Jan Bruintjes. We co-developed the idea of parameterizing convolutional kernels as the combination of an MLP with a learnable Gaussian, and of regularizing against aliasing through the use of Multiplicative Filter Networks \citep{fathony2021multiplicative}. We both contributed to theory, implementation, ablations, experiments and writing.

    \item[Chapter~\ref{chapter:dnarch}.] \textbf{Learning Convolutional Neural Architectures by Backpropagation.} This chapter is based on the paper:

    David W. Romero and Neil Zeghidour. DNArch: Learning convolutional neural architectures by backpropagation. In \textit{Differentiable Almost Everything: Differentiable Relaxations, Algorithms, Operators, and Simulators Workshop (ICML)}, 2023.\footnote{This paper was carried out during a Research Internship at Google Research.}

    \textit{Summary:} This chapter introduces an scalable method able to learn several architectural components of CNNs by propagation: (\textit{i}) the size of all convolutional kernels, (\textit{ii}) the width of all layers in the network, (\textit{iii}) the position and value of all downsampling layers, and (\textit{iv}) the depth of the network. This is achieved in an inexpensive way by viewing neural architectures as multidimensional continuous entities and using learnable differentiable masks to learn their sizes. Using a general-purpose CNN as backbone, e.g., a CCNN (Sec.~\ref{chapter:ccnn}), our method automatically finds high-performant architectures across several tasks and modalities. 

    \textit{Personal contribution:} Original idea to use differentiable masks to learn the whole architecture of CNNs, theory, implementation, ablations, experiments and writing.
\end{description}
\textbf{\sc{Part II: Efficiency Through Symmetry Preservation}}
\begin{description}
    \item[Chapter~\ref{chapter:coattentive}.] \textbf{Focusing Equivariance on Transformations Co-Occurring in Data.} This chapter is based on the paper:

    David W. Romero and Mark Hoogendoorn. Co-attentive equivariant neural networks: Focusing equivariance on transformations co-occurring in data. In \textit{International Conference on Learning Representations}, 2020.

    \textit{Summary:} This chapter proposes the use of a lightweight block cyclic permutation equivariant self-attention module to explicitly accentuate the responses of group convolutions whose relative geometrical poses align with geometric patterns seen in data. Due to this explicit system for the accentuation of co-occurring symmetries, resulting models that capture geometrical dependencies better, therefore enhancing the generalization of group equivariant models.%  To this end, a ciclically equivariant self-attention op operations on top of group convolutions to explicitly co the idea of incorporating self-attention operations on the orientation responses of group equivariant CNNs to  explicitly 

    \textit{Personal contribution:} Original idea to use block cyclic equivariant self-attention to accentuate geometric patterns co-occurring in data, formulation, theory, implementation, ablations, experiments and writing.

    \item[Chapter~\ref{chapter:attgconv}.] \textbf{Attentive Group Equivariant Convolutional Neural Networks.} This chapter is based on the paper:

    David W. Romero, Erik J. Bekkers, Jakub M. Tomczak and Mark Hoogendoorn. Attentive group equivariant convolutional networks. In \textit{International Conference on Machine Learning}, pp. 8188-8199. PMLR, 2020.

    \textit{Summary:} This chapter proposes the use of a generalized form of visual attention that is symmetry preserving. This mechanism allows for the accentuation of complex geometrical templates as well as the generation of attention maps that behave equivariantly, leading to improved generalization and interpretability.

    \textit{Personal contribution:} Original idea of symmetry-preserving visual attention, formulation, theory, implementation, ablations, experiments and writing. 

    \item[Chapter~\ref{chapter:g_selfatt}.] \textbf{Group Equivariant Stand-Alone Self-Attention.} This chapter is based on the paper:

    David W. Romero and Jean-Baptiste Cordonnier. Group equivariant stand-alone self-attention for vision. In \textit{International Conference on Learning Representations}, 2021.

    \textit{Summary:} Based on the success of Transformers, we provide a symmetry-preserving formulation of the self-attention operation. Since self-attention relies on positional encodings to handle geometrical information, our formulation is steerable, and thus can operate with groups whose action lives outside the data grid. Thanks to symmetry preservation, the resulting group equivariant Transformers exhibit improved data-efficiency and generalization than non-equivariant counterparts.% able to consider larger groups than discrete alternatives such as \citet{cohen2016group}.

    \textit{Personal contribution:} Original idea of symmetry-preserving self-attention, formulation, theory, implementation, ablations, experiments and writing. 

    \item[Chapter~\ref{chapter:waveletnets}.] \textbf{Scale-Translation Equivariant Learning From Raw Time-Series.} This chapter is based on the paper:

    David W. Romero, Erik J. Bekkers, Jakub M. Tomczak and Mark Hoogendoorn. Wavelet networks: Scale-translation equivariant learning from raw time-series. In \textit{Transactions
on Machine Learning Research}, 2023.

    \textit{Summary:} This chapter provides a translation scale equivariant formulation of convolution operations for the processing of raw time-series. The resulting mappings share an interesting relationship with the well-known Wavelet transform, providing interesting insights about the modus operandi of the resulting architectures. The preservation of translation and scale symmetries leads to improved data efficiency and generalization over existing CNNs that work on raw data.%, and that match the accuracy of highly engineered spectrogram-based approaches.
    
    \textit{Personal contribution:} Original idea of using scale translation equivariance for the processing of time-series, its connection to the Wavelet transform, formulation, theory, implementation, ablations, experiments and writing.

    \item[Chapter~\ref{chapter:partial_equiv}.] \textbf{Learning Equivariances and Partial Equivariances from Data.} This chapter is based on the paper:

    David W. Romero and Suhas Lohit. Learning equivariances and partial equivariances from data. In \textit{Advances in Neural Information Processing Systems 35}, pp. 36466-36478, 2022.\footnote{This paper was carried out during a research internship at Mitsubishi Electric Research Laboratories.}

    \textit{Summary:} While symmetry preservation is highly effective when the relevant symmetries exist in data, it can become too constraining if these symmetries are misspecified or only partially present. This chapter presents a framework that allows for the automatic adjustment of symmetry preservation during training to match to the actual symmetries present in data. The resulting models benefit from the data efficiency offered by symmetry preservation, while maintaining the flexibility to adjust when it becomes overly restrictive. This leads to more adaptable, data-efficient models that effectively self-manage the level in which this inductive bias affects the learning process.

    \textit{Personal contribution:} Original idea and formulation of partially equivariant models, theory, implementation, ablations, experiments and writing.
\end{description}

\textbf{\sc{Other Contributions}}

In addition to the papers explicitly considered in this dissertation, I was also involved in a number of projects for which I served in an advisory role. The papers associated with these projects are:
\begin{itemize}
    \item David M. Knigge, \textbf{David W. Romero} and Erik J. Bekkers. Exploiting redundancy: Separable group convolutional networks on lie groups. In \textit{International Conference on Machine Learning} pp. 11359-11386. PMLR, 2022.
    \item Tycho van der Ouderaa, \textbf{David W. Romero} and Mark van der Wilk. Relaxing equivariance constraints with non-stationary continuous filters. In \textit{Advances in Neural Information Processing Systems 35}, pp. 33818-33830, 2022.
    \item Stefano Massaroli, Michael Poli, Daniel Y. Fu, Hermann Kumbong, \textbf{David W. Romero}, Rom N. Parnichkun, Arman Timalsina, Quinn McIntyre, Beidi Chen, Atri Rudra, Ce Zhang, Christopher R\'{e}, Stefano Ermon and Yosua Bengio. Laughing hyena distillery: Extracting compact recurrences from convolutions. In \textit{Advances in Neural Information Processing Systems}, 2023.
    \item Erik J. Bekkers, Sharvaree Vadgama, Rob Hesselink, Putri A. van der Linden and \textbf{David W. Romero}. Fast, Expressive $\mathrm{SE}(n)$ Equivariant Networks trough Weight Sharing in Position-Orientation Space. In \textit{International Conference on Learning Representations}, 2024.
    %\item Riccardo Valperga, Samuele Papa, \textbf{David W. Romero} and Efstratios Gavves. Contrastive Implicit Representation Learning. \textit{Under review}, 2023.
    \item Alonso Urbano and \textbf{David W. Romero}. Self-Supervised Detection of Perfect and Partial Input-Dependent Symmetries. In \textit{ICML Workshop on Geometry-grounded Representation Learning and Generative Modeling}, 2024.
\end{itemize}
We briefly discuss these papers in Chapter~\ref{sec:12_limitations}, making explicit their connection to efficiency aspects of deep learning and the remaining limitations of this work.

\vspace{-7mm}
\subsection{How to read this dissertation}
Each chapter in this dissertation is based on one or more papers, which are either published or currently under review. To maintain the self-contained nature of each chapter, we have chosen to keep them as consistent as possible with the original publications, only making modifications to align with the formatting of this dissertation. This approach enables each chapter to function as an independent unit, at the cost of needing possible repetitions of key concepts across some chapters. For readers who plan to go through the dissertation sequentially, we recommend skimming past concepts previously introduced in the Preliminaries and Background sections of earlier chapters. 

Each chapter is accompanied by a corresponding Appendix situated after the References section. These appendixes serve to complement the main text by including, among other things, complementary theoretical definitions, proofs and detailed descriptions of the datasets used in the experimental section of each chapter. Readers are advised to skim past the concepts and datasets introduced in the Appendixes of previous chapters.

% \dataefficiencytag \\
% \computeefficiencytag \\
% \parameterefficiencytag \\
% \designefficiencytag \\
% \environmentalefficiencytag

%   \nobibliography{bibliography}
%   \bibliographystyle{plainnat}

%   \section*{Publications}
%   \begin{itemize}
%     \item \bibentry{romero2022ckconv}
%     \item \bibentry{romero2022flexconv}
%     \item \bibentry{knigge2023ccnn}
%     \item \bibentry{romero2023dnarch}
%     \item \bibentry{romero2020Co-Attentive}
%     \item \bibentry{romero2020attentive}
%     \item \bibentry{romero2020waveletNS}
%     \item \bibentry{romero2020group}
%     \item \bibentry{romero2022learning}
%   \end{itemize}

%=========================================================================

%% file: chapters/2_ckconv.tex
% path to figures directory
\graphicspath{{figures/2-ckconv/}}

%=========================================================================

% \begin{savequote}[75mm]
% Nulla facilisi. In vel sem. Morbi id urna in diam dignissim feugiat. Proin molestie tortor eu velit. Aliquam erat volutpat. Nullam ultrices, diam tempus vulputate egestas, eros pede varius leo.
% \qauthor{Quoteauthor Lastname}
% \end{savequote}
\chapter{Continuous Kernel Convolutions for Sequential Data}\label{chapter:ckconv}

\begin{flushright}
\textit{Based on the paper:}\break
\textit{CKConv: Continuous Kernel Convolution for Sequential Data \citep{romero2022ckconv}}
\end{flushright}
%=========================================================================
\vspace{-7mm}
\section{Introduction}

Recurrent Neural Networks (RNNs) have long governed tasks with sequential data \citep{rumelhart1985learning, hochreiter1997long, chung2014empirical}. Their main ingredient are \textit{recurrent units}: network components with a recurrence formulation which grants RNNs the ability to be unrolled for arbitrarily many steps and handle sequences of arbitrary size. In practice, however, the effective \textit{memory horizon} of RNNs, i.e., the number of steps the network retains information from, has proven to be surprisingly small, most notably due to the \textit{vanishing gradients problem} \citep{hochreiter1991untersuchungen, bengio1994learning}. Interestingly, it is the very recurrent nature of RNNs that allows them to be unrolled for arbitrarily many steps which is responsible for vanishing gradients \citep{pascanu2013difficulty}. This, in turn, hinders learning from the far past and induces a~small~effective~memory~horizon.

Convolutional Neural Networks (CNNs) \citep{lecun1998gradient} have proven a strong alternative to recurrent architectures as long as relevant input dependencies fall within their memory horizon \cite{conneau2016very, oord2016wavenet, dai2017very, dauphin2017language, bai2018empirical}. CNNs avoid the training instability and vanishing / exploding gradients characteristic of RNNs by avoiding \textit{back-propagation through time} \citep{werbos1990backpropagation} altogether.
However, these architectures model convolutional kernels as a sequence of independent weights. As a result, their memory horizon must be defined \textit{a-priori}, and larger memory horizons induce a proportional growth of the model size.

In this work, we provide a solution to these limitations. We propose to view a convolutional kernel as a continuous function parameterized by a small neural network instead of a sequence of independent weights. The resulting \textit{Continuous Kernel Convolution} (CKConv) %\footnote{Code available at \href{https://github.com/dwromero/ckconv}{\texttt{github.com/dwromero/ckconv}}}
enjoys the following properties:
\begin{itemize}
    \item CKConvs can define arbitrarily large memory horizons within a single operation. Consequently, \textit{Continuous Kernel Convolutional Neural Networks} (CKCNNs) detach their memory horizon from \emph{(i)} the depth of the network, \emph{(ii)} the dilation factors used, and \emph{(iii)} the size of the network. 
    \item CKConvs do not rely on any form of recurrence. As a result, CKCNNs (\textit{i}) can be trained in parallel, and (\textit{ii}) do not suffer from vanishing / exploding gradients or small effective memory horizons.
    \item Continuous convolutional kernels can be evaluated at arbitrary positions. Consequently, CKConvs and CKCNNs can be readily used on irregularly sampled data, and data at different resolutions.
\end{itemize}
We observe that continuous kernel parameterizations previously used to handle irregular data \textit{locally}, e.g., \citet{schutt2017schnet, wu2019pointconv}, are not adequate to model long-term dependencies. This is due to the inability of their kernels to model long spatial complex functions (Sec.~\ref{sec:2_ckconvkernel}). Contrarily, CKConvs perfectly describe long complex non-linear, non-smooth functions by parameterizing their kernels as SIRENs \citep{sitzmann2020implicit}: implicit neural representations with $\mathrm{Sine}$ nonlinearities. Shallow CKCNNs match or outperform state-of-the-art approaches on several tasks comprising stress tests, continuous, discrete and irregular data, as well as resolution changes. To the best of our knowledge, we are first to observe the potential of continuous convolutional kernels to model long-term dependencies, and to provide an useful parameterization to this end.
\begin{figure}
    \centering
    \hfill
         \begin{subfigure}[b]{0.115\textwidth}
         \centering
         \includegraphics[width=\textwidth]{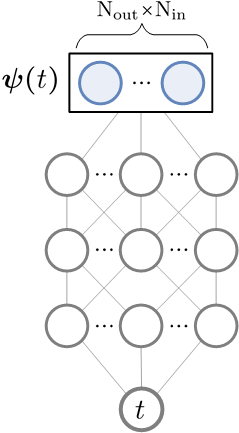}
         %\caption{}
         %\label{fig:2_centered_conv}
     \end{subfigure}
     \hfill
     \begin{subfigure}[b]{0.75\textwidth}
         \centering
         \includegraphics[width=\textwidth]{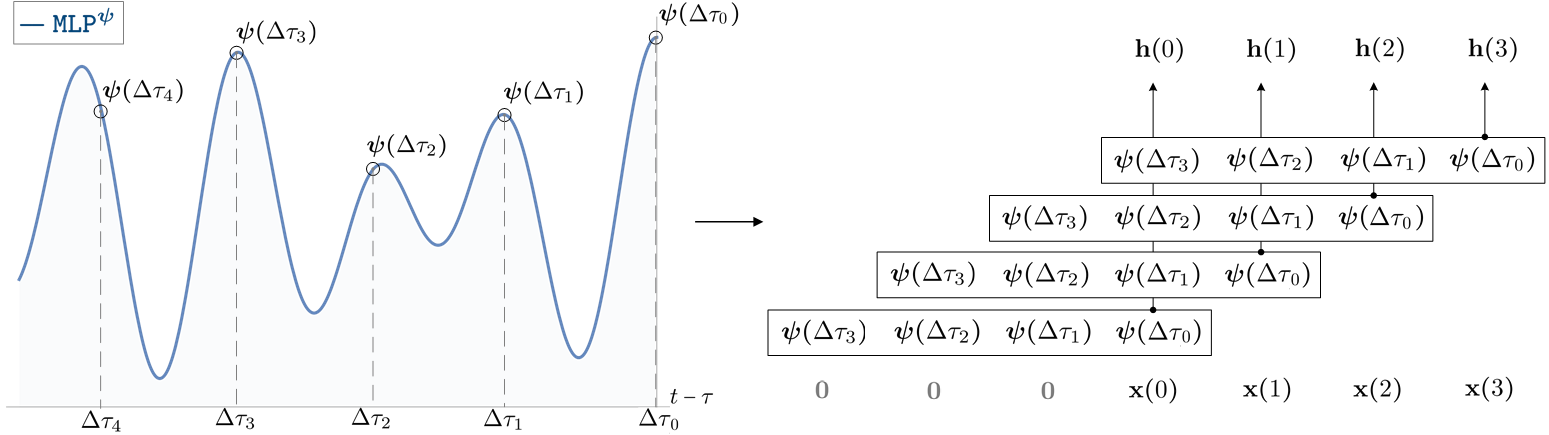}
         %\caption{}
         %\label{fig:2_causal_conv}
     \end{subfigure}
     \hfill
    \caption{Continuous Kernel Convolution (CKConv). CKConv views a convolutional kernel as a vector-valued continuous function $\boldsymbol{\psi}: \sR \rightarrow \sR^{\mathrm{N_{out}}\times \mathrm{N_{in}}}$ parameterized by a small neural network \mlppsi. \mlppsi\ receives a time-step and outputs the value of the convolutional kernel at that position. We sample convolutional kernels by passing a set of relative positions $\{ \Delta\tau_i\}$ to \mlppsi, and perform convolution with the sampled kernel next. Since \mlppsi\ is a continuous function, CKConvs can (\textit{i}) construct arbitrarily large kernels% by passing an equivalently large vector of positions
    , (\textit{ii}) generate kernels at different resolutions% by sampling grid of positions at the same resolutions
    , and (\textit{iii}) handle irregular data.% by providing input positions only at know places.
    }
    \label{fig:2_ckconv}
\end{figure}
% -----------SECTION----------------------------
\vspace{-7mm}
\section{Related Work}\label{sec:2_related_work}

\textbf{Continuous kernel formulation.} Continuous formulations for convolutional kernels were introduced to handle irregularly sampled 3D data \textit{locally} \citep{schutt2017schnet, simonovsky2017dynamic, wang2018deep, wu2019pointconv}. As discrete convolutions learn independent weights for specific relative positions, they cannot handle irregularly sampled data effectively. Following work focuses on point-cloud applications \citep{fuchs2020se, hu2020randla, shi2019points, thomas2018tensor}. Other approaches include Monte Carlo approximations of continuous operations \citep{finzi2020generalizing}.
Our work proposes a new broad flavor of applications for which continuous kernels are advantageous.

\textbf{Implicit neural representations.} Implicit neural representations construct continuous data representations by encoding the input in the weights of a neural network \citep{mescheder2019occupancy, park2019deepsdf, sitzmann2020implicit}. This leads to numerous advantages over conventional (discrete) data representations, e.g., memory efficiency, analytic differentiability, with interesting properties for several applications, e.g., generative modelling \citep{dupont2021generative, schwarz2020graf}.

Since we model convolutional kernels as continuous functions and parameterize them via neural networks, our approach can be understood as \textit{implicitly representing the convolutional kernels of a conventional CNN}.
Different is the fact that these convolutional kernels are not known a-priori, but learned as a part of the optimization task of the CNN. Making the connection between implicit neural representations and continuous kernel formulations explicitly brings substantial insights for the construction of these kernels. In particular, it motivates the use of $\mathrm{Sine}$ nonlinearities \citep{sitzmann2020implicit} to parameterize them, which leads to significant improvements over the $\mathrm{ReLU}$, $\mathrm{LeakyReLU}$, and $\mathrm{Swish}$ nonlinearities used so far for this purpose (Sec.~\ref{sec:2_ckconvkernel}).% To our best knowledge, this is the first work to make the connection between continuous kernel formulations and implicit neural representations explicit.% This allows us to provide a novel continuous parameterization with which, for the first time, long-term dependencies can be correctly described.
% -----------SECTION----------------------------
\vspace{-7mm}
\section{The Convolution and Common Kernel Parameterizations}\label{sec:2_convolution}

\textbf{Notation.} $[n]$ denotes the set $\{0, 1, \ldots, n\}$. Bold capital and lowercase letters depict vectors and matrices, e.g., $\xv$, $\Wm$, sub-indices index vectors, e.g., $\xv {=} \{x_{\ct}\}_{\ct = 1}^{\Nt_{\mathrm{in}}}$, parentheses index time, e.g., $\xv(\tau)$ is the value of $\xv$ at time-step $\tau$, and calligraphic letters depict sequences, e.g., $\gX {=} \{\xv(\tau) \}_{\tau {=} 0}^{\Nt_{\Xt}}$. 
% -----------sub----------------------------

\begin{figure}[t]
     \centering
     \begin{subfigure}[b]{0.26\textwidth}
         \centering
         \captionsetup{justification=centering}
         \includegraphics[width=\textwidth]{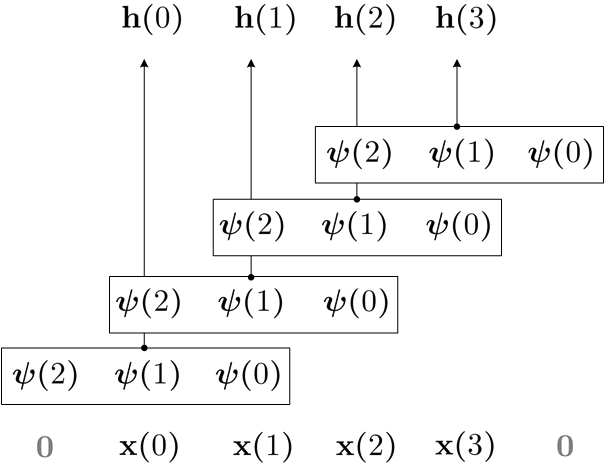}
         \caption{}
         \label{fig:2_centered_conv}
     \end{subfigure}
     \hfill
     \begin{subfigure}[b]{0.26\textwidth}
         \centering
         \captionsetup{justification=centering}
         \includegraphics[width=\textwidth]{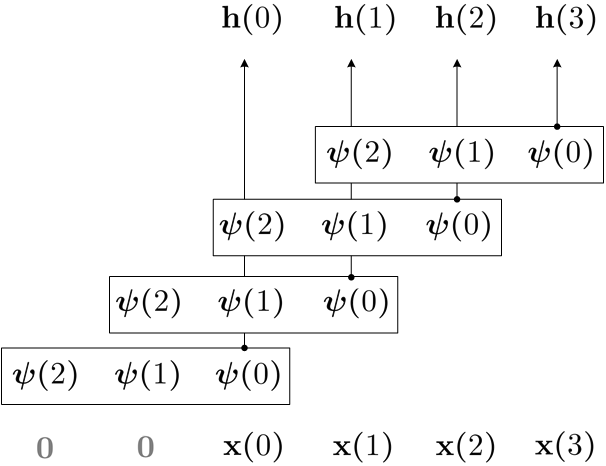}
         \caption{}
         \label{fig:2_causal_conv}
     \end{subfigure}
     \hfill
     \begin{subfigure}[b]{0.35\textwidth}
         \centering
         \captionsetup{justification=centering}
         \includegraphics[width=\textwidth]{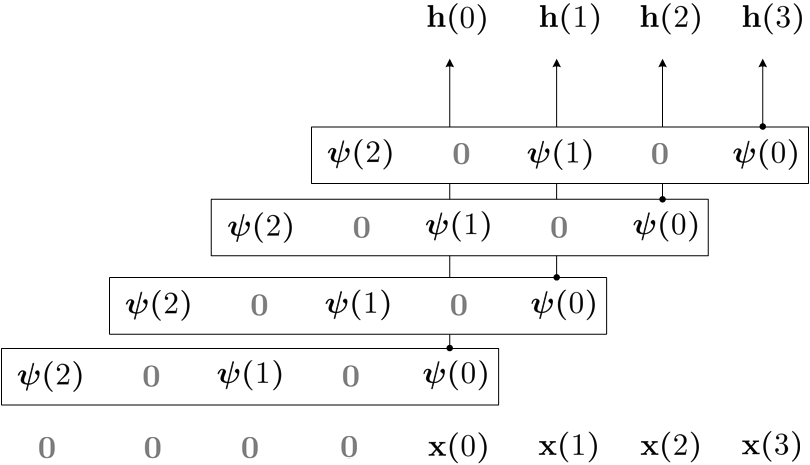}
         \caption{}
         \label{fig:2_dilated_conv}
     \end{subfigure}
     \vspace{-2mm}
     \captionsetup{justification=centering}
     \caption{Discrete centered, causal, and dilated causal convolutions.
     }
        \label{fig:2_causal_vs_centered}
\end{figure}

\textbf{Centered and causal convolutions.} 
Let $\xv: \sR \rightarrow \sR^{\Nt_{\mathrm{in}}}$ and $\boldsymbol{\psi}:\sR \rightarrow \sR^{\Nt_{\mathrm{in}}}$ be a vector valued signal and kernel on $\sR$, such that $\xv{=}\{x_{\ct}\}_{\ct {=} 1}^{\Nt_{\mathrm{in}}}$ and $\boldsymbol{\psi} {=} \{ \psi_{\ct} \}_{c {=} 1}^{\Nt_{\mathrm{in}}}$. The convolution is defined as:
\begin{equation}
    (\xv * \boldsymbol{\psi})(t) = \sum_{\ct = 1}^{\Nt_{\mathrm{in}}} \int_{\sR} x_{\ct}(\tau) \psi_{\ct}(t - \tau) \, \du \tau. \label{eq:2_conv_def1}
\end{equation}
In practice, the input signal $\xv$ is gathered via some sampling procedure. Resultantly, the convolution is effectively performed between the sampled input signal described as a sequence of finite length $\gX {=} \{\xv(\tau) \}_{\tau {=} 0}^{\Nt_{\Xt}}$ and a convolutional kernel $\gK {=} \{\boldsymbol{\psi}(\tau) \}_{\tau {=} 0}^{\Nt_{\Xt}}$ described the same way:
\begin{equation}
    (\xv * \boldsymbol{\psi})(t) = \sum_{\ct = 1}^{\Nt_{\mathrm{in}}} \sum_{\tau = -\nicefrac{\Nt_{\Xt}}{2}}^{\nicefrac{\Nt_{\Xt}}{2}} x_{\ct}(\tau) \psi_{\ct}(t - \tau). \label{eq:2_conv_centered}
\end{equation}
Values $\xv(\tau)$ falling outside of $\gX$ are often \textit{padded} by a constant value of zero (Fig.~\ref{fig:2_centered_conv}).

The convolutional kernel is commonly \textit{centered} around the point of calculation $t$. For sequence modeling this can be undesirable as future input values $\{ \xv(t - \tau)\}_{\tau {=} \nicefrac{-\Nt_{\Xt}}{2}}^{-1}$ are considered during the operation. This is solved by providing a \textit{causal formulation to the convolution}: a formulation in which the convolution at time-step $t$ only depends on input values at time-steps $(t - \tau) \leq t$ (Fig.~\ref{fig:2_causal_conv}):
\begin{equation}
    (\xv * \boldsymbol{\psi})(t) = \sum_{\ct = 1}^{\Nt_{\mathrm{in}}} \sum_{\tau = 0}^{t} x_{\ct}(\tau) \psi_{\ct}(t - \tau). \label{eq:2_causal_conv}
\end{equation}
In practice, causal convolutions are easily implemented via asymmetrical padding. In this work, we consider causal convolutions as default, but our analyses are valid for centered convolutions as well.

% -----------sub----------------------------
\textbf{Discrete convolutional kernels.}
By a large margin, most convolutional kernels $\boldsymbol{\psi}$ in literature are parameterized as a finite sequence of $\Nt_{\Kt} + 1$ independent learnable weights $\gK {=} \{ \boldsymbol{\psi}(\tau) \}_{\tau {=} 0}^{\Nt_{\Kt}}$ (Fig.~\ref{fig:2_causal_vs_centered}). As these weights are independent of one another, $\Nt_{\Kt}$ must be kept small to keep the parameter count of the model tractable. Hence, the kernel size is often much smaller than the input length: $\Nt_{\Kt} \ll \Nt_{\Xt}$. 
%As a result, the convolution operation is reduced to:
% \begin{equation}
% \setlength{\abovedisplayskip}{2pt}
% \setlength{\belowdisplayskip}{3pt}
%     (\xv * \boldsymbol{\psi})(t) = \sum_{\ct = 1}^{\Nt_{\Ct}} \sum_{\tau = 0}^{\Nt_{\Kt}} x_{\ct}(\tau) \psi_{\ct}(t - \tau).
% \end{equation}
This parameterization presents important limitations:
\begin{itemize}%[topsep=0pt, leftmargin=*]
\item The memory horizon $\Nt_{\Kt}$ must be defined a priori.
\item Since $\Nt_{\Kt} {\ll} \Nt_{\Xt}$, this parameterization implicitly assumes the convolution $(\xv * \boldsymbol{\psi})$\break at position $t$ \emph{only depends on input values at positions up to $\tau {=} \Nt_{\Kt}$ steps in the past}. Hence, no functions depending on inputs $\xv(t-\tau)$ for $\tau > \Nt_{\Kt}$ can be modeled.
\item The most general selection of $\Nt_{\Kt}$ is given by a \textit{global memory horizon}: $\Nt_{\Kt} {=} \Nt_{\Xt}$. Unfortunately, as discrete convolutional kernels are modeled as a sequence of independent weights, this incurs an extreme growth of the model size and rapidly becomes statistically unfeasible.
\end{itemize}
\textbf{Dilated convolutional kernels.}
To alleviate these limitations, previous works propose to interleave kernel weights with zeros in order to cover larger memory horizons without additional weights (Fig.~\ref{fig:2_dilated_conv}). %. The \textit{dilated convolution} is given by:
% \begin{equation}
% \setlength{\abovedisplayskip}{-4pt}
% \setlength{\belowdisplayskip}{3pt}
%     (\xv *_{\eta} \boldsymbol{\psi})(t)  = \sum_{\ct = 1}^{\Nt_{\Ct}} \sum_{\tau = 0}^{\Nt_{\Kt}} x_{\ct}(\eta \tau) \psi_{\ct}(t - \eta \tau).\label{eq:2_dilated_conv_2}
% \end{equation}
This formulation alleviates some of the previous limitations, but introduces additional ones:
\begin{itemize}%[topsep=0pt, leftmargin=*]
\item Dilated kernels are unable to model dependencies of input values falling in the interleaved regions.%among input values within $\xv(\eta \tau)$ and $\xv(\eta (\tau+ 1))$.
\item Several authors use dilated convolutions with varying dilation factors as a function of depth, e.g., \citet{bai2018empirical, dai2017very, oord2016wavenet}. By carefully selecting layer-wise dilation factors, one can assure that some kernel hits each input within the memory horizon of the network. However, due to the extreme sparsity of the formulation, it is difficult to estimate the effective amount of processing applied to the input. In addition, this layout ties together (\textit{i}) the memory horizon, (\textit{ii}) the depth, and (\textit{iii}) the layer-wise dilation factors of the network, which effectively constraints the flexibility of the neural architecture design.
\end{itemize}
In contrast to the (dilated) discrete convolutions presented in this section, our proposed formulation allows handling arbitrarily long sequences with arbitrarily large, dense memory horizons in a single layer and under a fixed parameter budget.
% -----------SECTION----------------------------
\vspace{-7mm}
\section{Continuous Kernel Convolution}\label{sec:2_continuous_kernel_conv}

In this section, we introduce our approach. First, we define it formally, analyze its properties, illustrate its connection to recurrent units, and elaborate on the functional family they can describe. Next, we discuss concrete parameterizations of continuous convolutional kernels, illustrate their connection to implicit neural representations, and show that our final kernels are able to fit complex functions.
%
% -----------SECTION----------------------------
\vspace{-7mm}
\subsection{Formulation and Properties}\label{sec:2_overview}
\textbf{Arbitrarily large convolutional kernels.} We formulate the convolutional kernel $\boldsymbol{\psi}$ as a continuous vector-valued function parameterized by a small neural network \mlp$^{\boldsymbol{\psi}}: \sR \rightarrow \sR^{\Nt_{\mathrm{out}} \times \Nt_{\mathrm{in}}}$ (Fig.~\ref{fig:2_ckconv}, left). \mlp$^{\boldsymbol{\psi}}$ receives a relative position $(t {-}\tau)$ and outputs the value of the convolutional kernel at that position $\boldsymbol{\psi}(t {-} \tau)$. As a result, an arbitrarily large convolutional kernel $\gK {=} \{\boldsymbol{\psi}(t {-} \tau)\}_{\tau {=} 0}^{\Nt_{\Kt}}$ can be constructed by providing an equally large sequence of relative positions $\{ t {-} \tau \}_{\tau = 0}^{\Nt_{\Kt}}$ to \mlp$^{\boldsymbol{\psi}}$.
For $\Nt_{\Kt} {=} \Nt_{\Xt}$, the size of the resulting kernel is equal to that of the input sequence $\gX$, and thus it is able to model (global) long-term dependencies.
The \textit{Continuous Kernel Convolution} (CKConv) is given by:
\begin{equation}
    (\xv * \boldsymbol{\psi}) (t) = \sum_{\ct = 1}^{\Nt_{\mathrm{in}}} \sum_{\tau = 0}^{t} x_{\ct}(\tau) \texttt{MLP}^{\boldsymbol{\psi}}_{\ct}(t - \tau).
\end{equation}
\textbf{Irregularly sampled data.} CKConvs are able to handle irregularly-sampled and partially observed data. To this end, it is sufficient to sample \mlp$^{\boldsymbol{\psi}}$ at positions for which the input signal is known and perform the convolution operation with the sampled kernel. For very non-uniformly sampled inputs, an inverse density function over the samples can be incorporated in order to provide an unbiased estimation of the convolution response (see Appx. \ref{appx:2_irregularly_sampled}, \citet{wu2019pointconv} for details). 
% \begin{figure}[t]
%     \centering
%     \includegraphics[width=0.99 \textwidth]{images/ckconv_2.png}
%     \vspace{-2mm}
%     \caption{Continuous kernel convolution. The continuous convolutional kernel, parameterized by an small neural network \mlp$^{\boldsymbol{\psi}}$, receives a sequence of relative positions $\{\Delta\tau_i {=} (t - \tau_i)\}_{i {=} 0}^{\Nt}$ as input and outputs the value of the kernel at those positions $\{\boldsymbol{\psi}(\Delta\tau_i)\}_{i = 0}^{\Nt}$. Consequently, arbitrarily large convolutional kernels can be constructed by providing an equally large sequence of relative positions.
%     \vspace{-7mm}}
%     \label{fig:2_ckconv}
% \end{figure}

\textbf{Data at different resolutions.} CKConvs can also process data at different resolutions. Consider the convolution $(\xv *\boldsymbol{\psi})_\mathrm{sr_1}$ between an input signal $\xv$ and a continuous convolutional kernel $\boldsymbol{\psi}$ sampled at a sampling rate $\mathrm{sr}_1$. Now, if the convolution receives the same input signal sampled at a different sampling rate $\mathrm{sr}_2$, it is sufficient to sample the convolutional kernel at the sampling rate $\mathrm{sr_2}$ in order to perform an \enquote{equivalent} operation:~$(\xv *\boldsymbol{\psi})_\mathrm{sr_2}$. As shown in Appx.~\ref{appx:2_diff_sampling_rates}, it holds that:
\begin{equation}\label{eq:2_diff_sampling_rates}
    (\xv *\boldsymbol{\psi})_\mathrm{sr_2}(t) \approx \frac{\mathrm{sr_2}}{\mathrm{sr_1}}(\xv *\boldsymbol{\psi})_\mathrm{sr_1}(t).
\end{equation}
That is, convolutions calculated at different resolutions $\mathrm{sr_1}$ and $\mathrm{sr_2}$ are approximately equal up to a factor given by the resolution change. As a result, CKCNNs \emph{(i)} can be trained in datasets with data at varying resolutions, and \emph{(ii)} can be deployed at resolutions other than those seen during training.

We note that, expect for RNNs with continuous-time interpretations, e.g., \citet{gu2020hippo,kidger2020neural}, the previous features are difficult to obtain.

%
% -----------SECTION----------------------------
\begin{figure}
     \centering
     \begin{subfigure}[b]{0.24\textwidth}
         \centering
         \includegraphics[width=\textwidth]{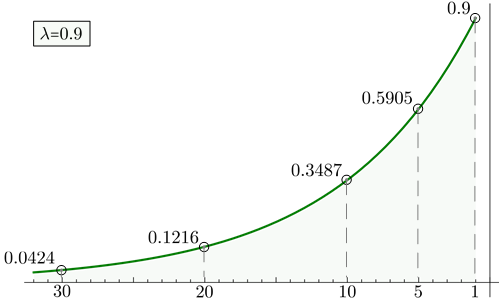}
         \caption{Recurrent ($\lambda {\leq} 1$).}
         \label{fig:2_vanishing_rnn}
     \end{subfigure}
     \hfill
     \begin{subfigure}[b]{0.24\textwidth}
         \centering
         \includegraphics[width=\textwidth]{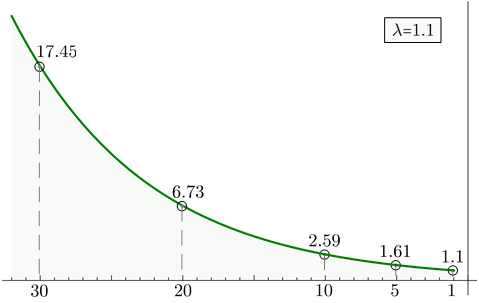}
         \caption{Recurrent ($\lambda {\geq} 1$).}
         \label{fig:2_exploding_rnn}
     \end{subfigure}
     \hfill
     \begin{subfigure}[b]{0.24\textwidth}
         \centering
         \includegraphics[width=\textwidth]{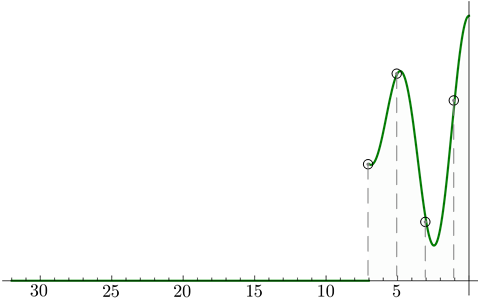}
         \caption{Discrete kernel.}
         \label{fig:2_exploding_conv}
     \end{subfigure}
     \hfill
     \begin{subfigure}[b]{0.24\textwidth}
         \centering
         \includegraphics[width=\textwidth]{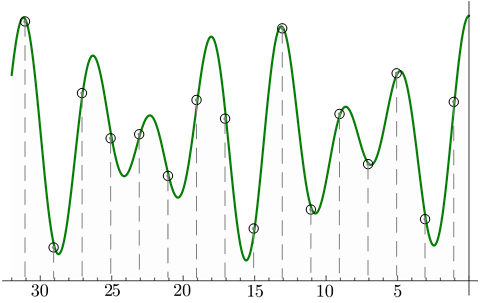}
         \caption{Continuous kernel.}
         \label{fig:2_exploding_ckconv}
     \end{subfigure}
     \vspace{-2mm}
        \caption{Functional family of recurrent units, discrete convolutions and CKConvs. For max. eigenvalues of $\Wm$, $\lambda{\neq}1$, recurrent units are restricted to exponentially decreasing ($\lambda {\leq} 1$) or increasing ($\lambda {\geq} 1$) functions  (Figs.~\ref{fig:2_vanishing_rnn},~\ref{fig:2_exploding_rnn}). Discrete convolutions can describe arbitrary functions within their memory horizon but are zero otherwise (Fig.~\ref{fig:2_exploding_conv}). Conversely, CKConvs define arbitrary long memory horizons, and thus are able to describe arbitrary functions upon the entire input sequence (Fig.~\ref{fig:2_exploding_ckconv}).}
        \label{fig:2_vanishing_exploding_analysis}
\end{figure}
\textbf{(Linear)~recurrent~units~are~continuous~kernel~convolutions.} Consider a recurrent unit
\begin{align}
 \hv(\tau) &= \sigma (\Wm \hv(\tau - 1) + \Um \xv(\tau))\label{eq:2_hidden_repr_formulation}\\[0 \jot]
 \tilde{\yv}(\tau) &= \mathrm{softmax}( \Vm \hv (\tau) ),
\end{align}
where $\Um, \Wm, \Vm$ depict the \textit{input-to-hidden}, \textit{hidden-to-hidden} and \textit{hidden-to-output} connections of the unit, $\hv(\tau)$,  $\tilde{\yv}(\tau)$ the hidden representation and the output at time-step $\tau$, and $\sigma$ a pointwise nonlinearity. As shown in Appx.~\ref{appx:2_linrecunitsasckconvs}, we can express the hidden representation $\hv$ of a linear recurrent unit, i.e., with $\sigma{=}\mathrm{Id}$, as a convolution between the input $\xv$ and a convolutional kernel $\boldsymbol{\psi}(\tau) {=} \Wm^{\tau} \Um$ of size equal to the input. That is, as a continuous kernel convolution with an exponentially increasing or decreasing kernel (Fig.~\ref{fig:2_vanishing_exploding_analysis}). Different authors show that nonlinear recurrent units are also restricted to the same functional family \citep{pascanu2013difficulty, arjovsky2016unitary, zhao2020rnn}.

\textbf{The functional family of continuous kernel convolutions.} From the previous observation, we can conclude that CKConvs are not only more general than discrete convolutions, but that the functional family they describe is also more general than that of (linear) recurrent units (Fig.~\ref{fig:2_vanishing_exploding_analysis}).
%
% -----------SECTION----------------------------
\vspace{-7mm}
\subsection{The Continuous Convolutional Kernel {\btt MLP}$^{\psi}$}\label{sec:2_ckconvkernel}

%
%So far we have discussed continuous convolutional kernel formulation without providing a concrete parameterization to \mlp$^{\boldsymbol{\psi}}$. In this section, we introduce our parameterization, analyze its properties and show that our parameterization is able to generate complex non-linear non-smooth kernels.
\textbf{Convolutional kernels as point-wise {\btt MLP}s.} Let $\{\Delta\tau_{i}{=}(t {-} \tau_{i})\}_{i= 0}^{\Nt}$ be a sequence of relative positions. The continuous vector-valued convolutional kernel $\boldsymbol{\psi}: \sR \rightarrow \sR^{\mathrm{N_{out}}\times \mathrm{N_{in}}}$ is parameterized as a neural network \mlppsi which maps each relative position $\Delta\tau_{i}$ to the value of the convolutional kernel at that position (Fig. \ref{fig:2_ckconv}, left). We refer to the nonlinearity used in \mlppsi as $\sigma$.
% \begin{align}
%     \hv^{(1)}(\Delta\tau_{i}) &= \sigma\big(\vec{w}^{(1)} \Delta\tau_{i} + \bv^{(1)}\big) \label{eq:2_first_layer} \\
%     \hv^{(l)}(\Delta\tau_{i}) &= \sigma\big(\Wm^{(l)} \hv^{(l-1)}(\Delta\tau_{i}) + \bv^{(l)}\big) \\
%     \boldsymbol{\psi}(\Delta\tau_{i}) &= \Wm^{(\Lt)} \hv^{(\Lt-1)}(\Delta\tau_{i}) + \bv^{(\Lt)},\label{eq:2_last_layer}
% \end{align}
% with $\sigma$ a pointwise nonlinearity, e.g., . %This can be interpreted as providing implicit neural representations to the unknown convolutional kernels of a convolutional neural network. 

\textbf{What kind of kernels can {\btt MLP}$^{\psi}$ generate?} Our method relies on the assumption that the neural network \mlppsi is able to model complex dependencies densely among all elements within the memory horizon. That is, it assumes that \mlppsi  is able to generate arbitrary convolutional kernels.
\begin{figure}
    \centering
    \includegraphics[width=0.71 \textwidth]{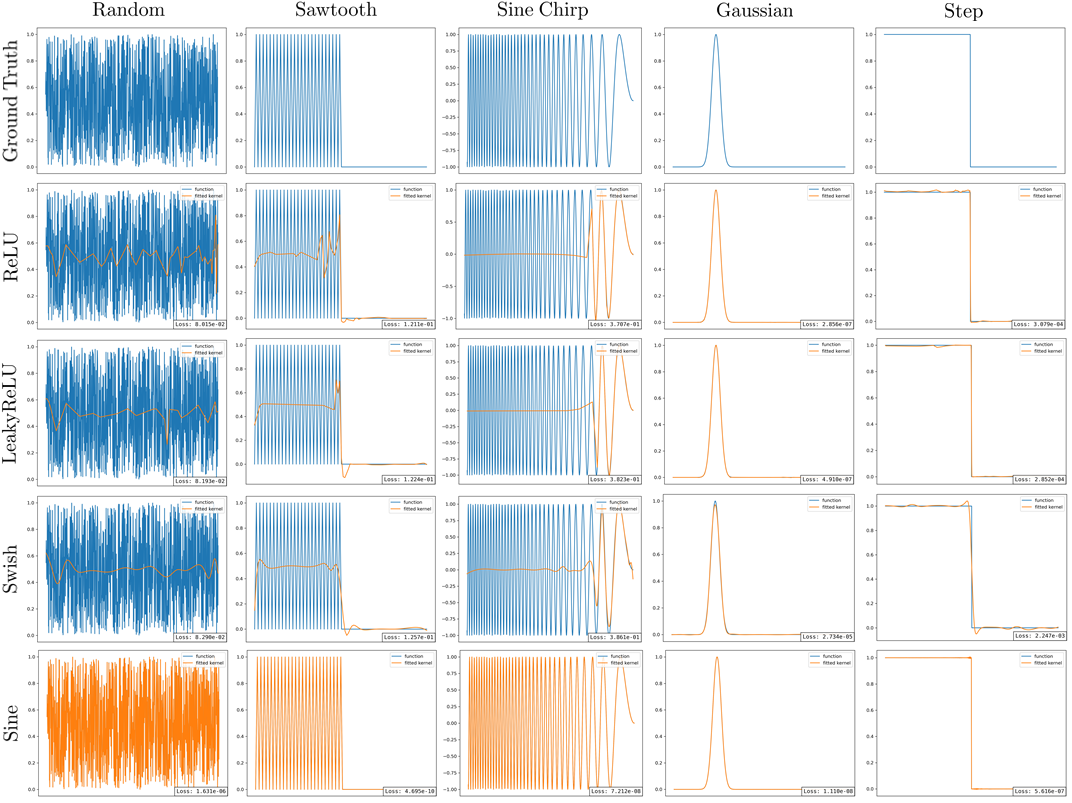}
    \caption{Approximation quality of \mlp s with $\mathrm{ReLU}$, $\mathrm{LeakyReLU}$, $\mathrm{Swish}$, and $\mathrm{Sine}$ nonlinearities. Networks with (smooth) piece-wise nonlinearities are unable to approximate non-smooth, non-linear functions. $\mathrm{Sine}$ networks, on the other hand, quickly approximate all target functions to near perfection. All networks share the same structure and vary only in the nonlinearity used.}
    \label{fig:2_summary_maintext_kernelfit}
\end{figure}

To test this hypothesis, we fit existing \mlp\ parameterizations, i.e., with $\mathrm{ReLU}$, $\mathrm{LeakyReLU}$ and $\mathrm{Swish}$ nonlinearities, to long target functions of varying level of smoothness and non-linearity (Fig.~\ref{fig:2_splineapprox}).
We observe that existing parameterizations can approximate simple functions, e.g., Gaussian, step functions, but fail by a large marking for increasing levels of non-linearity and non-smoothness. For our analysis, this means that CKConvs with $\mathrm{ReLU}$, $\mathrm{LeakyReLU}$ and $\mathrm{Swish}$ parameterizations are \textit{not} able to represent complex input dependencies. In our ablation studies (Appx.~\ref{appx:2_ablation}) we verify that CKCNNs with these kernels consistently perform worse than our proposed parameterization.% on all tasks considered.

\textbf{Convolutional kernels as implicit neural representations.} We notice that parameterizing a convolutional kernel with a neural network is equivalent to constructing an implicit neural representation of the kernel, with the subtle difference that our target objective is not known a-priori, but learned as part of the optimization task. Implicit neural representations study generic ways to represent data living in low-dimensional spaces, e.g., $\sR^{2}$, via neural networks, and thus, despite this difference, constitute an interesting starting point for the parameterization of continuous convolutional kernels.
In particular, recent works noticed that neural networks with piece-wise activation functions are unable to model high-frequencies. To alleviate this limitation, they introduce random Fourier features \citep{tancik2020fourier} and $\mathrm{Sine}$ nonlinearities \citep{sitzmann2020implicit}.

Based on these observations, we repeat the fitting experiment for a SIREN \citep{sitzmann2020implicit}: a \mlp\ with hidden layers of the form $\yv=\mathrm{Sine}(\omega_0 [\Wm\xv + \bv])$.  That is with $\mathrm{Sine}$ nonlinearities, and a non-learnable value $\omega_0$ that serves as a prior for the oscillations of the output. 
We observe that a SIREN quickly approximates \textit{all} target functions to near perfection regardless of their grade of smoothness or nonlinearity. Even a sequence of random noise. %We emphasize that all approximations shown in Fig.~\ref{fig:2_summary_maintext_kernelfit} share the same \mlp\ structure, and vary only in the nonlinearity used.
This implies that, contrary to other parameterizations, CKConvs with SIREN kernels have the ability to model complex input dependencies across large memory horizons. Our experimental results verify this statement. Our ablation studies in Appx.~\ref{appx:2_ablation} show that SIREN kernels consistently outperform all other variants. In addition, our experimental results in Sec.~\ref{sec:2_experiments} show that \textit{shallow} CKCNNs with SIREN kernels achieve state-of-the-art across datasets of different nature, i.e., with continuous and discrete data.
%In a subsequent study, we found that $\mathrm{ReLU}$ networks with Random Fourier Features \citep{tancik2020fourier} could outperform SIRENs for some of the approximation tasks illustrated in Fig.~\ref{fig:2_summary_maintext_kernelfit}. The experimental results as well as complementary ablation studies can be found in Appx.~\ref{}. 

\textbf{The success of $\mathrm{Sine}$ nonlinearites: A spline basis interpretation.} \citet{sitzmann2020implicit} motivates the usage of Sine nonlinearities for implicit neural representations. However, it is not clear \textit{why} Sine nonlinearities are better suited for this task than (smooth) piece-wise nonlinearities. For the interested reader, we provide an interpretation to this phenomenon from a spline function approximation perspective in Appx.~\ref{appx:2_spline_interpretation}.

Of most practical relevance from this analysis is the observation that proper initialization of the network parameters, particularly of the bias term $\{\bv^{(l)}\}$, is important to create a well-spread set of basis functions suited for function approximation. For SIRENs, this is achieved by initializing the bias term uniformly across the period of each of the $\mathrm{Sine}$ components: $\bv_{i} \sim \mathcal{U}(-\pi \| \Wm_{i,:}\|^{-1},\pi \| \Wm_{i,:}\|^{-1})$. We observe that this initialization leads to better results and faster convergence for all tasks considered.

\vspace{-7mm}
\section{Experiments}\label{sec:2_experiments}

We validate our approach against several existing models and across several tasks selected from the corresponding papers. Specifically, we benchmark its ability to handle long-term dependencies, data at different resolutions and irregularly-sampled data. A complete description of the datasets used as well as additional experiments and ablation studies can be found in the Appendix (Appx.~\ref{appx:2_datadescription},~\ref{appx:2_ablation}).

\textbf{Network details.} We parameterize our convolutional kernels as 3-layer SIRENs. Weight normalization \citep{salimans2016weight} leads to better and faster convergence when applied to the layers in \mlp, and we use it across all experiments. %in our kernels \mlp$^{\boldsymbol{\psi}}$: $\Wm_{i,:}= \vec{g}_{i} \tfrac{\Vm_{i,:}}{\| \Vm_{i,:}\|}$ \citep{salimans2016weight}, which allows for separate control of the magnitude $\vec{g}_{i}$ and the direction $\tfrac{\Vm_{i,:}}{\| \Vm_{i,:}\|}$ of the mappings, and leads to better and faster convergence.
All our CKCNNs follow the structure shown in Fig.~\ref{fig:2_ckcnn_structure} and vary only in the number of blocks and channels. We use two residual blocks for all experiments reported in this section. Specifications on the architectures and hyperparameters used are given in Appx.~\ref{appx:2_expdetails}.
We speed up the convolutions by using the \textit{convolution theorem} $(f * \psi)\hspace{-0.7mm} = \hspace{-0.3mm}\mathcal{F}^{-1}\hspace{-0.5mm}\big\{\mathcal{F}\{f\} \hspace{-0.3mm}\cdot\hspace{-0.3mm} \overline{\mathcal{F}\{\psi\}}\big\}$, with $\mathcal{F}$ the Fourier transform.
%  -=== Table

%\subsection{Long-range Memory Benchmark Tasks}
\textbf{Stress experiments.}
First, we validate that CKCNNs can readily model memory horizons of different lengths. To this end, we evaluate if a shallow CKCNN is able to solve the \textit{Copy Memory} and \textit{Adding Problem} tasks \citep{hochreiter1997long} for sequences of sizes in the range $[100, 6000]$. Success is achieved if 100\% accuracy, or a loss $\leq$ 1e-4 are obtained, for the copy memory and adding problem, respectively. Random predictions for the adding problem lead to a loss of approx. $0.17$.

Our results show that a shallow CKCNN is able to solve both problems for all sequence lengths considered without requiring structural modifications (Tab.~\ref{tab:2_addcopyprob}). Recurrent architectures are not able not solve the copy problem at all and could solve the sum problem up to 200 steps. TCNs with $k{=}7$, $n{=}7$ were able to solve both tasks for up to 1000 steps. However, larger lengths were out of reach as their memory horizon is constrained a priori. To handle larger sequences, TCNs must modify their network structure based on prior knowledge regarding the expected length of the input sequence. 

\textbf{Discrete sequences.} The continuous nature of our kernels might give the impression that CKCNNs are only suited for continuous data, i.e., time-series. However, $\mathrm{Sine}$ nonlinearities allow our convolutional kernels to model complex non-smooth non-linear functions (Fig.~\ref{fig:2_summary_maintext_kernelfit}). Consequently, we validate whether CKCNNs can be applied for discrete sequence modeling on the following tasks: \textit{sMNIST}, \textit{pMNIST} \citep{le2015simple}, \textit{sCIFAR10} \citep{trinh2018learning} and \textit{Char-level PTB} \citep{marcinkiewicz1994building}.

%We compare CKCNNs against recurrent: LSTM, GRU \citep{bai2018empirical}, IndRNN \citep{li2018independently}, DilRNN \citep{chang2017dilated}, HiPPO \citep{gu2020hippo}, r-LSTM: \citep{trinh2018learning}, self-attention \citep{trinh2018learning}, and convolutional architectures: TCN \citep{bai2018empirical}, TrellisNet \citep{bai2018trellis}. 
Shallow CKCNNs outperform recurrent, self-attention and convolutional models on sMNIST and pMNIST (Tab.~\ref{tab:2_smnist}). On sMNIST, a small CKCNN (100{\sc{k}} params.) achieves state-of-the-art results with a model 80$\times$ smaller than the current state-of-the-art. A wider CKCNN (1{\sc{m}} params.) slightly increases this result further. On pMNIST, we see an improvement of $0.8\%$ over the best model of size $\leq$100{\sc{k}}, and our wider shallow CKCNN achieves state-of-the-art on this dataset. For sCIFAR10, our small CKCNN obtains similar results to a self-attention model 5$\times$ bigger, and our wider variant improves performance by an additional $1\%$. Our best results are obtained with an even wider model (2.5{\sc{m}} params) with which an accuracy of $65.59\%$ is obtained. On Char-level PTB a CKCNN with 3{\sc{m}} parameters outperforms all models considered as well as the state-of-the-art: Mogrifier LSTMs \citep{melis2019mogrifier}, while being 13.3$\times$ smaller.
\begin{table}
\caption{Test results on discrete sequential datasets.}
\label{tab:2_smnist}
\vskip -4mm
\begin{small}
\scalebox{0.8}{
\begin{tabular}{cccccc}
\toprule
\multirow{2}{*}{\sc{Model}} & \multirow{2}{*}{\sc{Size}}  & \sc{sMNIST} & \sc{pMNIST} & \sc{sCIFAR10} & \sc{Char-PTB} \\
 &  & Acc (\%) & Acc (\%) & Acc (\%) & bpc\\
 \midrule
 TCN \citep{bai2018empirical} &70\sc{k}& \textbf{99.0} & \textbf{97.2}  & - & \textbf{1.31}$^{\dagger}$  \\
  LSTM \citep{bai2018empirical} &70\sc{k}& 87.2 & 85.7  & - & 1.36$^{\dagger}$ \\
 GRU \citep{bai2018empirical} &70\sc{k}&   96.2 & 87.3  & - & 1.37$^{\dagger}$ \\
 IndRNN \citep{li2018independently} & 83\sc{k} & \textbf{99.0} & 96.0& - & - \\
 DilRNN \citep{chang2017dilated} & 44\sc{k} & 98.0 & 96.1 & - & - \\
 \midrule
 HiPPO \citep{gu2020hippo} & 0.5\sc{m} & - & \textbf{98.30} & - & - \\
  r-LSTM \citep{trinh2018learning} & 0.5\sc{m} & 98.4 & 95.2 & 72.2 & - \\
 Self-Att. \citep{trinh2018learning} &0.5\sc{m} & 98.9 & 97.9 & 62.2 & - \\
 TrellisNet \citep{bai2018trellis}& 8\sc{m} & \textbf{99.20} & 98.13 & \textbf{\underline{73.42}} & \textbf{1.158}$^{\dagger}$\\
\midrule
CKCNN & 98\sc{k} & \textbf{99.31} & \textbf{98.00} & 62.25 & - \\
CKCNN-Big & 1\sc{m} & \textbf{\underline{99.32}} & \textbf{\underline{98.54}} & 63.74 & \textbf{\underline{1.045}}$^{\dagger}$\\
\bottomrule
\multicolumn{6}{l}{$^{\dagger}$ Model sizes are 3{\sc{m}} for TCN, LSTM and GRU, 13.4{\sc{m}} for TrellisNet and 1.8{\sc{m}} for CKCNN-Big.}
\end{tabular}}
\end{small}
\end{table}
\begin{table}
\RawFloats
\centering
\begin{minipage}{0.44 \textwidth}
\begin{center}
\caption{Evaluation on stress tasks. {\color{Green}\cmark} marks if the problem has been solved.}
\label{tab:2_addcopyprob}
\vskip -3mm
\begin{small}
\scalebox{0.8}{
\begin{tabular}{ccccccc}
\toprule
\multirow{2.5}{*}{\sc{Model}} & \multirow{2.5}{*}{\sc{Size}} & \multicolumn{5}{c}{\sc{Seq. Length}} \\
\cmidrule{3-7}
 & & $100$  & $200$  & $1000$ & $3000$ & $6000$  \\
 \toprule
 \multicolumn{7}{c}{\sc{{Copy Memory}}} \vspace{0.5mm}\\
 GRU &16\sc{k}& - & - & - & - & - \\
 TCN &16\sc{k}& {\color{Green}\cmark }  & {\color{Green}\cmark } & {\color{Green}\cmark } & - & -  \\
CKCNN & 16\sc{k} & {\color{Green}\cmark }& {\color{Green}\cmark}& {\color{Green}\cmark } & {\color{Green}\cmark } &{\color{Green}\cmark } \\
\toprule
\multicolumn{7}{c}{\sc{{Adding Problem (Loss)}}} \vspace{0.5mm}\\
 GRU &70\sc{k}& {\color{Green}\cmark } & {\color{Green}\cmark }  & - \\
  TCN &70\sc{k}& {\color{Green}\cmark } &{\color{Green}\cmark } & {\color{Green}\cmark } & - & - \\
CKCNN & 70\sc{k} &{\color{Green}\cmark } &{\color{Green}\cmark } & {\color{Green}\cmark } & {\color{Green}\cmark } & {\color{Green}\cmark }  \\
\bottomrule
\end{tabular}}
\end{small}
\end{center}
\end{minipage}%
\hfill
\begin{minipage}{0.53 \textwidth}
\centering
\caption{Test results on CT, SC and SC\_raw.}
\label{tab:2_timeseries}
\begin{center}
\vskip -3mm
\begin{small}
\scalebox{0.8}{
\begin{tabular}{cccccc}
\toprule
\sc{Model} & \sc{Size} & \sc{CT} & \sc{SC} & \sc{SC\_raw} \\
 \midrule
%  GRU-ODE & 89\sc{k} & 96.2 & 44.8 & $\sim$ 10.0 \\
  GRU-ODE &  \multirow{2}{*}{89\sc{k}} &  \multirow{2}{*}{96.2} &  \multirow{2}{*}{44.8} &  \multirow{2}{*}{$\sim$ 10.0} \\
 \citep{de2019gru} \\
%  GRU-$\Delta t$ & 89\sc{k} & 97.8 & 20.0 &  $\sim$ 10.0 \\
 GRU-$\Delta t$ & \multirow{2}{*}{89\sc{k}} & \multirow{2}{*}{97.8} & \multirow{2}{*}{20.0} &  \multirow{2}{*}{$\sim$ 10.0} \\
 \citep{kidger2020neural}\\
%  GRU-D & 89\sc{k} & 95.9 & 23.9 &  $\sim$ 10.0 \\
 GRU-D & \multirow{2}{*}{89\sc{k}} & \multirow{2}{*}{95.9} & \multirow{2}{*}{23.9} &  \multirow{2}{*}{$\sim$ 10.0} \\
 \cite{che2018recurrent} \\
%  ODE-RNN & 88\sc{k} & 97.1 & 93.2 &  $\sim$ 10.0\\
 ODE-RNN & \multirow{2}{*}{88\sc{k}} & \multirow{2}{*}{97.1} & \multirow{2}{*}{93.2} &  \multirow{2}{*}{$\sim$ 10.0}\\
 \citep{rubanova2019latent}\\
%  NCDE & 89\sc{k} & 98.8 & 88.5 &  $\sim$ 10.0 \\
 NCDE & \multirow{2}{*}{89\sc{k}} & \multirow{2}{*}{98.8} & \multirow{2}{*}{88.5} &  \multirow{2}{*}{$\sim$ 10.0} \\
  \citep{kidger2020neural}\\
 \midrule
 CKCNN & 100\sc{k}& \textbf{99.53} & \textbf{95.27} & \textbf{71.66}\\
\bottomrule
\end{tabular}}
\end{small}
\end{center}
\end{minipage}
\end{table}

\textbf{Time-series modeling.} Next, we evaluate CKCNNs on time-series data. To this end, we consider the \textit{CharacterTrajectories (CT)} \citep{bagnall2018uea} and the \textit{Speech Commands (SC)} \citep{warden2018speech} datasets. We follow \citet{kidger2020neural} to obtain a balanced classification dataset with precomputed mel-frequency cepstrum coefficients. In addition, we evaluate the ability of CKCNNs to model long-term dependencies by training on the raw SC dataset (\textit{SC\_raw}), whose records have length 16k.

We compare CKCNNs to representative sequential models with continuous-time interpretations: GRU-ODE \citep{de2019gru}, GRU-$\Delta t$ \citep{kidger2020neural}, ODE-RNN \citep{rubanova2019latent}, and NCDE \citep{kidger2020neural}. Sequential continuous-time models were selected as they are only sequential methods also able to handle irregularly-sampled data, and data at different resolutions. 
Our results show that shallow CKCNNs outperform all continuous-time models considered for both the CT and SC datasets (Tab.~\ref{tab:2_timeseries}). In addition, CKCNNs obtain promising results on SC\_raw, which validates their ability to handle very-long-term dependencies. In fact, CKCNNs trained on SC\_raw are able outperform several Neural ODE models trained on the preprocessed data (SC).

In addition, we observed that neural ODE methods considered in Tab.~\ref{tab:2_timeseries} were prohibitively slow for long sequences. For instance, NCDEs were $228\times$ slower than a CKCNN of equivalent size on SC\_raw, taking 17 hours per epoch to train. Consequently, training a NCDE on SC\_raw for a matching number of epochs would take more than $212$ days to conclude. In order to provide results for these models, we train them under the same computational budget than CKCNNs. This is enough to train them for a single epoch. All obtained results are at best only marginally better than random.

\textbf{Testing at different sampling rates.} We now consider the case where a network is trained with data at a sampling rate $\mathrm{sr}_1$, and tested with data at a different sampling rate $\mathrm{sr}_2$. Our results show that the performances of CKCNNs remains stable for large sampling rate fluctuations (Tab.~\ref{tab:2_diffsr}). This behaviour contrasts with most previous continuous-time models, whose performance rapidly decays upon these changes. CKCNNs outperform HiPPO \citep{gu2020hippo} and set a new state-of-the-art in this setting. Importantly, depending on the sampling, additional care may be needed to account for spatial displacements and high-frequencies of our kernels (see Appx.~\ref{appx:2_spatialdisplacement} for details).

\textbf{Irregularly-sampled data.} To conclude, we validate CKCNNs for irregularly-sampled data. To this end, consider the PhysioNet sepsis challenge \citep{reyna2019early} as well as the CT dataset with drops of 30\%, 50\% and 70\% of the data as in \citet{kidger2020neural}. In addition, we provide results under the same methodology for the SC\_raw dataset.% As in \citet{kidger2020neural}, we add an additional channel to the input to indicate whether the value at that position is known. %We omit the SC dataset as the preprocessed bins are calculated with overlapping windows and thus information from dropped points might still be present.
%  -=== Table
\begin{table}
\centering
\caption{Test results on irregular data.}
\label{tab:2_irregsampled}
\begin{center}
\vskip -3mm
\begin{small}
\scalebox{0.81}{
\begin{tabular}{cccccccccc}
\toprule
\multirow{2.5}{*}{\sc{Model}} &\multirow{1.5}{*}{\sc{PhysioNet}} & \multicolumn{4}{c}{\sc{{CharacterTrajectories}}} & \multicolumn{4}{c}{\sc{{SpeechCommands\_raw}}} \\
\cmidrule{3-6}
\cmidrule{7-10}
& AUC & \sc{(0\%)} & \sc{(30\%)} & \sc{(50\%)} & \sc{(70\%)} & \sc{(0\%)} & \sc{(30\%)} & \sc{(50\%)} & \sc{(70\%)} \\
 \midrule
 GRU-ODE & 0.852 & 96.2 & 92.6 & 86.7 & 89.9 & $\sim$ 10.0 & $\sim$ 10.0 & $\sim$ 10.0 & $\sim$ 10.0 \\
 GRU-$\Delta t$ & 0.878 & 97.8 & 93.6 & 91.3 & 90.4 & \multirow{4}{*}{$\vdots$} & \multirow{4}{*}{$\vdots$}  & \multirow{4}{*}{$\vdots$}  & \multirow{4}{*}{$\vdots$} \\
 GRU-D  & 0.871 & 95.9 & 94.2 & 90.2 & 91.9 \\
 ODE-RNN & 0.874 & 97.1 & 95.4 & 96.0 & 95.3 \\
 NCDE & 0.880 & 98.8  & 98.7 & \textbf{98.8} & \textbf{98.6}\\
 CKCNN & \textbf{0.895} & \textbf{99.53} & \textbf{99.30} & \textbf{98.83} & 98.14 & \textbf{71.66} & \textbf{63.46} & \textbf{60.55} & \textbf{57.50}  \\
\bottomrule
\end{tabular}}
\end{small}
\end{center}
\end{table}
\begin{table}
\centering
\caption{Results for different train and test resolutions. Fractions depict resolutions proportional to the original one of the dataset. The accuracy of all models on the original resolution surpasses $90\%$.}
\label{tab:2_diffsr}
\begin{center}
\vskip -3mm
\begin{small}
\begin{sc}
\scalebox{0.85}{
\begin{tabular}{ccccccc}
\toprule
 \multicolumn{7}{c}{\textbf{CKCNN} -- Size${=}${100k}}\\
 \toprule
 \multirow{2}{*}{Dataset} & \multirow{2}{*}{Train Freq.} & \multicolumn{5}{c}{Test Freq.}\\
 \cmidrule{3-7}
 & & $1$ & $\nicefrac{1}{2}$ & $\nicefrac{1}{4}$ & $\nicefrac{1}{8}$ & $\nicefrac{1}{16}$ \\
 \midrule
 \multirow{3}{*}{CT} & $1$ & \textbf{99.53} & 99.30 & 99.30 & 95.80 & 76.45 \\
 & $\nicefrac{1}{2}$ & 98.83 & \textbf{99.07} & 98.37 &96.97 & 80.42 \\
 & $\nicefrac{1}{4}$ & 96.74 & 96.97 & \textbf{99.30} &98.83 &84.85  \\
 & $\nicefrac{1}{8}$ & 96.97 & 97.44 & 97.20 & \textbf{99.30} & 73.43 \\
  \midrule
 \multirow{3}{*}{SC\_raw} & $1$ & \textbf{71.66} & 65.96 & 52.11 & 40.33 & 30.87\\
  & $\nicefrac{1}{2}$& 72.09 & \textbf{72.06} & 69.03 & 63.00 & 29.67 \\
 & $\nicefrac{1}{4}$ & 68.25 & 68.40 & \textbf{69.47} & 67.09 & 37.91\\
 & $\nicefrac{1}{8}$ & 40.48 & 42.00 & 54.91 & \textbf{66.44} & 22.29\\
 \bottomrule
\end{tabular}}
\vskip -0mm
\scalebox{0.81}{
\begin{tabular}{ccccccc}
\toprule
\multicolumn{7}{c}{\textbf{Model Comparison - Character Trajectories}}\\
\toprule
 Model  & GRU-D & ODE-RNN & LMU & NCDE & HiPPO & CKCNN \\
\midrule
$1 \rightarrow \nicefrac{1}{2}$  &23.1 & 41.8 & 44.7 & 6.0 & 88.8 & \textbf{99.30} \\
$\nicefrac{1}{2} \rightarrow 1$ & 25.5 & 31.5 & 11.3 & 13.1 & 90.1 & \textbf{98.83} \\
\bottomrule
\end{tabular}}
\end{sc}
\end{small}
\end{center}
\end{table}

Our results show that CKCNNs outperform NCDEs and obtain state-of-the-art performance on the PhysioNet dataset.  In addition, CKCNNs exhibit stable performance for varying quantities of missing data, and perform better than several models explicitly developed to this end (Tab.~\ref{tab:2_irregsampled}). On the CT dataset, NCDEs perform slightly better than CKCNNs for large data drop rates. However, we argue that our method is still advantageous due to the gains in training speed --see Section~\ref{sec:2_discussion} for details--.
% -----------SECTION----------------------------
\vspace{-7mm}
\section{Discussion and Limitations}\label{sec:2_discussion}

\textbf{Parameter-efficient large convolutional kernels.}  CKConvs construct large complex kernels with a fixed parameter budget. For large input sequences, this results in large savings in the number of parameters required to construct global kernels with conventional CNNs. For sequences from the pMNIST (length {=} 784) and SC\_raw (length {=} 16000) datasets, a conventional CNN with global kernels would require 2.14{\sc{m}} and 46.68{\sc{m}} of parameters, respectively, for a model equivalent to our CKCNN (100{\sc{k}}). In other words, our kernel parameterization allows us to construct CKCNNs that are $21,84$ and $445,71$ times smaller than corresponding conventional CNNs for these datasets. 
Detailed exploration on the effect of our efficient  continuous kernel parameterizations in optimization, overfitting and generalization is an interesting direction for future research.

\textbf{Is depth important? Shallow global memory horizons.} Our results are obtained with CKCNNs built with two residual blocks only. Additional experiments (Appx.~\ref{appx:2_depthvsshallow}) indicate that our models do not benefit from larger depth, and suggest that CKCNNs do not rely on very deep features. Though further analysis is required to draw consistent conclusions, it is intriguing to explore if it is sufficient to equip neural networks with global memory horizons even if this happens in a shallow manner.% This hypothesis is in line with the predominant design of recurrent architectures, for which a moderate number of layers are used, e.g., \cite{pascanu2013construct, graves2013speech, gu2020improving, gu2020hippo}. We consider corroborating or refuting this hypothesis as an important step towards the understanding of CKCNNs and deep learning in general.

\textbf{High-frequency components.} Interestingly, our kernels often contain frequency components higher than the resolution of the grid used during training (Fig.~\ref{fig:2_high_freq}). As a result, transitions to finer resolutions benefit from smoothing (see Appx.~\ref{appx:2_high_freq}). Nevertheless, we believe that, if tuned properly, these high-frequency components might prove advantageous for tasks such as super-resolution and compression.

\textbf{Faster continuous-time models.} CKCNNs rely on convolutions, and thus can be executed in parallel. As a result, CKCNNs can be trained faster than recurrent architectures. This difference becomes more pronounced with concurrent continuous-time models for sequential data, which are based on neural ODEs and require at least 5$\times$ slower than RNNs \citep{kidger2020neural}. At the cost of larger memory costs, CKCNNs can be further sped up by using the convolution theorem. 

\textbf{Neural networks parameterizing spatial functions should be able to model high-frequencies.} 
Our findings indicate that, common nonlinearities do not provide MLPs modelling spatial continuous functions the ability to model high-frequencies. Consequently, architectures that model continuous spatial functions via neural networks should transition towards models endowed with this ability, e.g., MLPs with $\mathrm{Sine}$ nonlinearities. These models encompass convolutional networks with continuous kernels \cite{schutt2017schnet, thomas2018tensor, wu2019pointconv}, positional encodings in transformers \cite{romero2021group, hutchinson2021lietransformer}, and graph neural networks \cite{defferrard2020deepsphere}. $\mathrm{Sine}$ nonlinearities can be used to reduce the number of parameters needed to model local functions, or to extend the receptive field of the operations efficiently.
%textbf{Neural networks parameterizing spatial functions should use $\mathrm{Sine}$ nonlinearities.} Our findings indicate that neural networks with $\mathrm{Sine}$ nonlinearities are much better suited to describe spatial continuous functions than all other nonlinearities commonly used for this purpose. This indicates that architectures modelling continuous spatial functions via neural networks should transition towards $\mathrm{Sine}$ nonlinearities. These models encompass convolutional networks with continuous kernels, e.g., \cite{schutt2017schnet, thomas2018tensor, wu2019pointconv}, as well as positional encodings in transformers, e.g., \cite{romero2020group, hutchinson2021lietransformer}, and graph neural networks, e.g., \cite{defferrard2020deepsphere}. $\mathrm{Sine}$ nonlinearities can either be used to reduce the number of parameters needed to model local functions, or to extend the receptive field of the operations efficiently.
% \vspace{-1mm}
% \subsection{Limitations}\label{sec:2_limitations}
% \vspace{-1mm}
%  \vspace{-2mm}
% \section{Limitations}
% \vspace{-2mm}

\textbf{Memory requirements.}  Although, CKCNNs can be deployed and trained in parallel, CKCNNs must store the convolution responses at each layer and for all input positions. This induces a linear memory complexity with regard to the sequence length, and largely contrasts with recurrent continuous-time models, whose memory complexity is constant. The memory consumption of the operation is further incremented if the convolution theorem is applied because it requires multiplying the Fourier transform of the convolution and the kernel, and taking them back to the temporal representation.
On the other hand, large convolutional kernels seem to allow CNNs to perform well without using many layers, which has a positive effect on memory consumption.

\textbf{Selection of $\boldsymbol{\omega_0}$.} We observe that CKCNNs are very susceptible to the selection of $\omega_0$. For instance, performance on pMNIST may vary from $98.54$ to $65.22$ for values of $\omega_0$ in $[1, 100]$. Consequently, finding a good value of $\omega_0$ induces an important cost in hyperpararameter search (see Appx.~\ref{appx:2_hyperparams}). 
$\omega_0$ acts as a prior on the variability of the target function. However, it is not obvious which value of $\omega_0$ is optimal for the internal (unknown) features of a network. Learning layer-wise $\omega_0$ values yielded sub-optimal results and the best results were obtained with a predefined $\omega_0$ value across all layers.
%  \vspace{-1mm}
% \subsection{Future work}
%  \vspace{-1mm}

 % -----------SECTION----------------------------
\vspace{-7mm}
 \section{Conclusion and Future Work}

We introduced the Continuous Kernel Convolution (CKConv), a simple, yet powerful approach able to model global long-term dependencies effectively in a parameter-efficient manner. Aside from the ability to get good accuracy, CKConvs are readily able to handle irregularly-sampled data, and data at different resolutions. CKCNNs achieve state-of-the-art results on multiple datasets, and often surpass neural architectures designed for particular settings, e.g., for irregularly-sampled data. 

%\textbf{Future work:}
 We are intrigued about the potential of CKCNNs for tasks in which (global) long-term dependencies play a crucial role, e.g., audio, video, reinforcement learning, (autoregressive) generative modeling. The usage of CKConvs to model long-term interactions in images is also very promising.
 In addition, CKConvs provide a convenient way to study the effect of the receptive field size of convolutional architectures, as no network modifications are required for different sizes. Our findings may also be useful for specific problems with irregularly-sampled data, e.g., medical, point clouds. We are also excited about structural advances of CKConvs. For instance, attentive versions of CKCNNs, or formulations that further improve computation and parameter efficiency%, e.g., attentive CKConvs, and further understanding of their underlying learning dynamics.

 \textbf{Alleviating limitations.} Reducing the memory consumption of CKConvs is vital for its application on a broad range of scenarios, e.g., embedded devices. Moreover, finding kernel parameterizations more stable to hyperparameter changes is desirable to reduce the need for hyperparameter search.
 
 \textbf{What is the best implicit kernel parameterization for convolutional kernels?} Despite the success of SIRENs, we believe that better kernel parameterizations might still be constructed, e.g., with Random Fourier Features \citep{tancik2020fourier}. Aside from improvements in implicit neural representations, which are directly transferable to CKConvs, we consider important to analyze the effect that having unknown, changing target objectives has on the approximation. A thorough empirical study of possible kernel parameterizations is an important direction for future research. A parameterization with which additional desiderata, e.g., smoothness, can be imposed is also desirable.
%CKConvs provide a simple yet powerful way to model global long-term interactions.
%=========================================================================

%% file: chapters/4_ccnn.tex
% path to figures directory
\graphicspath{{figures/4-ccnn/}}

%=========================================================================

% \begin{savequote}[75mm]
% Nulla facilisi. In vel sem. Morbi id urna in diam dignissim feugiat. Proin molestie tortor eu velit. Aliquam erat volutpat. Nullam ultrices, diam tempus vulputate egestas, eros pede varius leo.
% \qauthor{Quoteauthor Lastname}
% \end{savequote}

\chapter{Modelling Long Context in $N$D: From Task-Specific to a General Purpose CNN}
	\label{chapter:ccnn}
 
\begin{flushright}
\textit{Based on the papers:}\break
\textit{Towards a General Purpose CNN for Long Range Dependencies in $N$D \citep{romero2022towards}}\break
\textit{Modeling Long Context in $N$D: From Task-Specific to a General Purpose CNN \citep{knigge2023modelling}}
\end{flushright}
%=========================================================================
\vspace{-7mm}
\section{Introduction}

The vast popularity of Convolutional Neural Networks \citep{lecun1998gradient} (CNNs) is a result of their high performance and efficiency, which has led them to achieve state-of-the-art in applications across sequential \citep{abdel2014convolutional, van2016wavenet}, visual \citep{krizhevsky2012imagenet, simonyan2014very} and high-dimensional data \citep{schutt2017schnet, wu2019pointconv}. Nevertheless, a major limitation of CNNs --and other neural networks-- is that their architectures must be tailored to particular applications in order to consider the length, resolution and dimensionality of the input data. This has led to a plethora of task-specific architectures \citep{oord2016wavenet, bai2018empirical, simonyan2014very, szegedy2015going, ronneberger2015u, he2016deep, qi2017pointnet, wu2019pointconv} which (\textit{i}) hampers the selection of the most appropriate architecture for a particular task, and (\textit{ii}) obscures the transfer and generalization of insights across applications. In this work, we tackle the need for problem-specific CNN architectures and propose a generic CNN architecture that can be used independent of the length, resolution and dimensionality of the data.
%are \emph{de facto} the most successful neural network architecture to date, a feat attributable to their high performance and efficiency, as well as their phenomenal flexibility in model design choices. Although this allowed modelling specification in applications across sequential \citep{abdel2014convolutional, van2016wavenet}, visual \citep{krizhevsky2012imagenet, simonyan2014very} and high-dimensional data \citep{schutt2017schnet, wu2019pointconv}, the discrete nature of the classical CNN formulation led to severe fragmentation of the neural network formalism. CNN architectures must be tailored towards particular applications in order to handle different data lengths, resolutions and dimensionalities, leading to an extensive number of task-specific CNN architectures \citep{oord2016wavenet, bai2018empirical, simonyan2014very, szegedy2015going, ronneberger2015u, he2016deep, qi2017pointnet, wu2019pointconv}, making it hard to formalize and transfer insights between application domains. In this work, we introduce the \textit{Continuous Convolutional Neural Network} (CCNN), a network architecture independent of the aforementioned data characteristics, and show its applicability in a range of tasks in varying domains.
\begin{figure}[b]
    \centering
     \includegraphics[width=\textwidth]{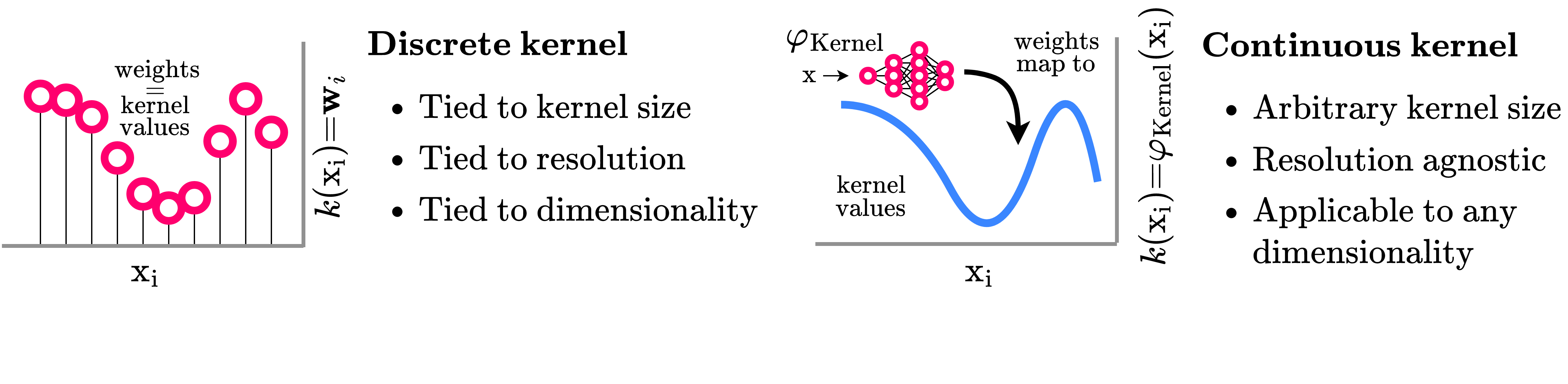}
     \vspace{-14mm}
     \caption{Discrete and continuous convolutional kernels. Discrete convolutional kernels assign a weight $\wv_i$ out of a discrete set of weights $\mat{W}$ to a relative offset ${\xv} - \tilde{\xv}$. This ties the kernel to the length, resolution and dimensionality of the input, limiting the general applicability of the CNN architectures. Instead, our \textit{Continuous Convolutional Neural Network} parameterizes kernel values as a continuous function $\varphi_{\rm Kernel}$ over the input domain $\mathbb{R}^d$, which decouples it from data characteristics.}
    \label{fig:4_continuous-kernels}
\end{figure}

\textbf{CNN architectures are data dependent.} Current CNN architectures are task-specific because they are tied to the \textit{length}, \textit{resolution}, and \textit{dimensionality} of the input. The \textit{length} of the data varies from task to task, e.g. audio fragments may span milliseconds to minutes. This requires carefully chosen strides and pooling to capture relevant dependencies across the entire input \citep{van2016wavenet, lee2017raw}. In addition, physical signals, e.g., audio, images, are continuous in nature. As such, their semantic meaning is independent of the \textit{resolution} at which they are sampled, e.g., the same audio may be expressed at different resolutions. Nevertheless, current CNN architectures are resolution-bound, and thus different resolutions require different CNNs. These limitations aggravate when considering \textit{multi-dimensional} data. Each input dimension can be defined at different lengths and resolutions, e.g., video, rectangular images, and each data modality brings its own conventions for each of these properties, e.g., the resolution of a second of audio ($16\mathrm{kHz}$) \citep{warden2018speech} strongly contrasts with that of images ($32\times32$) \citep{krizhevsky2009learning}.

\textbf{Towards a unified CNN architecture.} As discussed in Sec.~\ref{sec:4_from_discr_to_cont}, the core component that makes CNNs data-dependent are their \textit{discrete convolutional kernels}. Convolutional kernels are implemented via a one-to-one mapping between kernel values and model parameters (Fig.~\ref{fig:4_continuous-kernels} left), which (\textit{i}) binds them to the input resolution and length, and (\textit{ii}) makes them ill suited to model long-range dependencies. The latter results from the large number of parameters needed to construct large kernels. This is why standard CNNs favour local kernels in combination with task-dependent depths and pooling layers to model long-range dependencies, at the cost of making them task-dependent.

\begin{figure}
    \centering
         \begin{subfigure}[b]{0.3\textwidth}
         \centering
         \captionsetup{justification=centering}
         \includegraphics[width=\textwidth]{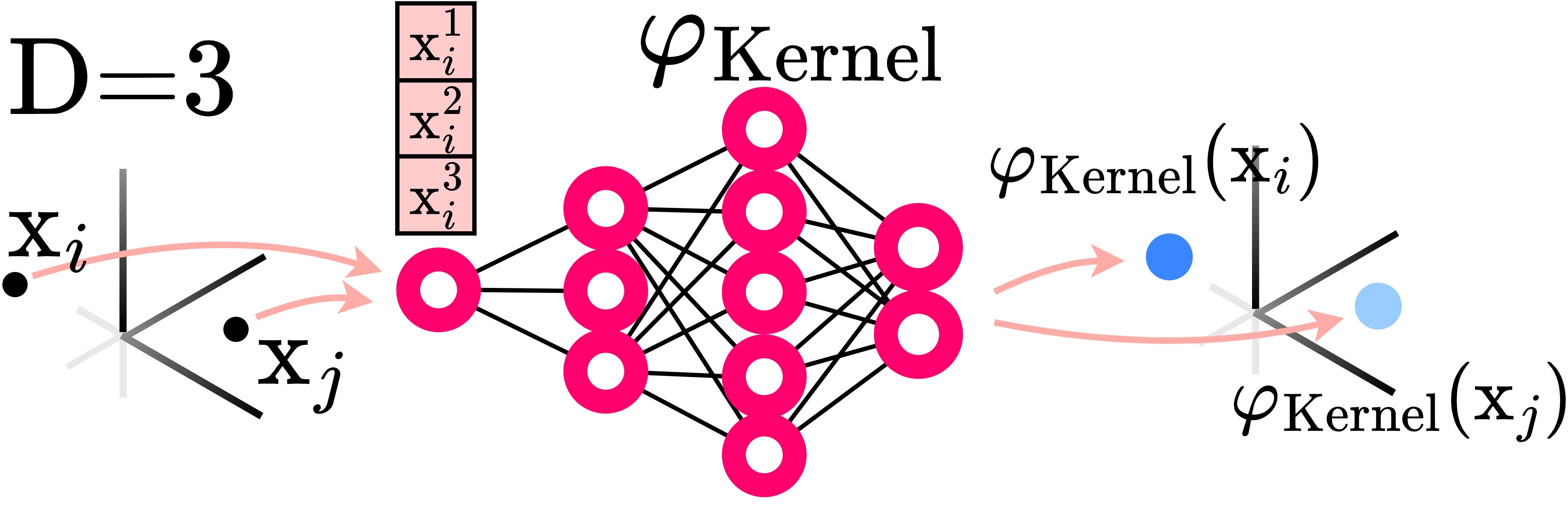}
         \caption{Kernel network.}
         \label{fig:4_ckconv_mlp}
     \end{subfigure}
     \hspace{2mm}
     \begin{subfigure}[b]{0.20\textwidth}
     \vspace{-5mm}
         \centering
         \captionsetup{justification=centering}
         \includegraphics[width=\textwidth]{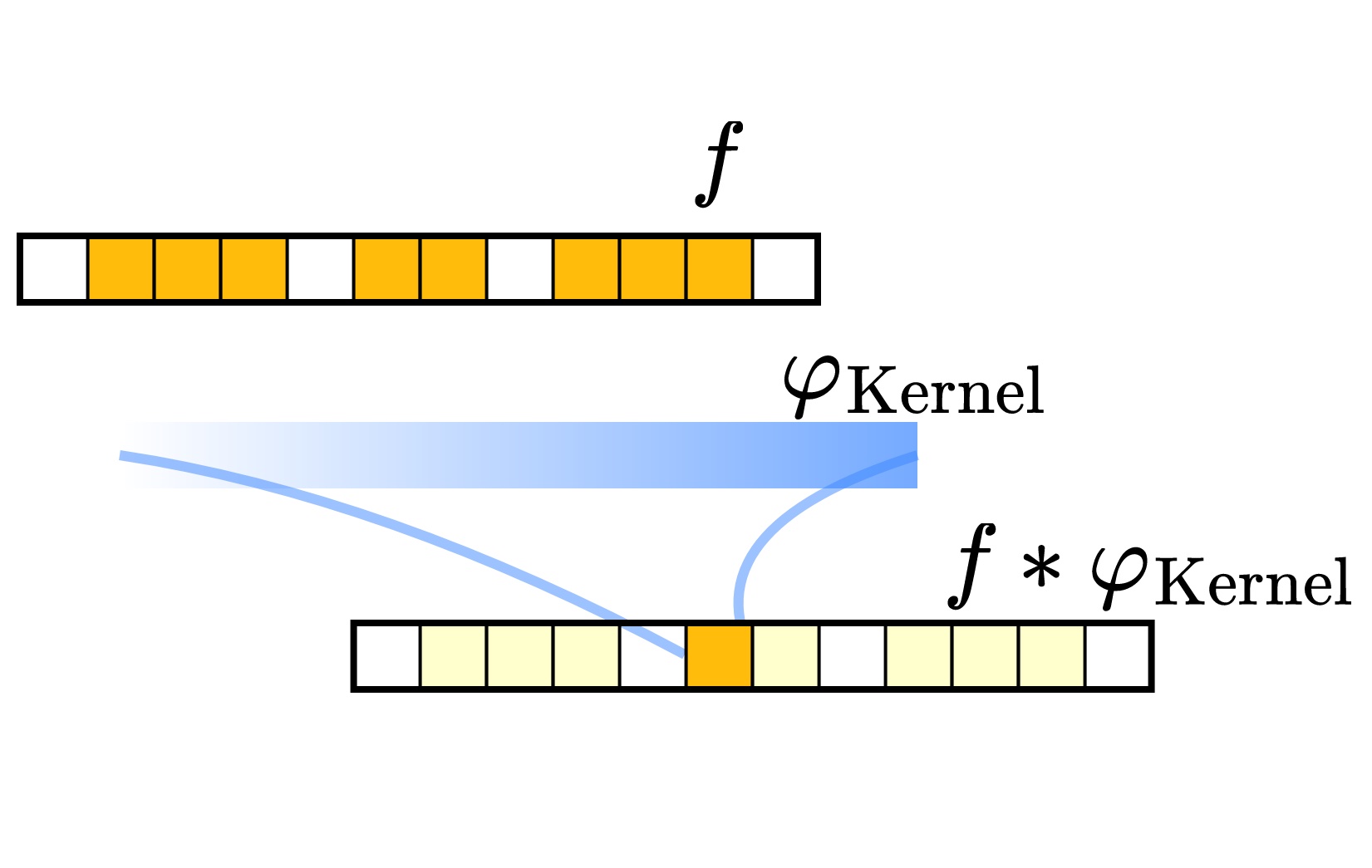}
         \caption{$\rm D{=}1$, sequences.}
         \label{fig:4_ckconv_1d}
     \end{subfigure}
     \hspace{1mm}
     \begin{subfigure}[b]{0.20\textwidth}
         \centering
         \captionsetup{justification=centering}
         \includegraphics[width=\textwidth]{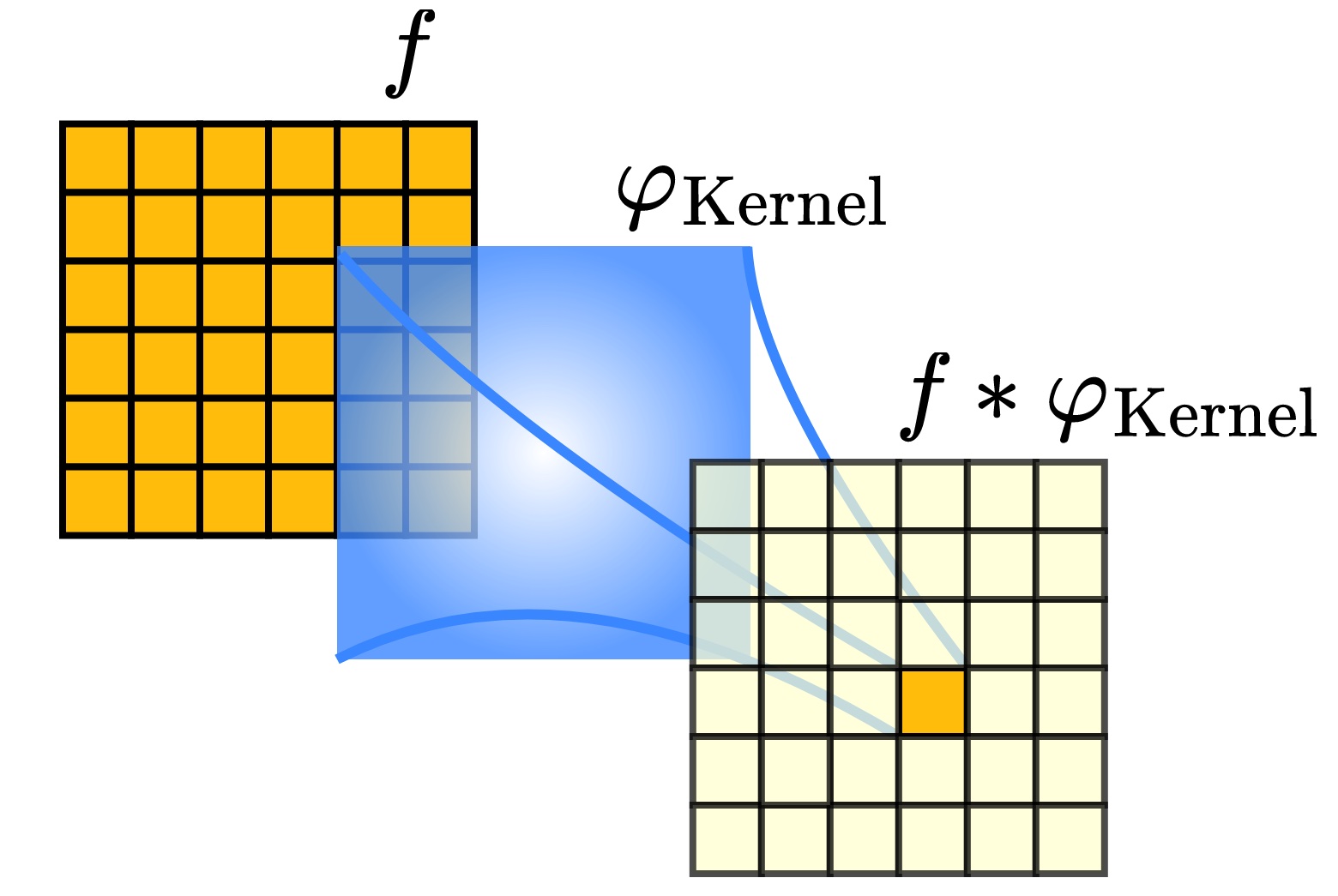}
         \caption{$\rm D{=}2$, images.}
         \label{fig:4_ckconv_2d}
     \end{subfigure}
     \hspace{1mm}
     \begin{subfigure}[b]{0.20\textwidth}
         \centering
         \captionsetup{justification=centering}
         \includegraphics[width=\textwidth]{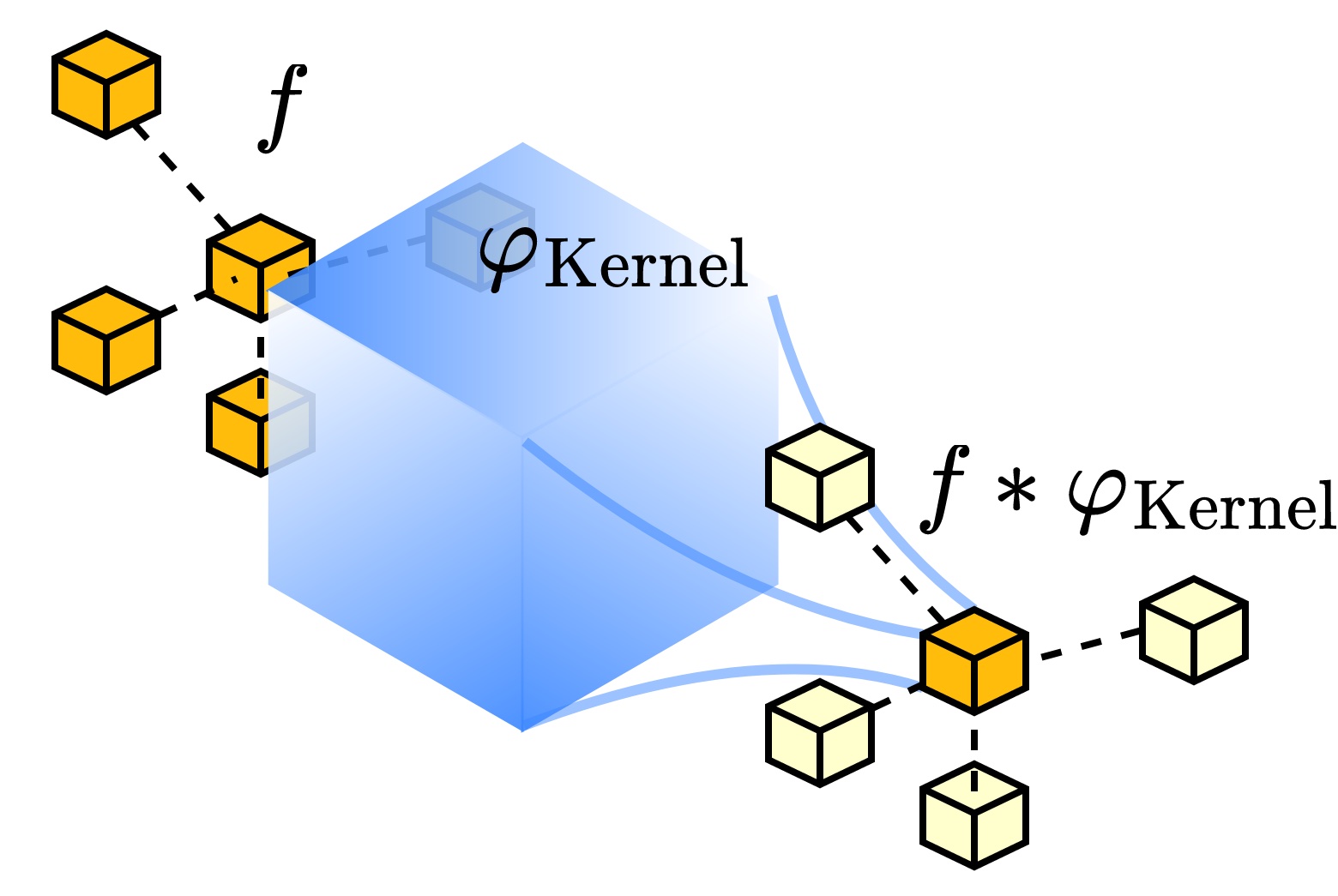}
         \caption{$\rm D{=}3$, volumes.}
         \label{fig:4_ckconv_3d}
     \end{subfigure}
    \vspace{-2mm}
    \caption{Continuous convolutional kernels: the key to a unified CNN architecture. The continuous parameterization of convolutional kernels used in this work consists of a small kernel network $\varphi_{\mathrm{kernel}}$ that receives coordinates as input and outputs the value of the convolutional kernel at that position (\ref{fig:4_ckconv_mlp}). By changing the dimensionality of the coordinates $\xv_i$, the same kernel network can render convolutional kernels for sequential (\ref{fig:4_ckconv_1d}), visual (\ref{fig:4_ckconv_2d}), and higher dimensional data (\ref{fig:4_ckconv_3d}).}
    \label{fig:4_ckconv}
\end{figure}

\textbf{The need for a continuous parameterization.} To overcome task-dependent architectures, it is crucial to define a kernel parameterization that decouples parameter count from kernel size. Following \citet{schutt2017schnet,romero2022ckconv}, we use a small neural network to define a continuous mapping from positions to the value of the kernel at those positions. The resulting \textit{Continuous Convolutional Kernels} (Fig.~\ref{fig:4_ckconv}), allow for the construction of convolutional kernels of arbitrary size in a parameter efficient manner. Consequently, the same convolutional layers --and thus the same CNN-- can be used regardless of the input length, resolution and dimensionality. We leverage this formulation to construct the \textit{Continuous Convolutional Neural Network} (CCNN): a single CNN architecture that can be applied regardless of input length, resolution or dimensionality.

\textbf{Empirical results.} To showcase the proposed CCNN, we deploy the same CCNN for several tasks on sequential (1D), visual (2D) and point-cloud (3D) data. Our CCNN matches and often outperforms the current state-of-the-art across all tasks considered. %sequential and visual tasks, and is a convincing and highly parameter-efficient method for point-cloud data. 
Importantly, the continuous parameterization of our CCNN allows it to handle irregularly sampled data natively. As a result, the CCNN is not restricted to grid data, e.g., 3D voxels, and can be used on point-clouds directly. 

\textbf{Contributions:}
\begin{itemize}
    \item We propose the \textit{Continuous Convolutional Neural Network}: a general purpose CNN architecture able to process data of arbitrary resolution, dimensionality and length without structural changes.
    \item We study the layers of CNNs, and demonstrate that the ability to model long-term dependencies on $N$D without the need of input dependent downsampling and depth values is \textit{necessary and sufficient} for the construction of a general purpose CNN architecture.
    \item In order to model long-term dependencies on $N$D without input dependent downsampling and depth values, we utilize and improve the Continuous Kernel Convolutions of \citet{romero2022ckconv}. Our proposed improvements allow the proposed Continuous CNN to achieve good empirical results on the tasks considered in 1D, 2D and 3D without structural changes.
\end{itemize}
\vspace{-7mm}
\section{Related Work}
A section with extended comparisons to related works is provided in Appx.~\ref{appx:4_extended}.

\textbf{General purpose architectures.} To the best of our knowledge, the only existing method aiming for a general purpose architecture is the Perceiver \citep{jaegle2021perceiver}, which uses a Transformer to lift restrictions regarding data characteristics and modalities. However, (\textit{i}) it must map inputs --regardless of their size-- to a small latent representation to reduce the quadratic complexity of self-attention, (\textit{ii}) decouples the depth of the network from its parameter count via recurrence, which requires tuning the number of unrolling steps per task, and (\textit{iii}) must use absolute positional encodings to encode the data structure, which break the translation equivariance of the self-attention operation \citep{romero2021group}. In contrast, CCNNs provide a general purpose architecture that: (\textit{i}) scales much more favorably than self-attention, (\textit{ii}) does not require a constant small latent representation, (\textit{iii}) does not require task-dependent depths, and (\textit{iv}) preserves translation equivariance. 

\textbf{Long range dependencies in $N\Dt$.} Based on our analysis (Sec.~\ref{sec:4_from_discr_to_cont}), any architecture able to model long range dependencies in $N\Dt$ without the need of input-dependent pooling or depth could be used as a general purpose architecture. To our best knowledge, the only existing convolutional methods able to construct global convolutional kernels are CKConvs \citep{romero2022ckconv, romero2022flexconv} and state-spaces \citep{gu2021combining, gu2022efficiently}. However, state-spaces rely on complex dynamical systems that are not easily defined in $N\Dt$ --aside from the combination of $1\Dt$ systems, equivalent to representing $N\Dt$ kernels as combinations of $N$ independent $1\Dt$ kernels-- \cite{nguyen2022s4nd}. Consequently, we select CKConvs as the building block of our approach given their advantages in terms of expressivity and simplicity.

%\textbf{Continuous convolutional kernels.} A multitude of previous works investigate a formulation of CNNs based on continuous convolutional kernels pursuing a variety of motivations. A number of these works apply specifically to pointcloud data, interpolating a set of weights to obtain a definition over the continuous input space \citep{hua2018pointwise, thomas2019kpconv}, expressing the kernels in a polynomial \citep{xu2018spidercnn} or neural network basis \citep{jia2016dynamic, schutt2017schnet, wu2019pointconv}. Other exploits desirable properties of spatially structured kernels on regular domains, for example to increase receptive fields efficiently \citep{su2021log}, learn more appropriate kernel geometries \citep{dai2017deformable}, or learn receptive field sizes \citep{tomen2021deep}. %, or implement equivariant convolutions \citep{sifre2014rigid, worrall2017harmonic, weiler20183d, weiler2019general, bekkers2019b, finzi2020generalizing}. 
%In these works, continuous kernels are used as drop-in replacement, hence the use of pooling operations in these architectures means they are still tied to data characteristics.
\textbf{Convolutional kernels as neural networks.} Modelling convolution kernels with small neural networks that map kernel positions to kernel values 
showed promising results for small kernels \citep{jia2016dynamic, schutt2017schnet, wu2019pointconv}. Subsequently, \cite{romero2022ckconv} realized that this parameterization decouples the size of the convolutional kernel from the number of parameters required to construct it, and thus can be used to construct arbitrarily large convolutional kernels in a parameter efficient manner. In this work, we show that this parameterization allows for the construction of a single CNN that can be used regardless of the input length, resolution and dimensionality.

\citet{romero2022ckconv} realized that the piece-wise MLPs used so far to parameterize convolutional kernels were unable to model complex long range dependencies due to their spectral bias \citep{tancik2020fourier}, and showed that implicit neural representations --specifically SIRENs \citep{sitzmann2020implicit}-- could be used to solve the issue. Subsequently, \citet{romero2022flexconv} parameterized their convolutional kernels with Multiplicative Anisotropic Gabor Nets (MAGNets), which provided them control over frequencies admitted in the convolutional kernel, thus preventing aliasing and aiding generalization across resolutions. In our work, we use FlexConvs parameterized by MAGNets. However, we observe that neural networks used to parameterize convolutional kernels are not correctly initialized for that purpose, and propose an initialization that solves the issue.
\vspace{-7mm}
\section{Toward Data-Independent CNN Architectures}\label{sec:4_from_discr_to_cont}

In this section, we study the components of CNN architectures and pinpoint the changes required in order to construct a CNN architecture independent of input lengths, resolutions and dimensionalities.
\vspace{-7mm}
\subsection{Pointwise Operations: Linear Layers, Dropout, Pointwise Nonlinearities and Residual Connections}

% \begin{wrapfigure}{r}{0.5\textwidth}
\begin{SCfigure}
\vspace{-2mm}
    \centering
    \begin{subfigure}[b]{0.3\textwidth}
         \centering
         \captionsetup{justification=centering}
         \includegraphics[scale=0.06]{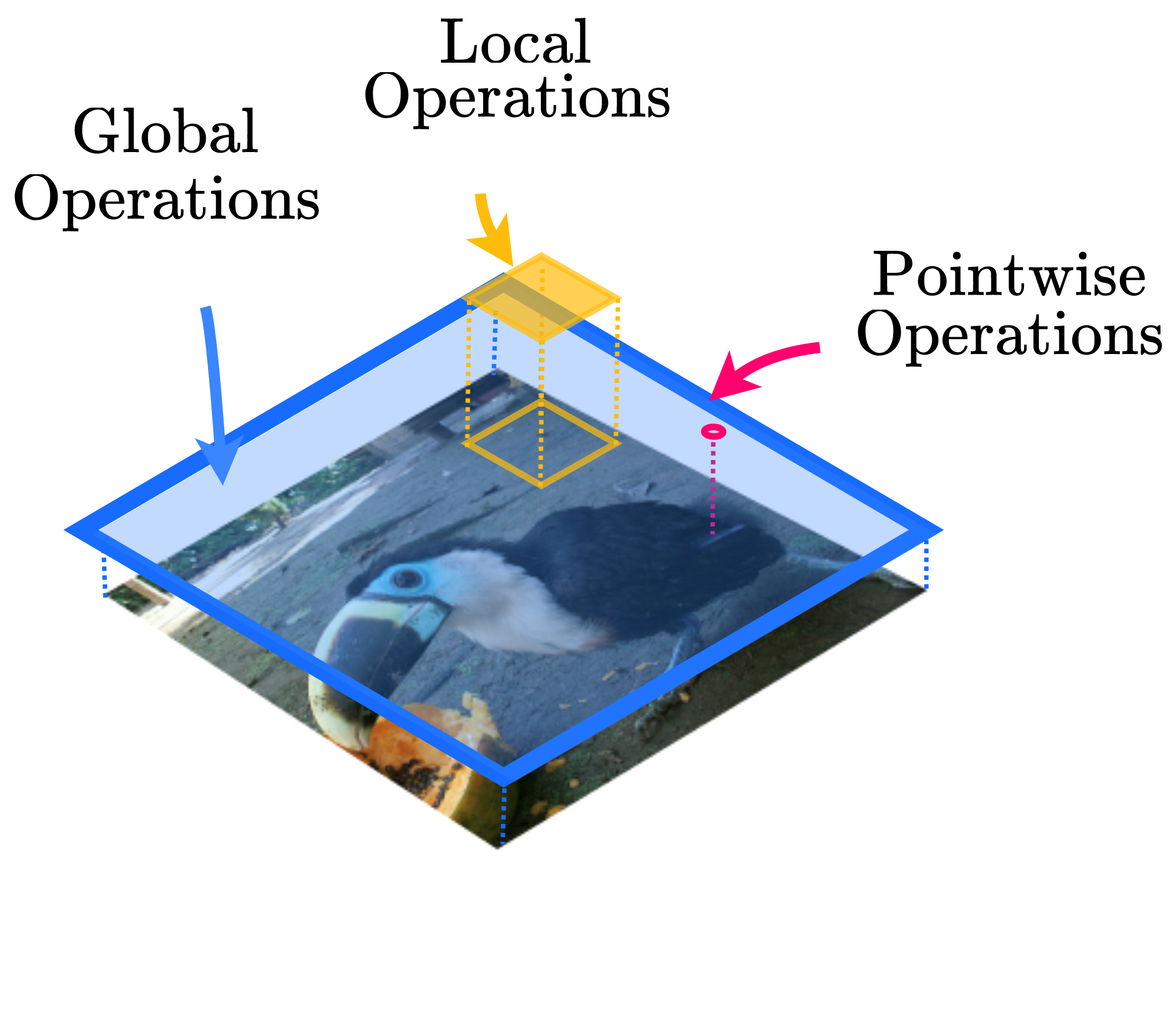}
         \vspace{-10mm}
         \caption{}
         \label{fig:4_operation-types-big}
     \end{subfigure}
     % \hfill
     \begin{subfigure}[b]{0.3\textwidth}
         \centering
         \captionsetup{justification=centering}
        \includegraphics[scale=0.07]{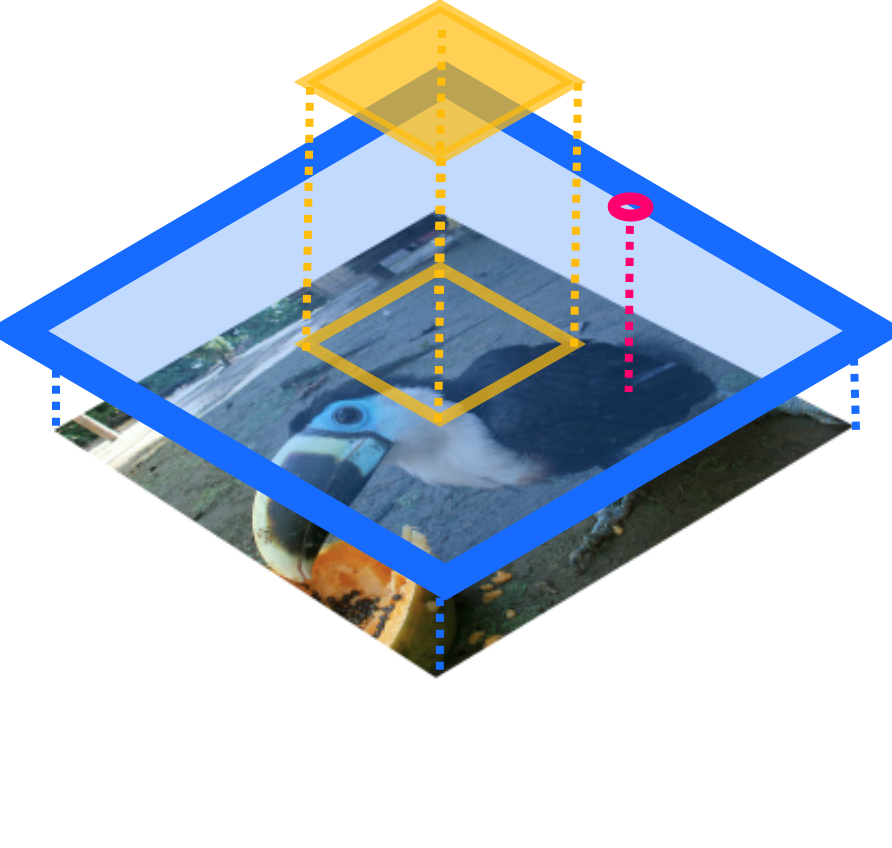}
        \vspace{-5mm}
         \caption{}
         \label{fig:4_operation-types-smaller}
     \end{subfigure}
     \vspace{-3mm}
    \caption{Operation types: global, pointwise and local. Local operations are resolution dependent. Transferring a local operation from (\ref{fig:4_operation-types-big}) to a lower resolution (\ref{fig:4_operation-types-smaller}) leads to an increased receptive field.}
    \label{fig:4_operation-types}
% \end{wrapfigure}
\end{SCfigure}
Pointwise operations are operations applied to each spatial element of the input separately (Fig.~\ref{fig:4_operation-types}), e.g., pointwise linear layers, dropout and pointwise nonlinearities. As such pointwise operations do not depend on the input shape and model the same function regardless of input length, resolution and dimensionality. Learnable parameters of pointwise operations, e.g., in pointwise linear layers and PReLU \citep{he2015delving}, are shared over the spatial domain; the same set of parameters is applied to inputs of any spatial shape.

\textbf{Conclusion.} Based on the previous observations, we can conclude that pointwise operations can be used without changes within a general-purpose CNN architecture. 
\vspace{-7mm}
\subsection{Global Operations: Normalization Layers and Global Pooling}
Global operations aggregate all spatial elements of the input for their processing, e.g., normalization layers, global pooling. As such, common global operations do not depend on the specific shape of the input signal and their effect is equivalent regardless of the input length, resolution and dimensionality. It is important to note, however, that common global CNN operations only define learnable parameters along the channel axes e.g., the scale and mean parameters of a normalization layer have shape $\wv \in \sR^{\mathrm{N_{in}}}$. Nevertheless, if one were to define a global operation for which spatial positions are also assigned weights, the shape of the parameters would be dependent on the input length, resolution and dimensionality, and thus the previous statement would no longer hold.

\textbf{Conclusion.} The previous observations indicate that global operations that only define channel-wise learnable parameters can be used without changes in order to construct a unified CNN architecture. Luckily, this is the case for all common global operations used in CNN architectures.\footnote{Global discrete convolutional kernels can be interpreted as global operations for which parameters along the spatial dimensions of the input are defined. Therefore, these require special treatment  (Sec.~\ref{sec:4_local_operations}).}
\vspace{-7mm}
\subsection{Local Operations: Convolutional Layers and Subsampling}\label{sec:4_local_operations}

Local operations are operations that rely on spatial portions of the input and are applied across its spatial dimensions, e.g., convolution, subsampling. In practice, the neighborhoods on which these operations are applied are hyperparameters, e.g., $3\times3$ kernels or pooling over $2\times2$ regions, and are selected based on the input size. Unfortunately, if the resolution of the input changes then the portion of the input that the operation considers changes and the effect of the operation changes  (Fig.~\ref{fig:4_operation-types}). Similarly, if the length of the input changes, then larger neighborhoods are required in order to model dependencies across the input. These effects exacerbate if one considers multi-dimensional inputs, as different dimensions might require different neighborhood sizes. Consequently, we can say that local operations \textit{depend on the resolution, length and dimensionality of the input}. 

A possible solution would be to adjust the size of the operations proportional to resolution and length changes. Unfortunately, if the local operation defines learnable parameters over its spatial dimensions --as in (discrete) convolutional kernels (Sec.~\ref{sec:4_independent_conv_layers})--, then increasing the size of the convolutional kernel is tied to a proportional increase in the number of parameters required to construct it. This, in turn changes the learning dynamics of the network and easily becomes prohibitive.

\textbf{The need for a continuous parameterization.} A better solution results from using a parameterization in which the number of parameters is \textit{independent} from the size of the kernel. By doing so, the same parameterization can be used independently from the input length, resolution and dimensionality.
\vspace{-7mm}
\subsubsection{From Data-Dependent to Data-Independent Convolutional Layers}\label{sec:4_independent_conv_layers}

Conventional CNNs implement a discrete version of the convolution operation
\begin{equation}
\label{eq:4_convolution-cont}
(f * k)^o (\xv) = 
    \int_{\mathbb{R}^\Dt} f(\xv-\tilde{\xv})k^o (\tilde{\xv}) \text{d}\tilde{\xv}, \quad o \in \left[1,...,{\rm N_{out}}\right),
\end{equation}
where $f:\mathbb{R}^\Dt {\rightarrow}\mathbb{R}^{\rm N_{in}}$ is a $\Dt-$dimensional input signal with $\rm N_{in}$ channels, and $k:\mathbb{R}^\Dt {\rightarrow} \mathbb{R}^{\rm N_{in} \times N_{out}} $ is a set of $\rm N_{out}$ convolutional kernels. For the types of data CNNs are commonly used for, e.g. images, the signal $f$ is generally sampled on a discrete grid of equidistant points, and thus it can be described as a function $f:\mathbb{Z}^\Dt \rightarrow \mathbb{R}^{\rm N_{in}}$. In practice, the signal $f$ is non-zero only on a finite subset of the grid $\Omega(f) \subset \mathbb{Z}^\Dt$ with limits given by the range of sampling, e.g., height and width of an image. Accordingly, the convolutional kernel $k$ is defined over the same grid of coordinates, of which generally a subset $\Omega(k) \subset \Omega(f)$ maps to nonzero values. Since both $\Omega(k)$ and $\Omega(f)$ are discrete and finite, the convolution is computed as the inner product of function and kernel values at each position:
\begin{equation}
\label{eq:4_convolution}
(f * k)^o (\xv) = 
    \sum_{\tilde{\xv} \in \Omega(k)} f(\xv - \tilde{\xv})k^o(\tilde{\xv}), \quad \xv \in \Omega(f).
\end{equation}
\textbf{Discrete convolutional kernels tie convolutional layers to data characteristics.} Conventionally, the convolutional kernel $k$ is implemented through a discrete set of randomly initialized weights $\Wm$, of which each entry $\wv_i \in \mathbb{R}^{\rm N_{in} \times N_{out}}$ corresponds to a point $\xv_i \in \Omega(k)$. Consequently, an increased kernel size is reflected in a larger set $\Omega(k)$, and thus in a correspondingly larger weight matrix $\Wm$. This directly ties a model's parameter count to its kernel size.

In order to model the long-range dependencies needed to extract high-level features in a parameter-efficient way, we must then resort to pooling operations and the stacking of layers that implicitly increase the receptive field of the kernel. This in turn, makes CNNs effective only on inputs of a certain size. For example, if we apply a network created to model long-range dependencies over images of size $256{\times}256$ to images of size $32{\times}32$, then intermediary pooling layers would make the spatial extent of the feature maps collapse long before all convolutional layers are applied. Similarly, if we use a network designed to model long-range dependencies on $32{\times}32$ on images of size $256\times256$, the network will not be able to model long-range dependencies in the input.

Additionally, the definition of the kernel $k$ through a discrete set of weights $\mat{W}$ ties the model to a given input resolution. Yet, for many applications, the input $f$ is a discretization of an underlying continuous function. Therefore, we would like our model to provide the same responses regardless of the resolution at which $f$ is provided. As discrete convolutional kernels live in a discrete domain, they cannot be easily represented at other resolutions. In fact, one can show that discrete CNNs do not generalize to unseen resolutions \cite{romero2022flexconv, nguyen2022s4nd}. Consequently, it is not possible to reliably apply trained discrete CNNs across resolutions.

In addition, note that in Eq.~\ref{eq:4_convolution}, the same discrete convolutional kernel $k$ can be used at every location $\xv \in \Omega(f)$ only because the input signal $f$ is defined over an equidistant grid, and the values of the discrete kernel $k(\tilde{\xv})$ align with the features $f(\xv), \forall \xv \in \Omega(f)$. This is not the case for irregular data. Consequently, discrete kernels are ill-suited to handle irregular data. With weights fixed to relative positions, an infinite number of weights would be needed to cover any continuous domain.

These limitations suggest a better approach to model convolutional kernels: \textit{using the model weights to parameterize $k$ as a continuous function over the data domain $\mathbb{R}^d$}.

\begin{figure}
    \centering
     \begin{subfigure}{0.27\textwidth}
         \centering
         \captionsetup{justification=centering}
         \includegraphics[scale=0.055]{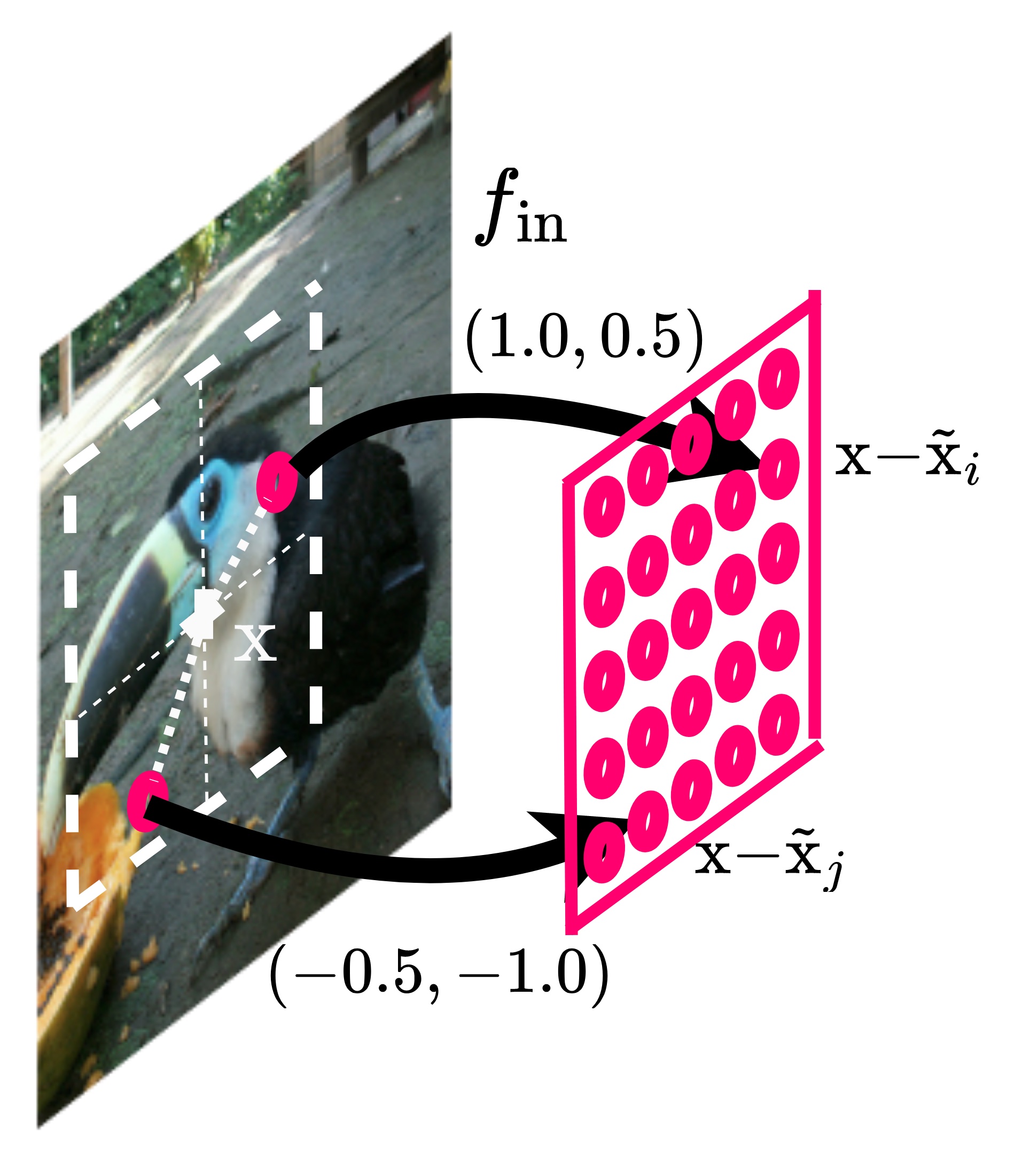}
         \vspace{-4mm}
         \caption{}
         \label{fig:4_grid}
     \end{subfigure}
     \hspace{1mm}
     \begin{subfigure}{0.45\textwidth}
         \centering
         \captionsetup{justification=centering}
         \includegraphics[scale=0.055]{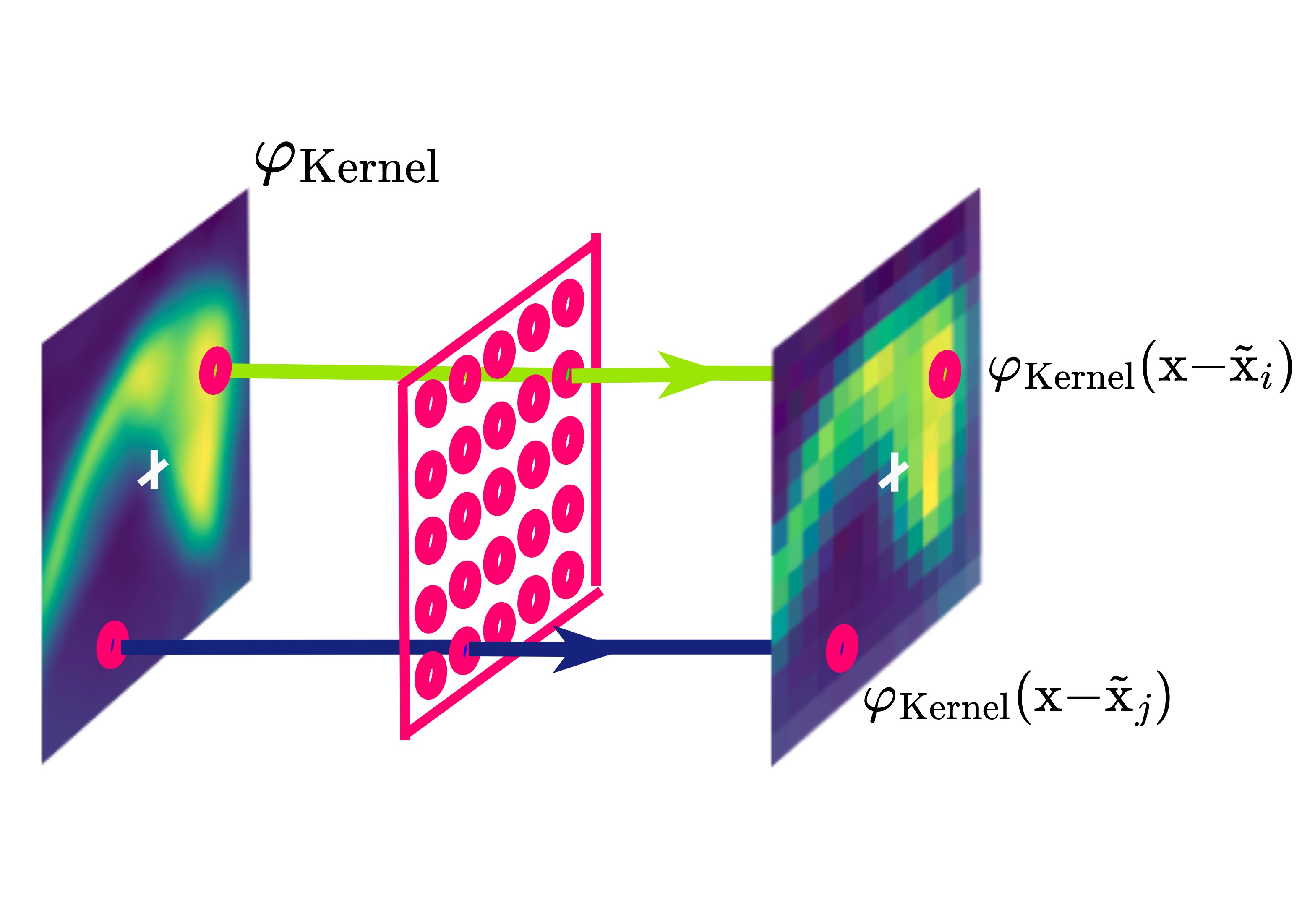}
         \vspace{-4mm}
         \caption{}
         \label{fig:4_sample}
     \end{subfigure}
     \hspace{1mm}
     \begin{subfigure}{0.2\textwidth}
         \centering
         \captionsetup{justification=centering}
         \includegraphics[scale=0.055]{ccnn-application.jpg}
         \vspace{-4mm}
         \caption{}
         \label{fig:4_apply}
     \end{subfigure}
     \vspace{-2mm}
    \caption{Applying a Continuous Kernel Convolution. Given a pixel-position in the input image $\xv \in \Omega({f})$, we obtain relative offsets to surrounding pixels $\{\xv-\tilde{\xv}\}_{\tilde{\xv} \in  \Omega({f})}$ (\ref{fig:4_grid}). Next, we pass each relative position to the kernel generator network in order to generate the kernel:  $\{\varphi_{\rm Kernel}(\xv-\tilde{\xv})\}_{\mathbf{\tilde{\xv}}\in \Omega(f)}$ (\ref{fig:4_sample}). Finally, we apply the convolution between the generated convolutional kernel and the input (\ref{fig:4_apply}).\label{fig:4_applying-continuous-kernels}}
\end{figure}
\textbf{A data independent parameterization.}
To obtain a formulation for a CNN applicable to arbitrary resolutions and sizes, we require a parameterization for convolutional layers that is invariant to the set $\Omega(f)$ over which $f$ is sampled. In other words, we must find a parameterization with which the kernel can be modelled over the underlying continuous domain of the input signal, i.e., $\mathbb{R}^d$. Moreover, to avoid models with different parameter count for different resolutions, it is necessary that the parameterization of the kernel decouples its parameter count from the size of the kernel.

Such a parameterization is provided by \textit{Continuous Kernel Convolutions} (CKConvs) \citep{romero2022ckconv, romero2022flexconv}. CKConvs provide a continuous parameterization for convolutional kernels by using a small neural network $\varphi_{\rm Kernel}: \sR^{\Dt} \rightarrow \sR^{\mathrm{N_{out}} \times \mathrm{N_{in}}}$ as a kernel generator network. This network maps coordinates in the domain of the kernel $\xv_ i \in \sR^{\Dt}$ to the values of the convolutional kernel at that position: $k(\xv) \in \mathbb{R}^{\rm N_{out} \times N_{in}}$ (Fig. \ref{fig:4_continuous-kernels}). A Continuous Kernel Convolution is illustrated in Fig. \ref{fig:4_applying-continuous-kernels}.

Since the parameter count of the kernel generator network is independent from the number of points in the neighborhood that determines the size of the kernel, CKConvs allow for construction of arbitrarily large kernels without increasing the parameter count of the layer. \cite{romero2022ckconv} shows that large kernels allow CNNs to model long range spatial dependencies at every layer, thus removing the need for downsampling and stacking of layers to increase receptive fields. This in turn allows us to build an architecture which does not contain resolution-, dimensionality-, and size-dependent layers. 

\textbf{Conclusion.} The previous observations indicate that local operations equipped with existing parameterizations are tied to the length, resolution and dimensionality of the input. Nevertheless, this limitation can be lifted if an alternative parameterization is used that detaches the neighborhood of action of the operation from the length, resolution and dimensionality of the input.
\vspace{-7mm}
\section{A General Purpose CNN Architecture}

\begin{wrapfigure}{r}{0.42\textwidth}
\vspace{-4mm}
         \centering
         \includegraphics[width=0.35\textwidth]{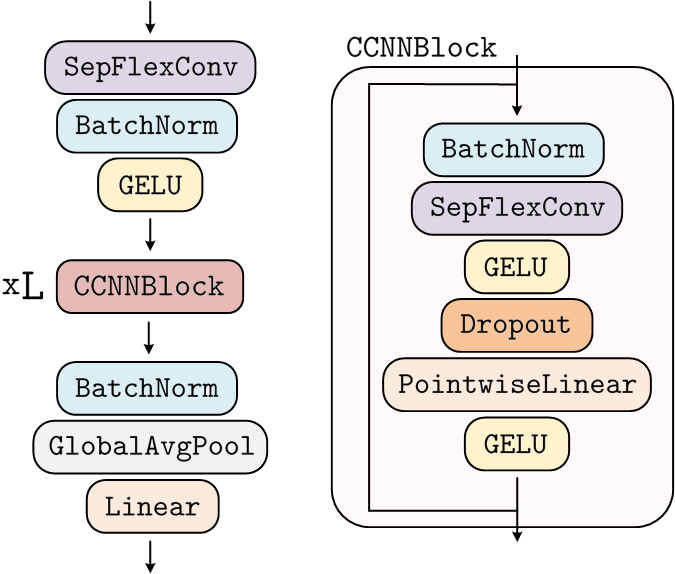}
    \vspace{-2mm}
    \caption{The CCNN architecture.\label{fig:4_ccnn}
    \vspace{-4mm}}
\end{wrapfigure}
\textbf{Starting point.} In principle, the CNN architectures with CKConv \citep{romero2022ckconv} and FlexConv \citep{romero2022flexconv} introduced previously fulfill the requirements posited in Sec.~\ref{sec:4_from_discr_to_cont}. Nevertheless, as depicted in these papers, these architectures still must be tailored to specific applications and domains in order to perform well, e.g., TCN \citep{bai2018empirical} and ResNet \citep{he2016deep} backbones for sequential and visual tasks, respectively in \citet{romero2022flexconv}.

In order to construct a single performant general purpose architecture that works well across all tasks considered, we start from a FlexNet \citep{romero2022flexconv}, and propose several structural changes (Sec.~\ref{sec:4_modifications}). The resulting CCNN architecture is shown in Fig.~\ref{fig:4_ccnn}.

\vspace{-7mm}
\subsection{Modifications and Improvements}\label{sec:4_modifications}

\textbf{Proper initialization of the kernel generator network $\varphi_{\rm Kernel}$.} First, we analize the kernel generator network $\varphi_{\rm Kernel}$, and observe that it is not initialized properly in previous works \citep{schutt2017schnet, wu2019pointconv, romero2022ckconv, romero2022flexconv} for the purpose of parameterizing convolutional kernels. 

Recall that, in order to ensure training stability it is desirable to retain a constant (unitary) variance throughout the activations of a neural network \citep{glorot2010understanding}. Hence, the discrete weights $\mat{W}$ conventionally used to construct a convolutional kernel are initialized to have variance inversely proportional to the number of elements over which the convolution is computed, i.e., the number of pixels $|\Omega(k)|$ times the number of channels ${\rm N_{in}}$. For instance, He initialization \citep{he2015delving} initializes $\mat{W}$ s.t. $\mathrm{Var}(\mat{W}){=}
g^{2}/({\rm N_{in}} |\Omega(k)|)$,
with a gain $g$ that depends on the nonlinearity used.

\begin{figure}
    \centering
     \begin{subfigure}[valign=T]{0.35\textwidth}
         \centering
         \includegraphics[width=\textwidth]{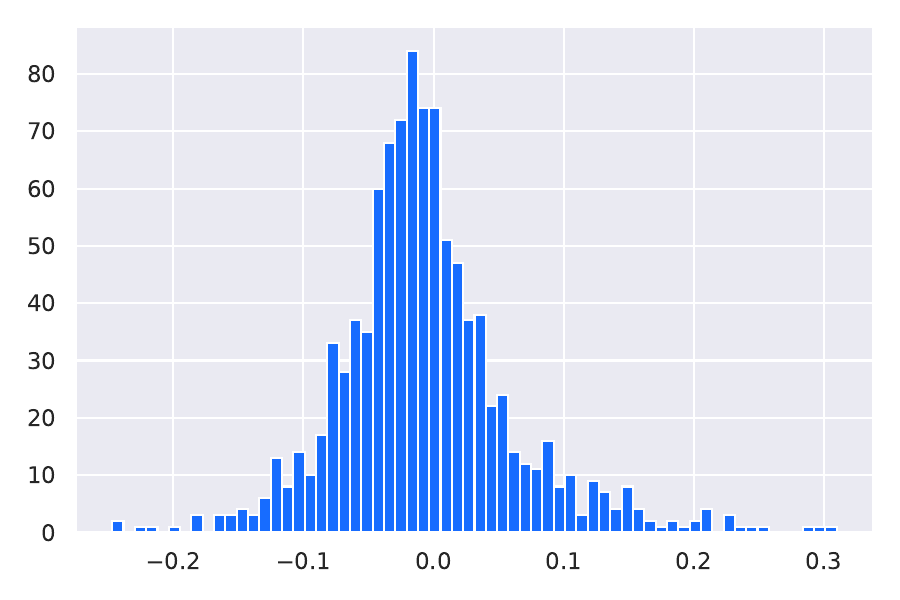}
     \end{subfigure}
     \begin{subfigure}[valign=T]{0.36\textwidth}
         \centering
         \includegraphics[width=\textwidth]{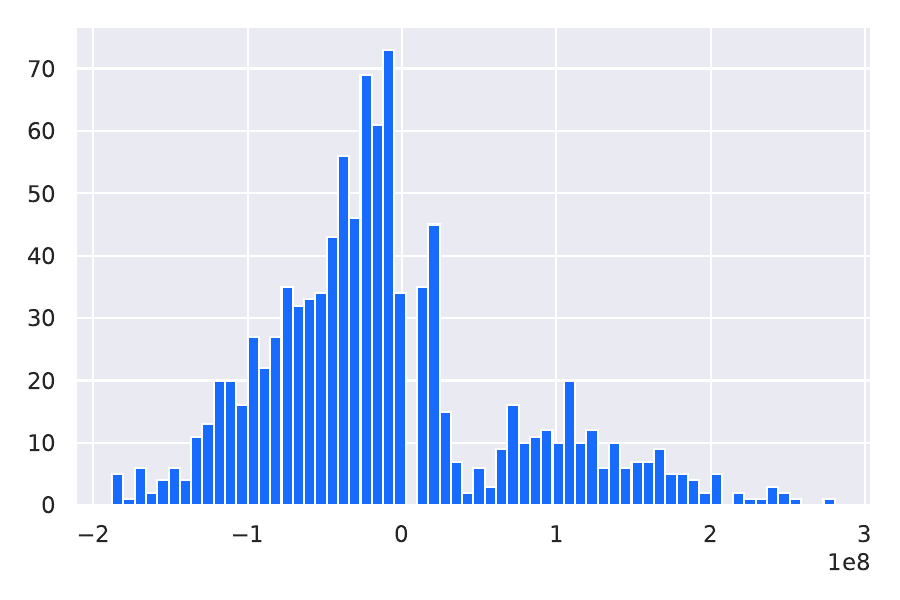}
     \end{subfigure}
     \vspace{-2mm}
    \caption{Histogram of the output of a CCNN with and without our proposed kernel initialization.\label{fig:4_initialization}}
\end{figure}
In related works, the networks used to parameterize convolutional kernels are themselves initialized to \textit{preserve a constant (unitary) variance throughout the network}. Consequently, when used as a kernel generator network, standard initializations lead the generated kernel to have unitary variance, i.e., $\mathrm{Var}(k){=}1$. This in turn, make CNNs using neural networks to parameterize their convolutional kernels experience a layer-wise growth in the variance of the feature maps proportional to $\Nin \cdot |\Omega(k)|$. This growth is of particular importance for kernel generator networks that generate large convolutional kernels, i.e., with large $|\Omega(k)|$. For instance, we observe that the logits of CNNs with CKConvs \citep{romero2022ckconv} and FlexConvs \citep{romero2022flexconv} lie in the order of $1\mathrm{e}^{19}$ on initialization: an undersirable property that might lead to unstable training and a need for low learning rates.

To address this issue, we must ensure that the variance at the output of the kernel generator network is inversely proportional to $\Nin \cdot |\Omega(k)|$. Inspired by \citet{Chang2020Principled}, we therefore re-weight the last linear layer of the kernel generator network $\varphi_{\rm Kernel}$ by $g^2/\sqrt{ \Nin \cdot |\Omega(k)|}$. With this modification, we observe that the variance of the generated convolutional kernels satisfies the desired constrains, and consequently, the logits of our CCNN show unitary variance upon initialization (Fig.~\ref{fig:4_initialization}).

\textbf{Depthwise Separable Continuous Convolutions.} Separable convolutions have long been used to improve the parameter and computational efficiency of CNNs \citep{rigamonti2013learning, sifre2014rigid}. Recent architectures have leveraged separability, and found CNNs with separable kernels to perform better than CNNs with conventional convolutions \citep{knigge2022exploiting, liu2022convnet}. This phenomenon results from the separation of spatial and channel dimensions, which allows for wider networks without additional computational and parameter complexity. 

Based on these observations, we construct a depth-wise separable version of FlexConv \citep{romero2022flexconv}, in which a channel-wise convolution is computed with a kernel generated by a kernel generator network $\varphi_{\rm Kernel}: \sR^{\Dt} \rightarrow \sR^{\mathrm{N_{in}}}$, followed by a pointwise linear layer from $\mathrm{N_{in}}$ to $\mathrm{N_{out}}$ dimensions. Separable FlexConvs allow  constructing a much wider CCNN --from 30 to 140 hidden channels-- with the same parameter and computation complexity.

\textbf{An improved residual block.} Residual connections \citep{he2016deep} provide training stability and improved performance. A residual block $\mathcal{R}(f) {=} \psi(f) + f$ is defined as the sum between the input and a so-called \textit{residual connection} $\psi$ composed of multiple layers: convolutional, normalization, etc.

\begin{figure}
    \centering
    \captionsetup{justification=centering}
     \includegraphics[width=0.65\textwidth]{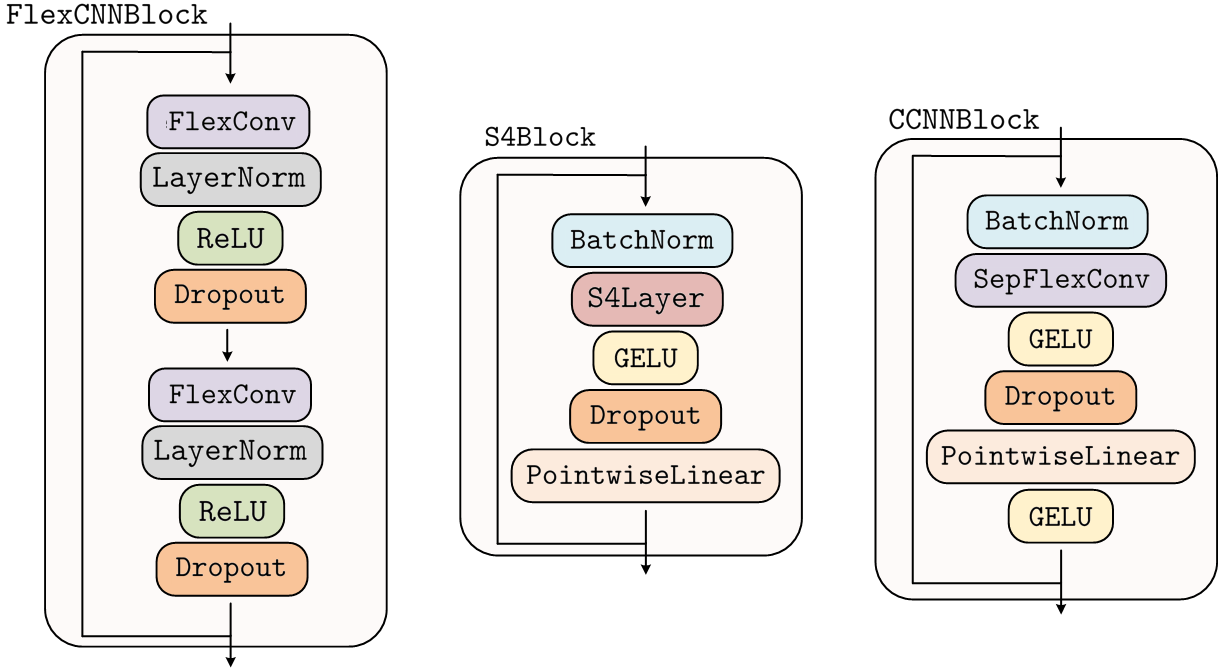}
     \vspace{-2mm}
     \caption{Residual blocks used in FlexCCNNs \citep{romero2022flexconv}, S4 \citep{gu2022efficiently} and CCNNs (ours).}
    \label{fig:4_block_comparison}
\end{figure}
Recent studies found improvements over the residual block of \citet{he2016deep} by changing the nonlinearities as well as the position and type of normalization layers within the blocks \citep{xiong2020layer, liu2022convnet}. Based on these advances and empirical evidence, we modify the residual blocks of \citet{romero2022flexconv} with residual blocks similar to those of S4 \citep{gu2022efficiently} complemented with a nonlinearity at the end of the block. A comparison of our residual block and those in \cite{gu2022efficiently} and \cite{romero2022flexconv} is given in Fig.~\ref{fig:4_block_comparison}.

\textbf{$L_2$ regularization of continuous kernels.} Weight decay penalizes the $L_2$ norm of learned kernel values $k$ by adding the norm of the weights $\mat{W}$ as an additional loss term \citep{krogh1991simple}. Since continuous kernels are not directly parameterized by weights $\mat{W}$, applying $L_2$ regularization on the parameters of $\varphi_{\rm Kernel}$ directly would not have the intended effect. Consequently, in order to produce the same effect, we amend the $L_2$ regularization term to penalize the \textit{generated convolution kernel} instead. Given $\varphi^{:,l}_{{\rm Kernel}}$ the set of generated kernel for all channels at a layer $l$, $\mathcal{L}_{\rm obj}$ the objective loss function, and $\lambda$ the regularizing parameter, we define a regularized loss as:
\begin{equation}
    \label{eq:4_decay}
    \mathcal{L}=\mathcal{L}_{\rm obj} + \mathcal{L}_{\rm reg} = \mathcal{L}_{\rm obj} +  \lambda \cdot \frac{1}{2}  \sum_{l=1}^L ||\varphi^{:,l}_{\rm Kernel}||^2.
\end{equation}

\vspace{-9mm}
\section{Experiments}\label{sec:4_experiments}

We aim to define an architecture that can be applied regardless of the specific data characteristics. To this end, we construct a CCNN and validate it on sequential ($1\Dt$), visual ($2\Dt$) and point-cloud ($3\Dt$) datasets --see Appx. \ref{appx:4_dataset_description} for a detailed description of each dataset--. We show that the \textit{same} CCNN obtains state-of-the-art results on several sequential tasks, competitive performance on image tasks, and surpasses the Perceiver on point-cloud processing. In addition, we show that learned CCNNs generalize from regular to irregular data by transferring a trained CCNN from voxels to point-clouds.

\textbf{Architecture specifications.} To study the scalability of our method, we create two CCNNs with differing numbers of hidden layers sizes: CCNN$_{4, 140}$ (4 blocks, 140 channels, 200{\sc{k}} params) and CCNN$_{6, 380}$ (6 blocks, 380 channels, 2{\sc{m}} params). The kernel network $\varphi_{\rm Kernel}$ is chosen to be a 3-layer MAGNet \citep{romero2022flexconv}, with 32 hidden units for CCNN$_{4,140}$, and 64 hidden units for the larger CCNN$_{6, 380}$. In each task, the input dimension of the kernel network $\varphi_{\rm Kernel}$ corresponds to the dimensionality of the data --1 for sequences, 2 for images, and 3 for point-clouds--. %Since we are using depthwise separable convolutions, the kernel network maps to a set of kernel values $\mathbb{R}^{\rm N_{in}}$. 
Additional details regarding hyperparameters, training regimes and experimental settings are given in Appx. \ref{appx:4_empirical_details}. An empirical assessment of the computational efficiency of our architectures is given in Appx \ref{appx:4_inferencetime}.

\textbf{CCNNs on sequential datasets.} In $1\mathrm{D}$ we consider Sequential MNIST, Permuted MNIST \citep{lecun1998gradient}, Sequential CIFAR10 \citep{krizhevsky2009learning}, the Long Range Arena \citep{tay2021long} and the Speech Commands dataset \citep{warden2018speech}. Our results (Tabs.~\ref{tab:4_experimental_results},~\ref{tab:4_lra_results}), demonstrate that CCNNs obtain state-of-the-art across several tasks, surpassing the performance of tailored architectures such as Recurrent Neural Networks and Transformers. The performance of CCNNs is explained by their ability to model long term dependencies with higher capabilities than tailored models. Interestingly, we observe that on tasks defined over flattened 2-dimensional signals, e.g., sMNIST, sCIFAR10, CCNNs learn to construct very large periodic kernels that model flattened local dependencies in 2D (Fig.~\ref{fig:4_learned_kernels}). This showcases the ability of CCNNs to model meaningful long range dependencies.

Additionally, we evaluate the capacity of CCNNs to classify speech \citep{warden2018speech} both from prepossessed (Speech-MFCC) and raw signals (Speech-Raw). Our results (Tab.~\ref{tab:4_experimental_results}) demonstrate state-of-the-art results on both preprocessed and raw signals --of length 16000--, thus demonstrating the ability of CCNNs to process very heterogeneous data types with a unified architecture.

\underline{\textit{Generalization across resolutions.}} CCNNs are defined on a continuous space. Consequently, it is possible to train a CCNN at one resolution and deploy it at other resolutions --a feat not achievable with discrete models \citep{romero2022flexconv, nguyen2022s4nd}--. To showcase this ability, we train CCNNs on Speech-Raw, and test them on a subsampled version of the dataset (Speech-0.5x). CCNNs not only generalize, but outperform existing models on zero-shot prediction over resolution changes.

\textbf{CCNNs on image datasets.} In $2\Dt$, we consider the CIFAR10, CIFAR100 \citep{krizhevsky2009learning} and the STL10 datasets \citep{coates2011analysis}. CCNNs outperform existing continuous convolutional vision architectures \citep{ruthotto2020deep, tomen2021deep, romero2022ckconv, romero2022flexconv} and are competitive with existing large scale discrete models, while being (much) smaller (Tab.~\ref{tab:4_experimental_results}).
\begin{table}
\begin{minipage}{\textwidth}
    \begin{center}
    \caption{Experimental results on sequence, image and point-cloud datasets. {\footnotesize $\times$ -  \textit{unable to apply, either due to the model's inability to handle the specified data, or to computational complexity.}}}
    \label{tab:4_experimental_results}
    \vspace{-3mm}
    \begin{small}
    \scalebox{0.55}{
    \begin{tabular}{ccccccccccccc}
    \toprule
    &&\multicolumn{6}{c}{\sc{Sequence}} & \multicolumn{3}{c}{\sc{Image}} & {\sc{Point-cloud}} \\
    \cmidrule(lr){3-8}\cmidrule(lr){9-11}\cmidrule(lr){12-12}
    & \sc{Size}  & \sc{sMNIST} & \sc{pMNIST} & \sc{sCIFAR10} & \sc{Speech-MFCC}& \sc{Speech-Raw} & \sc{Speech-0.5x} & CIFAR10 & CIFAR100 & STL10 & ModelNet40 \\
     \midrule
      % LSTM  &70\sc{k}& 87.2 & 85.7  & -\\
     % GRU &70\sc{k}&   96.2 & 87.3  & - \\
     % IndRNN  & 83\sc{k} & 99.0 & 96.0& -  \\
     % DilRNN  & 44\sc{k} & 98.0 & 96.1 & - \\
     % HiPPO-RNN & 500\sc{k} & - & 98.30 & - \\
      % r-LSTM  & 500\sc{k} & 98.4 & 95.2 & 72.2 \\
     %TCN &70\sc{k}& 99.0 & 97.2  & - \\
      %TrellisNet & 8\sc{m} & 99.20 & 98.13 & 73.42\\
     Transformer &500\sc{k} & 98.90 & 97.90 & 62.20 & 90.75 & $\times$ & $\times$ & $\times$ & $\times$ & $\times$ & $\times$ \\
    LSSL & 7.8\sc{m} & 99.53 & 98.76 & 84.65 & $\times$ & $\times$ & $\times$ & $\times$ & $\times$ & $\times$ & $\times$ \\
    S4 &  7.8\sc{m} & 99.63 & 98.70 & 91.13 & $\times$ & $\times$ & $\times$ & $\times$ & $\times$ & $\times$ &$\times$\\
    LSSL & 300k & N.A. & N.A. & N.A. & 93.58 & $\times$ & $\times$ & $\times$ & $\times$ & $\times$ & $\times$ \\
    S4 &  300k  & N.A. & N.A. & N.A. & 93.96 & 98.32 & 96.30  & $\times$ & $\times$ & $\times$ & $\times$ \\
    CKCNN-Seq & 98\sc{k} & 99.31 & 98.00 & 62.25 & 95.30 & 71.66 & 65.96  & $\times$ & $\times$ & $\times$ & $\times$ \\
    CKCNN-Seq-Big & 1\sc{m} & 99.32 & 98.54 & 63.74  & $\times$ & $\times$ & $\times$ & $\times$  & $\times$ & $\times$ & $\times$ \\
    FlexTCN-6 & 375\sc{k} & 99.62 & 98.63 & 80.82 & 97.67 & 91.73 & $\times$ & $\times$ & $\times$ & $\times$ & $\times$\\
    %CNN$_{4, 110}$ & \sc{k} & - & - &  \\
    %CNN$_{6, 380}$ & \sc{m} & - & - &   \\
    ResNet-44 & 660\sc{k} & $\times$ & $\times$ & $\times$ & $\times$ & $\times$ & $\times$ &  92.90 & 71.15 & NA & $\times$ \\
    ResNet-18 & 11.2\sc{M} & $\times$ & $\times$ & $\times$ & $\times$ & $\times$ & $\times$ & 94.92 & 77.50 & 81.04 & $\times$ \\
    ViT & 6.3\sc{M} & $\times$ & $\times$ & $\times$ & $\times$ & $\times$ & $\times$ & 90.92 & 66.54 & N.A. & $\times$ \\
    Swin-T/1 & 27.5\sc{M} & $\times$ & $\times$ & $\times$ & $\times$ & $\times$ & $\times$ & 94.46 & \textbf{78.07} & N.A. & $\times$ \\
    % ViT-P-T/2 & 5.4\sc{M} & $\times$ & $\times$ & $\times$ & $\times$ & $\times$ & $\times$ & \textbf{95.91} & \textbf{81.36} & N.A. & $\times$ \\
    Parabolic CNN & 502\sc{k} &  $\times$ &  $\times$ &  $\times$ &  $\times$ &  $\times$ &  $\times$  & 88.5 & 64.8 & 77.0 &  $\times$ \\
    Hamiltonian CNN & 264\sc{k}  & $\times$ &  $\times$ &  $\times$ &  $\times$ &  $\times$ &  $\times$ & 89.3 & 64.9 & 78.3 &  $\times$\\
    CKCNN & 630\sc{k} &  $\times$ &  $\times$ &  $\times$ &  $\times$ &  $\times$ &  $\times$ &  86.8 & N.A. & N.A. &  $\times$\\
    FlexNet-6 & 670\sc{k} &  $\times$ &  $\times$ &  $\times$ &  $\times$ &  $\times$ &  $\times$ &  92.2 & N.A. & N.A. &  $\times$\\
    SpiderCNN & N.A. &  $\times$ & $\times$ &$\times$ &$\times$ &$\times$ &$\times$ & 77.97 & N.A. & N.A. & 92.4 \\
    PointConv & N.A. &  $\times$ & $\times$ &$\times$ &$\times$ &$\times$ &$\times$ & 89.13 & N.A. & N.A. & 92.5 \\
    DGCNN & 1.84\sc{M} &  $\times$ & $\times$ &$\times$ &$\times$ &$\times$ &$\times$ &$\times$ &$\times$ &$\times$ & 89.0 \\
    PointNet & 3.48\sc{M} &  $\times$ & $\times$ &$\times$ &$\times$ &$\times$ &$\times$ &$\times$ &$\times$ &$\times$ & 89.2 \\
    PointNet++ & 1.99\sc{M} &  $\times$ & $\times$ &$\times$ &$\times$ &$\times$ &$\times$ &$\times$ &$\times$ &$\times$ & 90.0 \\
    RepSurf-U  & 1.48\sc{M} &  $\times$ & $\times$ &$\times$ &$\times$ &$\times$ &$\times$ &$\times$ &$\times$ &$\times$ & \textbf{94.7}\\
    \midrule 
    CCNN$_{4, 140}$ & 200\sc{k} & \textbf{99.72} & 98.82 &  90.30 & 95.01 & 98.34 & 96.22 & 92.78 & 66.86 & 81.80 & 84.44\\
    CCNN$_{6, 380}$ & 2\sc{m} & \textbf{99.72} & \textbf{98.84} & \textbf{93.08} & \textbf{97.98} & \textbf{98.44} & \textbf{96.44} & \textbf{95.20} & 73.16 & \textbf{83.00} & 85.70 \\
    \bottomrule
    \end{tabular}}
    \end{small}
\end{center}
\end{minipage}%
\end{table}

\begin{figure}
\RawFloats
    \centering
     \begin{minipage}{0.48\textwidth}
            \includegraphics[width=0.9\textwidth]{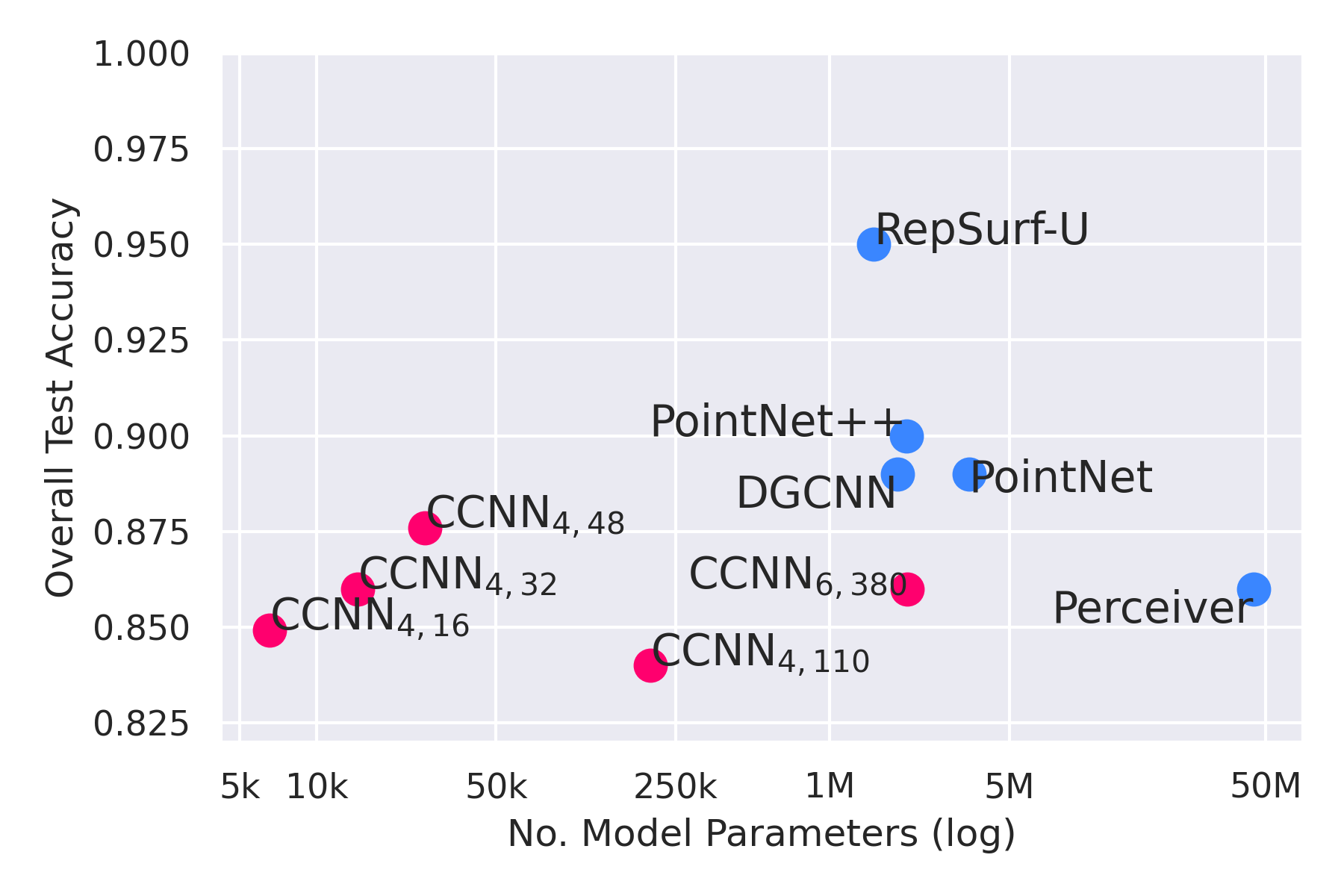}
            \vspace{-5mm}
            \caption{Parameter count versus performance on ModelNet40. Very small CCNN models obtain remarkable results.}
            \label{fig:4_modelnet40paramsize}
     \end{minipage}
     \hfill
    \begin{minipage}{0.48\textwidth}
         \includegraphics[width=0.95\textwidth, trim={0cm 0.5cm 0cm 0.5cm}, clip]{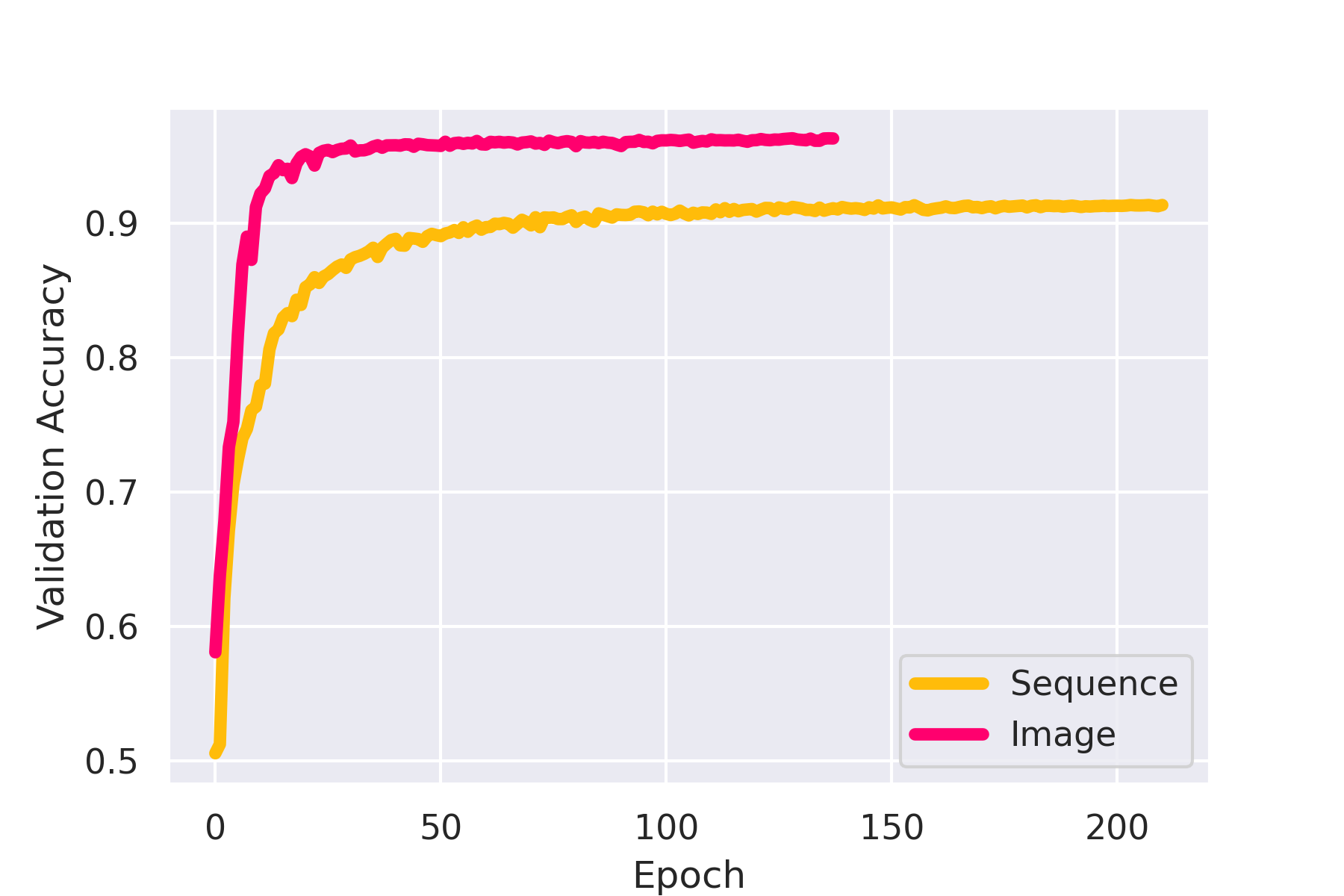}
         \vspace{-3mm}
         \caption{Performance of the CCNN$_{6,380}$ on the LRA Image task, where samples are either input as sequence or with the original image structure.}
         \label{fig:4_1dvs2d}
     \end{minipage}
\end{figure}

\underline{\textit{The importance of modelling long range dependencies in $N$D.}} In principle, all tasks can be\break treated as sequential tasks ignoring the $N$D structure --as done in S4 \citep{gu2022efficiently} for images due to the complexity of defining state spaces on $2\Dt$--, but this sacrifices structural information. Contrarily, CCNNs can be easily defined on multidimensional spaces simply by changing the dimension of the input coordinates of the kernel generator networks. We observe that by considering the 2D structure of the Image and Pathfinder tasks of the LRA benchmark, much better results can be obtained (Tab.~\ref{tab:4_lra_results}, right). In PathFinder with 2D images, the CCNN$_{6, 380}$ obtains an accuracy of $96.00$, outperforming the previous state-of-the-art by almost 10\% points and performing remarkably better than on flattened images. Additionally, we observe that models trained on the original 2D data converge faster than their sequential counterparts (Fig.~\ref{fig:4_1dvs2d}). Finally, we remark that discrete 2D CNNs with small convolutional kernels, e.g., ResNet-18 \citep{he2016deep}, are unable to solve Pathfinder due to the lack of fine-grained global context resulting from intermediate pooling layers. This was also seen by \citet{gu2022efficiently}.

\textbf{CCNNs on point-cloud datasets.} An additional advantage that comes from the continuous nature of CCNNs, is that --contrary to discrete CNNs-- CCNNs can be seamlessly applied on irregular data, e.g., point-clouds. To assess the performance of CCNNs in $3\Dt$, we consider the ModelNet40 dataset \citep{wu20153d}, which consists of 3D meshes of objects often treated as point-clouds \citep{wu2019pointconv}. Our results show that CCNNs are able to achieve a decent overall classification accuracy in comparison to point-cloud specific architectures with a relatively low parameter count (200{
\sc{k}}). In particular, CCNNs outperform the Perceiver \citep{jaegle2021perceiver} while being much smaller (Fig.~\ref{fig:4_modelnet40paramsize}).

\underline{\textit{From regular to irregular data.}} Interestingly, the continuous nature of CCNNs allows us to use CCNNs trained on regular data to process irregular data. We showcase this ability by training a CCNN$_{4, 140}$ on a version of ModelNet10 \citep{wu20153d} voxelized onto a grid of $40{\times}40{\times}40$ voxels --Fig.~\ref{fig:4_sample_voxelized} shows an example of the point-cloud and voxelized datasets--. After trained, we deploy the CCNN on the original (point-cloud) and voxelized test sets and observe respective test accuracies of $90.99$ and $90.64$. This result shows that CCNNs learn representations that reflect the continuous nature of the data, showcasing CCNNs as a truly general-purpose architecture even able to generalize across different data representations --a feat hardly obtainable by existing architectures--.

\underline{\textit{Large CCNNs overfit on point-clouds, but tiny CCNNs perform surprisingly well.}} In contrast to $1\Dt$ and $2\Dt$, we observe that larger CCNNs perform worse than smaller ones on $3\Dt$ point-clouds (Fig.~\ref{fig:4_modelnet40paramsize}). We hypothesize that this is a result of the sparsity of the task. Continuous convolutional kernels define a kernel function over the entire domain even if the number of kernel values sampled is sparse. As a result, sparse supervision makes it hard for the network to learn a kernel function that generalizes to unseen data, as sparsity exposes the kernel function to aliasing --high frequencies between sampled points--. To corroborate this hypothesis, we train extremely small CCNNs --with 6{\sc{k}}, 14{\sc{k}} and 48{\sc{k}} parameters-- on ModelNet40. We observe that these models achieve surprising results and even outperform the $200${\sc{k}} and $2${\sc{m}} CCNNs (Fig.~\ref{fig:4_modelnet40paramsize}). These results supports our previous hypothesis, but also show that the continuous kernel paradigm can be used to construct extremely small, performant CNNs.
\begin{table}
\centering
    \begin{minipage}{\textwidth}
    \centering
    \caption{Experimental results on the Long Range Arena benchmark. {\footnotesize $\times$ -  \textit{unable to apply due to the model's inability to handle the specified data.}}}
    \label{tab:4_lra_results}
    \vspace{-3mm}
    \begin{small}
    \scalebox{0.7}{
    \begin{tabular}{cccccc|cc}
    \toprule
     & \sc{ListOps}  & \sc{Text} & \sc{Image} & \sc{Pathfinder} & \sc{Avg.}& \sc{2DImage} & \sc{2DPathfinder} \\
     \midrule
    Transformer & 36.37 & 64.27 &  42.44 & 71.40 & 53.66 & $\times$&$\times$\\
    Reformer & 37.27 & 56.10 & 38.07 & 68.50 & 50.56 & $\times$&$\times$\\
    BigBird & 36.05 & 64.02 & 40.83 & 74.87 & 57.17  & $\times$&$\times$\\
    Linear Trans. & 16.13 & 65.90 & 42.34 & 75.30 & 50.46  & $\times$&$\times$\\
    Performer & 18.01 & 65.40 & 42.77 & 77.05 & 51.18  & $\times$&$\times$ \\
    FNet & 35.33 & 65.11 &38.67 &77.80 &54.52 & $\times$&$\times$\\
    Nystromförmer &37.15 &65.52 &41.58 &70.94 &57.46  & $\times$&$\times$ \\
    Luna-256 & 37.25 &64.57 &47.38 &77.72 &59.37  & $\times$&$\times$ \\
    S4 &\textbf{58.35} &76.02 &87.26 &86.05 & \textbf{80.48}  & $\times$&$\times$\\
    \midrule
    CCNN$_{4, 140}$  & 44.85 & 83.59 & 87.62 & 91.36 & 76.86  &89.48 & 94.80\\
    CCNN$_{6, 380}$ & 43.60 &  \textbf{84.08} & \textbf{88.90} & \textbf{91.51} & 77.02  & \textbf{91.12} & \textbf{96.00} \\
    \bottomrule
    \end{tabular}}
    \end{small}
    \end{minipage}%
\end{table} 
\vspace{-7mm}
\section{Limitations and Future Work}
\label{sec:4_limitations-and-conclusion}
\textbf{Computational efficiency.} Although convolutions with large convolutional kernels scale better than self-attention, e.g., Perceiver \citep{jaegle2021perceiver}, they can still be expensive. Luckily, the Fourier convolution can be used to strongly reduce their cost, e.g., \cite{romero2022ckconv, romero2022flexconv, gu2022efficiently}. Nevertheless, the Fourier convolutions by themselves are not sufficient to scale CCNNs to very large inputs. This problem is exacerbated if irregular data is considered, as it (\textit{i}) prevents the usage Fourier convolutions, and (\textit{ii}) requires rendering a different convolutional kernel for each spatial position of each sample in the batch. An important avenue for future research may look into reducing computational requirements, either via separability or self-adjusting architectures, e.g., downsampling \citep{riad2022learning}, for regular data, and \enquote{gridifying} techniques for irregular data.

\textbf{Larger models on point-cloud data.} Although CCNNs performs remarkably well on point-clouds with extremely small models, the current CCNN formulation is unable to achieve better performance with larger models. We hypothesize that this is due to increasing aliasing in the learned kernel generator networks. Although anti-aliasing techniques for kernel generator networks exist \citep{romero2022flexconv}, this assume an underlying grid and a corresponding Nyquist frequency. An additional avenue for future research may look into the aliasing issue on irregular data either by generalizations of \cite{romero2022flexconv}, or properly-defined smoothing techniques.

\textbf{Cross-modal training and data fusion.}
CCNNs present a potential solution for cross-modal training --currently challenging due to dissimilar per-modality architectures--. Additionally, the continuous properties of CCNNs allow them to be trained on a wide range of data sources, e.g., multiple resolutions, regular, irregular data, etc, which is interesting for several machine learning application.

%% file: chapters/X_gridifier.tex
% path to figures directory
\graphicspath{{figures/X-gridifier/}}

%=========================================================================

% \begin{savequote}[75mm]
% Nulla facilisi. In vel sem. Morbi id urna in diam dignissim feugiat. Proin molestie tortor eu velit. Aliquam erat volutpat. Nullam    ultrices, diam tempus vulputate egestas, eros pede varius leo.
% \qauthor{Quoteauthor Lastname}
% \end{savequote}

\chapter{Learnable Gridification for Efficient Point-Cloud Processing}\label{chapter:gridification}
\begin{flushright}
\textit{Based on the paper:}\break
\textit{Learnable Gridification for Efficient Point-Cloud Processing \citep{vanderlinden2023learned}}
\end{flushright}
%=========================================================================
\vspace{-7mm}
\section{Introduction}\label{sec:intro}
Point clouds provide sparse geometric representations of objects or surfaces equipped with signals defined over their structure, e.g., the surface normals of an underlying object \cite{wu20153d, qi2017pointnet} or the chemical properties of a molecule \cite{ramakrishnan2014quantum,schutt2017schnet}. Several neural operators have been developed that can be applied to such sparse representations provided by point clouds. These methods can be broadly understood as continuous generalizations of neural operators originally defined over regular discrete grids, e.g., convolution \cite{wu2019pointconv} and self-attention \cite{zhao2021point}.

\begin{figure}
  \centering
  \includegraphics[width=0.85\linewidth]{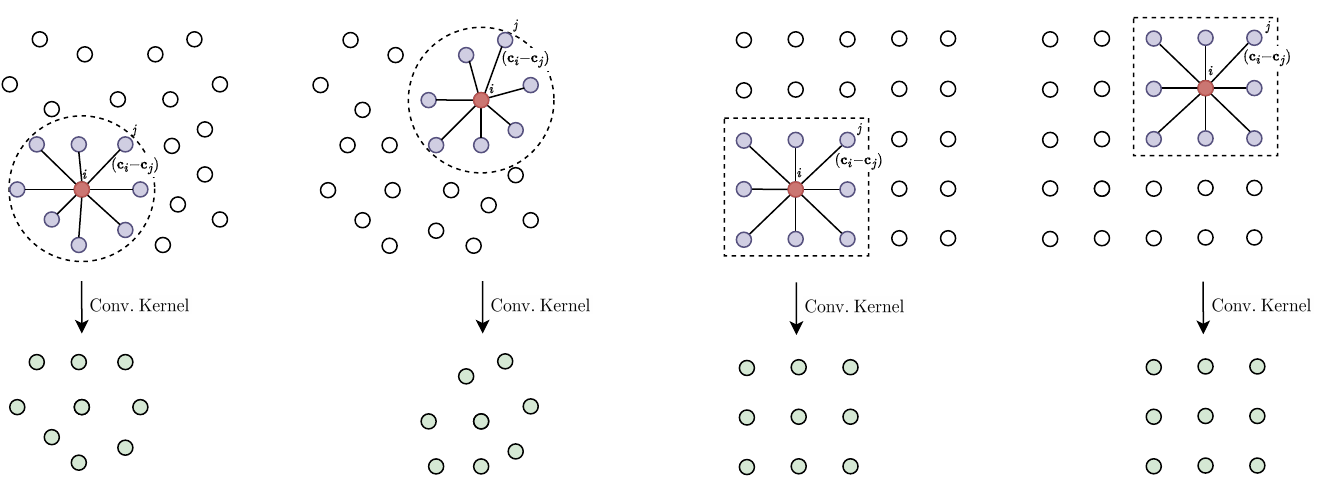}
  \vspace{-1mm}
  \caption{Convolution on point clouds and grids. Due to the irregular nature of point clouds, convolutional kernels --and other operations based on neighborhood information-- must be re-rendered for every query point in the point cloud (left). In contrast, grid data is regularly arranged, and thus pairwise distances are equal for any query point in the grid (right). As a result, the convolutional kernel can be computed once and reused for all query points.}
  \label{fig:convs}
\end{figure}

\textbf{The problem of learning on raw point clouds.} Unfortunately, the flexibility required from neural operators to accommodate irregular sparse representations like point clouds brings about important increases in time and memory consumption. This is especially prominent in neural operations that construct feature representations based on neighborhood information, e.g., convolution. In the case of point clouds, the irregular distances between points make these neural operations significantly more computationally demanding compared to regular grid representations like images or text. For instance, for convolution, the convolutional kernel needs to be recalculated for each point in a point cloud to account for irregular distances from the query point to other points in its neighborhood (Fig.~\ref{fig:convs} left). In contrast, grid representations standardize pairwise distances following a grid structure (Fig.~\ref{fig:convs} right). As a result, the distances from a point to all other points in its neighborhood are fixed for all points queried in the grid. Therefore, it is possible to compute the kernel once, and reuse it across all query positions. This difference illustrates that operations relying on neighborhood information scale much worse in terms of memory and time for point clouds than for grid data, specially for large inputs and large neighborhoods.
\begin{figure}
\captionsetup{justification=centering}
  \centering
  \includegraphics[width=0.7\linewidth]{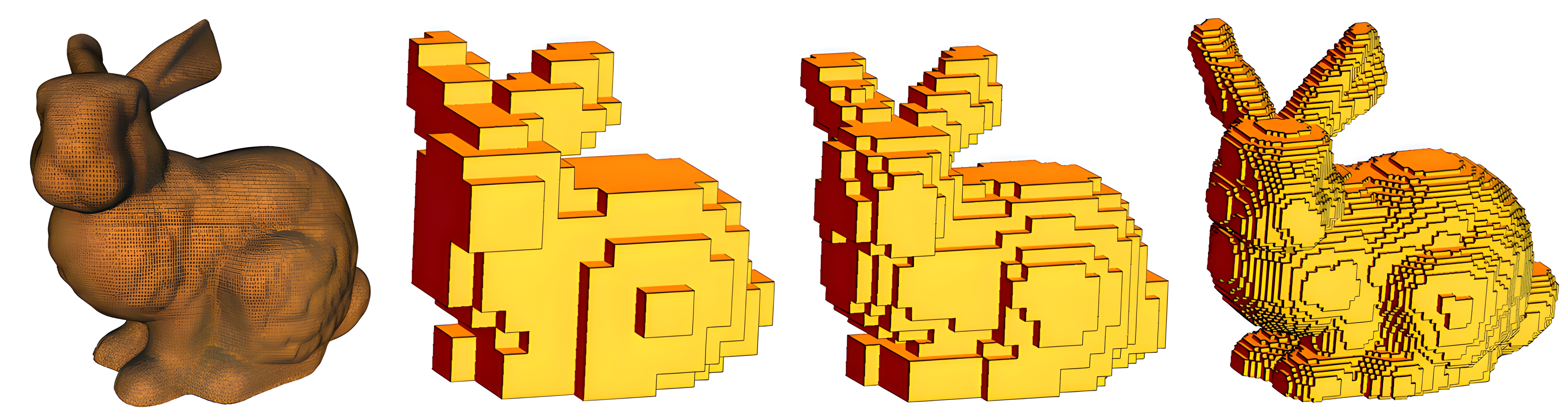}
  \vspace{-2mm}
  \caption{Voxelization of the Stanford Bunny \cite{turk1994zippered} for different resolutions \citep{karmakar2011construction}.}
  \label{fig:voxelization}
\end{figure}

\textbf{A potential solution: Voxelization.} A potential solution to address the challenges posed by point clouds lies in treating the point cloud as a continuous density that can be sampled on a dense regular grid: a process called \textit{voxelization} \cite{maturana2015voxnet, wu20153d}. The idea of voxelization is to create a grid that overlaps with the domain of the point cloud (Fig.~\ref{fig:voxelization}). Although voxelization methods create grid representations on which neural operations defined on grids can act, e.g., \texttt{Conv3D}, the grids resulting from voxelization are oftentimes much larger than the number of points in the original point cloud. This is a consequence of (\textit{i}) the high-resolution grids required to describe fine details from the point cloud, and \textit{(ii)} the low occupancy of the grid resulting from the sparse nature of point clouds which generally leads to many more points to process in the resulting grid than in the original point cloud.

\begin{figure*}
\centering
\hfill
\includegraphics[width=0.3\linewidth]{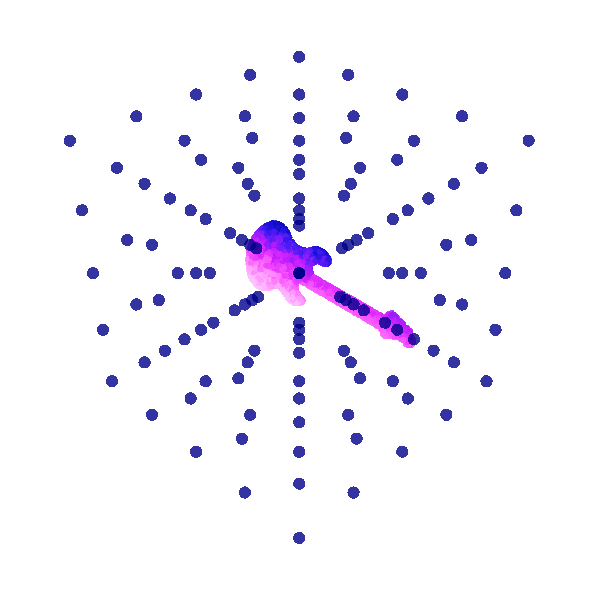}\hfill
\includegraphics[width=0.3\linewidth]{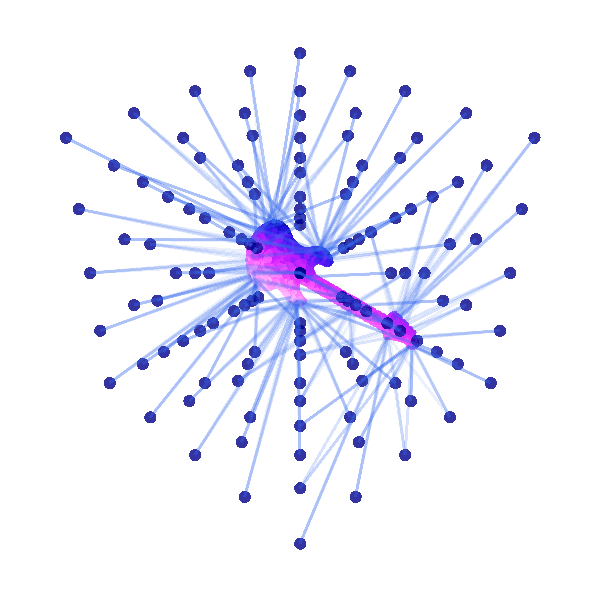}\hfill
\includegraphics[width=0.3\linewidth]{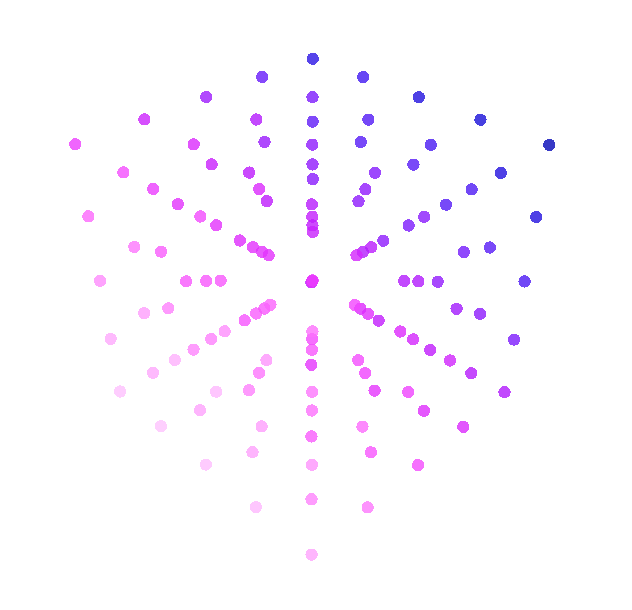}
\hfill
\vspace{-2mm}
\caption{Gridification. Gridification maps a point cloud $\gP$ onto a compact regular grid $\gG$. The method first constructs a $\Dt$-dimensional grid (left) that overlaps the point cloud. Then, it connects points on the point cloud to points in the grid given by a connectivity scheme $\gE_{\gP \rightarrow \gG}$, i.e., a set of edges from points in the point cloud to points in the grid,  determined by \textit{bilateral k-nearest neighbors connectivity} (middle). Finally, gridification propagates information from the point cloud onto the grid through a \textit{convolutional message passing} layer acting over the bipartite graph $(\gP, \gG,  \gE_{\gP \rightarrow \gG})$. By carefully selecting the different components of the gridification module, gridification is able to construct compact rich grid representations that can be subsequently processed with grid operations such as \texttt{Conv3D}.}
\label{fig:gridification}
\end{figure*}

\textbf{Our proposed solution: Gridification.} In this paper, we propose an alternative solution to address the memory and computational scalability of point cloud methods by addressing its root cause: \textit{the irregularity of the data}. We propose \textit{learnable gridification} as the first step in a point cloud processing pipeline to transform the point cloud into a \textit{compact, regular grid} (Fig.~\ref{fig:gridification}). Thanks to gridification, subsequent layers can use operations defined on grids, e.g., \texttt{Conv3D}, which scale much better than native point cloud methods. In a nutshell, gridification can be understood as a \textit{convolutional message passing} layer acting on a \textit{bipartite graph} that establishes connections between points in the point cloud to points in the grid given by a \textit{bilateral $k$-nearest neighbor connectivity}. The proposed bilateral $k$-nearest neighbor connectivity guarantees that all points both in the point cloud and in the grid are connected, therefore allowing for the construction of expressive yet compact grid representations.

In contrast to voxelization, gridification produces expressive compact grid representations in which the number of points in the resulting compact regular grid is roughly equal to the number of points in the original point cloud, yet the grid is able to preserve fine geometric details from the original point cloud. For instance, we observe that point clouds with $\Nt{=}1000$ points can be effectively mapped to a compact dense $\mathrm{10x10x10}$ grid without significant information loss. We show through theoretical and empirical analysis that the resulting grid representations scale much better in terms of memory and time than native point cloud methods. This is verified on several comparison studies for increasing number of points in the point cloud and increasing neighborhood sizes in the construction of convolutional kernels.

We demonstrate that gridification can also be used for tasks from point clouds to point clouds, e.g., segmentation. To this end, we introduce a \textit{learnable de-gridification} step at the end of the point cloud processing pipeline, which can be seen as an inverted gridification step that maps the compact, regular grid back to its original point cloud form. This extension allows for the construction of \textit{gridified networks} --networks that operate on grids-- to solve global prediction tasks, e.g., classification, as well as dense prediction tasks, e.g., segmentation and regression, on point cloud data.

%\textbf{Results.} {\color{green} TODO}%Through several experiments, we demonstrate that gridified networks are competitive to native point cloud methods while being much more efficient in terms of memory and computation speed. This is observed on both classification and segmentation tasks on ModelNet, ShapeNet, S3DIS and ScanObjectNN.

\vspace{-7mm}
\section{Method}\label{sec:method}
\vspace{-5mm}
\subsection{Point cloud and grid representations} 
\textbf{Point cloud.} A point cloud $\gP {=}\{(\cv^\gP_i, \xv^\gP_i)\}_{i{=}1}^{\Nt_\gP}$ is an \textit{unstructured} set of $\Nt_\gP$ pairs of coordinate-feature values $(\cv^\gP_i, \xv^\gP_i)$ scattered in space without any predefined pattern or connectivity. Point clouds sparsely represent geometric structures through pairs of coordinate vectors $\cv^\gP_i \in \sR^\Dt$ and corresponding function values over that geometric structure $\xv_i^\gP \in \mathbb{R}^{\Ft_{\gP}}$, e.g., surface normals, RGB-values, electric potentials, etc.

\textbf{Grid.} A grid $\gG{=}\{(\cv^\gG_i, \xv^\gG_i)\}_{i{=}1}^{\Nt_\gG}$ can be interpreted as a point cloud on which the coordinate-feature pairs $(\cv^\gG_i, \xv^\gG_i)$ are arranged in a regular pattern that form a lattice. In contrast to general point clouds, points in a grid are evenly spaced and align along predefined axes, e.g., $x$, $y$, $z$. The regular spacing between points leads to regular pairwise distances for all query points in the grid. As a result, we can calculate pairwise attributes once, and reuse them for all query points. 

\vspace{-7mm}
\subsection{Gridification: From a point cloud to a dense grid}
We seek to map the sparse point cloud $\gP {=}\{(\cv^\gP_i, \xv^\gP_i)\}_{j{=}1}^{\Nt_\gP}$ onto a compact regular grid $\gG {=}\{(\cv_i^\gG, \xv_i^\gG)\}_{i{=}1}^{\Nt_\gG}$ in $\sR^\Dt$. We formalize this process as an operation over a \textit{bipartite graph} that establishes connections between points in the point cloud $\gP$ to points in the grid $\gG$ given by a \textit{connectivity scheme} $\gE_{\gP \rightarrow \gG}$ defined as a set of edges $\ev_{j \rightarrow i} \in \gE_{\gP \rightarrow \gG}$.

\textbf{Learnable gridification as message passing.} We aim to learn a mapping from $\gP$ to $\gG$ such that the grid representation $\gG{=}\{(\cv^\gG_i, \xv^\gG_i)\}_{i{=}1}^{\Nt_\gG}$, $\cv_i \in \sR^\Dt$, $\xv^\gG_i \in \mathbb{R}^{\Ft_{\gG}}$, adequately represents the source point cloud $\gP$ for the downstream task. Given a source point cloud $\gP$, a target grid $\gG$ and a connectivity scheme $\gE_{\gP \rightarrow \gG}$, we define gridification as a \textit{convolutional message passing} layer \cite{gilmer2017neural} on the bipartite graph $(\gP, \gG, \gE_{\gP \rightarrow \gG})$ defined as:
\begin{equation}
    \xv^\gG_i = \phi_{\mathrm{upd}} \left( \bigoplus_{\ev_{j \rightarrow i} \ \in\  \gE_{\gP \rightarrow \gG}}\phi_{\mathrm{msg}}\Big(\phi_{\mathrm{node}}\left(\xv^\gP_j\right),  \phi_{\mathrm{pos}}\left(\cv^\gG_i - \cv^\gP_j\right) \Big) \right). 
\end{equation}
It consists of a node embedding network $\phi_\mathrm{node}: \sR^{\Ft_{\gP}} \rightarrow \sR^{\Ht}$ that processes the point cloud features $\xv_i^\gG$, a positional embedding network $\phi_\mathrm{pos}: \sR^{\Dt} \rightarrow \sR^{\Ht}$ that creates feature representations based on the pairwise distances between coordinates in $\gG$ and $\gP$ --thus resembling a convolutional kernel--, a message embedding network $\phi_\mathrm{msg}: \sR^{2\Ht} \rightarrow \sR^{\Ht}$ that receives both the node embedding and the relative position embedding to create the so-called \textit{message}. After the messages are created for all nodes described by connectivity of the node, these features are aggregated via the aggregation function $\bigoplus$, e.g., $\max$, $\mathrm{mean}$. Finally, the aggregated message is passed through the update network $\phi_\mathrm{upd}: \sR^{\Ht} \rightarrow \sR^{\Ft_{\gG}}$ to produce the grid feature representations $\xv_i^{\gG} \in \sR^{\Ft_{\gP}}$.

\vspace{-7mm}
\subsection{De-gridification: From a dense grid to a point cloud}
To extend the use of gridification to tasks from the point cloud $\gP$ to the point cloud $\gP$, e.g., segmentation, regression, we define a \textit{de-gridification} step that sends a grid representation $\gG$ back to its original point cloud form $\gP$. Formally, the de-gridification step is defined as:
\begin{equation}
    \xv^\gP_{i} = \phi_{\mathrm{upd}} \left( \bigoplus_{\ev_{j \rightarrow i} \ \in\  \gE_{\gG \rightarrow \gP}}\phi_{\mathrm{msg}}\Big( \phi_{\mathrm{node}}\left(\xv^\gG_j\right),  \phi_{\mathrm{pos}}\left(\cv^\gP_i - \cv^\gG_j\right) \Big) \right). 
\end{equation}
Intuitively, de-gridification can be interpreted as a gridification step from $\gG$ to $\gP$ given by an inverted connectivity scheme $\gE_{\gG \rightarrow \gP} {=} ( \gE_{\gP \rightarrow \gG})^{-1}$. Note that, it is not necessary to calculate the connectivity scheme for the de-gridification step. Instead, we can obtain it simply by taking the connectivity scheme from the gridification step $\gE_{\gG \rightarrow \gP}$ and inverting the output and input nodes of the edges.

\vspace{-7mm}
\subsection{Requirements and properties of gridification}\label{sec:requirements}
We desire to construct a compute and memory efficient grid representation $\gG$ that captures all aspects of the point cloud $\gP$ as good as possible. That is, a compact, yet rich grid representation $\gG$ that preserves the structure of the point cloud $\gP$ with as low loss of information as possible. With this goal in mind, we identify the following requirements:
\begin{enumerate}[label=(\textit{\roman*}), topsep=0pt, leftmargin=*, itemsep=0pt]
\item The number of points in the grid $\Nt_\gG$ should be at least as large as the number of points in the point cloud $\Nt_\gP$.
%\textcolor{blue}{We desire the gridification step to be able to represent \textit{at least} a one-to-one or one-to-many mapping. We wish gridification to be injective function so as to ensure no loss of information.}
\item The width of all hidden representations of the node embedding network $\phi_\mathrm{node}$ should be \textit{at least as large} as the width of the point cloud features $\xv_i^\gP$, i.e., $\Ft_{\gP}$. %\textcolor{blue}{The reasoning behind this property follows that of $(i)$.}
\item The width of all hidden representations of the position embedding network $\phi_\mathrm{pos}$ should be at least as large as the dimension of the domain $\Dt$.% \textcolor{blue}{This requirement follows from the desire for injectivity.}
\item The width of all hidden representations of the embedding networks $\phi_\mathrm{upd}$, $\phi_\mathrm{msg}$ should be \textit{at least as large} as the width of the point cloud features $\xv_i^\gP$ plus the dimension of the domain $\Dt$.% \textcolor{blue}{This requirement follows from the desire for injectivity.}
\item Each point-cloud point $\cv^\gP$ should be connected to \textit{at least} one point $\cv^\gG$ in the grid. %\textcolor{blue}{This requirement ensures that all input points contribute to the grid representation, and no information is discarded.}
\item The positional embedding network $\phi_\mathrm{pos}$ should describe high frequencies. %to avoid oversmooted representations of the point cloud onto the grid \gG.
\item Each grid point $\cv^\gG$ should be connected to \textit{at least} one point $\cv^\gP$ in the point cloud. %\textcolor{blue}{Unlike voxelization methods, we desire full occupancy of the grid representation.}
%\item The domain of the grid $\gG$ should correspond to the domain of the source point cloud $\gP$. %\textcolor{magenta}{Not sure what is meant with this one?}
\end{enumerate}
\textbf{Preventing information loss.} To prevent information loss, we want to avoid any kind of compression either in the grid representation or in any intermediary representation during the gridification process. Consequently, we restrict the number of points as well as the width of all representations to be at least as big as the corresponding dimensions in the source point cloud $\gP$ --items (\textit{i})-(\textit{iv})--. In addition, we must make sure that all points in the point cloud are connected to points in the grid to prevent points from being disregarded during gridification --item (\textit{v})--. Finally, we must also make sure that the positional embedding network $\phi_\mathrm{pos}$ is able to represent high frequencies --item (\textit{vi})--. This is important as multilayer perceptrons ($\mathrm{MLP}$s) with piecewise nonlinearities, e.g., $\mathrm{ReLU}$, have been shown to have an implicit bias towards smooth functions \cite{tancik2020fourier, sitzmann2020implicit}. In the context of gridification, this means that using conventional $\mathrm{MLP}$s for the positional embedding network $\phi_\mathrm{pos}$ could result in over-smooth grid representations unable to represent fine details from the source point cloud. We circumvent this issue by using parameterizations for $\phi_\mathrm{pos}$ able to model high frequencies (Sec.~\ref{sec:pos_network}).

\textbf{Encouraging compact representations.} In addition to encouraging no information loss, we also identify requirements that encourage the resulting grid representation to be compact and expressive. First, we note that item (\textit{v}) is important for this end as well, as over-smooth representations implicitly require higher resolutions to be able to encode fine-grained details. Additionally, we impose all points in the grid to be connected to points in the point cloud --item (\textit{vii})-- to prevent the grid representation from having low occupancy. This restriction allows us to make sure that all the spatial capacity of the grid is being used. This in turn allows us to construct compact rich grid representations.

\vspace{-7mm}
\subsection{Materializing the gridification module}
Based on the previous requirements and properties, we define the components of the gridification module as follows:

\vspace{-7mm}
\subsubsection{The grid $\gG$}
 Let $[a, b]^\Dt$ be the domain of the point cloud $\gP$, i.e., $\cv^\gP_i \in [a, b]^{\Dt}$, $\forall\ \cv^\gP_i\in \gP$. Then, we define the regular grid $\gG$ over the same domain $[a, b]^\Dt$ with $\sqrt[\leftroot{-3}\uproot{3}\Dt]{\Nt^{\gG}}$ points along each dimension. By doing so, we guarantee that the grid $\gG$ is uniformly spaced over the domain of the point cloud, therefore (\textit{i}) preserving the statistics of the input point cloud, and (\textit{ii}) being able to represent the underlying signal in the same range. In practice, point clouds are normalized during preprocessing. As a result, we often have that $a{=}-1$, $b{=}1$, leading to a point cloud and a grid defined on $[-1, 1]^\Dt$.
 
 \vspace{-7mm}
\subsubsection{The connectivity scheme $\gE_{\gP \rightarrow \gG}$}
\begin{figure}
\captionsetup{justification=centering}
    \centering
    \includegraphics[width=0.6\textwidth]{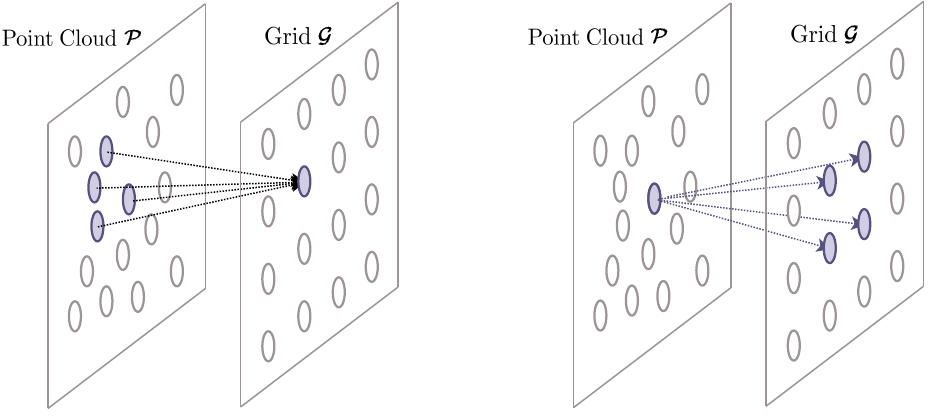}
    \vspace{-2mm}
    \caption{Bilateral $k$-nearest neighbor connectivity for $k{=}4$.}
    \label{fig:bilateral_conectivity}
\end{figure}

Motivated by the requirements in Sec.~\ref{sec:requirements}, we opt for \textit{bilateral $k$-nearest neighbor connectivity} over common alternatives such as radius connectivity \cite{qi2017pointnet, qi2017pointnet++} or one-way $k$-nearest neighbor connectivity \cite{barber1996quickhull, connor2010fast} for the construction of the connectivity scheme $\gE_{\gP \rightarrow \gG}$ to guarantee that no points either in the grid $\gG$ nor the point cloud $\gP$ are disconnected. Bilateral $k$-nearest neighbor connectivity consists of a two-way $k$-nearest neighbor approach in which first each point $\cv_i^{\gG}$ in the grid is linked to the $k$ nearest points $\cv_j^\gP$ in the point-cloud. Subsequently, connections are established from each point $\cv_i^{\gP}$ in the point cloud to its nearest $k$ points $\cv_j^{\gG}$ in the grid (Fig.~\ref{fig:bilateral_conectivity}). By following this procedure, bilateral $k$-nearest neighbor connectivity creates a \textit{complete} connectivity scheme from $\gP$ to $\gG$ with at least $k$ and at most $2k$ edges per point.

\vspace{-7mm}
\subsubsection{The positional embedding network $\phi_\mathrm{pos}$}\label{sec:pos_network}
In literature, the positional embedding network $\phi_\mathrm{pos}$ is often parameterized as an $\mathrm{MLP}$ with piecewise nonlinearities, e.g., $\mathrm{ReLU}$, that receives relative positions $(\cv_i {-} \cv_j)$ as input and retrieves the value of an spatial function at that position $\phi_\mathrm{pos}(\cv_i {-} \cv_j)$ \cite{schutt2017schnet, qi2017pointnet++, wu2019pointconv}. However, previous studies have shown that $\mathrm{MLP}$s with piecewise nonlinearities suffer from an spectral bias towards low frequencies, which limits their ability to represent functions with high frequencies \cite{tancik2020fourier, sitzmann2020implicit}. In the context of modelling spatial neural operators such as $\phi_\mathrm{pos}$, this implies that using piecewise $\mathrm{MLP}$s to parameterize spatial neural operators leads to inherently smooth operators. Consequently, applying such an operator over an input function, e.g., via a convolution operation, would implicitly perform a low-pass filtering of the input, causing the output representations to lack information regarding fine-grained details of the input.

To overcome this issue, we rely on the insights from \textit{Continuous Kernel Convolutions} \cite{romero2022ckconv} and parameterize the positional embedding network as a \textit{Neural Field} \cite{sitzmann2020implicit, tancik2020fourier}. In contrast to piecewise $\mathrm{MLP}$s, neural fields easily model high frequencies, and thus allow for powerful parameterizations of spatial neural operators that do not perform smoothing. In the context of gridification, using neural fields to parameterize $\phi_\mathrm{pos}$ allows gridification to project fine-grained geometric information from the point cloud onto the grid.

\vspace{-7mm}
\subsection{Gridified networks for global and dense prediction}\label{section:gridnetworks}

Gridification and de-gridification allow for the construction of \textit{gridified networks} able to\newpage

\begin{wrapfigure}{r}{0.48\linewidth}
    \centering
    \includegraphics[width=0.8\linewidth]{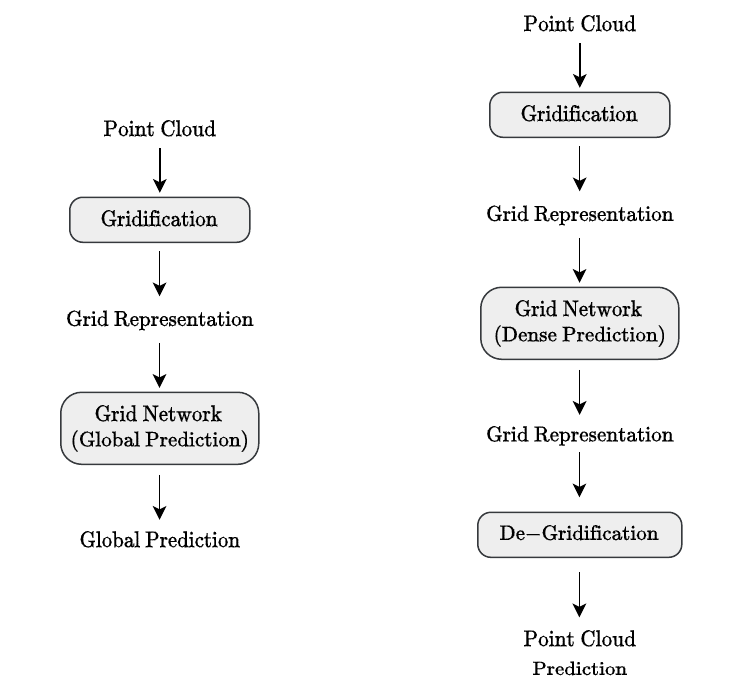}
    \vspace{-3mm}
    \caption{Point cloud pipeline for global (left) and dense prediction tasks (right).}
    \label{fig:gridified_nets}
\end{wrapfigure}
process point clouds both for global and dense prediction tasks (Fig.~\ref{fig:gridified_nets}). For global prediction tasks, e.g., classification, we construct a point cloud processing pipeline consisting of gridification, followed by a \textit{grid network}, i.e., a neural network that operates on grid data, designed for global prediction, e.g., a ResNet \cite{he2016deep} or a ViT \cite{dosovitskiy2020image}. For dense prediction tasks, e.g., segmentation, our proposed point cloud pipeline consists of gridification, followed by a grid network designed for dense predictions, e.g., a U-Net \cite{ronneberger2015u} or a CCNN \cite{knigge2023modelling}. After the processed grid representation is obtained, we utilize the de-gridification step to map back the grid representation to a point cloud with the output node predictions.
%The resulting practical architecture consists of the simple but effective recipe outlined in Fig \ref{fig:archrecipe}. 

\vspace{-7mm}
\section{Related Work}
Deep learning approaches for point cloud processing can be broadly classified in two main categories: (\textit{i}) native point cloud methods and (\textit{ii}) voxelization methods.

\textbf{Native point cloud methods.} Native point cloud methods operate directly on the raw, irregular point cloud data without any preprocessing steps such as voxelization. These methods leverage the inherent spatial distribution of the points to extract meaningful features. PointNet \cite{qi2017pointnet} introduced a pioneering framework for point cloud processing by employing shared multilayer perceptrons and symmetric functions to learn global and local features from unordered point sets. PointNet++ \cite{qi2017pointnet++} extended this work with hierarchical neural networks to capture hierarchical structures in point clouds. PointConv \cite{wu2019pointconv} introduced a convolution operation specifically designed for point clouds, incorporating local coordinate systems to capture local geometric structures. PointGNN \cite{shi2020point} utilized graph neural networks to model interactions between neighboring points. 

Despite the flexibility in handling irregular data that native point cloud methods provide, they suffer from scalability issues due to the increased computational and memory complexity of processing unstructured point sets.

\textbf{Voxelization methods.} Voxelization methods aim to convert the irregular point cloud data into a regular grid structure, enabling the utilization of neural architectures designed for regular grid data. VoxNet \cite{maturana2015voxnet} introduced the concept of voxelization for point clouds and employed 3D convolutions on the resulting grid representations. Volumetric CNN \cite{qi2016volumetric} extended this approach with an occupancy grid representation and achieved impressive performance on 3D shape classification tasks. Other works, such as VoxSegNet \cite{wang2019voxsegnet} explore variations of voxelization techniques to improve performance on tasks like object detection and segmentation. 

While voxelization methods offer a well-founded solution to the computational and memory complexity of native point cloud methods, in practice, they suffer from high memory consumption and information loss due to the discretization process. This is due to the inherent trade-off between the need to capture fine geometric details --which requires high resolution grids--, and the need for efficiency --which favors low resolution grids--. As a result, conventional voxelization methods struggle to strike a balance between resolution and speed. In contrast, gridification is able to generate compact yet expressive grid representations able to preserve fine geometric details on a low resolution grid with roughly the same number of points as the source point cloud. 

\textbf{Hybrid methods.} Aside from pure  point cloud and voxelization methods, there exist works that attempt combine the advantages of both categories. Their main idea is to combine point-wise and grid-wise operations to perform effective feature extraction while maintaining scalability and efficiency. PointGrid \cite{le2018pointgrid} uses a hybrid representation by voxelizing the point cloud and employing a combination of point-wise and grid-wise operations at each layer. Point-Voxel CNN \citep{liu2019point} combines grid convolutions with point-wise feature extraction. It uses low-resolution voxelization to aggregate neighborhoods with regular 3D convolutions and $\mathrm{MLP}$s to generate point-wise features that preserve fine-grained structure. These features are  then fused through interpolation. Point-Voxel Transformer \cite{zhang2022pvt} follows a similar two-branch structure, but replaces convolutions with windowed self-attention.

Although hybrid methods reduce the computational and memory complexity of native point cloud methods, their explicit use of voxelization still leads to a trade-off between information loss and efficiency on that branch. To compensate for the information lost during voxelization, they require a parallel raw point cloud branch, which does not scale well. In contrast, gridification does not make use of raw point cloud branches but instead focuses on the creation of descriptive compact grid representations that preserve the geometric information of the source point cloud. Hence, gridification offers a solution with better scalability properties than existing hybrid methods.  

\vspace{-7mm}
\section{Experiments}
To evaluate our approach, we first analyze the expressive capacity of gridification and de-gridification on a toy point cloud reconstruction task. Next, we construct gridified networks and evaluate them on classification and segmentation tasks. In addition, we provide empirical analyses on the computational and memory complexity of gridified networks which we then corroborate with theoretical analyses.
\begin{table*}[t]
\centering
    \begin{minipage}{\textwidth}
    \centering
    \caption{Classification performance on ModelNet40 benchmark.}
    \label{tab:modelnet_results}
    \vspace{-2.5mm}
    \begin{small}
    \scalebox{0.85}{
    \begin{tabular}{lllll}
    \toprule
    \sc{Model} & \sc{Input}  & \sc{Type} &  \sc{Accuracy} & \sc{Parameters} \\
    \midrule
    PointNet++ \cite{qi2017pointnet++} & $32 \times 1000$ & native & 89.64 &  1.5M \\
    VoxNet \cite{maturana2015voxnet} & $32\times 30^3$ & voxelization & 83.00 & 0.92M\\
    %Linear Trans. & 16.13 & 65.90 & 42.34 & 75.30 & 50.46  & $\times$&$\times$\\
    PointGrid \cite{le2018pointgrid} & $32\times16^3$ & voxelization & 92.00	 & - \\
    Point Voxel Transformer \cite{zhang2022pvt} & $32\times1024$ & hybrid & 94.00 & 2.76M\\
    Gridified Networks 3x3x3 (Ours) & $32\times1000 \rightarrow 32 \times 3^3$& voxelization & 90.86   & 0.28M  \\
    Gridified Networks 9x9x9 (Ours)  & $32\times1000 \rightarrow 32 \times 9^3 $ & voxelization & 92.28   & 0.47M  \\
    % \midrule
    % \midrule
    % CCNN$_{6, 380}$ (Global Kernels) & 59.60 & 88.10 & 90.59 & 86.70 & 95.24 \\
    % $\mathrm{DNArch}_\mathrm{K,R,W,D}$(CCNN$_{6, 380}$) & \textbf{61.05}$_{(1.00\times)}$ & \textbf{89.13}$_{(1.00\times)}$ & & \textbf{89.53}$_{(1.00\times)}$ \\
    \bottomrule
    \end{tabular}}
    \end{small}
    \end{minipage}%
\end{table*}

\textbf{Experimental setup.} For the position embedding function $\phi_\mathrm{pos}$ we use an Random Fourier Feature Network \cite{tancik2020fourier}, due to explicit control over the smoothness through the initial frequency parameter $\Omega$. The practical setup and instantiation of the convolution blocks can be found in Appendix \ref{app:networkarch}. We train our models without data augmentation using AdamW \cite{loshchilov2018decoupled} and a cosine scheduler \citep{loshchilov2017sgdr} with 10 epochs of linear warm-up. We follow the standard procedure and preprocess all objects in the datasets to be centered and normalized. For each dataset, we choose the grid resolution such that its number of points is roughly equal to the size of the original point cloud. For ModelNet40 we use surface normals in addition to positions as node features. Dataset specific hyperparameters can be found in Appendix \ref{app:hyperparams}. % Our code is publicly available at \href{https://github.com/computri/gridified-nets}{\texttt{github.com/computri/gridified-nets}}.

% \textbf{Experimental setup.} For the position embedding function $\phi_\mathrm{pos}$ we use an Random Fourier Feature Network \cite{tancik2020fourier}, due to explicit control over the smoothness through the initial frequency parameter $\Omega$, and set it to $\Omega=0.1$. The grid network consists of 3 convolution blocks with 128 channels. The practical setup and instantiation of the convolution blocks can be found in Appendix \ref{app:networkarch}. Through hyperparameter sweeps we set $k=9$ in the forward and backward knn-connectivity. We found that for most experiments using $L=4$ convolution blocks is sufficient. 

\begin{figure}
    \centering
    \includegraphics[width=0.65\textwidth]{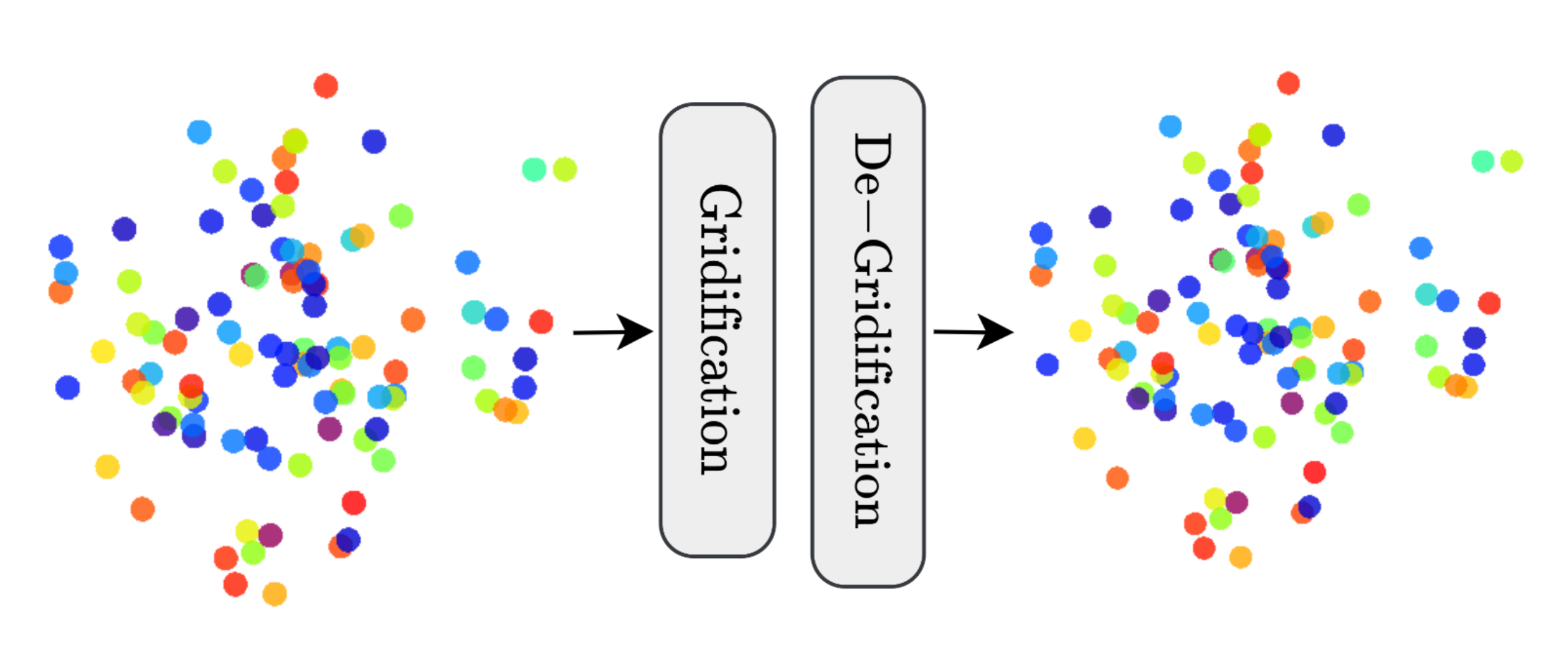}
    \vspace{-4mm}
    \caption{Random point clouds with random scalar node features are mapped to a grid representation and back via (de)-gridification.
    \vspace{-4mm}}
    \label{fig:recon_pipeline}
\end{figure}

% Each object is centered and normalized.  channels work best. For the position embedding function we set the smoothness parameter $\Omega=1.0$. 
\vspace{-7mm}
\subsection{Random point cloud reconstruction}
First, we evaluate the expressivity of our proposed gridification and de-gridification procedure. To this end, we construct a dataset with 1000 synthetic random graphs --800 for training and 200 for validation-- consisting of a predefined number of nodes $\Nt^\gP{=}1000$ randomly sampled on the unit cube, i.e., $\cv_i^\gP \sim \gU( [-1, 1]^3)$, accompanied with a random scalar feature $f_i^\gP \sim \gU(-1, 1)$ at each position. 

\textbf{Experimental setup.} To evaluate the expressiveness of our method, we set up a network consisting only of a gridification and a de-gridification step, i.e., no intermediary layers, in a point cloud reconstruction pipeline. In other words, the task consists of propagating the point cloud into a grid representation, and mapping the grid representation back to the original point cloud (see Fig. \ref{fig:recon_pipeline}). %Specifically, from the grid representation the task is to predict the point cloud features given the node positions. 
Therefore, to successfully reconstruct the original point cloud from the grid representation, the grid representation must be able to retain sufficient information from the input point cloud. 

\begin{figure}
    \centering
    \includegraphics[width=0.55\textwidth]{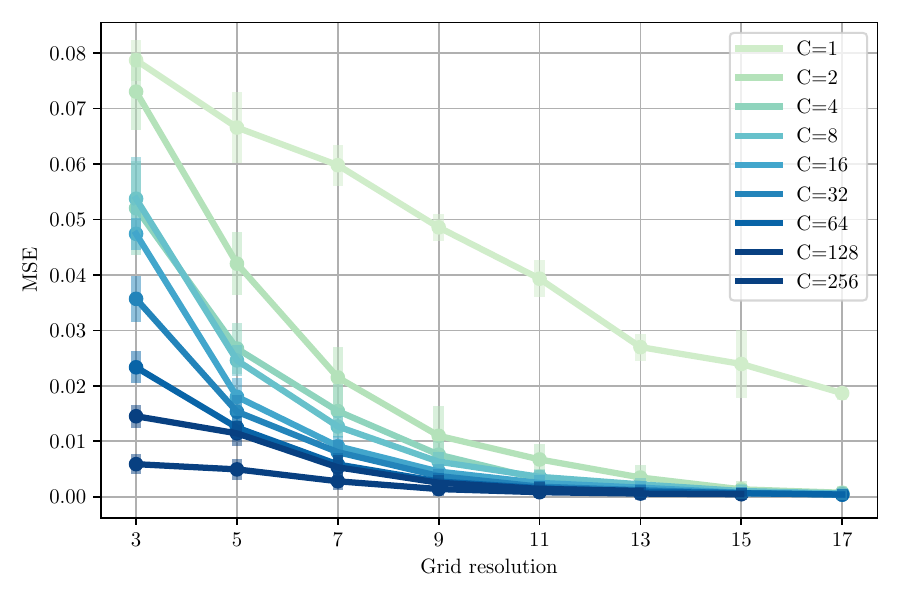}
    \vspace{-2mm}
    \caption{Random point cloud reconstruction error for varying grid resolution and number of channels on the grid representation.}
    \label{fig:recon}
\end{figure}

\textbf{Results.} Fig. \ref{fig:recon} shows reconstruction errors for different resolutions and different number of channels in the intermediary grid representation. We observe that it is possible to obtain good reconstructions by increasing the resolution of the grid or its number of channels. From an efficiency perspective, it is preferred to utilize low resolution representations with a larger number of channels due to the exponential growth in computational demands associated with higher grid resolutions, which instead scale linearly with the number of channels of the representation. Our experiments show that gridification is able to obtain compact grid representations that preserve the structure of the input point cloud. Furthermore, the quality of the grid representations can be efficiently improved by scaling the number of channels used. 
\vspace{-7mm}
\subsection{ModelNet40 classification} \label{sec:modelnet40exp}
Next, we evaluate gridification on point cloud classification. We deploy gridified networks on ModelNet40 \cite{wu20153d}: a synthetic dataset for 3D shape classification, consisting of 12,311 3$\Dt$ meshes of objects belonging to 40 classes. ModelNet40 is broadly used as a point cloud benchmark in which points are uniformly sampled from meshes.

\textbf{Results.} Our results (Tab.~\ref{tab:modelnet_results}) show that gridified networks achieve competitive performance while being significantly more efficient in terms of parameters, compute and memory. Interestingly, and in contrast to voxelization methods, we observe that gridified networks operate well even on extremely low resolution grids. For instance, on a $3{\times}3{\times}3$ grid, gridified networks attain an accuracy of $90.86\%$.

\vspace{-7mm}
\subsection{ShapeNet part segmentation} 
Next, we evaluate gridification on point cloud segmentation. To this end, we deploy gridified networks on ShapeNet \cite{shapenet2016}: a synthetic dataset with 16,000 point clouds of objects from 16 categories, each of which contains 2 to 6 parts. The objective of the task is to segment the point clouds into one of 50 possible part annotations.

\textbf{Results.} Our results (Tab.~\ref{tab:shapenet_results}) demonstrate that gridified networks are also able to achieve competitive performance in segmentation tasks, while being significantly more efficient in terms of parameters, compute and memory. This result validates the ability of gridification to handle dense prediction tasks via gridification and de-gridification.
\begin{table}
\centering
    \begin{minipage}{\textwidth}
    \centering
    \caption{Segmentation performance on ShapeNet-part benchmark.}
    \label{tab:shapenet_results}
    \vspace{-2.5mm}
    \begin{small}
    \scalebox{0.85}{
    \begin{tabular}{l|lll}
    \toprule
    \sc{Model} & Gridified Networks  & PointNet++  & PointGrid \\
    \midrule
   \sc{Type} & voxelization & native & hybrid
    \\
     \midrule
     instance average IoU & 87.07 & 85.1 & 86.4\\
     class average IoU & 81.68 & 81.9 & 82.2\\
     \midrule
airplane & 88.52  & 82.4 & 85.7\\
bag & 86.54  &  79.0 & 82.5 \\
cap & 74.09  & 87.7 & 81.8 \\
car & 80.46  & 77.3 & 77.9 \\
chair & 91.44  & 90.8 & 92.1 \\
earphone & 51.81 & 71.8 &  82.4 \\
guitar & 92.61 &  91.0 &  92.7 \\
knife & 89.44  & 85.9 & 85.8 \\
lamp & 82.07  &  83.7 & 84.2 \\
laptop & 96.07  &  95.3 & 95.3 \\
motor & 65.36  & 71.6 & 65.2 \\
mug & 92.99  &  94.1 & 93.4 \\
pistol & 86.72 & 81.3  & 81.7 \\
rocket & 58.57   & 58.7 & 56.9 \\
skateboard & 75.70   & 76.4 & 73.5 \\
table & 85.66 &  82.6 & 84.6 \\
    \bottomrule
    \end{tabular}}
    \end{small}
    \end{minipage}
\end{table}

\vspace{-7mm}
\subsection{Efficiency analysis of gridification} 
Finally, we investigate the scalability properties of gridification. Specifically, we analyze the time and memory consumption of gridified networks during inference on ModelNet40 for point clouds with increasing size, and compare the computation and memory complexity of convolutional operations on grid and point cloud data.

\begin{figure*}
\centering
\hfill
\includegraphics[width=0.46\linewidth]{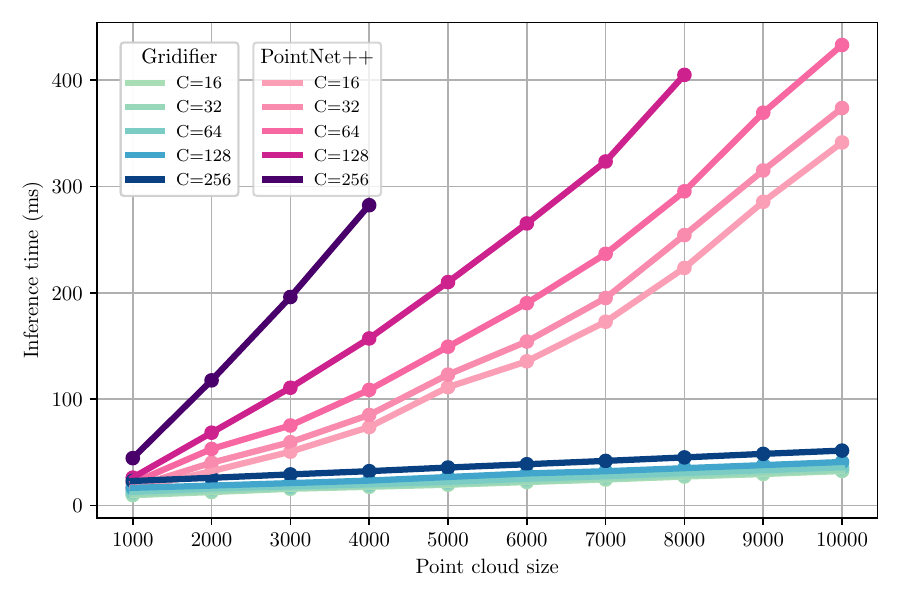}\hfill
\hspace{2mm}
\includegraphics[width=0.46\linewidth]{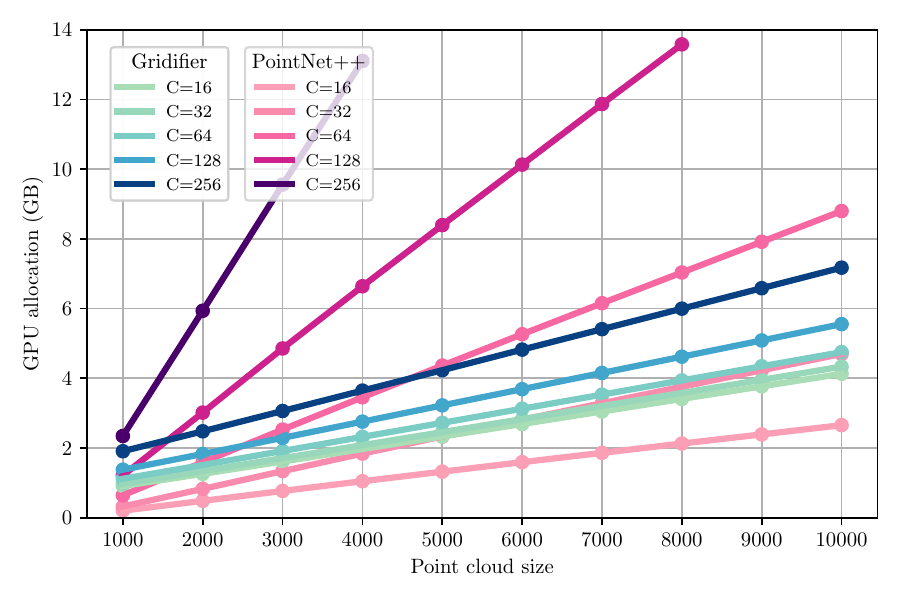}\hfill
\vspace{-2mm}
\caption{Average time (left) and GPU allocation (right) during inference on ModelNet40 for a batch size of $32$.
\vspace{-4mm}}
\label{fig:efficiency_modelnet}
\end{figure*}

\textbf{Scaling gridified networks to large point clouds.} Fig~\ref{fig:efficiency_modelnet} shows the average time and memory consumption during inference on ModelNet40 for gridified networks and PointNet++. We observe that gridified networks exhibit a much more favorable scalability both in terms of inference time and GPU allocation --linear vs. quadratic-- as the input size and number of channels increase. This demonstrates that gridified networks scale much better than native point cloud methods both for larger point clouds and networks. 

\textbf{Scaling the receptive field of neural operations.} Furthermore, we analyze the scalability properties of gridified networks relative to the size of its receptive fields. As illustrated in Fig.~\ref{fig:convs}), for native point cloud methods the convolutional kernel must be recomputed for all query points in the point cloud. As a consequence, the construction of the convolutional kernels of size $\Kt$ for all query points in a point cloud with $\Nt$ points incurs in $\gO(\Kt \Dt)$ memory and time complexity. In contrast, on grid data, we can compute the kernel once and reuse it at all positions. As a result, on a grid, this operation incurs in $\gO(\Dt)$ time and memory complexity.

\begin{figure*}
\centering
\hfill
\includegraphics[width=0.46\linewidth]{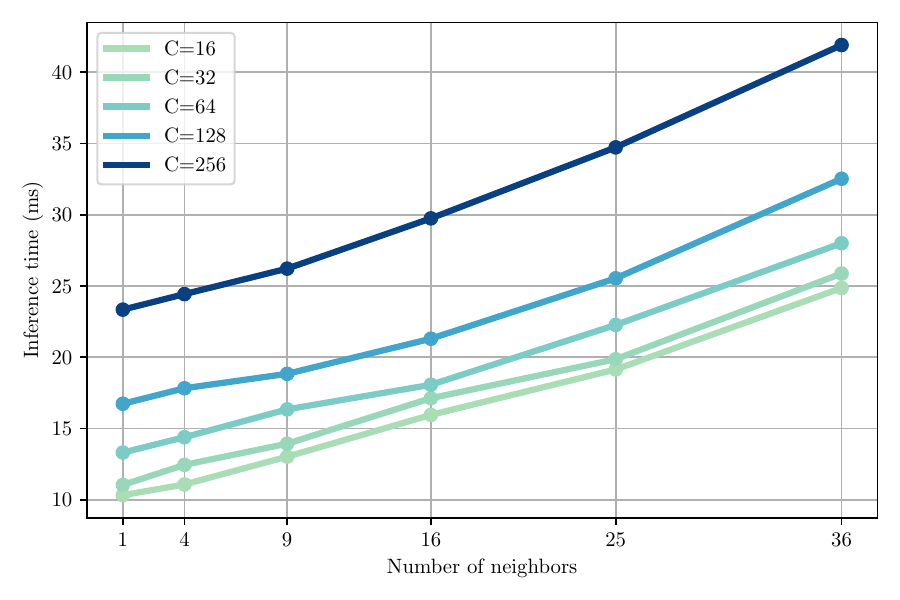}\hfill
\hspace{2mm}
\includegraphics[width=0.5\linewidth]{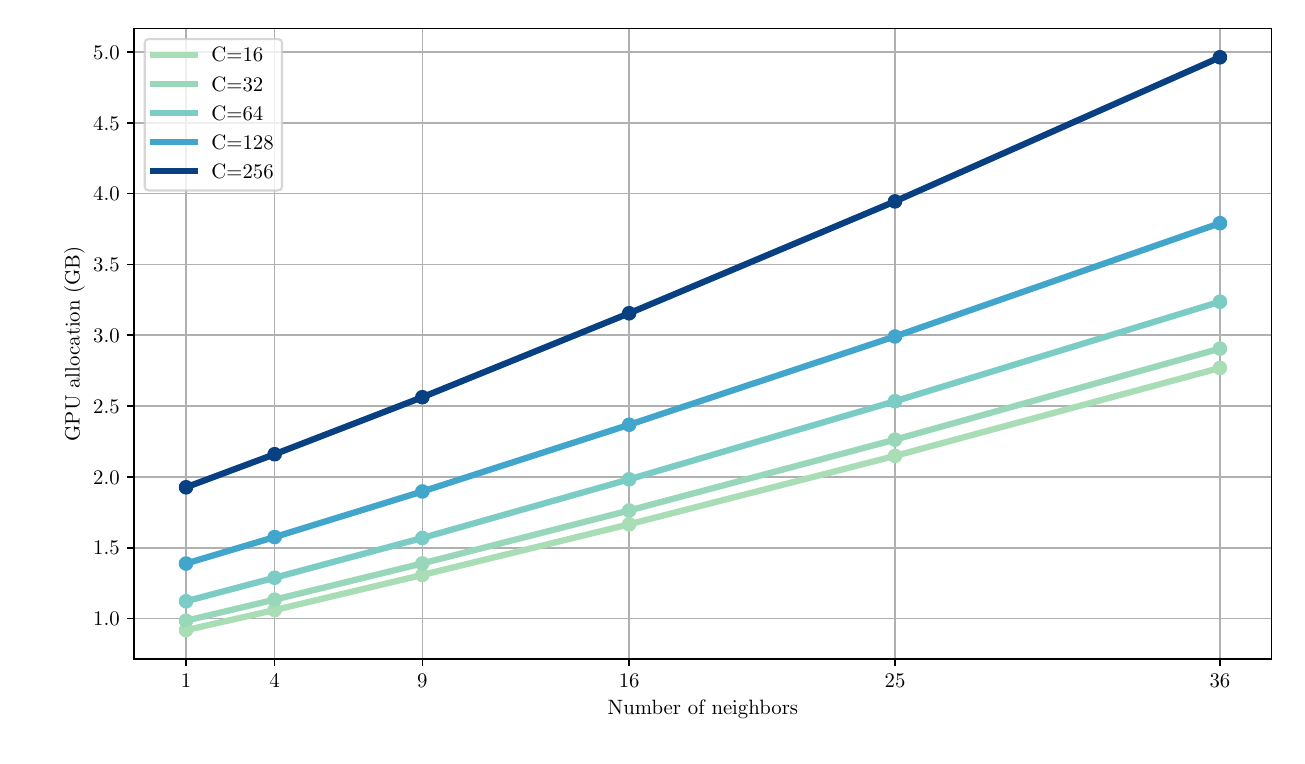}
\vspace{-2mm}
\caption{Average time (left) and GPU allocation (right) on ModelNet40 validation set per batch $B=32$ for various number of neighbors and number of channels $C$ on the grid representation.}
\label{fig:efficiency_neighbors}
\end{figure*}

Fig. \ref{fig:efficiency_neighbors} show the methods' potential to scale up the receptive field of the gridification module without introducing significant computational overhead.

\vspace{-7mm}
\section{Limitations and future work}
\textbf{The resolution of gridification depends on the size of the point cloud.} The main limitation of gridification is that the resolution on the grid is directly proportional to the size of point cloud in order to preserve information. This in turn means that the whole gridified architecture must be changed for point clouds of different sizes, even if they represent the same underlying signal. This is in contrast to native point cloud methods, which, due to their continuous nature, are, in principle, able to generalize to point clouds of different sizes as long as these exhibit the same structures.%    due to $k$-nearest neighbor connectivity scheme, the receptive field of each grid varies depending on the input point cloud density. Consequently, the number of input points in the point cloud implicitly defines the learned resolution of the network: the network has to be retuned for a different number of input points.
% \vspace{-3mm}
% \section{Future Work}
% \vspace{-1mm}
% We identify several future research directions for gridification in the context of point cloud processing.

\textbf{Towards no information loss.}  While gridification aims to produce compact grid representations with minimal information loss, our experiments reveal that some information still gets lost in the process. Loosely speaking, it should be possible to create grid representations that do not lose any information by ensuring that the grid representation has at least as many points and as many channels as the source point cloud representation. Gaining richer theoretical understanding of gridification, could therefore lead to grid representations with no information loss either by imposing other requirements on gridification, or by considering different functional families in the gridification process.

\textbf{Large scale point clouds and global context.} While we verify the scalability and efficiency of gridified networks for increasing point cloud sizes, we only carry on experiments on relatively small datasets. In future work, we aim to deploy gridification to large scale datasets. Furthermore, recent works have shown that using global receptive fields in convolutional operations consistently leads to better results across several tasks, even outperforming well-established Transformer architectures \cite{gu2022efficiently, knigge2023modelling, poli2023hyena}. Due to the computational complexity of native point cloud methods, networks with global context have not been explored for point cloud processing. With gridification this ability becomes computationally feasible. Exploring the effect of global context for point cloud processing is an exciting research direction.

\textbf{Generative tasks.} Gridification opens up the possibility of performing scalable generative tasks on large point clouds. Gridification can directly be extended to generative tasks if we assume that the point-cloud structure is preserved, i.e., if the coordinates of the output and input point clouds are equal. If this is not the case, e.g., for the generation of molecules \cite{xu2019deep, hoogeboom2022equivariant}, de-gridification module must be modified to predict both the features and positions of the new point cloud.% This  a particularly promising.% direction for future research. 

\textbf{Equivariant gridification.} In its current form, gridified networks do not respect symmetries which might be important for some applications, e.g., equivariance to 3D rotations for the prediction and generation of molecules \cite{schutt2017schnet, hoogeboom2022equivariant}. In future work, we aim to extend gridification to respect these symmetries by taking inspiration from equivariant graph neural networks \cite{fuchs2020se, satorras2021n}. It is important to note that not only gridification and de-gridification must be equivariant, but also that the grid operations in between. This can be achieved in an efficient yet expressive manner through the use of continuous Monte-Carlo convolutions on the regular representations of the group \cite{finzi2020generalizing, romero2022learning}.
% \textbf{Compactness of grid representation.} Even though the bilateral K-nearest neighbor connectivity scheme ensures every grid point is non-empty, it does not prevent neighboring grid points from potentially having identical neighborhoods, and thus having similar aggregate representations. The grid may be more optimally populated with feature values if we could ensure each grid point covers a diverse neighborhood.

% \textbf{Compactness of connectivity}. For simplicity, in the current work the amount of neighbours in the forward and backward connectivity scheme is set equally. In practice, the number of forward and backward edges can be different and are hyperparameters. Reducing the amount of edges in the forward or backward case could cut down on the bottleneck induced by the connectivity, which scales poorly with the number of grid points.

\vspace{-7mm}
\section{Conclusion}
This work presents gridification, a method that strongly reduces the computational requirements of point cloud processing pipelines by mapping input point clouds to a grid representation, and performing neural operations in there. We demonstrate that gridified networks are able to match the accuracy of native point cloud methods, while being much faster and memory efficient. Through empirical and theoretical analyses, we also show that gridified networks scale much more favorably than native point cloud methods to larger point clouds and larger neighborhoods.
%=========================================================================

%% file: chapters/3_flexconv.tex
% path to figures directory
\graphicspath{{figures/2-ckconv/}}

%=========================================================================

% \begin{savequote}[75mm]
% Nulla facilisi. In vel sem. Morbi id urna in diam dignissim feugiat. Proin molestie tortor eu velit. Aliquam erat volutpat. Nullam ultrices, diam tempus vulputate egestas, eros pede varius leo.
% \qauthor{Quoteauthor Lastname}
% \end{savequote}

\chapter{Continuous Kernel Convolutions with Differentiable Kernel Sizes}
	\label{chapter:flexconv}

\begin{flushright}
\textit{Based on the paper:}\break
\textit{FlexConv: Continuous Kernel Convolutions with Differentiable Kernel Sizes \citep{romero2022flexconv}}
\end{flushright}
%=========================================================================
\vspace{-7mm}
\section{Introduction}

The kernel size of a convolutional layer defines the region from which features are computed, and is a crucial choice in their design. % of a convolutional layer is the size of its kernel. 
Commonly, small kernels (up to 7px) are used almost exclusively and are combined with pooling to model long term dependencies \citep{simonyan2014very, szegedy2015going, he2016deep,  tan2019efficientnet}. Recent works indicate, however, that CNNs benefit from using convolutional kernels (\emph{i}) of varying size at different layers \citep{pintea2021resolution, tomen2021deep}, and (\emph{ii}) at the same resolution of the data \citep{peng2017large, Cordonnier2020On, romero2022ckconv}. Unfortunately, most CNNs represent convolutional kernels as tensors of discrete weights and their size must be fixed prior to training. This makes exploring different kernel sizes at different layers difficult and time-consuming due to (\textit{i}) the large search space, and (\textit{ii}) the large~number~of~weights~required~to~construct~large~kernels.

\begin{figure}
    \centering
     \begin{subfigure}[c]{0.28\textwidth}
         \centering
         \includegraphics[width=\textwidth]{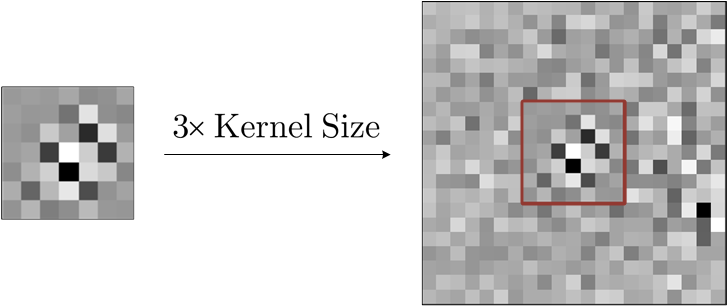}
         \caption{FlexConv kernels (ours)}
         \label{fig:3_flexconv_kernel}
     \end{subfigure}
     \hfill
     \begin{subfigure}[c]{0.28\textwidth}
         \centering
         \includegraphics[width=\textwidth]{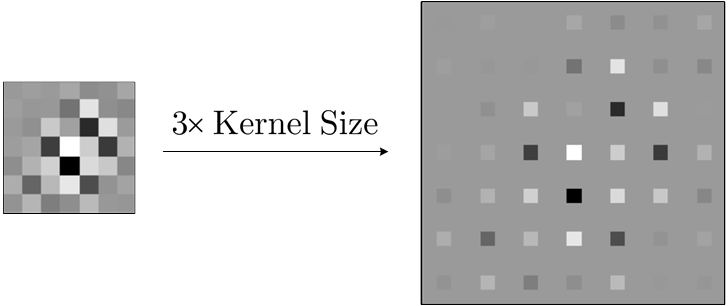}
         \caption{Dilation / deformation}
         \label{fig:3_dilated_kernel}
     \end{subfigure}
     \hfill
      \begin{subfigure}[c]{0.36\textwidth}
     \centering
     \includegraphics[width=\textwidth]{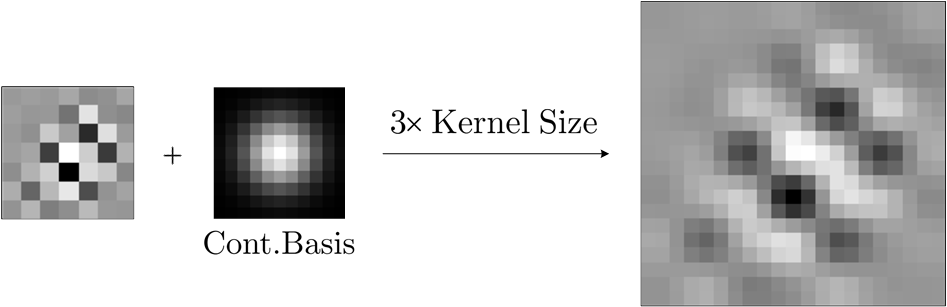}
     \caption{(Learnable) parametric dilation}
     \label{fig:3_parametric_dilation}
     \end{subfigure}
    \vspace{-2.5mm}
    \caption{Existing approaches increase the size of convolutional kernels via (learnable) parametric dilations, e.g., by deformation \citep{dai2017deformable} (b) or by Gaussian blur \citep{pintea2021resolution} (c). % or by pixel offsets (Fig.~\ref{fig:3_dilated_kernel}). 
    However, dilation limits the bandwidth of the dilated kernel and with it, the amount of detail it can describe. %This is due to the fact that the underlying discrete parameterization of the kerrnel is defined only at a fixed set of spatial positions. 
    Contrarily, FlexNets extend their kernels by passing a larger vector of positions to the neural network parameterizing them. As a result, FlexConvs are able to learn \textit{high bandwidth} convolutional kernels of varying size end-to-end (a).}
    \label{fig:3_dilations}
\end{figure}

A more efficient way to tune different kernel sizes at different layers is to \textit{learn} them during training.
Existing methods define a \textit{discrete} weighted set of basis functions, e.g., shifted Delta-Diracs (Fig.~\ref{fig:3_dilated_kernel}, \citet{dai2017deformable}) or Gaussian functions (Fig.~\ref{fig:3_parametric_dilation}, \citet{jacobsen2016structured, Shelhamer2019BlurringTL, pintea2021resolution}). During training they learn dilation factors over the basis functions to increase the kernel size, which crucially limits the bandwidth of the resulting kernels.
% Problematic with these solutions is that dilations implicitly reduce the resolution of the kernels, and with it, the amount of detail the kernels can describe. As a result, regardless of the basis used, the amount of detail these kernels can describe is determined by the tensor of free-form independent weights that combines the dilated bases (Fig.~\ref{fig:3_parametric_dilation}).

\begin{figure}[t]
    \centering
    \includegraphics[width=0.88\textwidth]{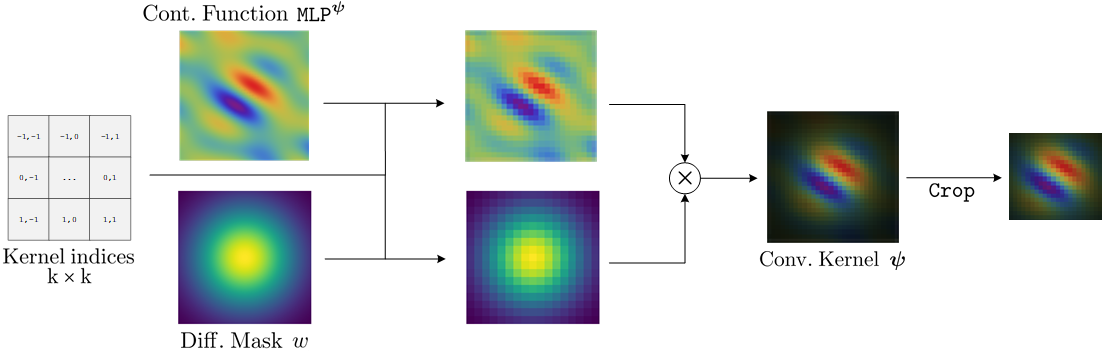}
    \caption{The Flexible Size Continuous Kernel Convolution (FlexConv). FlexConv defines convolutional kernels as the multiplication of a continuous convolutional kernel \mlppsi, with a Gaussian mask of local support $w_{\textrm{gauss}}$: $\boldsymbol{\psi}(x, y) = w_{\textrm{gauss}}( x, y ; \boldsymbol{\theta}_{\mathrm{mask}}) \cdot \boldsymbol{\text{\btt MLP}^{\psi}}(x, y)$. By learning the parameters of the mask, the size of the convolutional kernel can be optimized during training. See also Fig.~\ref{fig:3_app-flexconvexample}.
    \vspace{-2mm}}
    \label{fig:3_flexconv}
\end{figure}

In this work, we present the \textit{Flexible Size Continuous Kernel Convolution} (FlexConv), a convolutional layer able to learn \textit{high bandwidth} convolutional kernels of varying size during training (Fig.~\ref{fig:3_flexconv}). Instead of using discrete weights, we provide a \textit{continuous parameterization} of convolutional kernels via a small neural network \citep{romero2022ckconv}. This parameterization allows us to model continuous functions of arbitrary size with a fixed number of parameters. By multiplying the response of the neural network with a Gaussian mask, the size of the kernel can be learned during training (Fig.~\ref{fig:3_flexconv_kernel}). This~allows~us~to~produce~detailed~kernels~of~small~sizes~(Fig.~\ref{fig:3_capacity_tradeoff}),~and~tune~kernel~sizes~efficiently.
% To the best of our knowledge, FlexConv is the first method able to learn high bandwidth convolutional kernels of varying size end-to-end.
%Additionally, restricting the kernel area to be learned by continuous kernel parameterization allows it to achieve higher resolution in small kernels, as illustrated in Fig.~\ref{fig:3_capacity_tradeoff}.

FlexConvs can be deployed at higher resolutions than those observed during training, simply by using a more densely sampled grid of kernel indices. However, the high bandwidth of the kernel can lead FlexConv to learn kernels that show aliasing at higher resolutions, if the kernel bandwidth exceeds the Nyquist frequency. 
To solve this problem, we propose to parameterize convolutional kernels as \textit{Multiplicative Anisotropic Gabor Networks} (MAGNets). MAGNets are a new class of Multiplicative Filter Networks \citep{fathony2021multiplicative} that allows us to analyze and control the frequency spectrum of the generated kernels. We use this analysis to regularize FlexConv against aliasing. With this regularization, FlexConvs can be directly deployed at higher resolutions with minimal accuracy loss. Furthermore, MAGNets provide higher descriptive power and faster convergence speed than existing continuous kernel parameterizations \citep{schutt2017schnet, finzi2020generalizing, romero2022ckconv}. This leads to important improvements in classification accuracy (Sec.~\ref{sec:3_experiments}).%Aside from increased expressivity, and faster convergence speed

% We additionally improve the continuous kernel parameterization provided by \citet{romero2022ckconv}. We replace their SIREN \citep{sitzmann2020implicit} kernel parameterization with our proposed \textit{Multiplicative Anisotropic Gabor Networks (MAGNets)}: a novel implicit neural representation method with increased descriptive power, faster convergence speed, and the ability to \textit{analytically control} the properties of the approximation. We use the latter to analytically control the frequency components of MAGNets and prevent aliasing. % a novel problem introduced by the unconstrained nature of the frequency spectrum of MAGNets. 
% The resulting MAGNets generate convolutional kernels that generalize well across resolutions. As a result, FlexConvs can be reliably deployed at higher resolutions.

% \textit{Flexible Size Continuous Kernel CNNs} (FlexNets) learn the size of their convolutional kernels at every layer, and easily model long-term dependencies without the need of pooling.
Our experiments show that CNNs with FlexConvs, coined \textit{FlexNets}, achieve state-of-the-art across several sequential datasets, match performance of recent works with learnable kernel sizes with less compute, and are competitive with much deeper ResNets \citep{he2016deep} when applied on image benchmark datasets. Thanks to the ability of FlexConvs to generalize across resolutions, FlexNets can be efficiently trained at low-resolution to save compute, e.g., $16\times16$ CIFAR images, and be deployed on the original data resolution with marginal accuracy loss, e.g., $32\times32$ CIFAR images.% This training scheme results in important compute savin

In summary, our \textbf{contributions} are:
\begin{itemize}
    \item We introduce the \textit{Flexible Size Continuous Kernel Convolution} (FlexConv), a convolution operation able to learn \emph{high bandwidth} convolutional kernels of varying size end-to-end. 
    \item Our proposed \textit{Multiplicative Anisotropic Gabor Networks} (MAGNets) allow for analytic control of the properties of the generated kernels. This property allows us to construct analytic alias-free convolutional kernels that generalize to higher resolutions, and to train FlexNets at low resolution and deploy them at higher resolutions. %, so we can train networks at lower resolutions to save compute. 
    Moreover, MAGNets show higher descriptive power and faster convergence speed than existing kernel parameterizations.
    \item CNN architectures with FlexConvs (FlexNets) obtain state-of-the-art across several sequential datasets, and match recent works with learnable kernel size on CIFAR-10 with less compute.
    % FlexNets can be deployed across resolutions with minimal loss of precision.
    % Cross-resolution fine-tuned FlexNets match the performance of FlexNets trained on the original data resolution from scratch after limited fine-tuning.
\end{itemize}
% \begin{figure}
% \vspace{-2mm}
%     \centering
%      \begin{subfigure}[c]{0.375\textwidth}
%      \vspace{2.5mm}
%          \centering
%          \includegraphics[width=\textwidth]{images/discrete_conv.png}
%          \caption{Discrete Parameterization}
%          \label{fig:3_discrete_conv}
%      \end{subfigure}
%      \hfill
%      \begin{subfigure}[c]{0.58\textwidth}
%          \centering
%          %\includegraphics[width=\textwidth]{images/continuous_conv.png}
%          \includegraphics[width=\textwidth]{images/Figure1DiscreteVsContinuous.pdf}
%          \vskip -0.7cm
%          \caption{Continuous Parameterization}
%          \label{fig:3_continuous_conv}
%      \end{subfigure}
%      \vspace{-0.5mm}
%     \caption{Discrete and continuous parameterizations. Discrete parameterizations approximate functions with independent weights on a fixed grid, e.g., $\mathrm{5x5}$ in (a). Contrarily, continuous parameterizations provide a continuous approximation, e.g., via a small neural network \mlppsi. This approximation can be sampled at different resolutions by passing a grid of positions at the corresponding resolution.
%     \vspace{-2mm}}
%     \label{fig:3_continuous_param}
% \end{figure}
% ----------------- Section ---------------------------
\vspace{-7mm}
\section{Related Work}

\textbf{Adaptive kernel sizes.} \citet{loog2017supervised} regularize the scale of convolutional kernels for filter learning. For image classification, adaptive kernel sizes have been proposed via learnable pixel-wise offsets \citep{dai2017deformable}, learnable padding operations \citep{Han_2018_CVPR}, learnable dilated Gaussian functions \citep{Shelhamer2019BlurringTL, Xiong_2020_CVPR, tabernik2020spatially, nguyen2020robust} and scalable Gaussian derivative filters \citep{pintea2021resolution, tomen2021deep, lindeberg2021scale}.
These approaches either dilate discrete kernels (Fig.~\ref{fig:3_dilated_kernel}), or use discrete weights on dilated basis functions (Fig.~\ref{fig:3_parametric_dilation}). Using dilation crucially limits the bandwidth of the resulting kernels. In contrast, FlexConvs are able to construct high bandwidth convolutional kernels of varying size with a fixed parameter count. 
Larger kernels are obtained simply by passing more positions to the kernel network (Fig.~\ref{fig:3_flexconv}). 

\textbf{Continuous kernel convolutions.} Discrete convolutional kernel parameterizations assign an independent weight to each specific position in the kernel. Continuous convolutional kernels, on the other hand, view convolutional kernels as continuous functions parameterized via a small neural network \mlppsi$: \sR^{\mathrm{D}} \rightarrow \sR^{\mathrm{N}_{\mathrm{out}} \times \mathrm{N}_{\mathrm{in}}}$, with $\mathrm{D}$ the data dimensionality. This defines a convolutional kernel for which arbitrary input positions can be queried. Continuous kernels have primarily been used to handle irregularly-sampled data \textit{locally}, e.g., molecular data \citep{simonovsky2017dynamic, schutt2017schnet} and point-clouds \citep{thomas2018tensor, wang2018deep, shi2019points}. 

\begin{figure}
    \centering
     \begin{subfigure}[c]{0.267\textwidth}
         \centering
         \captionsetup{justification=centering}
         \includegraphics[width=\textwidth]{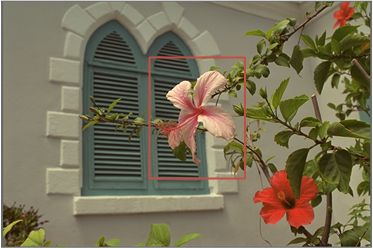}
         \caption{Ground Truth}
         \label{fig:3_gt_approx}
     \end{subfigure}
     \hspace{3.5mm}
     \begin{subfigure}[c]{0.677\textwidth}
         \centering
         \captionsetup{justification=centering}
         \includegraphics[width=\textwidth]{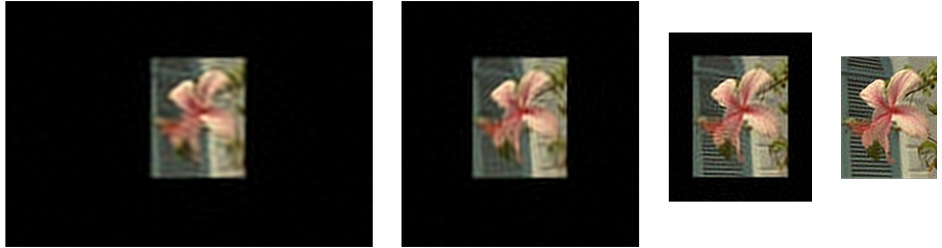}
         \caption{Reconstructions at varying degrees of localization}
         \label{fig:3_ckernel_approx}
     \end{subfigure}
     \vspace{-2mm}
     \caption{\label{fig:3_capacity_tradeoff} The importance of dynamic sizes in continuous kernel convolutions. Consider a neural network predicting pixel values at each position. If the entire image is considered, the network must use part of its capacity to learn to predict zeros outside of the flower region, which in turn degrades the quality of the approximation in the region of interest (b). Importantly, the better the localization of the flower, the higher the approximation fidelity becomes. FlexNets learn the size of their convolutional kernels at each layer during training, and thus \emph{(i)} use the capacity of the kernel efficiently, \emph{(ii)} converge faster to good approximations, and \emph{(iii)} are faster in execution.}
\end{figure}

Recently, \citet{romero2022ckconv} introduced the Continuous Kernel Convolution (CKConv) as a tool to model long-term dependencies. CKConv uses a continuous kernel parameterization to construct convolutional kernels as big as the input signal with a constant parameter cost. % This saves parameters compared to equivalent discrete kernels.
%CKConvs always use kernels as big as the input signal. 
Contrarily, FlexConvs jointly learn the convolutional kernel as well as its size. This leads to important advantages in terms of~expressivity~(Fig.~\ref{fig:3_capacity_tradeoff}),~convergence~speed~and~compute~costs~of~the~operation.

\textbf{Implicit neural representations.} Parameterizing a convolutional kernel via a neural network can be seen as learning an implicit neural representation of the underlying convolutional kernel \citep{romero2022ckconv}. Implicit neural representations construct continuous data representations by encoding data in the weights of a neural network \citep{park2019deepsdf, sitzmann2020implicit, fathony2021multiplicative}. 

We replace the SIREN \citep{sitzmann2020implicit} kernel parameterization used in \citet{romero2022ckconv} by our \textit{Multiplicative Anisotropic Gabor Networks}: a new class of Multiplicative Filter Networks \citep{fathony2021multiplicative}. MFNs allow for analytic control of the resulting representations, and allow us to construct analytic alias-free convolutional kernels. The higher expressivity and convergence speed of MAGNets lead to accuracy improvements in CNNs using them as kernel parameterization.% with this kernel parameterizations.%In addition, MAGNets show higher expressivity and convergence rate than SIRENs and MFNs, and lead to important improvements in classification accuracy.
% ----------------- Section ---------------------------
\vspace{-7mm}
\section{Method}

In this section, we introduce our approach. First, we introduce FlexConv and the Gaussian mask. Next, we introduce our Multiplicative Anisotropic Gabor Networks (MAGNets) and provide a description of our regularization technique used to control the spectral components of the generated kernel.
% ----------------- Subsection ---------------------------
\vspace{-7mm}
\subsection{Flexible Size Continuous Kernel Convolution (FlexConv)}
\label{sec:3_flexconv}

To learn the kernel size during training, FlexConvs define their convolutional kernels $\boldsymbol{\psi}$ as the product of the output of a neural network \mlppsi\ with a Gaussian mask of local support. The neural network \mlppsi\ parameterizes the kernel, and the Gaussian mask parameterizes its size (Fig.~\ref{fig:3_flexconv}).

\textbf{Anisotropic Gaussian mask.} Let $G(x ; \mu_{\Xt}, \sigma^2_{\Xt}) \coloneq \exp\big\{\hspace{-0.5mm}-\frac{1}{2}\sigma_\Xt^{-2}(x - \mu_{\Xt})^{2}\big\}$ 
be a Gaussian function parameterized by a mean-variance tuple $(\mu_{\Xt}, \sigma^2_{\Xt})$. The anisotropic Gaussian mask is defined as:
% \newpage
\begin{equation}
    w_{\mathrm{gauss}}(x, y; \{\mu_{\Xt}, \sigma^2_{\Xt}, \mu_{\Yt}, \sigma^2_{\Yt}\}) = G(x ; \mu_{\Xt}, \sigma_{\Xt}^2) \cdot G(y ; \mu_{\Yt}, \sigma_{\Yt}^2). \label{eq:3_gaussianmask}
\end{equation}
By learning $(\mu_{\Xt}, \sigma^2_{\Xt})$ and $(\mu_{\Yt}, \sigma^2_{\Yt})$, anisotropic non-centered windows are learned.
% ----------------- Subsection ---------------------------

\vspace{-7mm}
\subsection{Multiplicative Anisotropic Gabor Networks (MAGNets)}
\label{sec:3_magnets}

In this section, we formalize our proposed parameterization for the kernel \mlppsi. We introduce Multiplicative Filter Networks \citep{fathony2021multiplicative} and present MAGNets next.

\textbf{Multiplicative Filter Networks (MFNs).} Recently, \citet{fathony2021multiplicative} proposed to construct implicit neural representations as the linear combination of exponentially many basis functions $\boldsymbol{\mathrm{g}}$:
\begin{align}
    \label{eq:3_mfn}
    &\boldsymbol{\mathrm{h}}^{(1)} = \boldsymbol{\mathrm{g}}\big( [x,y]; \boldsymbol{\theta}^{(1)}\big) && \boldsymbol{\mathrm{g}}: \sR^{2} \rightarrow \sR^{\Nt_{\mathrm{hid}}}\\
    &\boldsymbol{\mathrm{h}}^{(l)} = \big(\Wm^{(l)} \boldsymbol{\mathrm{h}}^{(l-1)} + \boldsymbol{\mathrm{b}}^{(l)}\big) \cdot   \boldsymbol{\mathrm{g}}\big( [x,y] ; \boldsymbol{\theta}^{(l)}\big) && \Wm^{(l)} \in \sR^{\Nt_{\mathrm{hid}} \times \Nt_{\mathrm{hid}}}, \boldsymbol{\mathrm{b}}^{(l)} \in \sR^{\Nt_{\mathrm{hid}}} \quad \\
    &\boldsymbol{\psi}(x, y) = \Wm^{(\mathrm{L})} \boldsymbol{\mathrm{h}}^{(\mathrm{L}-1)} + \boldsymbol{\mathrm{b}}^{(\mathrm{L})} && \Wm^{(\mathrm{L})} \in \sR^{\Nt \times \Nt_{\mathrm{hid}}}, \boldsymbol{\mathrm{b}}^{(\mathrm{L})} \in \sR^{\Nt}
\end{align}
where $\big\{\boldsymbol{\theta}^{(l)}, \W^{(l)}$, $\boldsymbol{\mathrm{b}}^{(l)}\big\}$ depict the learnable parameters of the bases and the affine transformations, and $\Nt, \Nt_{\mathrm{hid}}$ depict the number of output and hidden channels, respectively. Depending on the selection of $\boldsymbol{\mathrm{g}}$, MFNs obtain approximations comparable to those of SIRENs \citep{sitzmann2020implicit} with faster convergence rate. %\citet{fathony2021multiplicative} demonstrate that MFNs are equivalent to a linear combination of several basis functions $\boldsymbol{\mathrm{g}}$.% and thus their mathematical analysis is much simpler.
The most successful instantiation of MNFs are the \textit{Multiplicative Gabor Network} (MGN): MFNs constructed with isotropic Gabor functions as basis $\boldsymbol{\mathrm{g}}$ (in Eq.~\ref{eq:3_mfn}):
\begin{gather}
    \boldsymbol{\mathrm{g}}\big( [x,y] ; \boldsymbol{\theta}^{(l)}\big) = \exp \bigg(-\frac{\boldsymbol{\gamma}^{(l)}}{2} \Big[ \big(x-\boldsymbol{\mu}^{(l)}\big)^{2} + \big(y-\boldsymbol{\mu}^{(l)}\big)^{2}\Big] \bigg)\, \mathrm{Sin} \big(\Wm_\mathrm{g}^{(l)} \cdot [x, y] + \boldsymbol{\mathrm{b}}_\mathrm{g}^{(l)} \big), \\
    \boldsymbol{\theta}^{(l)} {=} \big\{ \boldsymbol{\gamma}^{(l)} \in \sR^{\Nt_{\mathrm{hid}}}, \boldsymbol{\mu}^{(l)} \in \sR^{\Nt_{\mathrm{hid}}},\Wm_\mathrm{g}^{(l)} \in \sR^{\Nt_{\mathrm{hid}} \times 2}, \boldsymbol{\mathrm{b}}_\mathrm{g}^{(l)} \in \sR^{\Nt_{\mathrm{hid}}} \big\}.
\end{gather}
Note that, by setting $\Nt{=}\Nt_{\textrm{out}}{\times} \Nt_{\textrm{in}}$, an MFN can parameterize a convolutional kernel with $\Nt_{\textrm{in}}$ input and $\Nt_{\textrm{out}}$ output channels.
\citet{fathony2021multiplicative} show that MFNs are equivalent to a linear combination of exponentially many basis functions $\boldsymbol{\mathrm{g}}$. This allows us to analytically derive properties of MFN representations, and plays a crucial role in the derivation of alias-free MAGNets (Sec.~\ref{sec:3_crtraining}).%, becau as this results facilitate theis an important propety% and thus their mathematical analysis is much simpler.

\textbf{Multiplicative Anisotropic Gabor Networks (MAGNets).} Our MAGNet formulation is based on the observation that isotropic Gabor functions, i.e., with equal $\gamma$ for the horizontal and vertical directions, are undesirable as basis for the construction of MFNs. Whenever a frequency is required along a certain direction, an isotropic Gabor function automatically introduces that frequency in both directions. As a result, other bases must counteract this frequency in the direction where the frequency is not required, and thus the capacity of the MFN is not used optimally \citep{daugman1988complete}. 

Following the original formulation of the 2D Gabor functions \citep{daugman1988complete}, we alleviate this limitation by using anisotropic Gabor functions instead:%, as is common \citep{Chao2010GaborWT, movellan}:
\begin{gather}
    \label{eq:3_anisotropicgaussian}
    \boldsymbol{\mathrm{g}}\big( [x,y] ; \boldsymbol{\theta}^{(l)}\big) = \exp \bigg(-\frac{1}{2} \Big[\Big(\boldsymbol{\gamma}^{(l)}_\mathrm{X}\big(x - \boldsymbol{\mu}_\mathrm{X}^{(l)}\big)\Big)^2 \hspace{-1mm}+ \Big(\boldsymbol{\gamma}_\mathrm{Y}^{(l)}\big(y - \boldsymbol{\mu}_\mathrm{Y}^{(l)}\big)\Big)^2\Big] \bigg)\, \mathrm{Sin} \big(\Wm_\mathrm{g}^{(l)} [x, y] + \boldsymbol{\mathrm{b}}_\mathrm{g}^{(l)} \big)\\
    \boldsymbol{\theta}^{(l)} {=} \Big\{ \boldsymbol{\gamma}_{\Xt}^{(l)}\in \sR^{\Nt_{\mathrm{hid}}}, \boldsymbol{\gamma}_{\Yt}^{(l)}\in \sR^{\Nt_{\mathrm{hid}}}, \boldsymbol{\mu}_{\Xt}^{(l)}\in \sR^{\Nt_{\mathrm{hid}}}, \boldsymbol{\mu}_{\Yt}^{(l)}\in \sR^{\Nt_{\mathrm{hid}} },\Wm_\mathrm{g}^{(l)}\in \sR^{\Nt_{\mathrm{hid}} \times 2}, \boldsymbol{\mathrm{b}}_\mathrm{g}^{(l)}\in \sR^{\Nt_{\mathrm{hid}}} \Big\}.\label{eq:3_params_anisotropicgaussian}
\end{gather}
The resulting \textit{Multiplicative Anisotropic Gabor Network} (MAGNet) obtains better control upon frequency components introduced to the approximation, and demonstrates important improvements in terms of descriptive power and convergence speed (Sec.~\ref{sec:3_experiments}).

\textbf{MAGNet initialization.} \citet{fathony2021multiplicative} proposes to initialize MGNs by drawing the size of the Gaussian envelopes, i.e., the $\boldsymbol{\gamma}^{(l)}$ term, from a $\mathrm{Gamma(}\alpha \cdot \mathrm{L}^{-1}, \beta{\mathrm{)}}$ distribution at every layer $l \in [1, .., \mathrm{L}-1]$. We observe however that this initialization does not provide much variability on the initial extension of the Gaussian envelopes and in fact, most of them cover a large portion of the space at initialization.
%This is undesirable as the network must learn to counteract undesirable frequency components across large portions of the approximation upon initialization. 
To stimulate diversity, we initialize the $\{\boldsymbol{\gamma}_{\Xt}^{(l)}, \boldsymbol{\gamma}_{\Yt}^{(l)}\}$ terms by a $\mathrm{Gamma(}\alpha l^{-1}, \beta{\mathrm{)}}$ distribution at the $l$-th layer. We observe that our proposed initialization consistently leads to better accuracy than the initialization of \cite{fathony2021multiplicative} across all tasks considered. (Sec.~\ref{sec:3_experiments}).
\vspace{-7mm}
\subsection{Analytic Alias-free MAGNets}
\label{sec:3_crtraining}

FlexConvs can be deployed at higher resolutions than those observed during training, simply by sampling the underlying continuous representation of the kernel more densely, and accounting for the change in sampling rate. Consider a $\mathrm{D}$-dimensional input signal $f_{\mathrm{r}^{(1)}}$ with resolution $\mathrm{r}^{(1)}$. FlexConv learns a kernel $\boldsymbol{\psi}_{\mathrm{r}^{(1)}}$ that can be inferred at a higher resolution $\mathrm{r}^{(2)}$ \citep{romero2022ckconv}:  
%As shown in Appx.~\ref{}, the convolution $(f_{\mathrm{r}^{(2)}} * \boldsymbol{\psi}_{\mathrm{r}^{(2)}})$ of a continuous $\mathrm{D}$-dimensional input signal with a continuous convolutional kernel at resolution $\mathrm{r}^{(1)}$ is \textit{approximately} equivalent to a convolution $(f_{\mathrm{r}^{(2)}} * \boldsymbol{\psi}_{\mathrm{r}^{(2)}})$ between the same input signal and convolutional kernel at a different resolution $\mathrm{r}^{(2)}$, up to a normalization factor given by the relative resolution change. That is:
\begin{equation}
    \Big(f_{\mathrm{r}^{(2)}} * \boldsymbol{\psi}_{\mathrm{r}^{(2)}}\Big) \approx \left(\frac{\mathrm{r}^{(1)}}{\mathrm{r}^{(2)}}\right)^{\mathrm{D}} \Big(f_{\mathrm{r}^{(1)}} * \boldsymbol{\psi}_{\mathrm{r}^{(1)}}\Big). \label{eq:3_multires}
\end{equation}
%In other words, a continuous convolutional kernel learned at one resolution can be deployed at a different resolution by sampling the kernel in a corresponding grid, and normalizing by the relative resolution change. 
Note however, that Eq.~\ref{eq:3_multires} holds \textit{approximately}. This is due to aliasing artifacts which can appear if the frequencies in the learned kernel surpass the Nyquist criterion of the target resolution. Consequently, an anti-aliased parameterization is vital to construct kernels that generalize well to high resolutions.% is crucial to learn convolutional kernels continuous functions to be deployed at different resolutions, that aliasing is avoided.

\textbf{Towards alias-free implicit neural representations.} We observe that SIRENs as well as unconstrained MFNs and MAGNets exhibit aliasing when deployed on resolutions higher than the training resolution, which hurts performance of the model. An example kernel with~aliasing~is~shown~in~Fig.~\ref{fig:3_app-cifar10kernelfrequencies}.
%A similar finding was reported in \citet{romero2022ckconv}.

To combat aliasing, we would like to control the representation learned by MAGNets. MAGNets --and MFNs in general-- construct implicit neural representations that can be seen as a \textit{linear combination of basis functions}. This property allows us to analytically derive and study the properties of the resulting neural representation. Here, we use this property to derive the maximum frequency of MAGNet-generated kernels, so as to regularize MAGNets against aliasing during training. We analytically derive the maximum frequency of a MAGNet, and penalize it whenever it exceeds the Nyquist frequency of the training resolution. We note that analytic derivations are difficult for other implicit neural representations, e.g., SIRENs, due to stacked layer-wise nonlinearities. 

\textbf{Maximum frequency of MAGNets.}
The maximum frequency of a MAGNet~is~given~by:
\begin{equation}
    f^+_{\textrm{MAGNet}} = \sum_{l=1}^{\mathrm{L}} \max_{i_l}\left( \left( \max_{j} \frac{\Wm^{(l)}_{\mathrm{g}, i_l,j}}{2 \pi} \right) + \frac{\sigma_\mathrm{cut} \min\{\boldsymbol{\gamma}^{(l)}_{\Xt, i_l}, \boldsymbol{\gamma}^{(l)}_{\Yt,i_l}\}}{2 \pi}\right), \tag{\ref{eq:3_magnetfreq}}
\end{equation}
where $\mathrm{L}$ corresponds to the number of layers, $\Wm^{(l)}_{\mathrm{g}}, \boldsymbol{\gamma}^{(l)}_{\Xt}, \boldsymbol{\gamma}^{(l)}_{\Yt}$ to the MAGNet parameters as defined in Eq.~\ref{eq:3_params_anisotropicgaussian}, and $\sigma_\mathrm{cut}{=}2\cdot \mathtt{stdev}$ to the cut-off frequency of the Gaussian envelopes in the Gabor filters. The derivation of this term is provided in Appx.~\ref{sec:3_magnetanalysis}.
%where $W$ and $\gamma$ are parameters of the MAGNet (Sec.~\ref{sec:3_magnets}) and $h$ is a hyperparameter controlling the accuracy of the effect of the Gaussian mask on the frequency. The analytic derivation of this result is provided in Appx.~\ref{sec:3_magnetanalysis}.

\textbf{Effect of the FlexConv mask.} The Gaussian mask used to localize the response of the MAGNet 
%The gaussian mask that attenuates the kernel as generated by the MAGNet 
also has an effect on the frequency spectrum. Hence, the maximum frequency of a FlexConv kernel is:
\begin{equation}
    f^+_{\textrm{FlexConv}} = f^+_{\textrm{MAGNet}} + f^+_{w_\textrm{gauss}}, \ \ \text{with}\ \  f^+_{w_\textrm{gauss}}=
    \frac{\sigma_\mathrm{cut}}{\max\{\sigma_\Xt, \sigma_\Yt\} 2 \pi}. \tag{\ref{eq:3_flexconvfreq}}
\end{equation}
Here, $\sigma_{\Xt}, \sigma_{\Yt}$ correspond to the mask parameters (Eq.~\ref{eq:3_gaussianmask}). Intuitively, multiplication with the mask blurs in the frequency domain, as it is equivalent to convolution with the Fourier transform of the mask.

\textbf{Aliasing regularization of FlexConv kernels.} With the analytic derivation of $f^+_{\textrm{FlexConv}}$ we penalize the generated kernels to have frequencies smaller or equal to their Nyquist frequency $f_{\mathrm{Nyq}}(k)$ via:
\begin{equation}
    \mathcal{L}_{\mathrm{HF}} = ||\max\{f^+_{\textrm{FlexConv}}, f_{\mathrm{Nyq}}(k)\} - f_{\mathrm{Nyq}}(k)||^2, \ \ \text{with} \ \ f_{\textrm{Nyq}}(k) = \tfrac{k-1}{4}. \tag{\ref{eq:3_regularizeflexconv}}
\end{equation}
Here, $k$ depicts the size of the FlexConv kernel before applying the Gaussian mask, and is equal to the size of the input signal. In practice, we implement Eq.~\ref{eq:3_regularizeflexconv} by regularizing the individual MAGNet layers, as is detailed in Appx.~\ref{sec:3_magnetreg}. As verification, Fig.~\ref{fig:3_app-cifar10kernelfrequencies} shows that FlexNet kernels are properly regularized against aliasing.
\vspace{-7mm}
\section{Experiments}\label{sec:3_experiments}

We evaluate FlexConv across classification tasks on sequential and image benchmark datasets, and validate the ability of MAGNets to approximate complex functions. A complete description of the datasets used is given in Appx.~\ref{sec:3_datasets}. The hyperparameters used in our experiments are reported in Appx.~\ref{sec:3_flexnet-optimization}.

\vspace{-7mm}
\subsection{What kind of functions can MAGNets approximate?}
\label{sec:3_type1-experiment}

\begin{figure}
    \centering
     \begin{subfigure}[c]{0.49\textwidth}
         \centering
         \includegraphics[width=\textwidth]{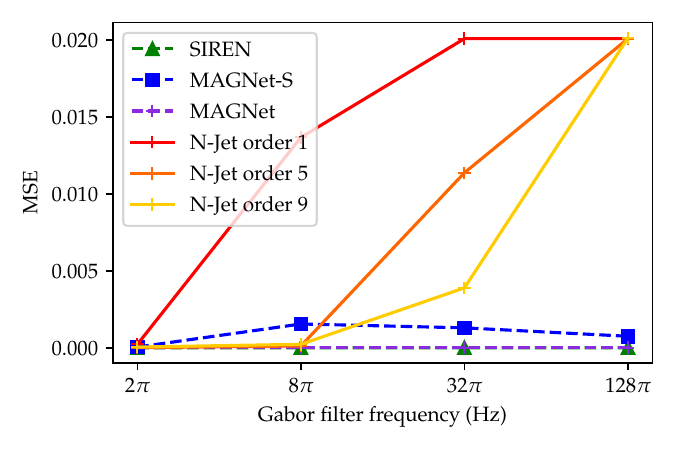}
        %  \caption{FlexConv kernels (ours)}
         \label{fig:3_exp-type1-gabor-graph}
     \end{subfigure}
     \hfill
     \begin{subfigure}[c]{0.49\textwidth}
         \centering
         \vspace{-8mm}
         \includegraphics[width=\textwidth]{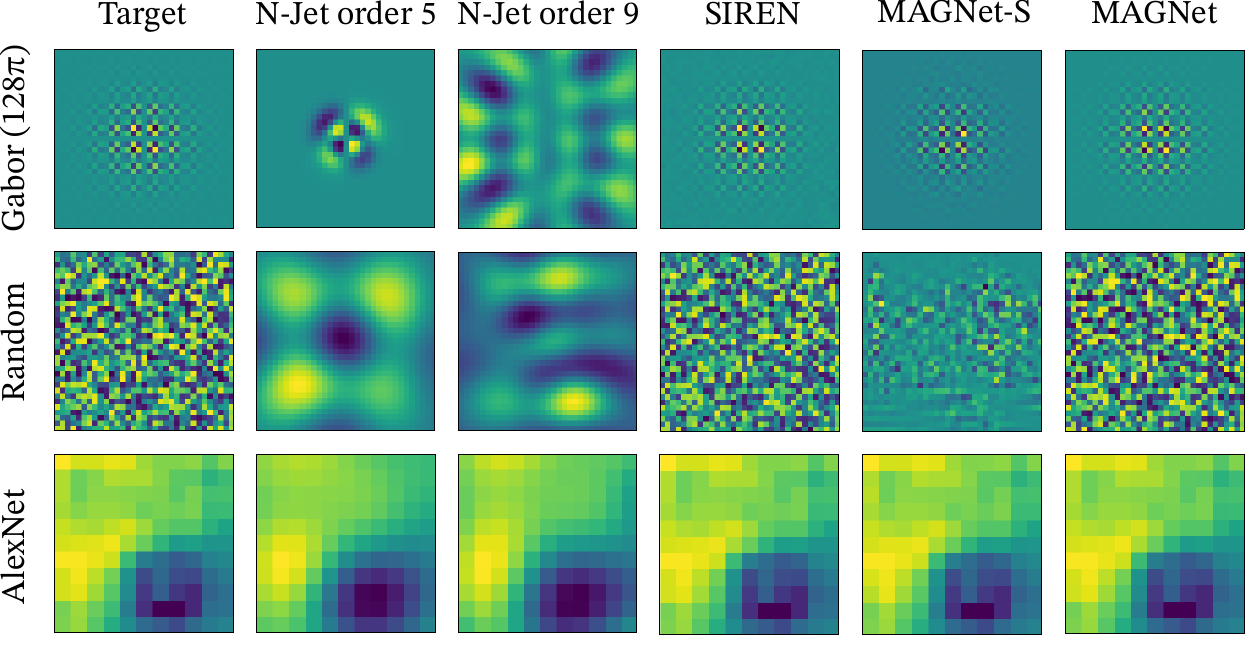}
        %  \caption{}
         \label{fig:3_exp-type1-gabor-kernels}
     \end{subfigure}
     \hfill
    \vspace{-9mm}
    \caption{%N-Jets cannot fit high frequency signals in large kernels, as the predefined Gaussian derivative order constraints the kernel bandwidth. 
    \textit{Left}: Final MSE after fitting each model to Gabor filters of different frequencies. N-Jets cannot fit high frequencies. \textit{Right}: Kernels learned by each model. SIREN and MAGNet can fit all targets. MAGNet-S: a small MAGNet of size akin to N-Jets, still does well on the Gabor and AlexNet targets.}
    \label{fig:3_exp-type1-gabor}
\end{figure}
\textbf{Bandwidth of methods with learnable sizes.} First, we compare the bandwidth of MAGNet against N-Jet \citep{pintea2021resolution} by optimizing each to fit simple targets: (i) Gabor filters of known frequency, (ii) random noise and (iii) an a $11\times11$ AlexNet kernel from the first layer \citep{krizhevsky2012imagenet}.
Fig.~\ref{fig:3_exp-type1-gabor} shows that, even with 9 orders of Gaussian derivatives, N-Jets cannot fit high frequency signals in large kernels. Crucially, N-Jet models require many Gaussian derivative orders to model high frequency signals in large kernels: a hyperparameter which proportionally increases their inference time and parameter count. % In addition, the order of Gaussian derivatives is a hyperparameter and must be selected prior to training. 
MAGNets, on the other hand, accurately model large high frequency signals. This allows FlexNets to learn large kernels with high frequency components.% , while N-Jet models must rely on pooling methods, such as the safe subsampling method proposed in \citet{pintea2021resolution}. 

\textbf{Expressivity of {\btt MLP}\ parameterizations.} Next, we compare the descriptive power and convergence speed of MAGNets, Gabor MFNs, Fourier MFNs and SIRENs for image approximation. To this end, we fit the images in the Kodak dataset \citep{kodak1991} with each of these methods. Our results (Tab.~\ref{tab:3_kodak-results}) show that MAGNets outperform all other methods, and converge faster to good approximations.% In addition, Tab.~\ref{tab:3_cifar-10} illustrates the isolated advantages of our initialization strategy and the use of anisotropic Gabor functions. 

\vspace{-7mm}
\subsection{Classification Tasks}
\label{sec:3_classification}

\begin{table}
\centering
\caption{Test accuracy and ablation on sMNIST, pMNIST, sCIFAR10 and npCIFAR10.}
\label{tab:3_smnist}
\vspace{-3mm}
\begin{small}
\scalebox{0.75}{
\begin{tabular}{cccccc}
\toprule
\sc{Model} & \sc{Size} & \sc{sMNIST} & \sc{pMNIST} & \sc{sCIFAR10} & \sc{npCIFAR10} \\
 \midrule
 DilRNN \citep{chang2017dilated} & 44\sc{k} & 98.0 & 96.1 & - & -\\
  IndRNN \citep{li2018independently} & 83\sc{k} & 99.0 & 96.0& - & - \\
  %LSTM \citep{bai2018empirical} &70\sc{k}& 87.2 & 85.7  & - \\
 %GRU \citep{bai2018empirical} &70\sc{k}&   96.2 & 87.3  & - \\
   TCN \citep{bai2018empirical} &70\sc{k}& 99.0 & 97.2  & - & - \\
  r-LSTM \citep{trinh2018learning} & 0.5\sc{m} & 98.4 & 95.2 & 72.2 & -\\
 Self-Att. \citep{trinh2018learning} &0.5\sc{m} & 98.9 & 97.9 & 62.2 & -\\
 TrellisNet \citep{bai2018trellis}& 8\sc{m} & 99.20 & 98.13 & 73.42 & - \\
 URLSTM \citep{gu2020improving} & - & 99.28 & 96.96 & 71.00 & - \\
  URGRU + Zoneout \citep{gu2020improving} & - & 99.27 & 96.51 & \textbf{74.40} & - \\
  HiPPO \citep{gu2020hippo} & 0.5\sc{m} & - & \textbf{98.30} & - & - \\
  Lipschitz RNN \citep{erichson2020lipschitz} & 158\sc{k} & 99.4 & 97.3 & 64.2 & 59.0 \\
  coRNN \citep{rusch2020coupled} & 134\sc{k} & \textbf{99.4} & 97.3 & - & 59.0 \\
  UnICORNN \citep{rusch2021unicornn} & 135\sc{k} & - & 98.4 & - & \textbf{62.4} \\
  pLMU \citep{chilkuri2021parallelizing} & 165\sc{k} & - & 98.49 & - & -\\
\midrule
CKCNN-2 & 98\sc{k} & 99.31 & 98.00 & 62.25 & 60.5\\
CKCNN-2-Big & 1\sc{m} & 99.32 & 98.54 & 63.74 & 62.2 \\
CKTCN$_{\text{\sc{Fourier}}}$-2 & 105\sc{k} & 99.44 & 98.40 & 68.28 & 66.26 \\
CKTCN$_{\text{\sc{Gabor}}}$-2 & 106\sc{k} & 99.52 & 98.38 & 69.26 & 67.37 \\
CKTCN$_{\text{\sc{MAGNet}}}$-2 & 105\sc{k} & \textbf{99.55} &\textbf{ 98.57} & \textbf{74.58} & \textbf{67.52} \\
\midrule
FlexTCN-2 & 108\sc{k} & \textbf{99.60} & \textbf{98.61} & \textbf{78.99} &  \textbf{67.11}\\
FlexTCN-4 &  241\sc{k} & \textbf{99.60} & \textbf{98.72} & \textbf{80.26} & \textbf{67.42}\\
FlexTCN-6 & 375\sc{k} & \textbf{99.62} &  \textbf{98.63} & \textbf{80.82} & \textbf{69.87} \\
\midrule
FlexTCN$_{\text{SIREN}}$-6 & 343\sc{k} & 99.03 & 95.36 & 69.24 & 57.27 \\
FlexTCN$_{\text{Fourier}}$-6 & 370\sc{k} & 99.49 & 97.97 & 74.79 & 67.35 \\
FlexTCN$_{\text{Gabor}}$-6 & 373\sc{k} &  99.50 & 98.37 & 78.36 & 67.56 \\
FlexTCN$_{\text{MAGNet}}$-6 & 375\sc{k} & \textbf{99.62} &  \textbf{98.63} & \textbf{80.82} & \textbf{69.87} \\
\bottomrule
\end{tabular}}
\end{small}
\end{table}

\textbf{Network specifications.} Here, we specify our networks for all our classification experiments. We parameterize all our convolutional kernels as the superposition of a 3-layer MAGNet and a learnable anisotropic Gaussian mask. We construct two network instances for sequential and image datasets respectively: FlexTCNs and FlexNets. Both are constructed by taking the structure of a baseline network --TCN \citep{bai2018empirical} or CIFARResNet \citep{he2016deep}--, removing all internal pooling layers, and replacing convolutional kernels by FlexConvs. The FlexNet architecture is shown in Fig.~\ref{fig:3_flexnet-architecture} and varies only in the number of channels and blocks, e.g., FlexNet-16 has 7 blocks. Akin to \cite{romero2022ckconv} we utilize the Fourier theorem to speed up convolutions with large kernels.

\textbf{Mask initialization.} We initialize the FlexConv masks to be small. Preliminary experiments show this leads to better performance, faster execution, and faster training convergence. For sequences, the mask center is initialized at the last kernel position to prioritize the last information seen.

\textbf{Time-series and sequential data.} First we evaluate FlexTCNs on sequential classification datasets, for which long-term dependencies play a crucial role. 
We validate our approach on intrinsic discrete data: \textit{sequential MNIST}, \textit{permuted MNIST} \citep{le2015simple}, \textit{sequential CIFAR10} \citep{chang2017dilated}, \textit{noise-padded CIFAR10} \citep{chang2019antisymmetricrnn}, and time-series: \textit{CharacterTrajectories} (CT) \citep{bagnall2018uea}, \textit{SpeechCommands} \citep{warden2018speech} with raw waveform (SC\_raw) and MFCC representations (SC).

Our results are shown in Tables~\ref{tab:3_smnist}~and~\ref{tab:3_time-series}. FlexTCNs with two residual blocks obtain state-of-the-art results on all tasks considered. In addition, depth further improves performance. FlexTCN-6 improves the current state-of-the-art on sCIFAR10 and npCIFAR10 by more than 6\%. On the difficult SC\_raw dataset --with sequences of length 16000--, FlexTCN-6 outperform the previous state-of-the-art by 20.07\%.

%  -=== Table
\begin{table}
\RawFloats
\centering
\begin{minipage}{0.48 \textwidth}
\centering
\caption{Test accuracy on CT, SC and SC\_raw}
\label{tab:3_time-series}
\vspace{-2mm}
\begin{small}
\scalebox{0.75}{
\begin{tabular}{ccccc}
\toprule
\sc{Model} & \sc{Size} & \sc{CT} & \sc{SC} & \sc{SC\_raw} \\
\midrule
GRU-ODE & 89\sc{k} & 96.2 & 44.8 & $\sim$10.0 \\
GRU-$\Delta t$ & 89\sc{k} & 97.8 & 20.0 & $\sim$10.0 \\
GRU-D & 89\sc{k} & 95.9 & 23.9 & $\sim$10.0 \\
ODE-RNN & 89\sc{k} & 97.1 & 93.2 & $\sim$10.0 \\
NCDE  & 89\sc{k} & 98.8 & 88.5 & $\sim$10.0 \\
\midrule
CKCNN & 100\sc{k} &\textbf{ 99.53 }& 95.27 & 71.66 \\
CKTCN$_{\text{Fourier}}$ & & - & 95.65 & 74.90 \\
CKTCN$_{\text{Gabor}}$ &  & - & 96.66 &  78.10 \\
CKTCN$_{\text{MAGNet}}$ & 105\sc{k} &  \textbf{99.53} & \textbf{97.01} & \textbf{80.69} \\
\midrule
FlexTCN-2 & 105\sc{sk} & \textbf{99.53} & \textbf{97.10} & \textbf{88.03}  \\
FlexTCN-4 & 239\sc{k} & \textbf{99.53} & \textbf{97.73} & \textbf{90.45}   \\
FlexTCN-6 & 373\sc{k} & \textbf{99.53} & \textbf{97.67} & \textbf{91.73} \\
\midrule
FlexTCN$_{\text{SIREN}}$-6 & 370\sc{k} & - & 95.83 & 85.73\\
FlexTCN$_{\text{Fourier}}$-6 & 342\sc{k} & - & 97.62 & 91.02 \\
FlexTCN$_{\text{Gabor}}$-6 & 373\sc{k} & - & 97.35 & 91.50 \\
FlexTCN$_{\text{MAGNet}}$-6 & 373\sc{k} & - & \textbf{97.67} & \textbf{91.73} \\
\bottomrule
\end{tabular}}
\end{small}
\end{minipage}%
\hfill
\begin{minipage}{0.50 \textwidth}
\centering
\caption{Results on CIFAR-10. Results from *original works and $\dagger$ single run.}
\label{tab:3_cifar-10}
\vspace{-2mm}
\begin{small}
\scalebox{0.75}{
\begin{tabular}{cccc}
\toprule
    \multirow{2}{*}{\sc{Model}} & \multirow{2}{*}{\sc{Size}} & \sc{CIFAR-10} & \sc{Time} \\
    & & \sc{Acc.} & \sc{(sec/epoch)}\\ \midrule
    CIFARResNet-44 & 0.66\sc{m} & 92.9*\!\dagger & 22 \\
    DCN-$\sigma^{ji}$ & 0.47\sc{m} & 89.7 $\pm$ 0.3* & - \\
    N-Jet-CIFARResNet32 & 0.52\sc{m} & 92.3 $\pm$ 0.3* & - \\
    N-Jet-ALLCNN & 1.07\sc{m} & 92.5 $\pm$ 0.1* & - \\ \midrule
    FlexNet-16 w/ conv. ($k = 3$) & 0.17\sc{m} & 89.5 $\pm$ 0.3 & 41   \\
    FlexNet-16 w/ conv. ($k = 33$) & 20.0\sc{m} & 78.0 $\pm$ 0.3 & 242  \\
    FlexNet-16 w/ N-Jet & 0.70\sc{m} & 91.7 $\pm$ 0.1 & 409  \\ \midrule
    % \midrule
    CKCNN-16 & 0.63\sc{m} & 72.1 $\pm$ 0.2 & 68 \\ % 71.70
    CKCNN$_{\text{MAGNet}}$-16 & 0.67\sc{m} &  86.8 $\pm$ 0.6  & 102 \\ % 85.91
    FlexNet$_{\text{SIREN}}$-16 & 0.63\sc{m} & 89.0 $\pm$ 0.3  & 89 \\ % 88.88
    FlexNet$_{\text{Gabor}}$-16 & 0.67\sc{m} & 91.9 $\pm$ 0.2 & 161 \\ \midrule % 91.99
    FlexNet$_{\text{Gabor}}$-16 + anis. Gauss. & 0.67\sc{m} & 92.0 $\pm$ 0.1 & 147 \\
    FlexNet$_{\text{Gabor}}$-16 + Gabor init. & 0.67\sc{m} & 92.0 $\pm$ 0.2 & 150 \\
    \midrule
    FlexNet-16 & 0.67\sc{m} & 92.2 $\pm$ 0.1 & 127 \\
 \bottomrule
\end{tabular}}
\end{small}
\end{minipage}
\end{table}

Furthermore, we conduct ablation studies by changing the parameterization of \mlppsi, and switching off the learnable kernel size ("CKTCNs") and considering global kernel sizes instead. CKTCNs and FlexTCNs with MAGNet kernels outperform corresponding models with all other kernel parameterizations: SIRENs \citep{sitzmann2020implicit}, MGNs and MFNs \citep{fathony2021multiplicative}. Moreover, we see a consistent improvement with respect to CKCNNs \citep{romero2022ckconv} by using learnable kernel sizes. This shows that both MAGNets and learnable kernel sizes contribute to the performance of FlexTCNs.
Note that in 1D, MAGNets are equivalent to MGNs. However, MAGNets consistently perform better than MGNs. This improvement in accuracy is a result of our MAGNet initialization.

\textbf{Image classification.}
Next, we evaluate FlexNets on CIFAR-10 \citep{krizhevsky2009learning}.  Additional experiments on ImageNet-32, MNIST and STL-10 are given in Appx.~\ref{sec:3_appx-experiments}.

Table~\ref{tab:3_cifar-10} shows our results on CIFAR-10. 
FlexNets are competitive with pooling-based methods such as CIFARResNet 
\citep{he2016deep} and outperform learnable kernel size method DCNs \citep{tomen2021deep}.
In addition, we compare using N-Jet layers of order three (as in \citet{pintea2021resolution}) in FlexNets against using MAGNet kernels. We observe that N-Jet layers lead to worse performance, and are significantly slower than FlexConv layers with MAGNet kernels. The low accuracy of N-Jet layers is likely to be linked to the fact that FlexNets do not use pooling. Consequently, N-Jets are forced to learn large kernels with high-frequencies, which we show N-Jets struggle learning~in~Sec.~\ref{sec:3_type1-experiment}.%, shows that N-Jets struggle  
%With this setting, 
% N-Jet layers in FlexNets are significantly slower than FlexNets, while performing worse on CIFAR-10.
%This lower accuracy may have to do with FlexNets not using pooling, forcing N-Jet layers to learn high frequency large kernels, which we have shown N-Jets struggle with in .
% If this hypothesis holds, this explains the necessity of the safe subsampling method introduced in \citet{pintea2021resolution}.

To illustrate the effect of learning kernel sizes, we also compare FlexNets against FlexNets with large and small discrete convolutional kernels (Tab.~\ref{tab:3_cifar-10}).
Using small kernel sizes is parameter efficient, but is not competitive with FlexNets. Large discrete kernels on the other hand require a copious amount of parameters and lead to significantly worse performance. These results indicate that the best solution is somewhere in the middle and varying kernel sizes can~learn~the~optimal~kernel~size~for~the~task~at~hand.
%To illustrate the potential of learning kernel size, we compare FlexNet against FlexNets with regular convolutional kernels. Table~\ref{tab:3_cifar-10} shows that the FlexNet with small kernels is parameter efficient, but is not competitive with our FlexNets. The FlexNet with large kernels requires a copious amount of parameters and is significantly worse than our FlexNets. Both of these results show that the best solution is somewhere in the middle: varying kernel sizes can learn the optimal kernel size for the problem at hand.

Similar to the sequential case, we conduct ablation studies on image data with learnable, non-learnable kernel sizes and different kernel parameterizations. Table~\ref{tab:3_cifar-10} shows that FlexNets outperform CKCNNs with corresponding kernel parameterizations. %to do so. 
In addition, a clear difference in performance is apparent for MAGNets with respect to other parameterizations. These results corroborate that both MAGNets and FlexConvs contribute to the performance of FlexNets. Moreover, Tab.~\ref{tab:3_cifar-10} illustrates the effect of the two contributions of MAGNet over MGN: anisotropic Gabor filters, and our improved initialization. Our results are in line with our results for sequential data (Tabs.~\ref{tab:3_smnist},~\ref{tab:3_time-series}), illustrating the value of the proposed improvements of MAGNets.

\vspace{-7mm}
\subsection{Alias-free FlexNets}
\label{sec:3_crossresexperiments}

\begin{figure}[t]
        \centering
    \includegraphics[width=\textwidth]{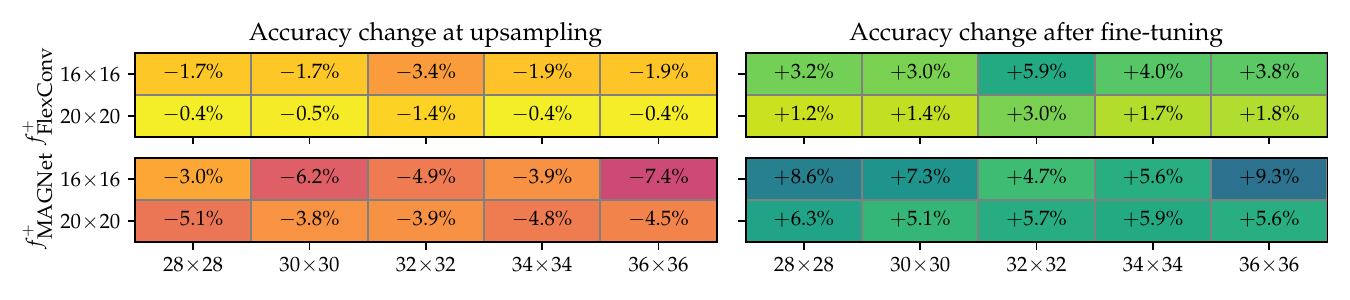}
    \vspace{-8mm}
    \caption{Alias-free FlexNet-16 on CIFAR-10. We report change in accuracy between source and target resolutions, directly after upsampling (left) and after fine-tuning (right) (means over five runs).}
    \label{fig:3_c10-crossres}
\end{figure}

\textbf{Regularizing the FlexConv mask.} Though including $f^+_{w_{\mathrm{gauss}}}$ in the frequency analysis of MAGNets is crucial for the accuracy of the derivation, including the FlexConv mask in aliasing regularization is undesirable, as it steers the model to learn large kernels in order to minimize the loss (see Eq.~\ref{eq:3_regularizeflexconv}). However, excluding the mask from regularization could compromise the ability of FlexNet to generalize to higher resolutions. Here, we experiment with this trade-off.

\begin{table}
\centering
\captionsetup{justification=centering}
\caption{Alias-free FlexNets on CIFAR-10.}
\label{tab:3_crossres}
\vspace{-2mm}
\begin{center}
\scalebox{0.75}{
\begin{tabular}{cccc}
\toprule
    \multirow{2}{*}{\sc{Model}} & \multirow{2}{*}{\sc{Size}} & \multicolumn{2}{c}{\sc{CIFAR-10 Acc.}} \\
    & & 16 px & $\Delta_{16 \textrm{px}}$ 32 px \\ \midrule % & fine-tuned \\ \midrule
    CIFARResNet-44 & 0.66\sc{m} & 85.8 $\pm$ 0.2 & -31.6 $\pm$ 1.3 \\ \midrule
    % 85.4 $\pm$ 0.3 & -31.2 $\pm$ 2.8 \\ \midrule % & 92.6 $\pm$ 0.2 \\ \midrule
    FlexNet-16 w/ conv. ($k = 3$) & 0.17\sc{m} & 85.3 $\pm$ 0.2 & -21.2 $\pm$ 1.0 \\ % & \textbf{90.1} $\pm$ 0.1  \\
    FlexNet-16 w/ conv. ($k = 33$) & 20.0\sc{m} & 67.7 $\pm$ 0.6 & -57.1 $\pm$ 1.6 \\ % & 79.2 $\pm$ 0.4  \\
    FlexNet-16 w/ N-Jets & 0.70\sc{m} & \textbf{86.4} $\pm$ 0.2 & -5.5 $\pm$ 1.3 \\ \midrule % & 89.5 $\pm$ 0.1  \\ \midrule
    CKCNN-16$_{\textrm{SIREN}}$ & 0.63\sc{m} & 45.9 $\pm$ 1.0 & -15.8 $\pm$ 1.2 \\
    FlexNet-16$_{\textrm{SIREN}}$ & 0.63\sc{m} & 70.4 $\pm$ 0.8 & -50.0 $\pm$ 16.9 \\ \midrule
    FlexNet-16 w/o reg. & 0.67\sc{m} & \textbf{86.4} $\pm$ 0.4 & -34.4 $\pm$ 14.3 \\ \midrule % & 88.8 $\pm$ 0.6 \\ \midrule
    FlexNet-16 w/ reg. $f^+_{\textrm{MAGNet}}$ & 0.67\sc{m} & \textbf{86.5} $\pm$ 0.1 & -3.8 $\pm$ 2.0 \\ % & 89.0 $\pm$ 0.3  \\
    FlexNet-16 w/ reg. $f^+_{\textrm{FlexConv}}$ & 0.67\sc{m} & 85.1 $\pm$ 0.3 & \textbf{-3.3} $\pm$ 0.3 \\ % & 87.6 $\pm$ 0.2 \\
 \bottomrule
\end{tabular}}
\end{center}
\end{table}

Figure~\ref{fig:3_c10-crossres} shows accuracy change between ten source and target resolution combinations on CIFAR-10, both for including and excluding the FlexConv mask in the aliasing regularization. We train at the source resolution for 100 epochs, before testing the model at the target resolution with the upsampling described in Sec.~\ref{sec:3_crtraining}. Next, we adjust $f_{\textrm{Nyq}}(k)$ to the target resolution, and finetune each model for 100 epochs at the target resolution.

We find that regularizing just $f^+_{\textrm{MAGNet}}$ yields a trade-off. It increases the accuracy difference between low and high resolution inference, but also increases the fine-tune accuracy at the target resolution.%, as the FlexConv masks are not constrained. 
We therefore regularize $f^+_{\textrm{MAGNet}}$ only by default.

Results of our alias-free FlexNet training on CIFAR-10 are in Table~\ref{tab:3_crossres}. We observe that the performance of a FlexNet trained without aliasing regularization largely breaks down when the dataset is upscaled. 
%Analogously, some cross-resolution performance degradation is observed for CKConvs on the 1D CT and \textsc{SC\_raw} datasets \citep{romero2022ckconv}. 
However, with our aliasing regularization most of the performance is retained. 

Comparatively, FlexNet retains more of the source resolution performance than FlexNets with N-Jet layers, while baselines degrade drastically at the target resolution. Fig.~\ref{fig:3_app-cifar10kernelfrequencies} shows the effect of aliasing regularization on the frequency components of FlexConv.

\textbf{Training at lower resolutions saves compute.} We can train alias-free FlexNets at lower resolutions. To verify that this saves compute, we time the first 32 batches of training a FlexNet-7 on CIFAR-10. We compare against training on $16 \times 16$ images (downsampled before training). On 16x16 images, each batch takes 179ms ($\pm$ 7ms). On 32x32 images, each batch takes 222ms ($\pm$ 9ms). Therefore, we save 24\% training time when training FlexNets alias-free at half the native CIFAR-10 resolution.
\vspace{-7mm}
\section{Discussion}
\label{sec:3_discussion}

\begin{figure}[t]
    \centering
    \includegraphics[width=\textwidth]{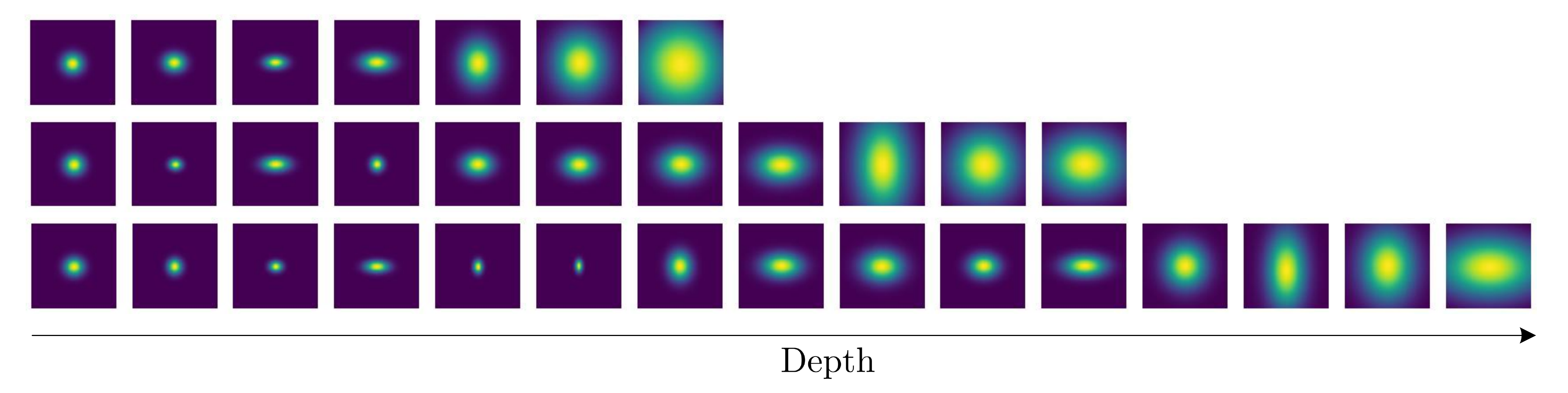}
    \vspace{-8mm}
    \caption{Learned masks for FlexNets with 3, 5 and 7 residual blocks. FlexNets learn very small kernels at shallow layers, which become larger as a function of depth.}
    \label{fig:3_c10-kernels}
\end{figure}

\textbf{Learned kernel sizes match conventional priors.} Commonly, CNNs use architectures of small kernels and pooling layers. This allows convolutions to build a progressively growing receptive field. With learnable kernel sizes, FlexNet could learn a different prior over receptive fields, e.g., large kernels first, and small kernels next. However, FlexNets learn to increase kernel sizes progressively (Fig.~\ref{fig:3_c10-kernels}), and match the network design that has been popular since AlexNet \citep{krizhevsky2012imagenet}.

\textbf{Mask initialization as a prior for feature importance.} The initial values of the FlexConv mask can be used to prioritize information at particular input regions. For instance, initializing the center of mask on the first element of sequential FlexConvs can be used to prioritize information from the far past. This prior is advantageous for tasks such as npCIFAR10. We observe that using this prior on npCIFAR10 leads to much faster convergence and better results (68.33\% acc. w/ FlexTCN-2).

\textbf{MAGNet regularization as prior induction.} MAGNets allow for analytic control of the properties of the resulting representations. We use this property to generate alias-free kernels. However, other desiderata could be induced, e.g., smoothness, for the construction of implicit neural representations. 
% Analyzing interesting properties of convolutional kernels, and implicit neural representations, in general.
% \textbf{Same architecture, different datasets.} STL-10, MNIST. No $w_0$ te tunen. 

%\textbf{FlexNets are able to exploit depth.} \cite{romero2022ckconv} indicate that CKCNNs cannot efficiently take advantage of depth, and thus CKCNNs contsain only two residual blocks. Our work shows that learnable kernel sizes allow FlexNets to take advantage of depth analogously to conventional CNNs.
\textbf{Benefits of cropping and the influence of {\btt PyTorch}.} Dynamic cropping adjust the computational cost of the convolutions on the fly. For a signal of size $\mathrm{M}$ and a cropped kernel size $\mathrm{k}$, this incurs in savings from $\mathrm{O}(\mathrm{M}^{2^D})$ to $\mathrm{O}(\mathrm{M^D k^D})$ relative to using global kernel sizes~($\mathrm{O}(\mathrm{M}^{4})$~to~$\mathrm{O}(\mathrm{M}^{2}\mathrm{k}^{2})$~in~2D).
We test this theoretical speed up in a controlled environment for the Speech Commands and CIFAR-10 datasets. Cropping reduces the per-epoch run time by a factor of 11.8$\mathrm{x}$ and 5.5$\mathrm{x}$ for Speech Commands and CIFAR-10, respectively. Interestingly, however, both run times become similar if the flag \texttt{torch.backends.cudnn.benchmark} is activated, with global kernel sizes being sometimes faster. This is because this flag tells \texttt{PyTorch} to optimize the convolution algorithms used under the hood, and some of these CUDA algorithms seem to be faster than our masking strategy on \texttt{Python}.
\vspace{-7mm}
\section{Limitations}
\label{sec:3_limitations}

\textbf{Dynamic kernel sizes: computation and memory cost of convolutions with large kernels.} Performing convolutions with large convolutional kernels is a compute-intensive operation. FlexConvs are initialized with small kernel sizes and their inference cost is relatively small at the start of training. However, despite the cropping operations used to improve computational efficiency (Figs.~\ref{fig:3_flexconv},~\ref{fig:3_capacity_tradeoff}, Tab.~\ref{tab:3_cifar-10}), the inference time may increase to up to double as the learned masks increase in size. At the cost of more memory, convolutions can be sped up by performing them in the frequency domain. % with large convolutional kernels. 
However, we observe that this does not bring gains on image data considered as FFT convolutions are only faster for very large convolutional kernels --in the order of hundreds of pixels.

% This leaves an efficiency gap for images in benchmark datasets, which cannot be fully exploited by our method.
% \textbf{Expensive parameterization for small kernels.} FlexConvs are very parameter efficient in modelling large kernels. However, as MAGNets parameterize kernels via a multi-layer perceptron \mlppsi, it can become more expensive for small kernels than discrete parameterizations. Nevertheless, this cost can be controlled by modifying the network structure. %use more parameters than conventional convolutions when using exclusively small kernels, though the exact trade-off depends on the parameterization of the MAGNet.
%FlexConv crops kernels using the window parameters to reduce the computational load. As FlexConvs are initialized with small kernel sizes, their inference cost is relatively small at the start of training. However, the inference time may increase to up to double as the learned masks increase in size. Performing the convolution in the frequency domain using Fourier transforms can make large kernel convolutions faster at the cost of larger memory requirements. 

\textbf{Remaining accuracy drop in alias-free FlexNets.} Some drop in accuracy is still observed when using alias-free FlexNets at a higher test resolutions (Tab.~\ref{tab:3_crossres}). Although more evidence is needed, this may be caused by aliasing effects introduced by $\mathrm{ReLU}$ \citep{vasconcelos2021impact} or by changes in the statistics of the feature maps passed to global pooling \citep{Touvron2019FixingTT}.
\vspace{-7mm}
\section{Conclusion}

We propose FlexConv, a convolutional operation able to learn high bandwidth convolutional kernels of varying size during training at a fixed parameter cost. We demonstrate that FlexConvs are able to model long-term dependencies without the need of pooling, and shallow pooling-free FlexNets achieve state-of-the-art performance on several sequential datasets, match performance of recent works with learned kernel sizes with less compute, and are competitive with much deeper ResNets on image benchmark datasets. In addition, we show that our alias-free convolutional kernels allow FlexNets to be deployed at higher resolutions than seen during training with minimal loss.
%Moreover, our proposed kernel parameterization shows higher descriptive power and faster convergence speed than existing parameterizations.

\textbf{Future work.} MAGNets give control over the bandwidth of the kernel. We anticipate that this control has more uses, such as fighting sub-sampling aliasing \citep{zhang2019making,Kayhan_2020_CVPR,karras2021alias}. With the ability to upscale FlexNets to different input image sizes comes the possibility of transfer learning representations between previously incompatible datasets, such as CIFAR-10 and ImageNet.
In a similar vein, the automatic adaptation of FlexConv to the kernel sizes required for the task at hand may make it possible to generalize the FlexNet architecture across different tasks and datasets. Neural architecture search \citep{zoph17} could see benefits from narrowing the search space to exclude kernel size and pooling layers. In addition, we envisage additional improvements from structural developments of FlexConvs such as attentive FlexNets.
% , and parameterizations with increased parameter efficiency. 

% Though their representational power is demonstrated in our experiments, the inference costs of FlexConvs is significant for large images at native resolution. We anticipate that the ability to learn long-term dependencies might benefit semantic segmentation tasks, but the large resolutions of the images in these tasks makes it impractical to use FlexNets. Future work could aim to reduce the computational load of FlexConvs on high resolutions.

%% file: chapters/5_dnarch.tex
% path to figures directory
\graphicspath{{figures/5-dnarch/}}

%=========================================================================

% \begin{savequote}[75mm]
% Nulla facilisi. In vel sem. Morbi id urna in diam dignissim feugiat. Proin molestie tortor eu velit. Aliquam erat volutpat. Nullam ultrices, diam tempus vulputate egestas, eros pede varius leo.
% \qauthor{Quoteauthor Lastname}
% \end{savequote}

\chapter{Learning Convolutional Neural Architectures by Backpropagation}
	\label{chapter:dnarch}

\begin{flushright}
\textit{Based on the paper:}\break
\textit{DNArch: Learning Convolutional Neural Architectures by Backpropagation \citep{romero2023dnarch}}
\end{flushright}
%=========================================================================
\vspace{-7mm}
\section{Introduction}\label{sec:5_intro}
Convolutional Neural Networks (CNNs) \cite{lecun1998gradient} are widely used for tasks such as image classification \cite{krizhevsky2012imagenet, he2016deep}, speech recognition \cite{sercu2016very, zeghidour2018fully}, text classification \cite{conneau2016very} and generative modeling \cite{oord2016wavenet, dhariwal2021diffusion} due to their performance and efficiency. However, tailoring a CNN architecture to a specific task or dataset typically requires substantial human intervention and cross-validation to design the architectures, e.g. to determine appropriate kernel sizes, width, depth, etc. This has motivated exploring the space of architectures in an automatic fashion, by developing architecture search algorithms~\citep{zoph17, liu18, zhou2019bayesnas}.

\begin{figure}
\centering
\begin{subfigure}{.58\textwidth}
  \centering
  \includegraphics[page=4, trim={0.5cm 22cm 16.5cm 0.5cm}, clip, width=\textwidth]{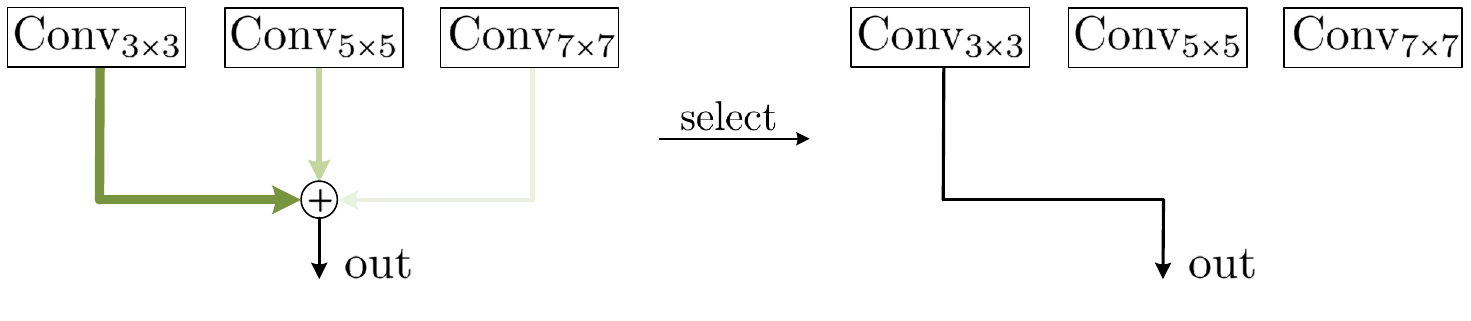}
  \captionsetup{justification=centering}
    \caption{DARTS \citep{liu2018darts}}
    \label{fig:5_darts}
\end{subfigure}
\hfill
\begin{subfigure}{.40\textwidth}
  \centering
  \begin{subfigure}[b]{0.32\textwidth}
         \centering
         \includegraphics[page=2, trim={0.5cm 19cm 13cm 0.5cm}, clip, width=\textwidth]{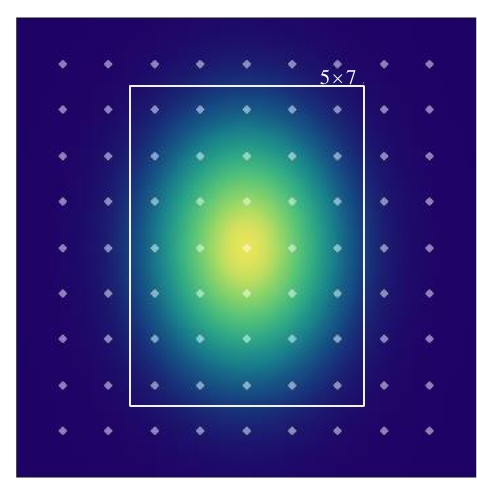}
     \end{subfigure}
     \hspace{3mm}
     \begin{subfigure}[b]{0.32\textwidth}
         \centering
         \includegraphics[page=3, trim={0.5cm 19cm 13cm 0.5cm}, clip, width=\textwidth]{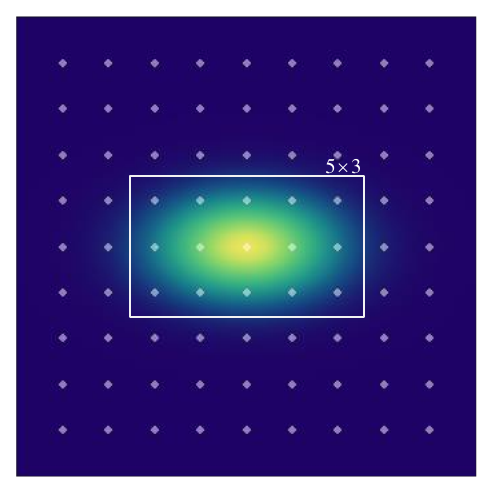}
     \end{subfigure}
     \captionsetup{justification=centering}
\caption{DNArch (ours)} \label{fig:5_dnarch}
\end{subfigure}
\vspace{-2mm}
\caption{DARTS vs. DNArch. DARTS learns the size of convolutional kernels using backpropagation to select among predefined options, e.g., DASH \cite{shen2022efficient}. DNArch, on the other hand, learns the size of convolutional kernels by modifying the parameters of the differentiable mask $m(\cdot \ ; \boldsymbol{\theta})$. Different $\boldsymbol{\theta}$ values lead to different sizes.
\vspace{-2mm}}\label{fig:5_DNArch_kernels}
\end{figure}

While these methods can find good architectures, they must solve an expensive discrete optimization problem that involves training and evaluating candidate architectures in each iteration, e.g. to optimize a reward with reinforcement learning \citep{zoph17}, or to evolve the model through a genetic algorithm \citep{liu18}. Differentiable Architecture Search (DARTS) \cite{liu2018darts} addresses this issue by allowing the network to consider a set of \textit{predefined} possible components in parallel, e.g., convolutions with kernels of size $3{\times}3$, $5{\times}5$, $7{\times}7$, and adjusting their contribution using learnable weights (Fig.~\ref{fig:5_darts}). Although DARTS is able to \textit{select} components via backpropagation, it requires (\textit{i}) defining a (small) set of possible components beforehand, (\textit{ii}) computing and keeping their responses in memory during training, and (\textit{iii}) retraining the found architecture from scratch to remove the effect of other components in the output.
\begin{figure}
    \centering
    \includegraphics[page=1, trim={0.5cm 15cm 12cm 0.5cm}, clip, width=0.95\textwidth]{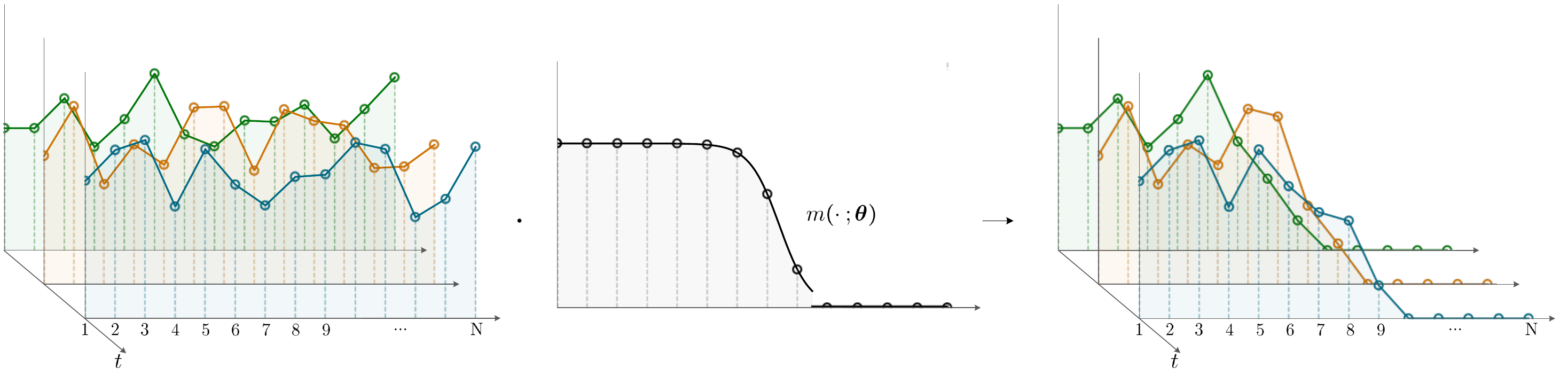}
    \vspace{-2mm}
    \caption{DNArch views neural architectures as entities in a continuous multidimensional space, and uses differentiable masks to learn their length by backpropagation. In this example, DNArch learns the width of a layer by applying a differentiable mask $m$ with learnable parameters $\boldsymbol{\theta}$ to the channel dimension. As a result, changes in the value of $\boldsymbol{\theta}$ effectively results in changes in the layer's width.
    \vspace{-4mm}
    }\label{fig:5_figure_1}
\end{figure}

In this paper, we introduce \textit{Differentiable Neural Architectures} (DNArch), a method that simultaneously learns the weights and the entire architecture of a CNN during training by backpropagation. Specifically, DNArch learns the weights as well as (\textit{i}) the size of convolutional kernels at each layer, (\textit{ii}) the number of channels at each layer, (\textit{iii}) the position and resolution of downsampling layers, and (iv) the number of layers of the network. %All of this without the need for a specific predefined set of components to choose from, or the need for a finetuning step once the architecture has been found.
To this end, DNArch takes a novel approach to learning neural architectures by viewing them as entities defined in a multidimensional continuous space with dimensions corresponding to network attributes, e.g., depth, width, etc., and using differentiable masks with learnable parameters along each dimension to control their length (Fig.~\ref{fig:5_figure_1}). Unlike DARTS methods, e.g., \cite{liu2018darts, shen2022efficient}, DNArch does \textit{not} require a predefined set of components to choose from, but instead is able to explore among \textit{all} feasible values, e.g., all kernel sizes between $1{\times}1$ and $\mathrm{N}{\times}\mathrm{N}$ for a $\mathrm{N}{\times}\mathrm{N}$ image. This is a result of the truly continuous nature of DNArch, which, unlike DARTS, does not require multiple instantiations of the same layer for different parameter values (Fig.~\ref{fig:5_darts}). Instead, DNArch explores the parameter space by modifying the learnable parameters of the differentiable masks (Fig.~\ref{fig:5_dnarch}), making it a much more scalable NAS method. Since both the architecture and the weights are optimized in a single run, no retraining is needed after training.

%In contrast to DARTS, DNArch does \textit{not} require a predefined set of components to select from, and rather views neural architectures as if they were defined in a multidimensional continuous space --with dimensions corresponding to the network's depth, width, etc.--, using a differentiable mask with learnable parameters along each dimension to define their length (Fig.~\ref{fig:5_figure_1}). As a consequence, it is able to choose from \textit{all} admissible values via backpropagation, e.g., all kernel sizes between $1{\times}1$ and $\mathrm{K_{max}}{\times}\mathrm{K_{max}}$. This is shown in Fig.~\ref{fig:5_DNArch_kernels} where two possible kernel sizes are obtained by adjusting the parameters of a learnable differentiable mask. In addition, DNArch does not require a finetuning step as both the architecture and the weights are optimized over a single training run. While such differentiable masks have been previously proposed e.g. by~\citet{riad2022learning} to learn the stride of convolutional layers, DNArch extends this approach to the entire structure of the network rather than targeting a specific operator.% To prevent the formation of poor architectures due to insufficiently large receptive fields,  we use Continuous Kernel Convolutions \cite{romero2022ckconv} to model the global context regardless of specific architectural choices. CKConvs are convolutional layers able to model (global) long range dependencies on inputs of any length, resolution and dimensionality regardless of the network architecture \cite{knigge2023modelling}.

\textbf{Results.} We empirically show that DNArch is able to find performant CNN architectures across several classification and dense prediction tasks on sequential and image datasets. The architectures found by DNArch consistently surpass the general-purpose convolutional architecture on which DNArch is applied, and often outperform specialized task-specific architectures. Moreover, we show that DNArch can be easily combined with a regularization term that controls the computational complexity of candidate networks. By doing so, DNArch explores among neural architectures that respect a predefined computational budget during the entire training process. As a result, finding architectures with DNArch is roughly as expensive as a single training loop of the underlying baseline.
%\textbf{Contribution.} To the best of our knowledge, DNArch is the first method that jointly learns the weights and the architecture of a CNN by backpropagation, exploring all combinations of kernel sizes, widths, depths and downsampling within a large admissible range.

% \begin{figure}
%     \centering
%     \includegraphics[page=4, trim={0.5cm 21cm 16.5cm 0.5cm}, clip, width=\textwidth]{images.pdf}
%     \vspace{-15mm}
%     \caption{Use of DARTS to learn the size of convolutional kernels. DARTS uses backpropagation to select among predefined options.
%     \vspace{-7.5mm}}
%     \label{fig:5_darts}
% \end{figure}
% \begin{figure}
%     \centering
%     \hfill
%     \begin{subfigure}[b]{0.32\textwidth}
%          \centering
%          \includegraphics[page=2, trim={0.5cm 19cm 13cm 0.5cm}, clip, width=\textwidth]{images.pdf}
%      \end{subfigure}
%      \hfill
%      \begin{subfigure}[b]{0.32\textwidth}
%          \centering
%          \includegraphics[page=3, trim={0.5cm 19cm 13cm 0.5cm}, clip, width=\textwidth]{images.pdf}
%      \end{subfigure}
%      \hfill
%      \vspace{-6mm}
%     \caption{Use of DNArch to learn the size of convolutional kernels. By modifying the parameters of the differentiable mask $m(\cdot \ ; \boldsymbol{\theta})$, convolutional kernels of different sizes can be obtained.
%     \vspace{-4mm}}
%     \label{fig:5_DNArch_kernels}
% \end{figure}
% \vspace{-3mm}
% \section{Related work}
% \vspace{-2mm}
\vspace{-7mm}
\section{Method}
DNArch has two key components: Differentiable Masking and Continuous Kernel Convolutions. Here, we introduce these concepts and show how they can be used to learn CNN architectures next.
\vspace{-7mm}
\subsection{Differentiable Masking}\label{sec:5_diff_masking}
Let us consider an arbitrary function $f: [\mathrm{a}, \mathrm{b}] \rightarrow \sR$, which we want to be non-zero only in a subset $[\mathrm{c}, \mathrm{d}] \subseteq [\mathrm{a}, \mathrm{b}]$. To this end, we can multiply $f$ with a mask $m$ whose values are non-zero only on $[\mathrm{c}, \mathrm{d}]$, e.g., a rectangular mask $\Pi_{[\mathrm{c}, \mathrm{d}]}(x) {=} \mathbb{1}_{[\mathrm{c}, \mathrm{d}]}$. However, as its gradient is either zero or non-defined, it is not possible to learn the interval in which it is non-zero by backpropagation.
To overcome this limitation, we can instead use a parametric differentiable mask $m(\cdot\ ; \boldsymbol{\theta})$ whose interval of non-zero values is defined by its parameters $\boldsymbol{\theta}$. As the mask $m(\cdot\ ; \boldsymbol{\theta})$ is differentiable with regard to its parameters $\boldsymbol{\theta}$, we can learn the interval on which it is non-zero using backpropagation.

In this work, we consider two types of  masks: a Gaussian mask $m_{\mathrm{gauss}}\left(\cdot \ ; \{\mu, \sigma^2\}\right)$ parameterized by its mean and variance $\boldsymbol{\theta}{=}\{\mu, \sigma^2\}$; and a Sigmoid mask $m_\mathrm{sigm}\left(\cdot \ ;\{\mu, \tau\}\right)$ parameterized by its offset and its temperature $\boldsymbol{\theta}{=}\{\mu, \tau\}$ defined as:
\begin{align}
 m_{\mathrm{gauss}}(x \ ; \{\mu, \sigma^2\})&=\left\{ \exp\hspace{-0.5mm}\left( - \tfrac{1}{2} \tfrac{(x - \mu)^2}{\sigma^2}\right) \ \ \text{if}\  \exp\hspace{-0.5mm}\left( - \tfrac{1}{2} \tfrac{(x - \mu)^2}{\sigma^2}\right) \geq T_m; \  0 \  \text{otherwise} \right\}, \label{eq:5_gauss_mask}\\
m_\mathrm{sigm}\left(x \ ;\{\mu, \tau\}\right)&=\left\{1 - \mathrm{sigm}\left(\tau (x - \mu) \right) \ \ \text{if}\ 1 - \mathrm{sigm}\left(\tau (x - \mu) \right)  \geq T_m; \  0 \  \text{otherwise} \right\}, \label{eq:5_sigm_mask}
\end{align}

\begin{wrapfigure}{r}{0.48 \textwidth}
    \centering
    \begin{subfigure}[b]{0.23\textwidth}
         \centering
         \includegraphics[page=5, trim={0.5cm 19cm 6.2cm 2cm}, clip, width=\textwidth]{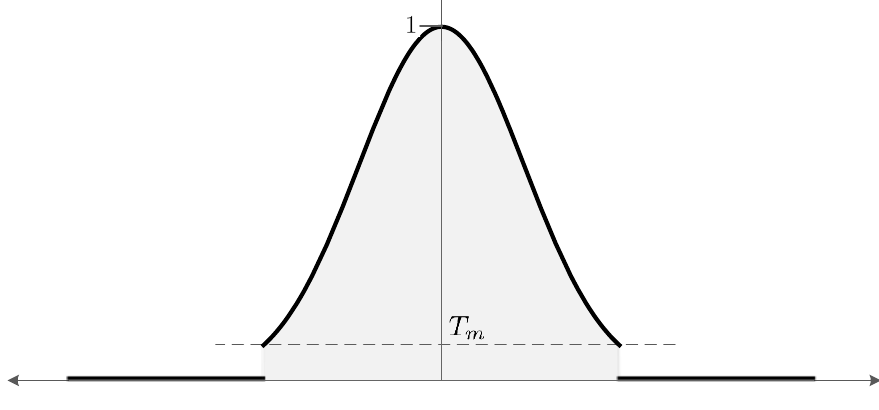}
     \end{subfigure}
     \hfill
     \begin{subfigure}[b]{0.20\textwidth}
         \centering
         \includegraphics[page=6, trim={1.2cm 19cm 6.2cm 2cm}, clip, width=\textwidth]{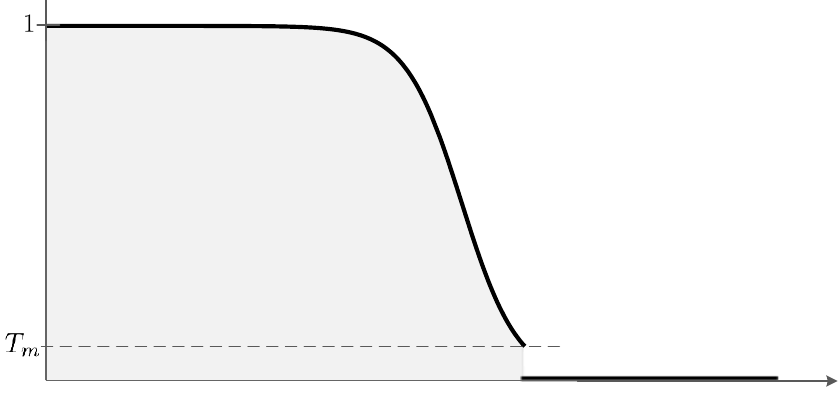}
     \end{subfigure}
     \vspace{-4mm}
     \caption{Gaussian and sigmoid masks.}
    \label{fig:5_diff_masks}
\end{wrapfigure}
where $T_m$ is a predefined threshold below which the mask is zero. These masks are illustrated in Fig.~\ref{fig:5_diff_masks}. To avoid clutter, in the rest of the document we will refer to these masks as $m_{\mathrm{gauss}}$ and $m_{\mathrm{sigm}}$, and will provide specific instantiations when needed.

\textbf{Multidimensional masks.} $\mathrm{N}$-dimensional masks can be constructed by combining $\mathrm{N}$ 1D masks, each with their own parameters. For example, the Gaussian mask used to learn the size of convolutional kernels in Fig.~\ref{fig:5_DNArch_kernels} is constructed as:
\begin{equation}
\setlength{\abovedisplayskip}{4pt}
\setlength{\belowdisplayskip}{4pt}
m_\mathrm{gauss}(x, y; \left\{\{\mu_X, \mu_Y\}, \{\sigma^2_X, \sigma^2_Y\} \right\}) = m_{\mathrm{gauss}}\left(x \ ; \{\mu_X, \sigma^2_X\}\right) \cdot m_{\mathrm{gauss}}\left(y \ ; \{\mu_Y, \sigma^2_Y\}\right)
\end{equation}

\vspace{-7mm}
\subsubsection{Materializing parameters only for non-zero mask values}\label{sec:5_only_nonzero_mask}
Parts of differentiable masks will map to zero based on the value of the parameters $\boldsymbol{\theta}$. Therefore, it would be a waste of compute and memory to materialize the mask --and the corresponding network parameters, e.g., channels $\mathrm{ch} \in [\mathrm{10}, \mathrm{N}]$ in Fig.~\ref{fig:5_figure_1}-- to zero them out next. Luckily, we can take advantage of the invertible form of the Gaussian and Sigmoid masks to materialize parameters only for values for which the mask is non-zero. To this end, we find the value $x_{T_m}$ for which the mask is equal to the threshold $T_m$, i.e., $x_{T_m} {=} x \ \text{such that}\ m(x;  \boldsymbol{\theta}){=} T_m$, and only materialize the mask and the corresponding network parameters for values of $x$ for which the value of the mask is greater than $T_m$. By inverting the mask equations (Eqs.~\ref{eq:5_gauss_mask},~\ref{eq:5_sigm_mask}),~we~obtain~$x_{T_m}$~as:
\begin{center}
\vspace{-7mm}
    \begin{tabular}{p{7cm}p{7cm}}
    %\vspace{1.5mm}
    \begin{equation}
        \pm x_{T_m} = \mu \pm \sqrt{-2 \sigma^2 \log(T_m)}, \label{eq:5_inv_gauss_mask}
    \end{equation}
    &
    \vspace{-3mm}
    \begin{equation}
         x_{T_m} = \mu - \tfrac{1}{\tau}\log\left( \tfrac{1}{1 - T_m} - 1\right),\label{eq:5_inv_sigm_mask}
    \end{equation}
    \end{tabular}
    \vspace{-8mm}
\end{center}
for Gaussian and Sigmoid masks, respectively. Consequently, we can make sure that all rendered values will be used by only materializing the mask and related network parameters for values of $x$ within the range $[-x_{T_m}, x_{T_m}]$ for Gaussian masks and $[x_\mathrm{min}, x_{T_m}]$ for Sigmoid masks, where $x_\mathrm{min}$ depicts the lowest coordinate indexing the mask.
\vspace{-7mm}
\subsection{Continuous Kernel Convolutions}
To prevent finding poor architectures due to insufficiently large receptive fields, it is important for a network to be able to model the global context regardless of specific architectural choices. We rely on Continuous Kernel Convolutions (CKConvs) \cite{romero2022ckconv} to model global dependencies on inputs of arbitrary length, resolution and dimensionality regardless of the network architecture \cite{knigge2023modelling}. CKConvs view convolutional kernels $\boldsymbol{\psi}$ as continuous functions parameterized by a small neural network $\text{\mlp}_{\psi}: \sR^{\Dt} \rightarrow \sR^{\Nin \times \Nout}$ that receives coordinates $\cv_i \in \sR^{\Dt}$ as input and predicts the value of the convolutional kernel at those coordinates: $\cv_i \mapsto \text{\mlp}_{\psi}(\cv_i){=}\psi(\cv_i)$. To construct a kernel of size $\mathrm{K_x}{\times}\mathrm{K_y}$, a CKConv layer constructs a grid of $\mathrm{K_x}{\times}\mathrm{K_y}$ coordinates $[\cv_{(1,1)}, \cv_{(1,2)}, ..., \cv_{(\mathrm{K_x}, \mathrm{K_y})}]$, and passes each coordinate through $\text{\mlp}_{\psi}$ (Fig.~\ref{fig:5_learnable_kernel_sizes}). As a result, CKConvs construct large kernels with few parameters by detaching the size of the kernel from its parameter count.
\vspace{-7mm}
\subsection{The need for learnable architectures}\label{sec:5_need_for_learnable_archs}
General-purpose architectures like Perceiver \cite{jaegle2021perceiver} and the Continuous CNN (CCNN) \cite{romero2022towards} make few assumptions about their input signals, and thus require few architectural changes to handle different tasks. However, the architectures of general-purpose models are static, and thus they are likely not optimal among all the tasks the model might need to solve.
For instance, Perceiver maps inputs to a hidden representation of constant size regardless of the complexity of the task and the input length, resolution and dimensionality. Consequently, it will likely not be able to represent tasks on large inputs correctly. CCNNs, on the other hand, avoid pooling and always perform (global) convolutions on the original input resolution. While this addresses the issue of having a hidden representation of fixed size for inputs with different characteristics, CCNNs can lead to unnecessarily high computational complexity by always performing convolutions at the input resolution. In addition, both the architectures of Perceivers and CCNNs are controlled by non-differentiable hyperparameters, and thus adapting them to a new task requires hyperparameter search across many configurations. 
To address these limitations, we propose to construct neural architectures that tune themselves to fit the requirements of a particular task in a single training run.
\vspace{-7mm}
\subsection{Learning CNN architectures by backpropagation}
Most components of DNArch, such as the learning of the network's width and depth, are not limited to convolutional architectures. However, this research aims to show \textit{how DNArch can be used to learn as many components of a neural architecture as possible}. To that end, we use a general-purpose convolutional architecture: the CCNN \cite{knigge2023modelling}, and make all its architectural components learnable.

\begin{wrapfigure}{r}{0.48\textwidth}
\vspace{-2mm}
    \centering
    \includegraphics[page=8, trim={0.5cm 13cm 3.8cm 0.5cm}, clip, width=0.39\textwidth]{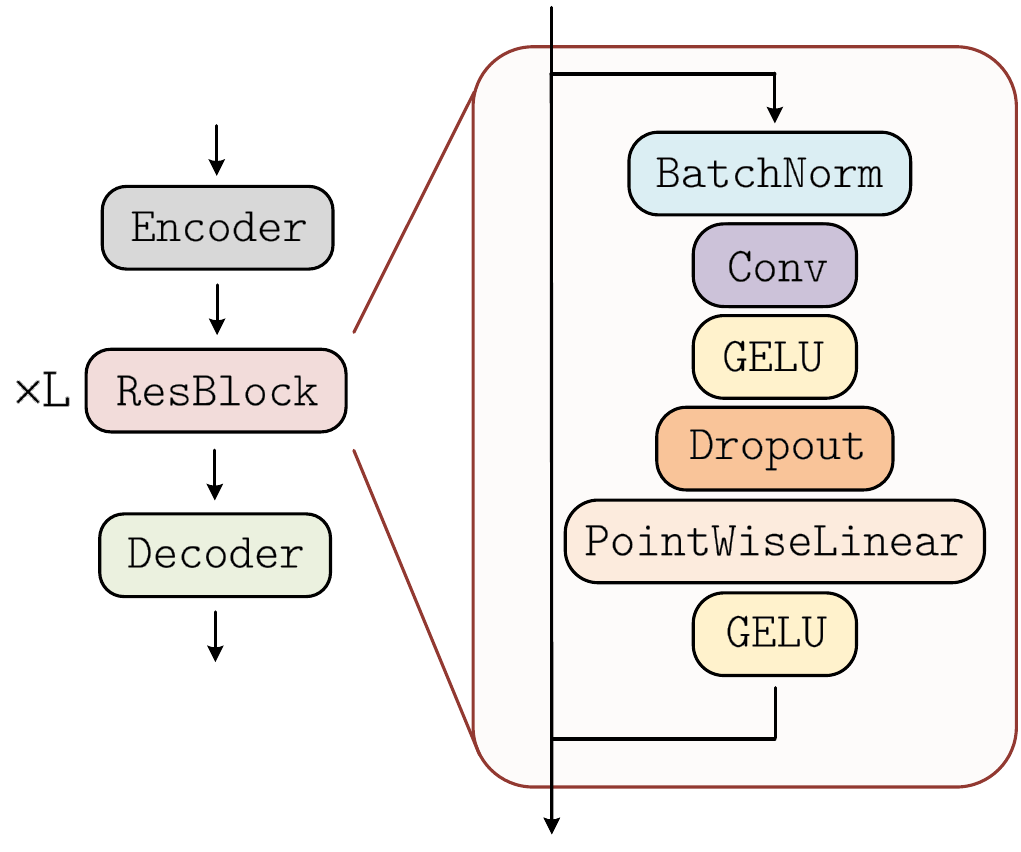}
    \vspace{-3mm}
    \caption{The CCNN architecture \citep{romero2022towards}.
    \vspace{-4mm}}
    \label{fig:5_ccnn_architecture}
\end{wrapfigure}

\textbf{The Continuous CNN \cite{knigge2023modelling}.} The Continuous CNN (CCNN) is a general-purpose convolutional model able to handle inputs of arbitrary dimension, length and resolution without changes. It consists of an {\tt Encoder}, a {\tt Decoder}, and many residual blocks (Fig.~\ref{fig:5_ccnn_architecture}). We refer to the branch on which residual blocks modify the input as \textit{residual branch}, and to the branch connecting the input and the output directly as the \textit{identity branch}. The {\tt Encoder} and {\tt Decoder} adapt the input and output of the model to the goal of the task, e.g., dense / global predictions. Importantly, CCNN's ability to model global context on inputs of any resolution, length and dimensionality makes it an ideal base network for DNArch as (\textit{i}) it prevents the formation of poor architectures due to insufficient receptive fields, and (\textit{ii}) it allows DNArch to be used on tasks on data of arbitrary length and dimensionality without changing the base network --which is needed in existing methods (see NAS-Bench-360 \cite{tu2022bench} for several examples).
\vspace{-7mm}
\subsubsection{Learning the size of convolutional kernels}\label{sec:5_learn_kernel_sizes}
First introduced in FlexConv \cite{romero2022flexconv}, differentiable masking can be combined with CKConvs to learn the size of convolutional kernels by backpropagation. This is done by

\begin{wrapfigure}{r}{0.48 \textwidth}
\vspace{-6mm}
    \centering
    \includegraphics[page=14, trim={2.5cm 15.5cm 20cm 0.5cm}, clip, width=0.45\textwidth]{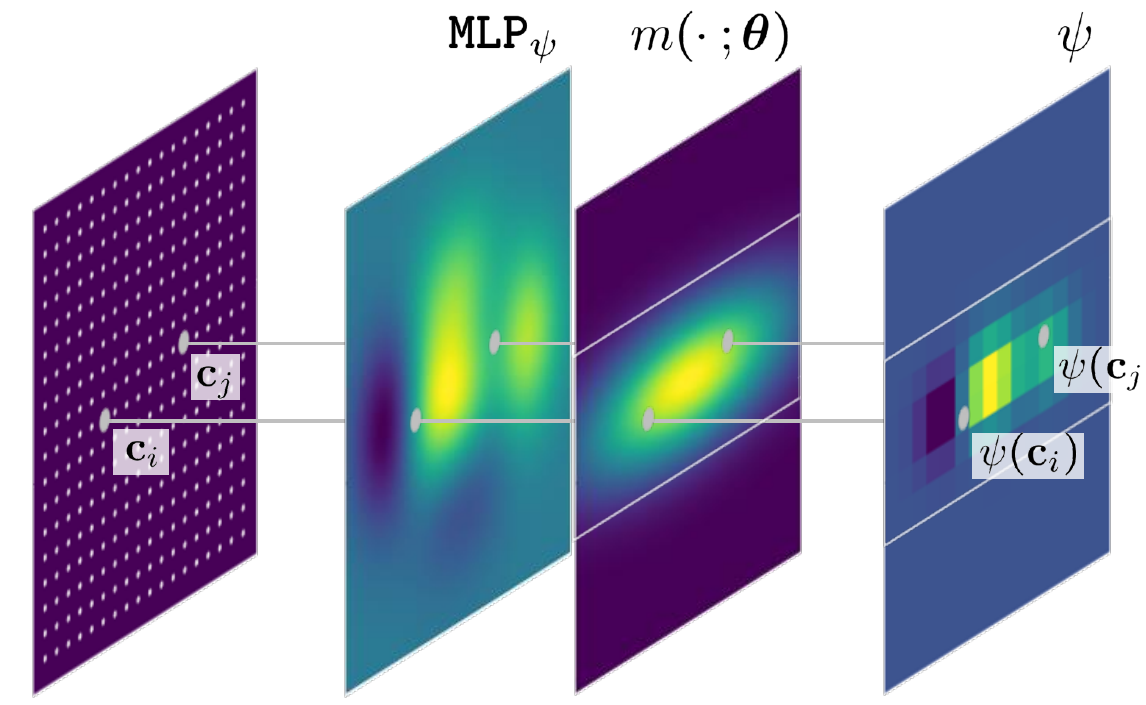}
    \vspace{-4mm}
    \caption{Learning kernel sizes with differentiable masking and CKConvs \citep{romero2022ckconv}.
    \vspace{-5mm}}
    \label{fig:5_learnable_kernel_sizes}
\end{wrapfigure}

  modelling convolutional kernels $\psi$ as the product of a small neural network $\text{\mlp}_\psi$, i.e., a Continuous Kernel Convolution, and a differentiable mask $m(\cdot\ ;\boldsymbol{\theta})$ with learnable parameters, i.e., $\psi(\cv_i) {=} {\text{\mlp}}_\psi(\cv_i) \cdot m(\cv_i; \boldsymbol{\theta})$  (Fig.~\ref{fig:5_learnable_kernel_sizes}). Note that, it is possible to construct the convolutional kernel only for non-zero values of the mask $m(\cv_i; \boldsymbol{\theta})$ by following the method outlined in Sec.~\ref{sec:5_only_nonzero_mask}.

\vspace{-7mm}
\subsubsection{Learning downsampling layers}
We can also use differentiable masking to learn downsampling by applying a differentiable mask on the Fourier domain. The Fourier transform $\gF$ represents a function $f: \sR^{\Dt} \rightarrow \sR$ in terms of its \textit{spectrum} $\tilde{f}: \sR^{\Dt} \rightarrow \sC$, which map frequencies $\omega$ to the amount of that frequency in the input $\tilde{f}(\omega)$. A useful identity in this context is that cropping high frequencies in the Fourier domain equals downsampling in the spatial domain.%By modifying the amplitude $\tilde{f}(\omega)$ of an specific frequency $\omega$, one can directly alter the  \enquote{amount} of that frequency in the input function. 

\begin{figure}
    \centering
    \includegraphics[page=12, trim={0cm 5cm 15cm 0.5cm}, clip, width=0.75\textwidth]{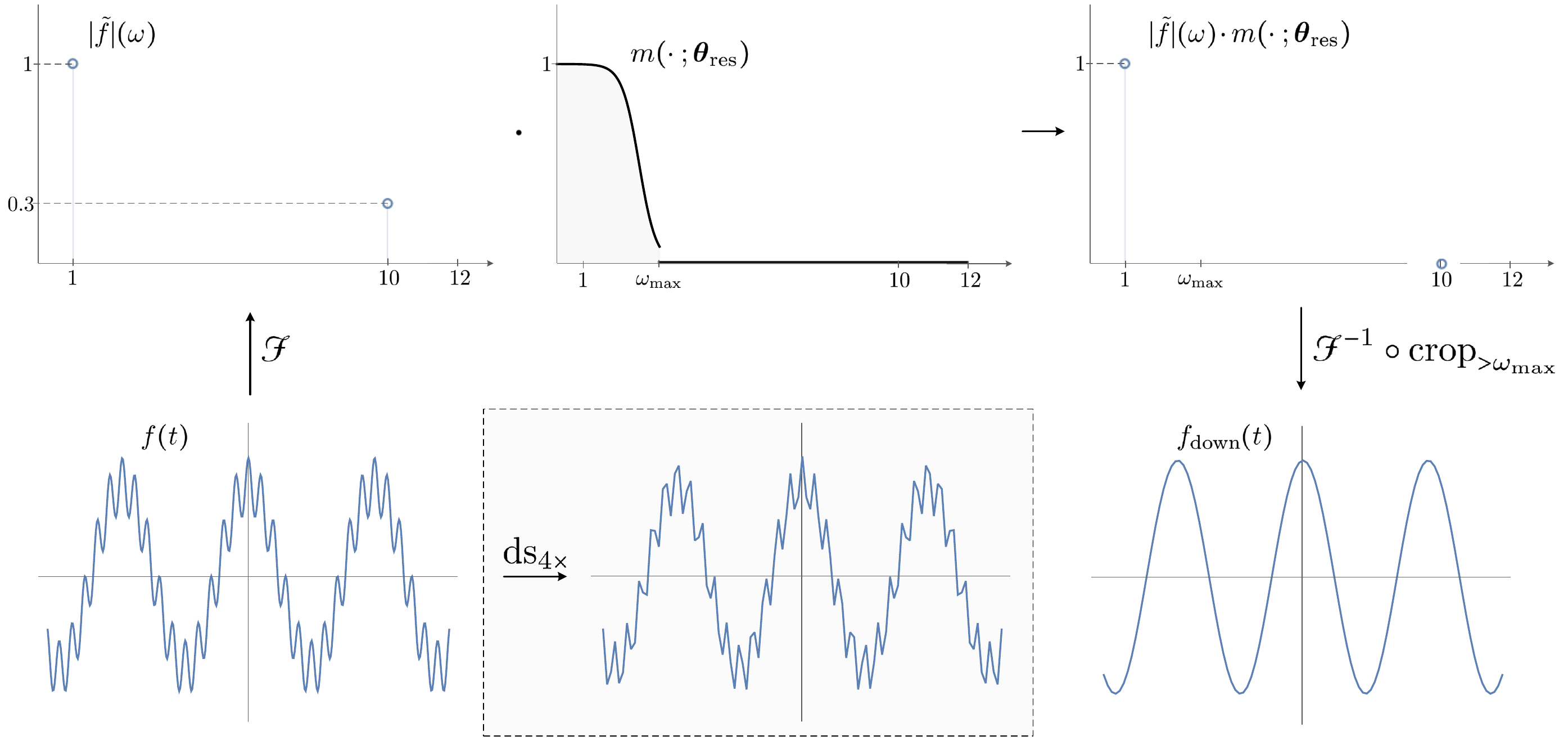}
    \vspace{-2mm}
    \caption{Learning downsampling with differentiable masking on the Fourier domain.}
    \label{fig:5_learnable_downsample}
\end{figure}
To learn downsampling, we use a learnable sigmoid mask $m_\mathrm{sigm}$ to perform a \emph{learnable low-pass filtering} on the input. This is achieved by multiplying the spectrum of the input with the mask $m_\mathrm{sigm}$. By doing so, all frequencies above the mask's cutoff frequency $\omega_\mathrm{max}{=}T_m$ becomes zero (Fig.~\ref{fig:5_learnable_downsample}). An important consequence of low-pass filtering is that as the spectrum of the signal becomes zero above $\omega_\mathrm{max}$, the low-passed signal can be faithfully represented at a lower resolution determined by $\omega_\mathrm{max}$. Letting $\gF$, $\gF^{-1}$ be the Fourier and inverse Fourier transform, $\mathrm{crop}_{> \omega_\mathrm{max}}$ be an operator that crops all values above $\omega_\mathrm{max}$, and $f_\mathrm{down}$ represent the downsampled signal $f$, we have that: 
\begin{equation}
f_\mathrm{down} {=} \gF^{-1}\left[ \mathrm{crop}_{>\omega_\mathrm{max}}\left(\gF[f] \cdot m_\mathrm{sigm}( \ \cdot ; \boldsymbol{\theta}) \right) \right].
\end{equation}
Fig.~\ref{fig:5_learnable_downsample} shows an example of downsampling by a factor of $4\mathrm{x}$. Unlike conventional downsampling, e.g., max-pooling, \textit{spectral downsampling} \cite{rippel2015spectral, riad2022learning} considers the spectral content of the input during downsampling, and thus prevents \textit{aliasing} --where the output resolution is insufficient to accurately represent the underlying signal (Fig.~\ref{fig:5_learnable_downsample}, middle down)--. This is important since it has been shown that aliasing has negative effects on robustness \cite{zhang2019making}, generation \cite{karras2021alias}  and generalization \cite{vasconcelos2021impact}.

\textbf{Combining learnable downsampling and convolution.} The previous method requires mapping inputs to the Fourier domain and back to learn downsampling. Fortunately, CCNNs as well as most methods that rely on global convolutions, e.g., CKConv \citep{romero2022ckconv}, S4 \citep{gu2022efficiently}, rely on the \textit{Fourier convolution theorem}: $(f * \psi) {=} \gF^{-1}\left[\gF[f] \cdot \gF[\psi] \right]$ to compute convolutions with large kernels efficiently. This means that CCNNs already use a Fourier and inverse Fourier transforms in each residual block to compute convolutions. Hence, we can avoid recomputing these steps by placing the learnable downsampling operation \textit{within the Fourier convolution}. Specifically, we simultaneously compute downsampling and convolution by applying the differentiable mask $m_\mathrm{sigm}$ and the cropping operations $\mathrm{crop}_{>\omega_\mathrm{max}}$ before returning from the Fourier domain back to the spatial domain:\footnote{We note that the Fourier transform is not strictly necessary to learn downsampling, e.g., for CNNs with local kernels. Leveraging the Fourier convolution theorem, equivalent downsampling can be achieved by convolving the input with the inverse Fourier transform of the mask in the spatial domain (see Appx.~\ref{appx:5_downsample_no_fourier}).}
\begin{equation}
    (f * \psi)_\mathrm{down} = \gF^{-1}\left[\mathrm{crop}_{>\omega_\mathrm{max}}(m_\mathrm{sigm}(\cdot\ ;\boldsymbol{\theta}) \cdot \gF[f] \cdot \gF[\psi]) \right].\label{eq:5_conv_and_downsampling}
\end{equation}
% 
% \textbf{Computational implications of the Fourier convolution.} The \textit{Fourier convolution theorem} states that the convolution between a function $f$ and a convolutional kernel $\psi$ is equivalent to the pointwise multiplication of their Fourier transforms $\gF[f]$, $\gF[\psi]$. This means that we can compute a convolution by taking the Fourier transform of both signals, multiplying them in the Fourier domain, and then transforming them back to the spatial domain:
% \begin{equation}
% \setlength{\abovedisplayskip}{4pt}
% \setlength{\belowdisplayskip}{4pt}
%     (f * \psi) {= } \gF^{-1}\left[\gF[f] \cdot \gF[\psi] \right].
% \end{equation}
% The computational complexity of this method is $\Lt \log(\Lt)$, which is faster than the complexity of a standard convolution ($\Lt \cdot \Kt$) when $\Kt>\log(\Lt)$. This is why Fourier convolutions are widely used in methods that rely on global convolutions, e.g., CKConv \cite{romero2022ckconv}, S4 \cite{gu2022efficiently}.
\textbf{Materializing functions only on the output resolution.} Note that Eq.~\ref{eq:5_conv_and_downsampling} computes the convolution on the resolution of the input and downsamples next. This incurs in an unnecessary overhead as the output of the convolution will be downsampled directly after. A more efficient approach comes from inverting the order of these operations to compute the convolution at the downsampled resolution. Luckily, this can be achieved by using the method outlined in Sec.~\ref{sec:5_only_nonzero_mask}. Since the cutoff frequency of the mask corresponds to the coordinate at which the mask equals the threshold, i.e., $\omega_\mathrm{max} {=} x_{T_m}$, it can be calculated using Eq.~\ref{eq:5_inv_sigm_mask}. Next, since the cutoff frequency defines the minimum resolution required to faithfully represent the input, we can downsample the input and the kernel to that resolution before the convolution to compute it on the output resolution.

\textbf{Learning subsampling for dense tasks.}  \citet{riad2022learning} apply learned downsampling on both the identity and the residual branches of a residual block to limit resolution of all representations after a specific residual block. In the context of DNArch, this is undesirable for two reasons: First, as we learn the whole network architecture during training, it is not known a priori what resolution mappings will require at each layer. However, forcing the identity branch to have the same resolution as the corresponding residual branch restricts all subsequent mappings to be of maximum that resolution. Secondly, dense prediction tasks, e.g., segmentation, require the learned architecture to produce outputs that share the same resolution as the input. However, if the identity branch is also downsampled, the output of the network would be of lower resolution even for a single level of downsampling in the network. This in turn, would result in over-smoothed predictions.

Based on these observations, we use downsampling \textit{only} on the residual branch and upsample features at the end of each residual block back to the resolution of the input. This allows us to (\textit{i}) have features at same resolution as the input in the last layer, and (\textit{ii}) learn U-Net \cite{ronneberger2015u} like architectures. %It is important to note that upsampling can be done efficiently in Deep Learning frameworks, e.g., in \href{https://jax.readthedocs.io/en/latest/index.html}{{\tt JAX}} via \texttt{jax.image.resize}.

\vspace{-7mm}
\subsubsection{Learning the width of the network at each layer}
We learn the width of a layer by applying a differentiable mask $m(\cdot\ ;\boldsymbol{\theta})$ along the channel dimension of feature representations (see Fig.~\ref{fig:5_figure_1}). The network can then adjust the width of its representations by changing the value of $\boldsymbol{\theta}$. By using the method in Sec.~\ref{sec:5_only_nonzero_mask} only non-masked channels are rendered.

\begin{wrapfigure}{r}{0.48\textwidth}
\vspace{-3mm}
    \centering
    \includegraphics[page=9, trim={13.8cm 2.5cm 16cm 0.5cm}, clip, width=0.39\textwidth]{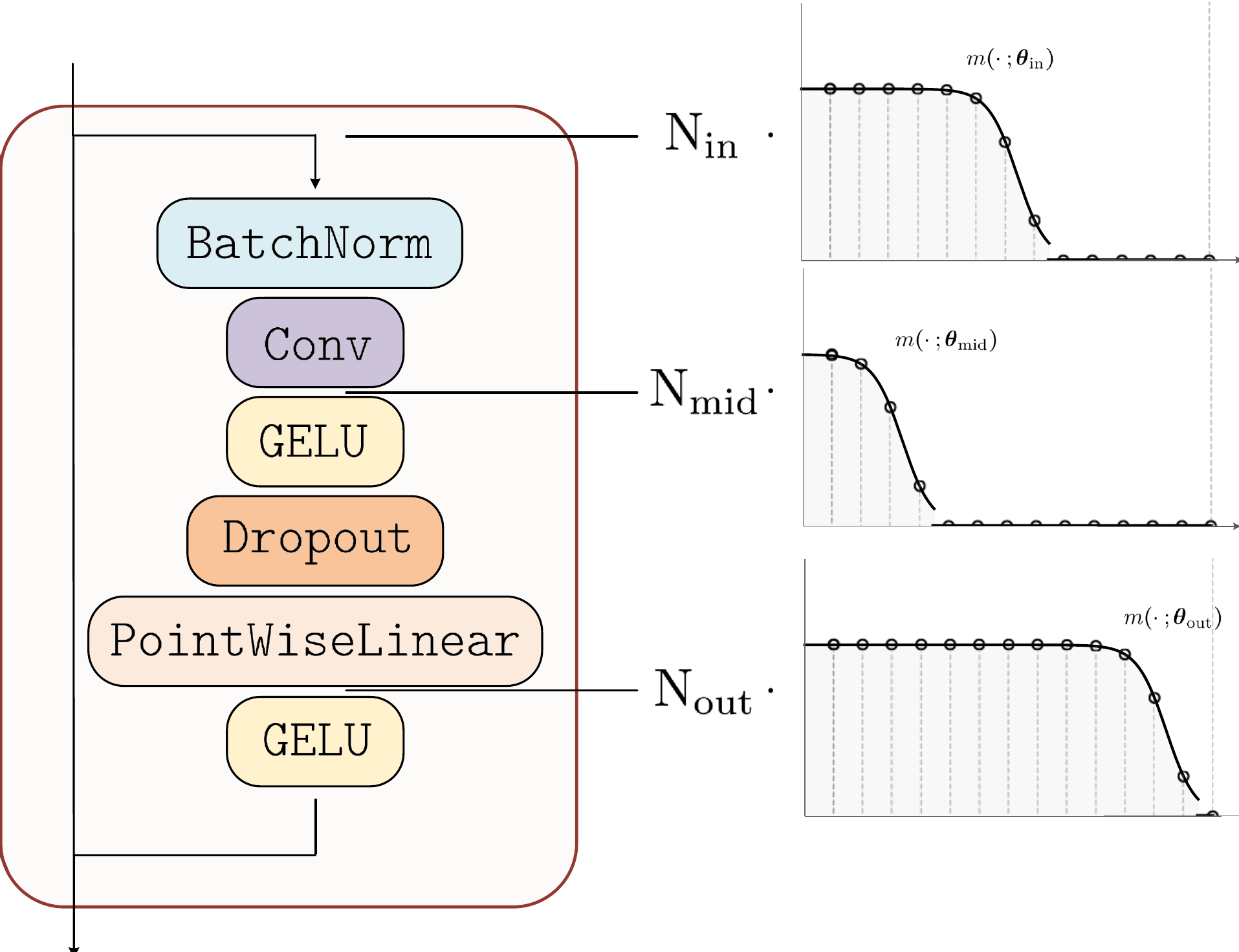}
    \vspace{-2mm}
    \caption{Positioning of width masks.}
    \label{fig:5_width_learning}
\end{wrapfigure}

\textbf{Positioning of the width masks.} We aim to learn the width of all layers in a network. To this end, we apply differentiable masks with independent learnable parameters along the channel dimensions of all the network components that change the network's width, i.e., all {\tt Conv} and {\tt PWLinear} layers. This corresponds to learning three independent masks for each residual block in the network, which correspond to the input ($\mathrm{N_{in}}$), the middle ($\mathrm{N_{mid}}$) and the output ($\mathrm{N_{out}}$) channels of the residual block (Fig.~\ref{fig:5_width_learning}). 
Components that do not change the width, e.g., {\tt BatchNorm}, {\tt GELU}, have their width determined by the preceding mask.

\textbf{The advantage of avoiding masks on the identity branch.} Applying a differentiable mask on the output of the entire residual block to constrain its width, i.e., after the sum of the residual and identity branches, would accumulate the effect of all masks applied before that block. As a result, the $l$-th block would be effectively masked by the combination of all masks before that block, i.e., $\prod_{i \leq l} m_i$, with $m_i$ the mask after the $i$-th block. Since the values of the masks live in the $[0, 1]$ interval, this would result in an exponential decrease in the magnitude of the hidden feature presentations. To avoid this, we apply the masks on the residual branch only. In addition, keeping the identity branch intact allows DNArch to construct DenseNet-like architectures \cite{huang2017densely}, where blocks can reuse channels that only have been modified by some --or none-- of the previous blocks.
\vspace{-7mm}
\subsubsection{Learning the depth of the network}\label{sec:5_learning_depth}

\begin{wrapfigure}{r}{0.4 \textwidth}
\vspace{-3mm}
    \centering
    \includegraphics[page=10, trim={0.5cm 6cm 8.5cm 0.5cm}, clip, width=0.175\textwidth]{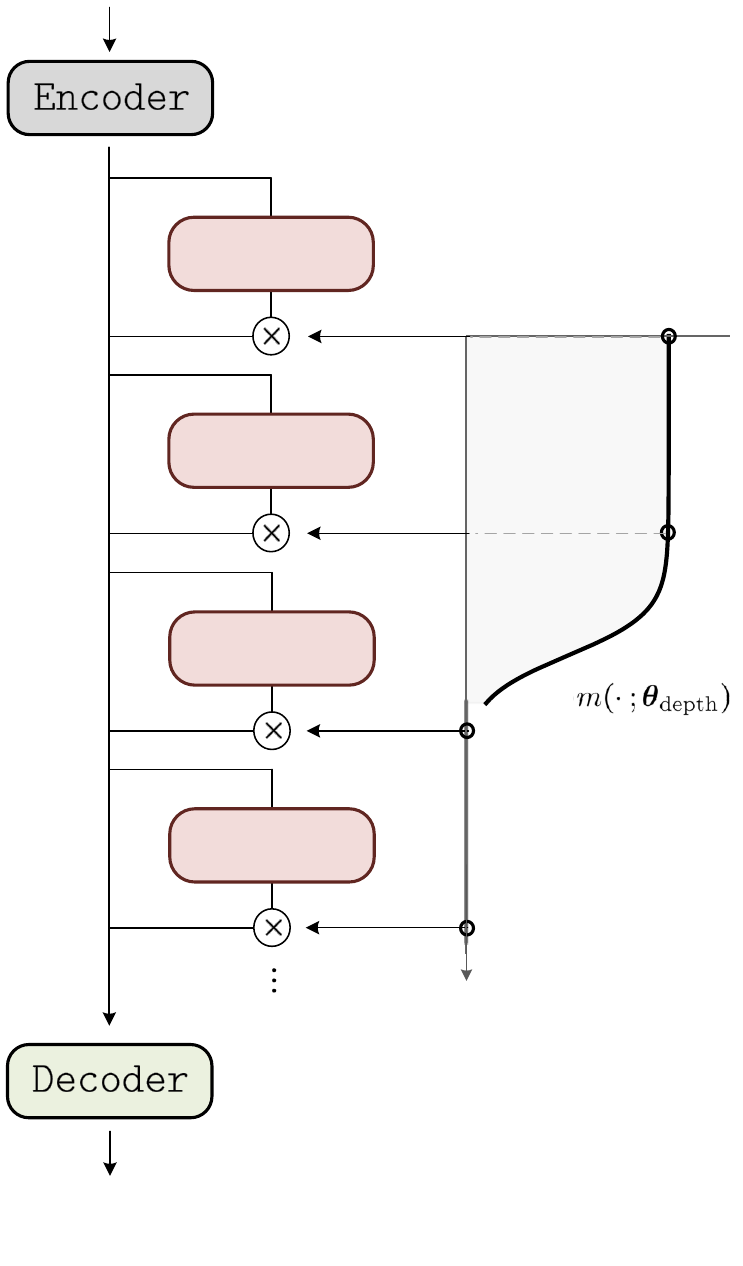}
    \vspace{-4mm}
    \caption{Learning network's depth with differential masking.
    \vspace{-4mm}}
    \label{fig:5_depth_learning}
\end{wrapfigure}

We learn the network's depth by viewing the number of residual blocks as a continuous axis with values $[1, 2, .., \Dt]$ corresponding to the index of each block, and using a differentiable mask $m(\cdot, \boldsymbol{\theta})$ along this axis to dynamically mask out blocks based on the value of the mask parameters $\boldsymbol{\theta}$ (Fig.~\ref{fig:5_depth_learning}).\newpage

\textbf{Positioning of the depth mask.} To ensure that information flows from the input to the output of the network regardless of the value of the mask parameters, we \textit{only} apply the mask on the residual branch. If the mask were also applied on the identity branch, feature representations at the end of the network could become zero, and the network would only be able to output random predictions.% Maintaining an active path between inputs and outputs through the identity branch allows training the model with a loss computed only at the output. This is unlike early-exit architectures~\citep{saeed_multi_exit,pondernet,schuster2022confident} that can adapt their depth on a per-example basis, but at the cost of adding an extra output layer (which can be computationally expensive e.g. for large-scale classification) at every layer.

\vspace{-7mm}
\subsubsection{Learning entire convolutional architectures by backpropagation}
By simultaneously using the methods outlined in this section, DNArch uses backpropagation to learn the weights, the size of convolutional kernels at each layer, the number of channels at each layer, the position and resolution of downsampling layers, and the depth of a convolutional network.
\vspace{-7mm}
\subsection{Learning neural architectures under computational constraints}\label{sec:5_constrained_DNArch}
We can ensure that the architectures searched by DNArch respect a predefined computational complexity by including an additional regularization term $\gL_\mathrm{comp}$ that reflects the complexity of the current candidate architecture based on its mask parameters. To this end, we define the optimization loss $\gL$ as the sum of the task objective loss $\gL_\mathrm{obj}$ and the complexity loss $\gL_\mathrm{comp}$~weighted~by~a~factor~$\lambda$:
\begin{equation}
    \gL = \gL_\mathrm{obj} + \lambda \ \gL_\mathrm{comp}.\label{eq:5_optim_loss}
\end{equation}
By minimizing this loss, DNArch is encouraged to find architectures that meet the desired computational budget while still achieving good performance on the end task.

\vspace{-7mm}
\subsubsection{Defining the complexity loss $\gL_\mathrm{comp}$} 
The purpose of $\gL_\mathrm{comp}$ is to use the size of the learned masks to estimate the total computation needed for a forward pass of the network. Its construction is outlined below.

\textbf{Layer-wise complexities.} Let $\gC_{\mathrm{layer}}(\mathrm{L}, \Nin, \Nout)$ be the number of operations required in a given layer with an input of length $\mathrm{L}$ and $\Nin$ and $\Nout$ input and output channels. To estimate the number of computations required based on the current size of the masks, we can substitute the lenght of each dimension with the size of the corresponding masks: $\gC_{\mathrm{layer}}\left(\mathrm{size}(m_\mathrm{res}), \mathrm{size}(m_\Nin), \mathrm{size}(m_\Nout)\right)$. As an example, consider a pointwise linear layer $\mathrm{lin}: \sR^{\Nin} \rightarrow \sR^{\Nout}$. It takes an input $f$ of length $\Lt$ and $\Nin$ channels and multiplies each element along the spatial dimensions of the input with a matrix of dimensions $[\Nin, \Nout]$ to produce an output of the same length, but with $\Nout$ number of channels.
The total operations required in this layer is given by $
\gC_{\mathrm{lin}}(f) {=} \Lt \cdot \Nin \cdot \Nout$.

Now, let us use three differentiable masks $m_\mathrm{res}$, $m_\Nin$ and $m_\Nout$ to mask the resolution, input and output channels of the linear layer. The total number of computations is now:
\begin{equation*}
   \gC_{\mathrm{lin, masked}} =  \mathrm{size}(m_\mathrm{res}) \cdot \mathrm{size}(m_\Nin) \cdot \mathrm{size}(m_\Nout).\nonumber
\end{equation*}
Since the size of the masks is now involved in the computation of the operations required, we can utilize it as an additional source of feedback to update the masks by making the function $\mathrm{size}$ differentiable with regard to the mask parameters. The same concept is used to calculate the cost of other layers based on the size of the masks. A summary of these costs can be found in Appx.~\ref{appx:5_masking_other_layers}.

\textbf{Effect of the depth mask.} To take into account the effect of the depth mask, we use it to determine the number of residual blocks in the network. If the number of operations of a network with $\mathrm{D}$ residual blocks is denoted as $\gC_\mathrm{net, D}$, the complexity of a network with masked depth is given by $\gC_\mathrm{net, \mathrm{size}(m_\mathrm{depth})}$ with $\mathrm{size}(m_\mathrm{depth})$ the size of the depth mask.

\textbf{Computing the $\mathrm{size}$ of the masks.} The size of a mask can be calculated in a differentiable manner by determining the length of the mask in continuous space and using that length to estimate the change in size of the corresponding network dimension. Specifically, the length at a time $t$ is $2 x_{T_m}^{t}$ and $x_{T_m}^{t} {-} x_\mathrm{min}$, for Gaussian and Sigmoid masks, respectively (see Fig.~\ref{fig:5_diff_masks}). For some initial $x^{0}_{T_m}$ and corresponding initial length $\Nt$, the size of a Gaussian and a Sigmoid mask at time $t$ is respectively:
\begin{center}
\vspace{-5mm}
    \begin{tabular}{p{5cm}p{6cm}}
    \begin{equation}
        \mathrm{size}(m_\mathrm{gauss}) = \tfrac{2 x^{t}_{T_m}}{2 x^{0}_{T_m}} \Nt, \label{eq:5_size_gauss}
    \end{equation}
    &
    \begin{equation}
         \mathrm{size}(m_\mathrm{sigm}) = \tfrac{x^{t}_{T_m} - x_\mathrm{min}}{ x^{0}_{T_m} - x_\mathrm{min}} \Nt.\label{eq:5_size_sigm}
    \end{equation}
    \end{tabular}
    \vspace{-5mm}
\end{center}
\textbf{Computational constraints as an additional loss.} Let $\gC_{\mathrm{curr}}$ be the current complexity of the network and $\gC_{\mathrm{target}}$ be the desired target complexity. We define the computational loss $\gL_\mathrm{comp}$ as the relative $\mathcal{l}^2$ difference between the relative complexity of the current network and the target:
\begin{equation}
    \mathcal{l}^{2}\left(\tfrac{\gC_{\mathrm{curr}}}{\gC_{\mathrm{target}}}, 1\right) = \left\| \tfrac{\gC_{\mathrm{curr}}}{\gC_{\mathrm{target}}} - 1.0 \right\|^2_2.
\end{equation}
This form has two advantages over the alternative form $\mathcal{l}^2(\gC_{\mathrm{curr}}, \gC_{\mathrm{target}})$. It (\textit{i}) prevents overflow that might occur when comparing large values --$\gC_{\mathrm{curr}}$ and $\gC_{\mathrm{target}}$ may easily be of order $1\mathrm{e}^{10}$--, and (\textit{ii}) allows for consistent tuning of $\lambda$ for different tasks and complexities. In the alternative form $\mathcal{l}^2(\gC_{\mathrm{curr}}, \gC_{\mathrm{target}})$, $\lambda$ might need to be tuned independently for different complexity regimes.

\vspace{-7mm}
\section{Experiments} \label{sec:5_exps}
\begin{table*}
\centering
    \begin{minipage}{\textwidth}
    \centering
    \caption{Performance on the LRA benchmark. $\times$ denotes random guessing. Highest per-section scores are in bold and the overall best scores are underlined. For DNArch, values in parenthesis indicate the computational cost of the architecture relative to the target complexity.}
    \label{tab:5_lra_results}
    \vspace{-2.5mm}
    \begin{small}
    \scalebox{0.78}{
    \begin{tabular}{llllllll}
    \toprule
    \sc{Model} & \sc{ListOps}  & \sc{Text} &  \sc{Retrieval} & \sc{Image} & \sc{Pathfinder} & \sc{Path-X} & \sc{Avg.} \\
     \midrule
    Transformer \cite{vaswani2017attention} & \textbf{36.37} & 64.27 &  57.46 & 42.44 & 71.40 & $\times$& 53.66\\
    Reformer \cite{kitaev2020reformer} & 37.27 & 56.10 & 53.40 & 38.07 & 68.50 & $\times$& 50.56\\
    %Linear Trans. & 16.13 & 65.90 & 42.34 & 75.30 & 50.46  & $\times$&$\times$\\
    Performer \cite{choromanski2020rethinking} & 18.01 &\textbf{ 65.40} & 53.82 & \textbf{42.77} & \textbf{77.05}  & $\times$& 51.18 \\
    BigBird \cite{zaheer2020big} & 36.05 & 64.02 & \textbf{59.29} & 40.83 & 74.87  & $\times$& \textbf{54.17} \\
    \midrule
    Mega ($\mathcal{O}(L^2)$) \cite{ma2022mega} & \textbf{\underline{63.14}} & \underline{\textbf{90.43}} & \textbf{91.25} & \textbf{\underline{90.44}} &  \textbf{\underline{96.01}} & \textbf{97.98} & \textbf{\underline{88.21}}\\
    Mega-chunk ($\mathcal{O}(L)$) \cite{ma2022mega} & 58.76 & 90.19 & 90.97 &  85.80 & 94.41 & 93.81 &  85.66 \\
    \midrule
    S4D \cite{gu2022parameterization} & 60.47 & 86.18 & 89.46 & 88.19 & 93.06 & 91.95 & 84.89\\
    S4 \cite{gu2022efficiently} & 59.60 & 86.82 & 90.90 & \textbf{88.65} & 94.20 & 96.35 & 86.09\\
    S5 \cite{smith2022simplified} & \textbf{61.50} & \textbf{89.31} & \textbf{\underline{91.40}} & 88.00 & \textbf{95.33} & \textbf{\underline{98.58}} & \textbf{87.35}\\
    % Liquid S4 & \textbf{62.75} & 89.02 & 91.20 & 89.50 & 94.80 & 96.66 & 87.32\\
    % \midrule
    % CCNN$_{4, 140}$ \cite{romero2022towards} & 44.85 & 83.59 & $\times$ & 87.62 & 91.36 & $\times$ & 76.86\\
    \midrule
        FNet \cite{lee2021fnet} &  35.33 & 65.11 & 59.61 & 38.67 & 77.80 & $\times$ & 54.42\\
    %Nystromförmer &37.15 &65.52 &41.58 &70.94 &57.46  & $\times$&$\times$ \\
    Luna-256 \cite{ma2021luna} & 37.25 & 64.57 & 79.29 & 47.38 & 77.72 & $\times$ & 59.37 \\
    CCNN$_{4, 140}$ \cite{romero2022towards} & \textbf{44.85} & \textbf{83.59}  & $\times$ & \textbf{87.62} & \textbf{91.36} & $\times$ & \textbf{76.86} \\
    % SGConv & \textbf{61.45} & \textbf{89.20} &  \textbf{91.11} &  \textbf{87.97} & \textbf{95.46} & \textbf{97.83} & \textbf{87.17} \\
    \midrule
    \midrule
    CCNN$_{4, 140}$ (Global Kernels) & 55.65 & 87.80 & 90.55 & 85.51 & 94.26 & 91.15 & 84.15\\
    $\mathrm{DNArch}_\mathrm{K}$(CCNN$_{4, 140}$) & 59.90 & 88.28& 90.66  & 86.07 & 93.46 & 89.93 & 84.72\\
    $\mathrm{DNArch}_\mathrm{K,R}$(CCNN$_{4, 140}$) & 60.15$_{(0.80\times)}$ & 88.50$_{(0.75\times)}$ & 91.08$_{(0.78\times)}$  & 86.55$_{(0.82\times)}$ & 94.05$_{(0.89\times)}$ & 91.15$_{(0.82\times)}$ & 85.25\\
    % $\mathrm{DNArch}$(T: CCNN$_{4, 140}$)$_{\mathrm{K,R,C}}$ & \\
    $\mathrm{DNArch}_\mathrm{K,R,W,D}$(CCNN$_{4, 140}$) & \textbf{60.55}$_{(1.01\times)}$ & \textbf{89.03}$_{(1.00\times)}$ & \textbf{91.22}$_{(1.02\times)}$ & \textbf{87.20}$_{(1.02\times)}$ & \textbf{94.95}$_{(1.00\times)}$ & \textbf{91.71}$_{(1.01\times)}$ & \textbf{85.78} \\
    %\midrule
    %CCNN$_{6, 380}$ (Global Kernels) & 59.60 & 88.10 & 90.59 & 86.70 & 95.24 & $\times$ & $\times$\\
    %$\mathrm{DNArch}_\mathrm{K,R,W,D}$(CCNN$_{6, 380}$) & \textbf{61.05}$_{(1.00\times)}$ & \textbf{89.13}$_{(1.00\times)}$ & \textbf{91.32}$_{(1.00\times)}$ & \textbf{89.53}$_{(1.00\times)}$ & \textbf{96.00}$_{(1.00\times)}$ & $\times$ & $\times$\\
    \bottomrule
    \end{tabular}}
    \end{small}
    \end{minipage}
\end{table*}
We evaluate DNArch on sequential and image datasets for classification and dense prediction tasks. On 1D, we use the Long Range Arena (LRA) benchmark \citep{tay2021long}, which includes six sequence modelling tasks with sequence lengths ranging from 1024 to over 16000. On 2D, we perform image classification on the CIFAR10 and CIFAR100 datasets \citep{krizhevsky2009learning} and report results on two dense prediction tasks from the NAS-Bench-360 benchmark \citep{tu2022bench}: {\tt DarcyFlow} \citep{li2020fourier} and {\tt Cosmic} \citep{zhang2020deepcr}. A detailed description of the datasets used can be found in Appx.~\ref{appx:5_datset_description}.

\textbf{Experimental setup.} We use two CCNNs of different capacity as base networks: a CCNN$_{4,140}$ --4 blocks, 140 channels, 200{\sc{k}} parameters--, and a CCNN$_{6,380}$ --6 blocks, 380 channels, 2{\sc{m}} parameters--, and use DNArch to learn their architectures. To understand the impact of learning each network component, we also report results learning some and none of the neural architecture components. 

\textbf{Mask configurations.} We initialize all the masks to match the architecture of the baseline CCNNs at the beginning of training. We use a Gaussian mask to learn kernel sizes as in FlexConv \citep{romero2022flexconv}, and Sigmoid masks to learn width, depth and downsampling. All masks use a threshold of $T_m{=}0.1$. All kernel masks are centered, i.e., $\mu{=}0$, and initialized to either be small or global, i.e., $\sigma \in [0.0325, 0.5]$. Resolution masks are initialized to weight the highest input frequency by $0.85$, and the width and depth masks are initialized to match the size of the base network's architecture. More information on hyperparameters, training regimes, and experimental settings can be found in Appx.~\ref{appx:5_exp_details}. 

\textbf{Notations.} We use $\mathrm{DNArch}$ as an operator acting on a base network and specify the learned components with indices $\mathrm{K, R, W, D}$ representing kernel sizes, downsampling, width and depth. $\mathrm{DNArch}_\mathrm{K}$(CCNN$_{4, 140}$) indicates using DNArch to learn only the kernel sizes of a CCNN$_{4, 140}$.
\vspace{-7mm}
\subsection{Using DNArch without computational constraints}\label{sec:5_no_constr}
First, we use DNArch to improve the expressiveness and computational efficiency of a CCNN$_{4, 140}$. We start using DNArch to learn the receptive field of all convolutional layers, and then we learn both the kernel sizes and downsampling layers to simultaneously improve the expressiveness and the computational efficiency of the CCNN$_{4,140}$. It is worth noting that in this scenario learning is solely driven by the objective loss $\gL_\mathrm{obj}$, i.e., the regularization term $\gL_\mathrm{comp}$ is not used. In addition, as we use Fourier convolutions, the learned kernel sizes do not impact computational efficiency.

\textbf{Results.} Except for {\tt PathFinder} and {\tt Path-X}, we find that using DNArch to learn kernel sizes consistently improves the accuracy of the base architecture (DNArch$_\mathrm{K}$ models in Tabs.~\ref{tab:5_lra_results}-\ref{tab:5_image_classif}). Interestingly, found DNArch architectures perform on par, and even surpass, architectures specifically designed for each tasks, e.g., S4 \citep{gu2022efficiently} for sequential tasks and NFOs \cite{li2020fourier} for PDEs on 2D with a remarkably lower number of trainable parameters. In contrast to DARTS methods, e.g., DASH \cite{shen2022efficient}, DNArch can be applied across \textit{all tasks} without the need to \textit{manually change} the base architecture. When additionally learning downsampling, we observe that DNArch finds high-performant architectures with improved computational efficiency (DNArch$_\mathrm{K, R}$ models in Tabs.~\ref{tab:5_lra_results}-\ref{tab:5_image_classif}). Interestingly, we observe that the found models often exhibit slight accuracy improvements. This is explained by low resolution kernels being easier to model and construct than higher resolution ones.
\begin{table}
\RawFloats
\begin{minipage}{.48\textwidth}
  \centering
  \caption{Dense prediction results.}
    \label{tab:5_nas_bench_360}
    \vspace{-3mm}
    \begin{small}
    \scalebox{0.8}{
    \begin{tabular}{lll}
    \toprule
     \multirow{2}{*}{\sc{Model}} & \sc{DarcyFlow}  & \sc{Cosmic} \\
     &  rel. $\mathcal{l}^2$ loss & 1 - \sc{auroc} \\
     \midrule
     Expert* & \textbf{0.008} & \textbf{ 0.13}\\
     \midrule
     WRN \cite{zagoruyko2016wide} & 0.073 & 0.24 \\
     DenseNAS \cite{fang2020densely} & 0.100 & 0.38\\
     DARTS \cite{liu2018darts} & \textbf{0.026} & 0.229 \\
     Auto-DL \cite{liu2019auto} & 0.049 & 0.495 \\
     DASH \cite{shen2022efficient} & 0.060 & \textbf{0.190}\\
    \midrule
    \midrule
    % CCNN$_{4, 140}$ \cite{romero2022towards}&  & $\times$ & $\times$  \\
    CCNN$_{4, 140}$ (Global Kernels) &  0.002989 & 0.059\\
    $\mathrm{DNArch}_\mathrm{K}$(CCNN$_{4, 140}$) & 0.002970 & 0.058\\
    $\mathrm{DNArch}_\mathrm{K,R}$(CCNN$_{4, 140}$) & 0.002929$_{(0.79\times)}$ & 0.056$_{(0.82\times)}$ \\
    %$\mathrm{DNArch}_\mathrm{K,R,W,D}$(CCNN$_{4, 140}$) \\
    $\mathrm{DNArch}_\mathrm{K,R,W,D}$(CCNN$_{4, 140}$) & \textbf{0.002285}$_{(1.01\times)}$ & \textbf{0.055}$_{(1.01\times)}$\\
    \midrule
    CCNN$_{6, 380}$ (Global Kernels) &  0.004521 & 0.059\\
    $\mathrm{DNArch}_\mathrm{K,R,W,D}$(CCNN$_{6, 380}$) & \textbf{\underline{0.001763}}$_{(1.00\times)}$ & \textbf{\underline{0.048}}$_{(1.00\times)}$ \\
    \bottomrule
    \multicolumn{3}{l}{$^*$ FNO \cite{li2020fourier} and deepCR \cite{zhang2020deepcr}.}
    \end{tabular}}
    \end{small}
\end{minipage}\hspace{6mm}
\begin{minipage}{0.48\textwidth}
    \centering
    \caption{Image classification results.}
    \label{tab:5_image_classif}
    \vspace{-3mm}
    \begin{small}
    \scalebox{0.8}{
    \begin{tabular}{lll}
    \toprule
     \sc{Model} & \sc{CIFAR10} & \sc{CIFAR100} \\
     \midrule
     WRN \cite{zagoruyko2016wide} &  - & \textbf{\underline{76.65}} \\
     DenseNAS \cite{fang2020densely} & - & 74.51 \\
     DARTS \cite{liu2018darts} & - & 75.98 \\
     Auto-DL \cite{liu2019auto} & - & - \\
     DASH \cite{shen2022efficient} & - & 75.63 \\
    \midrule
    \midrule
    CCNN$_{4, 140}$ (Global Kernels) & 90.52 & 64.72\\
    $\mathrm{DNArch}_\mathrm{K}$(CCNN$_{4, 140}$) & 92.51 & 69.01\\
    $\mathrm{DNArch}_\mathrm{K,R}$(CCNN$_{4, 140}$) & 92.77$_{(0.82\times)}$ & 68.96$_{(0.85\times)}$  \\
    % $\mathrm{Diff}$(T: CCNN$_{4, 140}$)$_{\mathrm{K,R,C}}$ \\
    $\mathrm{DNArch}_\mathrm{K,R,W,D}$(CCNN$_{4, 140}$) & \textbf{93.47}$_{(1.01\times)}$ & \textbf{72.98}$_{(1.03\times)}$  \\
    \midrule
    CCNN$_{6, 380}$ (Global Kernels) & 94.18 & 72.29\\
    $\mathrm{DNArch}_\mathrm{K,R,W,D}$(CCNN$_{6, 380}$) & \textbf{\underline{95.03}}$_{(1.00\times)}$ & \textbf{76.37}$_{(1.02\times)}$  \\
    \bottomrule
    \end{tabular}}
    \end{small}
\end{minipage}
\end{table}
\vspace{-7mm}
\subsection{Using DNArch under computational constraints}
Next, we utilize DNArch to learn entire convolutional architectures that respect a predefined computational budget. To this end, we start with base  CCNN$_{4,140}$ and CCNN$_{6,380}$ networks, and allow DNArch to learn their width, depth, kernel sizes and downsampling. We define the target complexity $\mathcal{L}_{\mathrm{comp}}$ as the complexity of the base CCNN networks. In other words, we use DNArch to find better convolutional architectures of computational complexity roughly equal to that of the base networks.% Importantly, as the architecture and the weights are simultaneously learned, no fine-tuning is needed.% Note that we only use Fourier convolutions and thus, the learned kernel sizes learned do not impact computational efficiency.

\textbf{Results.} Our results (DNArch$_\mathrm{K, R, W, D}$ models in Tabs.~\ref{tab:5_lra_results}-\ref{tab:5_image_classif}) show that DNArch finds neural architectures that achieve higher accuracy than the base CCNN networks while keeping the same computational complexity. In addition, we observe that learning more neural architecture components consistently leads to better results, therefore supporting the claim that using gradient-steered architectures can be more beneficial than using handcrafted ones. Furthermore, we observe that using base architectures with larger complexity and capacity consistently leads to better results. This result is encouraging for the application of DNArch to large architectures, e.g., LLMs \cite{brown2020language, chowdhery2022palm}.

\begin{figure}
    \centering
    \includegraphics[width=0.6\textwidth]{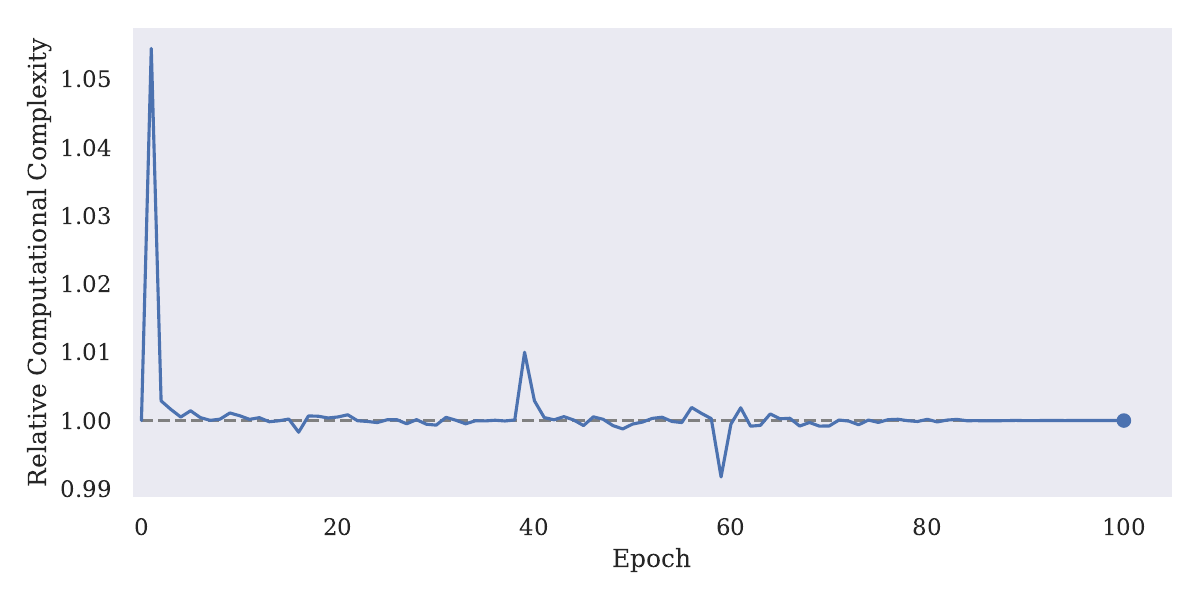}
    \vspace{-5mm}
    \caption{Relative complexity during the course of training on the {\tt Text} task. This behavior is consistent across all tasks.
    \vspace{-2mm}}
    \label{fig:5_rel_complexity}
\end{figure}

\textbf{Computational complexity of DNArch.} To assess the applicability of DNArch, it is important to analyze its computational overhead. To this end, we analyze the behavior of the relative complexity ($\gC_\mathrm{curr}/\gC_\mathrm{target}$) during training (Fig.~\ref{fig:5_rel_complexity}). Interestingly, we observe that the theoretical complexity of candidate architectures $\gC_\mathrm{curr}$ stays close to the target complexity $\gC_\mathrm{target}$ \textit{during the whole training}. This indicates that: (\textit{i}) DNArch only searches among architectures that share the target computational complexity, and that (\textit{ii}) the computational overhead of DNArch is \textit{negligible}. As a result, the cost of using DNArch on top of a CCNN is comparable to the cost of training the base CCNN network. Note that, for the experiments in Sec.~\ref{sec:5_no_constr}, the cost of training can be even lower than that of the base network since the complexity of found architectures are up to $25\%$ faster.
\vspace{-7mm}
\subsection{Architectures found by DNArch} 
The architectures found by DNArch are listed in Tabs.~\ref{tab:5_found architectures_lra}-\ref{tab:5_found architectures_2d}. Interestingly, we observe that found architectures are very diverse, even within each architecture. For instance, some residual blocks have a bottleneck structure, some an expanded structure, and others have monotonically decreasing or increasing widths. Interestingly, the resolution of found architectures for classification tasks, e.g., {\tt Text}, often follow the style of U-Nets, and not the monotonically decreasing pattern commonly seen in handcrafted networks. On dense tasks, we observe architectures that resemble U-Nets, and even concatenated U-Nets, e.g., the 1.5$\times$ U-Net like architecture found for the {\tt Cosmic} task. 

The kernel sizes found by DNArch are also very diverse. In 1D tasks, found kernels are often large, which would make them parameter intensive with traditional parameterizations. In 2D tasks, we often see rectangular kernels that do not follow a monotonic pattern of increasing or decreasing sizes. Instead, found architectures often perform interleaved low-level and high-level feature extraction.% Interestingly, this behaviour is also observed in the human visual system \cite{kietzmann2019recurrence}
\vspace{-7mm}
\section{Limitations}\label{sec:5_limitations}
\textbf{Training on {\tt TPU} requires static shapes.} We train our models on {\tt TPUs}, a type of accelerator that requires a static computational graph derived for specific input and network shapes via the {\tt XLA} (Accelerated Linear Algebra) compiler. As a result, {\tt TPUs} do not support operations that change the shapes of arrays during training. This means that on {\tt TPUs}, DNArch can only perform masking modifications to the network during training, i.e., setting certain channels to zero but still computing their outputs. At inference, however, the masks are fixed. Consequently, we can effectively trim unused values to remove useless computations in a way that is compatible with {\tt XLA}. It is important to note that this limitation is solely an implementation issue caused by nature of {\tt TPUs}' hardware and can be avoided by using libraries and hardware that support dynamic computational graphs, e.g., {\tt PyTorch} and {\tt GPUs}. While our results were obtained using {\tt TPUs}, we also provide a {\tt PyTorch} implementation that avoids this issue, making it more flexible and accessible, especially in scenarios where one needs to keep candidate networks close to the target complexity $\gC_\mathrm{target}$ during training.

\textbf{DNArch requires instantiating the largest possible architecture.} While masking weights through a gradient update is straightforward, increasing the number of active weights requires those weights to be instantiated in memory. This means that even with dynamic computational graphs, it is necessary to instantiate the largest possible architecture learnable by DNArch in memory. To overcome this limitation, we set the maximum kernel size to the size of the input, and limit the maximum network size along the depth and width dimensions to double the number of blocks and channels of the base network. While this trick allows DNArch to easily shrink and grow representations within that range, this restricts the potential sizes of optimal architectures and can restrict the applicability of DNArch to very large models, e.g., LLMs \cite{brown2020language, chowdhery2022palm}.
\vspace{-7mm}
\section{Outlook and future work}
\textbf{Input-dependent neural architectures.}
In this work, the mask parameters are constant for all inputs within a task. An alternative approach could use an additional neural network \mlp$_\mathrm{mask}$ to predict the mask parameters based on context, e.g., the current input, current task, etc. This would enable the creation of context-dependent neural architectures such as early-exit systems \cite{teerapittayanon2016branchynet, ghodrati2021frameexit, schuster2022confident}, but where the whole network architecture is context-dependent. Consequently, resulting architectures would providing finer control of per-sample / per-modality complexity than existing methods. %Given the success of early-exit systems,  we consider this a promising direction for future research.

\textbf{Dynamic weighting of $\gL_\mathrm{comp}$ during training.}
DNArch explores architectures with complexity similar to target complexity throughout training. This results from using a constant $\lambda$ in Eq.~\ref{eq:5_optim_loss}. Alternatively, one could use a dynamic value of $\lambda$ during training to induce a different training behavior. For example, gradually increasing $\lambda$ would allow DNArch to explore architectures with larger complexity at first, and progressively encourage it to converge to networks with the desired target complexity. Such a weighting scheduling of $\lambda$ could lead DNArch to find better architectures.% This methodology could lead to yet better results by performing architecture search and distillation simultaneously. %Finding forms of $\gL$ to induce interesting training behaviours is an interesting area for future research.

\textbf{Training DNArch with additional / multiple constraints.} Here, we only consider computational complexity as a constraint when training with DNArch. However, other properties such as memory efficiency, hardware-awareness and robustness are equally important. Designing regularization terms that encourage other properties in DNArch as well as exploring how different properties can be optimized in unison are important directions for further research.% This research direction is of particular practical relevance.

%\textbf{Are handcrafted architectures optimal?} Our results raise the question of whether current handcrafted architectures are truly optimal or rather a result of discrete parameterizations and predefined downsampling / upsampling steps in the architectures. Further analysis of the patterns exhibited by DNArch could provide deeper insights into what makes a neural architecture effective and how to construct them.

%\textbf{From architecture design to loss function design?} Deep Learning has shifted the paradigm of machine learning from feature design to architecture design. Could DNArch take this a step further and shift the focus from architecture design to the design of the optimization loss function? That is, could DNArch leads us to a landscape where researchers focus on the design of optimization loss terms like $\gL_\mathrm{comp}$ to generate neural architectures via DNArch that meet certain desired properties, e.g., memory and computational requirements, symmetry equivariance, robustness, etc.? This is perhaps the most promising outcome of our work and we look forward to exploring this research direction further.

%% file: chapters/5_1_coattentive.tex
% path to figures directory
\graphicspath{{figures/chapter-5/}}

%=========================================================================

% \begin{savequote}[75mm]
% Nulla facilisi. In vel sem. Morbi id urna in diam dignissim feugiat. Proin molestie tortor eu velit. Aliquam erat volutpat. Nullam    ultrices, diam tempus vulputate egestas, eros pede varius leo.
% \qauthor{Quoteauthor Lastname}
% \end{savequote}

\chapter{Focusing Equivariance on Transformations Co-Occurring in Data}\label{chapter:coattentive}

\begin{flushright}
\textit{Based on the paper:}\break
\textit{Co-Attentive Equivariant Neural Networks: Focusing Equivariance On Transformations Co-Occurring In Data
\citep{Romero2020Co-Attentive}}
\end{flushright}
%=========================================================================
\vspace{-7mm}
\section{Introduction}
\label{sec:10_intro}
Thorough experimentation in the fields of psychology and neuroscience has provided support to the intuition that our visual perception and cognition systems are able to identify familiar objects despite modifications in size, location, background, viewpoint and lighting \citep{bruce1994recognizing}. 
Interestingly, we are not just able to recognize such modified objects, but are able to characterize which modifications have been applied to them as well. As an example, when we see a picture of a cat, we are not just able to tell that there is a cat in it, but also its position, its size, facts about the lighting conditions of the picture, and so forth.
Such observations suggest that the human visual system is \textit{equivariant} to a large \textit{transformation group} containing translation, rotation, scaling, among others. In other words, the mental representation obtained by seeing a transformed version of an object, is equal to that of seeing the original object and
transforming it mentally next.

These fascinating abilities exhibited by biological visual systems have inspired a large field of research towards the development of neural architectures able to replicate them. Among these, the most popular and successful approach is the Convolutional Neural Network (CNN) \citep{lecun1989backpropagation}, which incorporates equivariance to translation via convolution.
Unfortunately, in counterpart to the human visual system, CNNs do not exhibit equivariance to other transformations encountered in visual data (e.g., rotations).
Interestingly, however, if an ordinary CNN happens to learn rotated copies of the same filter, the stack of feature maps becomes equivariant to rotations even though individual feature maps are not \citep{cohen2016group}. Since ordinary CNNs must learn such rotated copies independently, they effectively utilize an important number of network parameters suboptimally to this end (see Fig.~3 in \citet{krizhevsky2012imagenet}). 
Based on the idea that equivariance in CNNs can be extended to larger transformation groups by stacking convolutional feature maps, several approaches have emerged to extend equivariance to, e.g., planar rotations \citep{dieleman2016exploiting,marcos2017rotation,weiler2018learning,DREN}, spherical rotations \citep{cohen2018spherical,worrall2018cubenet, cohen2019gauge}, scaling \citep{marcos2018scale, worrall2019deep} and general transformation groups \citep{cohen2016group}, such that transformed copies of a single entity are not required to be learned independently. 

Although incorporating equivariance to arbitrary transformation groups is conceptually and theoretically similar\footnote{It is achieved by developing feature mappings that use the transformation group in the feature mapping itself, e.g., translating a filter during a feature transformation is used to obtain translation equivariance.}, evidence from real-world experiences motivating their integration might strongly differ.
Several studies in neuroscience and psychology have shown that our visual system does not react equally to all transformations we encounter in visual data. 
Take, for instance, translation and rotation. Although we easily recognize objects independently of their position of appearance, a large corpus of experimental research has shown that this is not always the case for in-plane rotations. \citet{yin1969} showed that \textit{mono-oriented objects}, i.e., complex objects such as faces which are customarily seen in one orientation, are much more difficult to be accurately recognized when presented upside-down. This behaviour has been reproduced, among others, for magazine covers \citep{dallett1968picture}, symbols \citep{henle1942} and even familiar faces (e.g., from classmates) \citep{brooks1963recognition}. Intriguingly, \citet{schwarzer2000development} found that this effect exacerbates with age (adults suffer from this effect much more than children), but, adults are much faster and accurate in detecting mono-oriented objects in usual orientations.
%Note, however, that although the recognition of mono-oriented objects is more difficult when presented upside-down, we are still able to recognize them at the cost of a longer processing time. 

Based on these studies, we draw the following conclusions: 
\begin{itemize}[topsep=0pt]
    \item The human visual system does not perform (fully) equivariant feature transformations to visual data. Consequently, it does not react equally to all possible input transformations encountered in visual data, even if they belong to the same transformation group (e.g., in-plane rotations).
    \item The human visual system does not just encode familiarity to objects but seems to learn through experience the poses in which these objects customarily appear in the environment to assist and improve object recognition \citep{freire2000face, riesenhuber2004face, sinha2006face}.
\end{itemize}
\begin{figure}
    \centering
    \vspace{-2mm}
    \includegraphics[width=0.45\linewidth]{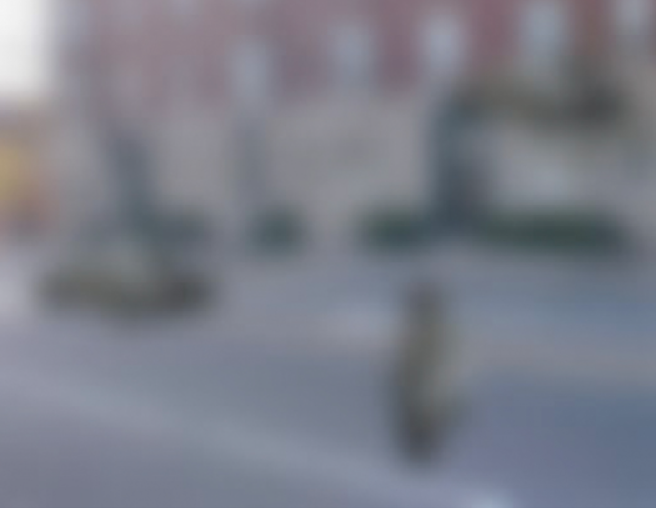}
    \caption{Our visual system infers object identities according to their size, location and orientation. In this blurred picture, observers describe the scene as containing a car and a pedestrian in the street. However, the pedestrian is in fact the \textit{same shape} as the car, except for a 90$^{\circ}$ rotation. The atypicality of this orientation \textit{within the context defined by the street scene} causes the car to be recognized as a pedestrian. Taken from \cite{oliva2007role}.
    \vspace{-2mm}} 
    \label{fig:10_context}
\end{figure}
Complementary studies \citep{tarr1989mental, oliva2007role} suggest that our visual system encodes orientation atypicality relative to the context rather than on an absolute manner (Fig. \ref{fig:10_context}). Motivated by the aforementioned observations we state \textit{the co-occurrence envelope hypothesis}:

\textbf{The Co-occurrence Envelope Hypothesis.} \textit{By allowing equivariant feature mappings to detect transformations that co-occur in the data and focus learning on the set formed by these co-occurrent transformations (i.e., the co-occurrence envelope of the data), one is able to induce learning of more representative feature representations of the data, and, resultantly, enhance the descriptive power of neural networks utilizing them. We refer to one such feature mapping as \textbf{co-attentive equivariant.}}

\begin{figure}
\begin{subfigure}{.12\textwidth}
  \centering
  % include first image
  \includegraphics[width=\linewidth]{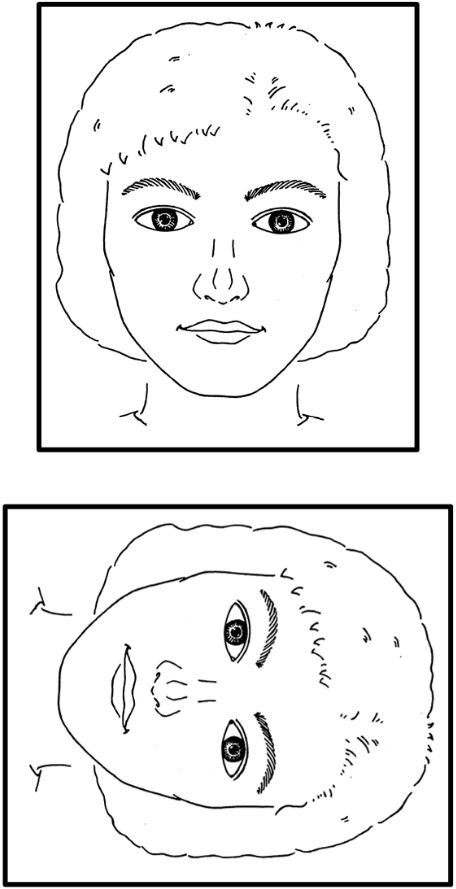}  
  \captionsetup{justification=centering}
  \caption{}
  \label{fig:10_input}
\end{subfigure}
\begin{subfigure}{.28\textwidth}
  \centering
  % include second image
  \includegraphics[width=0.9\linewidth]{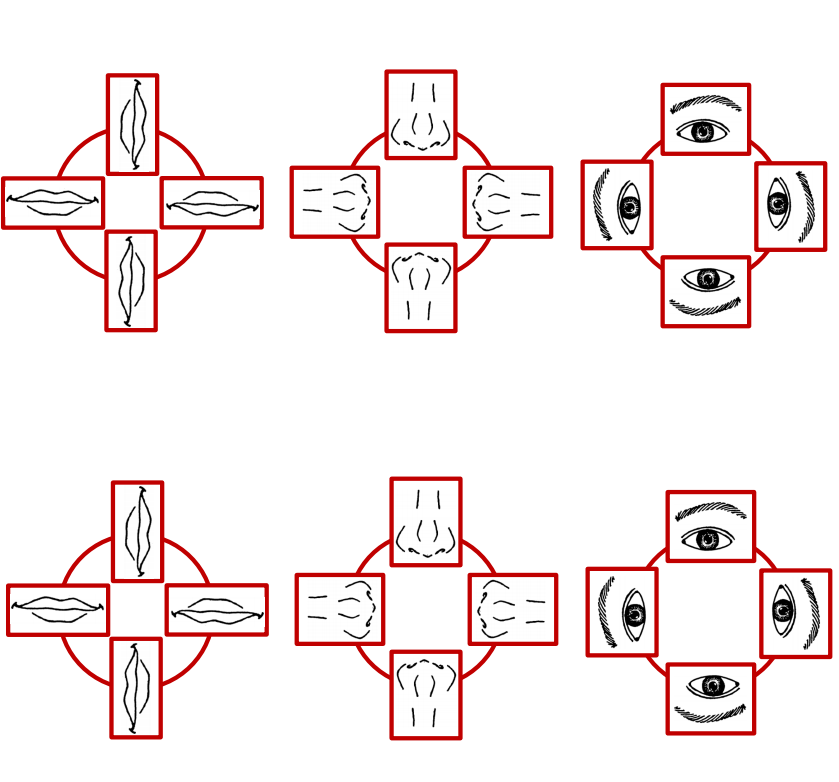}  
  \captionsetup{justification=centering}
  \caption{}
  \label{fig:10_full_equivariant}
\end{subfigure}
\begin{subfigure}{.28\textwidth}
  \centering
  % include third image
  \includegraphics[width=0.9\linewidth]{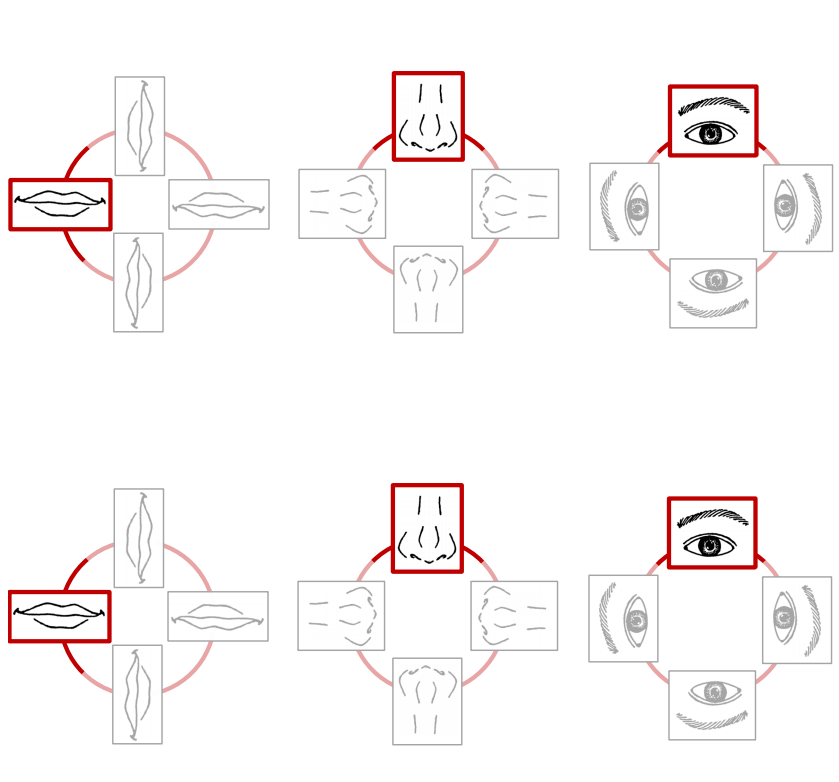}  
  \captionsetup{justification=centering}
  \caption{}
  \label{fig:10_partially_equivariant}
\end{subfigure}
\begin{subfigure}{.28\textwidth}
  \centering
  % include fourth image
  \includegraphics[width=0.9\linewidth]{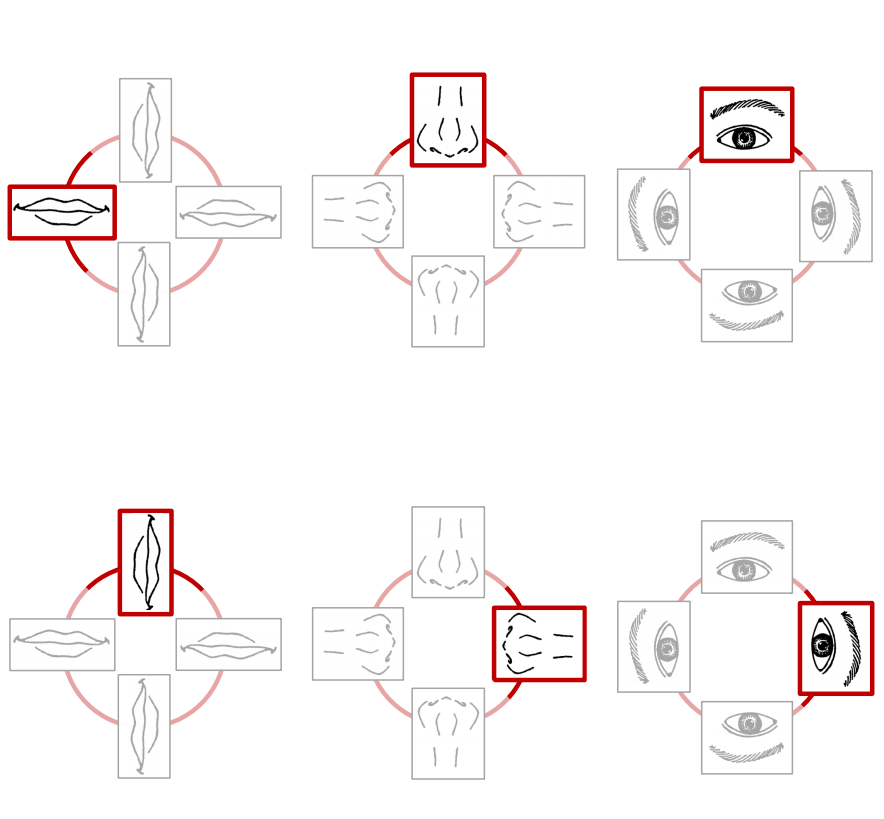}  
  \captionsetup{justification=centering}
  \caption{}
  \label{fig:10_adaptively_equivariant}
\end{subfigure}
\vspace{-2mm}
\caption{ Effect of multiple attention strategies for the prioritization of relevant pattern orientations in rotation equivariant networks for the task of face recognition. Given that all attention strategies are learned exclusively from upright faces, we show the set of relevant directions for the recognition of faces in two orientations (Fig. \ref{fig:10_input}) obtained by: no attention (Fig. \ref{fig:10_full_equivariant}), attending to the pattern orientations of appearance independently (Fig. \ref{fig:10_partially_equivariant}) and, attending to the pattern orientations of appearance relative to one another (Fig. \ref{fig:10_adaptively_equivariant}). Built upon Figure 1 from \citet{schwarzer2000development}.
}
\label{fig:10_fig}
\end{figure}

\textbf{Identifying the co-occurrence envelope.} Consider a rotation equivariant network receiving two copies of the same face (Fig. \ref{fig:10_input}). A conventional rotation equivariant network is required to perform inference and learning on the set of all possible orientations of the visual patterns constituting a face regardless of the input orientation (Fig. \ref{fig:10_full_equivariant}). However, by virtue of its rotation equivariance, it is able to recognize rotated faces even if it is trained on upright faces only. A possible strategy to simplify the task at hand could be to restrict the network to react exclusively to upright faces (Fig. \ref{fig:10_partially_equivariant}). In this case, the set of relevant visual pattern orientations becomes much smaller, at the expense of disrupting equivariance to the rotation group. Resultantly, the network would risk becoming unable to detect faces in any other orientation than those it is trained on. A better strategy results from restricting the set of relevant pattern orientations by defining them relative to one another (e.g., mouth orientation relative to the eyes) as opposed to absolutely (e.g., upright mouth) (Fig. \ref{fig:10_adaptively_equivariant}). In such a way, we are able to exploit information about orientation co-occurrences in the data without disrupting equivariance. The set of co-occurrent orientations in Fig. \ref{fig:10_adaptively_equivariant} corresponds to the co-occurrence envelope of the samples in Fig. \ref{fig:10_input} for the transformation group defined by rotations.

In this work, we introduce \textit{co-attentive equivariant feature mappings} and apply them on existing equivariant neural architectures. To this end, we leverage the concept of \textit{attention} \citep{bahdanau2014} to modify existing mathematical frameworks for equivariance, such that co-occurrent transformations can be detected. It is critical not to disrupt equivariance in the attention procedure as to preserve it across the entire network. To this end, we introduce \textit{cyclic equivariant self-attention}, a novel attention mechanism able to preserve equivariance to cyclic groups.

\textbf{Experiments and results.} We explore the effects of co-attentive equivariant feature mappings for single and multiple symmetry groups. Specifically, we replace conventional rotation equivariant mappings in $p4$-CNNs \citep{cohen2016group} and DRENs \citep{DREN} with co-attentive ones. We show that \textit{co-attentive rotation equivariant neural networks} consistently outperform their conventional counterparts in fully (rotated MNIST) and partially (CIFAR-10) rotational settings. Subsequently, we generalize cyclic equivariant self-attention to multiple similarity groups and apply it on $p4m$-CNNs \citep{cohen2016group} (equivariant to rotation and mirror reflections). Our results are in line with those obtained for single symmetry groups and support our stated hypothesis.

\textbf{Contributions.}
\begin{itemize}[topsep=0pt]
\item We propose the \textit{co-occurrence envelope hypothesis} and demonstrate that conventional equivariant mappings are consistently outperformed by our proposed \textit{co-attentive equivariant} ones.
\item We generalize co-attentive equivariant mappings to multiple symmetry groups and provide, to the best of our knowledge, the first attention mechanism acting generally on symmetry groups.
\end{itemize}
\vspace{-7mm}
\section{Preliminaries}\label{sec:10_preliminary}
%{\color{red}In this section we present key preliminary concepts that serve as foundations for our approach.\break We begin by introducing the notion of equivariance and motivate its importance for deep representation learning. Subsequently, we illustrate the construction and training procedures of conventional equivariant neural networks to conclude with some key implications that motivate our approach. }

\textbf{Equivariance.} We say that a feature mapping $f: \gX \rightarrow \gY$ is equivariant to a (transformation) group $\gG$ (or $\gG$-equivariant) if it commutes with actions of the group $\gG$ acting on its domain and codomain:
\begin{equation}
\label{eq:10_equi_def}
f\left(T^{\gX}_{g}(x)\right) = T^{\gY}_{g}\left(f(x)\right) \quad \forall g \in \gG, x\in \gX,
\end{equation}
where $T^{(\cdot)}_{g}$ denotes a \textit{group action} in the corresponding space. In other words, the ordering in which we apply a group action $T_{g}$ and the feature mapping $f$ is inconsequential.
There are multiple reasons as of why equivariant feature representations are advantageous for learning systems. %how to write this?
Since group actions $T^{X}_{g}$ produce predictable and interpretable transformations $T^{Y}_{g}$ in the feature space, the \textit{hypothesis space of the model} is reduced \citep{weiler2018learning} and the learning process simplified \citep{worrall2017harmonic}. Moreover, equivariance allows the construction of $L$-layered networks by stacking several equivariant feature mappings $\{f^{(1)}, ..., f^{(l)} ,..., f^{(L)}\}$ such that the input structure as regarded by the group $G$ is preserved (e.g., CNNs and input translations). As a result, any intermediate network representation $(f^{(l)}\circ ... \circ f^{(1)})(x)$, $l \in L$, is able to take advantage of the structure of $x$ as well.
\textit{Invariance} is an special case of equivariance in which $T^{Y}_{g} {=} \text{Id}_{Y}$, the identity, and thus all group actions in the input space are mapped to the same feature representation.

\textbf{Equivariant neural networks.} In neural networks, the integration of equivariance to arbitrary groups $G$ has been achieved by developing feature mappings $f$ that utilize the actions of the group $G$ in the feature mapping itself. %(e.g. convolution makes usage of $T_{(0,1)}$ and $T_{(1,0)}$ to achieve translation equivariance). 
Interestingly, \textit{equivariant feature mappings} encode equivariance as \textit{parameter sharing} with respect to $G$, i.e., the same weights are reused for all $g \in G$. This makes the inclusion of larger groups extremely appealing in the context of parameter efficient networks. 

Conventionally, the $l$-th layer of a neural network receives a signal $x^{(l)}(u,\lambda)$ (where $u \in \sZ^{2}$ is the spatial position and $\lambda \in \Lambda_{l}$ is the unstructured channel index, e.g., RGB channels in a color image), and applies a feature mapping $f^{(l)}: \sZ^{2}\times \Lambda_{l} \rightarrow \sZ^{2}\times \Lambda_{l+1}$ to generate the feature representation $x^{(l+1)}(u,\lambda)$. In CNNs, the feature mapping $f^{(l)}:=f_{T}^{(l)}$ is defined by a \textit{convolution}\footnote{Formally it is as a correlation. However, we hold on to the standard deep learning terminology.} ($\star_{\sR^{2}}$) between the input signal $x^{(l)}$ and a learnable convolutional filter $W^{(l)}_{\lambda',\lambda}$, $\lambda' \in \Lambda_{l}$, $\lambda \in \Lambda_{l+1}$:
\begin{equation}
\label{eq:10_conv}
x^{(l+1)}(u,\lambda) = [x^{(l)} \star_{\sR^2} W^{(l)}_{\lambda',\lambda}](u, \lambda) = \sum_{\lambda', u'}x^{(l)}(u + u', \lambda')W^{(l)}_{\lambda',\lambda}(u').
\end{equation}
By sliding $W^{(l)}_{\lambda',\lambda}$ across $u$, CNNs preserve the spatial structure of the input through the mapping $f^{l}_{T}$ and successfully provide equivariance to the translation group $T{=}(\sZ^{2},+)$.

The underlying idea for the extension of equivariance to larger groups in CNNs is conceptually equivalent to the strategy utilized by \cite{lecun1989backpropagation} for translation equivariance.  
Consider, for instance, the inclusion of equivariance to the set of rotations by $\theta_{r}$ degrees\footnote{The reader may easily verify that $\Theta$ (and hence $\sZ^{2} \rtimes\Theta$, with ($\rtimes$) the semi-direct product) forms a group.}: $\Theta {=} \{\theta_{r} {=} r \frac{2\pi}{r_{\text{max}}}\}_{r=1}^{r_{\text{max}}}$. 
To this end, we modify the feature mapping $f^{(l)}{\coloneqq}f^{(l)}_{R}: \sZ^{2} \times \Theta \times \Lambda_{l}\rightarrow \sZ^{2}  \times \Theta \times \Lambda_{l+1}$ to include the rotations defined by $\Theta$. Let $x^{(l)}(u,r,\lambda)$ and $W^{(l)}_{\lambda',\lambda}(u,r)$ be the input and the convolutional filter with an affixed index $r$ for rotation.  
The \textit{roto-translational convolution} ($\star_{\sR^{2} \rtimes \Theta}$) $f^{(l)}_{R}$ is defined as:% \citep{cheng2018rotdcf}: %\footnote{Note that $f_{R}$ makes additional usage of $\theta_{r}$, the generator of the group formed by $\Theta$ in the feature mapping.}:
\begin{equation}
\label{eq:10_rot_conv_1}
x^{(l+1)}(u, r, \lambda) = [x^{(l)}\star_{\sR^{2}\rtimes \Theta} W^{(l)}_{\lambda',\lambda}](u, r, \lambda) =\sum_{\lambda', r', u'}x^{(l)}(u + u', r', \lambda')W^{(l)}_{\lambda',\lambda}(\theta_{r}u', r'-r).
\end{equation}
Since $f^{(l)}_{R}$ produces ($\text{dim}(\Theta){=}r_{\text{max}})$ times more output feature maps than $f^{(l)}_{T}$, we need to learn fewer convolutional filters $W^{(l)}_{\lambda',\lambda}$ to produce the same number of output channels. 

\textbf{Learning equivariant neural networks.} Consider the change of variables $g=u$, $G=\sZ^2$, $g \in G$ and $g=(u,r)$, $G=\sZ^2\rtimes\Theta$,  $g \in G$ in Eq. \ref{eq:10_conv} and Eq. \ref{eq:10_rot_conv_1}, respectively. In general, neural networks are learned via backpropagation \citep{lecun1989backpropagation} by iteratively applying the chain rule of derivation to update the network parameters. Intuitively, the networks outlined in Eq. \ref{eq:10_conv} and Eq. \ref{eq:10_rot_conv_1} obtain feedback from all $g \in G$ and, resultantly, are inclined to learn feature representations that perform optimally on the entire group $G$. 
However, as outlined in Fig. \ref{fig:10_fig} and Section \ref{sec:10_intro}, several of those feature combinations are not likely to simultaneously appear. Resultantly, the\textit{ hypothesis space of the model} as defined by \citet{weiler2018learning} might be further reduced. 

Note that this reasoning is tightly related to existing explanations for the large success of \textit{spatial} \citep{xu2015show, woo2018cbam, zhang2018self} and \textit{temporal} \citep{luong2015effective, vaswani2017attention, mishra2017simple, zhang2018self} attention in neural architectures.    

\vspace{-7mm}
\section{Co-Attentive Equivariant Neural Networks} \label{sec:10_co_att_nns}
\begin{figure}
    \centering
    \includegraphics[width=0.72\linewidth]{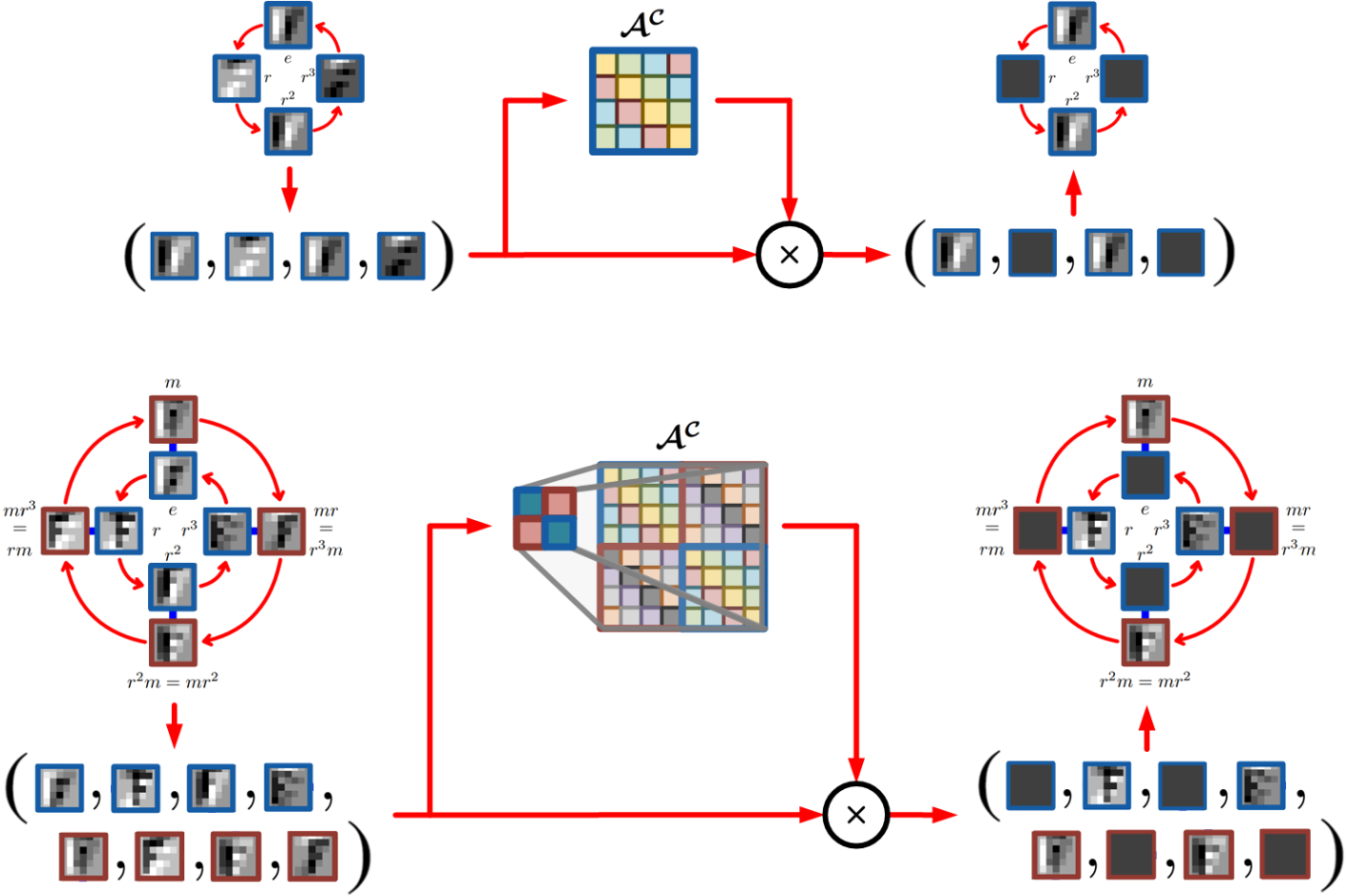}
    \caption{Co-attentive equivariant feature mappings acting on the groups $p4$ (top) and $p4m$ (bottom). In order to learn co-attentive equivariant representations, \textit{cyclic equivariant self-attention} $\mathcal{A^{C}}$ is applied on top of the output of a conventional equivariant feature mapping (here $p4$ and $p4m$ group convolutions, respectively). Resultantly, the group convolution responses are modulated based on their assessed relevance. For multiple symmetry groups, the group convolution responses must be rearranged in a vector structure so that the permutation laws of $\mathcal{A^{C}}$ correspond to those of the composing group symmetries. Same colors in $\mathcal{A^{C}}$ denote equal weights. The \textit{circulant (block) structure} of $\mathcal{A^{C}}$ ensures that group equivariance is preserved through the course of attention.  Therefore, if the input is rotated --or mirrored in $p4m$--, the attention mask will transform accordingly. Built upon Figures~1 and~2 of \citet{cohen2016group}.}
    \label{fig:10_attention_mech}
\end{figure}

 In this section we define co-attentive feature mappings and apply them in the context of equivariant neural networks (Figure \ref{fig:10_attention_mech}). To this end, we introduce cyclic equivariant self-attention and utilize it to construct co-attentive rotation equivariant neural networks. Subsequently, we show that cyclic equivariant self-attention is extendable to larger symmetry groups and make use of this fact to construct co-attentive neural networks equivariant to rotations and mirror reflections.

 \vspace{-7mm}
\subsection{Co-Attentive Rotation Equivariant Neural Networks} \label{subsec:10_co_attentive_rot}
To allow rotation equivariant networks to utilize and learn \textit{co-attentive equivariant representations}, we introduce an attention operator $\mathcal{A}^{(l)}$ on top of the roto-translational convolution $f^{(l)}_{R}$ with which discernment along the rotation axis $r$ of the generated feature responses $x^{(l)}(u,r,\lambda)$ is possible. Formally, our \textit{co-attentive rotation equivariant feature mapping} $f_{\mathcal{R}}^{(l)}$ is defined as follows:
\begin{equation}
\label{eq:10_att_conv}
x^{(l+1)} = f_{\mathcal{R}}^{(l)}(x^{(l)}) = \mathcal{A}^{(l)}(f_{R}^{(l)}(x^{(l)}))  =\mathcal{A}^{(l)}\big([x^{(l)}\star_{\sR^{2}\rtimes \Theta} W^{(l)}_{\lambda',\lambda}]).
\end{equation}
Theoretically, $\mathcal{A}^{(l)}$ could be defined globally over $f^{(l)}_{R}(x^{(l)})$ (i.e., simultaneously along $u$, $r$, $\lambda$) as depicted in Eq. \ref{eq:10_att_conv}. However, we apply attention locally to: (1) grant the algorithm enough flexibility to attend locally to the co-occurrence envelope of feature representations and, (2) utilize attention exclusively along the rotation axis $r$, such that our contributions are clearly separated from those possibly emerging from spatial attention. To this end, we apply attention pixel-wise on top of $f^{(l)}_{R}(x^{(l)})$ (Eq. \ref{eq:10_att_conv_local}). Furthermore, we assign a single attention instance  $\mathcal{A}^{(l)}_{\lambda}$ to each learned feature representation and utilize it across the spatial dimension of the output feature maps\footnote{For a more meticulous discussion on how Eq. \ref{eq:10_att_conv_local} attains co-occurrent attention, see Appendix \ref{sec:10_appendix}.}: 
\begin{equation}
\label{eq:10_att_conv_local}
x^{(l+1)}(u,r,\lambda) = \mathcal{A}^{(l)}_{\lambda}(\{x^{(l+1)}(u,\hat{r},\lambda)\}_{\hat{r}=1}^{r_{\text{max}}})(r).
\end{equation}
\textbf{Attention and self-attention.} Consider a source vector $x=(x_{1}, ...,x_{n})$ and a target vector $y=(y_{1}, ..
. , y_{m})$.
In general, an attention operator $\mathcal{A}$ leverages information from the source vector $x$ (or multiple feature mappings thereof)
to estimate an attention matrix $A \in [0,1]^{n \times m}$, such that: (1) the element $A_{i,j}$ provides an importance assessment of the source element $x_{i}$ with reference to the target element $y_{j}$ and (2) the sum of importance over all $x_{i}$ is equal to one: $\sum_{i}{\emA_{i,j}} = 1$. Subsequently, the matrix $A$ is utilized to modulate the original source vector $x$ as to \textit{attend} to a subset of relevant source positions with regard to $y_{j}$: $\tilde{x}^{j}=(A_{:,j})^{T}\odot x$ (where $\odot$ is the Hadamard product). A special case of attention is that of \textit{self-attention} \citep{cheng2016long}, in which the target and the source vectors are equal ($y:=x)$. In other words, the attention mechanism estimates the influence of the sequence $x$ on the element $x_{j}$ for its weighting.% {\color{red}Check row-col order}

In general, the attention matrix\footnote{Technically, each column of $A$ is restricted to a simplex and hence $A$ lives in a subspace of $[0,1]^{n \times m}$.} $A\in [0,1]^{n \times m}$ is constructed via nonlinear space transformations $f_{\tilde{A}}:\sR^{n}\rightarrow\sR^{n\times m}$ of the source vector $x$, on top of which the softmax function is applied: $A_{:,j} = \mathrm{softmax}(f_{\tilde{A}}(x)_{:,j})$. This ensures that the properties previously mentioned hold.
Typically, the mappings $f_{\tilde{A}}$ found in literature take feature transformation pairs of $x$ as input (e.g., $\{s,H\}$ in RNNs \citep{luong2015effective}, $\{Q,K\}$ in self-attention networks \citep{vaswani2017attention}), and perform (non)-linear mappings on top of it, ranging from multiple feed-forward layers \citep{bahdanau2014} to several operations between the transformed pairs \citep{luong2015effective, vaswani2017attention, mishra2017simple, zhang2018self}. Due to the computational complexity of these approaches and the fact that we do extensive pixel-wise usage of $f_{\tilde{A}}$ on every network layer, their direct integration in our framework is computationally prohibitive. To circumvent this problem, we modify the usual self-attention formulation as to enhance its descriptive power in a much more compact setting.

\textbf{Compact local self-attention.} Initially, we relax the range of values of $A$ from $[0,1]^{n\times n}$ to $\sR^{n\times n}$. This allows us to encode much richer relationships between element pairs $(x_{i}, x_{j})$ at the cost of less interpretability. Subsequently, we define $A = x^{T}\odot\tilde{A}$, where  $\tilde{A} \in \sR^{n\times n}$ is a matrix of learnable parameters. Furthermore, instead of directly applying softmax on the columns of $A$, we first sum over the contributions of each element $x_{i}$ to obtain a vector $a = \{\sum_{i}{A_{i,j}}\}_{j=1}^{n}$, which is then passed to the softmax function. Following \cite{vaswani2017attention}, we prevent the softmax function from reaching regions of low gradient by scaling its argument by $(\sqrt{\text{dim}(A)})^{-1} = (1 \mathbin{/} n)$: $\tilde{a}=\mathrm{softmax}((1\mathbin{/}n)\ a)$. Lastly, we counteract the contractive behaviour of the softmax function by normalizing $\tilde{a}$ before weighting $x$ as to preserve the magnitude range of its argument. This allows us to use $\mathcal{A}$ in deep architectures. Our \textit{compact self-attention mechanism} is summarized as follows: 
\begin{gather}
\label{eq:10_gral_att_1}
    a = \{{\textstyle\sum}_{i}{A_{i,j}}\}_{j=1}^{n} = {\textstyle\sum}_{i}(x^{T}\odot\tilde{A})_{i,j}=x\tilde{A}\\
    \tilde{a} = \mathrm{softmax}((1\mathbin{/}n) \ a) \\
    \label{eq:10_gral_att_2}
    \hat{x}=\mathcal{A}(x)=(\tilde{a}\mathbin{/}\max(\tilde{a}))\odot x.
\end{gather}
% --------------------------- old version -----------------------------
%\textbf{Compact local self-attention.} Initially, we relax the range of values of $\mA$ from $[0,1]^{n\times n}$ to $\sR^{n\times n}$. This allows us to encode much richer relationships between the pair of elements $(x_{i}, x_{j})$ at the cost of less interpretability. Subsequently, we define $\mA = \vx^{T}\odot\tilde{\mA}$, where  $\tilde{\mA} \in \sR^{n\times n}$ is a matrix of learnable parameters. Instead of applying column-wise softmax directly to $\mA$, we first sum over the contributions of each element $x_{i}$ to obtain a vector $\va = (\sum_{i}{A_{i,1}}, ..., \sum_{i}{A_{i,n}})$, which is then passed to the softmax function. Following \cite{vaswani2017attention}, we prevent the softmax function from reaching regions of low gradient by scaling its argument by $(\sqrt{\text{dim}(\mA)})^{-1} = (1 \mathbin{/} n)$: $\tilde{\va}=\mathrm{softmax}((1\mathbin{/}n)\ \va)$. Lastly, we counteract the softmax contractive behaviour by normalizing $\tilde{\va}$ before weighting $\vx$ to conserve the magnitude range of the input values. This allows us to use $\mathcal{A}$ in deep architectures. Formally, our compact modified self-attention mechanism is defined as: 
%\begin{gather}
%\label{eq:10_gral_att_1}
%    \tilde{\va} = \mathrm{softmax}((1\mathbin{/}n) \ \vx \tilde{\mA})\\
%    \label{eq:10_gral_att_2}
%    \hat{\vx}=\mathcal{A}(\vx)=(\tilde{\va}\mathbin{/}\max(\tilde{\va}))\odot \vx
%\end{gather}
\textbf{The cyclic equivariant self-attention operator $\boldsymbol{\mathcal{A^{C}}}$.}
Consider $\{x(u,r,\lambda)\}_{r=1}^{r_{\text{max}}}$,  the vector of responses generated by a roto-translational convolution $f_{R}$ stacked along the rotation axis $r$. By applying self-attention along $r$, we are able to generate an importance matrix $A \in \sR^{r_{\text{max}}\times r_{\text{max}}}$ relating all pairs of $(\theta_{i}, \theta_{j})$-rotated responses in the rotational group $\Theta$ at a certain position. We refer to this attention mechanism as \textit{full self-attention} ($\mathcal{A^{F}}$). Although $\mathcal{A^{F}}$ is able to encode arbitrary linear source-target relationships for each target position, it is not restricted to conserve equivariance to $\Theta$. Resultantly, we risk incurring into the behavior outlined in Fig. \ref{fig:10_partially_equivariant}. 
Before we further elaborate on this issue, we introduce the \textit{cyclic permutation operator} $\mathcal{P}^{i}$, which induces a cyclic shift of $i$ positions on its argument: $\sigma^{\mathcal{P}^{i}}(x_{j})=x_{(j+ i)\text{mod}(\text{dim}(x))}$, $ \forall x_{j} \in x$. 

Consider a full self-attention operator $\mathcal{A^{F}}$ acting on top of a roto-translational convolution $f_{R}$. Let $p$ be an input pattern to which $f_{R}$ only produces a strong activation in the feature map $x(\hat{r})=f_{R}(p)(\hat{r})$, $\hat{r}\in\{r\}_{r=1}^{r_{\text{max}}}$. Intuitively, during learning, only the corresponding attention coefficients $\tilde{A}_{;, \hat{r}}$ in $\mathcal{A^{F}}$ would be significantly increased. Now, consider the presence of the input pattern $\theta_{i}p$, a $\theta_{i}$-rotated variant of $p$. By virtue of the rotational equivariance property of the feature mapping $f_{R}$, we obtain (locally) an exactly equal response to that of $p$ up to a cyclic permutation of $i$ positions on $r$, and thus, we obtain a strong activation in the feature map $\mathcal{P}^{i}(x(\hat{r}))= x(\sigma^{\mathcal{P}^{i}}(\hat{r}))$. We encounter two problems in this setting: $\mathcal{A^{F}}$ is not be able to detect that $p$ and $\theta_{i}p$ correspond to the exact same input pattern and, as each but the attention coefficients $\tilde{A}_{:,j}$ is small, the network might considerably damp the response generated by $\theta_{i}p$. As a result, the network might (1) squander important feedback information during learning and (2) induce learning of repeated versions of the same pattern for different orientations. In other words, $\mathcal{A^{F}}$ does not behave equivariantly as a function of $\theta_{i}$.

Interestingly, we can introduce prior-knowledge into the attention model by restricting the structure of $\tilde{A}$. By leveraging \textit{equivariance to the cyclic group} $\mathcal{C}_{n}$, we are able to solve the problems exhibited by $\mathcal{A^{F}}$ and simultaneously reduce the number of additional parameters required by the self-attention mechanism (from $r_{\text{max}}^{2}$ to $r_{\text{max}}$). 
Consider again the input patterns $p$ and $\theta_{i}p$. We incorporate the intuition that $p$ and $\theta_{i}p$ are one and the same entity, and thus, $f_{R}$ (locally) generates the same output feature map up to a cyclic permutation $\mathcal{P}^{i}$: $f_{R}(\theta_{i}p){=}\mathcal{P}^{i}(f_{R}(p))$. Consequently, the attention mechanism should produce the \textit{exact same} output for both $p$ and $\theta_{i}p$ up to the same cyclic permutation $\mathcal{P}^{i}$. In other words, $\mathcal{A}$ (and thus $\tilde{A}$) should be \textit{equivariant to cyclic permutations}.

A well-known fact in mathematics is that a matrix $A$ is equivariant with respect to cyclic permutations of the domain i\textit{f and only if} it is \textit{circulant} \citep{alkarni2001statistical, aahlander2005applications}. We make use of this fact and leverage \textit{circulant matrices} to impose cyclic equivariance to the structure of $\tilde{A}$. Formally, a circulant matrix $C \in \sR^{n\times n}$ is composed of $n$ cyclic permutations of its defining vector $c {=} \{c_{i}\}_{i=1}^{n}$, such that its $j$-th column is a cyclic permutation of $j-1$ positions of $c$: $C_{:,j} {=} \mathcal{P}^{j-1}(c)^{T}$. We construct our \textit{cyclic equivariant self-attention operator} $\mathcal{A^{C}}$ by defining $\tilde{A}$ as a circulant matrix specified by a learnable attention vector $a^{\mathcal{C}}{=}\{a^{\mathcal{C}}_{i}\}_{i=1}^{r_{\text{max}}}$:
\begin{equation}
\label{eq:10_circ_matrix}
\tilde{A}=\{\mathcal{P}^{j-1}(a^{\mathcal{C}})^{T}\}_{j=1}^{n},
\end{equation}
and subsequently applying Eqs.~\ref{eq:10_gral_att_1}-\ref{eq:10_gral_att_2}.
Resultantly, $\mathcal{A^{C}}$ is able to assign the responses generated by $f_{R}$ for rotated versions of an input pattern $p$ to a unique entity: $f_{R}(\theta_{i}p)=\mathcal{P}^{i}(f_{R}(p))$, and dynamically adjust its output to the angle of appearance $\theta_{i}$, such that the attention operation does not disrupt its propagation downstream the network: $\mathcal{A^{C}}(f_{R}(\theta_{i}p))=\mathcal{P}^{i}(\mathcal{A^{C}}(f_{R}(p)))$. Consequently, the attention weights $a^{\mathcal{C}}$ are updated equally regardless of specific values of $\theta_{i}$. 
Due to these properties, $\mathcal{A^{C}}$ does not incur in any of the problems outlined earlier in this section. 
Conclusively, our \textit{co-attentive rotation equivariant feature mapping} $f_{\mathcal{R}}^{(l)}$ is defined as follows:
\begin{equation}
\label{eq:10_att_conv_last}
x^{(l+1)}(u,r,\lambda) = f_{\mathcal{R}}^{(l)}(x^{(l)})(u,r,\lambda) =\mathcal{A}^{\mathcal{C}(l)}_{\lambda}\big([x^{(l)}\star_{\sR^{2}\rtimes \Theta} W^{(l)}_{\lambda',\lambda}])(u, r, \lambda).
\end{equation}
Note that a co-attentive equivariant map $f_{\mathcal{R}}$ is approximately equal --up to a normalized softmax operation (Eq.~\ref{eq:10_gral_att_2})-- to a conventional equivariant one $f_{R}$, if $\tilde{A}{=}\alpha I$ for any $\alpha \in \sR$.

\vspace{-7mm}
\subsection{Extending $\mathcal{A^{C}}$ To Multiple Symmetry Groups}

The self-attention mechanisms outlined in the previous section are easily extendable to larger groups consisting of multiple symmetries. Consider, for instance, the group $\theta_{r}m$ of rotations by $\theta_{r}$ degrees and mirror reflections $m$ defined analogously to the group $p4m$ in \citet{cohen2016group}. Let $p(u,r,m,\lambda)$ be an input signal with an affixed index $m \in \{m_0,m_1\}$ for mirror reflections ($m_1$ indicates mirrored) and $f_{\theta_{r}m}$ be a \textit{group convolution} \citep{cohen2016group} on the $\theta_{r}m$ group. The group convolution $f_{\theta_{r}m}$ produces two times as many output channels ($2r_{\text{max}}:m_0 r_{\text{max}} + m_1 r_{\text{max}}$) as those generated by the roto-translational convolution $f_{R}$ (Eq. \ref{eq:10_rot_conv_1}, Fig. \ref{fig:10_attention_mech}).

Full self-attention $\mathcal{A^{F}}$ can be integrated directly by modulating the output of $f_{\theta_{r}m}$ as depicted in Sec. \ref{subsec:10_co_attentive_rot} with $\tilde{A} \in \sR^{2r_{\text{max}}\times2r_{\text{max}}}$. Here, $\mathcal{A^{F}}$ relates each of the group convolution responses with one another. However, just as for $f_{R}$, $\mathcal{A^{F}}$ disrupts the equivariance property of $f_{\theta_{r}m}$ to the $\theta_{r}m$ group.

Similarly, the cyclic equivariant self-attention operator $\mathcal{A^{C}}$ can be extended to multiple symmetry groups as well. Before we continue, we introduce the \textit{cyclic permutation operator} $\mathcal{P}^{i,t}$, which induces a cyclic shift of $i$ positions on its argument along the transformation axis $t$. Consider the input patterns $p$ and $\theta_{i}p$ outlined in the previous section and $mp$, a mirrored instance of $p$. Let $x(u,r,m,\lambda) {=} f_{\theta_{r}m}(p)(u,r,m,\lambda)$ be the response of the group convolution $f_{\theta_{r}m}$ for the input pattern $p$. By virtue of the rotation equivariance property of $f_{\theta_{r}m}$, the generated response for $\theta_{i}p$ is equivalent to that of $p$ up to a cyclic permutation of $i$ positions along the rotation axis $r$: $f_{\theta_r m}(\theta_{i}p)(u,r,m,\lambda)=\mathcal{P}^{i, r}(f_{\theta_r m}(p))(u,r,m,\lambda)=x(u,\sigma^{\mathcal{P}^{i}}(r),m,\lambda)$.
Similarly, by virtue of the mirror equivariance property of $f_{\theta_{r}m}$, the response generated by $mp$ is equivalent to that of $p$ up to a cyclic permutation of one position along the mirroring axis $m$: $f_{\theta_{r}m}(mp)(u,r,m,\lambda)=\mathcal{P}^{1,m}(f_{\theta_r m}(p))(u,r,m,\lambda)=x(u,r,\sigma^{\mathcal{P}^{1}}(m),\lambda)$. Note that if we take two elements from a group $g$, $h$, their composition $(gh)$ is also an element of the group. Resultantly, $f_{\theta_{r}m}((m\theta^{i})p)(u,r,m,\lambda)=( \mathcal{P}^{1,m}\circ\mathcal{P}^{i,r})(f_{\theta_{r}m}(p))(u,r,m,\lambda)=\mathcal{P}^{1,m}(\mathcal{P}^{i,r}(x))(u,r,m,\lambda)=\mathcal{P}^{1,m}(x)(u,\sigma^{\mathcal{P}^{i}}(r),m,\lambda)=x(u,\sigma^{\mathcal{P}^{i}}(r),\sigma^{\mathcal{P}^{1}}(m),\lambda)$. 

In other words, in order to extend $\mathcal{A^{C}}$ to the $\theta_{r}m$ group, it is necessary to restrict the structure of $\tilde{A}$ such that it respects the \textit{permutation laws} imposed by the equivariant mapping $f_{\theta_{r}m}$. 
Let us rewrite $x(u,r,m,\lambda)$ as $x(u,g,\lambda)$, $g = (mr) \in \{m_0,m_1\}\times \{\hat{r}\}_{\hat{r}=1}^{r_{\text{max}}}$. In this case, we must impose a \textit{circulant block matrix} structure on $\tilde{A}$ such that: (1) the composing blocks permute internally as defined by $\mathcal{P}^{i, r}$ and (2) the blocks themselves permute with one another as defined by $\mathcal{P}^{1, m}$. Formally, $\tilde{A}$ is defined as:
\begin{equation}
\label{eq:10_a_multiple}
    \tilde{A} = \begin{bmatrix}\tilde{A_{1}} & \tilde{A_{2}} \\ \tilde{A_{2}} & \tilde{A_{1}}
    \end{bmatrix},
\end{equation}
where $\{\tilde{A_{i}} \in \sR^{r_{\text{max}}\times r_{\text{max}}}\}$, $i \in \{1,2\}$ are circulant matrices (Eq. \ref{eq:10_circ_matrix}). Importantly, the ordering of the permutation laws in $\tilde{A}$ is interchangeable if the input vector is modified accordingly, i.e., $g=(rm)$. 

In summary, cyclic equivariant self-attention $\mathcal{A^{C}}$ can be directly extended to act on any $G$-equivariant feature mapping $f_{G}$, and for any symmetry group $G$, if the group actions $T^{Y}_{g}$ produce cyclic permutations on the codomain of $f_{G}$. To this end, one must restrict the structure of $\tilde{A}$ to that of a circulant block matrix, such that \textit{all} permutation laws of $T^{Y}_{g}$ hold: $T^{Y}_{g}(\mathcal{A^{C}}(f_{G}))=\mathcal{A^{C}}(T^{Y}_{g}(f_{G}))$, $\forall g \in G$.

\vspace{-7mm}
\section{Experiments}
\textbf{Experimental Setup.} 
We validate our approach by exploring the effects of co-attentive equivariant feature mappings for single and multiple symmetry groups on existing equivariant architectures. Specifically, we replace conventional rotation equivariant mappings in $p4$-CNNs \citep{cohen2016group} and DRENs \citep{DREN} with co-attentive equivariant ones and evaluate their effects in fully (rotated MNIST) and partially (CIFAR-10) rotational settings. Similarly, we evaluate co-attentive equivariant maps acting on multiple similarity groups by replacing equivariant mappings in $p4m$-CNNs \citep{cohen2016group} (equivariant to rotation and mirror reflections) likewise.
Unless otherwise specified, we replicate as close as possible the same data processing, initialization strategies, hyperparameter values and evaluation strategies utilized by the baselines in our experiments. Note that the goal of this paper is to study and evaluate the relative effects obtained by co-attentive equivariant networks with regard to their conventional counterparts. Accordingly, we do not perform any additional tuning relative to the baselines. We believe that improvements to our reported results are feasible by further parameter tuning of the proposed co-attentive equivariant networks.

The additional learnable parameters, i.e., those associated to the cyclic self-attention operator ($\tilde{A}$) are initialized identically to the rest of the layer. Subsequently, we replace the values of $\tilde{A}$ along the diagonal by $1$ (i.e., $\text{diag}(\tilde{A}_{\text{init}})=1$) such that $\tilde{A}_{\text{init}}$ approximately resembles the identity $I$ and, hence, co-attentive equivariant layers are initially approximately equal to equivariant ones.

\textbf{Rotated MNIST.} The rotated MNIST dataset \citep{larochelle2007empirical} contains 62000 gray-scale 28x28 handwritten digits uniformly rotated on the entire circle $[0,2\pi)$. The dataset is split into training, validation and tests sets of 10000, 2000 and 50000 samples, respectively.
We replace rotation equivariant layers in $p4$-CNN \citep{cohen2016group}, DREN and DRENMaxPooling \citep{DREN} with co-attentive ones. Our results show that co-attentive equivariant networks consistently outperform conventional ones (see Table \ref{tab:10_results}).% without any additional parameter tuning (see Table \ref{tab:10_results}).

\textbf{CIFAR-10.} The CIFAR-10 dataset \citep{krizhevsky2009learning} consists of 60000 real-world 32x32 RGB images uniformly drawn from 10 classes. Contrarily to the rotated MNIST dataset, this dataset does not exhibit rotation symmetry. The dataset is split into training, validation and tests sets of 40000, 10000 and 10000 samples, respectively.
We replace equivariant layers in the $p4$ and $p4m$ variations of the All-CNN \citep{springenberg2014striving} and the ResNet44 \citep{he2016deep} proposed by \cite{cohen2016group} with co-attentive ones. Likewise, we modify the r\_x4-variations of the NIN \citep{lin2013network} and ResNet20 \citep{he2016deep} models proposed by \cite{DREN} in the same manner. Our results show that co-attentive equivariant networks consistently outperform conventional ones in this setting as well (see Table \ref{tab:10_results}).
\begin{table}
\caption{Comparison of conventional equivariant and co-attentive equivariant neural networks.
Values between parenthesis correspond to relevant results obtained from our own experiments.
}
\vspace{-3mm}
    \centering
    \scalebox{0.8}{
    \begin{tabular}{rccrcc}
    \toprule
    \multicolumn{3}{c}{\textbf{Rotated MNIST}} & \multicolumn{3}{c}{\textbf{CIFAR-10}}\\
    \midrule
        Model & Test Error ($\%$) & $\#$ Params. & Model & Test Error ($\%$) & $\#$ Params.\\
        \midrule
        Z2CNN & 5.03 $\pm$ 0.002 & 21.75k & All-CNN & 9.44 & 1.372M \\
        $p4$-CNN & 2.28 $\pm$ 0.0004 & 19.88k  & $p4$-All-CNN & 8.84 & 1.371M \\ 
        $a$-$p4$-CNN & \textbf{2.06 $\pm$ 0.0429} & 20.76k & $a$-$p4$-All-CNN & \textbf{7.68} & 1.373M \\
        DREN & 1.78 (1.99) & 19.88k & $p4m$-All-CNN & 7.59 & 1.219M \\ 
        $a$-DREN & \textbf{1.674} & 20.76k & $a$-$p4m$-All-CNN & \textbf{6.92}  & 1.223M \\
        DRENMaxPool. & 1.56 (1.60) & 24.68k & ResNet44 & 9.45 (9.85) & 2.639M\\
        $a$-DRENMaxPool. & \textbf{1.34}  & 25.68k & $p4m$-ResNet44 &  6.46 (9.47) & 2.623M \\ 
        & & &  $a$-$p4m$-ResNet44 & \textbf{9.12} & 2.632M \\
        & & & NIN & 10.41 (15.92) & 0.967M \\
        & & & r-NINx4& 14.96 & 0.958M \\ 
        & & & $a$-r-NINx4 & \textbf{13.67}  & 0.968M \\
        & & & ResNet20 & 9.00 (12.32)  & 0.335M  \\
        & & & r-ResNet20x4 & 12.31 & 0.333M \\
        & & & $a$-r-ResNet20x4 & \textbf{11.32} & 0.339M \\
        \bottomrule
    \end{tabular}
    }
    \label{tab:10_results}
\end{table}

\textbf{Training convergence of equivariant networks.} \cite{DREN} reported that adding too many rotational equivariant (isotonic) layers decreased the performance of their models on CIFAR-10. As a consequence, they did not report results on fully rotational equivariant networks for this setting and attributed this behaviour to the non-symmetricity of the data. We noticed that, with equal initialization strategies, rotational equivariant CNNs were much more prone to divergence than ordinary CNNs. This behaviour can be traced back to the additional feedback resulting from roto-translational convolutions (Eq. \ref{eq:10_rot_conv_1}) compared to ordinary ones (Eq. \ref{eq:10_conv}). 
After further analysis, we noticed that the data preprocessing strategy utilized by \citet{DREN} leaves some very large outlier values in the data ($|x|>$100), which strongly contribute to the behaviour outlined before. 

In order to evaluate the relative contribution of co-attentive equivariant neural networks we constructed fully equivariant DREN architectures based on their implementation. Although the obtained results were much worse than those originally reported in \citet{DREN}, we were able to stabilize training by clipping input values to the $99$ percentile of the data ($|x|\leq$2.3) and reducing the learning rate to 0.01, such that the same hyperparameters could be used across all network types. The obtained results (see Table \ref{tab:10_results}) signalize that DREN networks are \textit{comparatively better than CNNs both in fully and partially rotational settings}, contradictorily to the conclusions drawn in \citet{DREN}. 

This behaviour elucidates that although the inclusion of equivariance to larger transformation groups is beneficial both in terms of accuracy and parameter efficiency, one must be aware that such benefits are directly associated with an increase of the network susceptibility to divergence during training.% This is caused due to an increase of the information flow relative to the number of parameters in the network.

\vspace{-7mm}
\section{Discussion and Future Work}
Our results show that co-attentive equivariant feature mappings can be utilized to enhance conventional equivariant ones. Interestingly, co-attentive equivariant mappings are beneficial both in partially and fully rotational settings. We attribute this to the fact that a set of co-occurring orientations between patterns can be easily defined (and exploited) in both settings. 
It is important to note that we utilized attention independently over each spatial position $u$ on the codomain of the corresponding group convolution. Resultantly, we were restricted to mappings of the form $xA$, which, in turn, constraint our attention mechanism to have a circulant structure in order to preserve equivariance (since group actions acting in the codomain of the group convolution involve cyclic permutations and cyclic self-attention is applied in the codomain of the group convolution). 

In future work, we want to extend the idea presented here to act on the entire group simultaneously (i.e., along $u$ as well). By doing so, we lift our current restriction to mappings of the form $xA$ and therefore, may be able to develop attention instances with enhanced descriptive power. Following the same line of though, we want to explore incorporating attention in the convolution operation itself. Resultantly, one is not restricted to act exclusively on the codomain of the convolution, but instead, is able to impose structure in the domain of the mapping as well. Naturally, such an approach could lead to enhanced descriptiveness of the incorporated attention mechanism.
%In the scope of this work, . In future work, we want to expand the idea presented here to act on the entire group simultaneously. By doing so, we are not restricted to mappings of the form $Ax$, which constraint our attention mechanism to have a cyclic structure, and thus, we could possibly encode richer relationships. Analyzed from a different perspective, attention itself could be encoded in the convolution operation itself. Resultantly, one is not restricted to act on the codomain of the convolution, but impose structure in the mapping and the domain of the mapping itself. Such an approach would result in much stronger descriptive power of the function.
Moreover, we want to utilize and extend more complex attention strategies (e.g., \citet{bahdanau2014, luong2015effective, vaswani2017attention, mishra2017simple}) such that they can be applied to large transformation groups without disrupting equivariance. As outlined earlier in Section \ref{subsec:10_co_attentive_rot}, this becomes very challenging from a computational perspective as well, as it requires extensive usage of the corresponding attention mechanism. Resultantly, an efficient implementation thereof is mandatory. Furthermore, we want to extend co-attentive equivariant feature mappings to continuous (e.g., \citet{worrall2017harmonic}) and 3D space (e.g., \citet{cohen2018spherical, worrall2018cubenet, cohen2019gauge}) groups, and for applications other than visual data (e.g., speech recognition). 

Finally, we believe that our approach could be refined and extended to a first step towards dealing with the enumeration problem of large groups \citep{gens2014deep}, such that functions acting on the group (e.g., group convolution) are approximated by evaluating them on the set of co-occurring transformations as opposed to on the entire group. Such approximations are expected to be very accurate, as non-co-occurrent transformations are rare. This could be though of as sharping up co-occurrent attention to co-occurrent restriction.

\vspace{-7mm}
\section{Conclusion}
We have introduced the concept of co-attentive equivariant feature mapping and applied it in the context of equivariant neural networks. By attending to the co-occurrence envelope of the data, we are able to improve the performance of conventional equivariant ones on fully (rotated MNIST) and partially (CIFAR-10) rotational settings. We developed cyclic equivariant self-attention, an attention mechanism able to attend to the co-occurrence envelope of the data without disrupting equivariance to a large set of transformation groups (i.e., all transformation groups $G$, whose action in the codomain of a $G$-equivariant feature mapping produce cyclic permutations). Our obtained results support the proposed co-occurrence envelope hypothesis.

%% file: chapters/6_attgconv.tex
% path to figures directory
\graphicspath{{figures/6-attgconv/}}

%=========================================================================

% \begin{savequote}[75mm]
% Nulla facilisi. In vel sem. Morbi id urna in diam dignissim feugiat. Proin molestie tortor eu velit. Aliquam erat volutpat. Nullam ultrices, diam tempus vulputate egestas, eros pede varius leo.
% \qauthor{Quoteauthor Lastname}
% \end{savequote}

\chapter{Attentive Group Equivariant Convolutional Neural Networks}\label{chapter:attgconv}

\begin{flushright}
\textit{Based on the paper:}\break
\textit{Attentive Group Equivariant Convolutional Networks \citep{romero2020attentive}}
\end{flushright}
%=========================================================================
\vspace{-7mm}
\section{Introduction} \label{sec:6_intro}
Convolutional Neural Networks (CNNs) \cite{lecun1989backpropagation} have shown impressive performance in a wide variety of domains. The developments of CNNs as well as of many other machine learning approaches have been fueled by intuitions and insights into the composition and \textit{modus operandi} of multiple biological systems \cite{wertheimer1938gestalt, biederman1987recognition, delahunt2019insect, blake2005role,v1hypothesis, delahunt2019insect}. 
Though CNNs have achieved remarkable performance increases on several benchmark problems, their training efficiency as well as generalization capabilities are still open for improvement.
One concept being exploited for this purpose is that of \emph{equivariance}, again drawing inspiration from human beings. 

Humans are able to identify familiar objects despite modifications in location, size, viewpoint, lighting conditions and background \cite{bruce1994recognizing}. In addition, we do not just recognize them but are able to describe in detail the type and amount of modification applied to them as well \cite{vonHelmholtz1868, cassirer1944concept, schmidt2016perception}. Equivariance is strongly related to the idea of \textit{symmetricity}. As these modifications do not modify the essence of the underlying object, they should be treated (and learned) as a single concept. Recently, several approaches have embraced these ideas to preserve symmetries including translations \cite{lecun1989backpropagation}, planar rotations \cite{dieleman2016exploiting, marcos2017rotation, worrall2017harmonic, weiler2018learning, DREN, cheng2018rotdcf, hoogeboom2018hexaconv, bekkers2018roto, veeling2018rotation, lenssen2018group, smets2020pde}, spherical rotations \cite{cohen2018spherical, worrall2018cubenet, weiler20183d, thomas2018tensor, cohen2019gauge}, scaling \cite{marcos2018scale,worrall2019deep, sosnovik2020scaleequivariant} and general symmetry groups \cite{cohen2016group, kondor2018generalization, weiler2019general, cohen2019general, bekkers2020bspline,Romero2020Co-Attentive, Venkataraman2020Building}. 
\begin{figure}
    \centering
    \hfill
        \begin{subfigure}{0.20\textwidth}
        \includegraphics[width=0.9\textwidth]{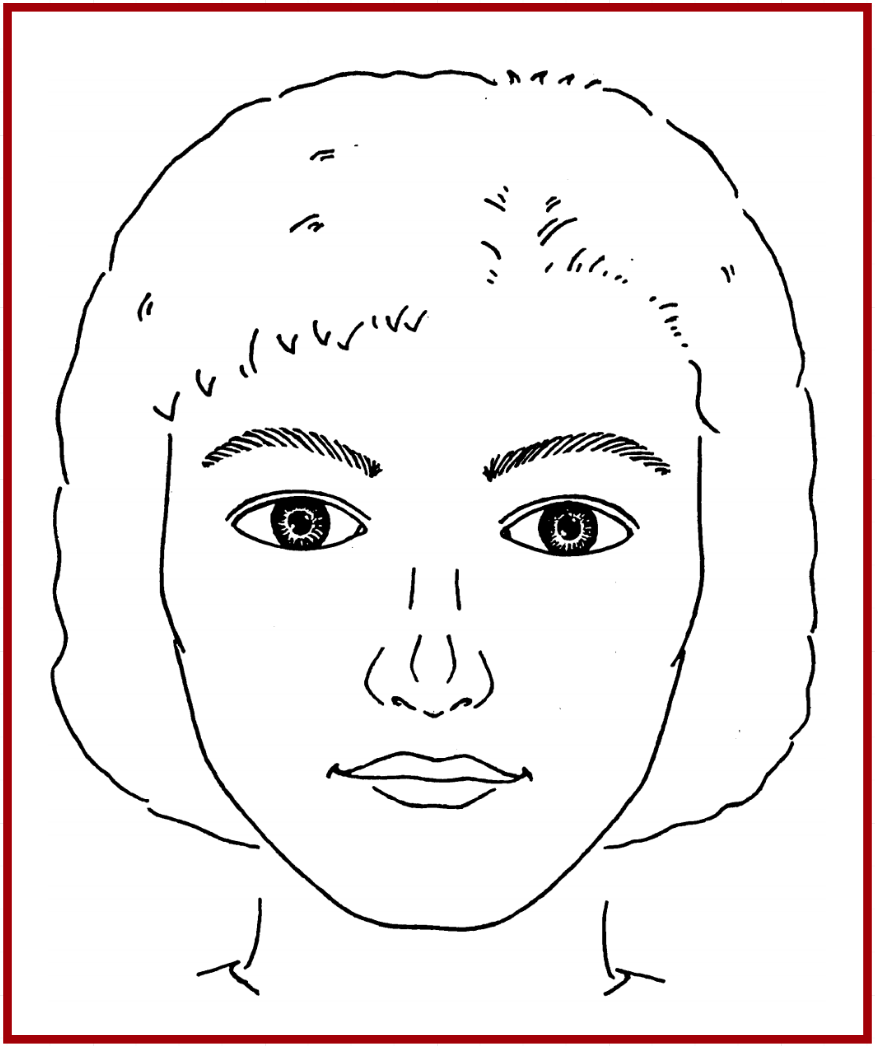}
    \end{subfigure}
    \hfill
        \begin{subfigure}{0.21\textwidth}
        \includegraphics[angle=90, width=0.9\textwidth]{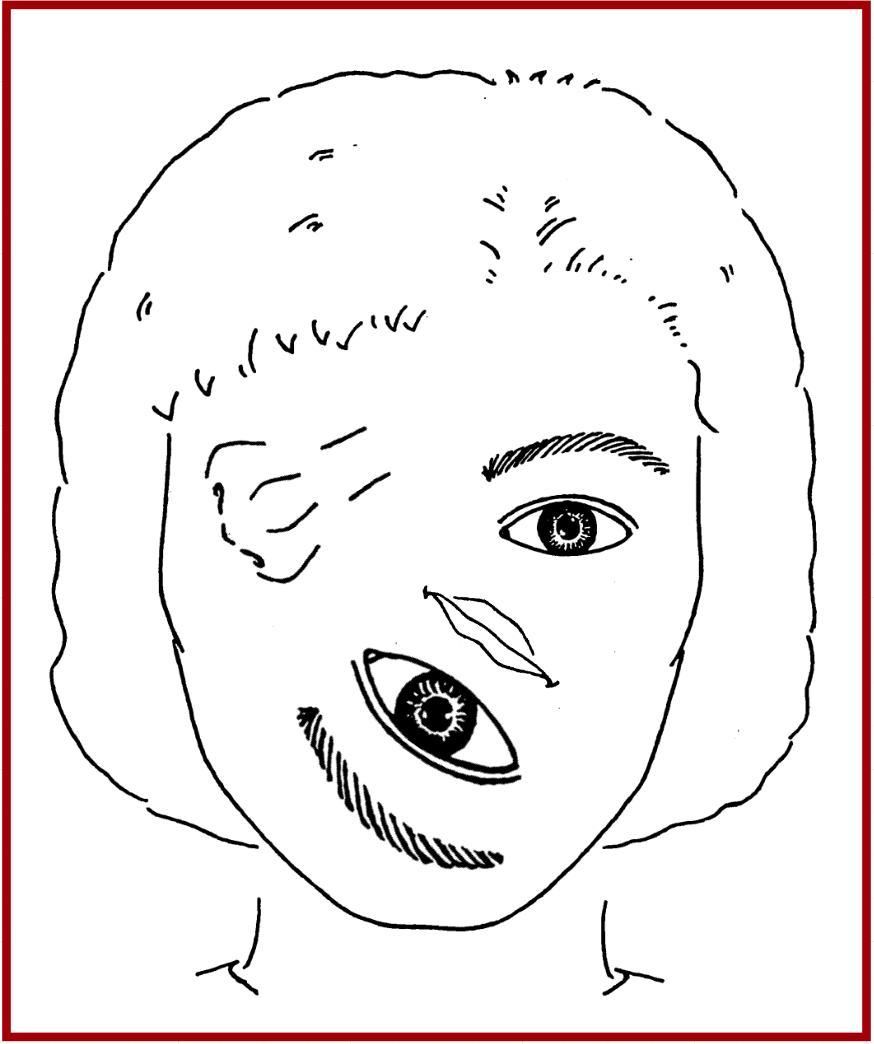}
    \end{subfigure}
    \hfill
    \begin{subfigure}{0.20\textwidth}
        \includegraphics[width=0.9\textwidth]{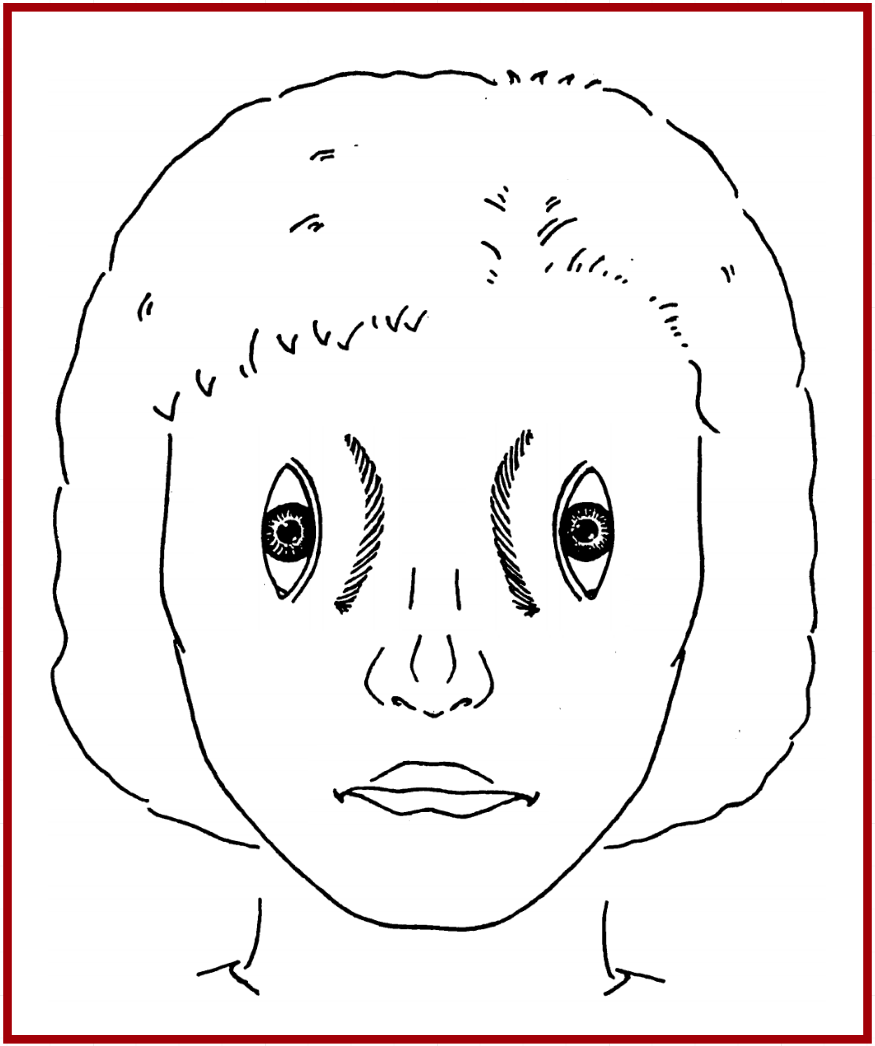}
    \end{subfigure}
    \hfill
        \begin{subfigure}{0.21\textwidth}
        \includegraphics[angle=90, width=0.9\textwidth]{good_face.png}
    \end{subfigure}
    \vspace{-2mm}
    \caption{Meaningful relationships among object symmetries. Though every figure is composed by the same elements, only the outermost examples resemble faces. The relative positions, orientations and scales of elements in the innermost examples do not match any meaningful face composition and hence, should not be labelled as such. Built upon Fig. 1 from \citet{schwarzer2000development}.}\label{fig:6_picasso}
\end{figure} 

While  group convolutional networks are able to learn powerful representations based on symmetry patterns, they lack any explicit means to learn meaningful relationships among them, e.g., relative positions, orientations and scales (Fig.~\ref{fig:6_picasso}). In this paper, we draw inspiration from another promising development in the machine learning domain driven by neuroscience and psychology, e.g., \citet{pashler2016attention}, \emph{attention}, to learn such relationships. The notion of attention is related to the idea that not all components of an input signal are \textit{per se} equally relevant for a particular task. As a consequence, given a task and a particular input signal, task-relevant components of the input should be focused during its analysis while irrelevant, possibly misleading ones should be suppressed. Attention has been broadly applied to fields ranging from natural language processing \cite{bahdanau2014, cheng2016long, vaswani2017attention}, visual understanding \cite{xu2015show, ilse2018attention, park2018bam}, and graph analysis \cite{velivckovic2017graph, zhang2020hypersagnn}.

Specifically, we present \textit{attentive group convolutions}, a generalization of the group convolution, in which attention is applied during convolution to accentuate meaningful symmetry combinations and suppress non-plausible, possibly misleading ones. We indicate that prior work on visual attention can be described as special cases of our proposed framework and show empirically that our \textit{attentive group equivariant group convolutional networks} consistently outperform conventional group equivariant ones on rot-MNIST and CIFAR-10 for the $\mathrm{SE(2)}$ and $\mathrm{E(2)}$ groups. In addition, we provide means to interpret the learned concepts trough the visualization of the equivariant attention maps.

\textbf{Contributions:}
\begin{itemize}
  \item We propose a general group theoretical framework for equivariant visual attention, \textit{the attentive group convolution}, and show that prior works on visual attention are special cases of our framework.
    \item We introduce a specific type of network referred to as \textit{attentive group convolutional networks} as an instance of this theoretical framework. 
    \item We show that our \textit{attentive group convolutional networks} consistently outperform plain group equivariant ones.
  \item We provide means to interpret the learned concepts via visualization of the predicted equivariant attention maps.
\end{itemize}
\vspace{-7mm}
\section{Preliminaries} \label{sec:6_prelim}
Before describing our approach, we first define crucial prior concepts: (group) convolutions and attention mechanisms.
\vspace{-7mm}
\subsection{Spatial Convolution and Translation Equivariance}
Let $f$, $\psi:\mathbb{R}^{d}\rightarrow \mathbb{R}^{N_{\tilde{c}}}$ be a vector valued signal and filter on $\mathbb{R}^{d}$, such that $f=\{f_{\tilde{c}}\}_{\tilde{c}=1}^{N_{\tilde{c}}}$ and $\psi=\{\psi_{\tilde{c}}\}_{\tilde{c}=1}^{N_{\tilde{c}}}$. The spatial convolution ($\star_{\mathbb{R}^{d}}$) is defined as:
\begin{equation} \label{eq:6_norm_conv}
    [f \star_{\mathbb{R}^{d}} \psi](y) = \sum_{\tilde{c}=1}^{N_{\tilde{c}}}\int_{\mathbb{R}^{d}}f_{\tilde{c}}(x)\psi_{\tilde{c}}(x - y) \, {\rm d}x
\end{equation}
Intuitively, Eq. \ref{eq:6_norm_conv} resembles a collection of $\mathbb{R}^{d}$ inner products between the input signal $f$ and $y$-translated versions of $\psi$.
Since the continuous integration in Eq. \ref{eq:6_norm_conv} is usually performed on signals and filters captured in a discrete grid $\mathbb{Z}^{d}$, the integral on $\mathbb{R}^{d}$ is reduced to a sum on $\mathbb{Z}^{d}$. In our derivations, however, we stick to the continuous case as to guarantee the validity of our theory for techniques defined on continuous spaces, e.g., steerable and Lie group convolutions \cite{cohen2016steerable, worrall2017harmonic, bekkers2018roto}.

To study (and generalize) the properties of the convolution, we rewrite Eq. \ref{eq:6_norm_conv} using the translation operator
$\mathcal{L}_{y}$: 
\begin{equation}\label{eq:6_norm_conv_goperator}
    [f \star_{\mathbb{R}^{d}} \psi](y) = \sum_{\tilde{c}=1}^{N_{\tilde{c}}}\int_{\mathbb{R}^{d}}f_{\tilde{c}}(x)\mathcal{L}_{y}[\psi_{\tilde{c}}](x) \,{\rm d}x
\end{equation}
where $\mathcal{L}_{y}[\psi_{\tilde{c}}](x) = \psi_{\tilde{c}}(x-y)$. Note that the translation operator $\mathcal{L}_{y}$ is indexed by an amount of translation $y$. Resultantly, we actually consider a set of operators $\{\mathcal{L}_{y}\}_{y \in \mathbb{R}^{d}}$ that indexes the set of all possible translations $y \in \mathbb{R}^{d}$. A fundamental property of the convolution is that it commutes with translations:
\begin{equation} \label{eq:6_equiv}
    \mathcal{L}_{y}[f \star_{\mathbb{R}^{d}} \psi](x) = \big[\mathcal{L}_{y}[f] \star_{\mathbb{R}^{d}} \psi \big](x), \ \  x,y \in \mathbb{R}^{d}.
\end{equation}
In other words, convolving a $y$-translated signal $\mathcal{L}_{y}[f]$ with a filter is equivalent to first convolving the original signal $f$ with the filter $\psi$, and $y$-translating the obtained response next. This property is referred to as \textit{translation equivariance} and, in fact, convolution (and reparametrizations thereof) is the \textit{only} linear \textit{translation equivariant} mapping \cite{ kondor2018generalization, cohen2019general, bekkers2020bspline}.
% =====SubSECTION=====

\vspace{-7mm}
\subsection{Group Convolution and Group Equivariance}
The convolution operation can be extended to general transformations by utilizing a larger set of transformations $\{\mathcal{L}_{g}\}_{g \in G}$, s.t. $\{\mathcal{L}_{y}\}_{y \in \mathbb{R}^{d}} \subseteq \{\mathcal{L}_{g}\}_{g \in G}$. However, in order to preserve equivariance, we must restrict the class of transformations allowed in $\{\mathcal{L}_{g}\}_{g \in G}$. To formalize this intuition, we first present some important concepts from \textit{group theory}.
% =====SubSubSECTION=====
% \vspace{-7mm}
% \subsubsection{Preliminaries from Group Theory} \label{sec:6_group_theory}

\textbf{Groups.} A \textit{group} is a tuple ($G$, $\cdot$) consisting of a set $G$, $g \in G$, and a binary operation $\cdot: G\times G \rightarrow G$, referred to as the \textit{group product}, that satisfies the following axioms:
\begin{itemize}
    \item \textit{Closure:} For all $h$, $g \in G$, $h \cdot g \in G$. 
    \item \textit{Identity:} There exists an $e \in G$, such that $e\cdot g = g \cdot e = g$.
    \item \textit{Inverse:} For all $g \in G$, there exists an element $g^{-1} \in G$, such that $g \cdot g^{-1} = g^{-1} \cdot g = e$.
    \item \textit{Associativity:} For all $g,h,k \in G$, $(g\cdot h) \cdot k = g \cdot (h \cdot k)$.
\end{itemize}
\textbf{Group actions.} Let $G$ and $X$ be a group and a set, respectively. The (left) \textit{group action} of $G$ on $X$ is a function $\odot: G \times X \rightarrow X$ that satisfies the following axioms:% the axioms:
\begin{itemize}
    \item \textit{Identity:} If $e$ is the identity of $G$, then, for any $x \in X$, $e \odot x=x$. 
    \item \textit{Compatibility:} For all $g$, $h \in G$, $x \in X$, $g \odot (h \odot x) = (g \cdot h)\odot x$.
\end{itemize}
In other words, the action of $G$ on $X$ describes how the elements $x\in X$ are transformed by $g \in G$. 
For brevity, we omit the operations $\cdot$ and $\odot$ and refer to the set $G$ as a group, to elements $g \cdot h$ as $gh$ and to actions $(g \odot x)$ as $gx$. %It should be clear from context what operation is being referred to.

\textbf{Semi-direct product and affine groups.} In practice, one is mainly interested in the analysis of data (and hence convolutions) defined on $\mathbb{R}^{d}$. Consequently, groups of the form $G = \mathbb{R}^{d} \rtimes H$, resulting from the \textit{semi-direct product} ($\rtimes$) between the translation group $\mathbb{R}^{d}$ and an arbitrary (Lie) group $H$ that acts on $\mathbb{R}^{d}$ (e.g., rotation, scaling, mirroring), are of main interest. This family of groups is referred to as \textit{affine groups} and their group product is defined as:
\begin{equation}\label{eq:6_semidirect}
    g_{1}g_{2}=(x_{1},h_{1})(x_{2},h_{2})=(x_{1}+h_{1}x_{2}, h_{1} h_{2})
\end{equation}
where $g_{1}=(x_{1}, h_{1})$, $g_{2}=(x_{2}, h_{2}) \in G$, $x_{1}, x_{2} \in \mathbb{R}^{d}$ and $h_1, h_2 \in H$. Some important affine groups are the roto-translation ($\mathrm{SE(d)} = \mathbb{R}^{d} \rtimes \mathrm{SO(d)}$), the scale-translation ($\mathbb{R}^{d} \rtimes \mathbb{R}^{+}$) and the euclidean ($\mathrm{E(d)}=\mathbb{R}^{d} \rtimes \mathrm{O(d)}$) groups.

\textbf{Group representations.} Let $G$ be a group and $\mathbb{L}_{2}(X)$ be a space of functions defined on some vector space $X$. The (left) regular \textit{group representation} of $G$ on functions $f \in \mathbb{L}_{2}(X)$ is a transformation  $\mathcal{L}: G \times \mathbb{L}_{2}(X) \rightarrow \mathbb{L}_{2}(X)$, $(g,f) \mapsto \mathcal{L}_{g}[f]$, such that it shares the group structure via:
\begin{gather}
    \mathcal{L}_{g}\mathcal{L}_{h}[f](x) = \mathcal{L}_{gh}[f](x) \\
    \mathcal{L}_{g}[f](x) 
    := f(g^{-1}x) 
\end{gather}
for any $g, h \in G$, $f \in \mathbb{L}_{2}(X)$, $x \in X$. That is, concatenating two such transformations, parametrized by $g$ and $h$, is equivalent to one transformation parametrized by $gh \in G$.
Intuitively, the representation of $G$ on a function $f \in \mathbb{L}_{2}(X)$ describes how the function as a whole, i.e., $f(x)$, $\forall \ x \in X$, is transformed by the effect of group elements $g \in G$.

If the group $G$ is affine, i.e., $G = \mathbb{R}^{d} \rtimes H$, the (left) group representation $\mathcal{L}_{g}$ can be split as:
\begin{equation} \label{eq:6_repr_decomp}
\mathcal{L}_{g}[f](x) = \mathcal{L}_{y} \mathcal{L}_{h}[f](x)
\end{equation}
with $g = (y, h) \in G$, $y \in \mathbb{R}^{d}$ and $h \in H$. This property is key for the efficient implementation of functions on groups.
% =====SubSubSECTION=====
\vspace{-7mm}
\subsubsection{The Group Convolution} 
Let $f$, $\psi:G\rightarrow \mathbb{R}^{N_{\tilde{c}}}$ be a vector valued signal and kernel on $G$. %, and $\mathcal{L}_{g}[\psi]$ be the $g$-transformed kernel. 
The group convolution ($\star_{G}$) is defined as: 
\begin{align}\label{eq:6_g_conv}
    [f\star_{G} \psi ](g)
    &=\sum_{\tilde{c}=1}^{N_{\tilde{c}}}\int_{G}f_{\tilde{c}}(\tilde{g})\psi_{\tilde{c}}(g^{-1}\tilde{g}) \,{\rm d}\tilde{g}\\ \label{eq:6_g_conv2}
    &=\sum_{\tilde{c}=1}^{N_{\tilde{c}}}\int_{G}f_{\tilde{c}}(\tilde{g})\mathcal{L}_{g}[\psi_{\tilde{c}}](\tilde{g}) \,{\rm d}\tilde{g}
\end{align} 
\begin{figure}
    \centering
    \includegraphics[width=0.7\textwidth]{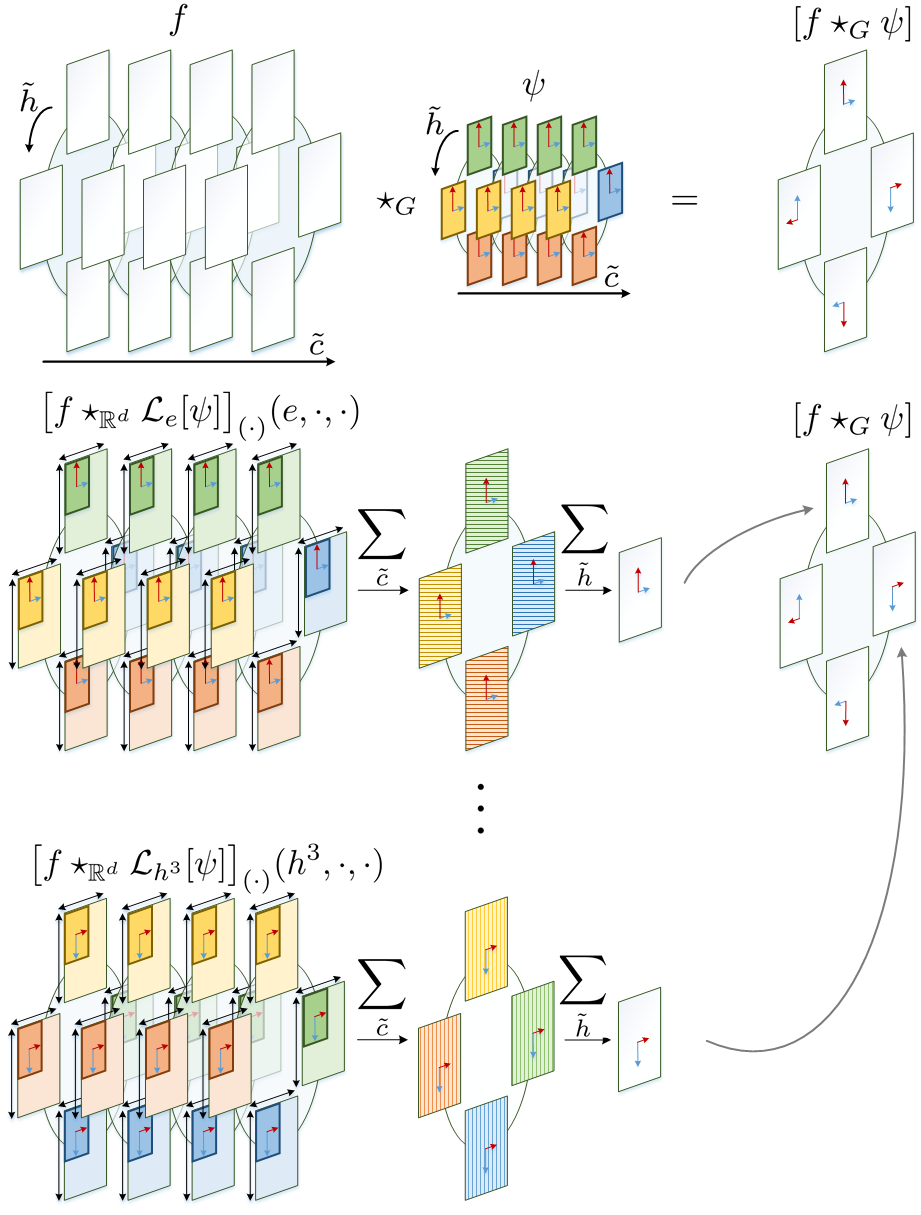}
    \vskip -2mm
    \caption{Group convolution on the roto-translation group $\mathrm{SE(2)}$ for discrete rotations by 90 degrees (also called the $p4$ group). The $p4$ group is defined as $H = \{e, h, h^{2}, h^{3}\}$, with $h$ depicting a 90$^{\circ}$ rotation. The group convolution corresponds to $|H|=4$ convolutions between the input $f$ and $h$-transformations of the filter $\psi$, $\mathcal{L}_{h}[\psi]$, $\forall \ h \in H$. Each of these convolutions is equal to the sum over group elements $\tilde{h} \in H$ and channels $\tilde{c} \in [N_{\tilde{c}}]$ of the spatial channel-wise convolutions $\big[f_{\tilde{c}} \star_{\mathbb{R}^{2}} \mathcal{L}_{h}[\psi_{\tilde{c}}]\big]$ among $f$ and $\mathcal{L}_{h}[\psi]$.}
    \label{fig:6_affine_conv}
\end{figure}
Differently to Eq. \ref{eq:6_norm_conv_goperator}, the domain of the signal $f$, the filter $\psi$ and the group convolution itself $[f \star_{G} \psi]$ are now defined on the group $G$.\footnote{Note that Eq. \ref{eq:6_norm_conv_goperator} matches Eq. \ref{eq:6_g_conv2} with the substitution $G=\mathbb{R}^{d}$. It follows that $\mathcal{L}_{g}[f](x)=f(g^{-1}x)=f(x-y)$, where $g^{-1}=-y$ is the inverse of $g$ in the translation group $(\mathbb{R}^{d}, +)$ for $g = y$.} Intuitively, the group convolution resembles a collection of inner products between the input signal $f$ and $g$-transformed versions of $\psi$.
A key property of the group convolution is that it generalizes equivariance (Eq. \ref{eq:6_equiv}) to arbitrary groups, i.e., it commutes with $g$-transformations:
\begin{equation}\label{eq:6_group_equiv}
        \mathcal{L}_{\bar{g}}[f \star_{G} \psi](g) = \big[\mathcal{L}_{\bar{g}}[f] \star_{G} \psi\big](g),\ \   g,\bar{g} \in G.
\end{equation}
In other words, group convolving a $\bar{g}$-transformed signal $\mathcal{L}_{\bar{g}}[f]$ with a filter $\psi$ is equivalent to first convolving the original signal $f$ with the filter $\psi$, and $\bar{g}$-transforming the obtained response next. This property is referred to as \textit{group equivariance} and, just as for spatial convolutions, the group convolution (or reparametrizations thereof) is the \textit{only} linear \textit{$G$-equivariant} map \cite{kondor2018generalization,cohen2019general,bekkers2020bspline}. 

% =====SubSubSECTION=====
\textbf{Group convolution on affine groups.} For affine groups, the group convolution (Eq. \ref{eq:6_g_conv2}) can be decomposed, without modifying its properties, by taking advantage of the group structure and the representation decomposition (Eq. \ref{eq:6_repr_decomp}) as:% in Eqs. \ref{eq:6_gconv_groupstruct} and \ref{eq:6_gconv_groupstruct_reprdecomp}, respectively,
\begin{align}
\hspace{-1mm}[f &\star_{G} \psi](g) =\sum_{\tilde{c}=1}^{N_{\tilde{c}}}  \int \limits_H \int \limits_{\mathbb{R}^{2}} f_{\tilde{c}}(\tilde{x},\tilde{h})\mathcal{L}_{g}[\psi_{\tilde{c}}](\tilde{x},\tilde{h}) \,{\rm d}\tilde{x}\,{\rm d}\tilde{h}\\[-1\jot] \label{eq:6_gconv_groupstruct_reprdecomp}
&\quad\quad\ \ \vspace{1mm}  =\sum_{\tilde{c}=1}^{N_{\tilde{c}}}\int \limits_H \int \limits_{\mathbb{R}^{2}} f_{\tilde{c}}(\tilde{x},\tilde{h})\mathcal{L}_{x}\mathcal{L}_{h}[\psi_{\tilde{c}}](\tilde{x},\tilde{h}) \,{\rm d}\tilde{x}\,{\rm d}\tilde{h}
\end{align}
where $g = (x,h)$, $\tilde{g}=(\tilde{x},\tilde{h}) \in G$, $x$, $\tilde{x}\in \mathbb{R}^{d}$ and $h$, $\tilde{h} \in H$. By doing so, the group convolution can be separated into $|H|$ spatial convolutions of the input signal $f$ for each $h$-transformed filter $\mathcal{L}_{h}[\psi]$ (Fig. \ref{fig:6_affine_conv}):
\begin{equation}
 \label{eq:6_gconv_groupstruct_smallconvs}
[f \star_{G} \psi](x, h)=\sum_{\tilde{c}=1}^{N_{\tilde{c}}}\int_H \big[f_{\tilde{c}} \star_{\mathbb{R}^{2}} \mathcal{L}_{h}[\psi_{\tilde{c}}]\big](x, \tilde{h})\,{\rm d}\tilde{h}
\end{equation}
Resultantly, the computational cost of a group convolution is roughly equivalent
to that of a spatial convolution with a filter bank of size $N_{\tilde{c}}\times |H|$ \cite{cohen2016group, worrall2019deep, cohen2019gauge}.
% =====SubSECTION=====
\vspace{-7mm}
\subsection{Attention, Self-Attention and Visual Attention} \label{sec:6_attention}
\begin{figure}
\floatbox[{\capbeside\thisfloatsetup{capbesideposition={right},capbesidewidth=8.5cm}}]{figure}[\FBwidth]
{\caption{English to French translation. Brighter depicts stronger influence. Note how relevant parts of the input sentence are highlighted as a function of the current output word during translation. Taken from \citet{bahdanau2014}.}\label{fig:6_att}}
{\includegraphics[width=3.2cm]{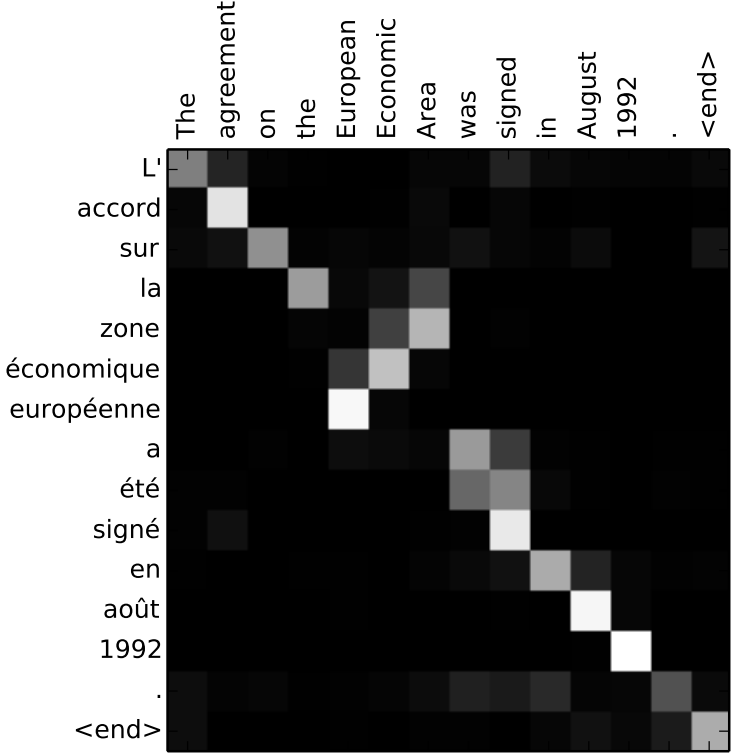}}
\end{figure}

Attention mechanisms find their roots in recurrent neural network (RNN) based machine translation. Let  $\varphi(\cdot)$ be an arbitrary non-linear mapping (e.g., a neural network), $\underline{y}=\{y_{j}\}_{j=1}^{m}$ be a sequence of target vectors $y_i$, and $\underline{x}=\{x_{i}\}_{i=1}^{n}$ be a source sequence, whose elements influence the prediction of each value $y_{j} \in \underline{y}$. 
In early models (e.g., \citet{kalchbrenner2013recurrent, cho2014learning}), features in the input sequence are aggregated into a context vector $c = \sum_{i}\varphi(x_{i})$ which is used to augment the hidden state in RNN layers. These models assume that source elements $x_{i}$ contribute equally to \textit{every} target element $y_{j}$ and hence, that the same context vector $c$ can be utilized for all target positions $y_{j}$, which does not generally hold (Fig. \ref{fig:6_att}). 

\citet{bahdanau2014} proposed the inclusion of \textit{attention coefficients} $\alpha_{i}=\{\alpha_{i,j}\}$, $[n] = \{1, ..., n\}$, $i \in [n]$, $j \in [m]$, $\sum_{i}\alpha_{i,j}=1$, to modulate the contributions of the source elements $x_{i}$ as a function of the current target element $y_{j}$ by means of an adaptive context vector $c_{j} = \sum_{i}\alpha_{i, j}\varphi(x_{i})$. Thereby, they obtained large improvements both in performance and interpretability. 
Recently, attention has been extended to several other machine learning tasks, e.g., \citet{vaswani2017attention,velivckovic2017graph, park2018bam}). The main development behind these extensions was \textit{self-attention} \cite{cheng2016long}, where, in contrast to conventional attention, the target and source sequences are equal, i.e., $\underline{x}=\underline{y}$. Consequently, the attention coefficients $\alpha_{i,j}$ encode correlations among input element pairs $(x_{i}, x_{j})$. For vision tasks, self-attention has been proposed to encode visual co-occurrences in data \cite{hu2018squeeze, wang2018non, park2018bam, woo2018cbam, cao2019gcnet,bello2019attention, ramachandran2019stand, Romero2020Co-Attentive}. Unfortunately, its application on visual and, in general, on high-dimensional data is non-trivial.

\vspace{-7mm}
\subsubsection{Visual Attention}\label{sec:6_visual_att}

In the context of visual attention, consider a feature map $f:{X}\rightarrow\mathbb{R}^{N_c}$ to be the source \enquote{sequence}\footnote{In the machine translation context, we can think of $f$ as a sequence $\underline{x} = \{f(x_{i})\}_{i=1}^{n}$, with $n {=} |X|$ elements.}. Self-attention then imposes the learning of a total $n^{2} = |X|^2$ attention vectors $\alpha_{i,j} \in \mathbb{R}^{N_{\tilde{c}}}$, which rapidly becomes unfeasible with increasing feature map size.
Interestingly, \citet{cao2019gcnet} and \citet{zhu2019empirical} empirically demonstrated that, for visual data, the attention coefficients $\{\alpha_{i,j}\}$ are approximately invariant to changes in the target position $x_j$. 
Consequently, they proposed to approximate the attention coefficients $\{\alpha_{i,j}\} \in \mathbb{R}^{|X|^{2} \times N_{\tilde{c}}}$ by a single vector $\{\alpha_{i}\} \in \mathbb{R}^{|X| \times N_{\tilde{c}}}$ which is independent of target position $x_{j}$. 
Despite this significant reduction in complexity, the dimensionality of $\{\alpha_i\}$ is still very large and further simplifications are mandatory. To this end, existing works \cite{hu2018squeeze, woo2018cbam} replace the input $f$ with a much smaller vector of input statistics $s$ that summarizes relevant information from $f$.

For instance, the SE-Net \cite{hu2018squeeze} utilizes global average pooling to produce a vector of channel statistics of $f$, $s^{\mathcal{C}} \in \mathbb{R}^{N_{\tilde{c}}}$, $s^{\mathcal{C}} = \frac{1}{|\mathbb{R}^{d}|}\int_{\mathbb{R}^{d}}f_{\tilde{c}}(x)\,dx$, which is subsequently passed to a small fully-connected network $\varphi^{\mathcal{C}}(\cdot)$ to compute channel attention coefficients $\alpha^{\mathcal{C}} = \{\alpha^{\mathcal{C}}_{\tilde{c}}\}_{\tilde{c}=1}^{N_{\tilde{c}}} = \varphi^{\mathcal{C}}(s^{\mathcal{C}})$. These attention coefficients are then utilized to modulate the corresponding input channels $f_{\tilde{c}}$.

Complementary to channel attention akin to that of the SE-Net, \citet{park2018bam} utilize a similar strategy for spatial attention. Specifically, they utilize channel average pooling to generate a vector of spatial statistics of $f$, $s^{\mathcal{X}} \in \mathbb{R}^{d}$, $s^{\mathcal{X}}=\frac{1}{N_{\tilde{c}}}\sum_{\tilde{c}=1}^{N_{\tilde{c}}}f_{\tilde{c}}(x)$, which is subsequently passed to a small convolutional network $\varphi^{\mathcal{X}}(\cdot)$ to compute spatial attention coefficients $\alpha^{\mathcal{X}}=\{\alpha^{\mathcal{X}}(x)\}_{x \in \mathbb{R}^{2}} = \varphi^{\mathcal{X}}(s^{\mathcal{X}})$. These attention coefficients are then utilized to modulate the corresponding spatial input positions $f(x)$. 
Recent works include more statistical information, e.g., max responses \cite{woo2018cbam}, or replace pooling by convolutions \cite{cao2019gcnet}.

% =====SECTION=====
\vspace{-7mm}
\section{Attentive Group Equivariant Convolution}
In this section, we propose our generalization of visual self-attention, discuss its properties and relations to prior work. 
\begin{figure}
\floatbox[{\capbeside\thisfloatsetup{capbesideposition={right},capbesidewidth=9cm}}]{figure}[\FBwidth]
{\caption{Same colors depict equal weights. The first column of $\mathcal{A^{C}}$ corresponds to $\psi$ and the following ones to $\mathcal{L}_{h}[\psi]$, obtained via cyclic permutations. See how $\{\mathcal{L}_{h}[\psi]\}_{h \in H}$ resembles a circulant matrix. Taken from \citet{Romero2020Co-Attentive}.}\label{fig:6_circulant}}
{\includegraphics[width=2.2cm]{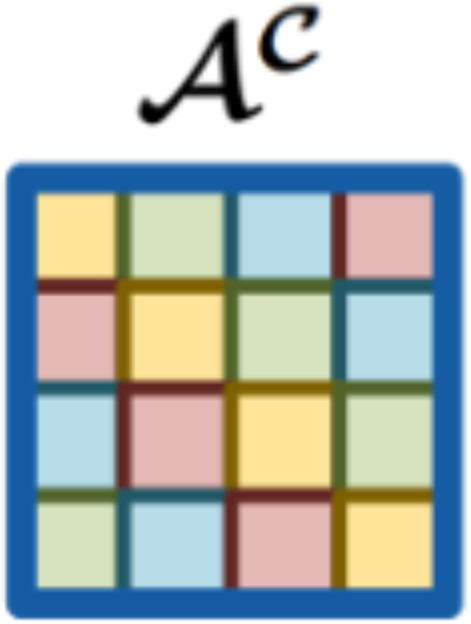}}
\end{figure}

Let $f, \psi: G \rightarrow \mathbb{R}^{N_{\tilde{c}}}$ be a vector valued signal and kernel on $G$, and let $\alpha: G \times G \rightarrow [0,1]^{N_{\tilde{c}}}$ be an \textit{attention map} that takes target and source elements $g,\tilde{g} \in G$, respectively, as input. We define the \textit{attentive group convolution} ($\star_{G}^{\alpha}$) as:
\begin{equation} \label{eq:6_att_g_conv}
    [f \star^{\alpha}_{G} \psi](g)
    =\sum_{\tilde{c}=1}^{N_{\tilde{c}}}\int_{G}\alpha_{\tilde{c}}(g,\tilde{g})f_{\tilde{c}}(\tilde{g})\mathcal{L}_{g}[\psi_{\tilde{c}}](\tilde{g}) \,{\rm d}\tilde{g}
\end{equation}
with $\alpha {=} \mathcal{A}[f]$ computed by some \textit{attention operator} $\mathcal{A}$. As such, the attentive group convolution modulates the contributions of group elements $\tilde{g} \in G$ at different channels $\tilde{c} \in [N_{\tilde{c}}]$ during pooling.\footnote{Note that Eq. \ref{eq:6_att_g_conv} is equal to Eq. \ref{eq:6_g_conv2} up to a multiplicative factor $\alpha_{\tilde{c}}(g,\tilde{g})^{-1}$, if $\alpha_{\tilde{c}}(g,\tilde{g})$ is constant for every $g,\tilde{g} \in G$, $\tilde{c} \in [N_{\tilde{c}}]$.} The properties and conditions on $\mathcal{A}$ are summarized in Thm.~\ref{thm1}. An extensive motivation as well as its proof are provided in Appx.~\ref{appx:6_gralized_visual_self_att}.

\begin{theorem}
\label{thm1}
The attentive group convolution is an equivariant operator if and only if the attention operator $\mathcal{A}$ satisfies:
\begin{equation}
\label{eq:6_equivconstraint}
\forall_{\overline{g},g,\tilde{g} \in G}: \;\;  \mathcal{A}[\mathcal{L}_{\overline{g}}f](g,\tilde{g}) = \mathcal{A}[f](\overline{g}^{-1} g, \overline{g}^{-1} \tilde{g})
\end{equation}
If, moreover, the maps generated by $\mathcal{A}$ are invariant to one of its arguments, and, hence, exclusively attend to either the input or the output domain (Sec. \ref{sec:6_att_sequence}), then $\mathcal{A}$ satisfies Eq.~\ref{eq:6_equivconstraint} if and only if it is equivariant and thus, based on group convolutions.
\end{theorem}
% \begin{corollary}
% \label{cor1}
% Each attention attention map $\alpha:G\times G \rightarrow [0,1]$ that is left-invariant to either one of arguments, and thus exclusively attends either the input or output domain, satisfies Thm.~\ref{thm1} iff its attention operator is equivariant.
% \end{corollary}
% \begin{proof}
% Proofs of Thm.~\ref{thm1} and Col.~\ref{cor1} are given in the supplementary materials.
% \end{proof}

\vspace{-7mm}
\subsection{Tying Equivariance and Visual Attention Together} \label{sec:6_tyingtogether}
Interestingly, and, perhaps in some cases unaware of it, \textit{all} of the visual attention approaches outlined in Section \ref{sec:6_visual_att}, as well as all of those we are aware of \cite{xu2015show, hu2018squeeze, park2018bam, woo2018cbam, wang2018non, ilse2018attention, hu2019local, ramachandran2019stand, cao2019gcnet, chen2019graph, bello2019attention, lin2019contextgated, diaconu2019affine, Romero2020Co-Attentive} \textit{exclusively utilize translation (or group) equivariance preserving maps for the generation of the attention coefficients and, hence, constitute altogether group equivariant networks by which they satisfy Thm.~\ref{thm1}}.

As will be explained in the following sections, all these works resemble special cases of Eq.~\ref{eq:6_att_g_conv} by substituting $G$ with the corresponding group and modifying the specifications about how $\alpha$ is calculated (Sec. \ref{sec:6_equiv_attcoefs} - \ref{sec:6_att_sequence}).

% =====SubSECTION=====
\vspace{-7mm}
\subsubsection{Translation Equivariant Visual Atention}
Since convolutions as well as popular pooling operations are translation equivariant, the visual attention approaches outlined in Sec. \ref{sec:6_visual_att} are translation equivariant as well.\footnote{In fact, conventional pooling operations (e.g., max, average) can be written as combinations of convolutions and pointwise non-linearities, which are translation equivariant, as well.} 
One particular case worth emphasising is that of SE-Nets. Here, a fully-connected network $\varphi^{\mathcal{C}}$, a non-translation equivariant map, is used to generate the channel attention coefficients $\alpha^{\mathcal{C}}$. However, $\varphi^{\mathcal{C}}$ \textit{is} indeed translation equivariant. Recall that $\varphi^{\mathcal{C}}$ receives $s^{\mathcal{C}}$ as input, a signal obtained via global average pooling (a convolution-like operation). Resultantly, $s^{\mathcal{C}}$ can be interpreted as a $\mathbb{R}^{N_{\tilde{c}} \times 1 \times 1}$ tensor and hence, applying a fully connected layer to $s^{\mathcal{C}}$ equals a pointwise convolution between $s^{\mathcal{C}}$ and a filter $\psi_{\text{fully}}\in \mathbb{R}^{N_o \times N_{\tilde{c}} \times 1 \times 1}$ with $N_o$ output channels.\footnote{This resembles a depth-wise separable convolution \cite{chollet2017xception} with the first convolution given by global average pooling.} 

% =====SubSECTION=====
\vspace{-7mm}
\subsubsection{Group Equivariant Visual Attention}
To the best of our knowledge, the only work that provides a group theoretical approach towards visual attention  is that of \citet{Romero2020Co-Attentive}. Here, the authors consider affine groups $G$ with elements $g {=} (x,h)$, $x \in \mathbb{R}^{d}$, $h \in H$ and cyclic permutation groups $H$. Consequently, they utilize a cyclic permutation equivariant map, $\varphi^{\mathcal{H}}(\cdot)$, to generate attention coefficients $\alpha^{\mathcal{H}}(h)$, $h \in H$, with which the corresponding elements $h$ are modulated. As a result, their proposed attention strategy is $H$-equivariant. 
To preserve translation equivariance, and hence, $G$-equivariance, $\varphi^{\mathcal{H}}$ is re-utilized at every spatial position $x \in \mathbb{R}^{d}$. This is equivalent to combining $\varphi^{\mathcal{H}}$ with a pointwise filter on $\mathbb{R}^{d}$.
\citet{Romero2020Co-Attentive} found that equivariance to cyclic groups $H$, can \textit{only} be achieved by constraining $\varphi^{\mathcal{H}}$ to have a \textit{circulant structure}. This is equivalent to a convolution with a filter $\psi$, whose group representations $\mathcal{L}_{h}$ induces cyclical permutations of itself (Fig.~\ref{fig:6_circulant}). Hence, it resembles a group convolution by which Thm.~\ref{thm1} is satisfied.

The work of \citet{Romero2020Co-Attentive} exclusively performs attention on the $h$ component of the group elements $g = (x,h) \in G$ and is only defined for (block) cyclic groups. Consequently, it does not consider spatial relationships during attention (Fig. \ref{fig:6_picasso}) and is not applicable to general groups. Conversely, our proposed framework allows for simultaneous attention on both components of the group elements $ g = (x, h)$ in a $G$ equivariance preserving manner.

% =====subSECTION=====
\vspace{-7mm}
\subsection{Efficient Group Equivariant Attention Maps}\label{sec:6_equiv_attcoefs}
Attentive group convolutions impose the generation of an additional attention map $\alpha: G \times G \rightarrow [0,1]^{N_{\tilde{c}}}$, which is computationally demanding. To reduce this computational burden, we exploit the fact that visual data is defined on $\mathbb{R}^{d}$ and, hence, relevant groups are affine, to provide an efficient factorization of the attention map $\alpha$. 

In Sec.~\ref{sec:6_visual_att} we indicated that attention coefficients $\alpha$ can be equivariantly factorized into spatial and channel components. We build upon this idea and factorize attention via:
$$
\alpha_{\tilde{c}}(g,\tilde{g}):=\alpha^{\mathcal{X}}(({x},h),(\tilde{{x}},\tilde{h}))\alpha_{\tilde{c}}^{\mathcal{C}}(h,\tilde{h})
$$
where $\alpha^{\mathcal{X}}$ attends for spatial relations without considering channel characteristics and $\alpha^{\mathcal{C}}$ attends for patterns in the channel- and $H$-axis, but ignores spatial patterns. We thus factorize $\alpha$ into a \emph{spatial attention map} $\alpha^{\mathcal{X}}: G \times G \rightarrow [0,1]$ and a \emph{channel attention map} $\alpha^{\mathcal{C}}: H \times H \rightarrow [0,1]^{N_{\tilde{c}}}$. 
Findings in literature have shown that, for visual data, attention maps are almost equivalent for different query positions and thus, only query-independent dependencies are learnt \cite{cao2019gcnet,zhu2019empirical}. Based on this observation, we further simplify $\alpha^{\mathcal{X}}$ to be invariant over spatial positions either at the input or output space. Since separate convolutional filters $\psi$ could possibly benefit from different attention maps, we omit spatial positions in the input space (see Sec.~\ref{sec:6_att_operator} for details). In other words, we replace $\alpha^{\mathcal{X}}(g,\tilde{g})$ with $\alpha^{\mathcal{X}}(g,\tilde{h})$, an spatial position invariant attention map over the input space: $\alpha^{\mathcal{X}}: G \times H \rightarrow [0,1]$.

Conveniently, attention coefficients of type $\alpha: \mathbb{R}^d \times H \rightarrow [0,1]^{N_{\tilde{c}}}$ can be interpreted as functions on $\mathbb{R}^d$ with pointwise visualizations $\tilde{x} \mapsto \alpha(\tilde{x},\tilde{h})$ for each $\tilde{x}\in \mathbb{R}^{d}$. Resultantly, we are able to aid the interpretability of the learned concepts and of the attended symmetries (e.g., Figs.~\ref{fig:6_examples}, \ref{fig:6_pcam_examples}, \ref{fig:6_good_examples}).
% Conveniently, attention coefficients of type $\alpha_{\tilde{c}}: \mathbb{R}^d \rtimes H \rightarrow [0,1]$ can be visualized as functions on $\mathbb{R}^d$ (e.g., with for each $h$ a point-wise visualizations of $\tilde{x} \mapsto \alpha_{\tilde{c}}(\tilde{x},\tilde{h})$) which aids the interpretability of the learned concepts and attended symmetries (Fig.~\ref{fig:6_examples}).
% In channel attention, however, we choose to focus attention on both input and target poses $h$ and $\tilde{h}$, and in the following subsection design a particular attention network which is based on that of \citet{woo2018cbam}, but which guarantees that the equivariance condition of Thm.~\ref{thm1} is satisfied. In Sec.~\ref{sec:6_att_sequence} we consider a simplification hereof that only depends on $\tilde{h}$ and thereby significantly reduces cost of computation.

% =====subSECTION=====
\vspace{-7mm}
\subsubsection{The Attention Operator $\mathcal{A}$}\label{sec:6_att_operator}
Recall that the attention map $\alpha$ is computed via an attention operator $\mathcal{A}$. In the most general case, $\alpha$ and, hence $\mathcal{A}$, is a function of both the input signal $f$ and the filter $\psi$. In order to define $\mathcal{A}$ as such, we generalize the approach of \citet{woo2018cbam} such that: (1) equivariance to general symmetry groups is preserved and (2) the attention maps depend on the filter $\psi$ as well.

\begin{figure}
    \centering
    \includegraphics[width=0.7\textwidth]{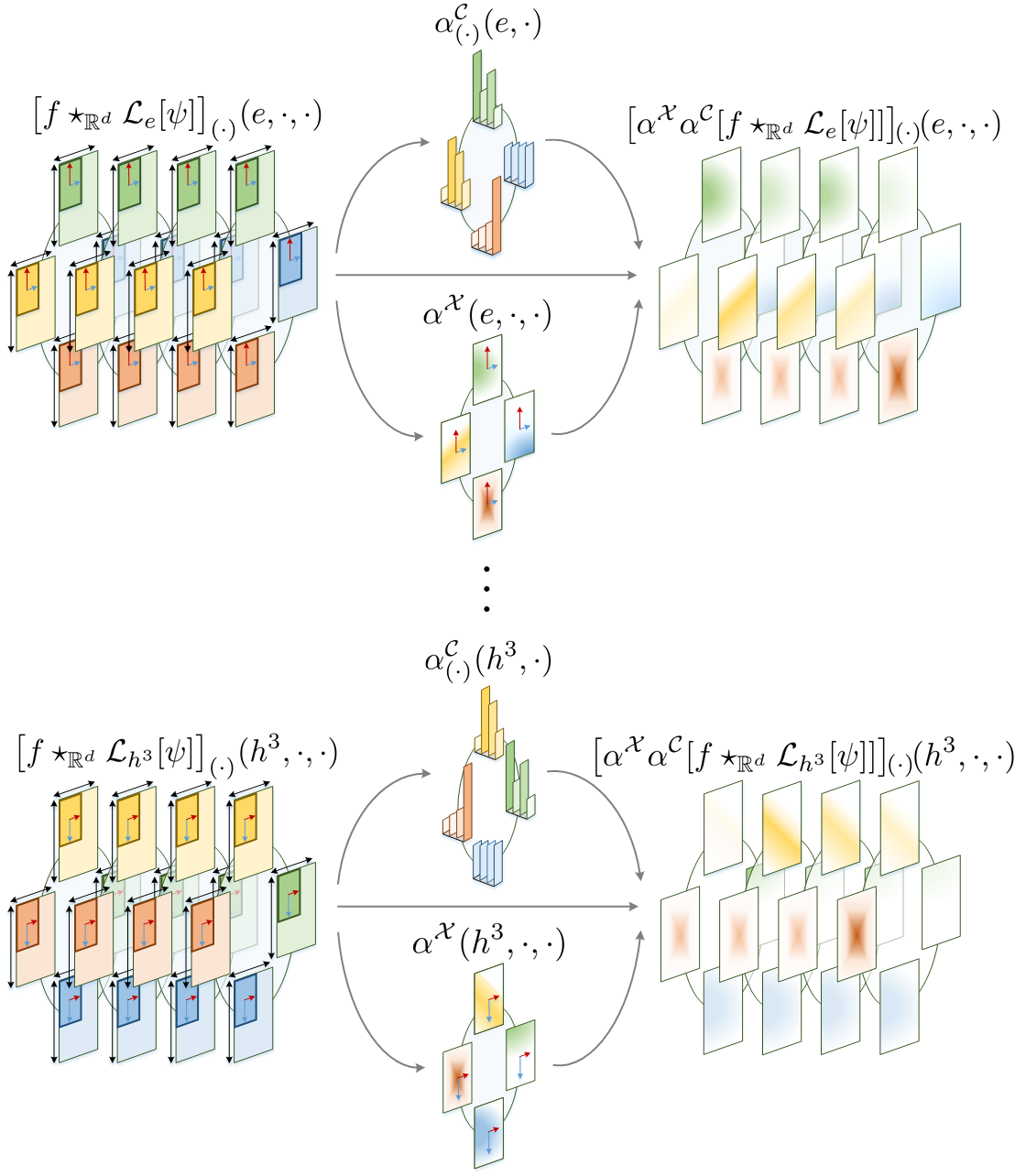}
    \vskip -4mm
    \caption{Attentive group convolution on the roto-translation group $\mathrm{SE(2)}$. In contrast to group convolutions (Fig. \ref{fig:6_affine_conv}, Eq. \ref{eq:6_gconv_groupstruct_smallconvs}), attentive group convolutions utilize channel $\alpha^{\mathcal{C}}$ and spatial $\alpha^{\mathcal{X}}$ attention to modulate the intermediary convolutional responses $[f \star_{\mathbb{R}^{2}} \mathcal{L}_{h}[\psi]]$ before pooling over the $\tilde{c}$ and $\tilde{h}$ axes.}
    \label{fig:6_att_affine_conv}
\end{figure}

Let $\phi^{\mathcal{C}}: \tilde{f} \mapsto s^{\mathcal{C}} = \{s^{\mathcal{C}}_{\text{avg}}, s^{\mathcal{C}}_{\text{max}}\}$, $s^{\mathcal{C}}_{i}: H \times H \rightarrow \mathbb{R}^{N_{\tilde{c}}}$ and $\phi^{\mathcal{X}}: \tilde{f} \mapsto s^{\mathcal{X}} = \{s^{\mathcal{X}}_{\text{avg}}, s^{\mathcal{X}}_{\text{max}}\}$, $s^{\mathcal{X}}_{i}: G \times G \rightarrow \mathbb{R}$ be functions that generate channel ($s^{\mathcal{C}}$) and spatial statistics ($s^{\mathcal{X}}$), respectively, from an intermediary vector valued signal $\tilde{f}: G \times G \rightarrow \mathbb{R}^{N_{\tilde{c}}}$ containing information both from the input and output spaces. Analogously to \citet{woo2018cbam}, we compute spatial and channel statistics to reduce the dimensionality of the input. However, in contrast to them, we compute these statistics from intermediary convolutional maps $\tilde{f}$ rather than from the input signal $f$ directly.\footnote{This is why the statistics $s^{\mathcal{C}}_{i}$, $s^{\mathcal{X}}_{i}$ receive tuples $(h, \tilde{h})$, $(g, \tilde{g})$, respectively, as input, as opposed to single argument inputs which often emerge in several prior works on visual attention.} As a result, we take the influence of the filter $\psi$ into account during the computation of the attention maps. Following the simplifications proposed in Sec.~\ref{sec:6_equiv_attcoefs} for $\alpha^{\mathcal{X}}$, we can further reduce $s^{\mathcal{X}}_{i}$ and $\tilde{f}$ to functions of the form $s^{\mathcal{X}}_{i}: G \times H \rightarrow \mathbb{R}$ and $\tilde{f}: G \times H \rightarrow \mathbb{R}^{N_{\tilde{c}}}$, respectively. Consequently, we define:
\begin{equation}\label{eq:6_def_ftilde}
    \tilde{f} = \{\tilde{f}_{\tilde{c}}\}_{\tilde{c}=1}^{N_{\tilde{c}}}, \ \ \tilde{f}_{\tilde{c}}(x,h,\tilde{h}) := \big[f_{\tilde{c}} \star_{\mathbb{R}^{d}} \mathcal{L}_{h}[\psi_{\tilde{c}}]\big](x, \tilde{h}),
\end{equation} 
which is the intermediary result of the convolution between the input $f$ and the $h$-transformation of the filter $\psi$, $\mathcal{L}_{h}[\psi]$ before pooling over $\tilde{c}$ and $\tilde{h}$ (Fig. \ref{fig:6_att_affine_conv}, Eq. \ref{eq:6_gconv_groupstruct_smallconvs}).

\textbf{Channel Attention.} Let $\varphi^{\mathcal{C}}: s^{\mathcal{C}} \mapsto \alpha^{\mathcal{C}}$ be a function that generates a channel attention map $\alpha^{\mathcal{C}}: H \times H \rightarrow [0,1]^{N_{\tilde{c}}}$ from a vector of channel statistics $s^{\mathcal{C}}: H \times H \rightarrow \mathbb{R}^{N_{\tilde{c}}}$ of the intermediate representation $\tilde{f}$. Our channel attention computation is analogous to that of \citet{woo2018cbam} based on two fully connected layers. However, in our case, each linear layer is parametrized by a \textit{matrix-valued kernel} $\mathbf{W}_i:H \rightarrow \mathbb{R}^{N_{out} \times N_{in}}$, which we shift via left-regular representations $\mathcal{L}_h\left[\mathbf{W}_i\right](\tilde{h}) {=} \mathbf{W}_i(h^{-1} \tilde{h})$ to guarantee equivariance (Thm.~\ref{thm1}):
\begin{align}
    \alpha^{\mathcal{C}}(h,\tilde{h})&=\varphi^{\mathcal{C}}\left[s^{\mathcal{C}}\right](h,\tilde{h}) \nonumber \\
    &= \sigma\Big(\big[
    \mathbf{W}_{2}(h^{-1}\tilde{h})\cdot[\mathbf{W}_{1}(h^{-1}\tilde{h})\cdot s^{\mathcal{C}}_{ \text{avg}}(h,\tilde{h})]^{+}
    \big] + \big[
    \mathbf{W}_{2}(h^{-1}\tilde{h})\cdot[\mathbf{W}_{1}(h^{-1}\tilde{h})\cdot s^{\mathcal{C}}_{ \text{max}}(h,\tilde{h})]^{+}
    \big] \Big) \label{eq:6_compute_ak_2}
\end{align}
with $[\cdot]^{+}$ the ReLU function, $\sigma$ the sigmoid function, $r$ a reduction ratio and $\mathbf{W}_{1}: H \rightarrow \mathbb{R}^{\frac{N_{\tilde{c}}}{r} \times N_{\tilde{c}}}$, $\mathbf{W}_{2}: H \rightarrow \mathbb{R}^{N_{\tilde{c}} \times \frac{N_{\tilde{c}}}{r}}$ filters defined on $H$.

\textbf{Spatial Attention.} Let $\varphi^{\mathcal{X}}: s^{\mathcal{X}} \mapsto \alpha^{\mathcal{X}}$ be a function that generates a spatial attention map $\alpha^{\mathcal{X}}: G \times H \rightarrow [0,1]$ from channel statistics $s^{\mathcal{X}}:G \times H \rightarrow \mathbb{R}^2$, in which per input $\tilde{h} \in H$ and output $g \in G$, the mean and max value is taken over the channel axis. Similarly to \citet{woo2018cbam}, spatial attention $\alpha^{\mathcal{X}}$ is then defined as:
\begin{align}
   \alpha^{\mathcal{X}}(x, h, \tilde{h}) &= \varphi^{\mathcal{X}}(s^{\mathcal{X}})(x, h, \tilde{h})\nonumber \\ \label{eq:6_compute_ax} &= \sigma\left(\big[s^{\mathcal{X}} \star_{\mathbb{R}^{d}} \mathcal{L}_{h}[ \psi^{\mathcal{X}}]\big]\right)(x, \tilde{h})
\end{align}
with $\psi^{\mathcal{X}}: G \rightarrow \mathbb{R}^{2}$ a group convolutional filter.

\begin{figure}
    \centering
    \includegraphics[width=0.7\textwidth]{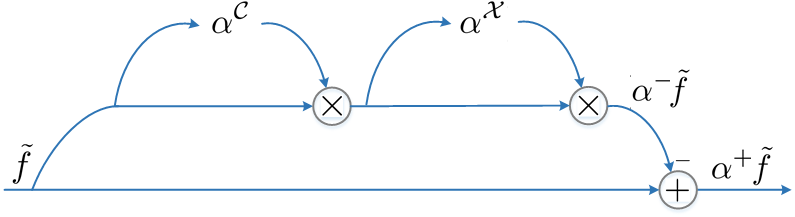}
    \vskip -2mm
    \caption{Sequential channel and spatial attention performed on the residual attention branch (Sec.~\ref{sec:6_res_att_branch}).}
    \label{fig:6_attention_branch}
\end{figure}

\textbf{Full Attention.} \citet{woo2018cbam} carried out extensive experiments to find the best performing configuration to combine channel and spatial attention maps for the $\mathbb{R}^{d}$ case, e.g., in parallel, serially starting with channel attention, serially starting with spatial attention. Based on their results we adopt their best performing configuration, i.e., \textit{serially starting with channel attention}, for the $G$ case (Fig. \ref{fig:6_attention_branch}). 

Recall that $\tilde{f}$ is the intermediary result from the convolution between the input $f$ and the $h$-transformation of the filter $\psi$ before pooling over $\tilde{c}$ and $\tilde{h}$. We perform attention on top of $\tilde{f}$ (Fig. \ref{fig:6_attention_branch}), where $\alpha^{\mathcal{C}}$ and $\alpha^{\mathcal{X}}$ are computed by Eqs. \ref{eq:6_compute_ak_2}, \ref{eq:6_compute_ax}, respectively. Resultantly, the attentive group convolution is computed as:
% \begin{align}
% &[f \star^{\alpha}_{G} \psi](x, h)\nonumber\\
% &= \sum_{\tilde{c}=1}^{N_{\tilde{c}}}\int_H \Big[\mathcal{L}_{h}[\alpha^{\mathcal{X}}]\mathcal{L}_{h}[\alpha^{\mathcal{C}}]\big[f_{k} \star_{\mathbb{R}^{d}} \mathcal{L}_{h}[\psi_{k}]\big]\Big](x,\tilde{h})\,d\tilde{h}
% \end{align}
\begin{equation}
[f \star^{\alpha}_{G} \psi](x, h)=
\sum_{\tilde{c}=1}^{N_{\tilde{c}}}\int_H \alpha^{\mathcal{X}}(x, h, \tilde{h}) \alpha_{\tilde{c}}^{\mathcal{C}}(h,\tilde{h})\tilde{f}(x,h,\tilde{h})\,{\rm d}\tilde{h} \label{eq:6_full_att}
\end{equation}

\vspace{-7mm}
\subsection{The Residual Attention Branch}\label{sec:6_res_att_branch}
Based on the findings of \citet{he2016deep}, several visual attention approaches propose to utilize residual blocks with direct connections during the course of attention to facilitate gradient flow \cite{hu2018squeeze, park2018bam, woo2018cbam, wang2018non, cao2019gcnet}. However, these approaches calculate the final attention map $\alpha^{+}$ as the sum of the direct connection $\boldsymbol{1}$ and the attention map obtained from the attention branch $\alpha$, i.e., $\alpha^{+}=\boldsymbol{1} + \alpha$. Consequently, the obtained attention map $\alpha^{+}: \mathbb{R}^{2} \rightarrow [1,2]^{N_{c}}$ is \textit{restricted} to the interval $[1, 2]$ and the network loses its ability to suppress input components. Inspired by the aforementioned works, we propose to calculate attention in what we call a \textit{residual attention branch} (Fig. \ref{fig:6_attention_branch}). Specifically, we utilize the attention branch to calculate a \textit{residual attention map} defined as $ \alpha^{-} = ( \boldsymbol{1} - \alpha^{+})$; $\alpha^{-}: G \times G \rightarrow [0,1]$. Next, we subtract the residual attention map $\alpha^{-}$ from the direct connection $\boldsymbol{1}$ to obtain the resultant attention map $\alpha^{+}$, i.e., $\alpha^{+} = \boldsymbol{1} - \alpha^{-}$. As a result, we are able to produce attention maps $\alpha^{+}$ that span the $[0,1]$ interval while preserving the benefits of the direct connections of \citet{he2016deep}.
% =====subSECTION=====
%\vspace{-8mm}
\vspace{-7mm}
\subsection{Attentive group convolution as a sequence of group convolutions and point-wise non-linearities}
\label{sec:6_att_sequence}

CNNs are usually organized in layers and hence, the input $f$ is usually convolved in parallel with a set of $N_o$ filters $\{\psi_{o}\}_{o=1}^{N_o}$. As outlined in the previous section, this implies that the attention maps can change as a function of the current filter $\psi_{o}$. One assumption broadly utilized in visual attention is that these maps do not depend on the filters $\{\psi_{o}\}_{o=1}^{N_o}$, and, hence, that $\alpha$ is a sole function of the input signal $f$ \cite{hu2018squeeze, park2018bam, woo2018cbam, diaconu2019affine, Romero2020Co-Attentive}. Consequently, the attention coefficients $\alpha$ are reduced from a function $\alpha: G \times G \rightarrow [0,1]^{N_{\tilde{c}}}$ (c.f.,~Eq.~\ref{eq:6_att_g_conv}) to a function $\alpha: G \rightarrow [0,1]^{N_{\tilde{c}}}$. In other words, attention becomes only dependent on $g$ (see Eqs. \ref{eq:6_compute_ak_2}-\ref{eq:6_full_att}) and thus, the generation of the attention maps $\alpha^{\mathcal{C}}$, $\alpha^{\mathcal{X}}$ can be shifted to the input feature map $f$. 
Resultantly, the attentive group convolution is reduced to a sequence of conventional group convolutions and point-wise non-linearities (Thm.~\ref{thm1}), which further reduces the computational cost of attention:
\begin{equation}
\label{eq:6_sec_attention}
    [f \star^{\alpha}_{G} \psi] = [f^{\alpha} \star_{G} \psi] = [(\alpha^{\mathcal{X}}\alpha^{\mathcal{C}}f) \star_{G} \psi]
\end{equation}
% =====SECTION=====
\vspace{-7mm}
\section{Experiments}
We validate our approach by exploring the effects of using attentive group convolutions in contrast to conventional ones. We compare the conventional group equivariant networks $p4$- and $p4m$-CNNs of \citet{cohen2016group} on the rotated MNIST and CIFAR-10 datasets with their corresponding attentive counterparts: $\alpha$-$p4$-CNNs and $\alpha$-$p4m$-CNNs, respectively; and the $p4$- and $p4m$-DenseNets of \citet{veeling2018rotation} on the PCam dataset with their corresponding attentive counterparts: $\alpha$-$p4$-DenseNet and $\alpha$-$p4m$- CNNs and DenseNets, respectively. Additionally, we explore the effects of only applying channel attention (e.g., $\alpha_{\text{CH}}$-$p4$-CNNs), spatial attention (e.g., $\alpha_{\text{SP}} $-$p4$-CNNs) and applying attention directly on the input (e.g., $\alpha_{F}$-$p4$-CNNs).%\footnote{Our code is publicly available at: \url{ https://github.com/dwromero/att_gconvs}}

We notice that the network architectures in \citet{cohen2016group} and \citet{Romero2020Co-Attentive} used for the CIFAR-10 experiments are equivariant only approximately. This results from using odd-sized convolutional kernels with stride $\geq$ 1 on even-sized feature maps (see Appx.~\ref{sec:6_approx_equiv} for a complete discussion). Since this effect distorts the equivariance property of our equivariant attention maps, i.e., they also become equivariant only approximately (Figs.~\ref{fig:6_bad_examples}, \ref{fig:6_good_examples}), this issue must be fixed. We achieve this by replacing strided convolutions in such regimes by conventional convolutions followed by a max-pooling layer.
%We exclude \citet{Romero2020Co-Attentive} from the experiments since it is not applicable to general groups and several of its experiments are carried out in an approximately equivariant regime (see Appx. \ref{sec:6_approx_equiv}). Furthermore, we explore the effects of only applying channel attention (e.g., $\alpha_{\text{CH}}$-$p4$-CNNs), spatial attention (e.g., $\alpha_{\text{SP}} $-$p4$-CNNs) and applying attention directly on the input (e.g., $\alpha_{F}$-$p4$-CNNs). 

For all our experiments we replicate as close as possible the training and evaluation strategies of the corresponding baselines, replace approximately equivariant networks by exact equivariant ones, and initialize any additional parameter in the same way as the corresponding baseline. Extended implementation details are provided in Appx.~\ref{appx:6_extended_details}.

\vspace{-7mm}
\subsection{rot-MNIST}
The rotated MNIST dataset \cite{larochelle2007empirical} contains 62$k$ gray-scale 28x28 handwritten digits uniformly rotated for $[0, 2\pi)$. The dataset is split into training, validation and test sets of 10$k$, 2$k$ and 50$k$ images respectively. We compare $p4$-CNNs with all the corresponding attention variants previously mentioned. For our attention models, we utilize a filter size of $7$ and a reduction ratio $r$ of $2$ on the attention branch. Since attentive group convolutions impose the learning of additional parameters, we also instantiate bigger $p4$-CNNs by increasing the number of channels uniformly at every layer to roughly match the number of parameters of the attentive versions. Furthermore, we compare our results with comparative attentive versions as defined in \citet{Romero2020Co-Attentive} ($\alpha_{\text{RH}}$), which perform attention exclusively over the axis of rotations. Our results show that (1) attentive versions consistently outperform non-attentive ones, and that (2) using attention over the whole group leads further improves accuracy (Tab.~\ref{tab:6_rot_mnist}).

\vspace{-7mm}
\subsection{CIFAR-10} 
\begin{figure}
    \centering
    \begin{subfigure}{0.24\textwidth}
        \includegraphics[width=\textwidth]{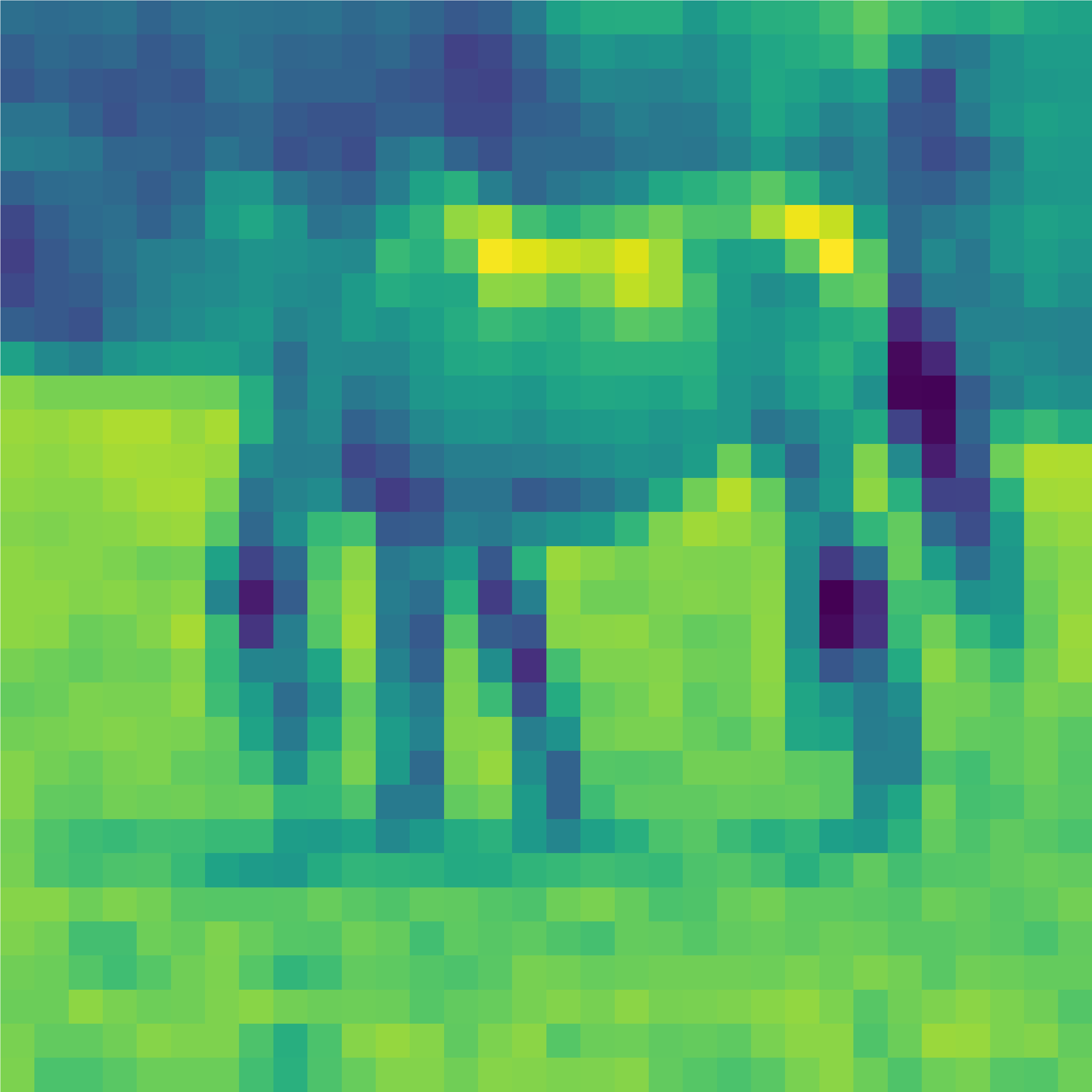}
    \end{subfigure}
    \begin{subfigure}{0.24\textwidth}
        \includegraphics[width=\textwidth]{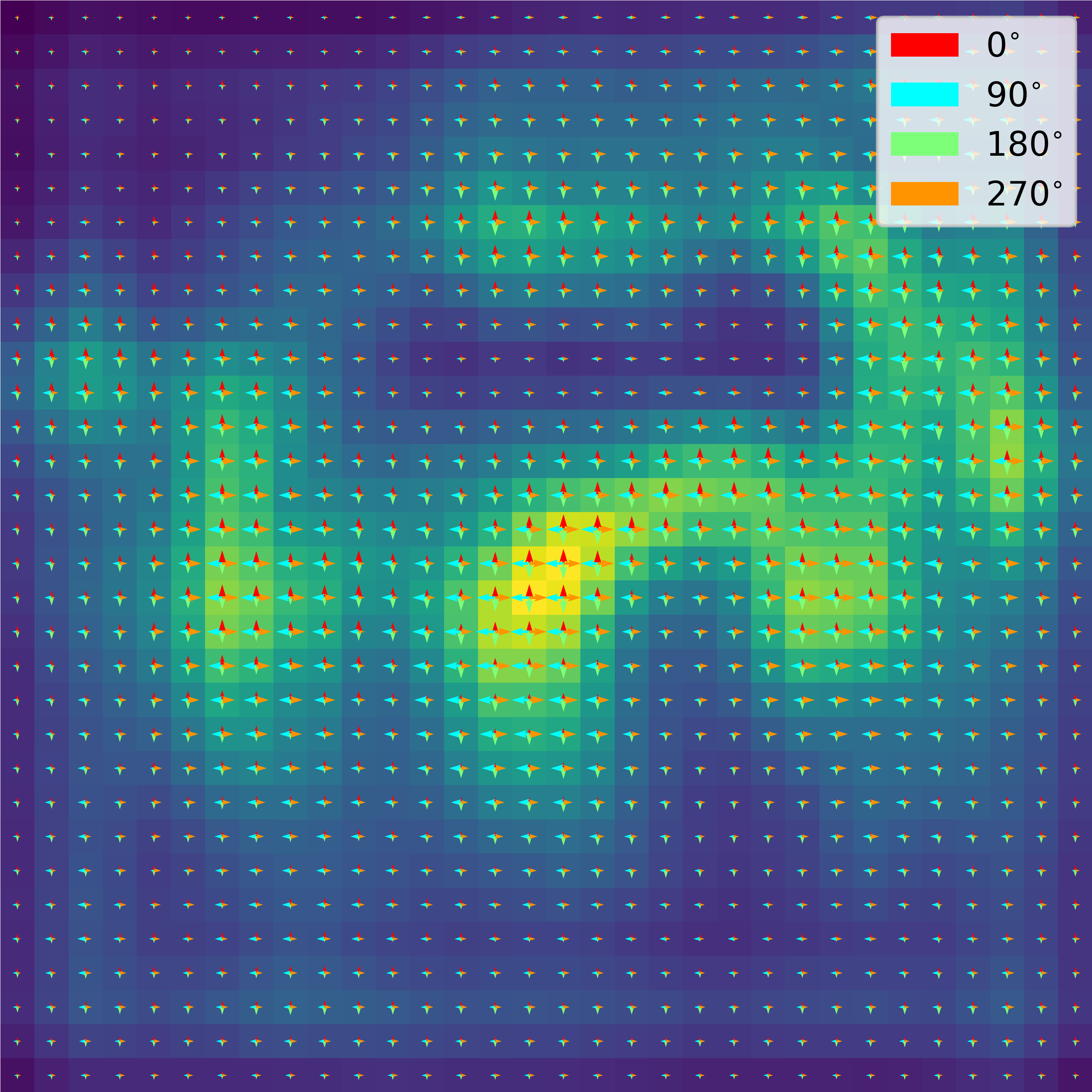}
    \end{subfigure}
    \hfill
    \begin{subfigure}{0.24\textwidth}
        \includegraphics[width=\textwidth]{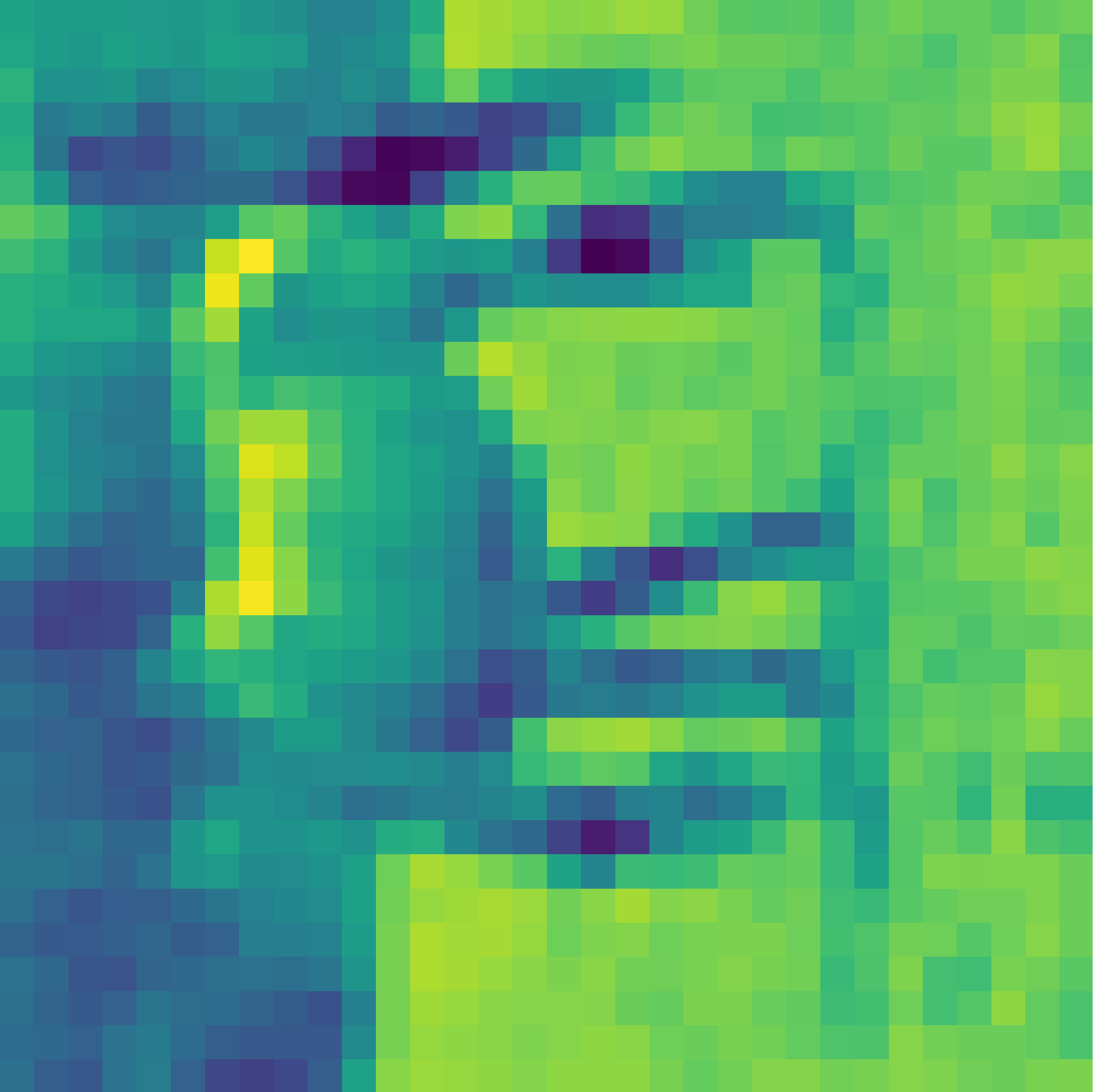}
    \end{subfigure}
    \begin{subfigure}{0.24\textwidth}
        \includegraphics[width=\textwidth]{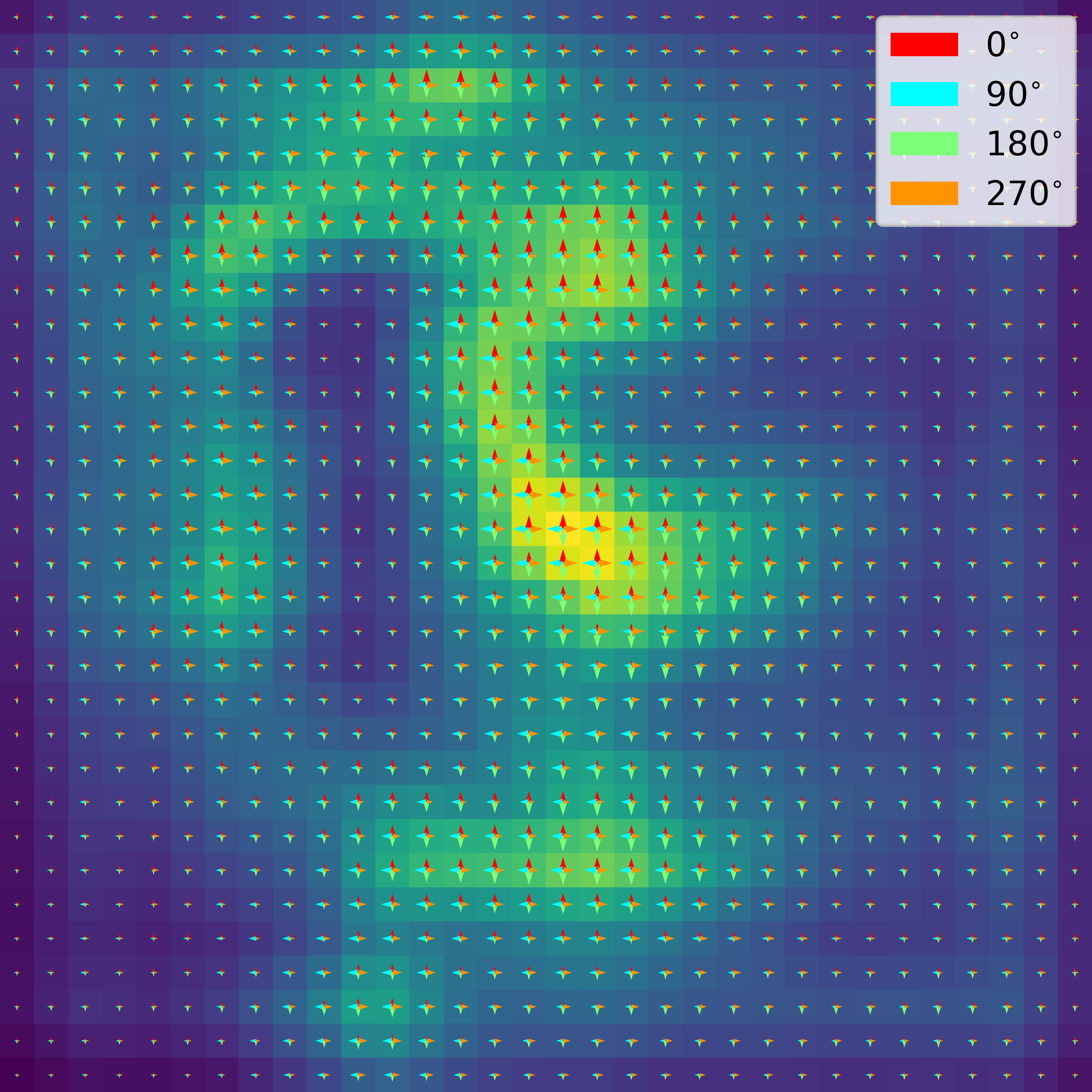}
    \end{subfigure}
    \vskip -2mm
    \caption{Equivariant attention maps on the roto-translation group $\mathrm{SE(2)}$. The predicted attention maps behave equivariantly under group symmetries. The arrows depict the strength of the filter responses at the corresponding orientations.}\label{fig:6_examples}
\end{figure}

The CIFAR-10 dataset \cite{krizhevsky2009learning} consists of 60$k$ real-world 32x32 RGB images uniformly drawn from 10 classes. The dataset is split into training, validation and test sets of $40k$, $10k$ and $10k$ images, respectively. We compare the $p4$ and $p4m$ versions of the All-CNN \cite{springenberg2014striving} and the Resnet44 \cite{he2016deep} in \citet{cohen2016group} with attentive variations. For all our attention models, we utilize a filter size of $7$ and a reduction ratio $r$ of $16$ on the attention branch. Unfortunately, attentive group convolutions impose an unfeasible increment on the memory requirements for this dataset.\footnote{the $\alpha$-$p4$ All-CNN requires approx. 72GB of CUDA memory, as opposed to 5GBs for the $p4$-All-CNN. This is due to the storage of the intermediary convolution responses required for the calculation of the attention weights (Eqs.~\ref{eq:6_compute_ak_2}- \ref{eq:6_full_att})} Resultantly, we are only able to compare the $\alpha_{\text{F}}$ variations of the corresponding networks. Our results show that attentive $\alpha_{\text{F}}$ networks consistently outperform non-attentive ones (Tab. \ref{tab:6_cifar}). Moreover, we demonstrate that our proposed networks focus on relevant parts of the input and that the predicted attention maps behave equivariantly for group symmetries (Figs.~\ref{fig:6_examples}, \ref{fig:6_good_examples}).

% ==== Figure =====
%  -=== Table
\begin{table}[t]
\centering
\caption{Test error rates on rot-MNIST (mean and std over 5 random seeds).}
\label{tab:6_rot_mnist}
\begin{center}
\vskip -3mm
\begin{small}
\begin{sc}
\scalebox{0.85}{
\begin{tabular}{rcc}
\toprule
Network & Test Error (\%) & Param.\\
\toprule
$p4$-CNN & 2.048 $\pm$ 0.045& 24.61k\\
$\alpha_{\text{RH}}$-$p4$-CNN & 1.980 $\pm$ 0.032 & 24.85k\\
\midrule
big$_{19}$-$p4$-CNN & 1.796 $\pm$ 0.035 & 77.54k\\
$\alpha$-$p4$-CNN & \textbf{1.696 $\pm$ 0.021} &  73.13k\\
\midrule
big$_{15}$-$p4$-CNN & 1.848 $\pm$ 0.019 & 50.42k\\
$\alpha_{\text{ch}}$-$p4$-CNN & \textbf{1.825 $\pm$ 0.048}& 48.63k \\
$\alpha_{\text{sp}}$-$p4$-CNN & \textbf{1.761 $\pm$ 0.027} & 49.11k \\
\midrule
big$_{11}$-$p4$-CNN & 1.996 $\pm$ 0.083  & 29.05k \\
$\alpha_{\text{f}}$-$p4$-CNN & \textbf{1.795 $\pm$ 0.028} & 29.46k \\
\bottomrule
\end{tabular}}
\end{sc}
\end{small}
\end{center}
\vspace{-2mm}
\end{table}
\begin{table}[t!]
\caption{Test error rates on CIFAR10 and augmented CIFAR10+.}
\vskip -3mm
\label{tab:6_cifar}
\begin{center}
\begin{small}
\begin{sc}
\scalebox{0.85}{
\begin{tabular}{r|cccc}
\toprule
Network & Type & CIFAR10 & CIFAR10+ & Param.\\
\toprule
\multirow{4}{*}{All-CNN} & $p4$ & 9.32 & 8.91 & 1.37M\\
& $\alpha_{\text{f}}$-$p4$ &\textbf{ 8.8} & \textbf{7.05} & 1.40M\\
& $p4m$ & 7.61 & 7.48 & 1.22M\\
& $\alpha_{\text{f}}$-$p4m$ & \textbf{6.93} & \textbf{6.53} & 1.25M\\
\midrule
\multirow{2}{*}{ResNet44} & $p4m$ & 15.72& 15.4 & 2.62M\\
& $\alpha_{\text{f}}$-$p4m$ &\textbf{ 10.82 }& \textbf{10.12} & 2.70M\\
\bottomrule
\end{tabular}}
\end{sc}
\end{small}
\end{center}
\end{table}
\vspace{-7mm}
\subsection{PCam} 
\begin{figure}[t]
    \centering
    \begin{subfigure}{0.24\textwidth}
        \includegraphics[width=\textwidth]{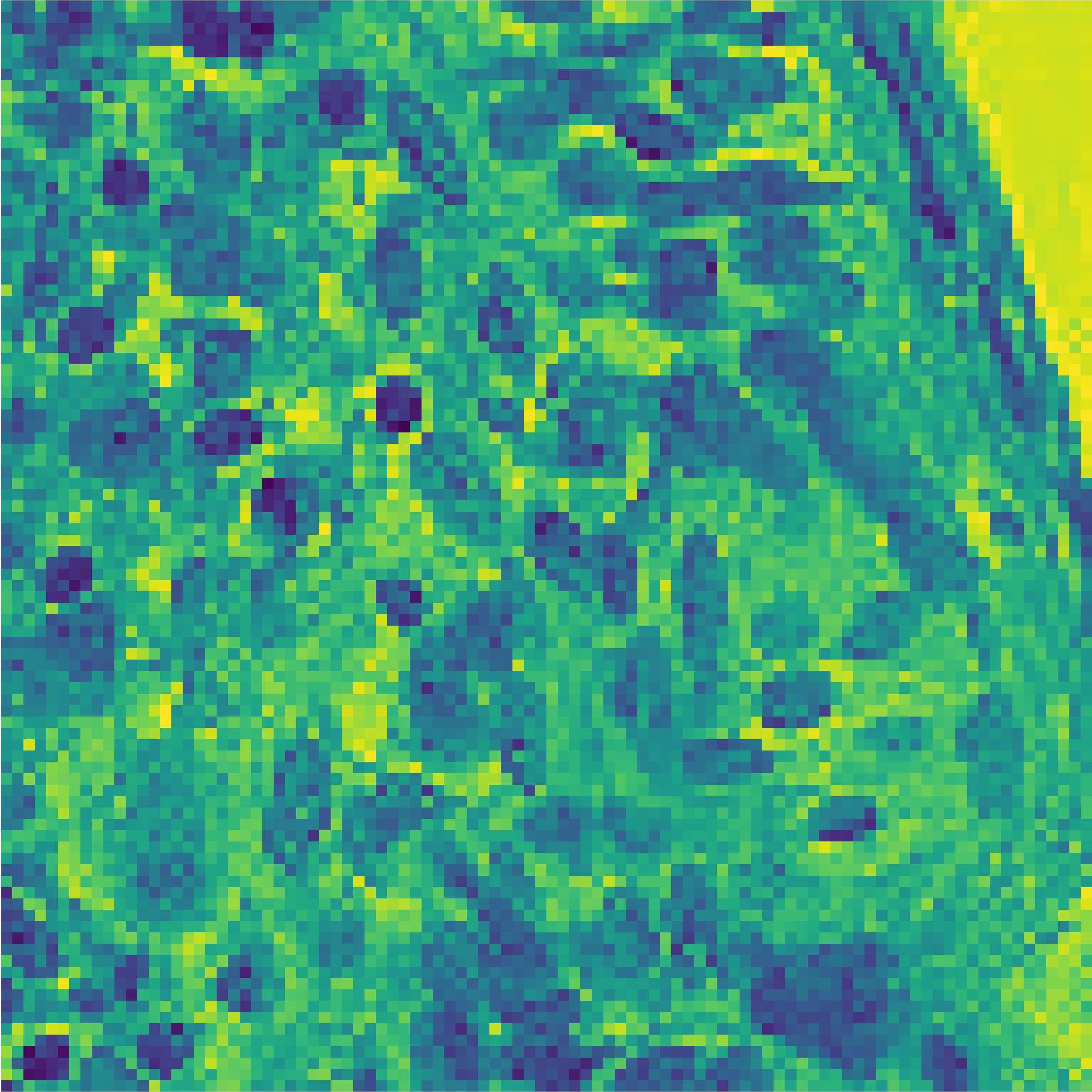}
    \end{subfigure}
    \begin{subfigure}{0.24\textwidth}
        \includegraphics[width=\textwidth]{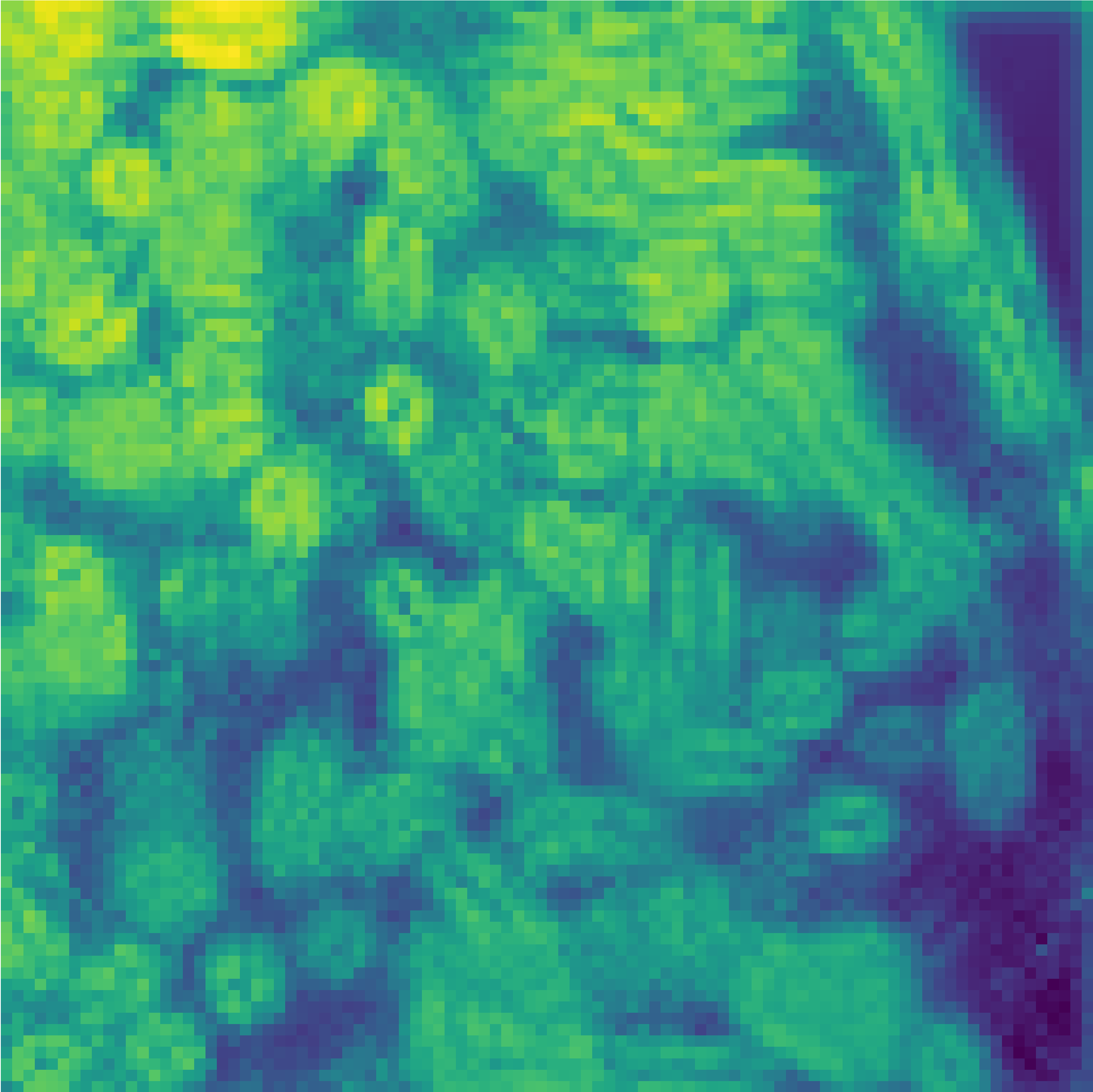}
    \end{subfigure}
    \hfill
    \begin{subfigure}{0.24\textwidth}
        \includegraphics[width=\textwidth]{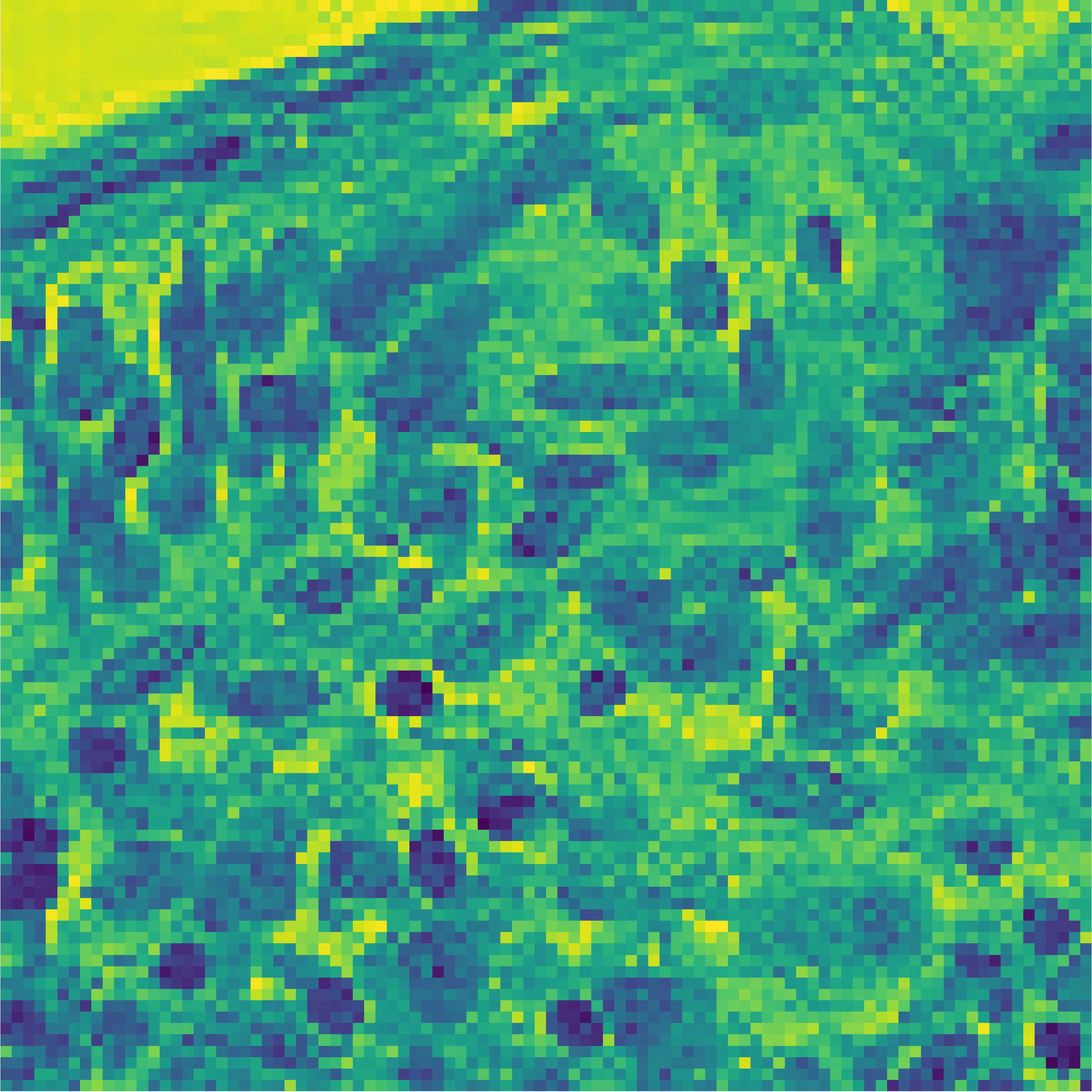}
    \end{subfigure}
    \begin{subfigure}{0.24\textwidth}
        \includegraphics[width=\textwidth]{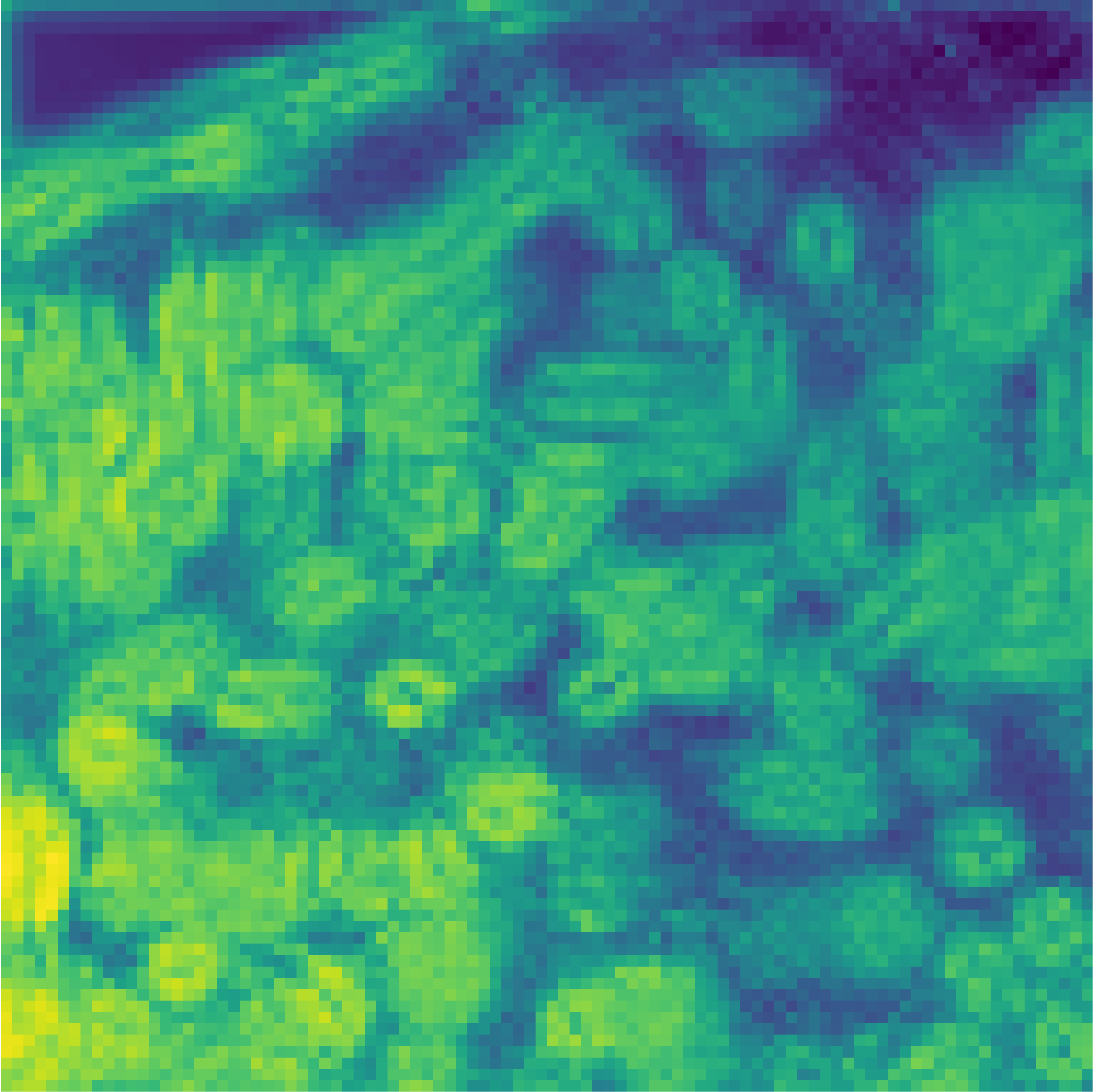}
    \end{subfigure}
    \vskip -2mm
    \caption{Equivariant attention maps on the PCam dataset. The predicted attention maps behave equivariantly for group symmetries. In addition, the network learns to focus on the nuclei of the cells and to remove background elements during training.}\label{fig:6_pcam_examples}
\end{figure}

The PatchCamelyon dataset \cite{veeling2018rotation} consists of 327$k$ 96x96 RGB image patches of tumorous / non-tumorous breast tissues extracted from the Camelyon16 dataset \cite{bejnordi2017diagnostic}, where each patch was labelled as tumorous if the central region (32x32) contained at least one tumour pixel as given by the original annotation in \citet{bejnordi2017diagnostic}. We compare the $p4$ and $p4m$ versions of the DenseNet \cite{huang2017densely} in \citet{veeling2018rotation} with attentive variants. For all our attention models, we utilize a filter size of $7$ and a reduction ratio $r$ of 16 on the attention branch. Similarly to the CIFAR-10 case, we restrict our experiments to $\alpha_{\text{F}}$ attentive networks due to computational constraints. Our results show that attentive $\alpha_{\text{F}}$ consistently outperform non-attentive ones (Tab.~\ref{tab:6_pcam}). Interestingly, the $\alpha_{\text{F}}$-$p4$-DenseNet is already able to outperform the $p4m$-DenseNet without attention. Furthermore, our equivariant attention maps show that the network learns to focus on the nuclei of the cells and to remove the background in a group equivariant fashion (Fig.~\ref{fig:6_pcam_examples}).
\begin{table}[t]
\caption{Test error rates on PCam.}
\vskip -3mm
\label{tab:6_pcam}
\begin{center}
\begin{small}
\begin{sc}
\scalebox{0.85}{
\begin{tabular}{r|cccc}
\toprule
Network & Type & Test Error (\%) & Param.\\
\toprule
\multirow{5}{*}{DenseNet}& $\mathbb{Z}^{2}$ & 15.93 &  130.60k\\
& $p4$ & 12.45 &  129.65k\\
& $\alpha_{\text{f}}$-$p4$ &\textbf{11.34} & 140.45k\\
& $p4m$ & 11.64  & 124.21k\\
& $\alpha_{\text{f}}$-$p4m$ & \textbf{10.88} &  141.22k\\
\bottomrule
\end{tabular}}
\end{sc}
\end{small}
\end{center}
\end{table}

% =====SECTION=====
\vspace{-7mm}
\section{Discussion and future work}
Our results show that attentive group convolutions can be utilized as a drop-in replacement for standard and group equivariant convolutions that simultaneously facilitates the interpretability of the network decisions. Similarly to convolutional and group convolutional networks, attentive group convolutional networks also benefit of data augmentation. Interestingly, however, we also see that including additional symmetries reduces the effect of augmentations given by group elements. This finding supports the intuition that symmetry variants of the same concept are learned independently for non-equivariant networks (see Fig. 2 in \cite{krizhevsky2012imagenet}). The main shortcoming of our approach is its computational burden. As a result, the application of $\alpha$-networks is computationally unfeasible for networks with several layers or channels. We believe, however, by extrapolation of our results on rot-MNIST, that further performance improvements are to be expected for $\alpha$ variations, should hardware requirements suffice. 

Group convolutional networks have recently been proven very successful in medical imaging applications \cite{bekkers2018roto, winkels20183d, lafarge2020rototranslation}. Since explainability plays a crucial role here, we believe that our attentive maps could be of high relevance to aid the explainability of the network decisions. Moreover, since our attention maps are guaranteed to be equivariant to transformations in the considered group, it is ensured that the predicted attention maps will be consistent across group symmetries. We believe this to be of crucial importance for rotation invariant tasks. Illustratively, in contrast to vanilla attentive CNNs, a malignant tissue will be ensured to generate consistent attention maps regardless of the orientation at which it has been provided to the network.

In future work, we want to explore ways to reduce the computational cost of full attention networks. If successful, we consider feasible to obtain a direct performance boost over our CIFAR-10 and PCam experimental results, without extensive additional memory requirements. Furthermore, we want to extend our work to symmetry groups defined on 3D. By doing so, we expect the range of possible applications of our work to reach several other important applications such as 3D medical imaging applications like CT-scans and other voxel-based representations.

% =====SECTION=====
\vspace{-7mm}
\section{Conclusion}
We introduced attentive group convolutions, a generalization of the group convolution in which attention is utilized to explicitly highlight meaningful relationships among symmetries. We provided a general mathematical framework for group equivariant visual attention and indicated that prior work on visual attention can be perfectly described as special cases of the attentive group convolution. Our experimental results indicate that attentive group equivariant networks consistently outperform conventional group equivariant ones. Furthermore, attentive group equivariant networks provide equivariant attention maps that behave predictively for symmetries of the group and with which the learned concepts can be visualized.

%% file: chapters/7_gselfatt.tex
% path to figures directory
\graphicspath{{figures/7-gselfatt/}}

%=========================================================================

% \begin{savequote}[75mm]
% Nulla facilisi. In vel sem. Morbi id urna in diam dignissim feugiat. Proin molestie tortor eu velit. Aliquam erat volutpat. Nullam ultrices, diam tempus vulputate egestas, eros pede varius leo.
% \qauthor{Quoteauthor Lastname}
% \end{savequote}

\chapter{Group Equivariant Stand-Alone Self-Attention}\label{chapter:g_selfatt}

\begin{flushright}
\textit{Based on the paper:}\break
\textit{Group Equivariant Stand-Alone Self-Attention \citep{romero2021group}}
\end{flushright}
%=========================================================================
\vspace{-7mm}
\section{Introduction}
Recent advances in Natural Language Processing have been largely attributed to the rise of the \textit{Transformer} \citep{vaswani2017attention}.
Its key difference with previous methods, e.g., recurrent neural networks, convolutional neural networks (CNNs), is its ability to query information from all the input words simultaneously. This is achieved via the \emph{self-attention operation} \citep{bahdanau2014, cheng2016long}, which computes the similarity between representations of words in the sequence in the form of \emph{attention scores}. Next, the representation of each word is updated based on the words with the highest attention scores.
Inspired by the capacity of transformers to learn meaningful inter-word dependencies, researchers have started applying self-attention in vision tasks. It was first adopted into CNNs by channel-wise attention \citep{hu2018squeeze} and non-local spatial modeling \citep{wang2018non}.
More recently, it has been proposed to replace CNNs with Transformers either partially \citep{bello2019attention} or entirely \citep{ramachandran2019stand}.
Contrary to discrete convolutional kernels, weights in self-attention are not tied to particular positions (Fig.~\ref{fig:7_comparison_att_conv}), yet self-attention layers can express any convolutional layer \citep{Cordonnier2020On}.
This flexibility allows leveraging long-range dependencies under a fixed parameter budget.

% \textit{Group equivariant CNNs} \citep{GCNN}, on the other hand, provide better generalization and sample-efficiency by hard-baking symmetries encountered in real-data into the structure of the model. A known example of such an architecture is the CNN, which can learn representations such that if a pattern is translated, its numerical descriptors are also translated, but not modified. This property, called \textit{translation equivariance}, has been further extended to larger symmetry groups (\S\ref{sec:7_related_work}).
An orthogonal advancement to deep learning architectures is the incorporation of symmetries into the model itself. The seminal work by \citet{cohen2016group} provides a recipe to extend the \textit{translation equivariance} of CNNs to other symmetry groups to improve generalization and sample- efficiency further (see Sec.~\ref{sec:7_related_work}). \textit{Translation equivariance} is key to the success of CNNs. It describes the property that if a pattern is translated, its numerical descriptors are also translated, but not modified. 

In this work, we introduce \textit{group self-attention}, a self-attention formulation that grants equivariance to arbitrary %compact 
symmetry groups. This is achieved by defining positional encodings invariant to the action of the group considered. In addition to generalization and sample-efficiency improvements provided by group equivariance, group equivariant self-attention networks (\gsa) bring important benefits over group convolutional architectures: (\emph{i}) \emph{Parameter efficiency:} contrary to conventional discrete group convolutional kernels, where weights are tied to particular positions of neighborhoods on the group, group equivariant self-attention leverages long-range dependencies on group functions under a fixed parameter budget, yet it is able to express any group convolutional kernel. This allows for very expressive networks with low parameter count. (\emph{ii}) \emph{Steerability:} since the group acts directly on the positional encoding, \gsa\ are \textit{steerable} \citep{weiler2018learning} by nature. This allows us to go beyond group discretizations that live in the grid without introducing interpolation artifacts. 

% In this work, we take inspiration from self-attention and group equivariance to provide a fully self-attention formulation for group equivariance. This allows for the construction of neural architectures that bring important advantages upon both transformer and group equivariant convolutional architectures: (\emph{i}) Group equivariance provides generalization and sample-efficiency improvements over vanilla self-attention, and, (\emph{ii}) group equivariant self-attention allows leveraging long-range dependencies on group functions under a fixed parameter budget, while being able to express any group equivariant convolutional layer. The latter is an important advantage over group convolutional kernels, where weights are tied to particular positions of neighborhoods on the group, typically much larger than those of conventional convolutional kernels.

\textbf{Contributions:}
\begin{itemize}
    \item We provide an extensive analysis on the equivariance properties of self-attention.
    \item We provide a general formulation to impose group equivariance to self-attention.
    \item We provide instances of self-attention equivariant to several symmetry groups.
    \item Our results show consistent improvements of \gsa\ over non-equivariant ones.
\end{itemize}
\begin{figure}
    \centering
    \includegraphics[width=0.7\textwidth]{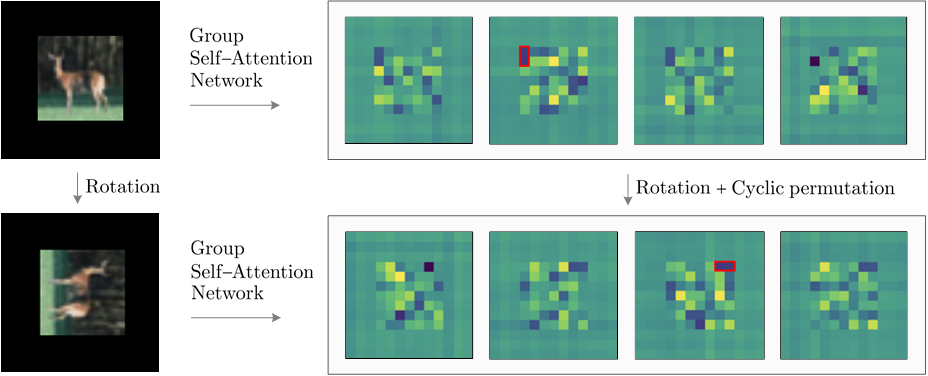}
        \caption{Behavior of feature representations in group self-attention networks. An input rotation induces a rotation plus a cyclic permutation to the intermediary feature representations of the network. Additional examples for all groups in this work and their usage are provided at \href{https://github.com/dwromero/g_selfatt/tree/master/demo}{\url{github.com/dwromero/g_selfatt/demo}}.}
        \label{fig:7_example_image}
\end{figure}

\vspace{-7mm}
\section{Related Work}
\label{sec:7_related_work}
\vspace{-1mm}
Several approaches exist which provide equivariance to various symmetry groups. The translation equivariance of CNNs has been extended to additional symmetries ranging from planar rotations \citep{dieleman2016exploiting, marcos2017rotation, worrall2017harmonic, weiler2018learning, DREN, cheng2018rotdcf, hoogeboom2018hexaconv, bekkers2018roto, veeling2018rotation, lenssen2018group, graham2020dense} to spherical rotations \citep{cohen2018spherical,cohen2019gauge, worrall2018cubenet, weiler20183d,  esteves2019cross, esteves2019equivariant, esteves2020spin}, scaling \citep{marcos2018scale, worrall2019deep, sosnovik2020scaleequivariant, romero2020wavelet} and more general symmetry groups \citep{cohen2016group, kondor2018generalization, tai2019equivariant, weiler2019general, cohen2019general, bekkers2020bspline, Venkataraman2020Building}. Importantly, all these approaches utilize discrete convolutional kernels, and thus, tie weights to particular positions in the neighborhood on which the kernels are defined. As group neighborhoods are (much) larger than conventional ones, the number of weights discrete group convolutional kernels require proportionally increases. This phenomenon is further exacerbated by \textit{attentive group equivariant networks} \citep{Romero2020Co-Attentive, romero2020attentive}. Since attention is used to leverage non-local information to aid local operations, non-local neighborhoods are required. However, as attention branches often rely on discrete convolutions, they effectively tie specific weights to particular positions on a large non-local neighborhood on the group. As a result, attention is bound to growth of the model size, and thus, to negative statistical efficiency. Differently, group self-attention is able to attend over arbitrarily large group neighborhoods under a fixed parameter budget. In addition, group self-attention is steerable by nature (Sec.~\ref{sec:7_steerability}) a property primarily exhibited by works carefully designed to that end.
% However, since attention branches utilize convolutional operations, particular weights are tied to specific positions in this non-local neighborhood. As a consequence, attention is bounded to an increment in the model size, and thus, to negative statistical efficiency.
% In fact, all the approaches mentioned above build upon discrete convolutional operations. %Hence, their weights are tied to particular positions.
% As their kernels are defined on neighborhoods on the group, typically larger than conventional ones, their kernels are bigger as well. Contrarily, our pure self-attention formulation allows attending over arbitrarily large neighborhoods under a fixed parameter budget.

Other way to detach weights from particular positions comes by parameterizing convolutional kernels as (constrained) neural networks \citep{thomas2018tensor, finzi2020generalizing}. Introduced to handle irregularly-sampled data, e.g., point-clouds, networks parameterizing convolutional kernels receive relative positions as input and output their values at those positions. In contrast, our mappings change as a function of the input content. Most relevant to our work are the $\mathrm{SE(3)}$ and \textit{Lie} Transformers \citep{fuchs2020se, hutchinson2021lietransformer}. However, we obtain group equivariance via a generalization of positional encodings, whereas \citet{hutchinson2021lietransformer} does so via operations on the Lie algebra and \citet{fuchs2020se} does so via irreducible representations. In addition, our work prioritizes applications on visual data and extensively analyses theoretical aspects and properties of group equivariant self-attention.
\vspace{-7mm}
\section{Stand-Alone Self-Attention}
\label{sec:7_self_att_formulation}

In this section, we recall the mathematical formulation of self-attention and emphasize the role of the positional encoding. Next, we introduce a functional formulation to self-attention which will allow us to analyze and generalize its equivariance properties.

\textbf{Definition.} Let $\X \in \sR^{N \times C_{\text{in}}}$ be an input matrix consisting of $N$ tokens of $C_{\text{in}}$ dimensions each.\footnote{We consequently consider an image as a set of $N$ discrete objects $\gi \in \{1, 2, ..., N\}$.} A self-attention layer maps an input matrix $\X \in \sR^{N \times C_{\text{in}}}$ to an output matrix $\Y \in \sR^{N \times C_{\text{out}}}$ as:
\begin{equation}\label{eq:7_self_att_matrix}
   \Y =
   \SA(\X) := \mathrm{softmax_{[\ ,\hspace{0.5mm}: \hspace{0.5mm}]}}(\A) \X \WV,
\end{equation}
with $\WV \in \sR^{C_{\text{in}} \times C_{\text{h}}}$ the \emph{value matrix}, $\A \in \sR^{N \times N}$ the \emph{attention scores matrix}, and $\mathrm{softmax_{[\ ,\hspace{0.5mm}: \hspace{0.5mm}]}}(\A)$ the \emph{attention probabilities}. The matrix $\A$ is computed as:
\begin{equation} \label{eq:7_att_scores}
    \A := \X \WQ (\X \WK)^\top,
\end{equation}
parameterized by \emph{query} and \emph{key matrices} $\WQ$, $\WK \in \sR^{C_{\text{in}} \times C_{\text{h}}}$. In practice, it has been found beneficial to apply multiple self-attention operations, also called \textit{heads}, in parallel, such that different heads are able to attend to different parts of the input. In this \textit{multi-head self-attention} formulation, the output of $H$ heads of output dimension $C_{h}$ are concatenated and projected to $C_{\text{out}}$ as:
\begin{equation}
    \label{eq:7_mhsa}
    \MHSA(\X) \coloneqq \cat_{h \in [H]}\big[\SA^{(h)}(\X)\big]\WO + \bv_{\text{out}},
\end{equation}
with a \textit{projection matrix} $\WO \in \sR^{H C_{h} \times C_{\text{out}}}$ and a bias term $\bv_{\text{out}} \in \sR^{C_{\text{out}}}$.
\vspace{-7mm}
\subsection{The role of the positional encoding}
Note that the self-attention operation defined in Eq.~\ref{eq:7_mhsa} is equivariant to permutations of the input rows of $\Xm$.
That is, a permutation of the rows of $\Xm$ will produce the same output $\Ym$ up to this permutation.
Hence, self-attention is blind to the order of its inputs, i.e., it is a set operation.
Illustratively, an input image is processed as a bag of pixels and the structural content is not considered.
To alleviate this limitation,
the input representations in self-attention are often enriched with a \emph{positional encoding} that provides positional information of the set elements.

\textbf{Absolute positional encoding.} \citet{vaswani2017attention} introduced a (learnable) positional encoding $\Pm \in \sR^{N \times C_{\text{in}}}$ for each input position which is added to the inputs when computing the attention scores:
\begin{equation}\label{eq:7_attention_scores_position_matrix_absolute}
    \Am := (\Xm + \Pm) \WQ ((\Xm + \Pm)\WK)^\top.
\end{equation}
More generally, $\Pm$ can be substituted by any function that returns a vector representation of the position and can be incorporated by means of addition or concatenation, e.g., \citet{Zhao_2020_CVPR}.
This positional encoding injects additional structural information about the tokens into the model, which makes it susceptible to changes in the token's positions.
Unfortunately, the model must learn to recognize similar patterns at every position independently as absolute positional encodings are \textit{unique} to each position.
This undesired data inefficiency is
addressed by \textit{relative positional encodings}.

\textbf{Relative positional encoding.} Introduced by \citet{shaw18relposenc}, relative encodings consider the \textit{relative distance} between the query token $\gi$ -- the token we compute the representation of --, and the key token $\gj$ -- the token we attend to --. The calculation of the attention scores (Eq.~\ref{eq:7_att_scores}) then becomes:
\begin{equation}\label{eq:7_relative_pos_attention_matrix}
    \Am^{\text{rel}}_{\gi, \gj}
    :=
    \Xm_{\gi} \WQ ((\Xm_{\gj} + \Pm_{x(\gj) - x(\gi)})\WK )^\top,
\end{equation}
where $\Pm_{x(\gj) - x(\gi)} \in \sR^{1 \times C_{\text{in}}}$ is a vector representation of the relative shift and $x(\gi)$ is the position of the token $\gi$ as defined in Sec.~\ref{sec:7_functional_attn}.
Consequently, similar patterns can be recognized at arbitrary positions, as relative query-key distances always remain equal.
%\jb{introduce function $x$ and space here.}
\vspace{-7mm}
\subsection{A functional formulation to self-attention}
\label{sec:7_functional_attn}

\textbf{Notation.} We denote by $[n]$ the set $\{1, 2, \ldots, n\}$. Given a set $\gS$ and a vector space $\gV$, $L_{\gV}(\gS)$ will denote the space of functions $\{f: \gS \rightarrow \gV \}$. Square brackets are used when functions are arguments, e.g., $\alpha[f]$ receives a function $f$ as argument.% and is indexed by $(\gi, \gj)$. 

Let $\gS = \{ \gi \}_{\gi = 1}^{N}$ be a set of $N$ elements. A matrix $\X \in \sR^{N \times C_{\text{in}}}$ can be interpreted as a vector-valued function $f: \gS \rightarrow \sR^{C_{\text{in}}}$ that maps element sets $\gi \in \gS$  to $C_{\text{in}}$-dimensional vectors: $f: \gi \mapsto f(\gi)$. Consequently, a matrix multiplication, $\X\W_{y}^\top$, of matrices $\X \in \sR^{N \times C_{\text{in}}}$ and $\W_{y} \in \sR^{C_{\text{out}} \times C_{\text{in}}}$ can be represented as a function $\varphi_{y}: L_{\gV_{C_\text{{in}}}}(\gS) \rightarrow  L_{\gV_{C_\text{{out}}}}(\gS)$,  $\varphi_{y}: f(\gi) \mapsto  \varphi_{y}(f(\gi)) $, parameterized by $\W_{y}$, between functional spaces  $L_{\gV_{C_\text{{in}}}}(\gS)=\{ f: \gS \rightarrow \sR^{C_{\text{in}}}\}$ and $L_{\gV_{C_\text{{out}}}}(\gS) = \{ f: \gS \rightarrow \sR^{C_{\text{out}}}\}$. Following this notation, we can represent the position-less attention scores calculation (Eq.~\ref{eq:7_att_scores}) as:
\begin{equation}
    \Am_{i,j} = \alpha[f](\gi, \gj) = \langle \varphi_{\text{qry}}(f(\gi)), \varphi_{\text{key}}(f(\gj)) \rangle.
\end{equation}
The function $\alpha[f]: \gS \times \gS \rightarrow \sR$ maps pairs of set elements $\gi, \gj \in \gS$ to the \textit{attention score} of $\gj$ relative to $\gi$. Therefore, the  self-attention (Eq.~\ref{eq:7_self_att_matrix}) can be written as:
\begin{align}
    \Ym_{i, : } &= \zeta[f](\gi) = \sum_{\gj \in \gS} \sigma_{\gj}(\alpha[f](\gi, \gj)) \varphi_{\text{val}}(f(\gj)) \nonumber \\
    &= \sum_{\gj \in \gS} \sigma\hspace{-0.5mm}_{\gj}\big(\langle \varphi_{\text{qry}}(f(\gi)), \varphi_{\text{key}}(f(\gj)) \rangle \big) \varphi_{\text{val}}(f(\gj)),
\end{align}
% \vspace{-0.5mm}
% \begin{align}
%     \Ym_{i, : } = \zeta[f](\gi) &= \sum_{\gj \in \gS} \sigma_{\gj}(\alpha[f](\gi, \gj)) \varphi_{\text{val}}(f(\gj)) \nonumber\\[-1\jot]
%     &= \sum_{\gj \in \gS} \sigma\hspace{-0.5mm}_{\gj}\big(\langle \varphi_{\text{qry}}(f(\gi)), \varphi_{\text{key}}(f(\gj)) \rangle \big) \varphi_{\text{val}}(f(\gj)),
% \end{align}
% \vskip -3mm
where~$\sigma\hspace{-0.5mm}_{\gj} = \mathrm{softmax}_{\gj}$~and $\zeta[f]: \gS \rightarrow \sR^{C_{\text{h}}}$.
Finally, multi-head self-attention (Eq.~\ref{eq:7_mhsa}) can be written as:
\begin{align}
    \MHSA(\X)_{i,:} = \gr{m}[f](\gi) &= \varphi_{\text{out}}\Big( \bigcup\nolimits_{h \in [H]} \zeta^{(h)}[f] (\gi) \Big)\nonumber \\[-1\jot] \label{eq:7_full_attention_func}
    &= \varphi_{\text{out}}\Big( \bigcup\nolimits_{h \in [H]} \sum_{\gj \in \gS} \sigma\hspace{-0.5mm}_{\gj}\big(\langle \varphi^{(h)}_{\text{qry}}(f(\gi)), \varphi^{(h)}_{\text{key}}(f(\gj)) \rangle \big) \varphi^{(h)}_{\text{val}}(f (\gj))  \Big),
\end{align}
where $\cup$ is the functional equivalent of the concatenation operator, and $\gr{m}[f]: \gS \rightarrow \sR^{C_{\text{out}}}$.

\textbf{Local self-attention.}
Recall that $\alpha[f]$ assigns an attention scores to every other set element $\gj \in \gS$ relative to the query element $\gi$.
The computational cost of self-attention is often reduced by restricting its calculation to a local neighborhood $\gN(\gi)$ around the query token $\gi$ %.
analogous in nature to the local receptive field of CNNs (Fig.~\ref{fig:7_conv_filters}).
Consequently, \textit{local self-attention} can be written as:
\begin{equation}\label{eq:7_selfatt_localregion}
    \gr{m}[f](\gi) = \varphi_{\text{out}}\Big( \bigcup_{h \in [H]} \sum_{\gj \in \gN(\gi)} \hspace{-1.5mm}\sigma\hspace{-0.5mm}_{\gj}\big(\langle \varphi^{(h)}_{\text{qry}}(f(\gi)), \varphi^{(h)}_{\text{key}}(f(\gj)) \rangle \big) \varphi^{(h)}_{\text{val}}(f(\gj)) \Big).
\end{equation}
Note that Eq.~\ref{eq:7_selfatt_localregion} is equivalent to Eq.~\ref{eq:7_full_attention_func} for $\gN(i) {=} \gS$, i.e. for global neighborhoods.

\textbf{Absolute positional encoding.} The absolute positional encoding is a function $\rho: \gS \rightarrow \sR^{C_{\text{in}}}$ that maps set elements $\gi \in \gS$ to a vector representation of its position: $\rho: \gi \rightarrow \rho(\gi)$. Note that this encoding is not dependent on functions defined on the set but \textit{only} on the set itself.\footnote{Illustratively, one can think of this as a function returning a vector representation of pixel positions in a grid. Regardless of any transformation performed to the image, the labeling of the grid remains equal.} Hence, absolute position-aware self-attention (Eq.~\ref{eq:7_attention_scores_position_matrix_absolute}) can be written as:
\begin{equation}\label{eq:7_attention_scores_func_absolute}
    \gr{m}[f, \rho](\gi) = \varphi_{\text{out}}\Big( \bigcup_{h \in [H]} \sum_{\gj \in \gN(i)}  \hspace{-1.5mm}\sigma\hspace{-0.5mm}_{\gj}\big(\langle \varphi^{(h)}_{\text{qry}}(f(\gi) + \rho(\gi)), \varphi^{(h)}_{\text{key}}(f(\gj) + \rho(\gj)) \rangle \big) \varphi^{(h)}_{\text{val}}(f(\gj)) \Big).
\end{equation}
 The function $\rho$ can be decomposed as two functions $\rho^{P} \circ x$: (\emph{i}) the \textit{position function} $x: \gS \rightarrow \gX$, which provides the position of set elements in the underlying space $\gX$ (e.g., pixel positions), and, (\emph{ii}) the \textit{positional encoding} $\rho^{P}: \gX \rightarrow \sR^{C_{\text{in}}}$, which provides vector representations of elements in $\gX$. This distinction will be of utmost importance when we pinpoint where exactly (group) equivariance must be imposed to the self-attention operation (Sec.~\ref{sec:7_where_equivariance}, Sec.~\ref{sec:7_group_equiv_stand_alone}).

\textbf{Relative positional encoding.} Here, positional information is provided in a relative manner. That is, we now provide vector representations of relative positions $\rho(\gi, \gj) \coloneqq \rho^{P}(x(\gj) -x(\gi))$ among pairs $(\gi, \gj)$, $\gi \in \gS, \gj \in \gN(i)$. Consequently, relative position-aware self-attention (Eq.~\ref{eq:7_relative_pos_attention_matrix}) can be written as:
\begin{equation}\label{eq:7_relaive_pos_att_func}
    \gr{m}^{r}[f, \rho](\gi) = \varphi_{\text{out}}\Big( \bigcup_{h \in [H]} \sum_{\gj \in \gN(i)} \hspace{-1.5mm}\sigma\hspace{-0.5mm}_{\gj}\big(\langle \varphi^{(h)}_{\text{qry}}(f(\gi)), \varphi^{(h)}_{\text{key}}(f(\gj) + \rho(\gi, \gj)) \rangle \big) \varphi^{(h)}_{\text{val}}(f(\gj)) \Big).
\end{equation}

\vspace{-7mm}
\section{Equivariance Analysis of Self-Attention}\label{sec:7_equiv_prop_selfatt}
In this section we analyze the equivariance properties of self-attention. Since the analysis largely relies on group theory, we provide all concepts required for proper understanding in Appx.~\ref{appx:7_group_concepts}.
\vspace{-7mm}
\subsection{Group equivariance and equivariance for functions defined on sets}

First we provide the general definition of group equivariance and refine it to relevant groups next. Additionally, we define the property of \textit{unique equivariance} to restrict equivariance to a given group.

\begin{definition}[\textbf{Group equivariance}]
\label{def:group_equivariance}
Let $\gG$ be a group (Def.~\ref{def:group}), $\gS, \gS'$ be sets, $\gV, \gV'$ be vector spaces, and $\gL_{g}[\cdot], \gL^{'}_{g}[\cdot]$ be the induced (left regular) representation (Def.~\ref{def:group_repr}) of $\gG$ on $L_{\gV}(\gS)$ and $L_{\gV'}(\gS')$, respectively. We say that a map $\varphi: L_{\gV}(\gS) \rightarrow L_{\gV'}(\gS')$ is equivariant to the action of $\gG$ -- or $\gG$-equivariant --, if it commutes with the action of $\gG$. That is, if:
\begin{equation*}
\varphi\big[\gL_{\gg}[f]\big] = \gL_{\gg}'\big[\varphi[f]\big], \ \ \forall f \in L_{\gV}(\gS), \ \forall \gg \in \gG.
\end{equation*}
\end{definition}

\begin{subdefinition}[\textbf{Permutation equivariance}] Let $\gS = \gS'=\{\gi\}_{\gi = 1}^{N}$ be a set of $N$ elements, and $\gG = \sS_{N}$ be the group of permutations on sets of $N$ elements. A map $\varphi: L_{\gV}(\gS) \rightarrow L_{\gV'}(\gS)$ is said to be equivariant to the action of $\ \sS_{N}$ -- or permutation equivariant --, if:
\begin{equation*}
\varphi\big[\gL_{\pi}[f]\big](\gi) = \gL_{\pi}'\big[\varphi[f]\big](\gi), \ \ \forall f \in L_{\gV}(\gS), \ \forall \pi \in \sS_{N}, \ \forall \gi \in \gS,
\end{equation*}
where $\gL_{\pi}[f](\gi) \coloneqq f(\pi^{-1}(\gi))$, and $\pi: \gS \rightarrow \gS$ is a bijection from the set to itself. The element $\pi(\gi)$ indicates the index to which the $\gi$-th element of the set is moved to as an effect of the permutation $\pi$.
I.o.w., $\varphi$ is said to be permutation equivariant if it commutes with permutations $\pi \in \sS_{N}$, i.e., if permutations of its input produce equivalent permutations on its output.
\end{subdefinition}
Several of the transformations of interest, e.g., rotations, translations, are not defined on sets. Luckily, as we consider sets gathered from homogeneous spaces $\gX$ where these transformations are well-defined, e.g., $\sR^{2}$ for pixels, there exists an injective map $x: \gS \rightarrow \gX$ that associates a position in $\gX$ to each set element, the \textit{position function}. % JB: the reference is not very helpful given how long is the section we point to
% (\S\ref{sec:7_functional_attn}).
%
In Appx.~\ref{appx:7_homospaces} we show that the action of $\gG$ on such a set is well-defined and induces a group representation to functions on it. With this in place, we are now able to define equivariance of set functions to groups whose actions are defined on homogeneous spaces.
\begin{definition}[\textbf{Equivariance of set functions to groups acting on homogeneous spaces}]
\label{def:group_equivariance}
Let $\gG$ be a group acting on two homogeneous spaces $\gX$ and $\gX'$, let $\gS, \gS'$ be sets and $\gV, \gV'$ be vector spaces. Let $x: \gS \rightarrow \gX$ and $x': \gS' \rightarrow \gX'$ be injective maps. % from the sets $\gS, \gS'$ to the corresponding homogeneous spaces $\gX, \gX'$.
We say that a map  $\varphi: L_{\gV}(\gS) \rightarrow L_{\gV'}(\gS')$ is equivariant to the action of $\gG$ -- or $\gG$-equivariant --, if it commutes with the action of $\gG$. That is, if:
\begin{equation*}
\varphi\big[\gL_{\gg}[f]\big] = \gL_{\gg}'\big[\varphi[f]\big], \ \ \forall f \in L_{\gV}(\gS), \ \forall \gg \in \gG,
\end{equation*}
where $\gL_{\gg}[f](\gi) \coloneqq f(x^{-1}(\gg^{-1}x(\gi)))$, $ \gL^{'}_{\gg}[f](\gi) \coloneqq f(x'^{-1}(\gg^{-1}x'(\gi)))$ are the induced (left regular) representation of $\gG$ on $L_{\gV}(\gS)$ and $L_{\gV'}(\gS')$, respectively. I.o.w., $\varphi$ is said to be $\gG$-equivariant if a transformation $\gg \in \gG$ on its input produces an equal transformation on its output.
\end{definition}
\begin{subdefinition}[\textbf{Translation equivariance}] Let $\gS$, $\gS'$ be sets and let $x: \gS \rightarrow \gX$ and $x': \gS' \rightarrow \gX'$ be injective maps from the sets $\gS, \gS'$ to the corresponding homogeneous spaces $\gX, \gX'$ on which they are defined, e.g., $\sR^{d}$ and $\gG$. With $(\gX, + )$ the translation group acting on $\gX$, we say that a map $\varphi: L_{\gV}(\gS) \rightarrow L_{\gV'}(\gS)$ is equivariant to the action of $(\gX, + )$ -- or translation equivariant --, if:
\begin{equation*}
\varphi\big[\gL_{y}[f]\big](\gi) = \gL_{y}'\big[\varphi[f]\big](\gi), \ \ \forall f \in L_{\gV}(\gS), \ \forall y \in \gX,
\end{equation*}
with $\gL_{y}[f](\gi) {\coloneqq} f( x^{-1}(x(\gi) - y))$, $\gL_{y}'[f](\gi) {\coloneqq} f( x'^{-1}(x'(\gi) - y))$. I.o.w., $\varphi$ is said to be translation equivariant if a translation on its argument produces an equal translation on its output.
\end{subdefinition}
\vspace{-7mm}
\subsection{Equivariance properties of self-attention}
In this section we analyze the equivariance properties of the self-attention. The proofs to all the propositions stated in the main text are provided in Appx.~\ref{appx:7_proofs}.
\begin{claim}\label{claim:perm_equiv}
The global self-attention formulation without positional encoding (Eqs.~\ref{eq:7_mhsa},~\ref{eq:7_full_attention_func}) is permutation equivariant. That is, it holds that: $\gr{m}[\gL_{\pi}[f]](\gi) = \gL_{\pi}[\gr{m}[f]](\gi)$.
\end{claim}

Note that permutation equivariance only holds for global self-attention.
The local variant proposed in Eq.~\ref{eq:7_selfatt_localregion} reduces permutation equivariance to a smaller set of permutations where neighborhoods are conserved under permutation, i.e., $\sS_{\gN} = \{ \pi \in \sS_{N} \mid \gj \in \gN(\gi) \rightarrow \pi(\gj) \in  \gN(\gi), \ \forall \gi \in \gS \}$.

\textbf{Permutation equivariance induces equivariance to important (sub)groups.} Consider the cyclic group of order 4, $\gZ_{4}=\{e, r, r^{2}, r^{3}\}$ which induces planar rotations by 90$^{\circ}$.\footnote{$e$ represents a 0$^{\circ}$ rotation, i.e., the identity.
The remaining elements $r^{j}$ represent rotations by $($90$\cdot j)^{\circ}$.} 
 As every rotation in $\gZ_{4}$ effectively induces a permutation of the tokens positions, it can be shown that $\gZ_{4}$ is a subgroup of $\sS_{N}$, i.e., $\sS_{N} \geq \gZ_{4}$. Consequently, maps equivariant to permutations are automatically equivariant to $\gZ_{4}$.
 However, as the permutation equivariance constraint is harder than that of $\gZ_{4}$-equivariance, imposing $\gZ_{4}$-equivariance as a result of permutation equivariance is undesirable in terms of expressivity. 
Consequently, \citet{ravanbakhsh2017equivariance} introduced the concept of \textit{unique $\gG$-equivariance} to express the family of functions equivariant to $\gG$ but not equivariant to other groups $\gG'\geq \gG$:
\begin{definition}[\textbf{Unique $\gG$-equivariance}]\label{def:unique_equiv} Let $\gG$ be a subgroup of $\gG'$ (Def.~\ref{def:subgroup}). We say that a map $\varphi$ is uniquely $\gG$-equivariant in and only if it is $\gG$-equivariant but not $\gG'$-equivariant for any group $\gG'$ from which $\gG$ is a subgroup.
\end{definition}

% REBUTTAL
In the following sections, we show that we can enforce unique equivariance not only to subgroups of $\sS_{N}$, e.g., $\gZ_{4}$, but also to other interesting groups not contained in $\sS_{N}$, e.g., groups of rotations finer than 90 degrees. This is achieved by enriching set functions with a proper positional encoding.
\begin{claim}\label{claim:no_perm_equiv_no_trans_equiv}
Absolute position-aware self-attention (Eqs.~\ref{eq:7_attention_scores_position_matrix_absolute},~\ref{eq:7_attention_scores_func_absolute}) is neither permutation nor translation equivariant. i.e., $\gr{m}[\gL_{\pi}[f], \rho](\gi) \neq \gL_{\pi}[\gr{m}[f, \rho]](\gi)$ and $\gr{m}[\gL_{y}[f], \rho](\gi) \neq \gL_{y}[\gr{m}[f, \rho]](\gi)$.
\end{claim}

Though absolute positional encodings do disrupt permutations equivariance, they are unable to provide translation equivariance. We show next that translation equivariance is obtained via relative encodings.
% REBUTTAL
% Absolute positional encodings do disrupt permutation equivariance but do not provide translation equivariance. We see next that translation equivariance can be obtained via relative encodings.

\begin{claim}\label{claim:trans_equiv}
Relative position-aware self-attention (Eq.~\ref{eq:7_relaive_pos_att_func}) is translation equivariant. That is, it holds that: $\gr{m}^{r}[\gL_{y}[f], \rho](\gi) = \gL_{y}[\gr{m}^{r}[f, \rho]](\gi)$.
\end{claim}
\vspace{-7mm}
\subsection{Where exactly is equivariance imposed in self-attention?}\label{sec:7_where_equivariance}

In the previous section we have seen two examples of successfully imposing group equivariance to self-attention. Specifically, we see that no positional encoding allows for permutation equivariance and that a relative positional encoding allows for translation equivariance. For the latter, as shown in the proof of Prop.~\ref{claim:trans_equiv} (Appx.~\ref{appx:7_proofs}), this comes from the fact that for all shifts $y \in \gX$,
\begin{equation}
    \rho(x^{-1}(x(\gi) + y), x^{-1}(x(\gj) + y)) = \rho^{P}(x(\gj) + y - (x(\gi) + y)) = \rho^{P}(x(\gj) -x(\gi)) = \rho(\gi, \gj).
\end{equation}
That is, from the fact that the relative positional encoding is invariant to the action of the translation group, i.e., $ \gL_{y}[\rho](\gi, \gj) = \rho(\gi, \gj)$, $\forall y \in \gX$. Similarly, the absence of positional encoding -- more precisely, the use of a constant positional encoding --, is what allows for permutation equivariance (Prop.~\ref{claim:perm_equiv}, Appx.~\ref{appx:7_proofs}). Specifically, constant positional encodings $\rho_{c}(\gi) = c$, $\forall \gi \in \gS$ are invariant to the action of the permutation group, i.e., $ \gL_{\pi}[\rho_{c}](\gi) = \rho_{c}(\gi)$, $\forall \pi \in \sS_{N}$.

From these observations, we conclude that $\gG$-equivariance is obtained by providing positional encodings which are \textit{\textbf{invariant to the action of the group}} $\boldsymbol{\gG}$, i.e., s.t., $\gL_{\gg}[\rho] = \rho$, $\forall \gg \in \gG$. Furthermore, unique $\gG$-equivariance is obtained by providing positional encodings which are invariant to the action of $\gG$ but \textbf{\textit{not invariant to the action of any other group}} $\boldsymbol{\gG'\geq \gG}$. This is a key insight that allows us to provide (unique) equivariance to arbitrary symmetry groups, which we provide next.%\jb{can we point to the proof?}

\vspace{-7mm}
\section{Group Equivariant Stand-Alone Self-Attention}\label{sec:7_group_equiv_stand_alone}
In Sec.~\ref{sec:7_where_equivariance} we concluded that unique $\gG$-equivariance is induced in self-attention by introducing positional encodings which are invariant to the action of $\gG$ but not invariant to the action of other groups $\gG'\geq \gG$. However, this constraint does not provide any information about the expressivity of the mapping we have just made $\gG$-equivariant. Let us first illustrate why this is important:

Consider the case of imposing rotation and translation equivariance to an encoding defined in $\sR^{2}$. Since translation equivariance is desired, a relative positional encoding is required.
For rotation equivariance,
we must further impose the positional encoding to be equal for all rotations. That is
% we need to additionally impose that for every possible angle, the relative positional encoding must be equal. That is,
$\gL_{\theta}[\rho](\gi, \gj) \overset{!}{=} \rho(\gi, \gj)$, $\forall \theta \in [0, 2\pi]$, where $\gL_{\theta}[\rho](\gi, \gj) \coloneqq \rho^{P}(\theta^{-1}x(\gj) - \theta^{-1}x(\gi))$, and $\theta^{-1}$ depicts a rotation by $-\theta$ degrees. This constraint leads to an isotropic positional encoding
% only allowed to change in the radial direction
unable to discriminate among orientations, which in turn enforces rotation invariance instead of rotation equivariance.% . This is disadvantageous as the capacity to discriminate among orientations is entirely lost. In other words, it imposes an isotropic positional encoding, which in turn
% This
%
\footnote{This phenomenon arises from the fact that $\sR^{2}$ is a quotient of the roto-translation group. Consequently, imposing group equivariance in the quotient space is equivalent to imposing an additional homomorphism of constant value over its cosets.
Conclusively, the resulting map is of constant value over the rotation elements and, thus, unable to discriminate among them.
See Ch.~3.1 \citep{dummit2004abstract} for an intuitive description.}
This is alleviated by \textit{lifting} the underlying function on $\sR^{2}$ to a space where rotations are explicitly encoded (Fig.~\ref{fig:7_lifting}). To this end, one performs self-attention operations for positional encodings $\gL_{\theta}[\rho]$ of varying values $\theta$ and indexes their responses by the corresponding $\theta$ value. Next, as rotations are now explicitly encoded, a positional encoding can be defined in this space which is able to discriminate among rotations (Fig.~\ref{fig:7_group_selfatt}). This in turn allows for rotation equivariance instead of rotation invariance.
% \vspace{-3mm}
% \subsection{Constructing group equivariant self-attention networks.} 
% \vspace{-1mm}

It has been shown both theoretically \citep{ravanbakhsh2020universal} and empirically \citep{weiler2019general} that the most expressive class of $\gG$-equivariant functions is given by functions that follow the regular representation of $\gG$. In order to obtain feature representations that behave that way, we introduce a lifting self-attention layer (Fig.~\ref{fig:7_lifting}, Eq.~\ref{eq:7_lifting_selfatt}) that receives an input function on $\sR^{d}$ and produces a feature representation on $\gG$. Subsequently, arbitrarily many group self-attention layers (Fig.~\ref{fig:7_group_selfatt}, Eq.~\ref{eq:7_group_selfatt}) interleaved with optional point-wise non-linearities can be applied. At the end of the network a feature representation on $\sR^{d}$ can be provided by pooling over $\gH$. In short, we provide a pure self-attention analogous to \cite{cohen2016group}. However, as the group acts directly on the positional encoding, our networks are \textit{steerable} as well \citep{weiler2018learning}. This allows us to go beyond group discretizations that live in the grid without introducing interpolation artifacts (Sec.~\ref{sec:7_steerability}).

Though theoretically sound, neural architectures using regular representations are unable to handle continuous groups directly in practice. This is a result of the summation over elements $\tilde{\gh} \in \gH$ in Eq.~\ref{eq:7_group_selfatt_compressed}, which becomes an integral for continuous groups. Interestingly, using discrete groups does not seem to be detrimental in practice. Our experiments indicate that performance saturates for fine discrete approximations of the underlying continuous group (Tab.~\ref{tab:7_results}). In fact, \cite[Tab.~3]{weiler2019general} show via extensive experiments that networks using regular representations and fine enough discrete approximations consistently outperform networks handling continuous groups via irreducible representations. We conjecture this is a result of the networks receiving discrete signals as input. As the action of several group elements fall within the same pixel, no further improvement can be obtained.
\vspace{-7mm}
\subsection{Group self-attention is an steerable operation}\label{sec:7_steerability}

\begin{figure}
     \centering
     \begin{subfigure}[b]{0.33\textwidth}
         \centering
         \captionsetup{justification=centering}
         \includegraphics[width=\textwidth]{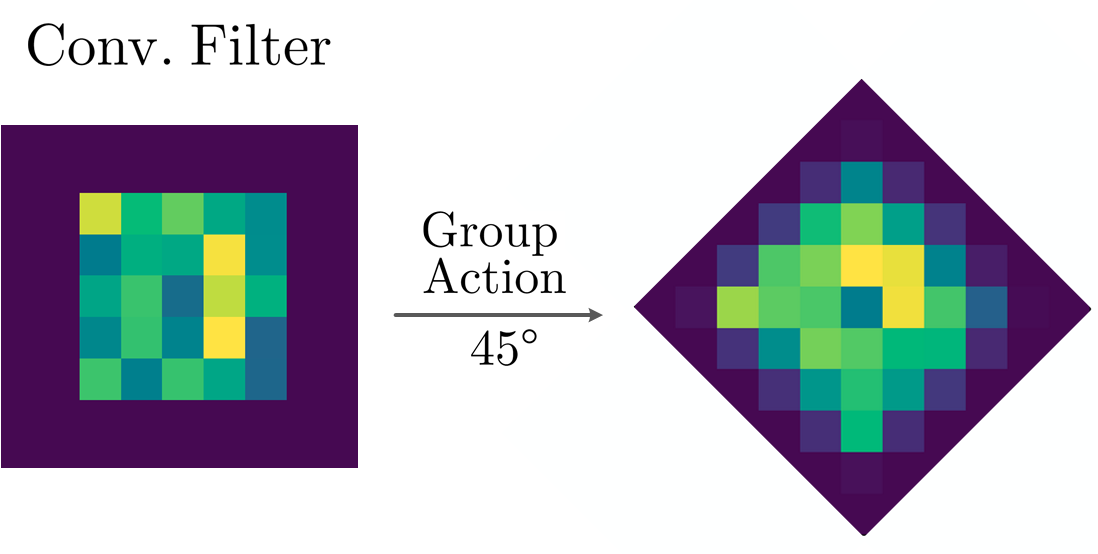}
         \caption{Discrete convolution}
         \label{fig:7_steerability_conv}
     \end{subfigure}
     \hfill
     \begin{subfigure}[b]{0.59\textwidth}
         \centering
         \captionsetup{justification=centering}
         \includegraphics[width=\textwidth]{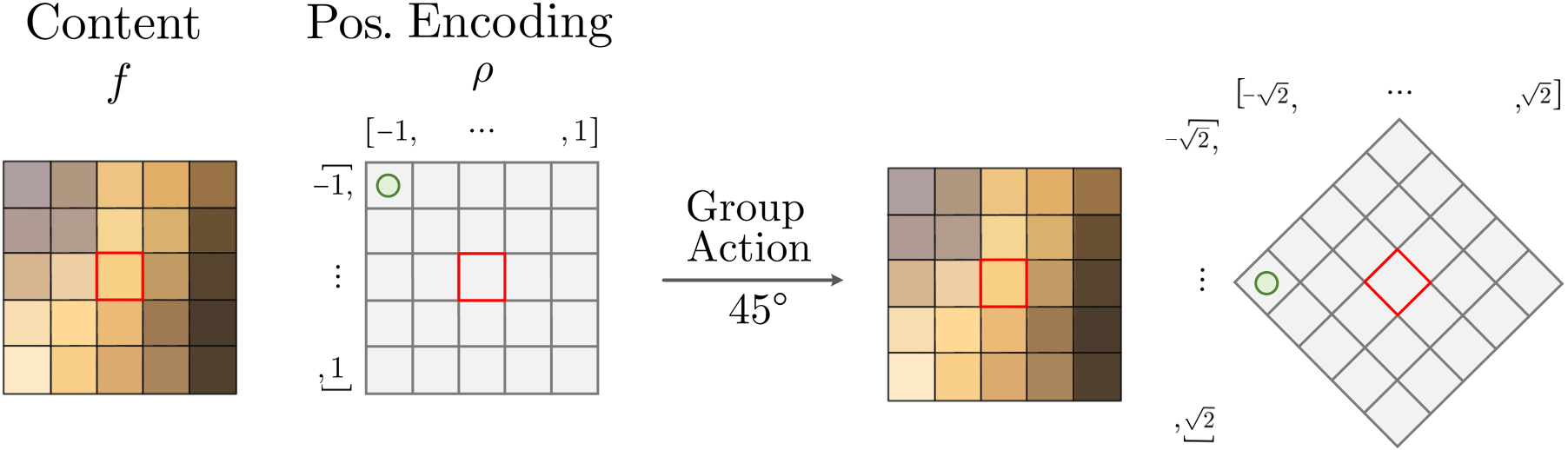}
         \caption{Group self-attention}
         \label{fig:7_steerability_selfatt}
     \end{subfigure}
     \vspace{-1mm}
        \caption{Steerability analysis of discrete convolutions and group self-attention.
        \vspace{-1.0em}
        }
        \label{fig:7_steerability}
\end{figure}
Convolutional filters are commonly parameterized by weights on a discrete grid, which approximate the function implicitly described by the filter at the grid locations. Unfortunately, for groups whose action does not live in this grid, e.g., 45$^{\circ}$ rotations, the filter must be interpolated. This is problematic as these filters are typically small and the resulting interpolation artifacts can be severe (Fig.~\ref{fig:7_steerability_conv}). \textit{Steerable CNNs} tackle this problem by parameterizing convolutional filters on a continuous basis on which the action of the group is well-defined, e.g., \textit{circular harmonics} \citep{weiler2018learning}, \textit{B-splines} \citep{bekkers2020bspline}. In group self-attention, the action of the group leaves the content of the image intact and only modifies the positional encoding (Figs.~\ref{fig:7_lifting},~\ref{fig:7_group_selfatt}). As the positional encoding lives on a con- tinuous space, it can be transformed at an arbitrary grade of precision without interpolation (Fig.~\ref{fig:7_steerability_selfatt}).
\vspace{-7mm}
\subsection{Lifting and group self-attention}\label{sec:7_lifting_self_att}

\begin{figure}
    \centering
    \includegraphics[width=1.0\textwidth]{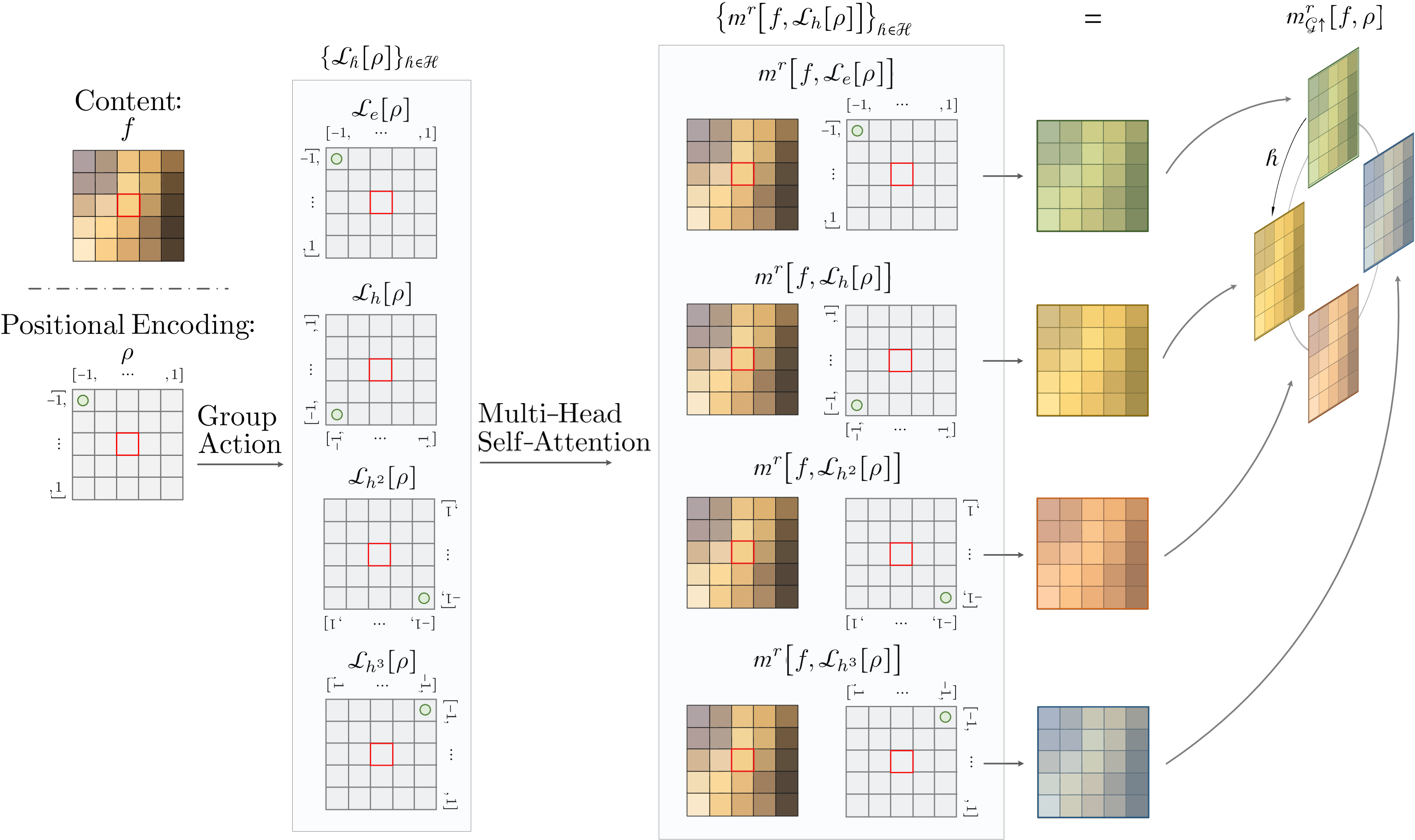}
    \caption{Lifting self-attention on the roto-translation group for discrete rotations by 90 degrees (also called the $\gZ_{4}$ group). The $\gZ_{4}$ group is defined as $\gH = \{e, \gh, \gh^{2}, \gh^{3}\}$, where $\gh$ depicts a 90$^{\circ}$ rotation. The lifting self-attention corresponds to the concatenation of $\mathopen| \gH \mathclose| = 4$ self-attention operations between the input $f$ and $\gh$-transformed versions of the positional encoding $\gL[\rho]$, $\forall \gh \in \gH$. As a result, the model \enquote{sees} the input $f$ at each of the rotations in the group at once. Since $\gZ_{4}$ is a cyclic group, i.e., $\gh^{4} = e$, functions on this group are often represented as responses on a ring (right side of the image).% This is a self-attention analogous to the regular lifting group convolution broadly utilized in group equivariant learning literature, e.g., \cite{GCNN, romero2020attentive}.
    \vspace{-3mm}}
    % Lifting self-attention computes multiple conventional self-attention operations for positional encodings $\{ \gL_{\gh}[\rho]\}_{\gh \in \gH}$ modified by the action of the affine group $\gH$ considered. This operation induces a regular group representation to the self-attention operation, which incorporates the structure of the group acting on the homogeneous space from which the data is gathered. This is a self-attention analogous to the regular lifting group convolution broadly utilized in group equivariant learning literature, e.g., \cite{GCNN}.}
    \label{fig:7_lifting}
\end{figure}

 \textbf{Lifting self-attention (Fig.~\ref{fig:7_lifting}).} Let $\gG = \sR^{d} \rtimes \gH$ be an affine group (Def.~\ref{def:affine_group}) acting on $\sR^{d}$. The \textit{lifting self-attention} $\gr{m}^{r}_{\gG\uparrow}[f, \rho]: L_{\gV}(\sR^{d}) \rightarrow L_{\gV'}(\gG)$ is a map from functions on $\sR^{d}$ to functions on $\gG$ obtained by modifying the relative positional encoding $\rho(\gi, \gj)$ by the action of group elements $\gh \in \gH$: $\{\gL_{\gh}[\rho](\gi, \gj)\}_{\gh \in \gH}$, $\gL_{\gh}[\rho](\gi, \gj) = \rho^{P}(\gh^{-1}x(\gj) - \gh^{-1}x(\gi))$. It corresponds to the concatenation of multiple self-attention operations (Eq.~\ref{eq:7_relaive_pos_att_func}) indexed by $\gh$ with varying positional encodings $\gL_{\gh}[\rho]$ :
\begin{align}
\setlength{\abovedisplayskip}{3.5pt}
\setlength{\belowdisplayskip}{4pt}
    \hspace{-3mm}\gr{m}^{r}_{\gG\uparrow}&[f, \rho](\gi, \gh) = \gr{m}^{r}\big[f, \gL_{\gh}[\rho]\big](\gi) \label{eq:7_lifting_selfatt_compressed} \\ %&= \Big\{\gr{m}^{r}\big[f, \gL_{\gr{v}}[\rho]\big]\Big\}_{\gr{v} \in \gH}(\gi, \gh) \label{eq:7_lifting_selfatt_compressed} \\
    &= \varphi_{\text{out}}\Big( \bigcup_{h \in [H]} \sum_{\gj \in \gN(i)} \hspace{-2mm}\sigma\hspace{-0.5mm}_{\gj}\big(\langle \varphi^{(h)}_{\text{qry}}(f(\gi)), \varphi^{(h)}_{\text{key}}(f(\gi) + \gL_{\gh}[\rho](\gi, \gj)) \rangle \big) \varphi^{(h)}_{\text{val}}(f(\gj)) \Big).\label{eq:7_lifting_selfatt}
\end{align}
\begin{claim}\label{claim:lifting_gequiv}
Lifting self-attention is $\gG$-equivariant. That is: $\gr{m}^{r}_{\gG\uparrow}[\gL_{\gg}[f], \rho](\gi, \gh) = \gL_{\gg}[\gr{m}^{r}_{\gG\uparrow}[f, \rho]](\gi, \gh)$.
\end{claim}

\begin{figure}
    \centering
    \includegraphics[width=1.0\textwidth]{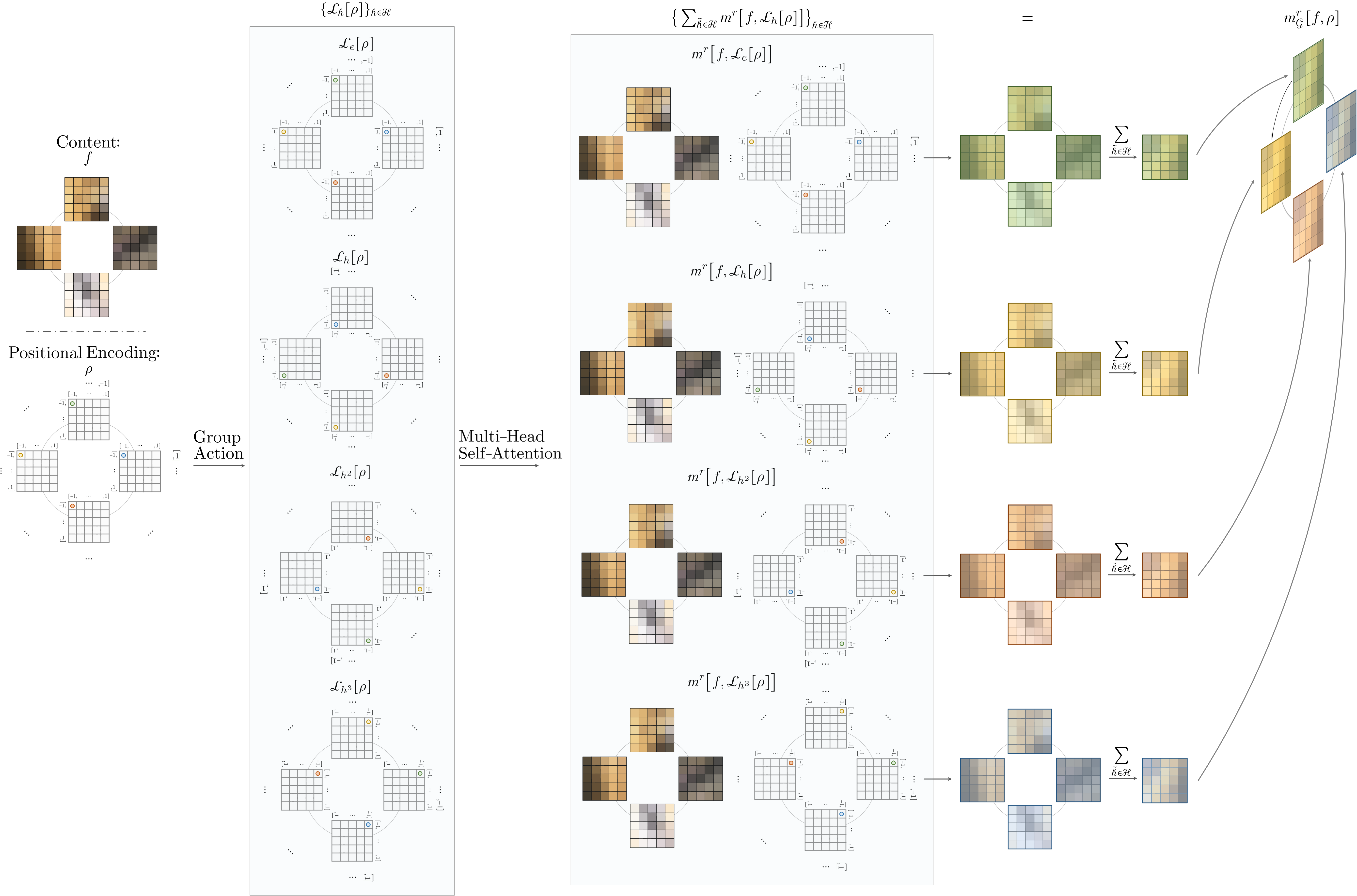}
    \caption{Lifting self-attention on the roto-translation group for discrete rotations by 90 degrees (also called the $\gZ_{4}$ group). The $\gZ_{4}$ group is defined as $\gH = \{e, \gh, \gh^{2}, \gh^{3}\}$, where $\gh$ depicts a 90$^{\circ}$ rotation. Analogous to lifting self-attention (Fig.~\ref{fig:7_lifting}), group self-attention corresponds to a concatenation of $\mathopen| \gH \mathclose| = 4$ self-attention operations between the input $f$ and $\gh$-transformed versions of the positional encoding $\gL[\rho]$, $\forall \gh \in \gH$. However, in contrast to lifting self-attention, both $f$ and $\rho$ are now defined on the group $\gG$. Consequently, an additional sum over $\tilde{\gh}$ is required during the operation (c.f., Eq.~\ref{eq:7_group_selfatt}). Since $\gZ_{4}$ is a cyclic group, i.e., $\gh^{4} = e$, functions on $\gZ_{4}$ are often represented as responses on a ring (right side of the image).% This is a self-attention analogous to the regular group convolution broadly utilized in group equivariant learning literature, e.g., \cite{GCNN, romero2020attentive}.
    \vspace{-3mm}}
    \label{fig:7_group_selfatt}
\end{figure}

\textbf{Group self-attention (Fig.~\ref{fig:7_group_selfatt}).} Let $\gG = \sR^{d} \rtimes \gH$  be an affine group acting on itself and $f(\gi, \tilde{\gh}) \in L_{\gV}(\gG)$, $\gi \in \gS$, $\tilde{\gh} \in \gH$, be a function defined on a set immersed with the structure of the group $\gG$. That is, enriched with a positional encoding $\rho((\gi, \Tilde{\gh}), (\gj, \hat{\gh}))\coloneqq \rho^{P}((x(\gj) - x(\gi), \tilde{\gh}^{-1}\hat{\gh}))$, $\gi, \gj \in \gS$, $\tilde{\gh}, \hat{\gh} \in \gH$. The \textit{group self-attention} $\gr{m}^{r}_{\gG}[f, \rho]: L_{\gV}(\gG) \rightarrow L_{\gV'}(\gG)$ is a map from functions on $\gG$ to functions on $\gG$ obtained by modifying the group positional encoding by the action of group elements $\gh \in \gH$: $\{\gL_{\gh}[\rho]((\gi, \Tilde{\gh}), (\gj, \hat{\gh}))\}_{\gh \in \gH}$, $\gL_{\gh}[\rho]((\gi, \Tilde{\gh}), (\gj, \hat{\gh})) = \rho^{P}(\gh^{-1}(x(\gj) - x(\gi)), \gh^{-1}(\tilde{\gh}^{-1}\hat{\gh}))$. It corresponds to the concatenation of multiple self-attention operations (Eq.~\ref{eq:7_relaive_pos_att_func}) indexed by $h$ with varying positional encodings $\gL_{\gh}[\rho]$ and followed by a summation over the output domain along $\tilde{\gh}$:
\begin{align}
    \hspace{-1.5mm}\gr{m}^{r}_{\gG}[f,& \rho](\gi, \gh)
    = \sum\nolimits_{\tilde{\gh} \in \gH}\gr{m}^{r}\big[f, \gL_{\gr{h}}[\rho]\big](\gi, \tilde{\gh}) \label{eq:7_group_selfatt_compressed} \\ % &= \Big\{\sum\nolimits_{\tilde{\gh} \in \gH}\gr{m}^{r}\big[f, \gL_{\gr{v}}[\rho]\big](\ \cdot\ , \tilde{\gh})\Big\}_{\gr{v} \in \gH}(\gi, \gh) \label{eq:7_group_selfatt_compressed} \\
    &= \varphi_{\text{out}}\Big( \bigcup_{h \in [H]}
    \sum_{\tilde{\gh} \in \gH}\hspace{-1.05cm} \sum_{\qquad \quad (\gj, \hat{\gh}) \in \gN(i, \tilde{\gh})} \hspace{-0.6cm}\hspace{-5mm}\sigma\hspace{-0.5mm}_{\gj , \hat{\gh}}\big(\langle \varphi_{\text{qry}}^{(h)}(f(\gi, \tilde{\gh})), \varphi_{\text{key}}^{(h)}(f(\gj, \hat{\gh}) \nonumber\\[-5\jot]
    &\hspace{6cm}+ \gL_{\gh}[\rho]((\gi, \tilde{\gh}), (\gj, \hat{\gh})) \rangle \big)
    \varphi_{\text{val}}^{(h)}(f(\gj, \hat{\gh}))
    \Big).\label{eq:7_group_selfatt}
\end{align}
In contrast to vanilla and lifting self-attention, the group self-attention neighborhood $\gN(i, \tilde{\gh})$ is now defined on the group. This allows distinguishing across group transformations, e.g., rotations.
\begin{claim}\label{claim:g_equivariance_g_attention}
Group self-attention is $\gG$-equivariant. That is: $\gr{m}^{r}_{\gG}[\gL_{\gg}[f], \rho](\gi, \gh) =\break \gL_{\gg}[\gr{m}^{r}_{\gG}[f, \rho]](\gi, \gh)$.
\end{claim}
Non-unimodular groups, i.e., groups that modify the volume of the objects they act upon, e.g., the scale group, require a special treatment, which we provide in Appx.~\ref{appx:7_non_unimodular}.
\vspace{-7mm}
\subsection{Group self-attention is a generalization of the group convolution}\label{sec:7_conv_generalization}

We have demonstrated that it is sufficient to define self-attention as a function on the group $\gG$ and ensure that $\gL_{\gg}[\rho] = \rho \ \forall \gg \in \gG$ in order to enforce $\gG$-equivariance. Interestingly, this observation is inline with the main statement of \citet{kondor2018generalization} for (group) convolutions: \enquote{\emph{the group convolution on $\gG$ is the only (unique) $\gG$-equivariant linear map}}. In fact, our finding can be formulated as a generalization of \citet{kondor2018generalization}'s statement as:
\begin{center}
    \textbf{\enquote{\emph{Linear mappings on $\gG$ whose positional encoding is $\gG$-invariant are $\gG$-equivariant.}}}
\end{center}
This statement is more general than that of \citet{kondor2018generalization}, as it holds for data structures where (group) convolutions are not well-defined, e.g., sets, and it is equivalent to \citet{kondor2018generalization}'s statement for structures where (group) convolutions are well-defined. It is also congruent with results complementary to \cite{kondor2018generalization, cohen2019general, bekkers2020bspline} as well as several works on group equivariance handling set-like structures like point-clouds \citep{thomas2018tensor, defferrard2020deepsphere, finzi2020generalizing, fuchs2020se} and symmetric sets \citep{maron2020learning}.

In addition, we can characterize the expressivity of group self-attention. It holds that (\emph{i}) group self-attention generalizes the group convolution and (\emph{ii}) regular global group self-attention is an equivariant universal approximator. Statement~(\emph{i}) follows from the fact that any convolutional layer can be described as a multi-head self-attention layer provided enough heads \citep{Cordonnier2020On}, yet self-attention often uses larger receptive fields. As a result, self-attention is able to describe a larger set of functions than convolutions, e.g., Fig.~\ref{fig:7_comparison_att_conv}. \citet{Cordonnier2020On}'s statement can be easily extended to group self-attention by incorporating an additional dimension corresponding to $\gH$ in their derivations, and defining neighborhoods in this new space with a proportionally larger number of heads. Statement (\emph{ii}) stems from the finding of \citet{ravanbakhsh2020universal} that functions induced by regular group representations are equivariant universal approximators provided \emph{full kernels}, i.e., global receptive fields. Global receptive fields are required to guarantee that the equivariant map is able to model any dependency among input components. Global receptive fields are readily utilized by our proposed regular global group self-attention and, provided enough heads, one can ensure that any such dependencies is properly modelled.

\vspace{-7mm}
\section{Experiments}\label{sec:7_experiments}

We perform experiments on three image benchmark datasets for which particular forms of equivariance are desirable. We evaluate our approach by contrasting \gsa\ equivariant to multiple symmetry groups. Additionally, we conduct an study on rotMNIST to evaluate the performance of \gsa\ as a function of the neighborhood size. All our networks follow the structure shown in Fig.~\ref{fig:7_net_structure} and vary only in the number of blocks and channels. We emphasize that both the architecture and the number of parameters in \gsa\ is left unchanged regardless of the group used. Our results illustrate that \gsa\ consistently outperform equivalent non-equivariant attention networks. We further compare \gsa\ with convolutional architectures. Though our approach does not build upon these networks, this comparison provides a fair view to the yet present gap between self-attention and convolutional architectures in vision tasks, also present in their group equivariant counterparts.

 \textit{\textbf{Efficient implementation of lifting and group self-attention.}} Our self-attention implementation takes advantage of the fact that the group action only affects the positional encoding $\Pm$ to reduce the total computational cost of the operation. Specifically, we calculate self-attention scores w.r.t. the content $\Xm$ once, and reuse them for all transformed versions of the positional encoding $\{\gL_{\gh}[\rho]\}_{\gh \in \gH}$.%\footnote{A na\"{i}ve implementation would require recomputing content attention values for each $\gL_{\gh}[\rho], \gh \in \gH$.}

\textit{\textbf{Nomenclature.}} We refer to translation equivariant self-attention models as {\btt Z2\_SA}. Reflection equivariant models receive the keyword {\btt M}, e.g., \msa, and rotation equivariant models the keyword {\btt Rn}, where {\btt n} depicts the angle discretization. For example, \resa\  depicts a model equivariant to rotations by 45 degrees. Specific model architectures are provided in Appx.~\ref{appx:7_exp_details}.

\textbf{RotMNIST.} The rotated MNIST dataset \citep{larochelle2007empirical} is a classification dataset often used as a standard benchmark for rotation equivariance. It consists of 62$k$ gray-scale 28x28 uniformly rotated handwritten digits, divided into training, validation and test sets of 10$k$, 2$k$ and $50$k images.
First, we study the effect of the neighborhood size on classification accuracy and convergence time. We train \rfsa\ networks for 300 epochs with vicinities \texttt{NxN} of varying size (Tab.~\ref{tab:7_neighborhood_results}, Fig.~\ref{fig:7_earlytrain_progress}). Since \gsa\ optimize where to attend, the complexity of the optimization problem grows as a function of \texttt{N}. Consequently, models with big vicinities are expected to converge slower. However, as the family of functions describable by big vicinities contains those describable by small ones, models with big vicinities are expected to be at least as good upon convergence. Our results show that models with small vicinities do converge much faster (Fig.~\ref{fig:7_earlytrain_progress}). However, though some models with large vicinities do outperform models with small ones, e.g., \texttt{7x7} vs. \texttt{3x3}, a trend of this behavior is not apparent. We conjecture that 300 epochs are insufficient for all models to converge equally well. Unfortunately, due to computational constraints, we were not able to perform this experiment for a larger number of epochs. We consider an in-depth study of this behavior an important direction for future work.

\begin{minipage}[h!]{0.5\textwidth}
\begin{table}[H]
\caption{Accuracy vs. neighborhood size. }\label{tab:7_neighborhood_results}
\vspace{-3mm}
\begin{small}
\begin{sc}
\scalebox{0.76}{
\begin{tabular}[t]{cccc}
\toprule
\multicolumn{4}{c}{\textbf{rotMNIST}}\\
\midrule
\multirow{2}{*}{Model} & \multirow{2}{*}{Nbhd. size} & \multirow{2}{*}{Acc. (\%)} & Train. time  \\
& & & / Epoch\\
\midrule
\multirow{9}{*}{\rfsa} & {\btt 3x3} & 96.56 & 04:53 - 1GPU \\
 & {\btt 5x5} & \textbf{97.49} & 05:34 - 1GPU \\
 & {\btt 7x7} & 97.33 & 09:03 - 1GPU \\
 & {\btt 9x9} & 97.42 & 09:16 - 1GPU \\
 & {\btt 11x11} & 97.17 & 12:09 - 1GPU \\
 & {\btt 15x15} & 96.89 & 10:27 - 2GPU \\
 & {\btt 19x19} & 96.86 & 14:27 - 2GPU \\
 & {\btt 23x23} & 97.05 & 06:13 - 3GPU \\
 & {\btt 28x28} & 96.81 & 12:12 - 4GPU \\
\bottomrule
\end{tabular}
}
\end{sc}
\end{small}
\end{table}
\end{minipage}
\begin{minipage}[h!]{0.52\textwidth}
\begin{figure}[H]
    \centering
    \includegraphics[width=0.85\textwidth]{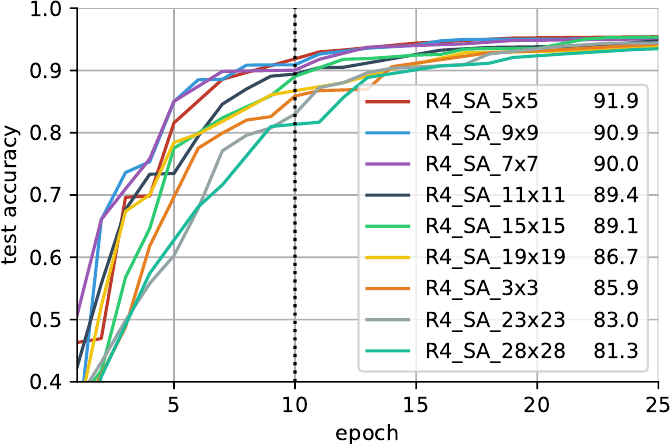}
    \vskip -3mm
    \caption{Test accuracy in early training.}
    \label{fig:7_earlytrain_progress}
\end{figure}
\end{minipage}
\vspace{1mm}

Next, we compare \gsa\ equivariant to translation and rotation at different angle discretizations (Tab.~\ref{tab:7_results}). Based on the results of the previous study, we select a {\btt 5x5}\ neighborhood, as it provides the best trade-off between accuracy and convergence time. Our results show that finer discretizations lead to better accuracy but saturates around {\btt R12}. We conjecture that this is due to the discrete resolution of the images in the dataset, which leads finer angle discretizations to fall within the same pixel.
\begin{table}
\caption{Classification results. All convolutional architectures use {\btt 3x3} filters.}\label{tab:7_results}
\vspace{-3mm}
\begin{small}
\begin{sc}
\scalebox{0.76}{
\begin{tabular}[t]{rcc}
\toprule
\multicolumn{3}{c}{\textbf{rotMNIST}}\\
\midrule
\multicolumn{1}{r}{Model \ \ \ } & Acc. (\%) & Params.  \\
\midrule
\zdsa\ \ \    & 96.37 & \multirow{5}{*}{44.67k} \\
\rfsa \ \ \    & 97.46 &  \\
\resa \ \ \   & 97.90 &  \\
\rtsa \ \ \   & \textbf{97.97} &  \\
\rssa \ \ \    & 97.66 &  \\
\midrule
\multicolumn{1}{r}{\zcnn$^{+}$} & 94.97 & 21.75k \\
\multicolumn{1}{r}{\rcnn$^{\dagger}$} & 98.21 & 77.54k \\
$\boldsymbol{\alpha}$-\rcnn$^{\dagger}$ & \textbf{98.31} & 73.13k \\
\bottomrule
\multicolumn{3}{l}{$^{+}${\rm \citet{cohen2016group}}}\\
\multicolumn{3}{l}{$^{\dagger}${\rm \citet{romero2020attentive}}}
\end{tabular}
}
\scalebox{0.76}{
\begin{tabular}[t]{rcc}
\toprule
\multicolumn{3}{c}{\textbf{CIFAR10}}\\
\midrule
Model\ \ \  & Acc. (\%) & Params.  \\
\midrule
\zdsa\ \ \  & 82.3  & \multirow{2}{*}{2.99m} \\
\msa\ \ \   & \textbf{83.72} &  \\
\midrule
\zcnn$^{+}$ & \textbf{90.56} & 1.37m \\
\bottomrule
\multicolumn{3}{l}{$^{+}${\rm \cite{cohen2016group}.}}
\end{tabular}
}
\scalebox{0.76}{
\begin{tabular}[t]{rcc}
\toprule
\multicolumn{3}{c}{\textbf{PatchCamelyon}}\\
\midrule
\multicolumn{1}{r}{Model \ \ \ } & Acc. (\%) & Params.  \\
\midrule
\zdsa\ \ \  & 83.04 & \multirow{4}{*}{205.66k} \\
\rfsa\ \ \   & 83.44 &  \\
\resa\ \ \  & 83.58 &  \\
\rmfsa\ \ \  & \textbf{84.76} &  \\
\midrule
\multicolumn{1}{r}{{\btt Z2\_CNN}$^{\dagger}$} & 84.07 & 130.60k \\
\multicolumn{1}{r}{{\btt R4\_CNN}$^{\dagger}$} & 87.55 & 129.65k \\
\multicolumn{1}{r}{{\btt R4M\_CNN}$^{\dagger}$} & 88.36 & 124.21k \\
\multicolumn{1}{r}{$\boldsymbol{\alpha_F}$-{\btt R4\_CNN}$^{\dagger}$} & 88.66 & 140.45k \\
\multicolumn{1}{r}{$\boldsymbol{\alpha_F}$-{\btt R4M\_CNN}$^{\dagger}$} & \textbf{89.12} & 141.22k \\
%\rmesa & - &  \\
\bottomrule
\multicolumn{3}{l}{$^{\dagger}${\rm \citet{romero2020attentive}.}}
\end{tabular}
}
\end{sc}
\end{small}
\end{table}

\textbf{CIFAR-10.} The CIFAR-10 dataset \citep{krizhevsky2009learning} consists of $60k$ real-world 32x32 RGB images uniformly drawn from 10 classes, divided into training, validation and test sets of $40k$, $10k$ and $10k$ images. Since reflection is a symmetry that appears ubiquitously in natural images, we compare \gsa\ equivariant to translation and reflection in this dataset (Tab.~\ref{tab:7_results}). Our results show that reflection equivariance indeed improves the classification performance of the model.

\textbf{PCam.} The PatchCamelyon dataset \citep{veeling2018rotation} consists of $327k$ 96x96 RGB image patches of tumorous/non-tumorous breast tissues extracted from Camelyon16 \citep{bejnordi2017diagnostic}. Each patch is labeled as tumorous if the central region (32x32) contains at least one tumour pixel. As cells appear at arbitrary positions and poses, we compare \gsa\ equivariant to translation, rotation and reflection (Tab.~\ref{tab:7_results}). Our results show that incorporating equivariance to reflection in addition to rotation, as well as providing finer group discretization, improve classification performance.
\vspace{-7mm}
\section{Discussion and Future Work}\label{sec:7_discussion}

Though \gsa\ perform competitively to \gcnn\ for some tasks, \gcnn\ still outperforms our approach in general. We conjecture that this is due to the harder nature of the optimization problem in \gsa\ and the carefully crafted architecture design, initialization and optimization procedures developed for CNNs over the years. 
Though our theoretical results indicate that  \gsa\ can be more expressive than \gcnn\ (Sec.~\ref{sec:7_conv_generalization}), further research in terms of design, optimization, stability and generalization is required. These are in fact open questions for self-attention in
general \citep{xiong2020layer, liu2020understanding, Zhao_2020_CVPR} and developments in this direction are of utmost importance.

The main drawback of our approach is the quadratic memory and time complexity typical of self-attention. 
This is an active area of research, e.g., \citet{kitaev2020reformer, wang2020linformer, zaheer2020big, choromanski2020rethinking} and we believe that efficiency advances to vanilla self-attention can be seamlessly integrated in \gsa. % to the transformer architecture could be incorporated in our group equivarient counter-part.
% 
% In future work, we want to explore how to alleviate this issue, e.g., \citet{kitaev2020reformer, wang2020linformer, zaheer2020big}.
Our theoretical results indicate that \gsa\ have the potential to become the standard solution for applications exhibiting symmetries, e.g., medical imagery.
% data, just as \sa{\btt-Nets} are for NLP tasks. 
% 
In addition, as self-attention is a set operation, \gsa\ provide straightforward solutions to set-like data types, e.g., point-clouds, graphs, symmetric sets, which may benefit from additional geometrical information, e.g., \citet{fuchs2020se, maron2020learning}.
 Finally, we hope our theoretical insights serve as a support point to further explore and understand the construction of equivariant maps for graphs and sets, which often come equipped with spatial coordinates: a type of positional encoding.\newpage% where the positional encoding is often provided as an additional feature value.% the  can be exploited Finally, we hope our theoretical insights 

%% file: chapters/8_waveletnets.tex
% path to figures directory
\graphicspath{{figures/8-waveletnets/}}

%=========================================================================

% \begin{savequote}[75mm]
% Nulla facilisi. In vel sem. Morbi id urna in diam dignissim feugiat. Proin molestie tortor eu velit. Aliquam erat volutpat. Nullam ultrices, diam tempus vulputate egestas, eros pede varius leo.
% \qauthor{Quoteauthor Lastname}
% \end{savequote}

\chapter{Scale-Translation Equivariant Learning From Raw Time-Series}\label{chapter:waveletnets}

\begin{flushright}
\textit{Based on the paper:}\break
\textit{Wavelet Networks: Scale-Translation Equivariant Learning From Raw Time-Series \citep{romero2020wavelet}}
\end{flushright}
%=========================================================================
\vspace{-7mm}
\section{Introduction}

Leveraging the symmetries inherent to specific data domains for the construction of statistical models, such as neural networks, has proven highly advantageous, by restricting the model to the family of functions that accurately describes the data. A prime example or this principle are Convolutional Neural Networks (CNNs) \citep{lecun1989backpropagation}. CNNs embrace the translation symmetries in visual data by restricting their mappings to a \textit{convolutional} structure. Convolutions possess a distinctive property called \textit{translation equivariance}: if the input is translated, the output undergoes an equal translation. This property endows CNNs with better data efficiency and generalization than unconstrained models like multi-layered perceptrons.

Group equivariant convolutional neural networks (G-CNNs) \citep{cohen2016group} extend equivariance to more general symmetry groups through the use of \textit{group convolutions}. Group convolutions are \textit{group equivariant}: if the input is transformed by the symmetries described by the group, e.g., scaling, the output undergoes an equal transformation. Equivariance to larger symmetry groups endows G-CNNs with increased data efficiency and generalization on data exhibiting these symmetries. Existing group equivariance research primarily focuses on symmetries found in visual data, e.g., planar rotations, planar scaling, etc \citep{weiler2018learning, worrall2019deep, romero2020attentive}, and more recently, on 3D symmetries, e.g., for spherical and molecular data \citep{thomas2018tensor, fuchs2020se, satorras2021n}. Yet, an important category remains underexplored, which also exhibits symmetries: \textit{time-series}. Notably, their translation symmetry is a cornerstone in signal processing and system analysis, e.g., Linear Time-Invariant (LTI) systems.

In this work, we bridge this gap by constructing neural networks that embrace the symmetries inherent to time-series. We begin by asking: \enquote{\textit{What symmetries are inherently present in time-series?}} We identify two fundamental symmetries --\textit{scale and translation}--, whose combination elucidate several phenomena observed in time-series, e.g., temporal translations, phase shifts, temporal scaling, resolution changes, pitch shifts, seasonal occurrences, etc. By leveraging group convolutions equivariant to the \textit{scale-translation} group, we construct neural architectures such that when the input undergoes translation, scaling or a combination of the two, all intermediate layers will undergo an equal transformation in a hierarchical manner, akin to the methods proposed by \citet{sosnovik2020scaleequivariant, zhu2022scaling} for visual data. Interestingly, we observe that constructing convolutional layers equivariant to scale and translation results in layers that closely resemble the \textit{wavelet transform}. However, we find that in order to preserve these symmetries consistently across the whole network, the output of each layer must be processed by a layer that also behaves like the wavelet transform. This approach substantially deviates from common approaches that rely on spectro-temporal representations, e.g., the wavelet transform, which compute spectro-temporal representations once and pass their response to a 2D CNN for further processing. 

Inspired by the resemblance of scale-translation group equivariant convolutions with the wavelet transform, we term our scale-translation equivariant networks for time-series processing \textit{Wavelet Networks}. Extensive empirical results show that Wavelet Networks consistently outperform conventional CNNs operating on raw waveforms, and match strongly engineered spectogram-based approaches, e.g., on Mel-spectrograms, across several tasks and time-series types, e.g., audio, environmental sounds, electrical signals. To the best of our best knowledge, we are first to propose scale-translation equivariant neural networks for general time-series processing tasks.

\vspace{-7mm}
\section{Related Work}
\begin{figure}
    \centering
        \centering
        \begin{subfigure}{0.38\textwidth}
        \captionsetup{justification=centering}
        \includegraphics[width=\textwidth]{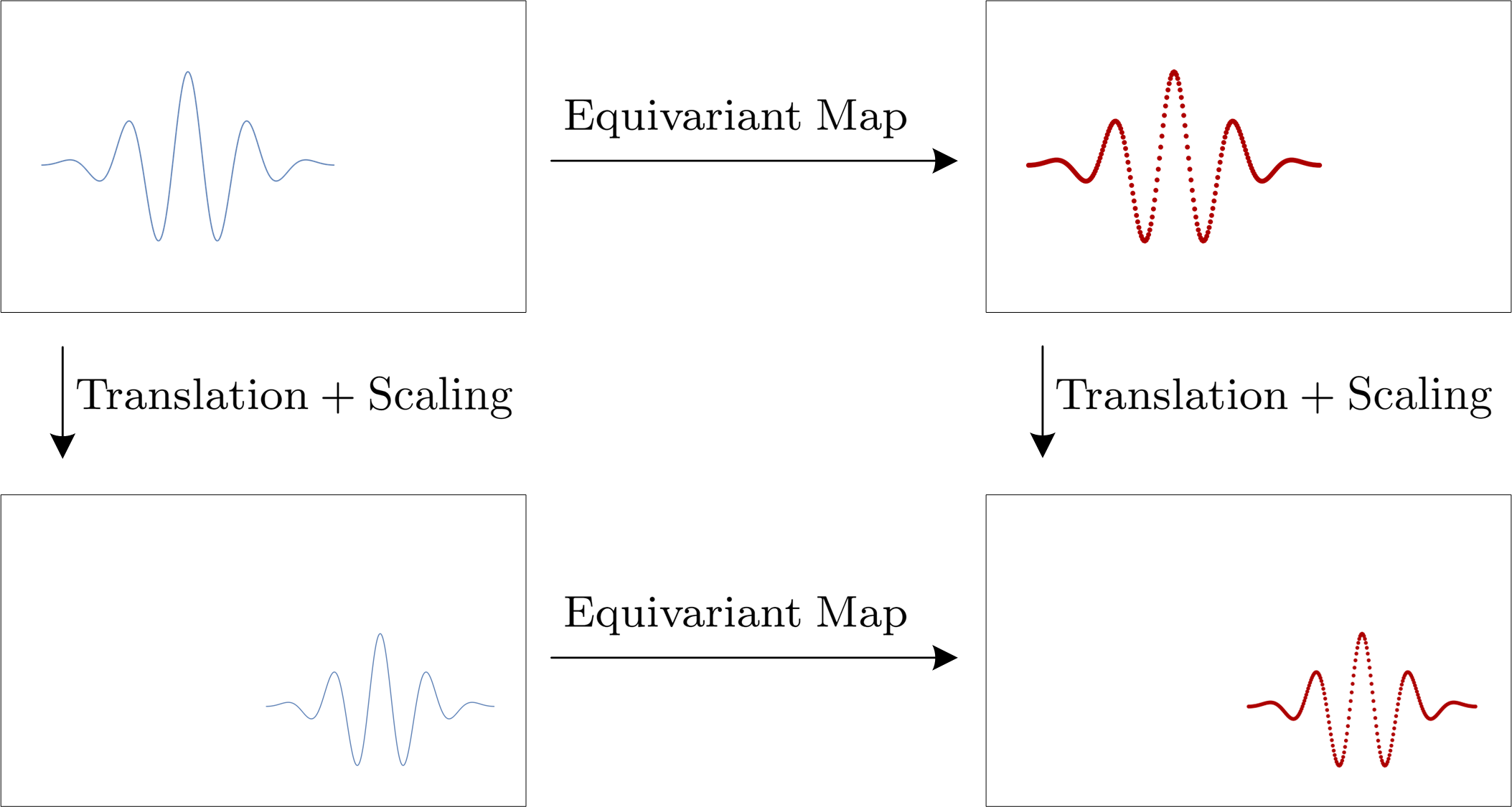}
        \caption{
        \vspace{-1mm}}
        \label{fig:equivariant_map}
    \end{subfigure}
    \hfill
        \begin{subfigure}{0.3615\textwidth}
        \captionsetup{justification=centering}
        \includegraphics[width=\textwidth]{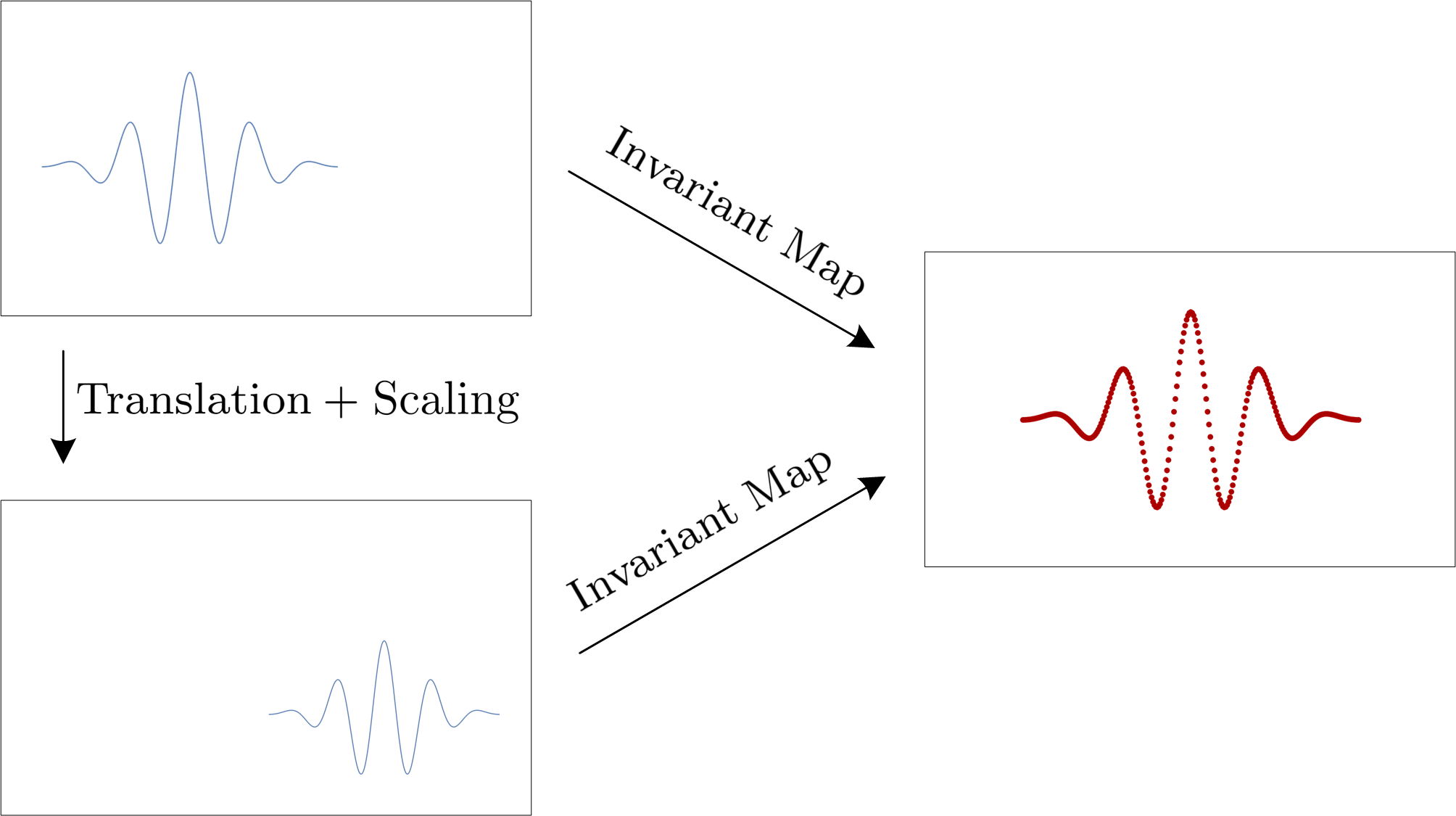}
        \caption{
        \vspace{-1mm}}
        \label{fig:invariant_map}
    \end{subfigure}
    \hfill
    \begin{subfigure}{0.22\textwidth}
        \captionsetup{justification=centering}
        \includegraphics[width=\textwidth]{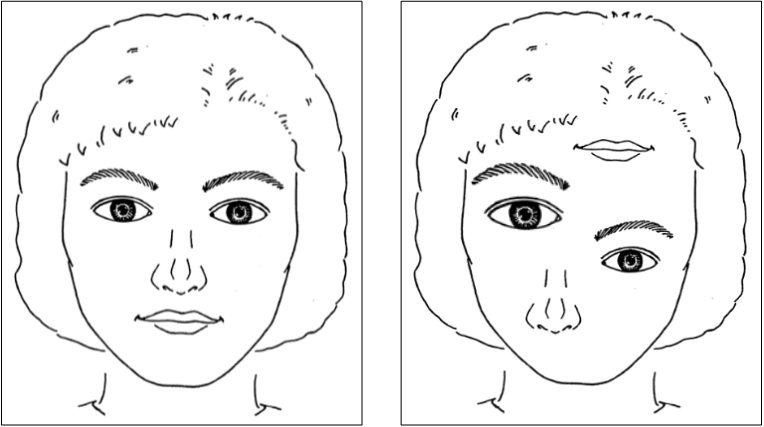}
        \caption{
        \vspace{-1mm}}
        \label{fig:faces_example}
    \end{subfigure}
    \vspace{-1.5mm}
    \caption{Equivariance, invariance and their impact on the hierarchical representations. In a group equivariant mapping, when the input is transformed by a group transformation, its output undergoes an equivalent transformation (Fig.~\ref{fig:equivariant_map}). In contrast, in group invariant maps, the output remains unchanged for all group transformations of the input (Fig.~\ref{fig:invariant_map}). This distinction holds significant implications in the construction of hierarchical feature representations. For example, a face recognition system built upon invariant eye, nose and mouth detectors would be unable to set the portraits in Fig.~\ref{fig:faces_example} apart. However, by leveraging equivariant mappings, information about the input transformations can be used to distinguish these portraits effectively. In essence, in contrast to equivariant maps, invariant maps permit senseless pattern combinations resulting for overly restraining constraints in their design.}
    \label{fig:invariance_equivariance}
\end{figure}
\textbf{Learning from raw time-series.}
Several end-to-end learning approaches for time-series exist \citep{dieleman2014end, dieleman2016exploiting, dai2017very, rethage2018wavenet, stoller2018wave}. Given the considerable high-dimensionality of time-series, existing works focus on devising techniques with parameter- and compute-efficient large memory horizons \citep{romero2022ckconv, goel2022s}. Due to small effective memory horizons and long training times, Recurrent Neural Networks (RNNs) \citep{rumelhart1985learning} have gradually been overshadowed by CNN backbones \cite{bai2018empirical}.

While CNNs are equivariant to translations, they do not inherently incorporate a distinct notion of scale. Although methods involving layer-wise multi-scale representations have been proposed, e.g., \citet{zhu2016learning, lu2019deep, von2019multi, guizzo2020multi}, these layers are not scale equivariant. As a result, networks incorporating them struggle to maintain consistent scale information across layers.

\textbf{Group-invariant time-series learning.} Learning invariant representations from raw speech and sound has been extensively studied in past. Scattering operators \citep{mallat2012group, bruna2013invariant} construct group \textit{invariant} feature representations that can be used to construct neural architectures invariant to scale and translation \citep{anden2014deep, peddinti2014deep, salamon2015feature}. 
In contrast to the \textit{invariant} feature representations developed by these works, Wavelet networks construct \textit{equivariant} feature representations. Since group equivariance is a generalization of group invariance (Fig.~\ref{fig:invariant_map}, Sec.~\ref{sec:equiv_invariance}), Wavelet Networks accommodate a broader functional family than previous works, while still upholding scale and translation preservation. Notably, equivariant methods shown superior performance compared to invariant methods across several tasks, even for intrinsically invariant tasks like classification \citep{cohen2016group, zaheer2017deep, maron2018invariant}. This phenomenon stems from the hierarchical form in which neural networks extract features. Enforcing invariance early in the feature extraction process imposes an overly restrictive constraint in the resulting models (Fig.~\ref{fig:faces_example}).

\textbf{Group-equivariant time-series learning.} The concept of equivariant learning for time-series has mainly been explored in two contexts. \citet{zhang2015discriminative} propose to learn feature representations equivariant to vocal tract length changes --an inherent symmetry of speech. However, vocal tract length changes do not conform to the mathematical definition of a group, making their equivariance definition only approximate. Interestingly, vocal tract length changes can be characterized by specific (scale, translation) tuples. Consequently, considering scale-translation equivariance implicitly describes vocal tract length changes as well as many other symmetries encountered in audio, speech and other time-series modalities. In addition, group equivariance has also been explored for the modeling and forecast of dynamical systems --an specific type of time-series \citep{wang2020incorporating, wang2020towards, walters2020trajectory}. Contrary to these methods, our work explores scale-translation equivariance for general time-series processing tasks, providing a framework that can enhance learning and generalization across diverse time-series applications.

\vspace{-7mm}
\section{Background}
This work assumes a basic familiarity with the concepts of a group, a subgroup and a group action. For those not familiar with these terms, we introduce them in Appx.~\ref{appx:group_and_action}.

\vspace{-7mm}
\subsection{Group equivariance, group invariance and symmetry preservation} \label{sec:equiv_invariance}
% \begin{tcolorbox}[enhanced, frame hidden, sharp corners, boxsep=0pt, before skip=5pt, after skip=8pt]
% \textbf{Group.} A group is an ordered pair $(\gG, \cdot)$ where $\gG$ is a set and $\cdot: \gG \times \gG \rightarrow \gG$ is a binary operation on $\gG$, such that \emph{(i)} the set is closed under this operation, \emph{(ii)} the operation is associative, i.e., $(g_1 \cdot g_2) \cdot g_3 = g_1 \cdot (g_2 \cdot g_3)$, $g_1, g_2, g_3 \in \gG$, \emph{(iii)} there exists an identity element $e \in \gG$ such that $\forall g \in \gG$ we have $e \cdot g = g \cdot e = g$, and \emph{(iv)} for each $g \in \gG$, there exists an inverse $g^{-1}$ such that $g \cdot g^{-1} = e$.
% \end{tcolorbox}
% %
% \begin{tcolorbox}[enhanced, frame hidden, sharp corners, boxsep=0pt, before skip=5pt, after skip=8pt]
% \textbf{Group action.}
% Let $\gG$ be a group and $\gX$ be a set. The (left) group action of $\gG$ on $\gX$ is a function 
% \begin{equation}
% 
% 
% \gA: \gG \times \gX \rightarrow \gX, \ \ \ \gA_{g}: x \rightarrow x',
% \end{equation}
% such that for any $g_1$, $g_2 \in \gG$, $\gA_{g_2 g_1} {=} \gA_{g_2} \circ \gA_{g_1}$.  
% In other words, the action of $\gG$ on $\gX$ describes how the elements in the set $x \in \gX$ are transformed by elements $g \in \gG$. For brevity, $\gA_{g}(x)$ is written as $gx$.
% \end{tcolorbox}
\textbf{Group equivariance.} Group equivariance is the property of a mapping to respect the transformations in a group. We say that a map is equivariant to a group if a transformation of the input by elements of the group leads to an equivalent transformation of the output (Fig.~\ref{fig:equivariant_map}). Formally, for a group $\gG$ with elements $g \in \gG$ acting on a set $\gX$, and a mapping $\phi: \gX \rightarrow \gX$, we say that $\phi$ is equivariant to $\gG$ if:
\begin{equation}
    \phi(gx) = g \phi(x), \quad \forall x \in \gX, \forall g \in \gG.
\end{equation}
For example, the convolution of a signal $f: \sR \rightarrow \sR$ and a kernel $\psi: \sR \rightarrow \sR$ is \textit{equivariant to the group of translations} --or \textit{translation equivariant}-- because when the input is translated, its convolutional descriptors are equivalently translated, i.e., $(\psi * \gL_t f){=} \gL_t (\psi *f)$, with $\gL_t$ a translation operator by $t$: $\gL_tf(x){=}f(x{-}t)$.

\textbf{Group invariance.} Group invariance is a special case of group equivariance in which the output of the map is equal for all transformations of the input (Fig.~\ref{fig:invariant_map}). Formally, for a group $\gG$ with elements $g \in \gG$ acting on a set $\gX$, and a mapping $\phi: \gX \rightarrow \gX$, we say that $\phi$ is invariant to $\gG$ if:
\begin{equation}
    \phi(gx) = \phi(x), \quad \forall x \in \gX, \forall g \in \gG.
\end{equation}
\underline{\textit{Relation to symmetry preservation}}. A symmetry-preserving mapping preserves the symmetries of the input. That is, if the input has certain symmetries, e.g., translation, rotation, scale, these symmetries will also be present in the output. Since symmetries are mathematically described as groups, it follows that group equivariant mappings preserve the symmetries of the group to which the mapping is equivariant. In contrast, invariant mappings \textit{do not} preserve symmetry, as they remove all symmetric information from the input. 

\vspace{-7mm}
\subsection{Symmetry-preserving mappings: Group and lifting convolutions}
When talking about (linear) symmetry-preserving mappings, we are obliged to talk about the group convolution. Previous work has shown that group convolutions are the most general class of group equivariant linear maps \citep{ cohen2019general}. Hence, it holds that any linear equivariant map is in fact a group convolution.

\textbf{Group convolution.}  Let $f: \gG \rightarrow \sR$ and $\psi: \gG \rightarrow \sR$ be a scalar-valued signal and convolutional kernel defined on a group $\gG$. Their group convolution ($*_{\gG}$) is given by:
\begin{equation}
    (f *_{\gG} \psi)(g) = \hspace{-1mm} \int_{\gG} f(\gamma)\gL_{g}\psi(\gamma)\ \du \mu_\gG(\gamma) =  \hspace{-1mm}  \int_{\gG} f(\gamma)\psi\left(g^{-1}\gamma\right)\ \du \mu_\gG(\gamma).\label{eq:gconv}
\end{equation}
where $g, \gamma \in \gG$, $\gL_{g}\psi(\gamma) {=}\psi\left(g^{-1}\gamma\right)$ is the action of the group $\gG$ on the kernel $\psi$, and $\mu_\gG(\gamma)$ is the (invariant) Haar measure of the group $\gG$ for $\gamma$. Notably, the group convolution generalizes the translation equivariance of convolutions to general groups. The group convolution is equivariant in the sense that for all $\gamma, g \in \gG$,
\begin{equation}
     \gL_g(f *_{\gG} \psi)(\gamma) = (\gL_g f *_{\gG} \psi)(\gamma), \ \text{with} \ \gL_{g}f(\gamma) {=}f\left(g^{-1}\gamma\right) .\label{eq:groupequiv_constraint}
\end{equation}
\textbf{The lifting convolution.} In practice, the input signals $f$ might not be readily defined on the group of interest $\gG$, but on a sub-domain thereof $\gX$, i.e., $f: \gX \rightarrow \sR$. For example, time-series are defined on $\sR$ although we might want to consider larger groups such as the scale-translation group. Hence, we require a symmetry-preserving mapping from $\gX$ to $\gG$ that \textit{lifts} the input signal to $\gG$ to use group convolutions. This operation is called a \textit{lifting convolution}. Formally, with $f: \gX \rightarrow \sR$ and $\psi: \gX \rightarrow \sR$ a scalar-valued signal and convolutional kernel defined on $\gX$, and $\gX$ a sub-group of $\gG$, the lifting convolution ($*_{\gG\uparrow}$) is a mapping from functions on $\gX$ to functions on $\gG$ defined as:
\begin{equation}
     (f *_{\gG\uparrow} \psi)(g) = \int_\gX f(x)\gL_{g}\psi(x)\ \du \mu_\gG(x) = \int_\gX f(x)\psi(g^{-1}x)\ \du \mu_\gG(x) 
\end{equation}
The lifting convolution is also group equivariant. That is, $\gL_g(f *_{\gG\uparrow} \psi) {=} (\gL_g f *_{\gG\uparrow} \psi)$.
\vspace{-7mm}
\section{The problem of learning 2D convolutional kernels on the time-frequency plane}
CNNs have been a major breakthrough in computer vision, yielding startling results in countless applications. Due to their success, several works have proposed to treat \textit{spectro-temporal representations} --representations on the time-frequency plane-- as 2D images and learn 2D CNNs on top. In this section, we delve into the differences between visual and spectro-temporal representations, and assess the suitability of training 2D CNNs directly on top of spectro-temporal representations. Our analysis suggest that treating spectro-temporal representations as images and learning 2D CNNs on top might not be adequate for effective time-series learning.

To enhance clarity, we define spectro-temporal representations in separate gray boxes throughout the section to avoid interrupting the reading flow. Those already familiar with these concepts may skip these boxes.
\begin{tcolorbox}[enhanced, breakable, frame hidden, sharp corners, boxsep=0pt, before skip=5pt, after skip=8pt]
\textbf{Spectro-temporal representations.} Let $f(t) \in \ltwo(\sR)$ be a square integrable function on $\sR$. An spectro-temporal representation $\Phi[f](t, \omega): \sR^2 \rightarrow \sC$ of $f$ is constructed by means of a \textit{linear time-frequency transform} $\Phi$ that correlates the signal $f$ with a dictionary $\gD$ of localized \textit{time-frequency atoms} $\mathcal{D}{=}\{\phi_{t, \omega}\}_{t \in \sR, \omega \in \sR}$, $\phi_{t, \omega}: \sR \rightarrow \sC$ of finite energy and unitary norm, i.e., $\phi_{t, \omega} \in \ltwo(\sR)$, $\| \phi_{t, \omega}\|^{2}{=}1$, $\forall t \in \sR, \omega \in \sR$. The resulting spectro-temporal representation $\Phi[f]$ is given by:
\begin{equation}\label{eq:time_freq_transf}
    \Phi[f](t, \omega) = \langle f, \phi_{t, \omega}\rangle = \int_\sR f(\tau)\phi^{*}_{t, \omega}(\tau) \, {\rm d}\tau,
\end{equation}
where $\phi^{*}$ depicts the complex conjugate of $\phi$, and $\langle \cdot, \cdot \rangle$ the dot product of its arguments. By selecting different parameterizations of the time-frequency components $\phi_{t, \omega}$, spectro-temporal representations with different properties can be obtained.
\end{tcolorbox}
\vspace{-7mm}
\subsection{Fundamental differences between visual representations and spectro-temporal representations}\label{sec:visualdata_timefreq} 
\begin{figure}[t]
    \centering
    \begin{subfigure}{0.4\textwidth}
    \centering
    \includegraphics[width=0.75\textwidth]{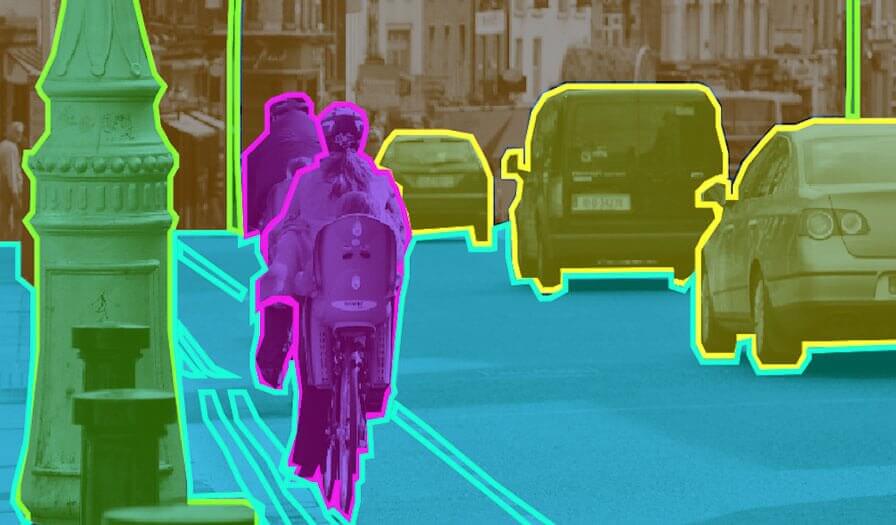}
    \end{subfigure}
    \hspace{5mm}
    \begin{subfigure}{0.55\textwidth}
    \centering
    \includegraphics[width=\textwidth]{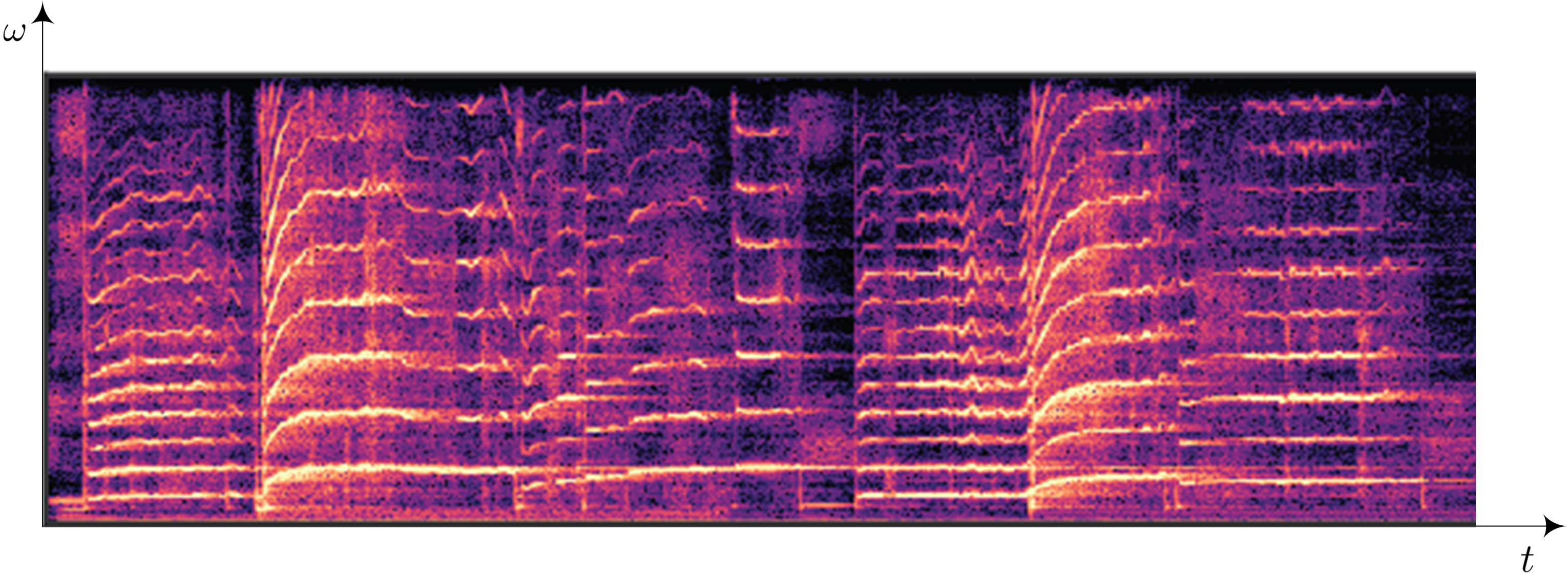}
    \end{subfigure}
    \vspace{-2mm}
    \caption{Locality of visual and auditory objects. Whereas visual objects are local (left), auditory objects are not. The latter often cover large parts of the frequency axis in a sparse manner (right).
    \vspace{-6mm}}
    \label{fig:locality}
\end{figure}
\begin{figure}[t]
    \centering
    \begin{subfigure}{0.5\textwidth}
    \centering
    \includegraphics[width=0.8\textwidth]{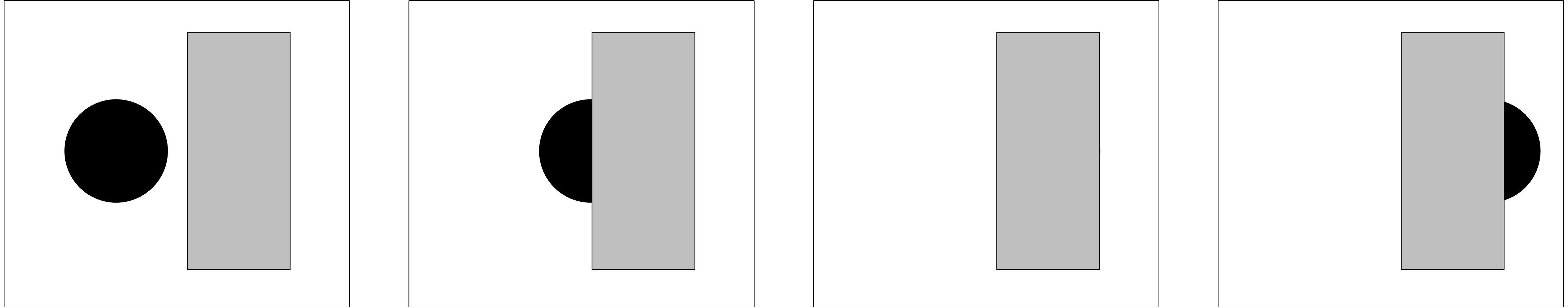}
    \end{subfigure}
    \hspace{5mm}
    \begin{subfigure}{0.3\textwidth}
    \centering
    \includegraphics[width=0.8\textwidth]{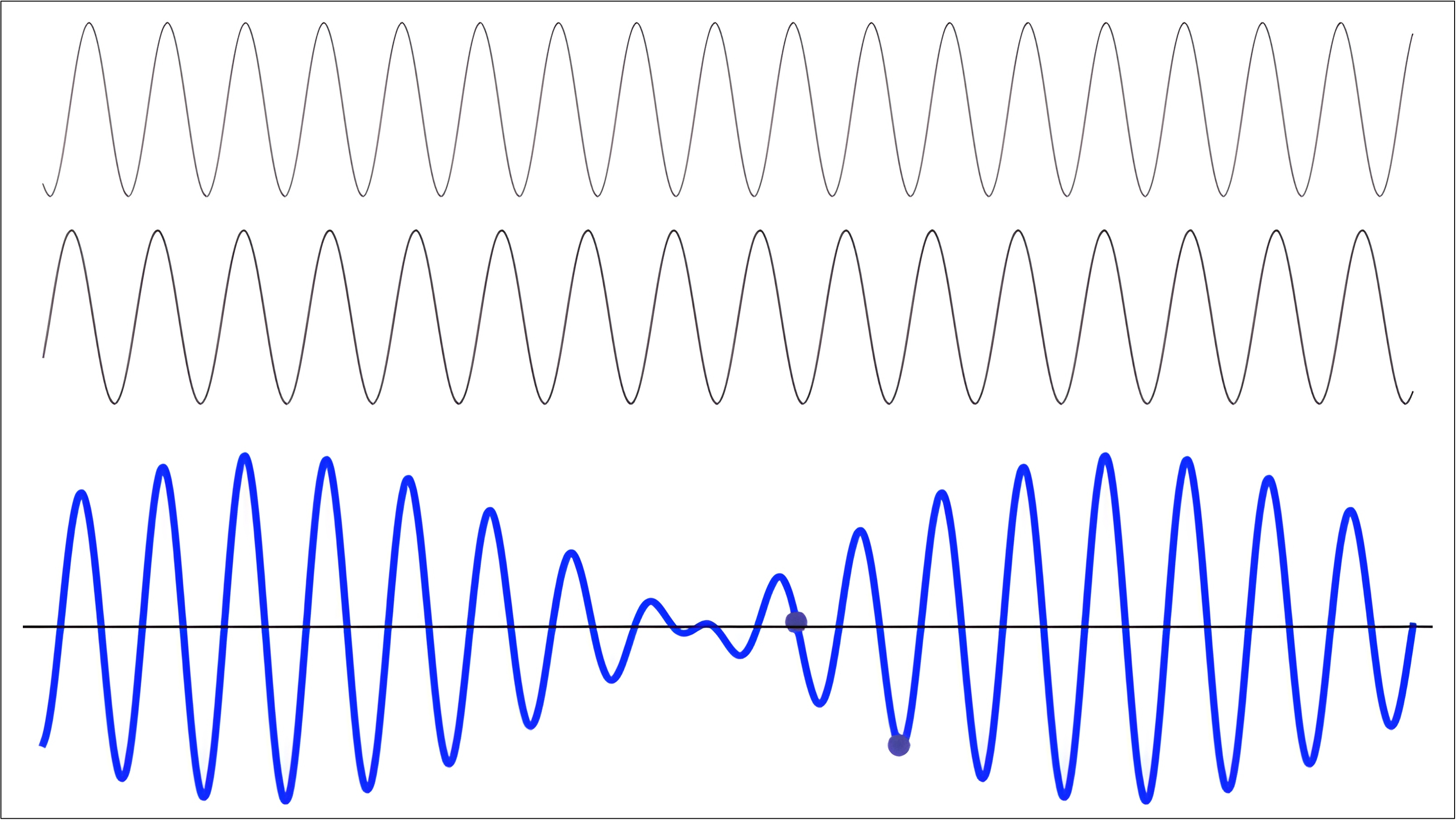}
    \end{subfigure}
    \vspace{-2mm}
    \caption{ Occlusion and superposition. Visual objects occlude each other when they appear simultaneously at a given position (left). Auditory objects, instead, superpose at all shared positions (right).}
    \label{fig:occlusion}
\end{figure}
There exist two fundamental distinctions between visual data and spectro-temporal representations, which are universal to all spectro-temporal representations: (\textit{i}) locality and (\textit{ii}) transparency. Unlike visual data, auditory signals exhibit strong \textit{non-local} characteristics. Auditory signals consist of auditory objects, e.g., spoken words, which contain components resonating at multiple non-local frequencies known as the \textit{harmonics of the signal}. Consequently, the spectro-temporal representations of auditory objects often occupy a significant portion of the time-frequency plane --particularly along the frequency axis ($\omega$)-- in a sparse manner (Fig.~\ref{fig:locality}).
Furthermore, when considering auditory signals comprising multiple auditory objects, these objects exhibit a phenomenon known as \textit{superposition}. This property is notably different from visual data, where visual objects in the same location \textit{occlude} one another, resulting in only the object closest to the camera being visible (Fig.~\ref{fig:occlusion}). This inherent property of sound is referred to as \textit{transparency}.

\vspace{-7mm}
\subsection{The problem of learning 2D kernels on short-time Fourier\\ spectro-temporal representations}
\begin{figure}
    \centering
    \begin{subfigure}{0.33\textwidth}
    \centering
    \captionsetup{justification=centering}
        \includegraphics[width=\textwidth]{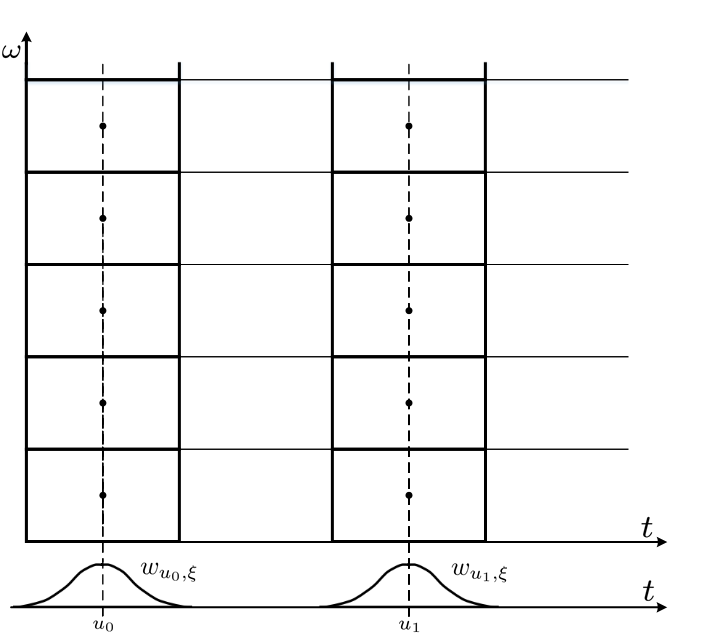}
        \caption{}
        \label{fig:wind_four}
    \end{subfigure}
    \hspace{5mm}
    \begin{subfigure}{0.33\textwidth}
    \captionsetup{justification=centering}
    \centering
        \includegraphics[width=1.02\textwidth]{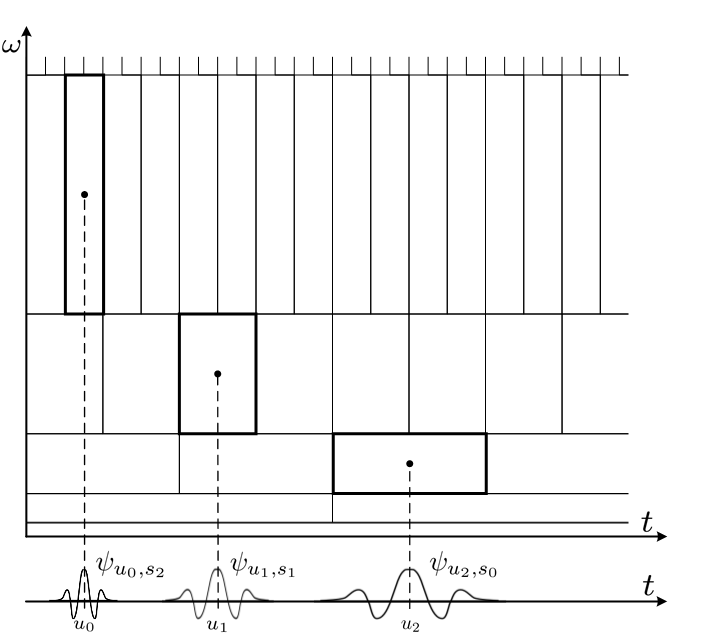}
        \caption{}
        \label{fig:wavelet}
    \end{subfigure}
    \vspace{-2mm}
        \caption{Tiling of the time-frequency plane for the short-time Fourier transform (Fig.~\ref{fig:wind_four}) and the wavelet transform (Fig.~\ref{fig:wavelet}). The short-time Fourier transform divides the time-frequency plane in tiles of equal resolution. This makes it adequate for signals without transient behaviour. The Wavelet transform, on the other hand, divides the time-frequency plane on tiless of changing spectro-temporal resolution. This allows it to represent detect highly localized events both on time and frequency.\label{fig:freq_time_tiling}}
\end{figure}

The short-time Fourier transform constructs a representation in which a signal is decomposed in terms of its correlation with time-frequency atoms of constant time and frequency resolution. As a result, it is effective as long as the signal $f$ does not exhibit \textit{transient behavior}  --components that evolve quickly over time-- with some waveform structures being very localized in time and others very localized in frequency. 
\begin{tcolorbox}[enhanced, frame hidden, sharp corners, boxsep=0pt, before skip=5pt, after skip=10pt]
\textbf{The short-time Fourier transform.} The short-time Fourier transform $\gS$ --also called \textit{windowed Fourier transform}-- is a linear time-frequency transform that uses a dictionary of time-frequency atoms $\phi_{t, \omega}(\tau){=}w(\tau - t) \eu^{-\iu \omega \tau}$, $t {\in} \sR$, $\omega {\in} \sR$, constructed with a symmetric window $w(\tau)$ of local support shifted by $t$ and modulated by the frequency $\omega$. The spectro-temporal representation $\gS[f]$ is given by:
\begin{equation}\label{eq:time_freq_transf}
    \gS[f](t, \omega) = \langle f, \phi_{t, \omega}\rangle = \int_\sR f(\tau)\phi^{*}_{t, \omega}(\tau) \, {\rm d}\tau = \int_\sR f(\tau)w(\tau - t) \eu^{-\iu \omega \tau} \, {\rm d}\tau.
\end{equation}
Intuitively, the short-time Fourier transform divides the time-frequency plane in tiles of equal resolution, whose value is given by the correlation between $f$ and the time-frequency atom $\phi_{t, \omega}$ (Fig.~\ref{fig:wind_four}).
\end{tcolorbox}

Nevertheless, decades of research in psychology and neuroscience have shown that humans largely rely in the transient behavior of auditory signals to distinguish auditory objects \citep{cherry1953some, van1975temporal, moore2012properties}. In addition, it has been shown that the human auditory system has high spectral resolution at low-frequencies and high temporal resolution at higher frequencies \citep{stevens1937scale, santoro2014encoding, bidelman2014spectrotemporal}. For example, a semitone at the bottom of the piano scale ($\sim\hspace{-1mm}30\si{\hertz}$) is of about $1.5$\si{\hertz}, while at the top of the musical scale ($\sim\hspace{-1mm}5\si{\kilo\hertz}$) it is of about $200$\si{\hertz}. These properties of the human auditory signal largely contrast both with (\textit{i}) the inability of the short-time Fourier transform to detect transient signals, as well as with (\textit{ii}) its constant spectro-temporal resolution.

To account for these differences, improved spectro-temporal representations on top of the short-time Fourier transform have been proposed such as log-Mel spectrograms \citep{stevens1937scale, furui1986speaker}. These developments revolve around transforming the frequency axis of the short-time Fourier transform in a logarithmic scale, thereby compressing the frequency axis and better aligning with the spectro-temporal resolution of the human auditory system. Consequently, this adjustment enables local structures, e.g., 2D convolutional kernels, to better capture non-local relationships \citep{ullrich2014boundary, choi2016automatic, xu2018large}. However, despite improvements, these spectro-temporal representations remain \textit{incomplete} due to their inability to modify the constant temporal resolution of the short-time Fourier transform.

\vspace{-7mm}
\subsection{The problem of learning 2D kernels on Wavelet\\ spectro-temporal representations}
In contrast to the short-time Fourier transform, the Wavelet transform constructs a spectro-temporal representation in terms of correlations with time-frequency atoms, \textit{whose time and frequency resolution change}. As a result, the resulting decomposition of the time-frequency plane allows the wavelet transform to \textit{correctly describe signals with transient behaviour with localized components both on time and frequency} (Fig.~\ref{fig:wavelet}).
\begin{tcolorbox}[enhanced, frame hidden, sharp corners, breakable, boxsep=0pt, before skip=5pt, after skip=8pt]
\textbf{The wavelet transform.} The wavelet transform $\gW$ is a linear time-frequency transform that uses a dictionary of time-frequency atoms $\phi_{t, \omega}(\tau){=}\frac{1}{\sqrt{\omega}}\psi\left(\frac{\tau - t}{\omega}\right)$, $t \in \sR$, $\omega \in \sR_{\geq 0}$. The function $\psi_{t, \omega}$ is called a \textit{Wavelet} and satisfies the properties of having zero mean, i.e., $\int \psi_{t, \omega}(\tau)\du \tau {=} 0$,s and being unitary, i.e., $\| \psi_{t, \omega}\|^{2}{=}1$, for any $t \in \sR$, $\omega \in \sR_{\geq 0}$. The resulting spectro-temporal representation $\gW[f]$ is given by:
\begin{equation}\label{eq:wavelet_transform}
    \gW[f](t, \omega) = \langle f, \phi_{t, \omega}\rangle = \int_\sR f(\tau)\phi^{*}_{t, \omega}(\tau) \, {\rm d}\tau = \int_\sR f(\tau) \frac{1}{\sqrt{\omega}}\psi\left(\frac{\tau - t}{\omega}\right) \, {\rm d}\tau.
\end{equation}
Intuitively, the Wavelet transform divides the time-frequency plane in tiles of different resolutions, with high frequency resolution and low spatial resolution at low frequencies, and low frequency resolution and high spatial resolution for high frequencies (Fig.~\ref{fig:wavelet}).\footnote{Importantly, it is not possible to have high frequency and spatial resolution at the same time due to the \textit{uncertainty principle} \citep{gabor1946theory}. It states that the joint time-frequency resolution of spectro-temporal representations is limited by a minimum surface $\sigma_{\phi, t} \sigma_{\phi, \omega} \geq \frac{1}{2}$, with $\sigma_{\phi, t}$, $\sigma_{\phi, \omega}$ the spread of the time-frequency atom $\phi$ on time and frequency.}  
\end{tcolorbox}
Interestingly, the modus operandi of the wavelet transform \textit{perfectly aligns} with the spectro-temporal resolution used by the human auditory system for the processing of auditory signals. Nevertheless, despite this resemblance, training local 2D structures, e.g., convolutional kernels, directly on the wavelet transform's output stil falls short in addressing the non-local, transparent characteristics inherent in auditory signals. Consequently, researchers have devised many strategies to overcome these challenges, e.g., by defining separable kernels that span large memory horizons along frequency and time independently \citep{pons2019randomly} or by prioritizing learning along the harmonics of a given frequency \citep{Zhang2020Deep}. 

As shown in the next section, a better alternative arises from considering the symmetries appearing in time-series data. Starting from this perspective, we are led to scale-translation equivariant mappings and find striking relationships between these family of mappings and the wavelet transform. Nevertheless, our analysis indicates that \textit{all layers within a neural network should be symmetry preserving} --a condition not met by the methods depicted in this section. By doing so, we devise neural architectures, whose convolutional layers process the output of previous layers in a manner akin to the wavelet transform. As a result, each convolutional layer performs spectro-temporal decompositions of the input in terms of localized time-frequency atoms able to process global and localized patterns both on time and frequency.

\vspace{-7mm}
\section{Wavelet networks: Scale-translation equivariant learning from raw waveforms}
We are interested in mappings that preserve the scale and translation symmetries of time-series. In this section, we start by tailoring lifting and group convolutions to the scale-translation group. Next, we outline the general form of Wavelet Networks and make concrete practical considerations for their implementation. At the end of this section, we formalize the relationship between Wavelet Networks and the wavelet transform, and provide a thorough analysis on the equivariance properties of common spectro-temporal transforms.  

\vspace{-7mm}
\subsection{Scale-translation preserving mappings: group convolutions on the\\ scale-translation group}
 We are interested in mappings that preserve scale and translation. By imposing equivariance to the \textit{scale-translation} group, we guarantee that if input patterns are scaled, translated, or both, their feature representations will transform accordingly.

\textbf{The scale-translation group.} From a mathematical perspective scale and translational symmetries are described by the affine \textit{scale-translation group} $\gG{=}\sR \rtimes \sR_{\geq0}$, which emerges from the semi-direct product of the translation group $\gT{=}(\sR, +)$ and the scale group $\gS{=}(\sR_{\geq0}, \times)$ acting on $\sR$. As a result, we have that the resulting group product is given by $ g \cdot \gamma {=} (t, s) \cdot (\tau, \varsigma) {=} (t + s \tau, s \cdot \varsigma)$, with $t, \tau \in \sR$ and $s, \varsigma \in \sR_{\geq0}$. In addition, by solving $g^{-1} \cdot g {=}e$, we obtain that the inverse of a group element $g{=}(t, s)$ is given by $g^{-1}{=}s^{-1}(-t, 1)$.
\begin{tcolorbox}[enhanced, frame hidden, sharp corners, boxsep=0pt, before skip=5pt, after skip=10pt]
\textbf{Semi-direct product and affine groups.} When treating data defined on $\sR^{d}$, one is mainly interested in the analysis of groups of the form $\gG {=} \sR^{d} \rtimes \gH$ resulting from the \textit{semi-direct product} ($\rtimes$) between the translation group ($\sR^{d}, +$) and an arbitrary (Lie) group $\gH$ acting on $\sR^{d}$, e.g., rotation, scale, etc. This kind of groups are called \textit{affine groups} and their group product is defined as:
\begin{equation}\label{eq:affine_product}
    g_{1} \cdot g_{2} = (x_{1}, h_{1}) \cdot (x_{2}, h_{2}) = (x_{1} + \gA_{h_1}(x_{2}), h_1 \cdot h_2), 
\end{equation}
with $g_1 {=} (x_{1}, h_{1})$, $g_{2}{=}(x_{2}, h_{2}) \in \gG$, $x_{1}$, $x_{2} \in \sR^{d}$ and $h_{1}, h_{2} \in \gH$. $\gA$ denotes the action of $\gH$ on $\sR^{d}$.
\end{tcolorbox}
\sidecaptionvpos{figure}{c}
\begin{SCfigure*}[10]
    \centering
    \includegraphics[width=0.55\textwidth]{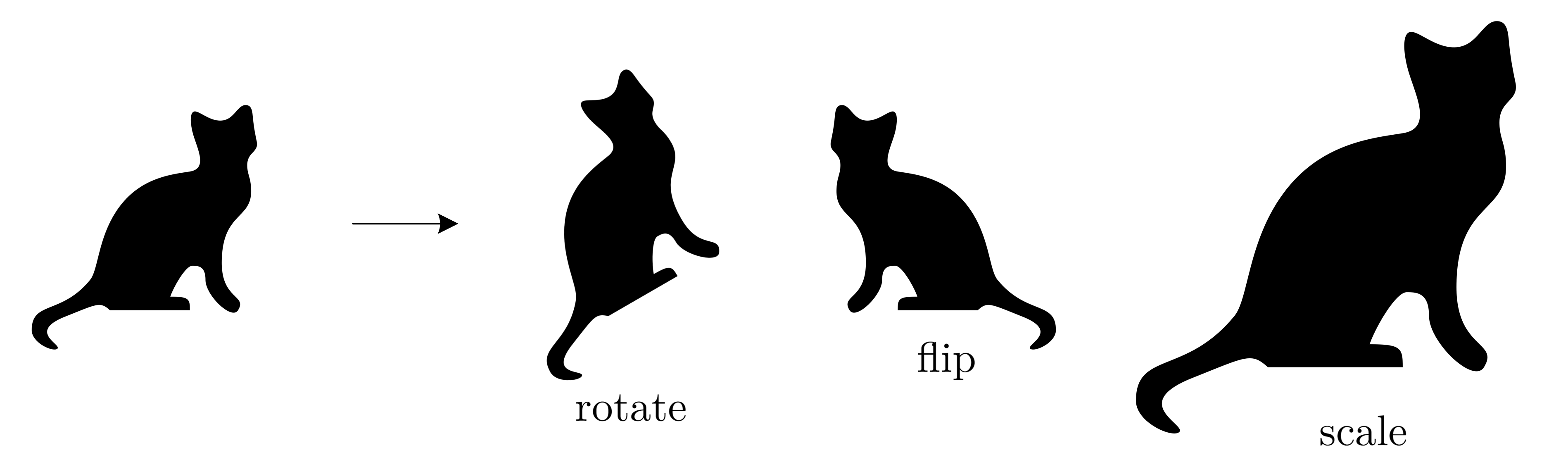}
    \caption{The action of unimodular and non-unimodular groups. Most unimodular groups, e.g., rotation, mirroring, keep the volume of the objects they act upon intact. In contrast, non-unimodular groups, e.g., scaling, change it through their action.}
    \label{fig:unimodular}
\end{SCfigure*}

\textbf{Unimodular and non-unimodular groups.} Unimodular groups, such as rotation, translation and mirroring, are groups whose action keeps the volume of the objects on which they act intact (Fig.~\ref{fig:unimodular}). Recall that a group convolution performs an integral over the whole group (Eq.~\ref{eq:gconv}). Hence, for its result to be invariant over different group actions, it is required for the Haar measure to be equal for all elements of the group --therefore the name \textit{invariant} Haar measure. Since the action of (most) unimodular groups does not alter the size of the objects on which they act, their action on infinitesimal objects keeps their size unchanged. As a consequence, for (most) unimodular groups, the Haar measure is equal to the Lebesgue measure, i.e., $\du \mu_\gG(\gamma) {=} \du \gamma$, $\forall \gamma \in \gG$, and therefore, it is often omitted in literature, e.g., in \citet{cohen2016group}.

In contrast, non-unimodular groups, such as the scale group and the scale-translation group, do modify the size of objects on which they act (Fig.~\ref{fig:unimodular} right). Consequently, their action on infinitesimal objects changes their size. As a result, the Haar measure must be treated carefully in order to obtain equivariance to non-unimodular groups \citep{bekkers2020bspline}. The Haar measure guarantees that $\du \mu_\gG (\gamma) {=} \du \mu_\gG (g \gamma)$, $\forall \ g, \gamma \in \gG$. For the scale-translation group, it is obtained as:
\begin{equation}
\du \mu_\gG (\gamma) = \du \mu_\gG (g \gamma) = \du \mu_\gG(t + s \tau, s \varsigma) = \du \mu_\gG(t + s \tau) \du \mu_\gG(s \varsigma) = \frac{1}{\mathopen|s\mathclose|} \du \tau \frac{1}{\mathopen|s\mathclose|} \du \varsigma,\label{eq:haar_meas}
\end{equation} 
where $g {=}(t, s), \gamma {=} (\tau, \varsigma) \in \gG$, $t, \tau \in \sR$, $s, \varsigma \in \sR_{>0}$; $ \du \tau$, $\du \varsigma$ are the Lebesgue measure of the respective spaces; and $\mathopen| s \mathclose|$ depicts the determinant of the matrix representation of the group element.\footnote{A member $s$ of the scale group $\sR_{>0}$ acting on a $N$-dimensional space is represented a matrix $\mathrm{diag}(s, ..., s)$. Since, its determinant $s^{N}$ depends on the value of the group element $s$, the factor $\tfrac{1}{\mathopen|s \mathclose|}{=}\tfrac{1}{s^N}$ in Eq.~\ref{eq:haar_meas} cannot be omitted.} 
Intuitively, the Haar measure counteracts the growth of infinitesimal elements resulting from the action of $s$ on $\sR {\times} \sR_{>0}$.

\begin{figure}
    \centering
    \includegraphics[width=0.9\textwidth]{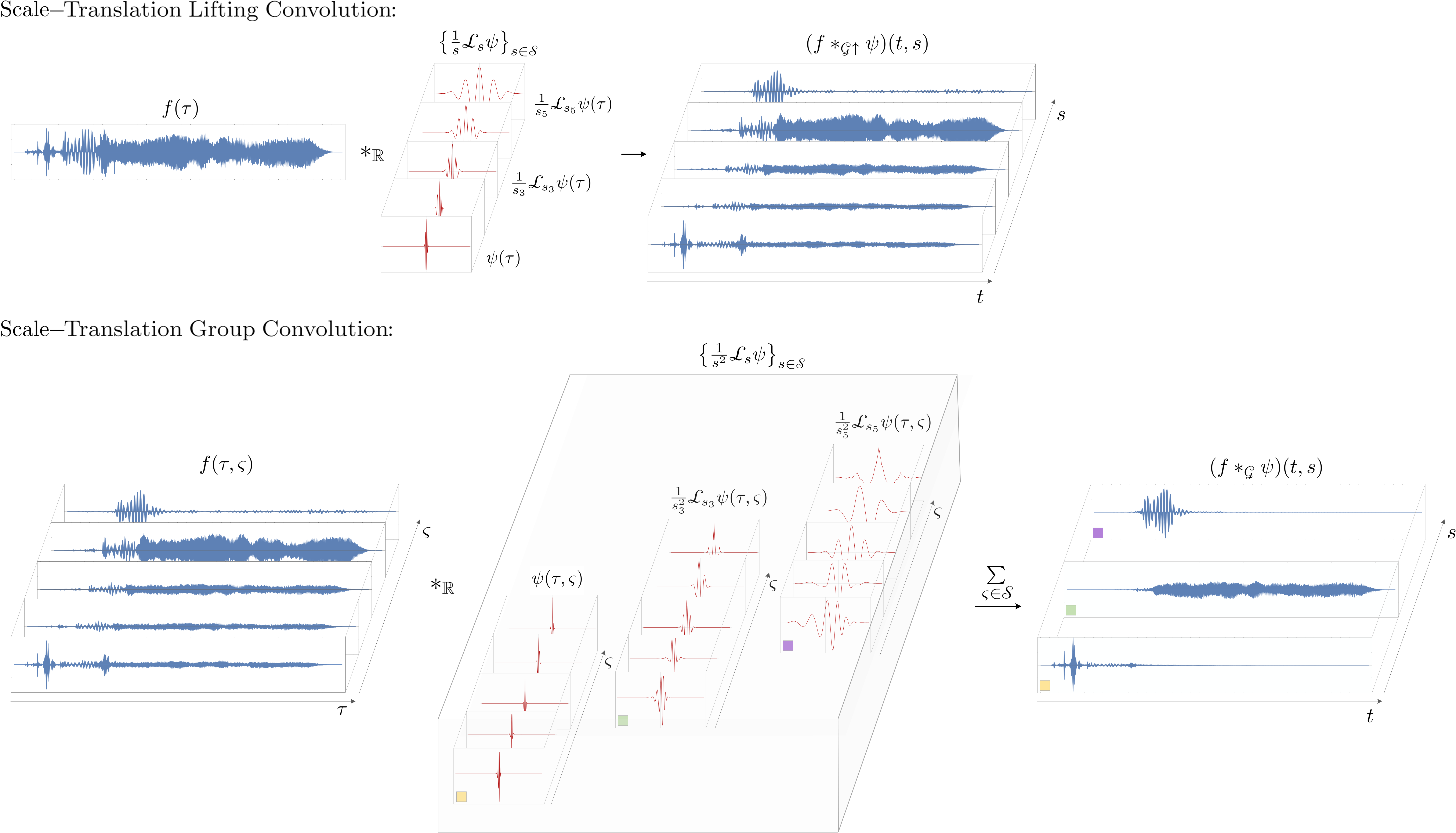}
    \vspace{-1mm}
    \caption{Scale-translation lifting and group convolution. The lifting convolution can be seen a set of 1$\mathrm{D}$ convolutions with a bank of scaled convolutional kernels $\frac{1}{s}\gL_s \psi$, and the group convolution can be seen as a set of 1$\mathrm{D}$ convolutions with a bank of scaled convolutional kernels $\frac{1}{s^2}\gL_s\psi$, followed by an integral over scales $\varsigma \in \sR$. Their main difference is that, for group convolutions, the input $f$ and the convolutional kernel $\psi$ are functions on the scale-translation group whereas for lifting convolutions these are functions on $\sR$. Lifting and group convolutions can be seen as spectro-temporal decompositions with large values of $s$ relating to coarse features and small values to finer features.
    \vspace{-4mm}}
    \label{fig:approach}
\end{figure}

\textbf{Scale-translation group convolutions.} The general formulation of the group convolution is given in Eq.~\ref{eq:gconv}. Interestingly, the scale-translation group has additional properties with which this formulation can be simplified. In particular, by taking advantage of the fact that the scale-translation group is an \textit{affine} group $\gG {=} \sR \rtimes \gS$, with $\gS {=} (\sR_{>0}, \times)$, as well as of the definition of the Haar measure for the scale-translation group in Eq.~\ref{eq:haar_meas} we can reformulate the group convolution for the scale-translation group as:  
\begin{align}
      (f *_{\gG} \psi)(g) &=  \hspace{-1mm}  \int_{\gG} f(\gamma)\psi(g^{-1}\gamma)\ \du \mu_\gG(\gamma) \nonumber \\
      (f *_{\gG} \psi)(t,s) 
     &=  \int_\gS \int_{\sR} f(\tau, \varsigma)\psi\left((t, s)^{-1} (\tau, \varsigma)\right)\ \frac{1}{\mathopen|s\mathclose|} \du \tau \ \frac{1}{\mathopen|s\mathclose|} \du \varsigma \nonumber \\
     &=  \int_\gS \int_{\sR} f(\tau, \varsigma)\ \frac{1}{s^2}\psi\left(s^{-1} (\tau - t , \varsigma)\right) \ \du \tau \ \du \varsigma \nonumber\\
     &=  \int_\gS \int_{\sR} f(\tau, \varsigma) \ \frac{1}{s^2} \gL_s\psi\left(\tau - t,  \varsigma\right) \ \du \tau \ \du \varsigma =  \int_\gS \left( f *_\sR \frac{1}{s^2} \gL_s\psi \right) (t, \varsigma)  \ \du \varsigma
\label{eq:groupconv_diltransgroup}
\end{align}
where $g{=}(t, h)$, $\gamma{=}(\tau, \varsigma) \in \gG$, $t, \tau \in \sR$, and $s, \varsigma \in \sR_{>0}$; and $\gL_{s}\psi(\tau, \varsigma){=}\psi\left(s^{-1}(\tau, \varsigma)\right)$ is the (left) action of the scale group $\gS$ on a convolutional kernel $\psi: \sR \times \sR_{>0} \rightarrow \sR$ defined on the scale-translation group. In other words, for the scale-translation group, \textit{the group convolution can be seen as a set of 1$\mathrm{D}$ convolutions with a bank of scaled convolutional kernels $\{\frac{1}{s^2}\gL_s\psi\}_{s\in\gS}$, followed by an integral over scales $\varsigma \in \sR$} (Fig.~\ref{fig:approach}, bottom).

\textbf{Scale-translation lifting convolution.} Like the group convolution, the lifting convolution can also be simplified by considering the properties of the scale-translation group. In particular, we can rewrite it as: 
\begin{align}
    (f *_{\gG\uparrow} \psi)(g) &= \int_\gX f(x)\psi(g^{-1}x)\ \du \mu_\gG(x) = \int_\sR f(\tau) \psi(g^{-1}\tau) \du_\gG(\tau) \nonumber \\
    (f *_{\gG\uparrow} \psi)(t, s) &= \int_\sR f(\tau)\psi((t, s)^{-1}\tau)\ \frac{1}{\mathopen|s\mathclose|}\du \tau = \int_\sR f(\tau)\ \frac{1}{s}\psi\left(s^{-1}(\tau - t)\right)\ \du \tau \nonumber \\%= \int_\sR f(\tau)\ \frac{1}{s}\gL_s\psi\left(\tau - t\right)\ \du \tau 
    &= \left(f *_\sR \frac{1}{s}\gL_s\psi \right)(t) \label{eq:lifting_diltransgroup}
\end{align}
where $g{=}(t, h)$, $\gamma{=}(\tau, \varsigma) \in \gG$, $t, \tau \in \sR$, and $s, \varsigma \in \sR_{>0}$; and $\gL_{s}\psi(\tau, \varsigma){=}\psi\left(s^{-1}(\tau, \varsigma)\right)$ is the (left) action of the scale group $\gS$ on a 1$\mathrm{D}$ convolutional kernel $\psi: \sR \rightarrow \sR$. In other words, for the scale-translation group, \textit{the lifting convolution can be seen as a set of 1$\mathrm{D}$ convolutions with a bank of scaled convolutional kernels $\{\frac{1}{s}\gL_s \psi\}_{s \in \gS}$} (Fig.~\ref{fig:approach}, top). Note that the Haar measure imposes a normalization factor of $\tfrac{1}{s^2}$ for group convolutions and of $\tfrac{1}{s}$ for the lifting convolution. This is because space on which the group convolution is performed $(\sR \rtimes \sR_{>0})$ has an additional dimension relative to the space on which the lifting convolution is performed ($\sR$).

\vspace{-7mm}
\subsection{Wavelet Networks: architecture and practical implementation}
\begin{figure}
    \centering
    \includegraphics[width=0.2\linewidth]{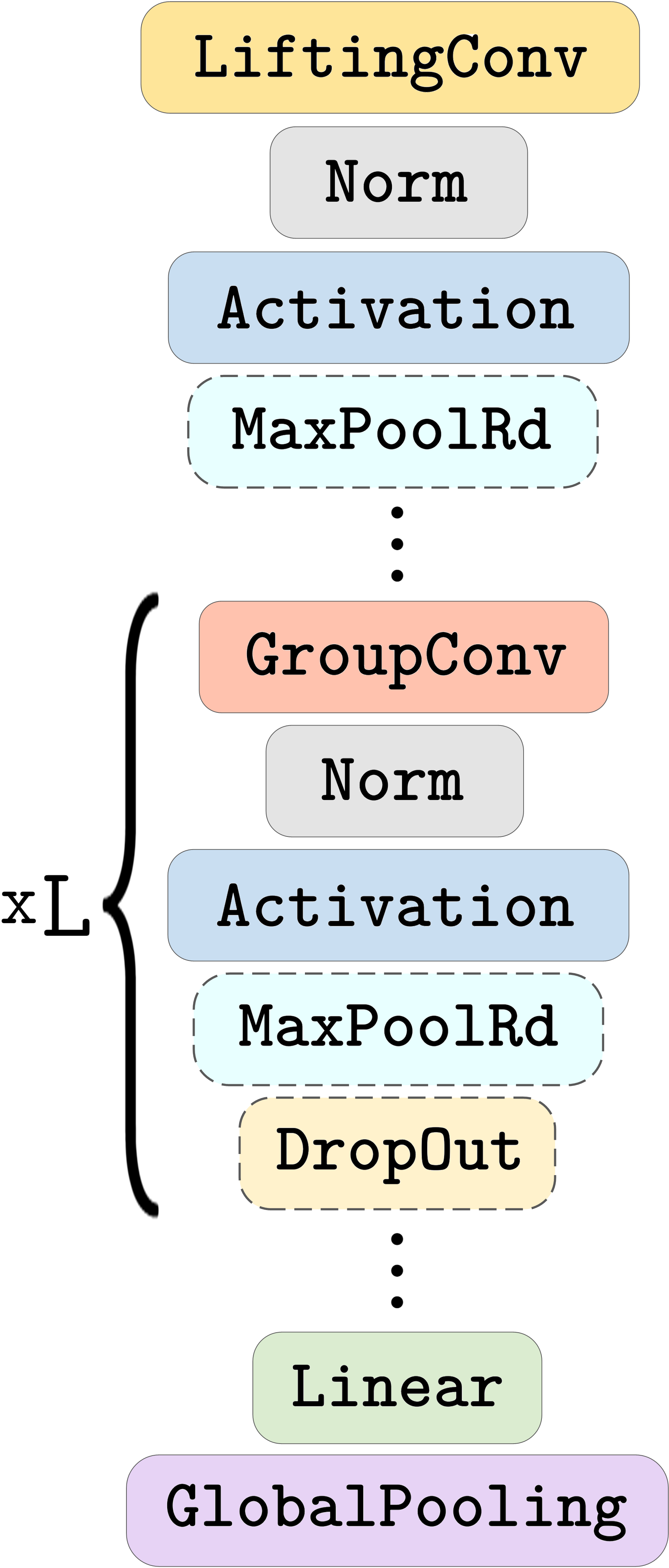}
    \vspace{-3mm}
    \captionsetup{justification=centering}
    \caption{Architecture of wavelet networks.}
    \label{fig:net_structure}
\end{figure}
The general architecture of our proposed Wavelet networks is shown in Fig.~\ref{fig:net_structure}. Wavelet networks consist of several stacked layers that respect scale and translation. They consist of a lifting group convolution layer that lifts input time-series to the scale-translation group, followed by arbitrarily many group convolutional layers. At the end of the network, a global pooling layer is used to produce scale-translation invariant representations. Due to their construction, Wavelet networks make sure that common neural operations, e.g., point-wise nonlinearities, do not disrupt scale and translation equivariance. This in turn, makes them broadly applicable and easily extendable to other existing neural architectures, e.g., ResNets \citep{he2016deep}, U-Nets \citep{ronneberger2015u}.

\vspace{-7mm}
\subsubsection{Group convolutional kernels on continuous bases}
\begin{figure}
    \centering
        \centering
        \begin{subfigure}{0.35\textwidth}
        \captionsetup{justification=centering}
        \includegraphics[width=0.9\textwidth]{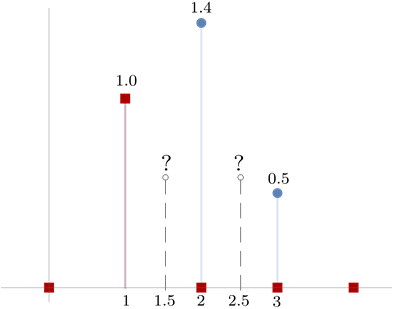}
        \caption{Discrete bases (Dirac deltas)}
        \label{fig:discr_basis}
    \end{subfigure}
    \hspace{5mm}
        \begin{subfigure}{0.35\textwidth}
        \captionsetup{justification=centering}
        \includegraphics[width=\textwidth]{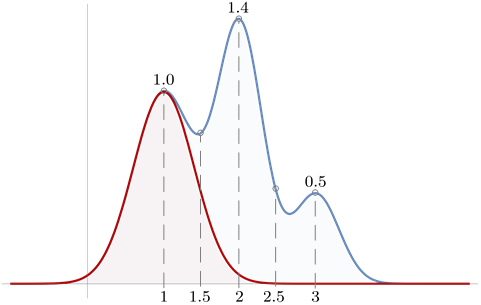}
        \caption{Continuous bases ($\mathrm{B}^2$-splines)}
        \label{fig:cont_basis}
    \end{subfigure}
    \vspace{-2mm}
    \caption{Convolutional kernels on discrete and continuous bases. The canonical basis used for the construction of the convolutional kernel is shown in red: a delta Dirac for the discrete case, and a $\mathrm{B}^{2}$-spline for the continuous case.}
    \label{fig:bases}
\end{figure}

Although our previous derivations build upon continuous functions, in practice, computations are performed on discretized versions of these functions. Continuous bases have proven advantageous for the construction of group convolutions as the action of relevant groups often impose transformations not well-defined for discrete bases \citep{weiler2018learning, bekkers2018roto, weiler2019general}. For instance, in the context of scale-translations, scaling a kernel $[w_1, w_2, w_3]$ by a factor of two results in a filter $[w_1, w_{1.5}, w_2, w_{2.5}, w_3]$ wherein the introduced values $[w_{1.5}, w_{2.5}]$ do not exist in the original basis (Fig.~\ref{fig:discr_basis}).

The most adopted solution to address this problem is interpolation, i.e., deriving the value of  $[w_{1.5}, w_{2.5}]$ based on the neighbouring known pixels. However, interpolation introduces spurious artifacts which are particularly severe for small kernels. Instead, we adopt an alternative approach: we define convolutional kernels directly on a continuous basis (Fig.~\ref{fig:cont_basis}). Drawing from the resemblance of gammatone filters --strongly motivated by the physiology of the human auditory system for the processing and recognition of auditory signals \citep{johannesma1972pre, hewitt1994computer, lindeberg2015idealized}-- to $\mathrm{B}^{2}$-splines, we parameterize our filters within a $\mathrm{B}^{2}$-spline basis as in \citet{bekkers2020bspline}. As a result, our convolutional filters are parameterized as a linear combination of shifted $\mathrm{B}^{2}$-splines $\psi(\tau) {\coloneqq} \sum_{i=1}^{\mathrm{N}}w_{i} \mathrm{B}^{2}(\tau - \tau_i)$, rather than the commonly used shifted Dirac delta's basis $\psi(\tau) {\coloneqq} \sum_{i=1}^{\mathrm{N}}w_{i} \delta(\tau - \tau_i)$.

\vspace{-7mm}
\subsubsection{Constructing a discrete scale grid}\label{sec:discrete_grid}
From the response of the lifting layers onward, the feature representations of wavelet networks possess an additional axis $s \in \sR_{>0}$. Just like the spatial axis, this axis must be discretized in order to perform computational operations. That is, we must approximate the scale axis $\sR_{>0}$ by a finite set of discrete scales $\{ s \}_{s {=} s_\mathrm{min}}^{s_\mathrm{max}}$. Inspired by previous work, we approximate the scale axis with a \textit{dyadic set} $\{ 2^{j}\}_{j = j_{\text{min}}}^{j_{\text{max}}}$ \citep{mallat1999wavelet, lindeberg2015scale, worrall2019deep}. Dyadic sets resemble the spectro-temporal resolution of the human auditory system, and are widely used for discrete versions of the wavelet transform.

\begin{figure}
    \centering
    \hfill
    \begin{subfigure}{0.26\textwidth}
    \centering
    \captionsetup{justification=centering}
        \includegraphics[width=\textwidth]{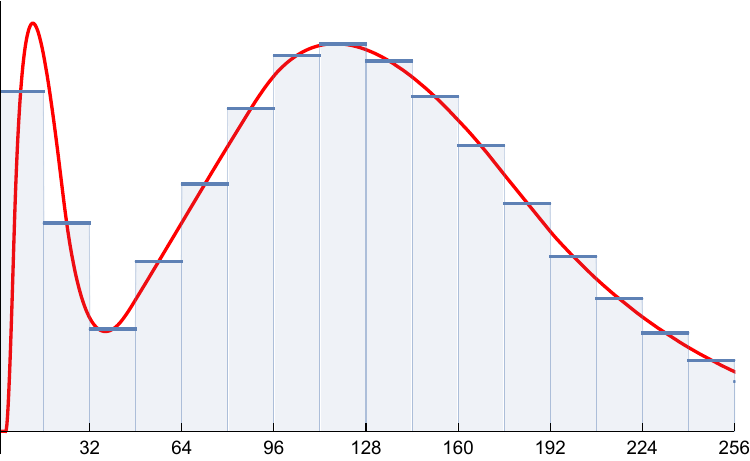}
        \caption{linear grid}
        \label{fig:riemannintegral_linlin}
    \end{subfigure}
    \hfill
    \begin{subfigure}{0.26\textwidth}
    \centering
    \captionsetup{justification=centering}
        \includegraphics[width=\textwidth]{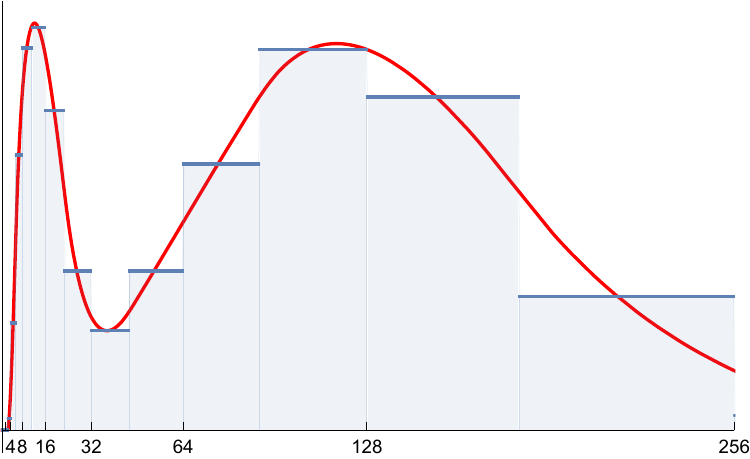}
        \caption{exponential grid}
        \label{fig:riemannintegral_linlog}
    \end{subfigure}
    \hfill
    \begin{subfigure}{0.30\textwidth}
    \centering
    \captionsetup{justification=centering}
        \includegraphics[width=0.9\textwidth]{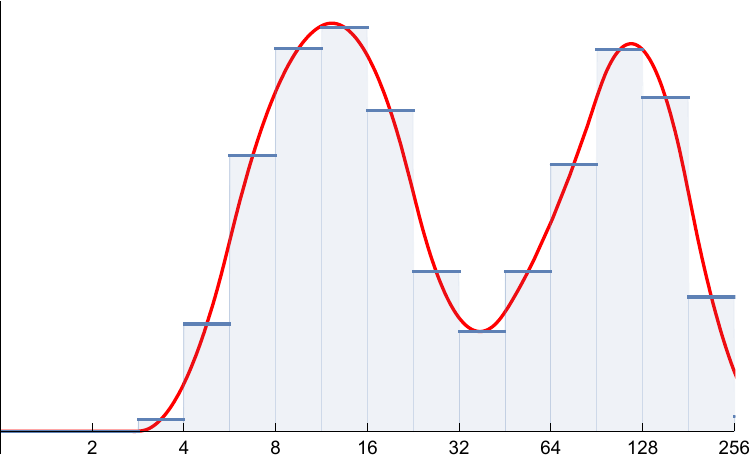}
        \caption{log-plot exponential grid}
        \label{fig:riemannintegral_log}
    \end{subfigure}
    \hfill
    \vspace{-2mm}
        \caption{Riemann integration of functions on $\mathbb{R}_{>0}$ using linear and exponential grids. \label{fig:riemannintegral}
        \vspace{-3mm}} 
\end{figure}

\textbf{Integrating on exponential grids.} A subtlety arises with respect to integrating over the scale axis when implementing the continuous theory in a discrete setting that is suitable for numerical computations. The group convolutions include scale correction factors as part of the Haar measure, which makes the integration invariant to actions along the scale axis. That is, the integral of a signal $f(s)$ over scale is the same as that of the same signal $f(z^{-1}s)$, whose scale is changed by a factor $z\in\mathbb{R}_{>0}$:
\begin{equation}
 \int_{\mathbb{R}_{>0}} \hspace{-2mm}f(z^{-1} s) \tfrac{1}{s}{\rm d}{s} \overset{s \rightarrow z s}{=}
\int_{\mathbb{R}_{>0}} \hspace{-2mm} f(z^{-1} s) \tfrac{1}{z s}{\rm d}{z s} = \int_{\mathbb{R}_{>0}} \hspace{-2mm} f(s) \tfrac{1}{s}{\rm d}{s}.  
\end{equation}
We can translate the scale integration to the discrete setting via Riemann integrals, where we sample the function on a grid and take the weighted sum of these values with weights given by the bin-width:
\begin{equation}
\int_{\mathbb{R}_{>0}} \hspace{-2mm}f(s) \tfrac{1}{s}{\rm d}{s} \approx
\sum_{i} f(s_i) \tfrac{1}{s_i} \Delta_i .
\end{equation}
When the scale grid is linear, the bin-widths $\Delta_i$ are constant, as depicted in Fig.~\ref{fig:riemannintegral_linlin}. When the scale grid is exponential, e.g., $s_i {=} b^{i-1}$ with $b$ some base factor, the bin widths are proportional to the scale values at the grid points, i.e., $\Delta_i \propto s_i$ (Fig.~\ref{fig:riemannintegral_linlog}). In this setting, the factor $\tfrac{1}{s_i}$ cancels out (up to some constant) with the bin width $\Delta_i$, and integration is simply done by summing the values sampled on the scale grid. Consequently, when working with an exponential grid along the scale axis, the factor in the group convolutions (Eq.~\ref{eq:groupconv_diltransgroup}) becomes $\tfrac{1}{s}$ instead of $\tfrac{1}{s^{2}}$. It is worth mentioning that using an exponential grid is the natural thing to do when dealing with the scale group. The scale group is a multiplicative group with a natural distance between group elements $z,s \in \mathbb{R}_{>0}$ defined by $\|\log{z^{-1} s} \|$. Consequently, on an exponential grid, the grid points are spaced uniformly with respect to this distance, as~illustrated~in~Fig.~\ref{fig:riemannintegral_log}.

\textbf{Defining the discrete scale grid.}
In practice, Wavelet networks must define the number of scales $\mathrm{N_s}$ to be considered in the dyadic set as well as its limits $s_\mathrm{min}, s_\mathrm{max}$. Fortunately, it turns out that these values are related to the spatial dimension of the input $f$ itself, and thus, we can use it to determine these values.

Let us consider a signal $f$ and a convolutional kernel $\psi$ sampled on discrete grids $[1,\mathrm{N}_f]$, $[1,\mathrm{N}_{\psi}] \subset \mathbb{Z}$ of sizes $\mathrm{N}_{f}$, and $\mathrm{N}_{\psi}$, respectively. When we re-scale the convolutional kernel $\psi$, we are restricted (\emph{i}) at the bottom of the scale axis by the Nyquist criterion, and (\emph{ii}) at the top of the scale by the scale for which the filter becomes constant in an interval of $\mathrm{N}_{f}$ samples. The Nyquist criterion is required to avoid aliasing and intuitively restricts us to a compression factor on $\psi$ such that it becomes as big as 2 grid samples. On the other hand, by having $\psi$ re-scaled to an extreme to which it is constant in the support of the input signal $f$, the kernel will only be able to perform average operations. 

\underline{\textit{Considerations regarding computational complexity}}. Note that the computational cost of Wavelet networks increases linearly with the number of scales considered. Hence, it is desirable to reduce the number of scales used as much as possible. To this end, we reason that using scales for which the sampled support of $\psi$ is smaller than $\mathrm{N}_{\psi}$ is unnecessary as the functions that can be described at those scales can also be described --and learned-- at the unscaled resolution of the kernel $s{=}1$. Therefore, we define the minimum scale as $s_\mathrm{min}{=}1$. Furthermore, we reason that using scales for which the support of the filter overpasses that of the input, i.e., $\mathrm{N}_{f} \leq \mathrm{N}_{\psi}$, is also suboptimal, as the values outside of the region $[1, \mathrm{N}_{f}]$ are unknown. Therefore, we consider the set of sensible scales to be given by the interval $[1, \frac{\mathrm{N}_{f}}{\mathrm{N}_\psi}]$. In terms of a dyadic set $\{ 2^j\}_{j{=}j_\mathrm{min}}^{j_\mathrm{max}}$, this corresponds to the $j$-values given by the interval $[0, 1, 2, ..., j_\mathrm{max}\ \text{s.t.} \ \mathrm{N}_{\psi}\  2^{j_\mathrm{max}} \leq \mathrm{N}_{f}]$.

\underline{\textit{Effect of downsampling on the scale grids used}}. Neural architectures utilize pooling operations, e.g., $\max$-pooling, to reduce the spatial dimension of the input as a function of depth. Following the rationale outlined in the previous paragraph, we take advantage of these reductions to reduce the number of scales that representations at a given depth should use. Specifically, we use the factor of downsampling as a proxy for the number of scales that can be disregarded. For example, if we use a pooling of 8 at a given layer, subsequent layers should reduce the number of scales considered by the same factor, i.e., $2^{3}$. For a set of dyadic scales before a pooling layer given by $\{ 2^j\}_{j{=}j_\mathrm{min}}^{j_\mathrm{max}}$ and a pooling layer of factor $2^p$, the set of dyadic scales considered after pooling will be given by $\{ 2^j\}_{j{=}j_\mathrm{min}}^{j_\mathrm{max} - p}$.

\vspace{-7mm}
\subsubsection{Imposing wavelet structure to the learned convolutional kernels}\label{sec:wavelet_structure}
In classical spectro-temporal analysis, wavelets are designed to have unit norm $\|\psi\|^2{=}1$ and zero mean $\int \psi(\tau) \ \du \tau {=} 0$. These constraints are useful for both theoretical and practical reasons including energy preservation, numerical stability and the ability to act as band-pass filters \citep{mallat1999wavelet}. Since Wavelet networks construct time-frequency representations of the input, we experiment with an additional regularization loss that encourages the learned convolutional kernels to behave like wavelets. First, we note that lifting and group convolutions inherently incorporate a normalization term --$\tfrac{1}{s}$, $\tfrac{1}{s^2}$-- in their definitions. Therefore, the normalization criterion is inherently satisfied. To encourage the learned kernels to have zero mean, we formulate a regularization term that promote this behaviour. Denoting $\psi_d$ as the convolutional kernel at the $d$-th layer of a neural network with $\mathrm{D}$ convolutional layers, the regularization term $\gL_{\mathrm{wavelet}}$ is defined as:
\begin{equation}
\gL_{\mathrm{wavelet}} =  \sum_{d=1}^{\mathrm{D}} \| \mathrm{mean}(\psi_d) \|^2.
\end{equation}
Interestingly, we observe that enforcing wavelet structure in the learned convolutional kernels consistently yields improved performance across all tasks considered (Sec.~\ref{sec:experiments}). This result underscores the potential value of integrating insights from classical signal processing, e.g., spectro-temporal analysis \citep{scharf1991statistical, mallat1999wavelet, daubechies2006fundamental}, in the design of deep models.

\vspace{-7mm}
\subsection{Wavelet networks perform nested non-linear time-frequency transforms}
Interestingly, we can use spectro-temporal analysis to understand the modus operandi of wavelet networks. Our analysis reveals that wavelet networks perform nested time-frequency transforms interleaved with point-wise nonlinearities. In this process, each time-frequency transform emerges as a linear combination of parallel wavelet-like transformations of the input computed with learnable convolutional kernels $\psi$.

\textbf{The relation between scale-translation equivariant mappings and the wavelet transform.} 
The wavelet transform shows many similarities to the scale-translation group and lifting convolutions \citep{grossmann1985transforms}. In fact, by analyzing the definition of the wavelet transform (Eq.~\ref{eq:wavelet_transform}), we obtain that the Wavelet transform is equivalent to a lifting group convolution (Eq.~\ref{eq:lifting_diltransgroup} with $\omega{=}s$) up to a normalization factor $\frac{1}{\sqrt{\omega}}$:
\begin{align}
    \gW[f](t, \omega) &= \int_\sR f(\tau) \frac{1}{\sqrt{\omega}}\psi\left(\frac{\tau - t}{\omega}\right) \, {\rm d}\tau \nonumber \\ 
    &= \int_\sR f(\tau) \frac{1}{\sqrt{\omega}}\psi\left(\omega^{-1}(\tau - t)\right) \, {\rm d}\tau \nonumber = \int_\sR f(\tau) \frac{1}{\sqrt{\omega}}\gL_\omega\psi\left(\tau - t\right) \, {\rm d}\tau \nonumber\\
    &= \left( f *_\sR \frac{1}{\sqrt{\omega}}\gL_\omega\psi\right)(t) = \frac{1}{\sqrt{\omega}}\left( f *_{\gG\uparrow} \psi\right)(t, \omega).
\end{align}
Furthermore, if we let the input $f$ be a function defined on the scale-translation group, and let $\omega$ act on this group according to the group structure of the scale-translation group, we have that the scale-translation group convolution is equivalent to a Wavelet transform whose input has been obtained by a previously applied Wavelet transform, , up to a normalization factor $\frac{1}{\omega\sqrt{\omega}}$:
\begin{align}
    \gW[f](t, \omega) &= \int_{\sR_{>0}}\int_\sR f(\tau, \varsigma) \frac{1}{\sqrt{\omega}}\psi\left(\omega^{-1}(\tau - t), \varsigma\right) \, {\rm d}\tau \ \du \varsigma \nonumber \\
    &= \int_{\sR_{>0}} \int_\sR f(\tau, \varsigma) \frac{1}{\sqrt{\omega}}\gL_\omega\psi\left(\tau - t, \varsigma\right) \, \du\tau \ \du\varsigma \nonumber \\
    &= \int_{\sR_{>0}} \left( f *_\sR \frac{1}{\sqrt{\omega}}\gL_\omega\psi\right)(t, \varsigma) \ \du \varsigma = \frac{1}{\omega\sqrt{\omega}} \left( f *_{\gG} \psi\right)(t, \omega)
\end{align}
In other words, lifting and group convolutions on the scale-translation group can be interpreted as linear time-frequency transforms that adopt time-frequency plane tiling akin wavelet transform (Fig.~\ref{fig:wavelet}), for which the group convolution accepts wavelet-like spectro-temporal representations as input.

\begin{tcolorbox}[enhanced, frame hidden, breakable, sharp corners, before skip=5pt, after skip=10pt]
\textbf{Equivariance properties of common time-frequency transforms.} For completeness, we also analyze the equivariance properties of common time-frequency transforms and their normalized representations, e.g., spectrogram. Careful interpretations and proofs are provided in Appx.~\ref{appx:equiv_gral}.\newline

Let $\gL_{t_0}f=f(t-t_{0})$ and $\gL_{s_{0}}f(t)=f(s_{0}^{-1}t)$, $t_{0} \in \sR$, $s_{0} \in \sR_{>0}$, be translation and scaling operators. The Fourier, short-time Fourier and Wavelet transform of $\gL_{t_0}f$ and $\gL_{s_0}f$, $f \in \ltwo(\sR)$, are given by:
\begin{itemize}[leftmargin=0.6cm]
    \item \textbf{Fourier Transform:} 
    \begin{align}
        &\mathcal{F}[\mathcal{L}_{t_{0}}f](\omega)= \eu^{- \iu \omega t_0}  \mathcal{F}[f](\omega) && \rightarrow \mathopen|\mathcal{F}[\mathcal{L}_{t_{0}}f](\omega) \mathclose|^{2} = \mathopen|\mathcal{F}[f](\omega)\mathclose|^{2}\hfill\\
        &\mathcal{F}[\mathcal{L}_{s_{0}}f](\omega) = s_{0}\mathcal{L}_{s_0^{-1}}\mathcal{F}[f](\omega) && \rightarrow  \mathopen|\mathcal{F}[\mathcal{L}_{s_{0}}f](\omega)\mathclose|^{2} = \mathopen|s_{0}\mathclose|^{2}\mathopen|\mathcal{L}_{s_0^{-1}}\mathcal{F}[f](\omega)\mathclose|^{2}
    \end{align}
    \item \vspace{-0.5mm}\textbf{Short-Time Fourier Transform:}\vspace{-0.5mm}
    \begin{align}
        &\mathcal{S}[\mathcal{L}_{t_{0}}f](t,\omega)=\eu^{- \iu \omega t_0}  \mathcal{L}_{t_{0}}\mathcal{S}[f](t, \omega)&&\rightarrow \mathopen|\mathcal{S}[\mathcal{L}_{t_{0}}f](t,\omega)\mathclose|^{2}=\mathopen|\mathcal{L}_{t_{0}}\mathcal{S}[f](t, \omega)\mathclose|^{2}\\
        &\mathcal{S}[\mathcal{L}_{s_{0}}f](t,\omega) \approx s_{0}\hspace{0.5mm}\mathcal{S}[f](s_{0}^{-1}t, s_{0}\omega)&&\rightarrow \mathopen|\mathcal{S}[\mathcal{L}_{s_{0}}f](t,\omega)\mathclose|^{2}\approx  \mathopen|s_{0}\mathclose|^{2}\mathopen|\mathcal{S}[f](s_{0}^{-1}t, s_{0}\omega)\mathclose|^{2}\ \boldsymbol{^\dagger} \label{eq:thm1_stft}
    \end{align}
    \item \textbf{Wavelet Transform:}
    \begin{align}
    &\mathcal{W}[\mathcal{L}_{t_{0}}[f]](t, \omega)=\mathcal{L}_{t_{0}}\mathcal{W}[f](t, \omega)&&\rightarrow \mathopen|\mathcal{W}[\mathcal{L}_{t_{0}}f](t, \omega)\mathclose|^{2}=\mathopen|\mathcal{L}_{t_{0}}\mathcal{W}[f](t, \omega)\mathclose|^{2}\\
    &\mathcal{W}[\mathcal{L}_{s_{0}}f](t, \omega)=\sqrt{s_{0}}\ \mathcal{L}_{s_{0}}\mathcal{W}[f](t, \omega) && \rightarrow \mathopen|\mathcal{W}[\mathcal{L}_{s_{0}}f](t, \omega)\mathclose|^{2}= \mathopen|\mathcal{L}_{s_{0}}\mathcal{W}[f](t, \omega)\mathclose|^{2} \label{eq:thm1_wavelet_scale}
    \end{align}
    \hspace{-4mm}$\boldsymbol{^{\dagger}}$ Eq.~\ref{eq:thm1_stft} only holds approximately for large windows (see Appx.~\ref{appx:equiv_wind_four}).
\end{itemize}

In other words, the Wavelet transform and the scalogram $\mathopen|\mathcal{W}[\cdot]\mathclose|^2$ are the only time-frequency representations that exhibit translation and scaling equivariance in a practical way.
\end{tcolorbox}

\textbf{Wavelet networks apply parallel time-frequency transforms with learned bases at\break every layer.} So far, our analysis has been defined for scalar-valued input and convolutional kernels. However, in practice, convolutional layers perform operations between inputs $f:\sR \rightarrow \sR^\Nin$ and convolutional kernels $\psi: \sR \rightarrow \sR^{\Nout \times \Nin}$ to produce outputs $(f * \psi): \sR \rightarrow \sR^\Nout$ as the linear combination along the $\Nin$ dimension of convolutions with several learned convolutional kernels computed in parallel:
\begin{equation}
    (f * \psi)_o {=} \sum_{i=1}^\Nin (f_i * \psi_i), \ \ o \in [1, 2, ..., \Nout].
\end{equation}
In practice, both lifting and group convolutional layers adhere to the same structure. In a dilation-translation convolutional layer with $\Nout$ output channels, $\Nout$ independent convolutional kernels, each consisting of $\Nin$ channels, are learned. During the forward pass, the input is group-convolved with each of these kernels in parallel. The $\Nout$ output channels are then formed by linearly combining the outcomes of the $\Nin$ channels. In other words, lifting and group convolutional layers produce linear combinations of distinct time-frequency decompositions  of the input computed in parallel at each layer.

 \textbf{Wavelet networks are scale-translation equivariant nested non-linear time-frequency transforms.} Just like in conventional neural architectures, the outputs of lifting and group convolutional layers are interleaved with point-wise nonlinearities. Therefore, wavelet networks compute nonlinear scale-translation equivariant feature representations that resemble nested nonlinear time-frequency transforms of the input.
 
\vspace{-7mm}
\section{Experiments}\label{sec:experiments}
In this section, we empirically evaluate wavelet networks. To this end, we take existing neural architectures designed to process raw signals and construct equivalent wavelet networks (W-Nets). We then compare the performance of W-Nets and the corresponding baselines on tasks defined on raw environmental sounds, raw audio and raw electric signals. We replicate as close as possible the training regime of the corresponding baselines and utilize their implementation as a baseline whenever possible. Detailed descriptions of the specific architectures as well as the hyperparameters used for each experiment are provided in Appx.~\ref{appx:exp_details}.

\vspace{-7mm}
\subsection{Classification of environmental sounds}
First, we consider the task of classifying environmental sounds on the UrbanSound8K\newpage

(US8K) dataset \citep{salamon2014dataset}. The US8K dataset consists of 8732 audio clips uniformly drawn from 10 environmental sounds, e.g., siren, jackhammer, etc, of 4 seconds or less, with a total of 9.7 hours of audio. 

\textbf{Experimental setup.} We compare the M$n$-Nets of \citet{dai2017very} and the 1DCNNs of \citet{abdoli2019end} with equivalent W-Nets in terms of number of layers and parameters. Contrarily to \citet{dai2017very} we sample the audio files at $22.05$\si{\kilo\hertz} as opposed to $8$\si{\kilo\hertz}. This results from preliminary studies of the data, which indicated that some classes become indistinguishable for the human ear at such low sampling rates.\footnote{See \url{https://github.com/dwromero/wavelet_networks/blob/master/experiments/UrbanSound8K/data_analysis.ipynb}.
} For the comparison with the 1DCNN of \citet{abdoli2019end}, we select the $50999$-1DCNN as baseline, as it is the network type that requires the less human engineering. We note, however, that we were unable to replicate the results reported in \citet{abdoli2019end}. In contrast to the 83$\pm$1,3\% reported, we were only able to obtain a final accuracy of 62.0$\pm$6.791. This inconsistency is further detailed in Appx.~\ref{appxx:details_us8k}. 

To compare to models other than M$n$-nets and 1DCNNs, e.g., \citet{pons2017end, tokozume2017learning}, we also provide 10-fold cross-validation results. This is done by taking 8 of the 10 official subsets for training, one for validation and one for test. We consistently select the $(n{-}1)\hspace{-2mm}\mod\hspace{-1mm}10$ subset for validation when testing on the $n{\text{-th}}$ subset. We note that this training regime might be different from those used in other works, as previous works often do not disclose which subsets are used for validation.

\textbf{Results.} Our results (Tab.~\ref{tab:US8Kresults}) show that wavelet networks consistently outperform CNNs on raw waveforms. In addition, they are competitive to spectrogram-based approaches, while using significantly fewer parameters and bypassing the need for preprocessing. Furthermore, we observe that encouraging wavelet structure to the convolutional kernels --denoted by the {\small\sc{WL}} suffix-- consistently leads to improved accuracy.

\begin{table}[t]
\captionsetup{justification=centering}
\caption{Experimental results on UrbanSound8K.}\label{tab:US8Kresults}
\vspace{-3.5mm}
\begin{small}
\begin{sc}
\scalebox{0.8}{
\begin{tabular}[t]{cccc}
\toprule
\multicolumn{4}{c}{\textbf{UrbanSound8K}}\\
\midrule
\multirow{2}{*}{Model} & 10$^\text{th}$ Fold   &  Cross-Val. & \multirow{2}{*}{\# Params.} \\
& Acc. (\%) & Acc. (\%) \\
\midrule
\midrule
M$3$-Net & 54.48 & - & 220.67k \\
W$3$-Net & 61.05 & -  & \multirow{2}[1]{*}{219.45k} \\
W$3$-Net-wl & \textbf{63.08} & -  &   \\
\midrule
M$5$-Net& 69.89  & - & 558.08k\\
W$5$-Net & 72.28 & - & \multirow{2}[1]{*}{558.03k} \\
W$5$-Net-wl & \textbf{74.55} & - &\\
\midrule
M$11$-Net& 74.43& - & 1.784m \\
W$11$-Net& 79.33 &  66.97 $\pm$ 5.178 & \multirow{2}[1]{*}{1.806m} \\
W$11$-Net-wl  & \textbf{\underline{80.41}} &  \textbf{\underline{68.47 $\pm$ 4.914}} & \\
\midrule
M$18$-Net& 69.65& - & 3.680m\\
W$18$-Net & 75.87  & 64.02 $\pm$ 4.645 & \multirow{2}{*}{3.759m} \\
W$18$-Net-wl & \textbf{78.26} & \textbf{65.01 $\pm$ 5.431} &\\
\midrule
M$34$-Net& 75.15& - & 3.978m \\
W$34$-Net & 76.22 & 65.69 $\pm$ 5.780 & \multirow{2}[1]{*}{4.021m}\\
W$34$-Net-wl & \textbf{78.38} & \textbf{66.77 $\pm$ 4.771}  &\\
\midrule
\midrule
1DCNN & - & 62.00 $\pm$ 6.791 & 453.42k\\
W-1DCNN& - & 62.47 $\pm$ 4.925 & \multirow{2}{*}{458.61k}\\
W-1DCNN-wl& - & \textbf{62.64 $\pm$ 4.979} & \\
\bottomrule
\vspace{-3mm}
\\
\end{tabular}}
\scalebox{0.8}{
\begin{tabular}[t]{cccc}
\toprule
\multicolumn{4}{c}{\textbf{Comparison With Other Approaches}}\\
\midrule
\multirow{2}{*}{Model} & \multirow{2}{*}{Type} &  Cross-Val. & \multirow{2}{*}{\# Params.} \\
& & Acc. (\%) \\
\midrule
\midrule
    W$11$-Net-wl  & Raw & 68.47 $\pm$ 4.914 & 1.806m \\
\midrule
PiczakCNN \cite{piczak2015environmental} &   \multirow{2}{*}{Mel Spectrogram} & 73.7 & 26m \\
%SB-CNN \cite{salamon2017deep} & & 73 & 241k \\
VGG \cite{pons2019randomly} &   & 70.74  & 77m \\
\midrule
EnvNet-v2 \cite{tokozume2017learning} & Raw (Bagging)  & \underline{\textbf{78}}  & 101m \\
\bottomrule
\end{tabular}}
\end{sc}
\end{small}
\vspace{-2mm}
\end{table}

\vspace{-9mm}
\subsection{Automatic music tagging}
Next, we consider the task of automatic music tagging on the MagnaTagATune (MTAT) dataset \citep{law2009evaluation}. The MTAT dataset consists of 25879 audio clips with a total of 170 hours of audio, along with several per-song tags. The goal of the task is to provide the right tags to each of the songs in the dataset.

\textbf{Experimental setup.} Following \citet{lee2017sample}, we extract the most frequently used $50$ tags and trim the audios to 29.1 seconds at a sample-rate of $22.05$\si{\kilo\hertz}. Following the convention in literature, we use ROC-curve (AUC) and mean average precision (MAP) as performance metrics. We compare the best performing model of \citet{lee2017sample}, the $3^{9}$-Net with a corresponding wavelet network denoted W$3^{9}$-Net.  

\textbf{Results.} Our results (Tab.~\ref{tab:results_magna}) show that wavelet networks consistently outperform CNNs on raw waveforms and perform competitively to spectrogram-based approaches in this dataset as well. In addition, we observe that encouraging the learning of wavelet-like kernels consistently results in increased accuracy as well.
\begin{table}[t]
\captionsetup{justification=centering}
\caption{Experimental results on MTAT.}\label{tab:results_magna}
\vspace{-3.5mm}
\begin{small}
\begin{sc}
\scalebox{0.82}{
\begin{tabular}[t]{cccccc}
\toprule
\multicolumn{6}{c}{\textbf{MagnaTagATune}}\\
\midrule
\multirow{2}[3]{*}{Model} & \multicolumn{2}{c}{Average AUC} & \multicolumn{2}{c}{MAP} & \multirow{2}[3]{*}{\# Params.}\\
\cmidrule{2-5}
  & Per-class & Per-clip & Per-class & Per-clip & \\
\midrule
\midrule
$3^{9}$-Net & 0.893 & 0.936 & 0.385 & 0.700 & 2.394m\\
W$3^{9}$-Net & 0.895 & 0.941 & 0.397 & 0.719 & \multirow{2}[1]{*}{2.404m}\\
 W$3^{9}$-Net-wl & \textbf{0.899} & \textbf{\underline{0.943}} & \textbf{0.404} & \textbf{\underline{0.723}} &\\
\bottomrule
\vspace{-3mm}
\\
\end{tabular}}
\scalebox{0.82}{
\begin{tabular}[t]{cccccc}
\toprule
\multicolumn{6}{c}{\textbf{Comparison With Other Approaches}}\\
\midrule
\multirow{2}[3]{*}{Model} & \multicolumn{2}{c}{Average AUC} & \multicolumn{2}{c}{MAP} & \multirow{2}[3]{*}{\# Params.}\\
\cmidrule{2-5}
  & Per-class & Per-clip & Per-class & Per-clip & \\
\midrule
\midrule
PCNN \cite{liu2016applying} & 0.9013 & \textbf{0.9365} & \textbf{ \underline{0.4267}} &\textbf{ 0.6902} & - \\
CNN \cite{pons2017end}$^{*}$ & \multirow{2}{*}{0.8905} & \multirow{2}{*}{-} & \multirow{2}{*}{0.3492} & \multirow{2}{*}{-} & \multirow{2}{*}{11.8m} \\
 (Raw) \\
CNN \cite{pons2017end}$^{*}$  & \multirow{2}{*}{\textbf{\underline{0.9040}}} & \multirow{2}{*}{-} & \multirow{2}{*}{0.3811} & \multirow{2}{*}{-} & \multirow{2}{*}{5m}\\
(Spect.) \\
CNN \cite{pons2017timbre} & \multirow{2}{*}{0.893} &\multirow{2}{*}{ -} & \multirow{2}{*}{-} & \multirow{2}{*}{-} &\multirow{2}{*}{191k} \\
(Spect.) \\
\bottomrule
\multicolumn{6}{l}{{\rm $^{*}$ Reported results are obtained in a more difficult version of this dataset.}}\\
\end{tabular}}
\end{sc}
\end{small}
\vspace{-3mm}
\end{table}

\vspace{-7mm}
\subsection{Bearing fault detection}
Finally, we also validate Wavelet networks for the task of condition monitoring in induction motors. To this end, we classify healthy and faulty bearings from raw data provided by \textit{[Anonymized Company]}. The dataset consists of 246 clips of 15 seconds sampled at $20$\si{\kilo\hertz}. The dataset is slightly unbalanced containing 155 healthy and 91 faulty recordings $[155, 91]$. The dataset is previously split into a training set of $[85, 52]$ and a test set of $[70, 39]$ samples, respectively. These splits are provided ensuring that measurements from the same motor are not included both in the train and the test set. We utilize 20$\%$ of the training set for validation. Each clip is composed of 6 channels measuring both current and voltage on the 3 poles of the motor. 

\textbf{Experimental setup.} We take the best performing networks on the US8K dataset: the {\small{\sc M-11}} and {\small{\sc W-11}} networks, and utilize variants of these architectures for this dataset.

\textbf{Results.} We observe that Wavelet networks outperform CNNs on raw waveforms and encouraging the learning of wavelet-like kernels consistently improves accuracy (Tab.~\ref{tab:sl}).
\begin{table}
\captionsetup{justification=centering}
\begin{center}
\vskip -3.5mm
\begin{small}
\begin{sc}
\scalebox{0.82}{
    \begin{tabular}{ccc}
    \toprule
        Model & Acc. (\%) & \# Params. \\
        \midrule
        \midrule
        M11-Net & 65.1376 & 1.806m\\
        W11-Net & \textbf{68.8073} & \multirow{2}{*}{1.823m}\\
        W11-Net-wl & \textbf{70.207} & \\
         \bottomrule
    \end{tabular}
    \caption{Experimental results on bearing fault detection.}
    \label{tab:sl}
}
\end{sc}
\end{small}
\end{center}
\vspace{-4mm}
\end{table}

\vspace{-10mm}
\subsection{Discussion} 
Our empirical results firmly establish wavelet networks as a promising avenue for learning from raw time-series data. Notably, these results highlight that considering the symmetries inherent to time-series data --namely translation and scale-- for the development of neural networks consistently leads to improved outcomes.
Furthermore, we observe that the benefits of wavelet networks extend beyond sound and audio domains. This result advocates for the use of wavelet networks and scale-translation equivariance for learning on time-series data from different sources, e.g., financial data, sensory data. Finally, we also note that promoting the learning of wavelet-like convolutional kernels consistently leads to improved outcomes. We posit that this discovery may hold broader implications for group equivariant networks in general.

\textbf{Relation to scale-equivariant models of images and $2\mathrm{D}$ signals.} In the past, multiple scale-equivariant models have been proposed for the processing of images and $2\mathrm{D}$ signals \citep{worrall2019deep, sosnovik2020scaleequivariant, sosnovik2021scale}. Interestingly, we find that the difference in the lengths of the inputs received by image and time-series models leads to very different insights per modality. For comparison, \citet{sosnovik2020scaleequivariant} considers images up to $96{\times}96$ pixels, whereas audio files in the US8K dataset are $32.000$ samples long. We find that this difference in input lengths has crucial implications for how scale interactions within scale-equivariant models function. \citet{sosnovik2020scaleequivariant} mentions that using inter-scale interactions introduces additional equivariance errors due to the truncation of the set $\gS$. Therefore, their networks are built with either no scale interaction or interactions of maximum 2 scales. This strongly contrasts with time-series where incorporating inter-scale interactions consistently leads to performance improvements. In our case, the number of scales and inter-scale interactions is rather constrained by the size and computational cost of convolutional kernels (Sec.~\ref{sec:discrete_grid}) rather than their potential negative impact on the model's accuracy.

\vspace{-7mm}
\section{Limitations and future work}\label{sec:limitations}
\textbf{Memory and time complexity grows with the number of scales.} The main limitation of our approach is the increase memory and time demands as the number of scales grows. A potential solution could be adopting Monte-Carlo approximations for the computation of group convolutions \citep{finzi2020generalizing}. This could not only make Wavelet networks equivariant to the continuous scale group --in expectation--, but also could dramatically reduce the number of scales used in each forward pass. Another promising direction involves extending the concept of partial equivariance \citep{romero2022learning} to the scale group. This would allow the model to learn the subset of scales to which it is equivariant, potentially improving execution and adaptability. Lastly, exploring separable group convolutions \citep{knigge2022exploiting} could help reduce the computational and memory requirements of wavelet networks.
% \textbf{TODO.} Wavelet transform discret.
 
\textbf{Convolutions with large convolutional kernels: parameterization and efficiency.} The foundation of our approach hinges on computing convolutions with banks of dilated convolutional kernels (Eq.~\ref{eq:lifting_diltransgroup},~\ref{eq:groupconv_diltransgroup}). Consequently, considering how these kernels are parameterized as well as how these convolutions are computed can unveil avenues for future improvement. Recently, \citet{romero2022ckconv} introduced an expressive continuous parameterization for (large) convolutional kernels that has proven advantageous for complex tasks such as large language modelling \citep{poli2023hyena} and processing DNA chains \citep{nguyen2023hyenadna}. Exploring the use of this parameterization for wavelet networks could lead to valuable insights and improvements, potentially surpassing the current utilization of $\mathrm{B}^2$-spline bases. Furthermore, convolutional networks that rely on convolutions with very large convolutional kernels, e.g., \cite{romero2022ckconv, poli2023hyena, nguyen2023hyenadna}, leverage the Fourier transform to compute convolutions in the frequency domain. In the context of wavelet networks, dynamically selecting between spatial and Fourier convolutions based on the size of convolutional kernels has the potential to significantly improve their efficiency. 

\vspace{-7mm}
\section{Conclusion}
In conclusion, this study introduces \textit{Wavelet Networks}, a new class of neural networks for raw time-series processing that harness the symmetries inherent to time-series data --scale and translation-- for the construction of neural architectures that respect them. We observe a clear connection between the wavelet transform and scale-translation group convolutions, establishing a profound link between our approach and classical spectro-temporal analysis. In contrast to the usual approach, which uses spectro-temporal representations as a frontend for the subsequent use of 2D CNNs, wavelet networks consistently preserve these symmetries across the whole network through the use of convolutional layers that resemble the wavelet transform. Our analysis reveals that wavelet networks combine the benefits of wavelet-like time-frequency decompositions with the adaptability and non-linearity of neural networks.   

Our empirical results demonstrate the superiority of Wavelet Networks over conventional CNNs on raw time-series data, achieving comparable performance to approaches that rely on engineered spectrogram-based methods, e.g., log-Mel spectrograms, with reduced parameters and no need for preprocessing. %We analyze the limitations of our work and characterize different potential directions for refining wavelet networks. 
This work pioneers the concept of scale-translation equivariant neural networks for time-series analysis, opening new avenues for time-series processing.

%% file: chapters/9_partialgcnn.tex
% path to figures directory
\graphicspath{{figures/9-partialgcnn/}}

%=========================================================================

% \begin{savequote}[75mm]
% Nulla facilisi. In vel sem. Morbi id urna in diam dignissim feugiat. Proin molestie tortor eu velit. Aliquam erat volutpat. Nullam ultrices, diam tempus vulputate egestas, eros pede varius leo.
% \qauthor{Quoteauthor Lastname}
% \end{savequote}

\chapter{Learning Equivariances and Partial Equivariances from Data}\label{chapter:partial_equiv}

\begin{flushright}
\textit{Based on the paper:}\break
\textit{Learning Partial Equivariances from Data \citep{romero2022learning}}
\end{flushright}
%=========================================================================
\vspace{-7mm}
\section{Introduction}\label{sec:9_intro}
The translation equivariance of Convolutional Neural Networks (CNNs) \citep{lecun1998gradient} has proven an important inductive bias for good generalization on vision tasks. This is achieved by restricting learned features to respect the translation symmetry encountered in visual data, such that if an input is translated, its features are also translated, but not modified. Group equivariant CNNs (G-CNNs) \citep{cohen2016group} extend equivariance to other symmetry groups. Analogously, they restrict the learned features to respect the symmetries in the group considered such that if an input is transformed by an element in the group, e.g., a rotation, its features are also transformed, e.g., rotated, but not modified.

Nevertheless, the group to which G-CNNs are equivariant must be fixed prior to training, and imposing equivariance to symmetries not present in the data leads to overly constrained models and worse performance \citep{chen2020group}. The latter comes from a difference in the data distribution, and the family of distributions the model can describe. Consequently, the group must be selected carefully, and it should correspond to the transformations that appear naturally in the data.

Frequently, transformations appearing in data can be better represented by a subset of a group than by a group as a whole, e.g., rotations in $[-90^{\circ}, 90^{\circ}]$. For instance, natural images much more likely show an elephant standing straight or slightly rotated than an elephant upside-down. In some cases, group transformations even change the desired model response, e.g., in the classification of the digits 6 and 9. % whose defining factor is their pose. 
In both examples, the data distribution is better represented by a model that respects rotation equivariance \textit{partially}. That is, a model equivariant to some, but not all rotations. 

Moreover, the optimal level of equivariance may change per layer. This results from changes in the likelihood of some transformations for low and high-level features. For instance, whereas the orientations of edges in an human face are properly described with full rotation equivariance, the poses of human faces relative to the camera are better represented by rotations in a subset of the circle.% Whereas detecting a face may benefit from partial rotation equivariance, detecting edges benefits from full rotation equivariance.
\begin{figure*}[t!]
    \centering
    \includegraphics[width=0.95\textwidth]{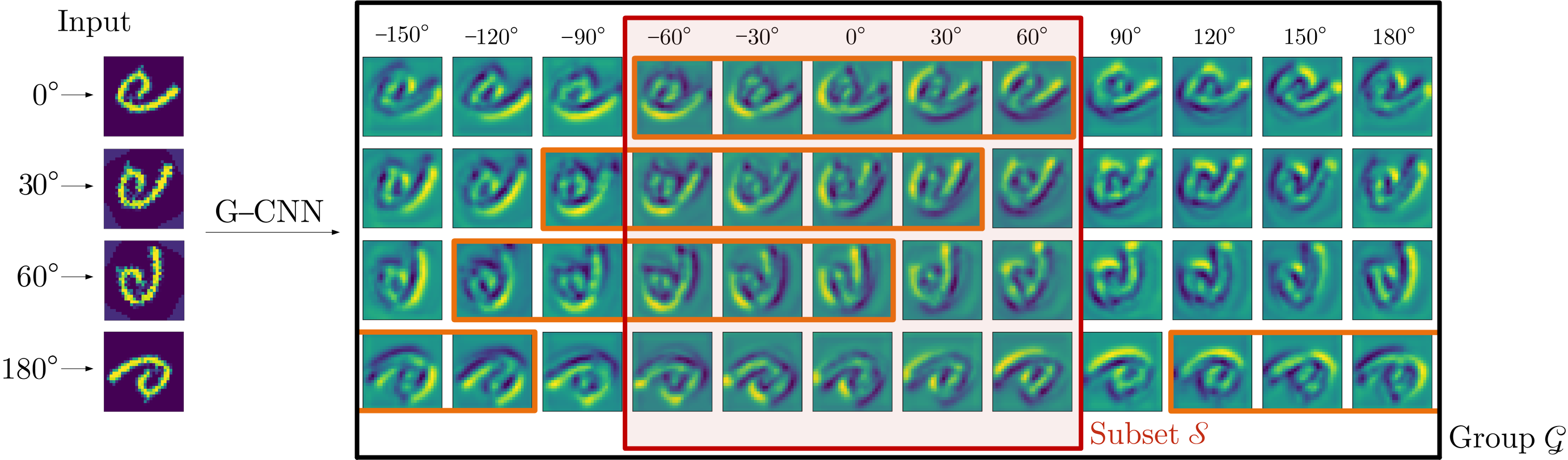}
    \vspace{-2mm}
    \caption{Partial group convolution. In a group convolution, the domain of the output is the group $\gG$. Consequently, all output components are part of the output for any group transformation of the input. In a partial group convolution, however, the domain of the output is a learned subset $\gS$ and all values outside of $\gS$ are discarded. As a result, parts of the output in a partial group convolution can change for different group transformations of the input. In this figure, the output components within $\gS$ for a $0^\circ$ rotation (outlined in orange) gradually leave $\gS$ for stronger transformations of the input. For strong transformations --here a $180^\circ$ rotation--, the output components within $\gS$ are \textit{entirely} different. This difference allows partial group convolutions to distinguish among input transformations. By controlling the size of $\gS$, the level of equivariance of the operation can be adjusted.}
    \label{fig:9_illustrated_partialequiv}
\end{figure*}

The previous observations indicate that constructing a model with different levels of equivariance at each layer may be advantageous. \citet{weiler2019general} empirically observed that manually tuning the level of equivariance at different layers leads to accuracy improvements for non-fully equivariant tasks. Nevertheless, manually tuning layer-wise levels of equivariance is not straightforward and requires iterations over several possible combinations of equivariance levels. Consequently, it is desirable to construct a model able to learn optimal levels of equivariance directly from data.

In this work, we introduce \textit{Partial Group equivariant CNNs} (Partial G-CNNs): a family of equivariant models able to \textit{learn} layer-wise levels of equivariance directly from data. %This is achieved by learning a probability distribution over group elements at each group convolutional layer in the network. 
Instead of sampling group elements uniformly from the group during the group convolution --as in G-CNNs--, Partial G-CNNs learn a probability distribution over group elements at each group convolutional layer in the network, and sample group elements during group convolutions from the learned distributions. By tuning the learned distributions, Partial G-CNNs adjust their level of equivariance at each layer during training.

%Note that our partial equivariance framework allows for equivariance forgetting, and full equivariance. The former is obtained by collapsing the distribution along a certain group dimension and the later results from keeping the learned distribution uniform over the entire group. 
We evaluate Partial G-CNNs on illustrative toy tasks and vision benchmark datasets. We show that whenever full equivariance is beneficial, e.g., for rotated MNIST, Partial G-CNNs learn to remain fully equivariant. However, if equivariance becomes harmful, e.g., for classification of 6~/~9 digits and natural images, Partial G-CNNs learn to adjust equivariance to a subset of the group to improve accuracy. Partial G-CNNs improve upon conventional G-CNNs when equivariance reductions are advantageous, and match their performance whenever their design is optimal.% To the best of our knowledge, this work is the first approach to explore partial group equivariances, and to provide a practical implementation to learn them directly from data end-to-end. 

In summary, our \textbf{contributions} are:

\begin{itemize}%[topsep=0pt, leftmargin=*]
    \item We present a novel design for the construction of equivariant neural networks able to learn layer-wise levels of partial or full equivariance from data.
    % \item We design a mechanism where partial group equivariances can be determined directly from data by learning probability distributions over group elements that minimize the loss function.
    \item We empirically show that Partial G-CNNs perform better than conventional G-CNNs for tasks for which full equivariance is harmful, and match their performance if full equivariance is beneficial.
\end{itemize}
\vspace{-7mm}
\section{Background}\label{sec:9_preliminaries}
This work expects the reader to have a basic understanding of concepts from group theory such as groups, subgroups and group actions. Please refer to Appx.~\ref{appx:9_background} if you are unfamiliar with these terms.

\textbf{Group equivariance.} Group equivariance is the property of a map to respect the transformations in a group. We say that a map is equivariant to a group if whenever the input is transformed by elements of the group, the output of the map is equally transformed but not modified. Formally, for a group $\gG$ with elements $g \in \gG$ acting on a set $\gX$, and a map $\phi: \gX \rightarrow \gX$, we say that $\phi$ is equivariant to $\gG$ if:
\begin{equation}
    \phi(g \cdot x) = g \cdot \phi(x), \ \ \forall x \in \gX, \ \forall g \in \gG.
\end{equation}
For example, the convolution of a signal $f:\sR \rightarrow \sR$ and a kernel $\psi: \sR \rightarrow \sR$ is \textit{translation equivariant} because $\gL_{t}(\psi * f) {=} \psi * \gL_{t}f$, where $\gL_{t}$ translates the function by $t$, i.e., $\gL_{t}f(x) = f(x - t)$. In other words, if the input is translated, its convolutional descriptors are also translated but not modified.

\textbf{The group convolution.} To construct neural networks equivariant to a group $\gG$, we require an operation that respects the symmetries in the group. The \textit{group convolution} is such a mapping. It generalizes the convolution for equivariance to general symmetry groups. Formally, for any $u \in \gG$, the group convolution of a signal $f: \gG \rightarrow \sR$ and a kernel $\psi: \gG \rightarrow \sR$ is given by:
\begin{equation}\label{eq:9_gconv}
   h(u) = (\psi * f)(u) = \hspace{-1mm} \int_{\gG} \psi(v^{-1}u)f(v) \, \du \mu_{\gG}(v),
\end{equation}
where $\mu_{\gG}(\cdot)$ is the (invariant) Haar measure of the group. The group convolution is $\gG$-equivariant in the sense that for all $u, v, w \in \gG$, it holds that:
\begin{equation*}
     (\psi * \gL_{w}f)(u) =  \gL_{w}(\psi * f)(u), \ \text{with} \ \gL_{w}f(u) {=} f(w^{-1} u).
\end{equation*}
\textbf{The lifting convolution.} Regularly, the input of a neural network is not readily defined on the group of interest $\gG$, but on a sub-domain thereof $\gX$, i.e., $f: \gX \rightarrow \sR$. For instance, medical images are functions on $\sR^{2}$ although equivariance to 2D-translations and planar rotations is desirable. In this case $\gX{=}\sR^2$, and the group of interest is $\gG{=}\mathrm{SE(2)}$. Consequently, we must first \textit{lift} the input from $\gX$ to $\gG$ in order to use group convolutions. This is achieved via the \textit{lifting convolution} defined as:
\begin{equation}\label{eq:9_lifting}
    (\psi *_{\text{lift}} f)(u) = \hspace{-1.5mm}\int_{\gX} \hspace{-1mm} \psi(v^{-1}u)f(v) \, \du \mu_{\gG}(v); \ \ u \in \gG, v \in \gX.
\end{equation}
\textbf{Practical implementation of the group convolution.}
The group convolution requires integration over a continuous domain and, in general, cannot be computed in finite time. As a result, it is generally approximated. Two main strategies exist to approximate group convolutions with regular group representations: group discretization \citep{cohen2016group} and Monte Carlo approximation \citep{finzi2020generalizing}. The former approximates the group convolution with a fixed group discretization. Unfortunately, the approximation becomes \textit{only} equivariant to the transformations in the discretization and not to the intrinsic continuous group.

A Monte Carlo approximation, on the other hand, ensures equivariance --in expectation-- to the continuous group. This is done by uniformly sampling transformations $\{ v_j \}$, $\{ u_i\}$ from the group during each forward pass, and using these transformations to approximate the group convolution as:
\begin{equation}\label{eq:9_monte_carlo}
    (\psi \ \hat{*}\  f)(u_i) = \sum\nolimits_{j} \psi(v_j^{-1}u_{i})f(v_j) \mu_{\gG}(v_j).
\end{equation}
Note that this Monte Carlo approximation requires the convolutional kernel $\psi$ to be defined on the \textit{continuous group}. As the domain cannot be enumerated, independent weights cannot be used to parameterize the convolutional kernel. Instead, \citet{finzi2020generalizing} parameterize it with a small neural network, i.e., $\psi(x)=\texttt{MLP}(x)$. This allows them to map all elements $v_j^{-1}u_i$ to a defined kernel value.
\vspace{-7mm}
\section{Partial Group Equivariant Networks}
\vspace{-5mm}
\subsection{(Approximate) partial group equivariance}
Before defining the partial group convolution, we first formalize what we mean by partial group equivariance. We say that a map $\phi$ is \textit{partially equivariant} to $\gG$, if it is equivariant to transformations in a subset of the group $\gS \subset \gG$, but not necessarily to all transformations in the group $\gG$. That is, if:
\begin{equation}
    \phi(g \cdot x) = g \cdot \phi(x) \ \ \ \forall x \in \gX, \forall g \in \gS.
\end{equation}
Different from equivariance to a \textit{subgroup} of $\gG$ --a subset of the group that also fulfills the group axioms--, we do not restrict the subset $\gS$ to be itself a group. 

As explained in detail in Sec.~\ref{sec:9_from_gconvs_to_partial}, partial equivariance holds, in general, \textit{only approximately}, and it is exact only if $\gS$ is a subgroup of $\gG$. This results from the set $\gS$ not being necessarily closed under group actions. In other words, partial equivariance is a relaxation of group equivariance similar to \textit{soft invariance} \citep{van2018learning}: the property of a map to be approximately invariant. We opt for the word \textit{partial} in the equivariance setting to emphasize that (approximate) partial group equivariance arises by restricting the domain of the signals in a group convolution to a subset, i.e., a part, of the group. % in the output of the group convolution. 
%With this definition, we are now able to generalize the group convolution to be partially equivariant.
\vspace{-7mm}
\subsection{The partial group convolution}

Let $\gS^{(1)}, \gS^{(2)}$ be subsets of a group $\gG$ and $\mathrm{p}(u)$ be a probability distribution on the group, which is non-zero only on $\gS^{(2)}$. The partial group convolution from a function $f:\gS^{(1)} \rightarrow \sR$ to a function $h: \gS^{(2)} \rightarrow \sR$ is given by:
\begin{equation}\label{eq:9_partial_gconv}
    h(u) = (\psi * f)(u) = \int_{\gS^{(1)}} \mathrm{p}(u) \psi(v^{-1}u)f(v) \, \du \mu_{\gG}(v); \ u \in \gS^{(2)}, v \in \gS^{(1)}.
\end{equation}
In contrast to group convolutions whose inputs and outputs are always defined on the entire group, i.e., $f, h: \gG \rightarrow \sR$, the domain of the input and output of the partial group convolution can also be \textit{subsets of the group}. By learning these subsets, the model can become (\emph{i}) fully equivariant $\left(\gS^{(1)}, \gS^{(2)} {=}\gG\right)$, (\emph{ii}) partially equivariant $\left(\gS^{(1)}, \gS^{(2)} {\neq}\gG\right)$, or (\emph{iii}) forget some equivariances $\left( \gS^{(2)}\text{\ a subgroup\ of\ } \gG\right)$. %In practice, partial group convolutions are computed via a Monte Carlo approximation as described in the supplement. 
\vspace{-7mm}
\subsection{From group convolutions to partial group convolutions}\label{sec:9_from_gconvs_to_partial}

In this section, we show how group convolutions can be extended to describe partial equivariances. Vital to our analysis is the equivariance proof of the group convolution \citep{cohen2016group, cohen2019general}. In addition, we must distinguish between the domains of the input and output of the group convolution, i.e., the domains of $f$ and $h$ in Eq.~\ref{eq:9_gconv}. This distinction is important because they may be different for partial group convolutions. From here on, we refer to these as the \textit{\textbf{input domain}} and the \textbf{\textit{output domain}}. 

\begin{proposition}\label{prop:equivariance} Let $\gL_{w}f(u) {=} f(w^{-1} u)$. The group convolution is $\gG$-equivariant in the sense that:
 \begin{equation}\label{eq:9_g_equiv}
 \setlength{\abovedisplayskip}{0pt}
\setlength{\belowdisplayskip}{3pt}
     (\psi * \gL_{w}f)(u) =  \gL_{w}(\psi * f)(u), \ \text{for all\ } u, v, w \in \gG .
 \end{equation}
\end{proposition}
\vspace{-3mm}
\begin{proof} \cite{cohen2019general}
\begin{align*}
    \hspace{10mm}(\psi * \gL_{w}f)(u) &= \int_{\gG} \psi(v^{-1}u) f(w^{-1}v) \, \du \mu_{\gG}(v) = \int_{\gG} \psi(\Bar{v}^{-1}w^{-1}u) f(\Bar{v}) \, \du \mu_{\gG}(\Bar{v}) \\
    &= (\psi * f)(w^{-1}u) = \gL_{w}(\psi * f)(u).
\end{align*}
In the first line, the change of variables $\Bar{v} {\coloneqq} w^{-1}v$ is used. This is possible because the group convolution is a map from the group to itself, and thus if $w, v \in \gG$, so does $w^{-1}v$. Moreover, as the Haar measure is an invariant measure on the group, we have that $\mu_{\gG}(v) {=} \mu_{\gG}(\Bar{v})$, for all $v, \Bar{v} \in \gG$.
\end{proof}

\textbf{Going from the group $\gG$ to a subset $\gS$.} Crucial to the proof of Proposition~\ref{prop:equivariance} is the fact the group convolution is an operation from functions on the group to functions on the group. As a result, $w^{-1} u$ is a member of the output domain for any $w \in \gG$ applied to the input domain. Consequently, a group transformation applied to the input can be reflected by an equivalent transformation on the output.% That is, if $w$ and $u$ belong to $\gG$, so does $w^{-1}u$.

Now, consider the case in which the output domain is not the group $\gG$, but instead an arbitrary subset $\gS \subset \gG$, e.g., rotations in $[-\frac{\pi}{2}, \frac{\pi}{2}]$. Following the proof of Proposition~\ref{prop:equivariance} with $u \in \gS$, and $v \in \gG$, we observe that the operation is equivariant to transformations $w \in \gG$ as long as $w^{-1} u$ is a member of $\gS$. However, if $w^{-1} u$ does not belong to the output domain $\gS$, the output of the operation cannot reflect an equivalent transformation to that of the input, and thus equivariance is not guaranteed (Fig.~\ref{fig:9_illustrated_partialequiv}). By tuning the size of $\gS$, partial group convolutions can adjust their level of equivariance.

%\textbf{\textit{How approximate is partial group equivariance?}} 
Note that equivariance is \textit{only} obtained if Eq.~\ref{eq:9_g_equiv} holds for \textit{all} elements in the output domain. That is, if $w^{-1} u$ is a member of $\gS$, for all elements $u \in \gS$. For partial group convolutions, this is, in general, not the case as the output domain $\gS$ is not necessarily closed under group transformations. Nevertheless, we can precisely quantify how much the output response will change for any input transformation given an output domain $\gS$. Intuitively, this difference is given by the difference in the parts of the output feature representation that go in and out of $\gS$ by the action of input group transformations. The stronger the transformation and the smaller the size of $\gS$, the larger the equivariance difference in the output is (Fig.~\ref{fig:9_illustrated_partialequiv}). The formal derivations are provided in Appx.~\ref{appx:9_from_g_to_s}.

\textbf{Going from a subset $\gS^{(1)}$ to another subset $\gS^{(2)}$.} 
Now, consider the case in which the domain of the input and the output are both subsets of the group, i.e., $v \in \gS^{(1)}$ and $u \in \gS^{(2)}$. Analogous to the previous case, equivariance to input transformations $w \in \gG$ holds at positions $u \in \gS^{(2)}$ for which $w^{-1}u$ are also members of $\gS^{(2)}$. Nevertheless, the input domain is not longer restricted to be closed, and thus the input can also change for different group transformations. %, which in turn induces an additional equivariance difference in the output response.

\begin{figure*}
    \centering
    \includegraphics[width=1.0 \textwidth]{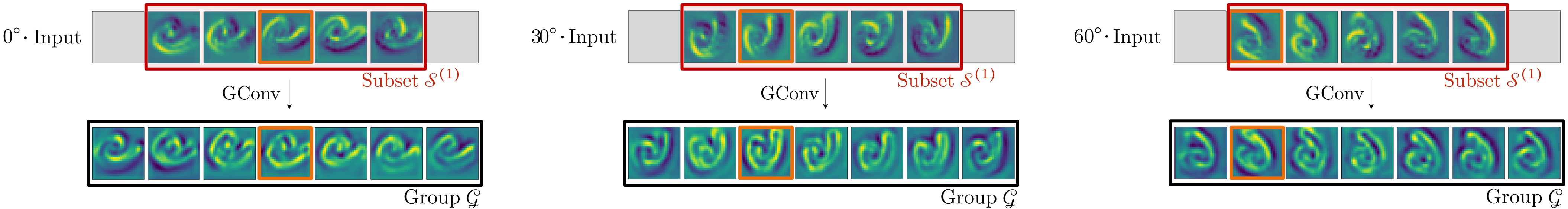}
    \vspace{-5mm}
    \caption{The effect of group subsets in the input domain. Partial group convolutions can receive input functions whose domain is not the group $\gG$ but a subset $\gS^{(1)}$, e.g., in a mid layer of a Partial G-CNN. Consequently, even if the output domain is the group, i.e., $\gS^{(2)}{=}\gG$, a partial group convolution does not produce exactly equivariant outputs by design. The difference from exact equivariance in the output response is proportional to the strength of the input transformation and the size of $\gS^{(1)}$.}
    \label{fig:9_subset_input}
\end{figure*}

To see this, consider a partial group convolution from an input subset $\gS^{(1)}$ to the group $\gG$ (Fig.~\ref{fig:9_subset_input}). Even if the output domain is the group, differences in the output feature map can be seen. This results from differences observed in the input feature map $f$ for different group transformations of the input.

%\textit{\textbf{Quantifying the equivariance difference.}} 
Similar to the previous case, we can precisely quantify how much the output response changes for an arbitrary subset $\gS^{(1)}$ in the input domain. Intuitively, the difference is given by the change in the parts of the input feature representation that go in and out of $\gS^{(2)}$ by the action of the input group transformation. The stronger the transformation and the smaller the size of $\gS^{(1)}$, the larger the difference in the output is (Fig.~\ref{fig:9_subset_input}). The formal treatment and derivation of this quantity is provided in Appx.~\ref{appx:9_from_s1_to_s2}.

\textbf{Special case: lifting convolutions.} To conclude, we note that the lifting convolution (Eq.~\ref{eq:9_lifting}) is a special case of a partial group convolution with $\gS^{(1)}{=}\gX$ and $\gS^{(2)}{=}\gG$. Note however, that the lifting convolution is fully group equivariant. This is because the domain of the input --$\gX$-- is closed under group transformations. Hence no input component leaves $\gX$ for group transformations of the input.
\vspace{-7mm}
\subsection{Learning group subsets through probability distributions on the group}

So far, we have discussed the properties of the partial group convolution in terms of group subsets without specifying how these subsets can be learned. Here, we describe how this can be done by learning a certain probability distribution on the group. We provide examples for discrete groups, continuous groups, and combinations thereof.

Vital to our approach is the Monte Carlo approximation to the group convolution presented in Sec.~\ref{sec:9_preliminaries}:
\begin{equation*}
    (\psi \hat{*} f)(u_i) = \sum\nolimits_{j} \psi(v_j^{-1}u_{i})f(v_j) \Bar{\mu}_{\gG}(v_j).
\end{equation*}
As shown in Appx.~\ref{appx:9_monte_carlo}, this approximation is equivariant to $\gG$ in expectation if the elements in the input and output domain are uniformly sampled from the Haar measure, i.e., $ u_i, v_j \sim \mu_{\gG}(\cdot)$.\footnote{\citet{finzi2020generalizing} show a similar result where $u_i$ and $v_j$ are the same points and thus $v_j \sim \mu_{\gG}(\cdot)$ suffices.}

\textbf{Approach.} Our main observation is that we can prioritize sampling specific group elements during the group convolution by learning a probability distribution $\mathrm{p}(u)$ over the elements of the group. When group convolutions draw elements uniformly from the group, each group element is drawn with equal probability and thus, the resulting approximation is fully equivariant in expectation. However, we can also define a different probability distribution that draws some samples with larger probability. For instance, we can sample from a certain region, e.g., rotations in $[-\frac{\pi}{2}, \frac{\pi}{2}]$, by defining a probability distribution on the group $\mathrm{p}(u)$ which is uniform in this range, but zero otherwise. The same principle can be used to forget an equivariance by letting this distribution collapse to a single point, e.g., the identity, along the corresponding group dimension.

In other words, learning a probability distribution $\mathrm{p}(u)$ on the group that is non-zero \textit{only} in a subset of the group can be used to effectively learn this subset. Specifically, we define a probability distribution $\mathrm{p}(u)$ on the output domain of the group convolution in order to learn a subset of the group $\gS^{(2)}$ upon which partial equivariance is defined. Note that we only need to define a distribution on the output domain of each layer. This is because neural networks apply layers sequentially, and thus the distribution on the output domain of the previous layer defines the input domain of the next layer.

\textbf{Distributions for one-dimensional continuous groups.} We take inspiration from Augerino \citep{benton2020learning}, an use the reparameterization trick \citep{kingma2013auto} to parameterize continuous distributions. In particular, we use the reparameterization trick on the Lie algebra of the group \citep{falorsi2019reparameterizing} to define a distribution which is uniform over a connected set of group elements\break $[u^{-1}, \dots, e, \dots, u]$, but zero otherwise. To this end, we define a uniform distribution $\gU(\mathfrak{u} \cdot [-1,1])$ with learnable $\mathfrak{u}$ on the Lie algebra $\mathfrak{g}$, and map it to the group via the push-forward of the exponential map $\mathrm{exp}: \mathfrak{g} \rightarrow \gG$. This give us a distribution which is uniform over a connected set of elements $[u^{-1}, \dots, e, \dots, u]$, but zero otherwise.\footnote{Note that an $\mathrm{exp}$-push-forwarded local uniform distribution is locally equivalent to the Haar measure, and thus we can still use the Haar measurement for integration on group subsets.}

For instance, we can learn a distribution on the rotation group $\mathrm{SO(2)}$, which is uniform between $[-\theta, \theta]$ and zero otherwise by defining a uniform probability distribution $\gU(\theta \cdot [-1, 1])$ with learnable $\theta$ on the Lie algebra, and mapping it to the group. If we parameterize group elements as scalars $g \in [-\pi, \pi)$, the exponential map is the identity, and thus $\mathrm{p}(g){=} \mathcal{U}(\theta \cdot [-1, 1))$. If we sample group elements from this distribution during the calculation of the group convolution, the output domain will only contain elements in $[-\theta, \theta)$ and the output feature map will be partially equivariant.

\textbf{Distributions for one-dimensional discrete groups.} We can define a probability distribution on a discrete group as the probability of sampling from all possible element combinations. For instance, for the mirroring group $\{1, -1\}$, this distribution assigns a probability to each of the combinations $\{0, 0\}$, $\{0, 1\}$, $\{1, 0\}$, $\{1, 1\}$ indicating whether the corresponding element is sampled (1) or not (0). For a group with elements $\{e, g_1, \dots, g_n\}$, however, this means sampling from $2^{n+1}$ elements, which is computationally expensive and potentially difficult to train. To cope with this, we instead define element-wise Bernoulli distributions over each of the elements $\{g_1, \dots, g_n\}$, and learn the probability $p_i$ of sampling each element $g_i$. The probability distribution on the group can then be formulated as the joint probability of the element-wise Bernoulli distributions $\mathrm{p}(e, g_1, \dots, g_n)=\prod_{i=1}^{n}\mathrm{p}(g_i)$. 

To learn the element-wise Bernoulli distributions, we use the Gumbel-Softmax trick \citep{jang2016categorical, maddison2016concrete}, and use the Straight-Through Gumbel-Softmax to back-propagate through sampling. If all the probabilities are equal to 1, i.e., $\{p_i{=}1\}_{i=1}^{n}$, the group convolution will be fully equivariant. Whenever probabilities start declining, group equivariance becomes partial, and, in the limit, if all probabilities become zero, i.e., $\{p_i{=}0\}_{i=1}^{n}$, then only the identity is sampled and this equivariance is effectively forgotten. 

\textbf{Probability distributions for multi-dimensional groups.} There exist several multi-dimensional groups with important applications, such as the orthogonal group $\mathrm{O(2)}$ --parameterized by rotations and mirroring--, or the dilation-rotation group --parameterized by scaling and rotations--. 

For multi-dimensional groups, we parameterize the probability distribution over the entire group as a combination of \textit{independent probability distributions along each of the group axes}. For a group $\gG$ with elements $g$ decomposable along $n$ dimensions $g{=}(d_1, ..., d_n)$, we decompose the probability distribution as: $\mathrm{p}(g){=}\prod_{i=1}^{n}\mathrm{p}(d_i)$, where the probability $\mathrm{p}(d_i)$ is defined given the type of space -- continuous or discrete--. For instance, for the orthogonal group $\mathrm{O(2)}$ defined by rotations $r$ and mirroring $m$, i.e., $g = (r,m)$, $r \in \mathrm{SO(2)}, m \in \{\pm1\}$, we define the probability distribution on the group as $\mathrm{p}(g){=}$ $\mathrm{p}(r)\cdot\mathrm{p}(m)$, where $\mathrm{p}(r)$ is a continuous distribution, and $\mathrm{p}(m)$ is a discrete one as defined above. 
\begin{algorithm}[t]
\caption{The Partial Group Convolution Layer}
\label{algo:patial_g_conv}
\begin{algorithmic}[1]
\State {\bfseries Inputs:}  position, function-value tuples on the group or a subset thereof $\{ v_j, f(v_j)\}$.
\State {\bfseries Outputs:}  convolved position, function-value tuples on the output group subset $\{u_i, (f\hat{*} \psi)(u_i)\}$.
\State $\{u_i\}\sim \mathrm{p}(u)$ \Comment{\texttt{Sample elements from p(u)}}
\For{$u_i \in \{u_i\}$}
    \State $h(u_i) = \sum_j \psi(v_j^{-1}u_i)f(v_j) \bar{\mu}_{\gG}(v_j)$ \Comment{\texttt{Compute group convolution}}
\EndFor
\State {\bfseries Return:}  $\{u_i, h(u_i)\}$
\end{algorithmic}
\end{algorithm}

\vspace{-7mm}
\subsection{Partial Group Equivariant Networks}

\begin{figure}
\captionsetup{justification=centering}
    \centering
    \includegraphics[width=0.4\textwidth]{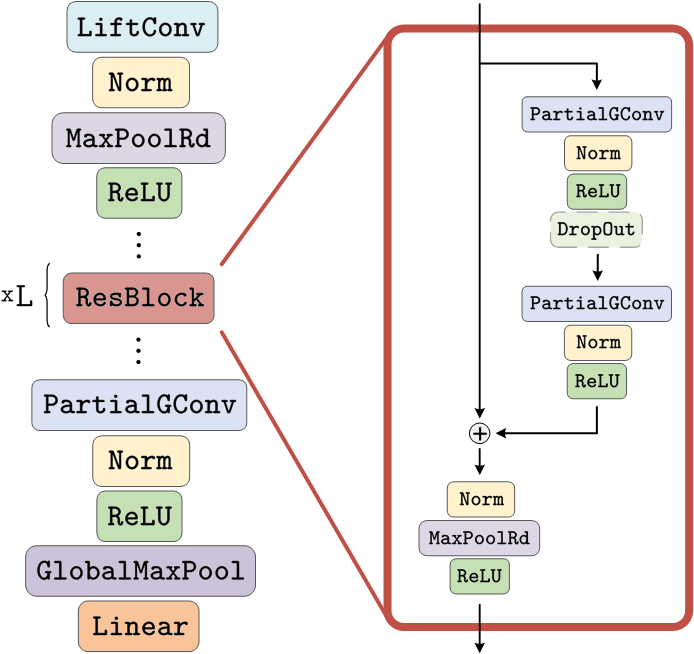}
    \caption{The Partial G-CNN.}
    \label{fig:9_network_arch}
\end{figure}

To conclude this section, we illustrate the structure of Partial G-CNNs. We build upon \citet{finzi2020generalizing} and extend their continuous G-CNNs to discrete groups. This is achieved by parameterizing the convolutional kernels on (continuous) Lie groups on their Lie algebra, and applying the action of discrete groups directly on the group representation of the kernels.

In addition, we replace the isotropic lifting of \citet{finzi2020generalizing} with lifting convolutions (Eq.~\ref{eq:9_lifting}). Inspired by \citet{romero2022ckconv}, we parameterize convolutional kernels as implicit neural representations with SIRENs \citep{sitzmann2020implicit}. This parameterization leads to higher expressivity, faster convergence, and better accuracy than $\mathrm{ReLU}$, \leakyrelu\ and $\mathrm{Swish}$ \texttt{MLP}\ parameterizations used so far for continuous G-CNNs, e.g., \cite{schutt2017schnet, finzi2020generalizing} --see Tab.~\ref{tab:9_kernel_comparison}, \cite{knigge2022exploiting}-. The architecture of Partial G-CNNs is shown in Fig.~\ref{fig:9_network_arch}.
\vspace{-7mm}
\section{Related work}
\textbf{Group equivariant neural networks.} The seminal work of G-CNNs \citep{cohen2016group} has inspired several methods equivariant to many different groups. Existing methods show equivariance to planar \citep{dieleman2016exploiting, worrall2017harmonic, weiler2018learning}, spherical rotations \citep{weiler20183d,cohen2018spherical, esteves2019cross,esteves2019equivariant,esteves2020spin}, scaling \citep{worrall2019deep, sosnovik2020scaleequivariant, romero2020wavelet}, and other symmetry groups \citep{romero2020attentive, bogatskiy2020lorentz, finzi2021practical}. Group equivariant self-attention has also been proposed \citep{fuchs2020se, romero2021group, hutchinson2021lietransformer}. Common to all these methods is that they are fully equivariant and the group must be fixed prior to training. Contrarily, Partial G-CNNs learn their level of equivariance from data and can represent full, partial and no equivariance.% , and are able to do during training is able to define partial equivariances  method allows for partial equivariances and us to define layers approximately equivariant to subsets of the group, and to learn these subsets during training.

\textbf{Invariance learning.}  Learning the right amount of global invariance from data has been explored by learning a probability distribution over continuous test-time augmentations \citep{benton2020learning}, or by using the marginal likelihood of a Gaussian process \citep{van2018learning, ouderaa2021}. Contrary to these approaches, Partial G-CNNs aim to learn the right level of equivariance at every layer and do not require additional loss terms. Partial G-CNNs relate to Augerino \citep{benton2020learning} in the form that probability distributions are defined on continuous groups. However, Partial G-CNNs are intended to learn partial layer-wise equivariances and are able to learn probability distributions on discrete groups. There also exist learnable data augmentation strategies \citep{lim2019fast, hataya2020faster, li2020dada, chatzipantazis2021learning} that can find transformations of the input that optimize the task loss. We can also view Augerino as learning a smart data augmentation technique which we compare with. In contrast to these methods, Partial G-CNNs find optimal partial equivariances at each layer.

\textbf{Equivariance learning.} Learning equivariant mappings from data has been explored by meta-learning of weight-tying matrices encoding symmetry equivariances \citep{zhou2020meta, alet2021noether} and by learning the Lie algebra generators of the group jointly with the parameters of the network \citep{dehmamy2021lie}. These approaches utilize the same learned symmetries across layers. MSR \citep{zhou2020meta} is only applicable to (small) discrete groups, and requires long training times. L-Conv \citep{dehmamy2021lie} is only applicable to continuous groups and is not fully compatible with current deep learning components, e.g., pooling, normalization. Unlike these approaches, Partial G-CNNs can learn levels of equivariance at every layer, are fully compatible with current deep learning components, and are applicable for discrete groups, continuous groups and combinations thereof. We note, however, that \citet{zhou2020meta, dehmamy2021lie} learn the structure of the group from scratch. Contrarily, Partial G-CNNs start from a (very) large group and allows layers in the network to constrain their equivariance levels to better fit the data. \citet{finzi2021residual} incorporate soft equivariance constraints by combining outputs of equivariant and non-equivariant layers running in parallel, which incurs in large parameter and time costs. Differently, Partial G-CNNs learns from data the level of partial equivariance directly on the group manifold.
\vspace{-7mm}
\section{Experiments}

\textbf{Experimental details.} We parameterize all our convolutional kernels as 3-layer SIRENs \citep{sitzmann2020implicit} with 32 hidden units. All our networks --except for the (partial) group equivariant 13-layer CNNs \citep{laine2016temporal} used in Sec.~\ref{sec:9_additional_experiments}-- are constructed with 2 residual blocks of 32 channels each, batch normalization \citep{ioffe2015batch} following the structure shown in Fig.~\ref{fig:9_network_arch}. Here, we intentionally select our networks to be simple as to better assess the effect of partial equivariance. We avoid learning probability distributions on the translation part of the considered groups, and assume all spatial positions to be sampled in order to use fast \texttt{PyTorch} convolution primitives in our implementation. %\footnote{We plan to release our code publicly at \url{github.com/merlresearch/partial-gcnn}} 
Additional experimental details, e.g., the specific hyperparameters used, as well as complementary results can be found in Appx.~\ref{appx:9_exp_details_sec},~\ref{appx:9_extra_results}.
\begin{table}
\RawFloats
\centering
\begin{minipage}{0.5 \textwidth}
\begin{center}
    \caption{Results on MNIST6-180 and MNIST6-M.}
\label{tab:9_toy_tasks}
\vspace{-3mm}
\begin{small}
\scalebox{0.75}{
    \centering
    \begin{tabular}{cccc}
    \toprule
     \sc{Base Group} & \sc{Dataset} & \sc{G-CNN} & \sc{Partial G-CNN} \\
    \midrule
       $\mathrm{SE(2)}$ & MNIST6-180  & 50.0 & \textbf{100.0} \\
       Mirroring & MNIST6-M &  50.0 & \textbf{100.0} \\
       \midrule
       \multirow{2}{*}{$\mathrm{E(2)}$} & MNIST6-180 &  50.0 & \textbf{100.0} \\
       & MNIST6-M  & 50.0 & \textbf{100.0} \\
    \bottomrule
    \end{tabular}
    }
    \end{small}
\end{center}
\end{minipage}%
\hfill
\begin{minipage}{0.49 \textwidth}
\begin{center}
\caption{Augerino vs. Partial G-CNNs. Terms in parentheses show accuracy of Partial G-CNNs.}
\label{tab:9_augerino}
\vspace{-3mm}
\begin{small}
\scalebox{0.75}{
\begin{tabular}{ccccc}
\toprule
\multirow{2}{*}{\sc{\shortstack{Base\\Group}}} & \multirow{2}{*}{\sc{\shortstack{No.\\ Elems}}}  & \multicolumn{3}{c}{\sc{Classification Accuracy (\%)}} \\
         \cmidrule{3-5}
&  &  \sc{RotMNIST} & \sc{CIFAR10} & \sc{CIFAR100} \\
\toprule
\multirow{2}{*}{$\mathrm{SE(2)}$} & 8 & 99.17 (\textbf{99.23}) & 82.38 (\textbf{88.59}) & 52.98 (\textbf{57.26}) \\
 & 16 & \textbf{99.25} (99.18) & 82.53 (\textbf{88.59}) & 51.47 (\textbf{57.31}) \\
\midrule
\multirow{2}{*}{$\mathrm{E(2)}$} & 8 & \textbf{99.12} (97.78) & 84.29 (\textbf{89.00}) & 52.59 (\textbf{55.22}) \\
 & 16 & \textbf{99.18} (98.35) & 83.54 (\textbf{90.12}) & 54.76 (\textbf{61.46}) \\
\bottomrule
\end{tabular}
}
\end{small}
\end{center}
\end{minipage}
\end{table}

\textbf{Toy tasks: MNIST6-180 and MNIST6-M.} First, we validate whether Partial G-CNNs can learn partial equivariances. To this end, we construct two illustrative datasets: \textit{MNIST6-180}, and \textit{MNIST6-M}. \textit{MNIST6-180} is constructed by extracting the digits of the class 6 from the MNIST dataset \citep{lecun1998gradient}, and rotating them on the circle. The goal is to predict whether the number is a six, i.e., a rotation in $[-90^{\circ}, 90^\circ]$ was applied, or a nine otherwise. Similarly, we construct \textit{MNIST6-M} by mirroring digits over the y axis. The the goal is to predict whether a digit~was~mirrored~or~not.

As shown in Tab.~\ref{tab:9_toy_tasks}, G-CNNs are unable to solve these tasks as discrimination among group transformations is required. Specifically, $\mathrm{SE(2)}$-CNNs are unable to solve MNIST6-180, and $\mathrm{Mirror}$-CNNs --G-CNNs equivariant to reflections-- are unable to solve MNIST6-M. Furthermore, $\mathrm{E(2)}$-CNNs cannot solve any of the two tasks, because $\mathrm{E(2)}$-CNNs incorporate equivariance to both rotations and reflections. Partial G-CNNs, on the other hand, easily solve both tasks with corresponding base groups. This indicates that Partial G-CNNs learn to adjust the equivariance levels in order to solve the tasks. 

In addition, we verify the learned levels of equivariance for a Partial $\mathrm{SE(2)}$-CNN on MNIST6-180. To this end, we plot the probability of assigning the label $6$ to test samples of MNIST6-180 rotated on the whole circle. Fig~\ref{fig:9_rotated_input} shows that the network learns to predict \enquote{6} for rotated samples in $[-90^{\circ}, 90^\circ]$, and \enquote{9} otherwise. Note that Partial G-CNNs learn the expected levels of partial equivariance without any additional regularization loss terms to encourage them --as required in \citet{benton2020learning}--.

\textbf{Benchmark image datasets.} Next, we validate Partial G-CNNs on classification datasets: RotMNIST \citep{larochelle2007empirical}, CIFAR-10 and CIFAR-100 \citep{krizhevsky2009learning}. Additional results on the PatchCam dataset \citep{veeling2018rotation} can be found in Appx.~\ref{appx:9_extra_results}.

We construct Partial G-CNNs with base groups $\mathrm{SE(2)}$ and $\mathrm{E(2)}$, and varying number of elements used in the Monte Carlo approximation of the group convolution and compare them to equivalent G-CNNs and ResNets (equivalent to $\mathrm{T(2)}$-CNNs). Our results (Tab.~\ref{tab:9_vision_tasks}) show that Partial G-CNNs are competitive with G-CNNs when full-equivariance is advantageous (rotated MNIST and PatchCamelyon). However, for tasks in which the data does not naturally exhibit full rotation equivariance (CIFAR-10 and CIFAR-100), Partial G-CNNs consistently outperform fully equivariant G-CNNs.

% \subsection{Ablation studies}

\textbf{The need for learning partial layer-wise equivariances.} Next, we evaluate (\emph{i}) the effect of learning partial equivariances instead of soft invariances, and (\emph{ii}) the effect of learning layer-wise levels of equivariances instead of a single level of partial equivariance for the entire network. 

For the former, we compare Partial G-CNNs to equivalent ResNets with Augerino \cite{benton2020learning} (Tab.~\ref{tab:9_augerino}). We extend our strategy to learn distributions on discrete groups to the Augerino framework to allow it to handle groups with discrete components, e.g., $\mathrm{E(2)}$. For the latter, we construct regular G-CNNs and replace either the final group convolutional layer by a $\mathrm{T(2)}$ convolutional layer, or the global max pooling layer at the end by a learnable \texttt{MLP} (Tab.~\ref{tab:9_final_layer_T2}). If determining the level of equivariance at the end of the network is sufficient, these models should perform comparably to Partial G-CNNs.

\begin{table}
\RawFloats
\centering
\begin{minipage}{0.51 \textwidth}
\begin{center}
\caption{Test image classification results.}
\label{tab:9_vision_tasks}
\vspace{-3mm}
\begin{small}
\scalebox{0.72}{
    \centering
    \begin{tabular}{cccccc}
    \toprule
         \multirow{2}{*}{\sc{\shortstack{Base\\ Group}}} & \multirow{2}{*}{\sc{\shortstack{No.\\ Elems}}} & \multirow{2}{*}{\sc{\shortstack{Partial\\ Equiv.}}}  & \multicolumn{3}{c}{\sc{Classification Accuracy (\%)}} \\
         \cmidrule{4-6}
        
        &  &  & \sc{RotMNIST} & \sc{CIFAR10} & \sc{CIFAR100} \\ 
        
        \toprule
        $\mathrm{T(2)}$ & 1 & - & 97.23 & 83.11 & 47.99\\
        \midrule
         
        \multirow{6}{*}{$\mathrm{SE(2)}$} & \multirow{2}{*}{4} & \xmark & 99.10 &  83.73 & 52.35 \\ 
         
        & & \cmark & \textbf{99.13} & \textbf{86.15} &  \textbf{53.91} \\
         \cmidrule{2-6}
         
        & \multirow{2}{*}{8} & \xmark & 99.17 & 86.08 & 55.55\\
        
        & & \cmark & \textbf{99.23} & \textbf{88.59} & \textbf{57.26}\\
        \cmidrule{2-6}
        
        & \multirow{2}{*}{16}  & \xmark & \underline{\textbf{99.24}} & 86.59 & 51.55 \\
        
        & & \cmark & 99.18 & \textbf{89.11} & \textbf{57.31} \\
        \midrule
        
        \multirow{4}{*}{$\mathrm{E(2)}$} & \multirow{2}{*}{8} & \xmark & \textbf{98.14} & 85.55 & 54.29\\
        
        & & \cmark & 97.78 & \textbf{89.00} & \textbf{55.22} \\
        \cmidrule{2-6}
        
        & \multirow{2}{*}{16}  & \xmark & 98.35 & 88.95 & 57.78 \\
        
        & & \cmark & \textbf{98.58} & \underline{\textbf{90.12}} &  \underline{\textbf{61.46}} \\
        \bottomrule
    \end{tabular}
    }
\end{small}
\end{center}
\end{minipage}%
\hfill
\begin{minipage}{0.43 \textwidth}
\begin{center}
    \includegraphics[width=\textwidth]{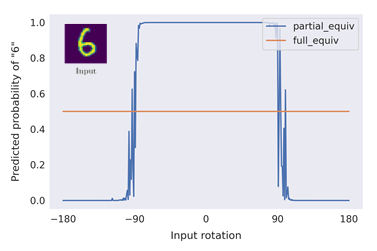}
    \captionof{figure}{Learned equivariances for a ``6" on MNIST6-180. Partial G-CNNs become equivariant to rotations on the semi-circle, while G-CNNs are unable to solve the task.}
    \label{fig:9_rotated_input}
\end{center}
\end{minipage}
\end{table}

Tab.~\ref{tab:9_augerino} shows that Augerino is competitive to Partial G-CNNs on rotated MNIST, but falls behind by a large margin on CIFAR-10 and CIFAR-100. This result can be explained by how these datasets are constructed. RotMNIST is constructed by rotating MNIST digits globally, thus it is not surprising that a model able to encode global invariances can match Partial G-CNNs. The invariance and equivariance relationships in natural images, however, are more complex, as they can be local as well. Consequently, tackling different levels of equivariance at each layer using Partial G-CNNs leads to benefits over using a single level of global invariance for the entire network.

Although the $\mathrm{T(2)}$ and \texttt{MLP} alternatives outlined before could solve the MNIST-180 and MNIST-M toy datasets, we observed that Partial G-CNNs perform consistently better on the visual benchmarks considered (Tab.~\ref{tab:9_final_layer_T2}). This indicates that learning layer-wise partial equivariances is beneficial over modifying the level of equivariance only at the end of the model. In addition, it is important to highlight that Partial G-CNNs can become fully-, partial-, and non-equivariant during training. Alternative models, on the other hand, are either unable to become fully equivariant ($\mathrm{T(2)}$ models) or very unlikely to do so in practice (\texttt{MLP} models). 

\textbf{SIRENs as group convolution kernels.} Next, we validate SIRENs as parameterization for group convolutional kernels. Tab.~\ref{tab:9_kernel_comparison} shows that $\mathrm{SE(2)}$-CNNs with SIREN kernels outperform $\mathrm{SE(2)}$-CNNs with \relu, \leakyrelu\ and \swish\ kernels by a large margin on all datasets considered. This result suggests that SIRENs are indeed better suited to represent continuous group convolutional kernels.
\vspace{-7mm}
\subsection{Experiments with deeper networks}\label{sec:9_additional_experiments}

In addition to the simple networks of the previous experiments, we also explore partial equivariance in a group equivariant version of the 13-layer CNN of \citet{laine2016temporal}. Specifically, we construct partial group equivariant 13-layer CNNs using $\mathrm{SE(2)}$ as base group, and vary the number of elements used in the Monte Carlo approximation of the group convolution. For each number of elements, we compare Partial 13-layer G-CNNs to their fully equivariant counterparts as well as equivalent 13-layer CNNs trained with Augerino. Our results are summarized in Tab.~\ref{tab:9_13layercnn}. 
\begin{table}
\RawFloats
\centering
\begin{minipage}{0.55 \textwidth}
\begin{center}
\caption{Accuracy of G-CNNs with a \texttt{MLP} instead of global pooling, G-CNNs with a final $\mathrm{T(2)}$-convolutional layer, and Partial G-CNNs.}
\label{tab:9_final_layer_T2}
\vspace{-3mm}
\begin{small}
\scalebox{0.75}{
\centering
\begin{tabular}{cccccc}
\toprule
\multirow{2}{*}{\sc{\shortstack{Base\\Group}}} & \multirow{2}{*}{\sc{\shortstack{No.\\ Elems}}} & \multirow{2}{*}{\sc{\shortstack{Net\\ Type}}} & \multicolumn{3}{c}{\sc{Classification Accuracy (\%)}} \\
         \cmidrule{4-6}
&  & & \sc{RotMNIST} & \sc{CIFAR10} & \sc{CIFAR100} \\
\toprule
\multirow{3}{*}{$\mathrm{SE(2)}$} & \multirow{3}{*}{16} & $\mathrm{T(2)}$ & 99.04	& 82.76	& 52.51 \\
 &  & \sc{MLP} &  99.00 & 86.25 & 56.29 \\
&  & \sc{Partial} &  \textbf{99.18} & \textbf{87.45} & \textbf{57.31} \\
\midrule
\multirow{3}{*}{$\mathrm{E(2)}$} & \multirow{3}{*}{16}  & $\mathrm{T(2)}$ & 97.98 & 86.68 & 57.61 \\
&  & \sc{MLP} & \textbf{99.02} & 87.43 & 58.87 \\
&  & \sc{Partial} & 98.58 & \textbf{90.12} &  \textbf{61.46}\\
\bottomrule
\end{tabular}
}
\end{small}
\end{center}
\end{minipage}%
\hfill
\begin{minipage}{0.41 \textwidth}
\begin{center}
    \includegraphics[width=\textwidth]{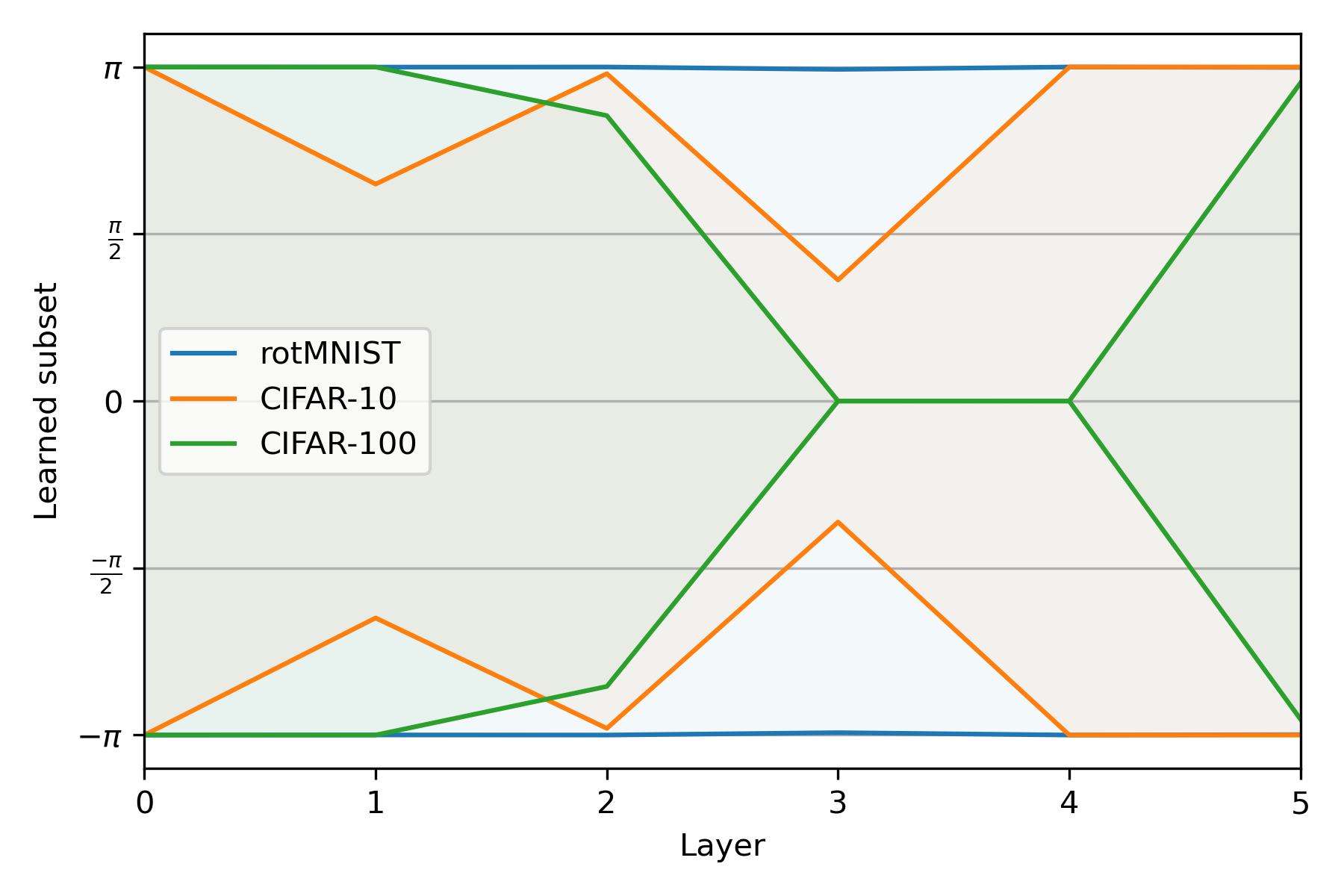}
    \captionof{figure}{Example group subsets learned by Partial G-CNNs.}
    \label{fig:9_learned_subsets}
\end{center}
\end{minipage}
\end{table}

\begin{figure}
    \centering
    \includegraphics[width=0.5\textwidth]{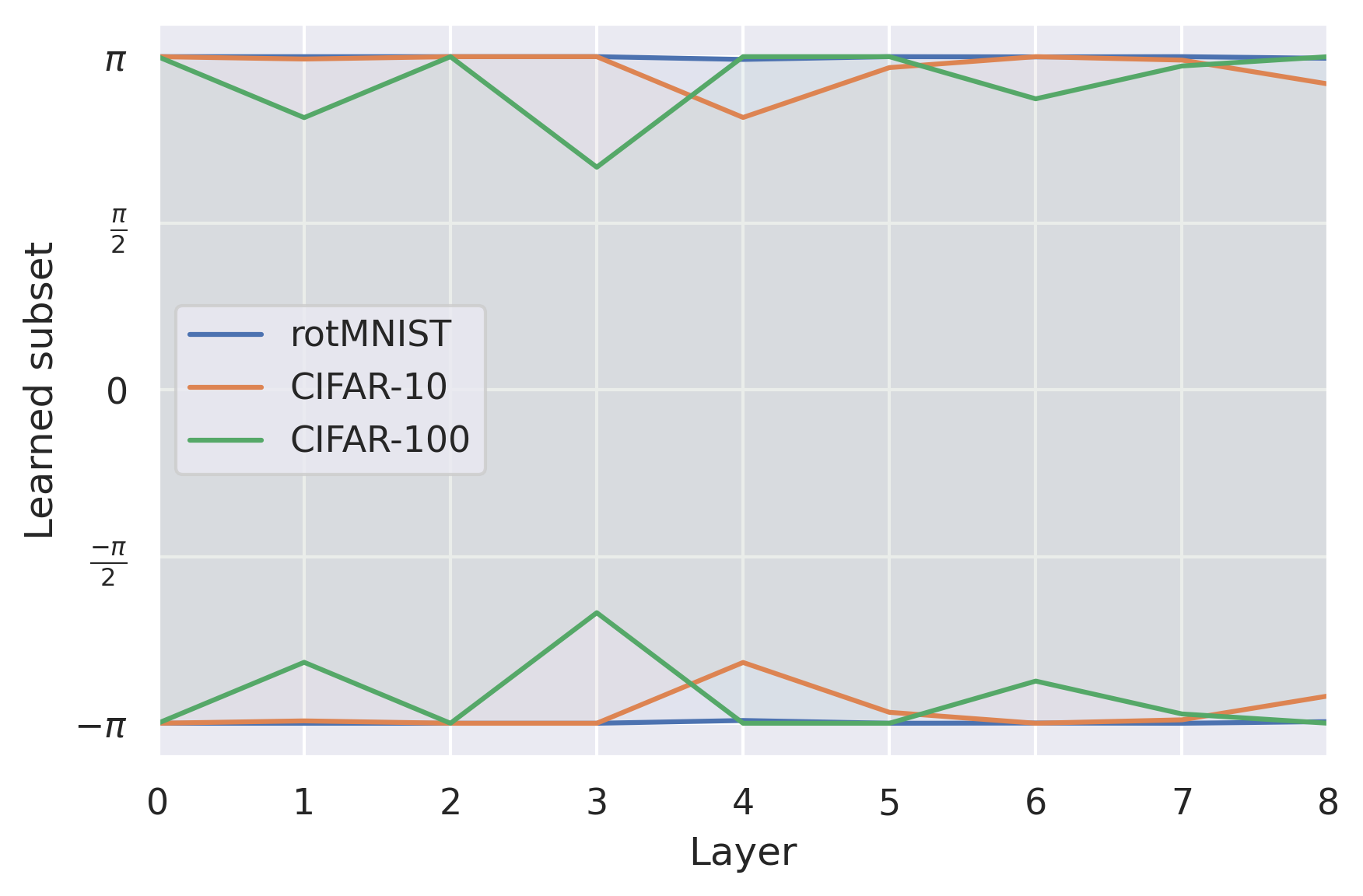}
    \vspace{-4mm}
    \captionsetup{justification=centering}
    \caption{Group subsets learned by 13-layer Partial G-CNNs.
    \vspace{-2mm}}
    \label{fig:9_subsets_13layer}
\end{figure}

On partial equivariant settings (CIFAR10 and CIFAR100), partial equivariant networks consistently outperform fully equivariant and Augerino networks for all number of elements used. Interestingly, translation equivariant CNNs outperform CNNs equivariant to $\mathrm{SE(2)}$ on CIFAR10 and CIFAR100. This illustrates that overly restricting equivariance constraints can degrade accuracy. In addition, partial equivariant CNNs retain the accuracy of full equivariant networks in fully equivariant settings. By looking at the group subsets learned by partial equivariant networks (Fig~\ref{fig:9_subsets_13layer}), we corroborate that partial equivariant networks learn to preserve full equivariance if full equivariance is advantageous, and learn to disrupt it otherwise.
\vspace{-7mm}
\section{Discussion}\label{sec:9_discussion}

\textbf{Memory consumption in partial equivariant settings.} G-CNNs fix the number of samples used to approximate the group convolution prior to training. In Partial G-CNNs we fix a maximum number of samples and adjust the number used at every layer based on the group subset learned. %We fix a maximum number of samples prior to training and adjust the number of samples used at every layer based on the subset of the group learned. 
Consequently, a partial group convolution with a learned distribution $\mathrm{p}(u){=}\gU(\frac{\pi}{2} [-1, 1])$ uses half of the elements used in a corresponding group convolution. This reduction in memory and execution time leads to improvements in training and inference time for Partial G-CNNs on partial equivariant settings. We observe reductions up to 2.5$\times$ in execution time and memory on CIFAR-10 and CIFAR-100.

\textbf{Sampling per batch element.} In our experiments, we sample once from the learned distribution $\mathrm{p}(u)$ at every layer, and use this sample for all elements in the batch. A better estimation of $\mathrm{p}(u)$ can be obtained by drawing a sample per batch element. Though this method may lead to faster convergence and better estimations of the learned distributions, it comes at a prohibitive memory cost resulting from independent convolutional kernels that must be rendered for each batch element. Consequently, we use a single sample per batch at each layer in our experiments.

\textbf{Better kernels with implicit neural representations.} We replace \leakyrelu\,, $\mathrm{ReLU}$ and $\mathrm{Swish}$ kernels used so far for continuous group convolution kernels with a SIREN \cite{sitzmann2020implicit}. Our results show that SIRENs are better at modelling group convolutional kernels than existing alternatives.

\textbf{Going from a small group subset to a larger one. What does it mean and why is it advantageous?} In Sec.~\ref{sec:9_from_gconvs_to_partial} we described that a partial group convolution can go from a group subset $\gS^{(1)}$ to a larger group subset $\gS^{(2)}$, e.g., the whole group $\gG$. Nevertheless, once a layer becomes partially equivariant, subsequent layers cannot become fully equivariant even for $\gS^{(2)}{=}\gG$. Interestingly, we observe that Partial G-CNNs often learn to disrupt equivariance halfway in the network, and return to the whole group afterwards (Fig.~\ref{fig:9_learned_subsets}). As explained below, this behavior is actually advantageous.%Doing so actually brings advantages as~described~below.
% \begin{wrapfigure}{r}{0.5\textwidth}
%     \centering
%     \includegraphics[width=\textwidth]{learned_subsets.png}
%     \caption{Example group subsets learned by Partial G-CNNs.}
%     \label{fig:9_learned_subsets}
% \end{wrapfigure}
% \vspace{-2mm}

Full equivariance restricts group convolutions to apply the same mapping on the entire group. As a result, once the input is transformed, the output remains equal up to the same group transformations. In partial equivariance settings, Partial G-CNNs can output different feature representations for different input transformations. Consequently, Partial G-CNNs can use the group dimension to encode different feature mappings. Specifically, some kernel values are used for some input transformations and other ones are used for other input transformations. This means that when Partial G-CNNs go back to a larger group subset from a smaller one, they are able to use the group axis to encode transformation-dependent features, resulting in increased model expressivity.% of the partial equivariant model.

\begin{table}
\RawFloats
\centering
\begin{center}
\caption{Classification accuracy with (partial) group equivariant 13-layer CNNs \citep{laine2016temporal}.}
\label{tab:9_13layercnn}
\vspace{-3mm}
\begin{small}
\scalebox{0.75}{
    \centering
    \begin{tabular}{ccccccc}
    \toprule
         \multirow{2}{*}{\sc{\shortstack{Base\\ Group}}} & \multirow{2}{*}{\sc{\shortstack{No.\\ Elems}}} & \multirow{2}{*}{\sc{\shortstack{Partial\\ Equiv.}}} & \multirow{2}{*}{\sc{Augerino}} & \multicolumn{3}{c}{\sc{Classification Accuracy (\%)}} \\
         \cmidrule{5-7}
        
        &  &  & & \sc{RotMNIST} & \sc{CIFAR10} & \sc{CIFAR100} \\ 
        
        \toprule
        $\mathrm{T(2)}$ & 1 & - & - & 96.90 & 91.21 &  67.14 \\
        \midrule
         
        \multirow{9}{*}{$\mathrm{SE(2)}$} & \multirow{3}{*}{2} & \multirow{2}{*}{\xmark} & \xmark & 98.70 &  85.51 &  62.06 \\ 
         
        & & & \cmark & \textbf{98.94} & 87.78 & 65.79\\
        %\cmidrule{3-7}
        & & \cmark &  - & 98.72 & \textbf{\underline{92.48}} & \textbf{66.72} \\
         \cmidrule{2-7}

        & \multirow{3}{*}{4} & \multirow{2}{*}{\xmark} & \xmark & 98.43 & 89.73  & 65.97 \\
        
        & & & \cmark & \textbf{98.94} & 91.66 & 68.99 \\
        & & \cmark & - & 98.78 & \textbf{92.28} & \textbf{69.83} \\
        \cmidrule{2-7}
         
        & \multirow{3}{*}{8} & \multirow{2}{*}{\xmark} & \xmark & 98.54  & 90.55 & 67.70 \\
        
        & & & \cmark & \textbf{99.28} & 89.96 & 69.66\\
        & & \cmark & - & 98.77 & \textbf{91.99} & \textbf{\underline{70.80}} \\
        \bottomrule
    \end{tabular}
    }
\end{small}
\vspace{-2mm}
\end{center}
\end{table}
In Appx.~\ref{appx:9_extra_results}, we evaluate the effect of enforcing monotonically decreasing group subsets as a function of depth. That is, Partial G-CNNs whose subsets at deeper layers are equal or smaller than those at previous ones. Our results show that this monotonicity leads to slightly worse results compared to the unconstrained case, thus supporting the use of unconstrained learning of group subsets.

\vspace{-7mm}
\section{Limitations and future work}
\textbf{Partial equivariances for other group representations.} The theory of learnable partial equivariances proposed here is only applicable to architectures using regular group representations, e.g., \citep{cohen2016group, romero2021group}. Nevertheless, other type of representations exist with which exact equivariance to continuous groups can be obtained: irreducible representations \citep{worrall2017harmonic, weiler2018learning, weiler2019general}. We consider extending the learning of partial equivariances to irreducible representations a valuable extension of our work.% However, this is a non-trivial endeavor which we leave out for future work.

\textbf{Unstable training on discrete groups.} Although we can model partial equivariance on discrete groups with our proposed discrete probability distribution parameterization, we observed that these distributions can be unstable to train. To cope with this, we utilize a 10x lower learning rate for the parameters of the probability distributions (See Appx.~\ref{appx:9_exp_details} for details). Nevertheless, finding good ways of learning discrete distributions is an active field of research \citep{hoogeboom2019integer, hoogeboom2021argmax}, and advances in this field could be used to further improve the learning of partial equivariances on discrete groups.

\textbf{Scaling partial equivariance to large groups.} Arguably the main limitation of G-CNNs with regular representations is their computational and memory complexity, which prevents the use of very large groups, e.g., simultaneous rotation, scaling, mirroring and translations. Partial equivariance is particularly promising for large groups as the network is initialized with a prior towards being equivariant to the entire group, but is able to focus on those relevant to the task at hand.  We consider learning partial equivariances on large groups an interesting direction for further research which orthogonal to other advances to scale group convolutions to large groups, e.g., via separable group convolutional kernels \citep{lengyel2021exploiting, knigge2022exploiting}.
%\textbf{Future work.} Our method relies on a good parameterization of the probability distribution defined over group elements. Finding better representations and learning methods for these distributions, specially for distributions on discrete variables, is an important direction for future research. Other interesting avenues include the extension of partial equivariances to irreducible group representations, and the definition of input-dependent partial equivariances.

%% file: chapters/conclusion.tex
% path to figures directory
\graphicspath{{figures/0-meta/}}

%=========================================================================

% \begin{savequote}[75mm]
% Nulla facilisi. In vel sem. Morbi id urna in diam dignissim feugiat. Proin molestie tortor eu velit. Aliquam erat volutpat. Nullam    ultrices, diam tempus vulputate egestas, eros pede varius leo.
% \qauthor{Quoteauthor Lastname}
% \end{savequote}

\chapter{Conclusion and Future Work} \label{chapter:conclusion}
This dissertation aimed to study the role of continuous modeling and symmetry preservation in improving Deep Learning efficiency among its different efficiency aspects. Our novel contributions affirm that both continuous modeling and symmetry preservation can be used to improve Deep Learning efficiency widely amongst its diverse aspects, while simultaneously improving the state-of-the-art in many cases. Nevertheless, we find that the benefits of symmetry preservation --particularly on data and parameter efficiency-- impose compromises in other efficiency aspects, specially on the computational efficiency side. However, as discussed below under \textit{Research Question 2}, it is worth noting that the increased computational costs are tied to interpretation. For convenience, we replicate Tab~\ref{tab:0_efficiency_contribution} here summarizing the efficiency contribution of each chapter:\\
{\color{White}space}

% \begin{table}[h]
% \captionsetup{justification=centering}
% \caption{Copy of Tab.~\ref{tab:0_efficiency_contribution}. Efficiency contributions as per Sec.~\ref{sec:0_efficiency_aspects} of each chapter.%\footnote{Note that Tab.~\ref{tab:0_efficiency_contribution} does not depict the environmental and financial efficiency aspects defined in Sec.~\ref{sec:0_efficiency_aspects}. From a technical standpoint, enhancements in environmental and financial efficiently are implicitly driven by advancements across the remaining efficiency axes. Consequently, we do not explicitly add them here. However, we consider important to acknowledge the importance of these efficiency aspects, even more as Deep Learning continuous to permeate everyday tools and applications.}
% }
\begin{center}
    \scalebox{0.9}{
    \begin{tabular}{cccccc}
    \toprule
         & & Compute  & Data & Parameter & Design  \\
         & & Efficiency & Efficiency & Efficiency & Efficiency \\
         \midrule
       \multirow{5}{*}{\rotatebox{90}{\thead{\sc{Part I:} \\ \sc{Continuous} \\ \sc{Modelling}}}} &  Chapter~\ref{chapter:ckconv} & {\color{Green}\cmark }  & - & {\color{Green}\cmark }  & {\color{Green}\cmark }  \\ 
          & Chapter~\ref{chapter:ccnn} & {\color{Green}\cmark }  & - & {\color{Green}\cmark }  & {\color{Green}\cmark }  \\ 
          & Chapter~\ref{chapter:gridification} & {\color{Green}\cmark }  & - & {\color{Green}\cmark }  & - \\ 
          & Chapter~\ref{chapter:flexconv} & {\color{Green}\cmark } & - & - & {\color{Green}\cmark }  \\ 
          &  Chapter~\ref{chapter:dnarch} & {\color{Green}\cmark }  & - & {\color{Green}\cmark }  & {\color{Green}\cmark }  \\ 
         \midrule
         \multirow{5}{*}{\rotatebox{90}{\thead{\sc{Part II:} \\ \sc{Symmetry} \\ \sc{Preservation}}}} & Chapter~\ref{chapter:coattentive} & - & {\color{Green}\cmark }  & - & - \\ 
         & Chapter~\ref{chapter:attgconv}  & - & {\color{Green}\cmark }  & - & - \\ 
          & Chapter~\ref{chapter:g_selfatt}  & - & {\color{Green}\cmark }  & {\color{Green}\cmark }  & -  \\ 
           & Chapter~\ref{chapter:waveletnets} & - & {\color{Green}\cmark }  & {\color{Green}\cmark } & - \\ 
            & Chapter~\ref{chapter:partial_equiv} & {\color{Green}\cmark }  & {\color{Green}\cmark }  & {\color{Green}\cmark }  & {\color{Green}\cmark }  \\ 
        \bottomrule
    \end{tabular}}
\end{center}

%For com compromise in other efficiency aspects such as c able to contribute significantly to various dimensions of efficiency  to various dimensions of efficiency and simultaneously are able to improve the state-of-the-art in many cases --albeit occasionally inducing compromises in other efficiency aspects.

\begin{description}
    \item[Research Question 1.] \textbf{\textit{Can continuous modeling improve Deep Learning Efficiency?\\ {\color{White}spacespacespace} If so, which specific efficiency aspects does it improve?}}
\end{description}
Our study indicates that continuous modeling is indeed a powerful inductive bias capable of bolstering multiple aspects of Deep Learning efficiency.

In Chapter~\ref{chapter:ckconv}, we showed how continuous modeling facilitates the creation of compact global convolutional kernels able to capture global long-term dependencies at a lower time complexity --and thus higher parameter and computational efficiency-- than existing global models like Transformers ($\mathrm{O(N \ logN)}$ vs. $\mathrm{O(N^2)}$). Due to the generality of their formulation, we show in Chapter~\ref{chapter:ccnn} that the resulting models can be straightforwardly used to process multi-dimensional data as well as inputs of different length and resolution --therefore aiding design efficiency--. Existing models that share these properties are based on self-attention, e.g., Perceivers \citep{jaegle2021perceiver}, and thus our proposed general-purpose convolutional architectures offer a more computationally efficient alternative. Furthermore, the use of a continuous parameterization in neural operations enables the treatment of irregularly-sampled data (Chapters~\ref{chapter:ckconv}-\ref{chapter:gridification}) and the development of resolution-agnostic architectures able to generalize to unseen resolutions (Chapter~\ref{chapter:flexconv}). These properties lead to improved design and computational efficiency improvements.

In Chapter~\ref{chapter:gridification}, we revealed that continuous modeling enables the creation of point-cloud processing pipelines that substantially improve the efficiency and scalability of native point-cloud methods such as Message Passing Networks \citep{gilmer2017neural, satorras2021n}. Leveraging the fact that point-cloud are in fact sparse representations of continuous functions, we can rely on continuous modeling to normalize the irregular nature of point-clouds by mapping them onto a regular grid. The result is a substantial gain in computational efficiency and scalability, both in terms of the size of the point-cloud and the size of the neighborhoods considered in their processing.% comparison to native point-cloud methods, this results in This results in a much better scaling both in terms of point-cloud sizes and the neighbourhoods considered in these operations.

Finally, we showed that continuous modeling facilitates the automatic tuning of structural network parameters typically treated as hyperparameters: the size of convolutional kernels (Chapter~\ref{chapter:flexconv}) as well as many more structural components like depth, width and downsampling layers (Chapter~\ref{chapter:dnarch}). By conceptualizing neural architectures as continuous entities, we are able to explore a vast multitude of possible network configurations with little computational overhead. This approach sidesteps the inherent limitations of existing Differentiable Neural Architecture methods, e.g., DARTS \citep{liu2018darts}, where the search space must be small, predefined and expensive to explore, leading to substantial improvements in terms of design and compute efficiency. 

Collectively, these developments depict the transformative potential of continuous modeling in crafting more efficient deep learning algorithms, and our contributions to it.

\textbf{Note about generalization and data-efficiency.} As a final note, we emphasize that our proposed methods oftentimes surpassed the state-of-the-art at the time of publication. We highlight that there exists a correlation between data efficiency and generalization. However, as this is rather a nuanced relationship, we refrain from categorizing the methods in this part as data efficient. 

\begin{description}
    \item[Research Question 2.] \textbf{\textit{Can symmetry preservation improve Deep Learning Efficiency? \\ {\color{White}spacespacespace} If so, which specific efficiency aspects does it improve?}}
\end{description}
Our study indicates that symmetry preservation is indeed a powerful inductive bias capable of bolstering multiple aspects of Deep Learning efficiency, specially in terms of data efficiency. Nevertheless, we find that these improvements impose computationally efficiency compromises.\footnote{It is worth noting that the increase in computational cost is tied to on interpretation. Specifically, one can understand the additional group dimension as additional channels --as in \citet{cohen2016group}. Under this interpretation, the computational cost of group convolutions is lower than that of vanilla convolutions. However, a different interpretation refers to the number of channels as the number of features that are independently learned --analogously to the vanilla convolution case. Under this interpretation, a group convolution is quadratically more expensive than a vanilla convolution (Sec.~\ref{sec:12_groupconv}).}

In Chapters~\ref{chapter:coattentive} and~\ref{chapter:attgconv}, we proposed the use of neural components that explicitly encourage coherent symmetrical combinations along the stabilizer group and the whole group, respectively. These components result in improved data efficiency and generalization of the underlying symmetry preserving models. In addition, we find that the group equivariant attention maps generated by these neural components also serve other purposes. They can be used to explain decisions taken by the model, and to validate the prediction reliability of the model across symmetrical modifications of the input.% insights into the model's decision making.

In Chapter~\ref{chapter:g_selfatt}, we broadened the applicability of symmetry preservation to the popular Transformer architecture. The resulting group equivariant Transformers show a noticeable increase in data and parameter efficiency over vanilla Transformer architectures, comparable to the gains seen in convolutional models. The versatility and generality of group equivariant self-attention has proven useful in diverse applications, notably in the processing and generation of $3$D molecular structures \citep{hutchinson2021lietransformer, hoogeboom2022equivariant}.

In Chapter~\ref{chapter:waveletnets}, we investigated the role of symmetry preservation in the context of time-series processing. We identify two inherent symmetries of time-series: translation and scale, and build neural architectures that respect them. Our findings confirm that the benefits of symmetry preservation carry over to the time-series domain, leading to improved data and parameter efficiency. In addition, we delve into the relationship between scale-translation equivariance and the Wavelet transform, shedding light on the modus-operandi of these networks, and on ways to improve their parameterization.% adjustments on the parameterizations of convolutional kernel, which lead to improved results. 

Finally, we address a key limitation of symmetry-preserving models in Chapter~\ref{chapter:partial_equiv}: the risk of symmetry misspecification resulting from the need to specify data symmetries prior to training. Typically, symmetry-preserving models require a predefined set of symmetries to adhere to. However, this can lead to overly restrictive models if these symmetries are misspecified or only approximately present in practice. To overcome this issue, we introduce a data-driven approach to dynamically adjust the symmetry specifications of a model based on the  symmetries observed in data. This formulation leads to models with enhanced generalization, data efficiency and design efficiency than conventional symmetry-preserving models. Furthermore, our formulation makes it possible to adapt the number of group samples used during inference based on the level of symmetries observed in data, leading to improved computational efficiency in settings with partial symmetries.%  the computational efficiency of these models can be optimized by adapting the number of group samples used during a group convolution, according to the level of symmetries observed in data.

Collectively, these developments depict the transformative potential of symmetry preservation in crafting more efficient deep learning algorithms, and our contributions to it.%various applications and benefits elucidate the potential of symmetry preservation for the development of more data efficiency deep learning algorithms. However, it is worth mentioning that all these models incur in an increased computational cost than non-symmetry preserving models.\footnote{Explain that this is a bit open to interpretation.}

\vspace{-7mm}
\section{Limitations and additional advances}\label{sec:12_limitations}
Despite the multiple advances made by this research, there are several important limitations that remain. In this section, we outline these limitations and discuss potential solutions coming both from recent advancements and avenues for future research.

\vspace{1mm}
\textbf{\large\sc{Continuous Modeling}} 
\vspace{-7mm}
\subsection{Computational implications of global convolutions} 
Despite the advancements in Chapters~\ref{chapter:ckconv}-\ref{chapter:dnarch}, there is one important overarching limitation that remains: the computational cost of using (global) long convolutional kernels. While the computation efficiency these methods excel in comparison to other global models like Transformers, they have a steeper cost relative to discrete CNNs that rely on local convolutions, which --despite an emerging change of paradigm towards global operations, e.g., self-attention-- have traditionally been the prime processing pipeline for multi-dimensional data like images and point-clouds.

It is worth noting that for $1$D problems, this computational limitation is not pronounced, as long convolutions can be efficiently computed using the Fast Fourier Transform (FFT) in $\mathrm{O(N\ logN)}$ time. Although the complexity of FFT convolutions on multiple dimensions should mirror that of $1$D in theory, e.g., $\mathrm{O(N^2\ logN^2)}{=}\mathrm{O(N^2\ logN)}$ for images of size $\mathrm{N}{\times}\mathrm{N}$, we find that existing FFT convolution implementations in \texttt{PyTorch} and \texttt{JAX} lag much behind these expectations in practice. As such, computing convolutions on the spatial domain --which have $\mathrm{O(N^4)}$ complexity for an image of size $\mathrm{N}{\times}\mathrm{N}$ and a global convolutional kernel-- is typically \textit{faster}. Given the growing interest in long convolutions, developing more efficient \texttt{CUDA} implementation of multidimensional FFT convolutions in Deep Learning frameworks would be very beneficial and impactful. Such implementations will likely result in significant speed-ups to existing methods.

From an algorithmic standpoint, FlexConv (Sec.~\ref{chapter:flexconv}) offers a partial remedy to this limitation by learning the size of convolutional kernels at each layer. However, it is likely --and desirable-- for some kernels to remain global. An additional mitigation strategy comes from DNArch (Sec.~\ref{chapter:dnarch}) by learning the resolution on which long convolutions are computed based on data. However, even at lower resolutions, global convolutions can still be costly. Other potential structural alternatives may involve employing dilated convolutions with learnable spacings, as illustrated by \citet{hassani2023dilated}. 

Other recent studies offer additional strategies to pare down the computational overhead of (global) long convolutions in $N$D by means of decomposition and sparsification. \citet{nguyen2022s4nd} decomposes $N$D convolutional kernels into $N$ $1$D convolutional kernels along each dimension, leading to important effficiency improvements. Recently, \citet{kirchmeyer2023convolutional} proposed the use of oriented $1$D kernels for the processing of complex $2$D problems, showing that oriented $1$D kernels are able to match architectures that rely on $2$D kernels at a much higher speed. % across several complex problems at .%learning the orientation of such $1$-dimensional kernels can be used to
Lastly, \citet{liu2023more} proposes a low-rank decomposition of convolutional kernels into kernels of smaller sizes whose response can be computed in parallel, e.g., $5{\times}5$, $5{\times}51$ and $51{\times}5$ for a kernel of size $51{\times}51$.% All these techniques could be used to reduce the computational complexity of long convolutions.
\begin{figure}
    \centering
    \includegraphics[width=0.8\textwidth]{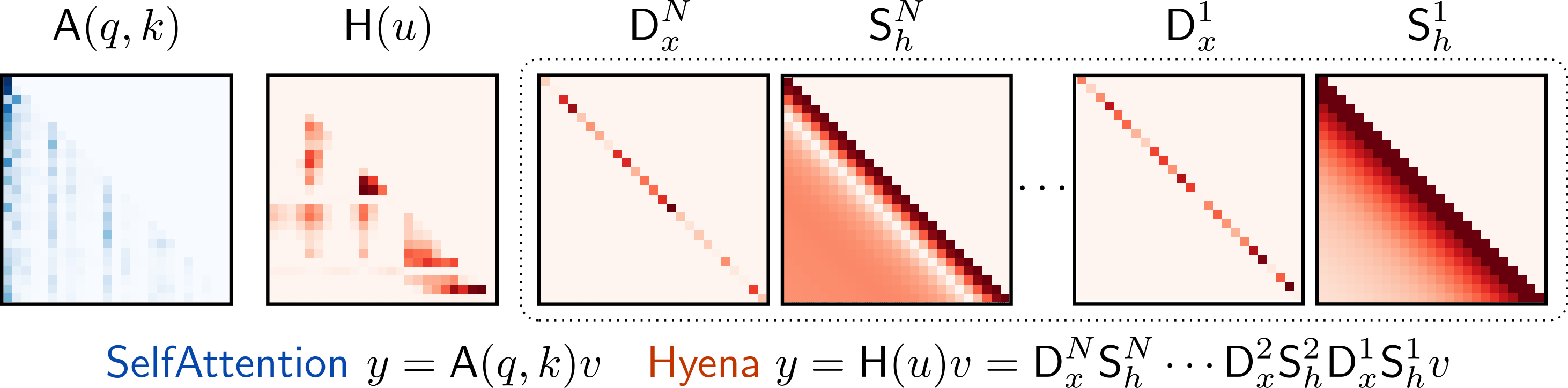}
    \vspace{-1mm}
    \caption{Global, input-dependent mappings through the use of interleaved long convolutions and input-dependent gating. Taken from \citet{poli2023hyena}.}
    \label{fig:0_attention}
\end{figure}

\vspace{-7mm}
\subsection{Input-dependence in long convolutional models} 
Although long convolutions are faster than self-attention, they lack a critical capability inherent to self-attention. Self-attention is not only able to capture global context, but it simultaneously is able to generate input-specific mappings --a feature absent in the long convolutional models discussed in this dissertation.

Interestingly, a recent study by \citet{poli2023hyena} demonstrated that long convolutional models, e.g., those based on CKConvs (Ch.~\ref{chapter:ckconv}), can be extended to produce input-dependent mappings. This is achieved through a combination of $\mathrm{D}{=}2$ interleaved long convolutions and input-depending gating operations. The resulting models are both global and input-dependent with a computational complexity of $\mathrm{O(D N\ logN)}$ (Fig.~\ref{fig:0_attention}). \citet{poli2023hyena} revealed that CKConv-based Hyenas match Transformers in intricate language processing tasks, marking them as the first attention-free model able to do so. Moreover, capitalizing on the computational benefits of long convolutions, \citet{nguyen2023hyenadna} showed that Hyenas can process sequences with 1 million tokens at orders of magnitude faster speed than Transformers, while still considering global context and input-dependence.
\begin{figure}
    \centering
    \includegraphics[width=0.8\textwidth]{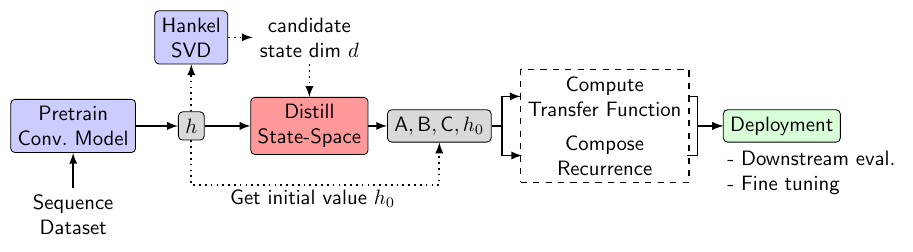}
    \vspace{-2mm}
    \caption{Distillation strategy to transform trained long convolutional networks into recurrent neural networks. Taken from \citet{massaroli2023laughing}.}
    \label{fig:0_distillation}
\end{figure}

\vspace{-7mm}
\subsection{Autoregressive inference with long convolutional models}\label{sec:12_autoregressive}
When using convolutions, it is necessary to retain the entire history of past activations in memory to predict the next value in the sequence. This inherent characteristic of convolutional models presents a significant challenge for autoregressive tasks like text generation, as the history grows linearly during the generation process. While this property is not an issue during training --owing to the vectorization across full training sequences-- it becomes a notable limitation during inference.

Addressing this challenge, our recent research presented in \citet{massaroli2023laughing} provides a solution. Specifically, we introduce a method with which a trained long convolutional model can be distilled into a Recurrent Neural Network without loss in accuracy by converging the learned model into a State-Space Model representation \citep{chen1984linear} (Fig.~\ref{fig:0_distillation}). Since State-Space Models can be deployed both in convolutional and recurrent form, it is possible to conduct autoregressive inference post-training with constant memory complexity, without compromising on prediction accuracy.

%In combination with the developments of the previous section, we consider the further development and exploration of these techniques a very inspiring and promising direction for further research. We are particularly
\vspace{2mm}
\textbf{\large\sc{Symmetry Preservation}} 
\vspace{-7mm}
\subsection{Computational implications of group convolutions and large groups}\label{sec:12_groupconv}
From a computational viewpoint, the main limitation of group convolutions stems from the additional complexity resulting from the use of additional axes to encode the symmetries of the group considered\footnote{This is the case when using regular group representations, as is done throughout this dissertation.}. Specifically, relating back to the seminal G-CNNs of \citet{cohen2016group}, given an image of size $\mathrm{N}{\times}\mathrm{N}$, a finite group with $\mathrm{G}$ elements, and a convolutional kernel of spatial dimensions $\mathrm{K}{\times}\mathrm{K}$, a group convolution has $\mathrm{O(N^2K^2G^2)}$ time complexity, which is $\mathrm{G^2}$ times higher than the complexity of the corresponding standard spatial convolution $\mathrm{O(N^2K^2)}$. This increased demand arises from the fact that both the input and the convolutional kernel are defined on the group, which introducing an extra dimension with $\mathrm{G}$ coordinates in this example.
\begin{figure}
    \centering
    \includegraphics[width=0.8\textwidth]{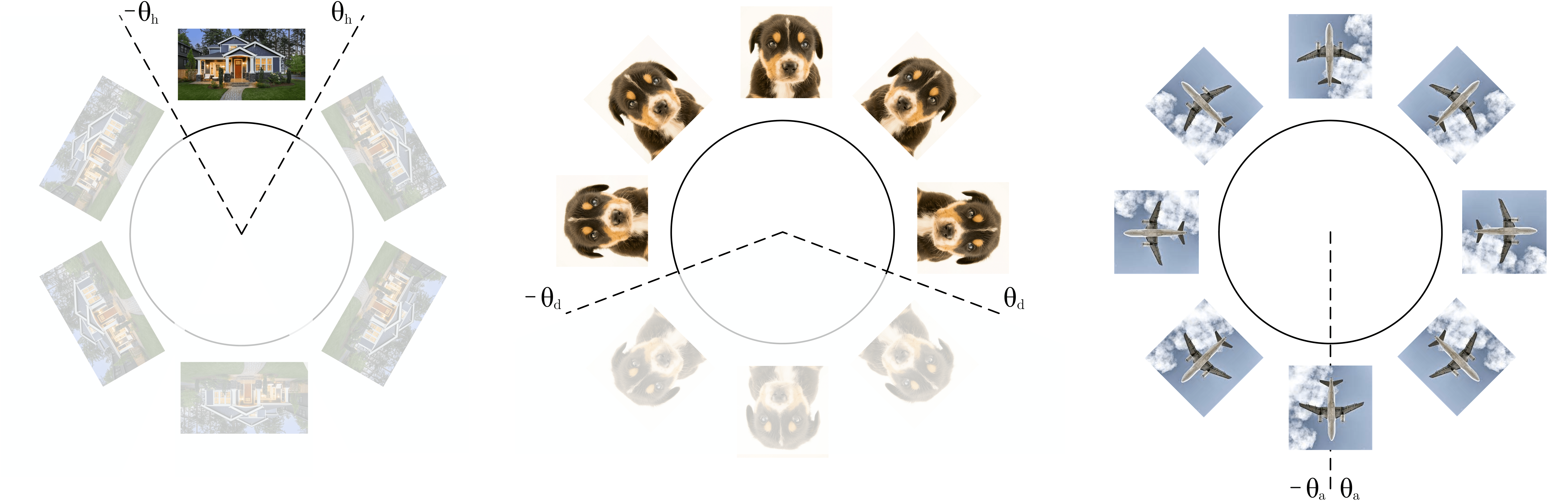}
    \vspace{-2mm}
    \caption{In real world scenarios, samples with different semantic meaning may have different symmetrical properties. Considering this input-dependency in the design of symmetry-preserving neural architectures facilitates the construction of more flexible neural architectures that accurately represent the symmetrical properties of data. Taken from \citet{urbano2023self}.}
    \label{fig:0_self-supervised}
\end{figure}

A promising avenue to reduce the complexity of these operations comes from the use of separable group convolutions \citep{knigge2022exploiting}. In \citet{knigge2022exploiting}, we show that decomposing group convolutional kernels into spatial and group components substantially reduces the computational demands of group convolutional networks without sacrificing accuracy. This reduction, additionally allows for the creation of neural architectures that are equivariant to very large groups, such as the $\mathrm{Sim}(2)$ group, consisting of $2$D translations, scaling and rotation. In addition, we utilize a similar strategy in \citet{bekkers2023fast} and rely on separability on the $3$D space of positions and orientations $\sR^3{\times}S^2$, to device efficient neural architectures for the equivariant processing and generation of molecules.

Another proposing approach leverages Monte-Carlo approximations of group convolutions \citep{finzi2020generalizing}. It entails approximating group convolutions by sampling different group elements during each forward pass of the network. This technique provides exact equivariance on expectation, and allows the construction of efficient equivariant neural architectures able to consider (infinite) continuous groups with regular group representations. Evidence suggests that using continuous parameterizations akin to CKConv (Ch.~\ref{chapter:ckconv}) for group convolutional kernels leads to enhanced expressivity \citep{knigge2022exploiting, romero2022learning, zhdanov2022implicit}.% equivariant toover the  

It is important to mention that regular group representations are not the only possible alternative. In fact, several works have explored the use of irreducible group representations, which allow for the construction of neural architectures equivariant to continuous groups \citep{cohen2016steerable, worrall2017harmonic, weiler2019general, cesa2021program, zhdanov2022implicit}. However, neural architectures based on irreducible group representations require a yet more intricate design than regular group representation. This imposes additional strong constraints over the kind of operations that can be used, e.g., the type of admissible nonlinearities, and the knowledge required for their construction. Furthermore, several studies indicate that regular representations lead to consistently better results than irreducible representations \citep{weiler2019general, cesa2021program, kuipers2023regular}, specially if continuous groups are considered \citep{finzi2020generalizing}. Based on these observations, we contend that pursuing efficiency improvements for neural architectures built on regular group representations may hold the highest potential for future progress.

\vspace{-7mm}
\subsection{Symmetry pre-specification in group CNNs} 
Typically, during the instantiation of group equivariant models, one must predefine the group of symmetries to which the model most be equivariant. This definition denotes a limiting factor in terms of design efficiency and has the risk of producing overly constrained models the symmetry group is not properly specified \citep{chen2020group, romero2022learning}.

A recent line of work aims to release researchers from the need to define these symmetries prior to training, by inferring them directly from data during training. Existing approaches propose this by means of meta-learning \citep{zhou2020meta}, generative modeling \citep{yang2023generative}, neural architecture search \citep{maile2023equivarianceaware}, probabilistic reasoning \citep{immer2022invariance, vanderouderaa2023learning} and the learning of infinitesimal group generators \citep{dehmamy2021lie}. This line of work has the potential to be particularly impactful, as it may help uncover the symmetries that complex systems should respect, and release researchers from the need to specify them by hand.

An additional scenario in which the group of symmetries may be misspecified is in cases in which the symmetries encoded in the network are only partially observed in data (Ch.~\ref{chapter:partial_equiv}). To address this challenge, the concept of partial --or soft-- equivariance has been introduced. Partially equivariant models aim to relax the symmetric constrains of a model during training based on the data observed. Existing approaches employ probabilistic methods to learn group subsets based on data \citep{romero2022learning, van2022relaxing} or combine equivariant and non-equivariant components in a learnable fashion \citep{finzi2021residual, wang2022approximately}.

Nevertheless, an important challenge persists in both cases: both the notions of equivariant and partial equivariance are inferred at a dataset level instead of at a sample level. This generalized approach is prone to overlooking nuanced symmetrical properties exhibited by individual samples within a single dataset (Fig.~\ref{fig:0_self-supervised}). As a result, such models can only predict the average level of symmetry seen in the dataset. In \citet{urbano2023self}, we recently proposed a methodology that enables the extraction of symmetric characteristics --both exact and partial-- on a per-sample basis. We achieve this behaviour through the learning of input dependent symmetry distributions that can be efficiently inferred from semantically similar objects in the dataset. We are optimistic about the potential of this approach, as it may empower the development of neural architectures able to consider a more flexible definition of symmetry preservation.  

\vspace{-7mm}
\section{Future Work}

\vspace{-5mm}
\subsection{Exploring other inductive biases for efficiency and beyond}
While this dissertation focuses on the role of continuous modeling and symmetry preservation, these are by no means the only inductive biases that serve this purpose. There exist several other inductive biases that also hold substantial promise for improving the efficiency of deep learning algorithms. For example, sparsity and causality.

Sparsity pertains to the selective activation and utilization of neurons within a neural network, offering significant advantages for reducing the computational costs of neural networks during training and deployment \citep{lecun1989optimal}. It aims at constructing compressed representations where only the most critical features and connections are preserved, thereby reducing the model's parameter, memory and computational complexity. Traditionally, sparsity has been primarily used post-training via \textit{pruning}, to distil a trained network into a smaller equally effective sub-network \citep{han2015learning}. Recent studies have shown that sparsity can also be effectively introduced early in training, as illustrated in the PhD theses by \citet{frankle2023lottery} and \citet{lee2020toward}. Sparsity holds great potential for the development of more compute- and parameter-efficient neural architectures.

Causality, on the other hand, pertains the structuring of models and algorithms around the concept of cause-and effect relationships between variables, thereby holding the potential to improve the data, compute and parameter efficiency of deep learning models \citep{pearl2009causality, peters2017elements}. Unlike traditional correlation-based approaches, which merely identify correlating patterns in the data, causal approaches strive to capture and understand the underlying data-generating mechanisms. As a result, causality-informed models typically exhibit simpler structures that require less training data, and are less prone to overfitting \citep{scholkopf2021toward, shen2022weakly}. Beyond efficiency gains, causality may also improve the explainability of the resulting models by interpreting the data-generating process captured \citep{ribeiro2016should}. Causality holds considerable promise not only for the development of more efficient neural architectures, but also for increasing the reliability and applicability of neural networks.% We therefore consider causality to be a particularly impactful area for future research.

We consider investigating the use of inductive biases for the development of Deep Learning architectures as a promising avenue for further research. Notably, inductive biases not only have implications for efficiency, but also hold value for the development of more robust, reliable, generalizable and explainable Deep Learning methods. 

\vspace{-7mm}
\subsection{Future work on continuous modeling and symmetry preservation}
Despite the progress in continuous modeling and symmetry preservation made in this dissertation, numerous open questions and avenues for future research remain. We conclude this dissertation highlighting research directions we consider most promising as a complement to addressing the limitations presented in Sec.~\ref{sec:12_limitations}.% For a detailed discussion on future work directions for each of the contributions made in this dissertation, we refer the reader to the Future Work sections of each chapter.

\vspace{-7mm}
\subsubsection{Better parameterizations of continuous spatial functions}
In this work, we have investigated several parameterization of spatial continuous functions such as continuous convolutional kernels. Specifically, we proposed parameterization based on multiple Neural Field families including SIRENs \citep{sitzmann2020implicit} (Ch.~\ref{chapter:ckconv}), RFNets \citep{tancik2020fourier} (Ch.~\ref{chapter:ccnn},~\ref{chapter:gridification}) and MFNs \citep{fathony2021multiplicative} (Ch.~\ref{chapter:flexconv}). However, numerous new families have emerged since the inception of this research, which may improve the parameterizations introduced herein, e.g., \citet{muller2022instant, yang2023tinc, saragadam2023wire, wu2023neural}. 

However, it is important to highlight that the way in which Neural Fields are used in this dissertation differs dramatically from the standard problem considered in Neural Field's literature. Specifically, we utilize Neural Fields to parameterize neural operations for which the ground truth \textit{is unknown and is constantly changing}. This is in contrast to conventional Neural Field's literature, where Neural Fields map  to \textit{known and fixed} Consequently, insights from Neural Field literature do not necessarily extrapolate to the scenarios considered in this dissertation, and Neural Fields that perform well in Neural Field literature will not necessarily work well in this setting. 

For instance, given that the \enquote{ground truth} of continuous neural operations are constantly changing during training, the speed at which a Neural Field comes to a rough approximation of this ground truth could be more valuable than its ability to fit this ground truth to perfection after several iterations. Additionally, certain properties deemed peripheral in Neural Fields literature may hold important advantages for the applications here considered. For example, the ability of MFNs \citep{fathony2021multiplicative} to provide analytic control over the frequency components of the approximation is of outmost importance for tasks like zero-shot predictions at higher resolutions (Ch.~\ref{chapter:flexconv}). Therefore, even though the fitting capabilities of MFNs have been surpassed by newer methods like InstantNGP \citep{muller2022instant}, this unique feature makes MFNs an incredibly compelling candidate for future exploration.

In summary, we deem a systematic evaluation of recent Neural Field parameterizations in the context of this dissertation to be a critical avenue for future research. Given the increasing interest in long convolutions parameterized by Neural Fields \citep{vasconcelos2023cuf, kim2023smpconv, poli2023hyena, nguyen2023hyenadna, oh2023schyena, zimerman2023multi}, such an investigation has the potential to have significant impact across multiple applications and disciplines.

\vspace{-7mm}
\subsubsection{From self-attention to (input-dependent) long convolutional models}
While self-attention excels in handling global dependencies, its computational overhead is a significant, increasingly prohibitive drawback. Recent research indicates that input-dependent long convolutional models can achieve comparable or even superior performance than Transformers, but with greater computational efficiency and scalability \citep{poli2023hyena, nguyen2023hyenadna, zimerman2023multi}. Further exploring the application of CKConvs \citep{romero2022ckconv} and Hyenas \citep{poli2023hyena} across applications where Transformers are currently prevalent offers the prospect of improving both efficiency and versatility. In addition, the increased computational efficiency of long convolutional models could open new horizons in computationally-intensive domains such as those involving truly long contexts. For example, for the construction of foundational audio, time-series and video models.\footnote{In contrast to text, where a 32.000-token context encompasses a very long piece of text, a single second of audio easily contains 48.000 samples. This makes evident the scale of context foundational models on this kind of data should be able to handle.}$^,$\footnote{It is worth noting that long convolutional models can also be distilled into RNNs during inference to achieve constant memory costs in autoregressive generation (see Sec.~\ref{sec:12_autoregressive}, \citet{massaroli2023laughing}). This is well below the linear cost of (cached) causal Transformers.}

Unlike self-attention, convolutional operations are translation preserving. Since translation is a symmetry that is ubiquitously present in natural data, the use of translation preserving operations for the processing of natural data, e.g., time-series, audio, video, may lead --as observed in this dissertation-- to a significant reduction in the model size and data requirements for training. In other words, this property may hold large potential for the scalability, generalizability and applicability of long convolutional models.

Lastly, the only method currently known for the implementation of input-dependency in long convolutional models is the one presented in \citet{poli2023hyena}, which involves an interleaved combination of input-dependent gating and long convolutions. Investigating alternative techniques could offer additional avenues for improvement. Since long convolutional models often rely on Neural Fields to parameterize their kernels, we consider the use of Neural Field conditioning techniques, e.g., \citet{perez2018film, park2021hypernerf, jiang2021cotr}, an interesting direction for future research.%. Since these long convolution models rely on Neural Fields for the parameterization of the convolutional kernels, exploring input-dependency through the use of conditioning may lead to interesting results.

\vspace{-7mm}
\subsubsection{Extending group convolutions to very large groups}
As discussed in Sec.~\ref{sec:12_groupconv}, one of the main limitations of group convolutions lies in their computational cost, which scales quadratically with the size of the group considered. This is particularly challenging for large complex group such as the projective $2$D group: $\mathrm{Proj(2)}$, and several other groups that act in $3$D, e.g., $\mathrm{Sim(3)}$: the group consisting of $3$D rotations, $3$D translations and homogeneous scaling. Nevertheless, considering equivariance to these larger groups offer rich structural prior information that could be invaluable for emerging applications, such as point-cloud processing and mesh generation, particularly for applications on which train data is scarse. Devising techniques that make the handling of such large groups manageable, e.g., through group decompositions \citep{knigge2022exploiting, bekkers2023fast} and Monte-Carlo approximations \citep{finzi2020generalizing, romero2022learning}, holds great promise for the development of more data-efficient Deep Learning methods.

%% file: backmatter/1_appx_ckconv.tex
% path to figures directory
\graphicspath{{figures/2-ckconv/}}

%=========================================================================

% \begin{savequote}[75mm]
% Nulla facilisi. In vel sem. Morbi id urna in diam dignissim feugiat. Proin molestie tortor eu velit. Aliquam erat volutpat. Nullam    ultrices, diam tempus vulputate egestas, eros pede varius leo.
% \qauthor{Quoteauthor Lastname}
% \end{savequote}

\chapter{Continuous Kernel Convolutions for Sequential Data}
		\label{appx:2_ckconv}

%=========================================================================
\vspace{-7mm}
\section{Properties of CKConvs}\label{appx:2_properties}
\vspace{-4mm}
\subsection{Very Irregularly Sampled Data}\label{appx:2_irregularly_sampled}
CKConvs can readily handle irregularly-sampled, and partially observed data. This is a result of the convolutional kernel \mlp$^{\boldsymbol{\psi}}$ being able to be sampled at arbitrary positions. For very non-uniformed sampled inputs, however, the corresponding sampling of the convolutional kernel can provide a biased estimation of the operation. To overcome this, one can follow the strategy proposed by \citet{wu2019pointconv}, which we summarize here for completeness.

For very non-uniformly sampled inputs, the continuous convolution $(x * \psi)(t) = \break \int_{\sR} x(\tau) \psi(t - \tau) \, \du \tau$, must account for the distribution of samples in the input. Specifically, it is reformulated as:
\begin{equation}
     (x * \psi)(t) = \int_{\sR} s(\tau) x(\tau) \psi(t - \tau) \, \du \tau, \label{eq:2_irreg_sampled}
\end{equation}
where $s(\tau)$ depicts the inverse sample density of the input at point $\tau$. Intuitively, $s(\tau)$ controls the contribution of points $x(\tau)$ to the output response. If multiple points are close to one another, their contribution should be smaller than the contribution of points in regions where the sample distribution is much sparser. This provides a Monte Carlo estimate of $(x * \psi)$ from biased samples. In particular, one has that:
\begin{equation*}
    \int \hspace{-1mm} f(\tau) \, \du \tau= \int \hspace{-1mm} \frac{f(\tau)}{p(\tau)} p(\tau) \, \du \tau \approx \sum_{i}\frac{f(\tau_i)}{p(\tau_i)}, \ \text{for}\  \tau_{i} \sim p(\tau). 
\end{equation*}
With  $s(\tau) = \tfrac{1}{p(\tau)}$, Eq.~\ref{eq:2_irreg_sampled} provides an unbiased estimation of $(x * \psi)$.
\vspace{-7mm}
\subsection{Data Sampled at Different Sampling Rates}\label{appx:2_diff_sampling_rates}

In addition, CKConvs are readily able to handle data at different resolutions. In particular, the continuous kernel convolution between an input signal $\xv$ and a continuous convolutional kernel $\boldsymbol{\psi}$ calculated at sampling rates $\mathrm{sr}_1$: $(\xv *\boldsymbol{\psi})_\mathrm{sr_1}$, and $\mathrm{sr_2}$: $(\xv *\boldsymbol{\psi})_\mathrm{sr_2}$, are approximately equal up to a normalization factor given by $\tfrac{\mathrm{sr_2}}{\mathrm{sr_1}}$:
\begin{equation*}
    (\xv *\boldsymbol{\psi})_\mathrm{sr_2}(t) \approx \frac{\mathrm{sr_2}}{\mathrm{sr_1}}(\xv *\boldsymbol{\psi})_\mathrm{sr_1}(t).
\end{equation*}
Consequently, CKCNNs \emph{(i)} can be deployed at sampling rates different than those seen during training, and \emph{(ii)} can be trained on data with varying spatial resolutions. The later is important for tasks in which data can be given at different resolutions such as super-resolution and segmentation. 

\textit{Proof.} To prove the previous statement, we start with the continuous definition of the convolution:
\begin{equation*}
    (x * \psi)(t) = \int_{\sR} x(\tau) \psi(t - \tau) \, \du \tau,
\end{equation*}
where we assume for simplicity and without loss of generality that the functions $x$, $\psi$ are scalar-valued. 

In practice, an integral on a continuous function $f: \sR \rightarrow \sR$ cannot be computed on finite time. Consequently, it is approximated via a Riemann integral defined on a finite grid $\{ \tau_{\mathrm{sr}, i} \}_{i = 1}^{\Nt_\mathrm{sr}}$ obtained by sampling $\tau$ at a sampling rate $\mathrm{sr}$:
\begin{equation*}
    \int f(\tau) \, \du \tau \approx \sum_{i = 1}^{\Nt_\mathrm{sr}} f(\tau_{\mathrm{sr}, i}) \Delta_{\mathrm{sr}},
\end{equation*}
where $\Delta_{\mathrm{sr}} = \tfrac{1}{\mathrm{sr}}$ depicts the distance between sampled points. For two sampling rates $\mathrm{sr}_1$, $\mathrm{sr}_2$, the convolution can be approximated through the corresponding Riemann integrals:
\begin{align}
   \int_{\sR} x(\tau) \psi\big(t - \tau) \,  \du \tau  & \approx \sum_{i = 1}^{\Nt_{\mathrm{sr}_{1}}} x(\tau_{\mathrm{sr}_{1}, i}) \psi\big(t - \tau_{\mathrm{sr}_{1}, i}) \Delta_{\mathrm{sr}_{1}}  \nonumber \\[-0.5\jot]
    & \approx \sum_{i = 1}^{\Nt_{\mathrm{sr}_{2}}} x(\tau_{\mathrm{sr}_{2}, i}) \psi\big(t - \tau_{\mathrm{sr}_{2}, i}) \Delta_{\mathrm{sr}_{2}}  \nonumber 
\end{align}
As a result, we have that both approximations are approximately equal to the continuous integral at positions $t$ defined on both discrete grids. By equating both approximations, we obtain that: 
\begin{align}
    \sum_{i = 1}^{\Nt_{\mathrm{sr_2}}} \hspace{-1mm} x(\tau_{\mathrm{sr_2}, i}) \psi\big(t - \tau_{\mathrm{sr_2}, i}) \Delta_{\mathrm{sr_2}} & \approx  \sum_{i = 1}^{\Nt_{\mathrm{sr_1}}} \hspace{-1mm} x(\tau_{\mathrm{sr_1}, i}) \psi\big(t - \tau_{\mathrm{sr_1}, i})  \Delta_{\mathrm{sr_1}} \nonumber \\[-0.5\jot]
    \underbrace{\sum_{i = 1}^{\Nt_{\mathrm{sr_2}}} \hspace{-1mm} x(\tau_{\mathrm{sr_2}, i}) \psi\big(t - \tau_{\mathrm{sr_2}, i})}_{(x * \psi)_{\mathrm{sr_2}}(t)} \tfrac{1}{\mathrm{sr_2}} & \approx \underbrace{\sum_{i = 1}^{\Nt_{\mathrm{sr}_{1}}} \hspace{-1mm} x(\tau_{\mathrm{sr_1}, i}) \psi\big(t - \tau_{\mathrm{sr_1}, i})}_{(x * \psi)_{\mathrm{sr_1}}(t)} \tfrac{1}{\mathrm{sr_1}} \nonumber \\[-0.5\jot]
    (x *\psi)_\mathrm{sr_2}(t) &\approx \frac{\mathrm{sr_2}}{\mathrm{sr_1}}(x *\psi)_\mathrm{sr_1}(t)  \nonumber
\end{align}
which concludes the proof. 

\vspace{-7mm}
\subsection{Linear Recurrent Units Are CKConvs}\label{appx:2_linrecunitsasckconvs}
Interesting insights can be obtained by drawing connections between convolutions and recurrent units. In particular, we can show that linear recurrent units are equal to a CKConv with a particular family of convolutional kernels: exponential functions. Besides providing a generalization to recurrent units, this equality provides a fresh and intuitive view to the analysis of vanishing and exploding gradients.

\textbf{Recurrent unit.} Given an input sequence $\gX = \{ \xv(\tau) \}_{\tau = 0}^{\Nt_{\Xt}}$, a recurrent unit is defined as:
\begin{align}
 \hv(\tau) &= \sigma (\Wm \hv(\tau - 1) + \Um \xv(\tau))\label{eq:2_hidden_repr_formulation}\\[0 \jot]
 \tilde{\yv}(\tau) &= \mathrm{softmax}( \Vm \hv (\tau) ),
\end{align}
where $\Um, \Wm, \Vm$ parameterize the \textit{input-to-hidden}, \textit{hidden-to-hidden} and \textit{hidden-to-output} connections of the unit. $\hv(\tau)$,  $\tilde{\yv}(\tau)$ depict the hidden representation and the output at time-step $\tau$, and $\sigma$ represents a point-wise non-linearity. 

The hidden representation $\hv$ of a \textit{linear} recurrent unit, i.e., with $\sigma{=}\mathrm{Id}$, can be written as a convolution. To see this, consider the hidden representation of the unit unrolled for $t$ steps, with $\hv(-1)$ the initial state of the hidden representation:
\begin{equation}
\setlength{\abovedisplayskip}{0pt}
\setlength{\belowdisplayskip}{1pt}
    \hv(t) = \Wm^{t+1} \hv(-1) + \sum_{\tau = 0}^{t} \Wm^{\tau}\Um \xv(t - \tau).\label{eq:2_unroll_t_steps}
\end{equation}
We see that in fact it corresponds to a convolution between an input signal $\xv$ and a convolutional kernel $\boldsymbol{\psi}$ given by:\footnote{We discard $\hv(-1)$ as it only describes the initialization of $\hv$.}
\vspace{0.0mm}
\begin{align}
     &\ \xv = [\xv(0) , \xv(1) , ... , \xv(t - 1), \xv(t) ] \label{eq:2_correspondence1} \\[0 \jot]
      &\ \ \boldsymbol{\psi} =  [ \Um  , \Wm \Um,  ... ,  \Wm^{t - 1} \Um, \Wm^{t} \Um] \label{eq:2_correspondence2}\\[-1.5\jot]
       \hv(t)& =  \sum_{\tau=0}^{t}\xv(\tau) \boldsymbol{\psi}(t-\tau)  = \sum_{\tau=0}^{t}\xv(t - \tau) \boldsymbol{\psi}(\tau) . \label{eq:2_correspondence3}      
\end{align}
Drawing this equality yields some important insights:

\textbf{The cause of the exploding and vanishing gradients.} Eqs.~\ref{eq:2_correspondence1}-\ref{eq:2_correspondence3} intuitively depict the root of the exploding and vanishing gradient problem. It stems from sequence elements $\xv(t - \tau)$ $\tau$ steps back in the past being multiplied with an effective convolutional weight $\boldsymbol{\psi}(\tau) {=} \Wm^{\tau} \Um$. 
For eigenvalues of $\Wm$, $\lambda$, other than one, the resulting convolutional kernel $\boldsymbol{\psi}$ can only represent functions that either grow ($\lambda{\geq}1)$ or decrease ($\lambda{\leq}1)$ exponentially as a function of the sequence length (Figs.~\ref{fig:2_vanishing_rnn},~\ref{fig:2_exploding_rnn}).
As a result, the contribution of input values in the past either rapidly fades away or governs the updates of the model parameters. As exponentially growing gradients lead to divergence, the eigenvalues of $\Wm$ for converging architectures are often smaller than 1. This explains the effective small effective memory horizon of recurrent networks.

\textbf{Linear recurrent units are a subclass of CKConvs.} Linear recurrent units can be described as a convolution between the input and a very specific class of convolutional kernels: exponential functions (Eq.~\ref{eq:2_correspondence2}). In general, however, convolutional kernels are not restricted to this functional class. This can be seen in conventional (discrete) convolutions, whose kernels are able to model complex functions within their memory horizon. Unfortunately, discrete convolutions use a predefined, small kernel size, and thus possess a restricted memory horizon. This is equivalent to imposing an effective magnitude of zero to all input values outside the memory horizon (Fig.~\ref{fig:2_exploding_conv}). CKConvs, on the other hand, are able to define arbitrary large memory horizons. For memory horizons of size equal to the input length, CKConvs are able to model complex functions upon the entire input (Fig.~\ref{fig:2_exploding_ckconv}).

In conclusion, we illustrate that CKConvs are also a generalization of (linear) recurrent architectures which allows for parallel training and enhanced expressivity.

\vspace{-7mm}
\section{An Spline Interpretation Of $\mathrm{ReLU}$ and $\mathrm{Sine}$ Networks}\label{appx:2_spline_interpretation}

\citet{sitzmann2020implicit} motivates the usage of Sine nonlinearities for implicit neural representations. However, there is no clear understanding as of \textbf{\textit{why}} Sine nonlinearities are better suited for this task than (smooth) piece-wise nonlinearities. Here, we provide an interpretation to this phenomenon from a spline function approximation perspective.

\vspace{-7mm}
\subsection{Kernel Parameterization via $\mathrm{ReLU}$ Networks} \label{appx:2_relu}
\textbf{The importance of initialization.} There is an important
distinction between implicit neural representations and conventional neural applications regarding the assumed distribution of the input. Conventional applications assume the distribution of the input features to be centered around the origin. This is orthogonal to implicit neural representations, where the spatial distribution of the output, i.e., the value of the function being implicitly represented, is \textit{uniformly
distributed}. 

\begin{figure}
    \centering
    \includegraphics[width=0.8\textwidth]{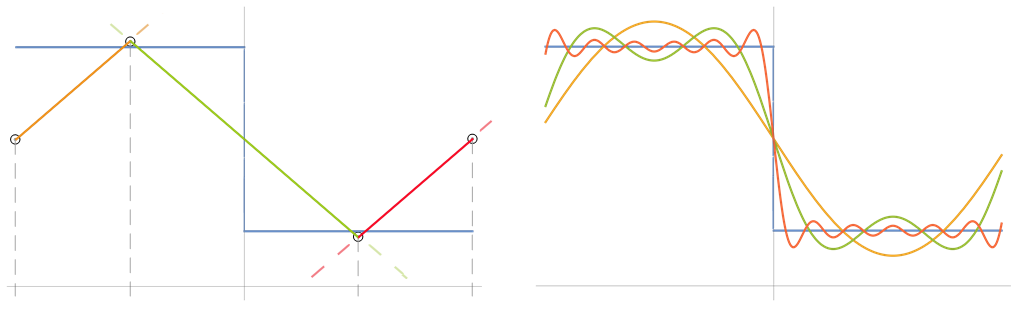}
    \caption{An step function approximated via a spline basis (left) and a periodic basis (right). As the target function is defined uniformly on a given interval, uniformly initializing the knots of the spline basis provides faster and better approximations. Periodic bases, on the other hand, periodically bend space, and thus can be tuned easier to approximate the target function at arbitrary points in space.}
    \label{fig:2_splineapprox}
\end{figure}
For $\mathrm{ReLU}$ networks, function approximation is equivalent to an approximation via a max-spline basis \citep{balestriero2018mad}, and its expressiveness is determined by the number of \emph{knots} the basis provides, i.e., places where a non-linearity bends the space. Naturally, the better the placing of these knots at initialization, the faster the approximation may converge. For applications in which the data is centered around zero, initializing the knots around zero is a good inductive bias.\footnote{This is why $\bv{=}\vec{0}$ is common in regular initialization schemes.} However, for spatially uniform distributed inputs, the knots should be uniformly distributed (Fig.~\ref{fig:2_splineapprox}). As a result, conventional initializations lead to very poor reconstructions ($\mathrm{ReLU}$ $\mathrm{0}$-$\mathrm{Init}$, Fig.~\ref{fig:2_summary_maintext_kernelfit}), and explicitly aggregating positional encodings to the mappings leads to important improvements, e.g, \cite{mildenhall2020nerf}.

For $\mathrm{ReLU}$ layers $\yv {=} \mathrm{max}\{\vec{0}, \Wm \xv + \bv\}$ knots appear at the point where $\vec{0} {=} \Wm \xv + \bv$. To place the knots at $\xv{=}\vec{0}$, it is sufficient to set the bias to zero: $\bv{=}\vec{0}$. For uniformly distributed knots in a range $[\xv_{\text{min}}, \xv_{\text{max}}]$, however, one must solve the $\mathrm{ReLU}$ equation for uniformly distributed points in that range: $\vec{0} {=} \Wm \xv_{\text{unif}} + \bv$. It results that $\bv {=} - \Wm \xv_{\text{unif}}$, for arbitrary values of $\Wm$.

In multilayered networks, the approximation problem can be understood as reconstructing the target function in terms of a basis $\hv^{(\Lt-1)}$. Consequently, the expressivity of the network is determined by the number of knots in $\hv^{(\Lt-1)}$. In theory, each $\mathrm{ReLU}$ layer is able to divide the linear regions of the previous layer in exponentially many sub-regions \citep{montufar2014number, serra2018bounding}, or equivalently, to induce an exponential layer-wise increase in the number of knots. For the first layer, the positions of the knots are described by the bias term, and for subsequent layers, these positions also depend on $\Wm^{(l)}$. Unfortunately, as depicted by \citet{hanin2019complexity}, slight modifications of $\{\Wm^{(l)}, \bv^{(l)}\}$ can strongly simplify the landscape of the linear regions, and thus the knots (Fig.~\ref{fig:2_sawtooth_hanin}). More importantly,  \citet{hanin2019complexity} showed that the number of linear regions at initialization is actually equal to a constant times the number of neurons in the network (with a constant very close to one in their experiments). In addition, they show that this behavior barely changes throughout training.

\textbf{An improved initialization scheme.} Following the previous reasoning, we explore inducing a uniformly distributed initialization of the knots. However, we observe that finding an initialization with an exponential number of knots is a cumbersome and unstable procedure. In fact, it is not always possible, and, whenever possible, it strongly restricts the values the weights $\Wm^{(l)}$ can assume.

Following the findings of \citet{hanin2019complexity}, we instead employ an initialization procedure with which the total number of knots is equal to the number of neurons of the network. 
This is obtained by replicating the initialization procedure of the first layer throughout the network: For randomly initialized weights $\Wm^{(l)}$, the bias term $\bv^{(l)}$ is given by the equality $\bv^{(l)} {=} - \Wm^{(l)} \hv^{(l)}(\xv_{\text{unif}})$, where $\xv_{\text{unif}}$ is a vector of uniformly distributed points in $[\xv_{\text{min}}, \xv_{\text{max}}]$.
Interestingly, we observe that this initialization strategy consistently outperforms the standard initialization for a large range of target functions ($\mathrm{ReLU}$ $\mathrm{Unif}$-$\mathrm{Init}$, Fig.~\ref{fig:2_kernelfitting_all}). However, we note that $\mathrm{ReLU}$ networks still show large difficulties in representing very nonlinear and non-smooth functions. In Fig.~\ref{fig:2_kernelfitting_all}, we illustrate that other popular nonlinearities: $\mathrm{LeakyReLU}$, $\mathrm{Swish}$, exhibit the same behavior.

\vspace{-7mm}
\subsection{Kernel Parameterization via $\mathrm{Sine}$ Networks} \label{appx:2_sine}
Recently, \citet{sitzmann2020implicit} proposed to replace $\mathrm{ReLU}$ nonlinearities by $\mathrm{Sine}$ for the task of implicit neural representation learning. Based on their relation with implicit neural representations, we explore using $\mathrm{Sine}$ networks to parameterize our continuous kernels. Intriguingly, we observe that this slight modification allows our kernels to approximate \textit{any} provided function to near perfection, and leads to a consistent improvement for all tasks considered in this paper (Appx.~\ref{appx:2_sinevsothers}, Fig.~\ref{fig:2_kernelfitting_all}).  

A possible explanation for these impressive results can be given via our prior analysis:
\begin{figure}[t]
\includegraphics[width=.25\textwidth]{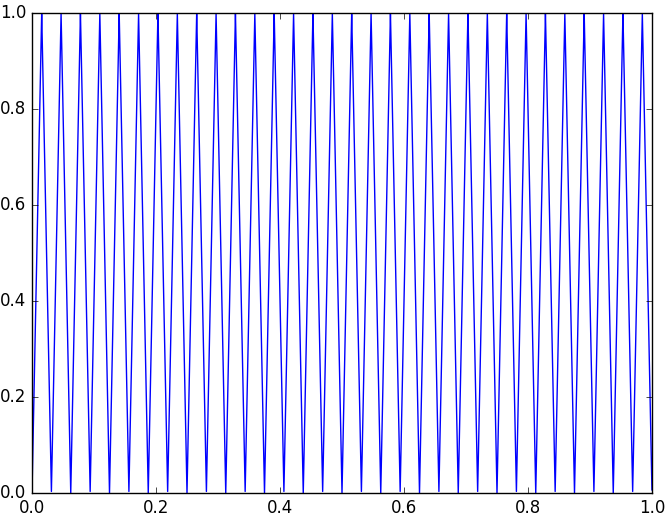}\hspace{2mm}
\includegraphics[width=.25\textwidth]{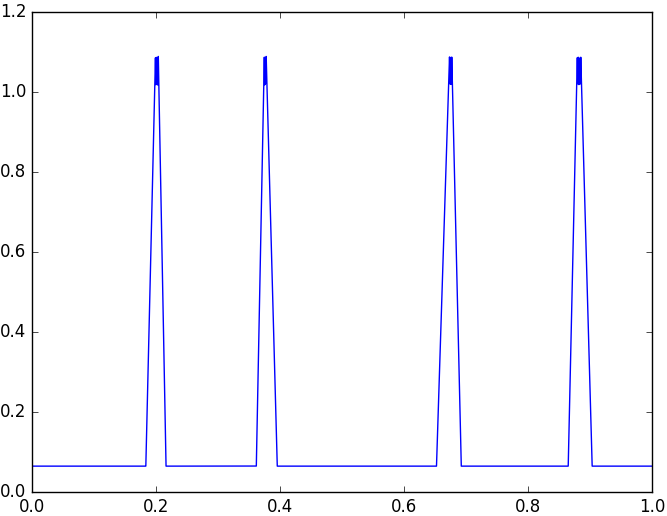}
\vskip.1in
\includegraphics[width=.45\textwidth]{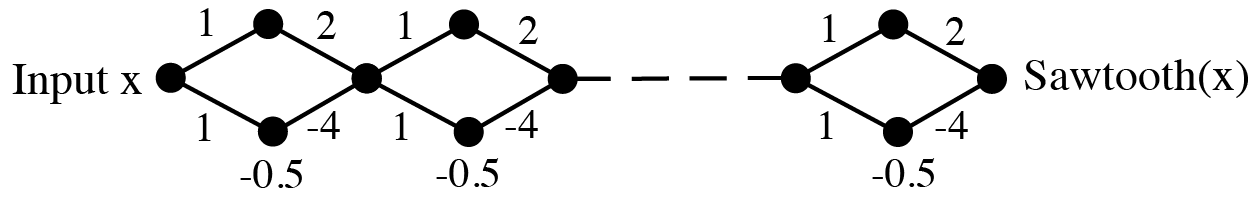}
\vspace{-2mm}
\caption{The sensitivity of networks with layer-wise exponential growing to slight changes. Taken from \citet{hanin2019complexity}. The sawtooth function with $2^n$ teeth (left) can be easily expressed via a $\mathrm{ReLU}$ network with $3n+4$ neurons (bottom). However, a slight perturbation of the network parameters --Gaussian noise with standard deviation $0.1$-- greatly simplifies the linear regions captured by the network, and thus the distribution of the knots in the basis (right).}
\label{fig:2_sawtooth_hanin}
\end{figure}

\textbf{Periodic bending of the space.} A $\mathrm{Sine}$ layer is given by: $\yv=\mathrm{Sin}(\omega_0[\Wm \xv + \bv])$, where $\omega_0$ works as a prior on the variability of the target function. Orthogonal to $\mathrm{ReLU}$ layers, $\mathrm{Sine}$ layers periodically bend the space. As a result, the same $\yv$ value is obtained for all bias values $\bv_{i}'{=} \bv_{i} + n 2\pi \| \Wm_{i,:}\|^{-1}$, $\forall n \in \sZ$. This is important from a spline approximation perspective. While for $\mathrm{ReLU}$ layers a unique value of $\bv$ exists that bends the space at a desired position, infinitely many values of $\bv$ do so for $\mathrm{Sine}$ ones. Resultantly, $\mathrm{Sine}$ layers are much more robust to parameter selection, and can be tuned to benefit pattern approximation at arbitrary --or even multiple-- positions in space (Fig.~\ref{fig:2_splineapprox}, right). We conjecture that this leads to more reliable approximations and faster convergence. 

\textbf{An exponentially big Fourier basis.} It is not surprising for a (large) basis of phase-shifted sinusoidal functions to be able to approximate arbitrary functions with high fidelity. This result was first observed over two centuries ago by \citet{fourier1807memoire} and lies at the core of the well-known \textit{Fourier transform}: any integrable function can be described as a linear combination of a (possibly) infinite basis of phase-shifted sinusoidal functions. \citet{sitzmann2020implicit} proposed an initialization of $\{ \Wm^{(l)} \}$ that allows for the construction of deep $\mathrm{Sine}$ networks able to periodically divide the space into exponentially many regions as a function of depth. Intuitively, approximations via $\mathrm{Sine}$ networks can be seen in terms of an exponentially large Fourier-like basis.
We conjecture that this exponential growth combined with the periodicity of sine is what allows for excellent approximations: the more terms in a Fourier transform, the better the approximation becomes.

Interestingly, we find that a uniformly distributed initialization of the bias term $\bv_{i} \sim \mathcal{U}(-\pi \| \Wm_{i,:}\|^{-1},\pi \| \Wm_{i,:}\|^{-1})$ also leads to better and faster convergence for $\mathrm{Sine}$ networks.

\vspace{-7mm}
\section{Dataset Description}\label{appx:2_datadescription}

\textbf{Copy Memory Problem.} The copy memory task consists of sequences of length $T +$20, for which the first 10 values are chosen randomly among the digits $\{1, . . . , 8\}$, the subsequent $T -$1 digits are set to zero, and the last 11 entries are filled with the digit 9. The goal is to generate an output of the same size of the input filled with zeros everywhere except for the last 10 values, for which the model is expected to predict the first 10 elements of the input sequence. 

\textbf{The Adding Problem.} The adding problem consists of input sequences of length $T$ and depth 2. The first dimension is filled with random values in $[0, 1]$, whereas the second dimension is set to zeros except for two elements marked by 1. The objective is to sum the random values for which the second dimension is equal to 1. Simply predicting the sum to be 1 results in a MSE of about 0.1767. 

\textbf{Sequential and Permuted MNIST.} The MNIST dataset \citep{lecun1998gradient} consists of 70{\sc{k}} gray-scale $28\times28$ handwritten digits divided into training and test sets of 60{\sc{k}} and 10{\sc{k}} samples, respectively. The sequential MNIST dataset (sMNIST) presents MNIST images as a sequence of 784 pixels for digit classification. Consequently, good predictions require preserving long-term dependencies up to 784 steps in the past: much longer than most language modelling~tasks~\citep{bai2018trellis}.

The permuted MNIST dataset (pMNIST) additionally permutes the order or the sMNIST sequences at random. Consequently, models can no longer rely on on local features to perform classification. As a result, the classification problem becomes more difficult and the importance of long-term dependencies more pronounced.   

\textbf{Sequential CIFAR10.} The CIFAR10 dataset \citep{krizhevsky2009learning} consists of 60{\sc{k}} real-world $32\times32$ RGB images uniformly drawn from 10 classes divided into training and test sets of 50{\sc{k}} and 10{\sc{k}} samples, respectively. Analogously to the sMNIST dataset, the sequential CIFAR10 (sCIFAR10) dataset presents CIFAR10 images as a sequence of 1024 pixels for image classification. This dataset is more difficult than sMNIST, as \emph{(i)} even larger memory horizons are required to solve the task, and \emph{(ii)} more complex structures and intra-class variations are present in the images \citep{trinh2018learning}. 

\textbf{CharacterTrajectories.} The CharacterTrajectories dataset is part of the UEA time series classification archive \citep{bagnall2018uea}. It consists of 2858 time series of different lengths and 3 channels representing the $x,y$ positions and the pen tip force while writing a Latin alphabet character in a single stroke The goal is to classify which of the different 20 characters was written using the time series data. The maximum length of the time-series is 182.

\textbf{Speech Commands.} The Speech Commands dataset \citep{warden2018speech} consists of 105809 one-second audio recordings of 35 spoken words sampled at $16$kHz. Following \citet{kidger2020neural}, we extract 34975 recordings from ten spoken words to construct a balanced classification problem. We refer to this dataset as \textbf{SC\_raw}. In addition, we utilize the preprocessing steps of \citet{kidger2020neural} and extract mel-frequency cepstrum coefficients from the raw data. The resulting dataset, named \textbf{SC}, consists of time series of length 161 and 20 channels. 

\textbf{PhysioNet.} The PhysioNet 2019 challenge on sepsis prediction \citep{goldberger2000physiobank, reyna2019early} is a irregularly sampled, partially observed dataset consisting of 40335 time series of variable length describing the stay of patients within an ICU. Time-series are made out of 5 static features, e.g., age, and 34 time-dependent features, e.g., respiration rate, creatinine blood concentration, and 10.3\% of the values are observed. We follow \citet{kidger2020neural} and consider the first 72 hours of a patient's stay to predict whether sepsis is developed over the course of their entire stay --which can extend for a month for some patients--. 

\textbf{PennTreeBank.} The PennTreeBank (PTB) \citep{marcinkiewicz1994building} is a language corpus which consists of 5,095K characters for training, 396K for validation and 446K for testing. On a char lever that we use in our experiment the vocabulary size is 50 characters (or the size of the alphabet, including end-of-string char). We follow \cite{bai2018empirical} in performing character-level language modeling task on this dataset.

\vspace{-7mm}
\section{Ablation Studies}\label{appx:2_ablation}
In this section, we perform an ablative study of our approach. Specifically, we analyze the effect of multiple components of our network, and provide additional comparisons with alternative architectures. Specifications on the architectures and hyperparameters used are given in Appx.~\ref{appx:2_expdetails}.

\vspace{-7mm}
\subsection{Using Sine Non-Linearities Over Popular Alternatives}\label{appx:2_sinevsothers}
As shown in Sec.~\ref{sec:2_ckconvkernel}, Sine nonlinearities provide astonishing improvements over equivalent networks with $\mathrm{ReLU}$ nonlinearities for function reconstruction. In this section, we provide additional experiments to highlight the suitability of Sine nonlinearities over other popular alternatives both for function approximation and the rest of the tasks considered in this work. The same architectures are used across all experiments and vary only in the nonlinearity used in \mlp$^{\boldsymbol{\psi}}$. We find that nonlinearities other than $\mathrm{Sine}$ benefit from layer normalization and thus we incorporate it in these variants.

\textbf{Case I: Function Approximation via \mlp$^{\boldsymbol{\psi}}$.} First, we evaluate the problem of function approximation in Sec.~\ref{sec:2_ckconvkernel}, Fig.~\ref{fig:2_summary_maintext_kernelfit}, for nonlinearities other than $\mathrm{ReLU}$ and $\mathrm{Sine}$. In particular, we approximate several functions with a \mlp$^{\boldsymbol{\psi}}$ network which varies only in the type of nonlinearity used: $\mathrm{ReLU}$ \citep{nair2010rectified}, $\mathrm{LeakyReLU}$ \citep{xu2015empirical}, $\mathrm{Swish}$ \citep{ramachandran2017searching}, and $\mathrm{Sine}$ \citep{sitzmann2020implicit}. 

\begin{figure*}
    \centering
    \includegraphics[width=0.98\textwidth]{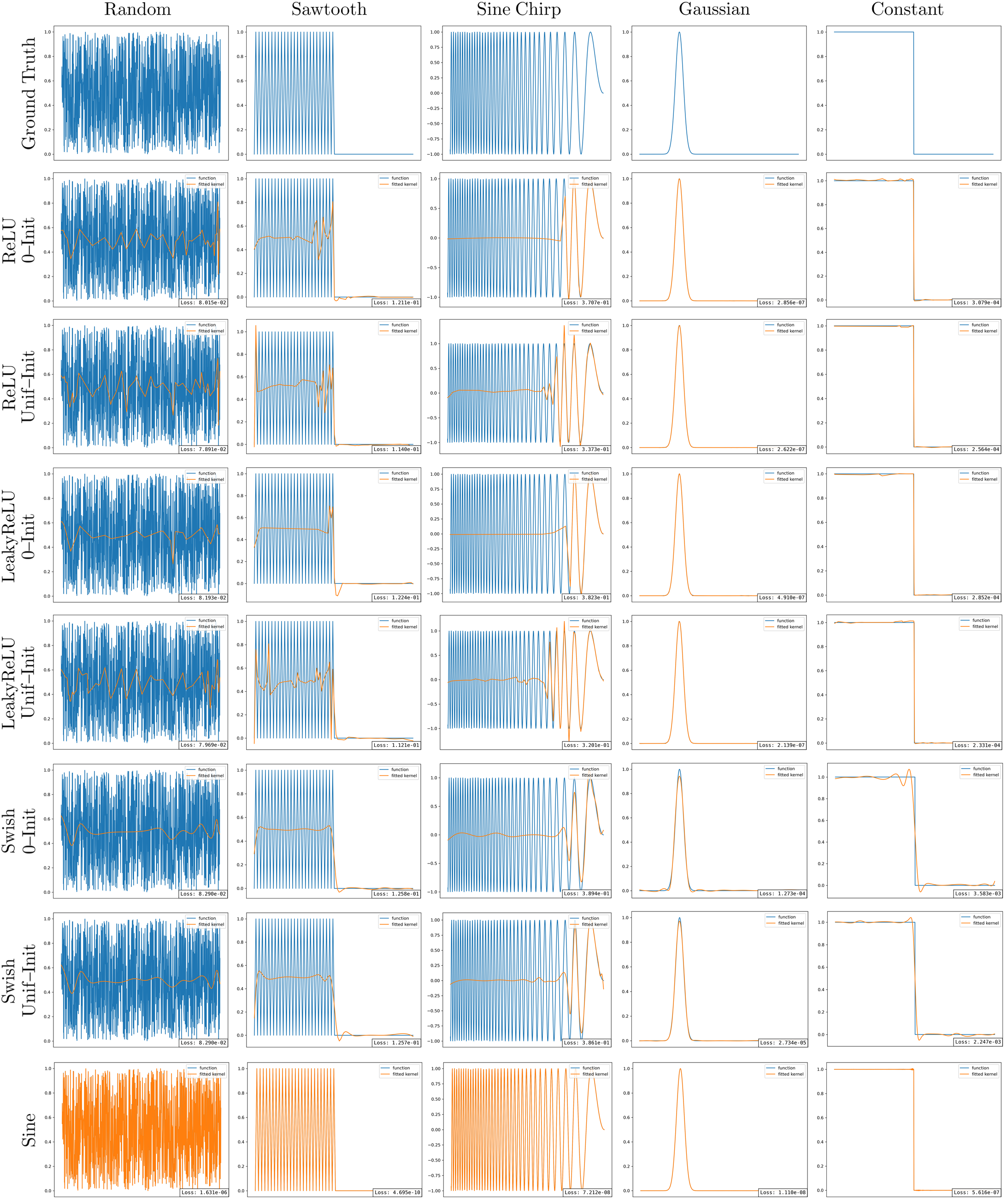}
    \vspace{-2mm}
    \caption{Function approximation via $\mathrm{ReLU}$, $\mathrm{LeakyReLU}$, $\mathrm{Swish}$ and $\mathrm{Sine}$ networks. All network variants perform a decent job in approximating simple functions. However, for non-linear, non-smooth functions, all networks using nonlinearities other than $\mathrm{Sine}$ provide very poor approximations. Interestingly, the uniform knot initialization proposed in Sec.~\ref{sec:2_ckconvkernel} provides consistent improvements for all network variants. However, despite this improvement, the approximation results remain insufficient. Contrarily, $\mathrm{Sine}$ networks quickly and seamlessly approximate all functions. All network configurations are equal up to the non-linearities used.}
    \label{fig:2_kernelfitting_all}
\end{figure*}

Our results (Fig.~\ref{fig:2_kernelfitting_all}), illustrate that Sine provides astonishing approximation capabilities over all other nonlinearities considered. In particular, we observe that Sine is the only nonlinearity able to reconstruct very nonlinear and very non-smooth functions, while all other alternatives fail poorly.

\textbf{Case II: CKCNNs with nonlinearities other than $\mathrm{Sine}$.} 
Next, we consider the case in which CKCNNs with nonlinearities other than $\mathrm{Sine}$ are used to solve the tasks considered in Sec.~\ref{sec:2_experiments}. In particular, we train CKCNNs on sMNIST, pMNIST, SC and SC\_raw for four different nonlinearities: $\mathrm{ReLU}$, $\mathrm{LeakyReLU}$, $\mathrm{Swish}$, $\mathrm{Sine}$. We utilize the same backbone architecture used in the main text for the corresponding dataset. % All other architectural parameters are kept intact.

Our results (Tab.~\ref{tab:2_nonlinearitiescomp}) indicate that CKCNNs relying on $\mathrm{Sine}$ nonlinearities outperform CKCNNs relying on all other nonlinearities. 

\textbf{Analysis of the results.} Our findings indicate $\mathrm{Sine}$ is much better suited to describe continuous spatial functions via neural networks than all other nonlinearities considered. This result motivates replacing popular nonlinearities by $\mathrm{Sine}$ for applications in which neural networks are used to describe continuous positional functions. This family of models encompass --but is not restricted to-- continuous types of convolutions, e.g., \citet{schutt2017schnet, thomas2018tensor, finzi2020generalizing, fuchs2020se}, as well as positional encodings in transformers, e.g., \citet{dai2019transformer, ramachandran2019stand, romero2021group}, and graph neural networks, e.g., \citet{defferrard2020deepsphere}. We consider this result to be of large relevance to the deep learning community.

\begin{table}
\RawFloats
\centering
\begin{minipage}{0.55 \textwidth}
\centering
\caption{Test accuracies of CKCNNs with different nonlinearities. Model size ${=}$ 100{\sc{k}}.}
\label{tab:2_nonlinearitiescomp}
\begin{center}
\vskip -3mm
\begin{small}
\scalebox{0.8}{
\begin{tabular}{ccccc}
\toprule
\multirow{2.5}{*}{\sc{Non-Linearity}} & \multicolumn{4}{c}{\sc{Dataset}} \\
\cmidrule{2-5}
& \sc{sMNIST} & \sc{pMNIST} & \sc{SC} & \sc{SC\_raw} \\
 \midrule
 \sc{ReLU} & 81.21 & 59.15 & 94.97 & 49.15 \\
 \sc{LeakyReLU} & 80.57 & 55.85 & 95.03 & 38.67 \\
 \sc{Swish} & 85.20 & 61.77 & 93.43 & 62.23 \\
 \sc{Sine} & \textbf{99.31} & \textbf{98.00} & \textbf{95.27} & \textbf{71.66} \\
\bottomrule
\end{tabular}}
\end{small}
\end{center}
\end{minipage}%
\hfill
\begin{minipage}{0.44 \textwidth}
\centering
\caption{Test accuracy of CKCNNs for various depths and widths.}
\label{tab:2_depthcomp}
\begin{center}
\vskip -3mm
\begin{small}
\scalebox{0.8}{
\begin{tabular}{ccccc}
\toprule
 \multicolumn{5}{c}{\sc{pMNIST}} \\
\midrule
\multirow{2}{*}{\sc{Depth}} & \multicolumn{2}{c}{\sc{Fixed Width}} &  \multicolumn{2}{c}{\sc{Fixed Size}}   \\
 & \sc{Size} & \sc{Acc.}(\%) &   \sc{Size} & \sc{Acc.}(\%) \\
 \midrule
2 Blocks   & 98k & 99.21& 98k  & \textbf{99.21}\\
4 Blocks   & 225k& \textbf{99.26}& 95k  & \textbf{99.19}\\
8 Blocks   & 480k& \textbf{99.29}& 105k & 99.12\\
16 Blocks  & 990k& 99.19& 107k & 99.02\\
\bottomrule
\end{tabular}}
\end{small}
\end{center}
\end{minipage}
\end{table}

\vspace{-7mm}
\subsection{Going Deeper with CKCNNs}\label{appx:2_depthvsshallow}
The experimental results shown in Sec.~\ref{sec:2_experiments} are obtained by shallow CKCNNs with 2 residual blocks. An interesting question is whether going deeper can be used to improve the performance of CKCNNs. To analyze this, we compare deep and shallow CKCNNs  with the same architecture for equal width, and equal number of parameters. 

Our results (Tab.~\ref{tab:2_depthcomp}) indicate that constructing deeper CKCNNs do not lead to accuracy improvements. In fact, we observe that deep CKCNNs underperform shallow CKCNNs of equal size. This is an interesting results as CKCNNs do not strongly rely on deep-wise compositionality of features: largely considered indispensable in deep learning. 

\textbf{Analysis of the results.} The dynamics governing these results are not fully understood. Our findings may lead to two different conclusions, both of which we consider important for the development and understanding of CKCNNs and deep learning in general:

\textbf{\textit{Outcome I: Deep CKCNNs.}} The first possible outcome is that our current parameterization does not correctly leverage depth. In this case, efforts to construct proper \textit{deep} CKCNNs will likely lead to performance improvements over the current architectures, and thus have the potential to advance the state-of-the-art further.

\textbf{\textit{Outcome II: Depth is not needed when global memory horizons are provided with shallow networks.}} The second possible outcome is that depth is used mainly as a means to construct global memory horizons. Consequently, neural networks do not have to be very deep at all provided that global memory horizons are defined by shallow neural networks. Interestingly, this conclusion is in line with the predominant design of recurrent architectures, for which a moderate number of layers are used, e.g., \cite{pascanu2013construct, graves2013speech, gu2020improving, gu2020hippo}. This possible outcome is very exciting as depth is largely considered indispensable in the deep learning community.

%  -=== Table Depth
\vspace{-7mm}
\section{Experimental Details}\label{appx:2_expdetails}
In this section, we provide extended details over our implementation as well as the exact architectures and optimization schemes used in our experiments.

\vspace{-7mm}
\subsection{General Remarks}
\begin{figure}
     \centering
     \includegraphics[width=0.8\textwidth]{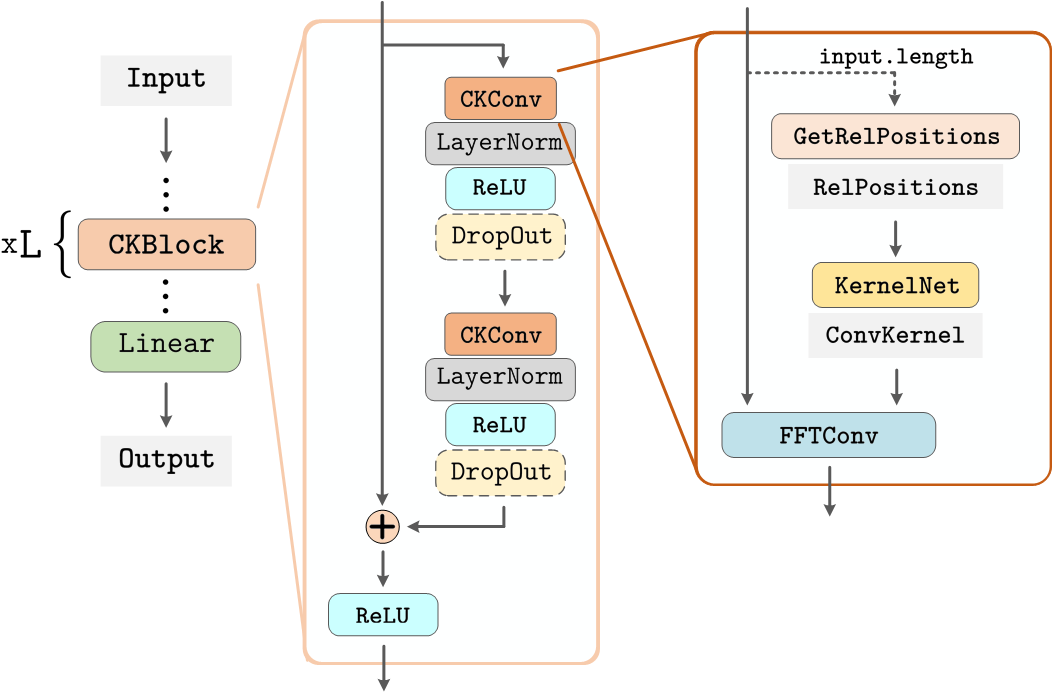}
     \vspace{-2mm}
     \caption{Graphical description of continuous kernel convolutional networks. Dot-lined blocks depict optional blocks, and blocks without borders depict variables. {\btt KernelNet} blocks use $\mathrm{Sine}$ nonlinearities. We replace spatial convolutions by Fourier Convolutions ({\btt FFTConv}), which leverages the convolution theorem to speed up computations.}
     \label{fig:2_ckcnn_structure}
 \end{figure}

 Our models follow the structure shown in Fig.~\ref{fig:2_ckcnn_structure} and vary only in the number of channels. We use layer normalization \citep{ba2016layer} in our backbone network, and use the Adam optimizer \citep{kingma2014adam} across all our experiments. Our code is implemented in {\btt PyTorch} and is publicly available at \textit{link removed for the sake of the double-blind review}.
 %\href{https://github.com/dwromero/ckconv}{\texttt{github.com/dwromero/ckconv}}. 
 We utilize {\btt wandb} \citep{wandb} to log our results, and use NVIDIA TITAN RTX GPUs throughout our experiments. 

\textbf{Continuous Convolutional Kernels \mlp$^{\boldsymbol{\psi}}$.}
All our convolutional kernels are parameterized by a 3-layer neural network with 32 hidden units and $\mathrm{Sine}$ nonlinearities: 
$$1 \rightarrow 32 \rightarrow 32 \rightarrow \Nt_{\mathrm{out}} \times \Nt_{\mathrm{in}}.$$
Here, $\Nt_{\mathrm{in}}$, $\Nt_{\mathrm{Cout}}$ are the number of input and output channels of the convolutional layer. We utilize weight normalization \citep{salimans2016weight} in our \mlp$^{\boldsymbol{\psi}}$ networks, and select a hidden size of 32 based on empirical evidence and findings from previous work \cite{finzi2020generalizing}.

\textbf{Normalized relative positions.} The \mlp s parameterizing our convolutional kenels receive relative positions as input. However, considering unitary step-wise relative positions, i.e., 0, 1, 2, ... , $\Nt$, can be problematic from a numerical stability perspective as $\Nt$ may grow very large, e.g., $\Nt{=}$16000 for the SC\_raw dataset. Consequently, we follow good practices from works modelling continuous functions with neural networks, and map the largest unitary step-wise relative positions seen during training $[0, \Nt]$ to a uniform linear space in $[-1, 1]$. 

\textbf{Hyperparameter tuning.} We tune the hyperparameters of our models via the {\btt bayes} method given in {\btt wandb} Sweeps, which selects hyperparameter values via a Gaussian process over the results obtained so far. We perform tuning on a validation dataset until a predefined maximum number of runs of $100$ is exhausted. Further improvements upon our results may be obtained by leveraging more sophisticated tuning methods as well as additional runs. 

\textbf{Selecting $\boldsymbol{\omega_0}$.} CKCNNs are very susceptible to the value of $\omega_0$. In order to obtain a reasonable $\omega_0$, we first perform a random search on a large interval $\omega_0 \in [0, 3000]$. After a few runs, we stop the random search and select the subinterval in which the validation accuracy is most promising. Next, we restart the random search on this sub-interval and repeat the process until a $\omega_0$ value is obtained, for which the validation accuracy is sufficiently high. Surprisingly, we found optimal values of $\omega_0$ to be always enclosed in the interval $[1, 70]$ even for very long sequences as in SC\_raw.

\vspace{-7mm}
\subsection{Accounting for Spatial Displacements of the Sampled Convolutional Kernels}\label{appx:2_spatialdisplacement}
We follow the sampling procedure of
\citet{gu2020hippo} throughout our test sampling rate discrepancy experiments. Specifically, for a sequence {\btt seq} of length $\Nt$, subsampling by a factor {\btt n} is performed by running {\btt seq[::n]}. That is, by taking the {\btt n}-th element of the sequence starting from its first element. For example, for a sequence of length $\Nt{=}182$, different values of {\btt n} would yield the following sequences:

{(\btt n = 1)} $\rightarrow$ {\btt [1, 2, 3, \ ... , 180, 181, 182]}\newline
{(\btt n = 2)} $\rightarrow$ {\btt [1, 3, 5, \   ... , 177, 179, 181]}\newline
{(\btt n = 4)} $\rightarrow$ {\btt [1, 5, 9, \  ... , 173, 177, 181]}\newline
{(\btt n = 8)} $\rightarrow$ {\btt [1, 9, 17, ... , 161, 169, 177]}

Recall that \mlp$^{\boldsymbol{\psi}}$ takes normalized relative positions in $[-1, 1]$ as input, which are computed based on the max input length seen during training. However, some of these subsampling transitions \textit{change} the max value of the sequence, e.g., for {(\btt n = 8)} the maximum is given by 177, whereas for {(\btt n = 1)} this value corresponds to 182. Consequently, a naive approach would consider the last position in each subsampled sequence to correspond to the maximum normalized relative position $1$. This effectively induces displacement and re-scaling of the sampled convolutional kernel used during training. 

This misalignment is automatically handled under the hood in our {\btt CKConv} implementation, but we highlight this subtlety to prevent it in future applications.

\vspace{-7mm}
\subsection{Dealing with High-Frequency Components}\label{appx:2_high_freq}
\begin{figure}
     \centering
     \includegraphics[width=0.95\textwidth]{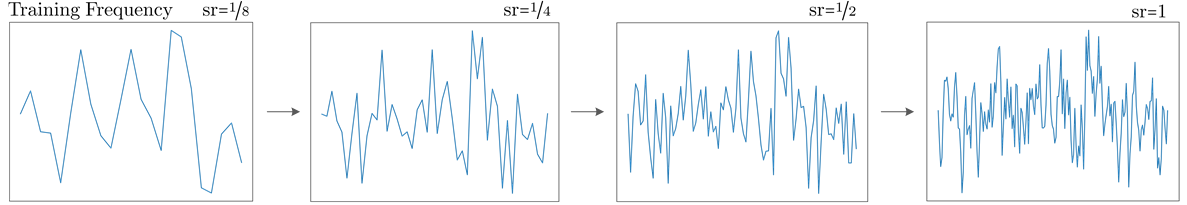}
     \vspace{-2mm}
     \caption{High-frequency components in $\mathrm{Sine}$ continuous kernels. We observe that continuous kernels parameterized by Sine networks often contain frequency components of frequency higher than the resolution of the grid used during training. Here, for instance, the kernel looks smooth on the training grid. However, several high-frequency components appear when sampled on a finer grid. Although this may be problematic, we believe that, if used properly, these high-frequency components can prove advantageous to model fine details in tasks such as super-resolution and compression.}
     \label{fig:2_high_freq}
 \end{figure}

Interestingly, our experiments revealed that our continuous kernels often contain frequency components of frequency higher than the resolution of the sampling grid used during training (Fig.~\ref{fig:2_high_freq}). As these high-frequency components are not observed during training, we observe that they hurt performance when evaluated at higher resolutions. 

In order to neutralize their influence, we filter these components before doing the convolution by means of blurring. This is done by applying a convolution upon the convolutional kernel with a Gaussian filter $G$ of length $2 \tfrac{\mathrm{sr}_\text{test}}{\mathrm{sr}_\text{train}} + 1$ and parameters $\mu{=}0$, $\sigma{=}0.5$:
$$\Big[G\big( - \tfrac{\mathrm{sr}_\text{test}}{\mathrm{sr}_\text{train}} \big) , G\big(-\tfrac{\mathrm{sr}_\text{test}}{\mathrm{sr}_\text{train}}+1\big), ... , G\big(0\big), ... , G\big(\tfrac{\mathrm{sr}_\text{test}}{\mathrm{sr}_\text{train}}-1\big), G\big(\tfrac{\mathrm{sr}_\text{test}}{\mathrm{sr}_\text{train}}\big)\Big]$$

Note that blurring is only used when the test sampling rate is higher than the train sampling rate, as opposed to the normalization factor $\tfrac{\mathrm{sr}_\text{test}}{\mathrm{sr}_\text{train}}$ discussed in Eq.~\ref{eq:2_diff_sampling_rates}, Appx.~\ref{appx:2_diff_sampling_rates}, which is applied whenever the sampling rates differ.

\vspace{-7mm}
\subsection{Hyperparameters and Experimental Details}\label{appx:2_hyperparams}
In this section, we specify the hyperparameter configurations with which our models are trained. An overview of these hyperparameters is provided in Tab.~\ref{tab:2_hyperparams}.

\begin{table*}[t!]
% \centering
\caption{Hyperparameter specifications of the best performing CKCNN models.}
\label{tab:2_hyperparams}
\begin{center}
\vskip -3mm
\begin{small}
\scalebox{0.6}{
\begin{tabular}{lccccccccc}
\toprule
\multirow{2}{*}{\sc{Params.}} & \multirow{2}{*}{\sc{Copy Memory}} & \multirow{2}{*}{\sc{Adding Problem}} & {\sc{sMNIST}} & {\sc{pMNIST}} & {\sc{sCIFAR10}} & \multirow{2}{*}{\sc{CT$^\dagger$}} & \multirow{2.25}{*}{\sc{SC}} & \multirow{2}{*}{\sc{SC\_raw}$^\dagger$}  & \multirow{2}{*}{\sc{PTB}}   \\
 & & & Small / Big & Small / Big & Small / Big \\
\midrule
Epochs & See Appx.~\ref{appx:2_hyperparams} &  See Appx.~\ref{appx:2_hyperparams} & 200 & 200 & 200 & 200 & 200 & 300 & 200\\
Batch Size & 32 & 32 & 64 & 64 & 64 & 32 & 64 & 32 & 24\\
Optimizer & Adam & Adam & Adam & Adam & Adam & Adam & Adam & Adam & Adam\\
Learning Rate & 5e-4 & 0.001 & 0.001 & 0.001 & 0.001 & 0.001 & 0.001 & 0.001 & 0.002\\
\# Blocks & 2 & 2 & 2 & 2 & 2 & 2 & 2 & 2 & 2\\
Hidden Size & 10 & 25 & 30 / 100  & 30 / 100  & 30 / 100 & 30 & 30 & 30 & 128\\
$\omega_0$ & See Appx.~\ref{appx:2_hyperparams} & See Appx.~\ref{appx:2_hyperparams} & 31.09 / 30.5 & 43.46 / 42.16  &  25.67 & 21.45 & 30.90 & 39.45 & 25.78\\
Dropout & -  & - & 0.1 / 0.2 & - & 0.2 / 0.3  & 0.1  & 0.2 & - & - \\
Input Dropout & -  & - & 0.1 / 0.2 & 0.1 / 0.2 & 0.0 / 0.0 & -  & - & - & 0.1\\
Embedding Dropout & - & - & - & - & - & - & - & - & 0.1 \\
Weight Dropout & -  & - & - & - & - / 0.1 & - & - & -& - \\
Weight Decay & -  & - & - & - & - / 1e-4 & - & - & 1e-4 & 1e-6\\
Scheduler & -  & - & Plateau & Plateau & Plateau & Plateau & Plateau & Plateau & Plateau \\
Patience & - & - & 20 & 20 & 20 & 20  & 15 & 20 & 5\\
Scheduler Decay & - & - & 5 & 5 & 5 & 5  & 5  & 5 & 5\\
\midrule
Model Size & 15.52{\sc{k}} & 70.59{\sc{k}} & 98.29{\sc{k}} / 1.03{\sc{m}} & 98.29{\sc{k}} / 1.03{\sc{m}} & 100.04{\sc{k}} / 1.04{\sc{m}} & 100.67{\sc{k}} & 118.24{\sc{k}} &  98.29{\sc{k}}& 1.8{\sc{m}}\\ 
\bottomrule
\multicolumn{9}{l}{$^\dagger$ Hyperparameter values for the classification and varying sampling rate tasks. For hyperparameters w.r.t. irregularly-sampled data please see Tab.~\ref{tab:2_hyperparams_irreg}. }
\end{tabular}
}
\end{small}
\end{center}
\end{table*}

 \textbf{Copy Memory.} We set the number of channels of our CKCNN as to roughly match the number of parameters of the GRU and TCN networks of \citet{bai2018empirical}. This is obtained with 10 hidden channels at every layer. We observe that the time to convergence grew proportional to the length of the sequence considered. Whereas for sequences of length 100 convergence was shown after as few as 10 epochs, for sequences of length 6000 approximately 250 epochs were required. The maximum number of epochs is set to 50, 50, 100, 200 and 300 for sequences of size 100, 200, 1000, 3000 and 6000. We observe that different values of $\omega_0$ are optimal for different sequence lengths. The optimal $\omega_0$ values found are 19.20, 34.71, 68.69, 43.65 and 69.97 for the corresponding sequence lengths.
 
\textbf{Adding Problem.} We set the number of channels of our CKCNN as to roughly match the number of parameters of the GRU and TCN networks of \citet{bai2018empirical}. This is obtained with 25 hidden channels at every layer. Similarly to the Copy Memory task, we observe that the time to convergence grew proportional to the length of the sequence considered. Interestingly, this task was much easier to solve for our models, with convergence for sequences of length 6000 observed after 38 epochs. The maximum number of epochs is set to 20, 20, 30, 50 and 50 for sequences of size 100, 200, 1000, 3000 and 6000. We observe that different values of $\omega_0$ are optimal for different sequence lengths. The optimal $\omega_0$ values found are 14.55, 18.19, 2.03, 2.23 and 4.3 for the corresponding sequence lengths. 

\textbf{sMNIST, pMNIST and sCIFAR10.} We construct two models of different sizes for these datasets: CKCNN and CKCNN-Big. The first is constructed to obtain a parameter count close to 100{\sc{k}}. The second model, is constructed to obtain a parameter count close to 1{\sc{m}}. The parameters utilized for these datasets are summarized in Tab.~\ref{tab:2_hyperparams}. Despite our efforts, we observed that our models heavily overfitted sCIFAR10. Combinations of weight decay, dropout and weight dropout were not enough to counteract overfitting. 

\textbf{CT, SC and SC\_raw.} The parameters utilized for classification on these datasets are summarized in Tab.~\ref{tab:2_hyperparams}. For hyperparameters regarding experiments with irregularly-sampled data please refer to Tab.~\ref{tab:2_hyperparams_irreg}. Any non-specified parameter value in Tab.~\ref{tab:2_hyperparams_irreg} can be safely consider to be the one listed for corresponding dataset in Tab.~\ref{tab:2_hyperparams}.

\textbf{PennTreeBank} For a character-level language modeling on PTB dataset we use hyperparameters specified in Tab.~\ref{tab:2_hyperparams}. We use embedding of size 100 following the TCN model from \cite{bai2018empirical}.

\begin{table}[t]
\centering
\caption{Hyperparameter values for experiments on irregularly sampled data. Non-listed parameters correspond to those in Tab.~\ref{tab:2_hyperparams}.}
\label{tab:2_hyperparams_irreg}
\begin{center}
\vskip -3mm
\begin{small}
\scalebox{0.69}{
\begin{tabular}{lccccccc}
\toprule
\multirow{2.5}{*}{\sc{Params.}}  & \multirow{2.5}{*}{\sc{PhysioNet}} & \multicolumn{3}{c}{\sc{CT}} & \multicolumn{3}{c}{\sc{SC\_raw}}  \\
\cmidrule{3-8}
 & & \sc{(30\%)} & \sc{(50\%)} & \sc{(70\%)} & \sc{(30\%)} & \sc{(50\%)} & \sc{(70\%)} \\
\midrule
$\omega_0$ & 4.38 & 17.24 & 12.00 & 4.24 & 35.66 & 31.70 & 25.29\\
Dropout & 0.0 & 0.2  & 0.2 & 0.0 & 0.1 & 0 & 0\\
Weight Decay & 0.0 & 0.0 & 1e-4 & 0.0 & 1e-4 & 1e-4 & 1e-4 \\ 
Batch Size & 1024 \\
\midrule
Model Size & 175.71{\sc{k}} & \multicolumn{3}{c}{101.75{\sc{k}}} & \multicolumn{3}{c}{99.34{\sc{k}}}  \\ 
\bottomrule
\end{tabular}}
\end{small}
\end{center}
\end{table}

%=========================================================================

%% file: backmatter/appx_4_ccnn.tex
% path to figures directory
\graphicspath{{figures/4-ccnn/}}

%=========================================================================

% \begin{savequote}[75mm]
% Nulla facilisi. In vel sem. Morbi id urna in diam dignissim feugiat. Proin molestie tortor eu velit. Aliquam erat volutpat. Nullam ultrices, diam tempus vulputate egestas, eros pede varius leo.
% \qauthor{Quoteauthor Lastname}
% \end{savequote}

\chapter{Modelling Long Context in $N$D: From Task-Specific to a General Purpose CNN}
	%\label{chapter:}

%=========================================================================
\vspace{-7mm}
\section{Extended Related Work}\label{appx:4_extended}

In this section, we provide a more extensive treatment of similar methods related to continuous reparameterizations of convolutional kernels and their limitations in use for general purpose architectures.

\textbf{Rethinking convolutional kernels.} Several previous works investigate a reformulation of CNNs based on modifications to the classical convolutional layer. These works provide formulations of continuous convolutional kernels pursuing several different motivations.
Several works apply specifically to point-cloud data --for which discrete kernels are not appropriate-- by interpolating a set of weights to obtain a definition over the continuous input space \citep{hua2018pointwise, thomas2019kpconv} or expressing the kernels in a polynomial \citep{xu2018spidercnn} or neural network basis \citep{jia2016dynamic, schutt2017schnet, wang2018deep, wu2019pointconv}. These architectures often rely on sophisticated data augmentation and feature engineering schemes, perform pooling or downsampling operations, and have not been shown to perform well on regular data, e.g. images \citep{wu2019pointconv}.

Other work focuses on continuous kernel formulations to address specific architectural considerations, for example to increase receptive fields efficiently \citep{su2021log}, or to learn more appropriate kernel geometries \citep{dai2017deformable, jeon2017active}. In these examples, a fixed set of weights is interpolated over the kernel domain, which implicitly defines a low-resolution kernel that impedes modeling long term dependencies in a dense manner \citep{romero2022flexconv}. 

A different line of work incorporating continuous kernel definitions focuses on exploiting desirable properties of spatially structured kernels for implementing convolutions equivariant to roto-translations \citep{sifre2014rigid, worrall2017harmonic, weiler20183d, weiler2019general, bekkers2020bspline, finzi2020generalizing}, or dilation-translation transformations \citep{sosnovik2020scaleequivariant, worrall2019deep, sosnovik2021disco}. Although these formulations also in principle give us handles for a continuous definition of convolutional kernels over the spatial domain, we choose not to restrict the kernel functions learned in our models to be equivariant, as this limits applicability to settings in which such priors are not warranted.

Note that graph convolutions  \citep{kipf2016semi} may similarly be viewed as providing a convolution operation invariant to arbitrary permutation symmetries on the input. Since this operation disregards positional information, it essentially shares the kernel function over the domain. In practice, the bias to permutation invariance needlessly impedes network expressivity for data where such assumptions do not hold, such as sequences, images or point-clouds. 

Continuous Kernel Convolutions \citep{romero2022ckconv} may be seen as a special case of neural message passing \citep{gilmer2017neural} on a fully connected graph over the input, with messages conditioned on relative positions of nodes. In the case of FlexConv \citep{romero2022flexconv}, graph connectivity is dynamic, conditioned on distance.

\vspace{-7mm}
\section{Dataset description}\label{appx:4_dataset_description}

\textbf{Sequential and Permuted MNIST.} The MNIST dataset \cite{lecun1998gradient} consists of 70{\sc{k}} gray-scale $28{\times}28$ handwritten digits divided into training validation and test sets of 60{\sc{k}} and 10{\sc{k}} samples, respectively. For validation purposes, the training dataset is further divided into training and validation sets of 55{\sc{k}} and 5{\sc{k}} samples, respectively. 

The sequential MNIST dataset (sMNIST) presents MNIST images as a sequence of 784 pixels for digit classification. Consequently, good predictions require the model to preserve long-term dependencies up to 784 steps in the past. The permuted MNIST dataset (pMNIST) incorporates an additional level of difficulty by permuting the order of all sMNIST sequences with a random permutation. Resultantly, models can no longer rely on local information for the construction of their features and the importance of modelling long-term dependencies becomes more pronounced.   

\textbf{CIFAR10, CIFAR100 and Sequential CIFAR10.} The CIFAR10 dataset \cite{krizhevsky2009learning} consists of 60{\sc{k}} real-world $32{\times}32$ RGB images uniformly drawn from 10 classes divided into training and test sets of 50{\sc{k}} and 10{\sc{k}} samples, respectively. The CIFAR100 dataset \cite{krizhevsky2009learning} is similar to the CIFAR10 dataset, with the difference that the images are now uniformly drawn from 100 different classes. For validation purposes, the training dataset of both CIFAR10 and CIFAR100 are further divided into training and validation sets of 45{\sc{k}} and 5{\sc{k}} samples, respectively. 

Analogously to the sMNIST dataset, the sequential CIFAR10 (sCIFAR10) dataset presents CIFAR10 images as a sequence of 1024 pixels for image classification. This dataset is more difficult than sMNIST, as \emph{(i)} larger memory horizons are required to solve the task, and \emph{(ii)} more complex structures and intra-class variations are present. 

\textbf{Speech Commands.} The Speech Commands dataset \cite{warden2018speech} consists of 105809 one-second audio recordings of 35 spoken words sampled at $16$kHz. Following \citet{kidger2020neural}, we extract 34975 recordings from ten spoken words to construct a balanced classification problem. We refer to this dataset as \textit{Raw Speech Commands}. In addition, we use the preprocessing steps of \citet{kidger2020neural} and extract mel-frequency cepstrum coefficients from the raw data. The resulting dataset, referred to as \textit{MFCC Speech Commands}, consists of time series of length 161 and 20 channels. 

\textbf{Long Range Arena.} The Long Range Arena benchmark \citep{tay2021long} consists of 6 tasks with lengths 1{\sc{k}}-16{\sc{k}} steps encompassing modalities and objectives that require similarity, structural, and visuospatial reasoning. The \texttt{Pathfinder}, \texttt{Path-X} and \texttt{Image} tasks are similar in nature to the sMNIST and sCIFAR10 tasks. These tasks consists of classification tasks performed on images that are treated as sequences.

\begin{wrapfigure}{r}{0.48\textwidth}
    \vspace{-4mm}
    \centering
    \hfill
         \begin{subfigure}[b]{0.23\textwidth}
         \centering
         \captionsetup{justification=centering}
         \includegraphics[width=0.9\textwidth]{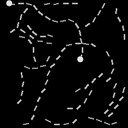}
         \caption{}
         \label{fig:4_positive}
     \end{subfigure}
     \hfill
         \begin{subfigure}[b]{0.23\textwidth}
         \centering
         \captionsetup{justification=centering}
         \includegraphics[width=0.9\textwidth]{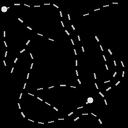}
         \caption{}
         \label{fig:4_negative}
     \end{subfigure}
    \caption{Positive and negative samples from the \texttt{Path-X} dataset.}
    \label{fig:4_pathx_samples}
\end{wrapfigure}

The \texttt{Image} task corresponds to the sequential CIFAR10 dataset with the only difference that the CIFAR10 images are treated as gray-scale images. The \texttt{Pathfinder} and \texttt{Path-X} tasks are binary tasks in which binary images are provided and the model must predict whether the two points in the images are connected with a line or not --see Fig.~\ref{fig:4_pathx_samples} for an example--. The difference between both datasets is their resolution. Whereas \texttt{Pathfinder} has images of size 32$\times$32, \texttt{Path-X} has images of size 128$\times$128. It is important to mention that these tasks are so difficult that even if treated as 2D signals, CNNs without global receptive fields cant solve them \citep{gu2022efficiently}.

\textbf{STL-10.} The STL-10 dataset \citep{coates2011analysis} is a subset of the ImageNet dataset \citep{krizhevsky2012imagenet} consisting of 13,000 $96{\times}96$ real-world RGB images uniformly drawn from 10 classes divided into training
and test sets of 5{\sc{k}} and 8{\sc{k}} images, respectively. For validation purposes, the training dataset is further divided into training and validation sets of 4,500 and 500 samples, respectively. 

\textbf{ModelNet40.} The ModelNet40 dataset \citep{wu20153d} contains 12,311 3D meshes of objects belonging to 40 classes. We use the official split with 9,843 training samples and 2,468 validation samples. In contrast with previous works which sample 1,024 points from these meshes, (e.g. \cite{wu2019pointconv}), we sample 512 points uniformly from the faces of each mesh, along with the normal vectors at each of these positions. These position and normal vectors serve as input to the model. 
\begin{figure}
    \centering
        \begin{subfigure}[b]{0.21\textwidth}
         \centering
         \captionsetup{justification=centering}
         \includegraphics[width=\textwidth]{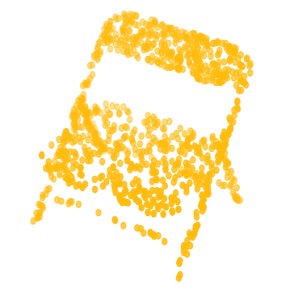}
         \caption{}
         \label{fig:4_sample_point_cloud}
     \end{subfigure}
     \hspace{3mm}
        \begin{subfigure}[b]{0.23\textwidth}
         \centering
         \captionsetup{justification=centering}
         \includegraphics[width=\textwidth]{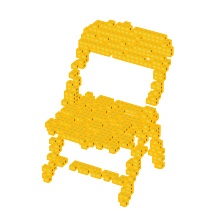}
         \caption{}
         \label{fig:4_sample_voxelized}
     \end{subfigure}
     \vspace{-2mm}
    \caption{An example of a point-cloud sampled from ModelNet10 (a), and a corresponding voxel representation of the same sample (b).
    \vspace{-3mm}}
    \label{fig:4_voxelized}
\end{figure}

\textbf{ModelNet10 Voxelized.} ModelNet10 is a subset of ModelNet40 containing orientation-aligned samples from 10 classes of objects. We voxelize ModelNet10 based on a subsampling of 4096 points from the meshes. First, we normalize these points to fall within the interval $\left[-1, 1\right]$. Next, we bin these points into a grid of $40\times40\times40$ voxels. All nonzero voxels get assigned their location as feature value. Afterwards, to align with our point-cloud configuration, we mask out the all but 512 nonzero voxels. Throughout the network, we only apply convolutions at the locations of each nonzero voxel (effectively masking out activations at all other voxel locations). As for point-clouds, we limit the convolution to integrate over the closest 256 points to limit its computational cost.

\vspace{-7mm}
\section{Experimental details}\label{appx:4_empirical_details}
\vspace{-5mm}
\subsection{General remarks}
\textbf{Code repository and logging.} Our code is written in \texttt{PyTorch}. We utilize \texttt{wandb} \citep{wandb} \texttt{hydra} \citep{Yadan2019Hydra} and \texttt{pytorch-lightning} \citep{falcon2019pytorch} for logging and code structuring. Our experiments are performed on NVIDIA TITAN RTX, A6000 and A100 GPUs, depending on the size of the datasets and inputs considered. Our code is publicly available at \textit{url hidden for the sake of the double blind review process.}

\textbf{Normalized relative positions.} The kernel network $\varphi_{\rm Kernel}$ can, in principle, receive arbitrary coordinates as input. However, considering unitary step-wise relative positions, i.e., 0, 1, 2, ... , $\Nt$, can be problematic from a numerical stability perspective as $\Nt$ may grow very large, e.g., $\Nt{=}$16000 for the Speech Commands dataset. Consequently, based on insights from the Implicit Neural Representations, e.g., \citet{sitzmann2020implicit, fathony2021multiplicative}, we normalize the coordinates such that they lie in the space $[-1, 1]^{\Dt}$ for $\Dt$-dimensional kernels. To this end, we map largest unitary positions seen during training $[0, N]$ to a uniform linear space in $[-1, 1]$. Any relative kernel positions outside of the trained kernel domain that may appear during inference are masked out.
\vspace{-7mm}
\subsection{Hyperparameters and training details}

\textbf{Optimizer and learning rate scheduler.} All our models are optimized with AdamW \citep{loshchilov2018decoupled} in combination with a cosine annealing learning rate scheduler \cite{loshchilov2017sgdr} and a linear learning rate warm-up stage of 10 epochs.

\textbf{Best hyperparameters found.} We perform hyperparameter search on the learning rate, dropout rate, weight decay, and $\omega_0$ of our CCNNs for each task considered.\footnote{$\omega_0$ serves as a prior on the variance of the data that is fitted with several types of implicit neural representations, e.g., SIRENs \cite{sitzmann2020implicit}, MFNs \cite{fathony2021multiplicative}, etc.} The best hyperparameters found are reported in Tables~\ref{tab:4_hyperparams_small} and \ref{tab:4_hyperparams_big}.

\begin{table}
    \centering
    \caption{Best hyperparameters found for CCNN$_{4, 140}$ on all tasks considered.}
    \vspace{-3mm}
    \label{tab:4_hyperparams_small}
    \begin{small}
    \scalebox{0.75}{
    \begin{tabular}{ccccccc}
    \toprule
         & $\boldsymbol{w_0}$ & \textbf{Dropout} & \textbf{Learning Rate} & \textbf{Weight Decay} & \textbf{Batch Size} & \textbf{Epochs}  \\
         \midrule
         \sc{sMNIST} & 2976.49 & 0.1 & 0.01 & 1e-6 & 100 & 210 \\
         \sc{pMNIST} & 2985.63 & 0.2 & 0.02 & 0 & 100 & 210 \\
         \sc{sCIFAR10} & 2386.49 & 0.0 & 0.02 & 0 & 50 & 210 \\
         \midrule
         \sc{Speech Commands (raw)} & 1295.61 & 0.2 & 0.02 & 1e-6 & 20 & 160 \\
         \sc{Speech Commands (mfcc)} & 750.18 & 0.2 & 0.02 & 1e-6 & 100 & 110 \\
         \midrule
         \sc{ListOps} & 784.66 & 0.1 & 0.001 & 1e-6 & 50 & 60\\
         \sc{Text} & 2966.60& 0.2 & 0.001 & 1e-5 & 50 & 60 \\
         \sc{Image} & 4005.15 & 0.2 & 0.01 & 0 & 50 & 210  \\
         \sc{Pathfinder} & 2272.56 & 0.2 & 0.01 & 0 & 100 & 210 \\
         \midrule 
         \sc{CIFAR10} & 1435.77 & 0.1 & 0.02 & 0.0001 & 50 & 210 \\
         \sc{CIFAR100} & 3521.55 & 0.1 & 0.02 & 0.0001 & 50 & 210 \\
         \sc{STL10} & 954.28 & 0.1 & 0.02 & 0 & 64 & 210 \\
         \sc{2DImage} & 2085.43& 0.2 & 0.02 & 1e-6 & 50 & 210  \\
         \sc{2DPathfinder}  & 1239.14 & 0.1 & 0.01 & 0 & 100 & 210 \\
         \midrule
         \sc{ModelNet40} & 50 & 0.0 & 0.02 & 1e-8 & 16 & 200\\
         \bottomrule
    \end{tabular}}
    \end{small}
    \vspace{-2mm}
    \end{table}

\begin{table}
    \caption{Best hyperparameters found for CCNN$_{6, 380}$ on all tasks considered.}
    \label{tab:4_hyperparams_big}
    \vspace{-3mm}
    \begin{small}
    \scalebox{0.75}{
    \begin{tabular}{ccccccc}
    \toprule
         & $\boldsymbol{w_0}$ & \textbf{Dropout} & \textbf{Learning Rate} & \textbf{Weight Decay} & \textbf{Batch Size} & \textbf{Epochs}  \\
         \midrule
         \sc{sMNIST} & 2976.49 & 0.1 & 0.01 & 0 & 100 & 210 \\
         \sc{pMNIST} & 2985.63 & 0.2 & 0.02 & 0 & 100 & 210 \\
         \sc{sCIFAR10} & 4005.15 & 0.25 & 0.01 & 0 & 50 & 210 \\
         \midrule
         \sc{Speech Commands (raw)} & 1295.61 & 0.2 & 0.02 & 1e-6 & 20 & 160 \\
         \sc{Speech Commands (mfcc)} & 750.18 & 0.2 & 0.02 & 1e-6 & 100 & 110 \\
         \midrule
         \sc{ListOps} & 784.66 & 0.25 & 0.001 & 0 & 50 & 60\\
         \sc{Text} & 2966.60& 0.3 & 0.02 & 0 & 50 & 60 \\
         \sc{Image} & 4005.15 & 0.1 & 0.01 & 0 & 50 & 210  \\
         \sc{Pathfinder} & 2272.56 & 0.1 & 0.01 & 1e-6 & 100 & 210 \\
         \midrule 
         \sc{CIFAR10} & 1435.77 & 0.15 & 0.02 & 0 & 50 & 210 \\
         \sc{CIFAR100} & 679.14 & 0.2 & 0.02 & 0 & 50 & 210 \\
         \sc{STL10} & 954.28 & 0.1 & 0.01 & 1e-6 & 64 & 210 \\
         \sc{2DImage} & 2306.08& 0.2 & 0.02 & 0 & 50 & 210  \\
         \sc{2DPathfinder}  & 3908.32 & 0.2 & 0.01 & 0 & 100 & 210 \\
         \midrule
         \sc{ModelNet40} & 50 & 0.0 & 0.02 & 1e-7 & 32 & 200\\
         \bottomrule
    \end{tabular}}
    \end{small}
\end{table}
\textbf{Parameter efficiency experiments on ModelNet40.} To further assess parameter efficiency of the CCNN on ModelNet40, we run a number of experiments with smaller model sizes. Results for the following models are shown in Fig. \ref{fig:4_modelnet40paramsize}: CCNN$_{4, 16}$ with 4 residual blocks, a channel size of 16 and a kernel network hidden size of 8, with 6,545 total parameters. CCNN$_{4, 32}$ with 4 residual blocks, a channel size of 32 and a kernel network hidden size of 8, with 14,401 total parameters. CCNN$_{4, 48}$ with 4 residual blocks, a channel size of 48 and a kernel network hidden size of 8, with 26,353 total parameters. We use a learning rate of $2e^{-2}$, no weight decay, and an $\omega_0$ of 50.

\begin{figure}
    \centering
     \begin{subfigure}{0.88\textwidth}
         \centering
         \captionsetup{justification=centering}
         \includegraphics[width=0.4\textwidth]{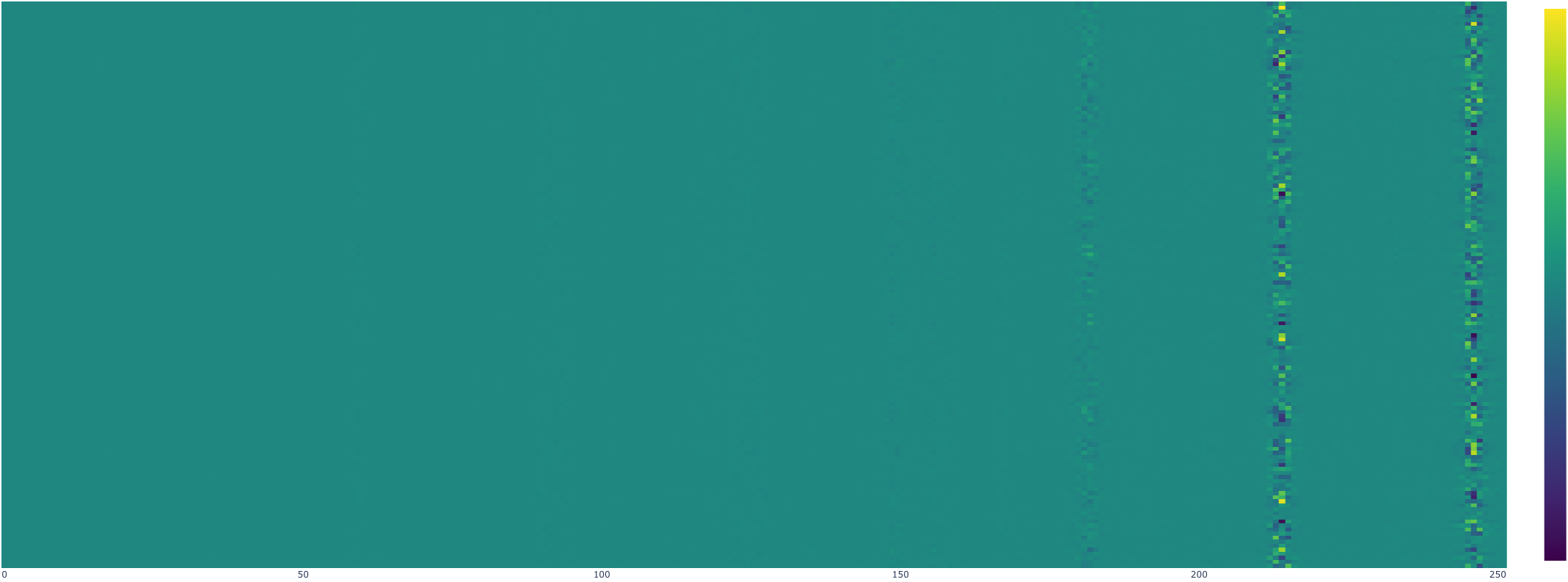}
         \caption{First block.}
     \end{subfigure}
     \begin{subfigure}{0.88\textwidth}
         \centering
         \captionsetup{justification=centering}
         \includegraphics[width=0.6\textwidth]{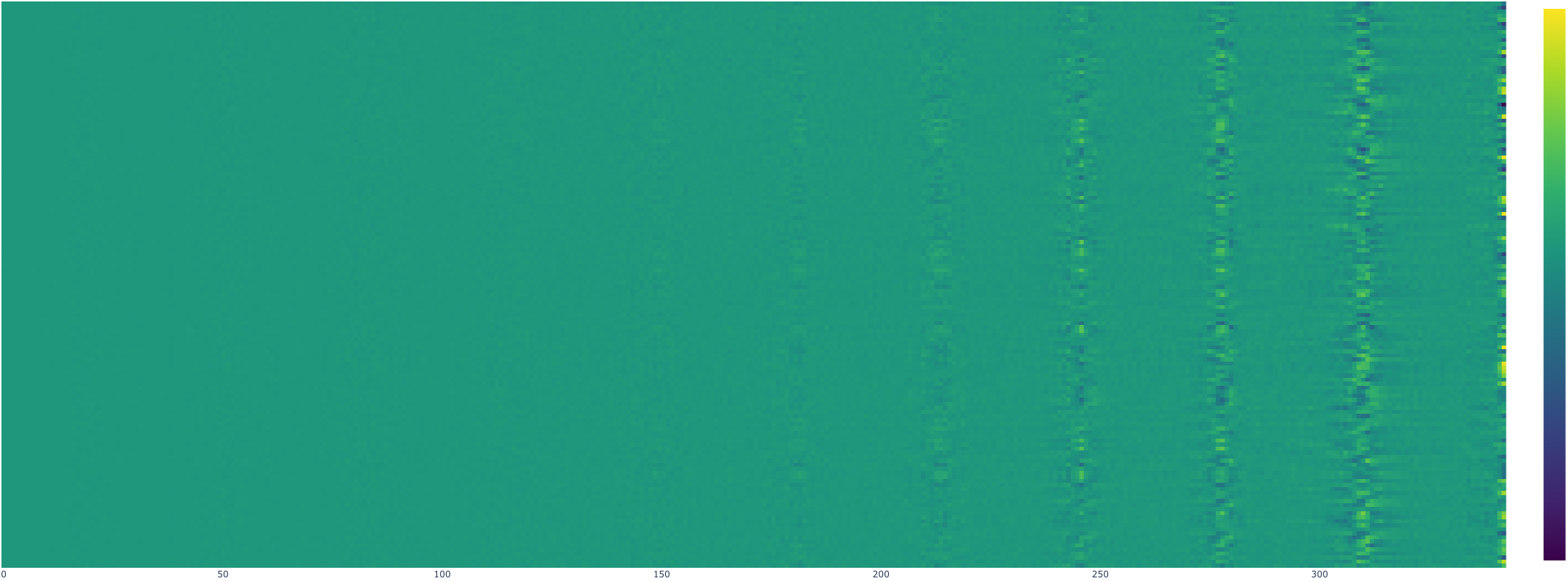}
         \caption{Second block.}
     \end{subfigure}
     \begin{subfigure}{0.88\textwidth}
         \centering
         \captionsetup{justification=centering}
         \includegraphics[width=\textwidth]{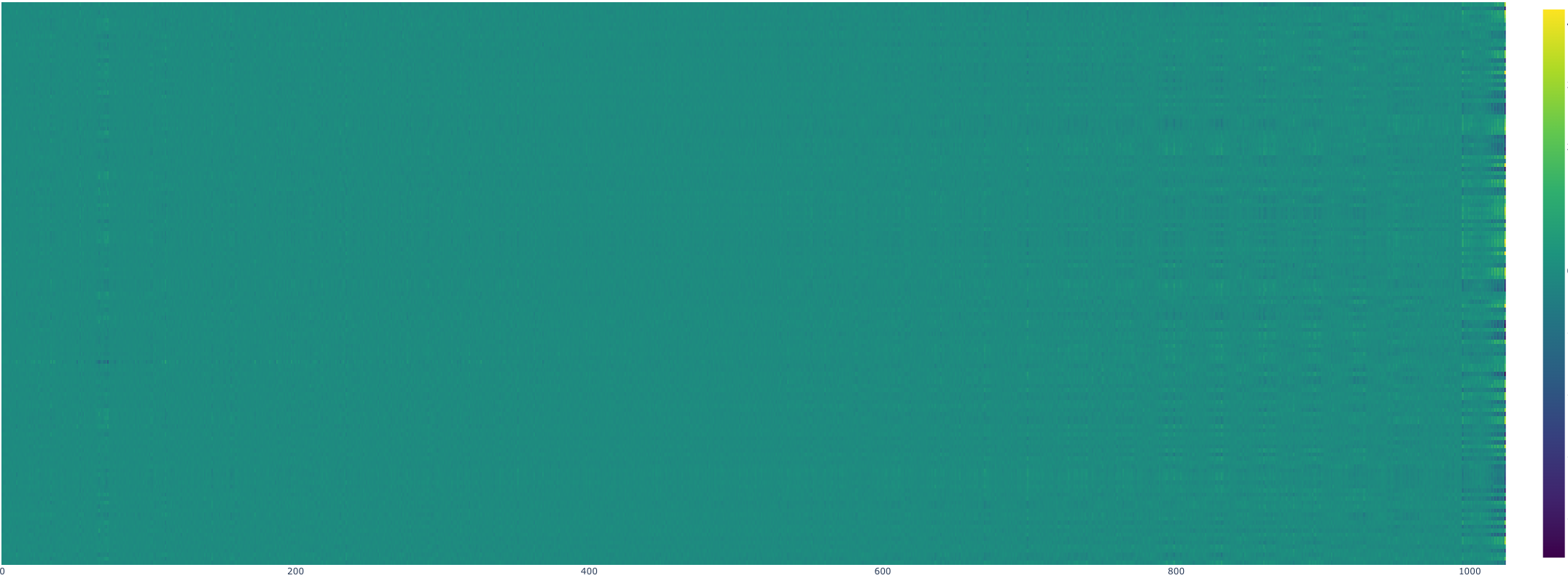}
         \caption{Third block.}
     \end{subfigure}
     \begin{subfigure}{0.88\textwidth}
         \centering
         \captionsetup{justification=centering}
         \includegraphics[width=\textwidth]{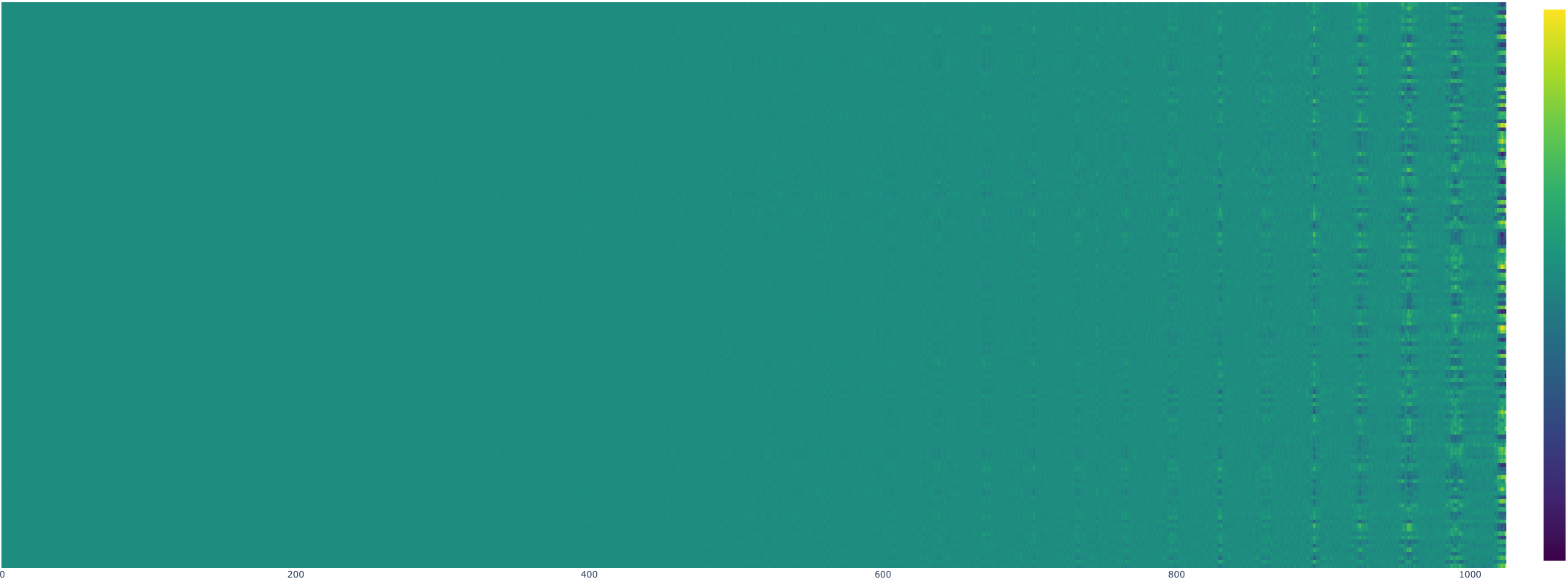}
         \caption{Fourth block.}
     \end{subfigure}
     \vspace{-4mm}
    \caption{Learned (causal) convolutional kernels in a CCNN$_{4, 140}$ trained on sequential CIFAR10. The vertical axis indexes the channels, and the horizontal axis the length of the kernels in a single layer. Interstingly, we observe a clear periodic pattern of 32 steps along the spatial dimension across all layers --a period that corresponds to the width of the underlying $32{\times}32$ CIFAR10 images--. This illustrates that CCNNs in fact learn to represent 2D structures on the flattened 1D space on which sequential CIFAR10 is defined. Despite these important capabilities, it is important to note that modelling $N$D signals as flattened $1$D signals poses a unnecessary burden to the model, and modelling the signal in the original 2D space leads to faster convergence and better accuracy (Fig.~\ref{fig:4_1dvs2d}).\label{fig:4_learned_kernels}}
\end{figure}

\vspace{-7mm}
\section{Additional Experiments and Details}\label{sec:4_additiona-exp}
\vspace{-5mm}
\subsection{Computational Efficiency}
\label{appx:4_inferencetime}
\vspace{-5mm}
\subsubsection{Efficiency of continuous convolutional kernels}
\textbf{Experimental setup}. We experimentally asses the computational complexity of our model. We measure the time of a single forward/backward pass for the two architectures used in our experiments: CCNN$_{4, 140}$ (4 blocks, 140 channels) and CCNN$_{6, 380}$ (6 blocks, 380 channels). It is important to note that we assume the "worst-case" scenario for CCNNs in which the learned kernel size equals the input length, i.e., the model performs convolutions with global kernels at each layer. FlexConvs \citep{romero2022flexconv} learn the kernel size during training and --as shown exemplarily in Fig~\ref{fig:4_learned_kernels}-- having global kernels at each layer is not something that we observe in practice. 

We compare to (\textit{i}) discrete convolutional networks which match in architecture to ours, with global depthwise separable convolutional kernels; Global CNN$_{4, 140}$, and Global CNN$_{6, 380}$, (\textit{ii}) more traditional CNN architectures with downsampling after each block (pooling of size 2), local depthwise separable kernels (kernels of size 5), and a depth tuned to the input size such that the receptive field of the final residual block covers the entire input. We take ${\rm depth} = \lceil {\rm d}\rceil$ with ${\rm d}$ given by:
\begin{equation}
    \frac{\mathrm{input\ length}}{{\rm pooling\ size}^{\mathrm{d}}} \geq \mathrm{kernel\ size}.\nonumber
\end{equation}
These architectures are denoted Local CNN$_{4, 140}$ and Local CNN$_{6, 380}$. For input lengths of 1024, 4096, 8192 and 16000 this results in depths of 8, 10, 11 and 12 blocks respectively.

We measure training time (forward and backward pass) for sequence data of different lengths between 1024 --the length of sCIFAR10--, and 16000 --the length of Speech-Raw--, the largest sequence length in our experiments. We measure the time of training over 1000 batches of size 8 on a RTX 3090.

\textbf{Results}. Results are summarized in Tab \ref{tab:4_inference}. Note that for the Global CNN$_{6, 380}$, we were unable to complete the experiment on data of lengths exceeding 8192 because of insufficient memory during the backward pass. This highlights another benefit of the continuous convolutional layer: its relative memory efficiency in computing weight updates, which results from the fact that gradients for individual kernel values can be discarded. As only the kernel network weights need to be updated, gradient values for the kernel weights do not need to be stored during the backward pass.

Our results show that our approach requires additional computation compared to more traditional discrete CNNs. We note that although the use of a kernel network to infer kernel values brings computational overhead, this overhead is not a limiting factor in practice. The global convolutional kernels used in our network architectures have an impact on computational efficiency, but the use of convolutions in the Fourier domain help address this issue.

\begin{table}
    \begin{center}
    \caption{Training time (ms). \textit{\footnotesize $\times$ - unavailable due to memory issues. $n$ - $\#$ blocks tuned per input length.}}
    \label{tab:4_inference}
    \vspace{-3mm}
    \scalebox{0.8}{
    \begin{tabular}{ccccc}
    \toprule
    \sc{Size} & 1024 & 4096 & 8192 & 16000 \\\midrule
    Local CNN$_{n, 140}$ & 24 & 32 & 43 &  65 \\
    Global CNN$_{4, 140}$ & 13 & 25 & 53 & 122 \\
    CCNN$_{4, 140}$& 52 & 56 & 77 & 145 \\
    \midrule
    Local CNN$_{n, 380}$ & 24 & 38 & 93 & 178 \\
    Global CNN$_{6, 380}$ & 31 & 102 & 220 & $\times$ \\
    CCNN$_{6, 380}$ & 71 & 136 & 255 & 537\\
    \bottomrule
    \end{tabular}
    }
\end{center}
\vspace{-2mm}
\end{table}

\vspace{-7mm}
\subsubsection{Efficiency of CCNNs with regard to other global methods}

\textbf{Experimental Setup.} Next, we benchmark the computational efficiency of CCNNs compared to other input-global methods: S4 \citep{gu2022efficiently} and LSSL \citep{gu2021combining}. 

We reproduce the efficiency benchmark in \cite{gu2022efficiently}. That is, we provide the runtime for a single layer of forward and backward pass for different hidden dimensionalities, measured for different input lengths. The kernel network used in these experiments is a MAGNet with 3 layers and a hidden dimensionality of 32. Again, we assume a "worst-case" setting in which the learned kernel size equals the input length, i.e. the layer performs convolutions with global kernels. Note that in this setting, we perform convolutions in the spatial domain. We average runtime over 10000 steps. Experiments are performed on a RTX 3090.

\textbf{Results}. Results are summarized in Tab. \ref{tab:4_time-vs-s4}. We show that, up to a sequence length of 16000, runtimes in comparison with S4 are marginally faster, but remain in the same order of magnitude. Differences in memory utilization become more pronounced for increases in hidden dimensionality and sequence length, but remain a constant factor. Next, note that in both memory and speed, we show that a CCNN layer is asymptotically more efficient than LSSL. For LSSL, only results present in \cite{gu2022efficiently} are shown, as we were unable to successfully reproduce the LSSL layer performance.

% \begin{table}[]
% \begin{minipage}{\textwidth}
%     \begin{center}
%     \caption{Forward and backward for a single layer of CCNN in comparison with S4 \citep{gu2022efficiently} and LSSL \citep{gu2021combining} in speed  (ms) and memory usage (Mb).}
%     \label{tab:4_time-vs-s4}
%     \begin{tabular}{cccccccc}
%     \toprule
%     &\multicolumn{3}{c}{\sc{Training Step (ms)}} & \multicolumn{3}{c}{\sc{Memory Alloc (Mb)}} \\
%     \cmidrule(lr){2-4}\cmidrule(lr){5-7}
%     \sc{Hidden Size} & 256 & 512 & 1024 & 256 & 512 & 1024 \\\midrule
%     LSSL & 9.32 & 20.60 & 140.70 & 222.1 & 1685 & 13140 \\
%     S4 & 4.77 & 3.07 & 4.75 & 5.3 & 12.6 & 33.5 \\
%     CCNN & 5.21 & 5.29 & 5.23 & 10.3 & 15.5 & 32.6 \\
%     \bottomrule
%     \end{tabular}
% \end{center}
% \end{minipage}
% \end{table}

\begin{table}
    \begin{center}
    \caption{Forward and backward for a single layer of CCNN in comparison with S4 \citep{gu2022efficiently} and LSSL \citep{gu2021combining} in speed (ms). \textit{\footnotesize $\times$ - result not available.}}
    \vspace{-3mm}
    \label{tab:4_time-vs-s4}
    \scalebox{0.8}{
    \begin{tabular}{ccccccccccccc}
    \toprule
    \sc{Input Length} & \multicolumn{3}{c}{128} & \multicolumn{3}{c}{1024} & \multicolumn{3}{c}{4096} & \multicolumn{3}{c}{16000}  \\
    \cmidrule(lr){2-4}\cmidrule(lr){5-7}\cmidrule(lr){8-10}\cmidrule(lr){11-13}
    \sc{Hidden Size} & 256 & 512 & 1024 & 256 & 512 & 1024 & 256 & 512 & 1024 & 256 & 512 & 1024 \\\midrule
    LSSL & 9.32 & 20.60 & 140.70 & $\times$ & $\times$& $\times$& $\times$& $\times$& $\times$ & $\times$& $\times$& $\times$  \\
    S4 & 4.16 & \textbf{3.41} & 4.16 & 4.11 & 6.39 & 9.06 & 6.86 & 10.57 & 25.20 & \textbf{18.10} & \textbf{35.72} & \textbf{96.64}  \\
    CCNN & \textbf{3.99} & 3.71 & \textbf{4.12} & \textbf{3.90} & \textbf{5.07} & \textbf{7.25} & \textbf{5.91} & \textbf{9.67} & \textbf{22.39} & 20.64 & 42.49 & 102.57  \\
    \bottomrule
    \end{tabular}}
\end{center}
\end{table}

\begin{table}
    \begin{center}
    \caption{Forward and backward for a single layer of CCNN in comparison with S4 \citep{gu2022efficiently} and LSSL \citep{gu2021combining} memory usage (Mb). \textit{\footnotesize $\times$ - result not available.}}
    \label{tab:4_memory-vs-s4}
    \vspace{-3mm}
    \scalebox{0.7}{
    \begin{tabular}{ccccccccccccc}
    \toprule
    \sc{Input Length} & \multicolumn{3}{c}{128} & \multicolumn{3}{c}{1024} & \multicolumn{3}{c}{4096} & \multicolumn{3}{c}{16000}  \\
    \cmidrule(lr){2-4}\cmidrule(lr){5-7}\cmidrule(lr){8-10}\cmidrule(lr){11-13}
    \sc{Hidden Size} & 256 & 512 & 1024 & 256 & 512 & 1024 & 256 & 512 & 1024 & 256 & 512 & 1024 \\\midrule
    LSSL & 222.1 & 1685 & 13140 & $\times$ & $\times$& $\times$& $\times$& $\times$& $\times$ & $\times$& $\times$& $\times$  \\
    S4 & \textbf{5.6} & \textbf{14.46} & \textbf{42.64} & \textbf{33.65} & \textbf{70.47} & \textbf{153.69} & \textbf{129.69} & \textbf{262.51} & \textbf{537.72} & \textbf{508.93} & \textbf{1014.08} & \textbf{2033.82} \\
    CCNN & 14.46 & 28.94 & 60.38 & 113.13 & 222.97 & 444.17 & 451.39 & 888.24 & 1763.43 & 1764.46 & 3467.34 & 6878.26 \\
    \bottomrule
    \end{tabular}}
\end{center}
\end{table}

% lengths 1024, 4096, 8192, 16000
\vspace{-7mm}
\subsection{Ablation: Performance of FlexCNN, S4 and CCNN Residual Blocks}
\label{appx:4_residual-block-ablation}
\textbf{Experimental setup}. We provide an ablation over the impact on performance of the proposed residual CCNNBlock compared to the FlexCNNBlock used in the architectures in \cite{romero2022flexconv} and the S4Block used in \cite{gu2022efficiently} (Fig. \ref{fig:4_block_comparison}).

To this end, we run experiments on a select number of datasets covering sequence and image data with a version of our architecture that includes the FlexCNNBlock (F-CCNN$_{4, 140}$ with 4 blocks, 140 channels, 233{\sc{k}} params and F-CCNN$_{6, 380}$ with 6 blocks, 380 channels, 2.24{\sc{m}} params), and a version of our architecture that includes the S4Block (S4-CCNN$_{4, 140}$ with 4 blocks, 140 channels, 200{\sc{k}} params and S4-CCNN$_{6, 380}$ with 6 blocks, 380 channels, 2{\sc{m}} params). We compare performance against the two architectures used throughout the experiments in Sec. \ref{sec:4_experiments}, which uses the CCNNBlock (CCNN$_{4, 140}$ with 4 blocks, 140 channels, 200{\sc{k}} params and CCNN$_{6, 380}$ with 6 blocks, 380 channels, 2{\sc{m}} params). To isolate the impact of the residual block architecture, we replace the FlexConv layers in the original formulation of the FlexCNNBlock with our proposed depthwise separable implementation SepFlexConv. Note that we do not parameter-match these models, which results in the architectures with FlexCNNBlocks having more parameters compared to ones with the CCNNBlock, due to the FlexCNNBlock having two convolutional layers instead of the one convolutional layer and one pointwise linear layer of the CCNNBlock. Training regimes and hyperparameters for all architectures are as per Sec. \ref{appx:4_empirical_details}.

\textbf{Results}. Results are summarized in table \ref{tab:4_blockablation}. First, note that without the nonlinearity added add the end of the block, the S4Block performs poorly over all tasks. Next, as shown, the CCNNBlock formulation improves performance of the architecture compared to FlexCNNBlock, in some cases by a significant margin. Note that architectures with the S4Block and CCNNBlock contain fewer convolutional layers and parameters compared to architecturally matched ones with FlexCNNBlocks, indicating higher parameter-efficiency as well as computational efficiency of CCNNBlocks (e.g. 8h 55m runtime for F-CCNN$_{6,380}$ vs. 8h 12m runtime for CCNN$_{6,380}$ on sCIFAR10). Because of the shown performance benefits, we use the CCNNBlock throughout our experiments.%, also greatly improving computational efficiency ()

\begin{table}
    \begin{center}
    \caption{Test accuracy of models with FlexCNNBlock (F-CCNN$_{4, 140}$, F-CCNN$_{6, 380}$), models with S4Block (S4-CCNN$_{4, 140}$, S4-CCNN$_{6, 380}$) and models with CCNNBlock (CCNN$_{4, 140}$, CCNN$_{6, 380}$).}
    \label{tab:4_blockablation}
    \vspace{-3mm}
    \scalebox{0.8}{
    \begin{tabular}{cccc}
    \toprule
     & \sc{SpeechCommands-MFCC} & \sc{sCIFAR10} & \sc{CIFAR10} \\\midrule
    F-CCNN$_{4, 140}$ & 94.38 & 85.54 & 86.34 \\
    S4-CCNN$_{4, 140}$ & 73.60 & 53.25  & 60.06 \\
    CCNN$_{4, 140}$ & \textbf{95.01} & \textbf{90.30} & \textbf{92.78} \\
    \midrule
    F-CCNN$_{6, 380}$ & 94.92 & 84.68 & 92.24 \\
    S4-CCNN$_{6, 380}$ & 60.82 & 51.50 & 67.01 \\
    CCNN$_{6, 380}$ & \textbf{97.98} & \textbf{93.08} & \textbf{95.20}\\
    \bottomrule
    \end{tabular}
    }
\end{center}
\end{table}

%% file: backmatter/appx_X_gridifier.tex
% path to figures directory
\graphicspath{{figures/X-gridifier/}}

%=========================================================================

% \begin{savequote}[75mm]
% Nulla facilisi. In vel sem. Morbi id urna in diam dignissim feugiat. Proin molestie tortor eu velit. Aliquam erat volutpat. Nullam    ultrices, diam tempus vulputate egestas, eros pede varius leo.
% \qauthor{Quoteauthor Lastname}
% \end{savequote}

\chapter{Learnable Gridification for Efficient Point-Cloud Processing}

\vspace{-7mm}
\section{Network structure and convolution blocks}
\label{app:networkarch}
\begin{figure}
  \centering
  \includegraphics[width=0.7\linewidth]{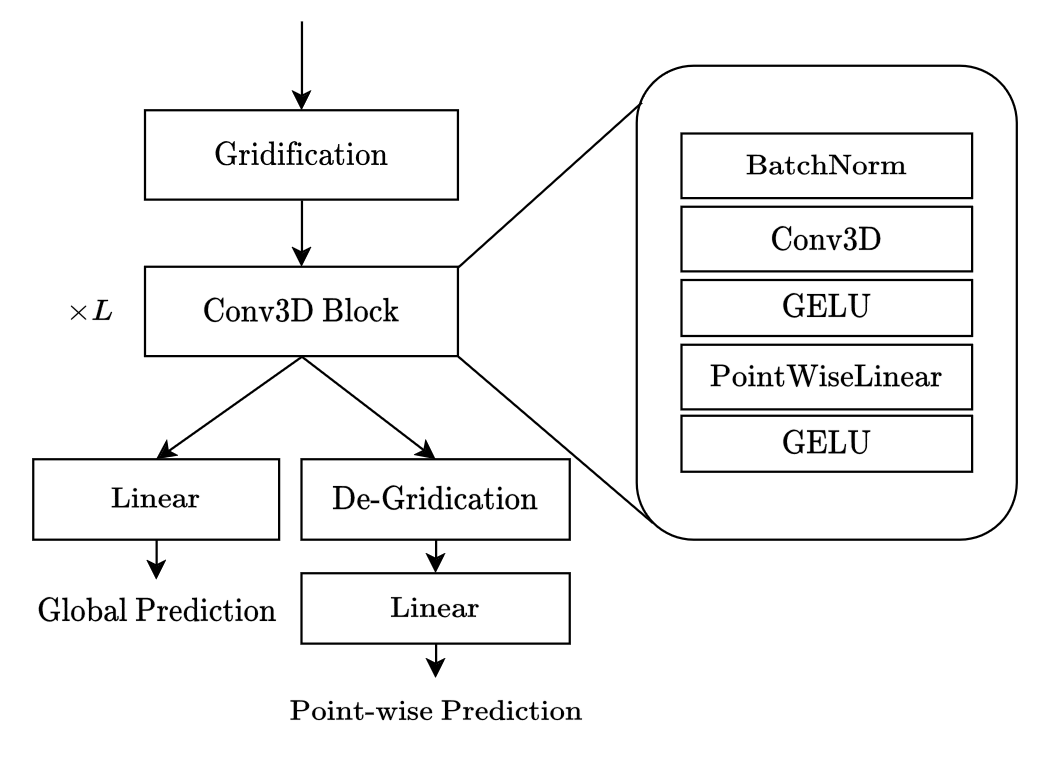}
   \vspace{-2mm}
   \captionsetup{justification=centering}
  \caption{Practical pipeline and convolution blocks.}
  \label{fig:networkarch}
\end{figure}

Fig. \ref{fig:networkarch} shows how we instantiated the grid network as described in section \ref{section:gridnetworks} in practice. The \texttt{Conv3D} blocks are CCNN blocks as in \citet{knigge2023modelling}.

\vspace{-7mm}
\section{Hyperparameters}
\label{app:hyperparams}
Table \ref{tab:hyperparams} contains the best hyperparameters for the specific datasets found through hyperparameter sweeps.

\begin{table}
\centering
    \begin{minipage}{\textwidth}
    \centering
    \captionsetup{justification=centering}
    \caption{Hyperparameter settings for the different datasets.}
    \label{tab:hyperparams}
    \vspace{-2.5mm}
    \begin{small}
    \scalebox{0.85}{
    \begin{tabular}{llll}
    \toprule
     & ModelNet40 & ShapeNet\\
     \midrule
batch size & 32 & 16 \\
nr. conv blocks & 3  &  6  \\
hidden channels & 128  & 256  \\
nr. epochs & 60 &  50  \\
nr. input points & 1000 & 2047 \\ 
$\Omega$ position embedding & 0.1  & 1.0 \\
optimizer & AdamW &  AdamW   \\
learning rate & 0.005  & 0.001  \\
learning rate scheduler & Cosine Annealing &  Cosine Annealing\\
learning rate warmup & 10  &  10 \\
nr. neighbors & 9  & 9 \\
grid resolution & 9 & 13 \\
conv. kernel size &9  & 9\\
dropout & 0.1 & 0.3\\
weight decay & 0 & 0.001\\
aggregation & mean & max \\
    \bottomrule
    \end{tabular}}
    \end{small}
    \end{minipage}%
    \vspace{-4mm}
\end{table}

%% file: backmatter/appx_3_flexconv.tex
% path to figures directory
\graphicspath{{figures/3-flexconv/}}

%=========================================================================

% \begin{savequote}[75mm]
% Nulla facilisi. In vel sem. Morbi id urna in diam dignissim feugiat. Proin molestie tortor eu velit. Aliquam erat volutpat. Nullam    ultrices, diam tempus vulputate egestas, eros pede varius leo.
% \qauthor{Quoteauthor Lastname}
% \end{savequote}
\vspace{-7mm}
\chapter{Continuous Kernel Convolutions with Differentiable Kernel Sizes} \label{appx:3_flexconv}

  %=========================================================================

% \begin{wrapfigure}{r}{5.7cm}
% %\vspace{-3.5mm}
%         \centering
%      \begin{subfigure}[c]{0.18\textwidth}
%          \centering
%          \includegraphics[width=\textwidth]{images/gauss_mask.png}
%          \caption{Gaussian}
%          \label{fig:3_gauss_mask}
%      \end{subfigure}
%      \hspace{2mm}
%      \begin{subfigure}[c]{0.18\textwidth}
%          \centering
%          \includegraphics[width=\textwidth]{images/sigmoid_mask.png}
%          \caption{Squared sigmoid}
%          \label{fig:3_sigm_mask}
%      \end{subfigure}
%      \vspace{-1.5mm}
%     \caption{Differentiable masks.}
%     \vspace{-5mm}
%     \label{fig:3_masks}
% \end{wrapfigure}

\begin{figure}
    \centering
     \begin{subfigure}[c]{0.24\textwidth}
    %  \vspace{2.5mm}
         \centering
         \captionsetup{justification=centering}
         \includegraphics[width=\textwidth]{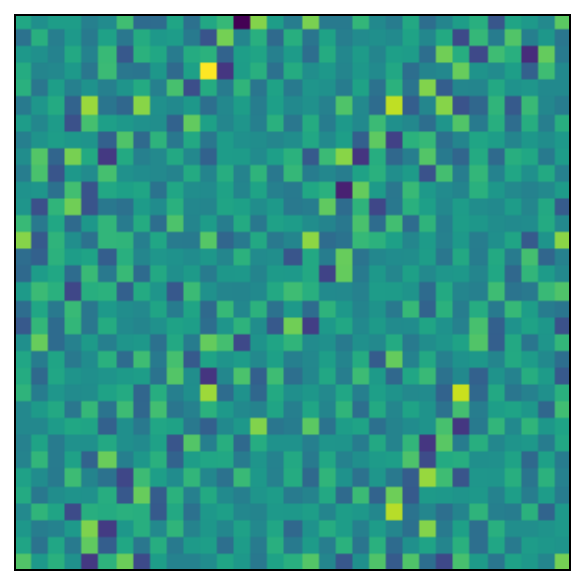}
         \caption{$\texttt{MLP}^{\boldsymbol{\psi}}$ output}
         %([-\mathrm{k},\mathrm{k}]^2)$}
         \label{fig:3_flexconvexample-raw}
     \end{subfigure}
    %  \hfill
     \begin{subfigure}[c]{0.24\textwidth}
        %  \vspace{2.5mm}
         \centering
         \includegraphics[width=\textwidth]{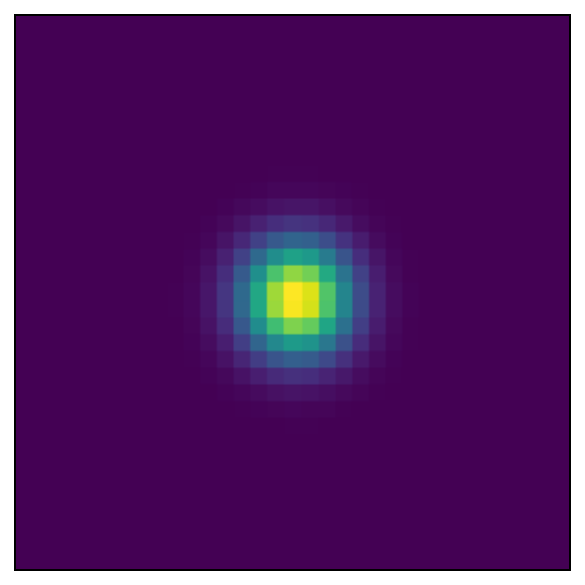}
        %  \vskip -0.7cm
        \captionsetup{justification=centering}
         \caption{$\gv\big( [x,y] ; \boldsymbol{\theta}^{(l)}\big)$ %(Eq.~\ref{eq:3_anisotropicgaussian})
         }
         \label{fig:3_flexconvexample-mask}
     \end{subfigure}
     \begin{subfigure}[c]{0.24\textwidth}
        %  \vspace{2.5mm}
         \centering
         \captionsetup{justification=centering}
         \includegraphics[width=\textwidth]{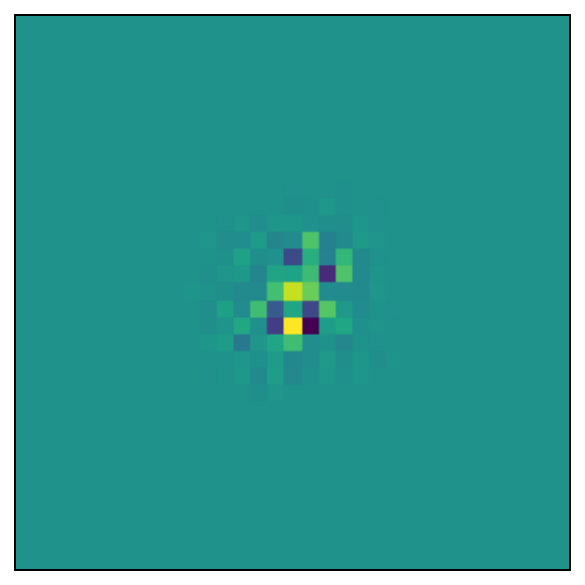}
        %  \vskip -0.7cm
         \caption{
         $\boldsymbol{\psi}(x, y) = $ (a) $\cdot$ (b)}
         \label{fig:3_flexconvexample-maskedkernel}
     \end{subfigure}
     \begin{subfigure}[c]{0.24\textwidth}
        %  \vspace{2.5mm}
         \centering
         \captionsetup{justification=centering}
         \includegraphics[width=\textwidth]{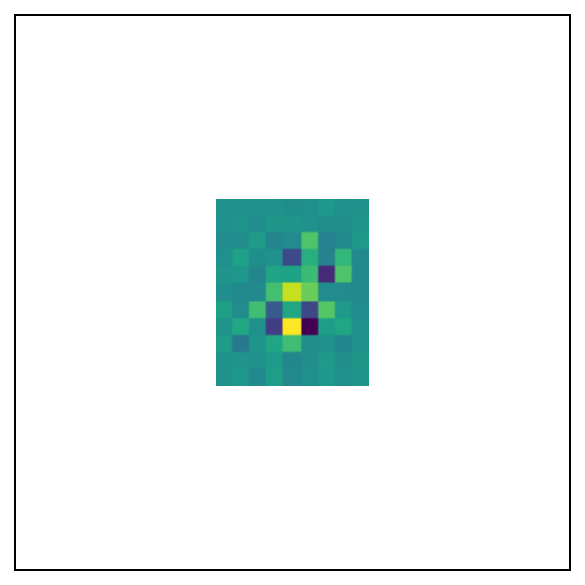}
        %  \vskip -0.7cm
         \caption{After cropping}
         \label{fig:3_flexconvexample-kernel}
     \end{subfigure}
    \caption{Example kernels, generated step by step. FlexConv samples a kernel from  $\texttt{MLP}^{\boldsymbol{\psi}}$ (a), which is attenuated by an anistropic Gaussian envelope with learned parameters $\boldsymbol{\theta}^{(l)}$ (b), creating (c) which is cropped to contain only values of $> 0.1$ (d).
    \vspace{-2mm}}
    \label{fig:3_app-flexconvexample}
\end{figure}

% \begin{figure}
%     \centering
%      \begin{subfigure}[c]{0.98\textwidth}
%     %  \vspace{2.5mm}
%          \centering
%          \includegraphics[width=\textwidth]{images/net_kernels.pdf}
%          %([-\mathrm{k},\mathrm{k}]^2)$}
%          \label{fig:3_flexconvexample-raw}
%      \end{subfigure}
%     \caption{Kernels learned by FlexConvs throughout a CIFAR-10 FlexNet (horizontal), for three different seeds (vertical).
%     \vspace{-2mm}}
%     \label{fig:3_app-learnedkernelsizes}
% \end{figure}

\begin{figure}
    \centering
    % \begin{subfigure}[c]{0.98\textwidth}
    %  \vspace{2.5mm}
    %      \centering
    %      \includegraphics[width=\textwidth]{images/c10-cres-16x32-noreg-kernels-1.pdf}
    %      \label{fig:3_c10-cres-gabor-kernels-1}
    %      \caption{No regularization, block 1 of 7.}
    %  \end{subfigure}
     \begin{subfigure}[c]{0.98\textwidth}
     \vspace{2.5mm}
         \centering
         \captionsetup{justification=centering}
         \includegraphics[width=\textwidth]{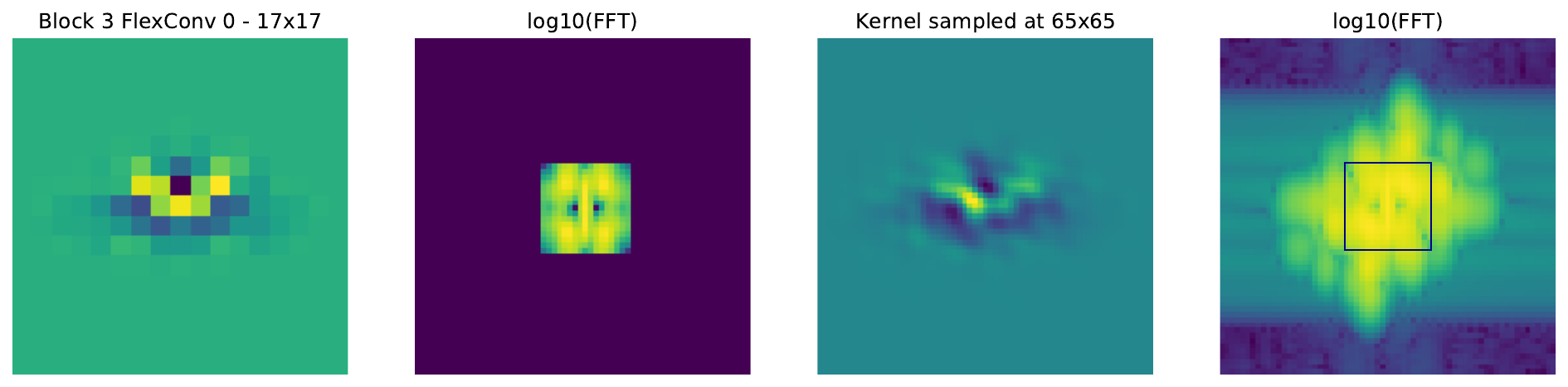}
         %([-\mathrm{k},\mathrm{k}]^2)$}
         \label{fig:3_c10-cres-noreg-kernels-1}
         \vspace{-3mm}
         \caption{No regularization, block 3 of 7.}
     \end{subfigure}
    %  \begin{subfigure}[c]{0.98\textwidth}
    %  \vspace{2.5mm}
    %      \centering
    %      \includegraphics[width=\textwidth]{images/c10-cres-16x32-gabortog-kernels-1.pdf}
    %      %([-\mathrm{k},\mathrm{k}]^2)$}
    %      \label{fig:3_c10-cres-gabor-kernels-1}
    %      \caption{Regularizing $f^+_{\textrm{MAGNet}}$, block 1 of 7}
    %  \end{subfigure}
     \begin{subfigure}[c]{0.98\textwidth}
     \vspace{2.5mm}
         \centering
         \captionsetup{justification=centering}
         \includegraphics[width=\textwidth]{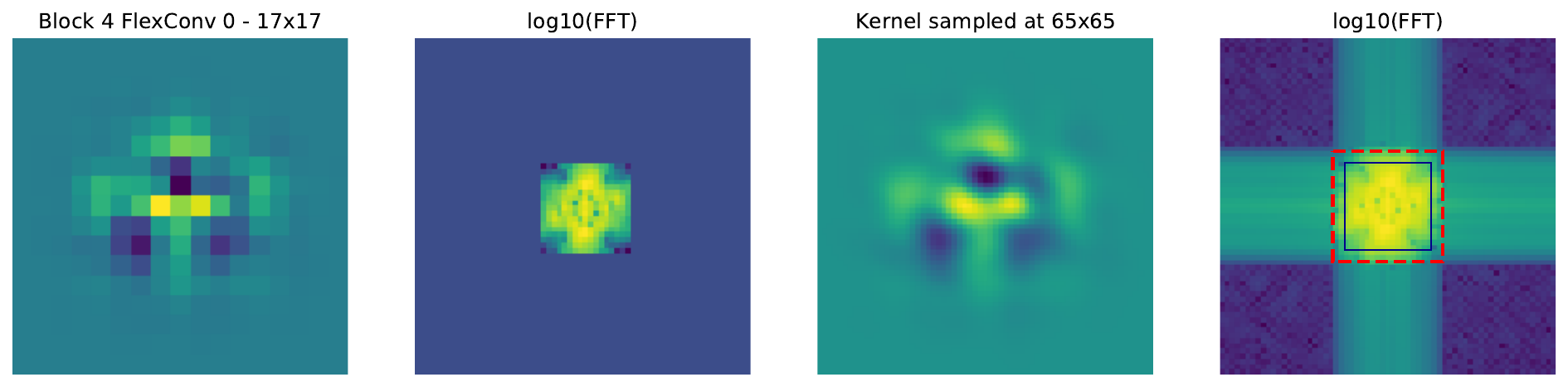}
         %([-\mathrm{k},\mathrm{k}]^2)$}
         \label{fig:3_c10-cres-gabor-kernels-1}
         \vspace{-3mm}
         \caption{Regularizing $f^+_{\textrm{MAGNet}}$, block 4 of 7.}
     \end{subfigure}
    %  \begin{subfigure}[c]{0.98\textwidth}
    %  \vspace{2.5mm}
    %      \centering
    %      \includegraphics[width=\textwidth]{images/c10-cres-16x32-gnmtog-kernels-1.pdf}
    %      %([-\mathrm{k},\mathrm{k}]^2)$}
    %      \label{fig:3_c10-cres-gabor-kernels-1}
    %      \caption{Regularizing $f^+_{\textrm{FlexConv}}$, block 1 of 7}
    %  \end{subfigure}
     \begin{subfigure}[c]{0.98\textwidth}
     \vspace{2.5mm}
         \centering
         \captionsetup{justification=centering}
         \includegraphics[width=\textwidth]{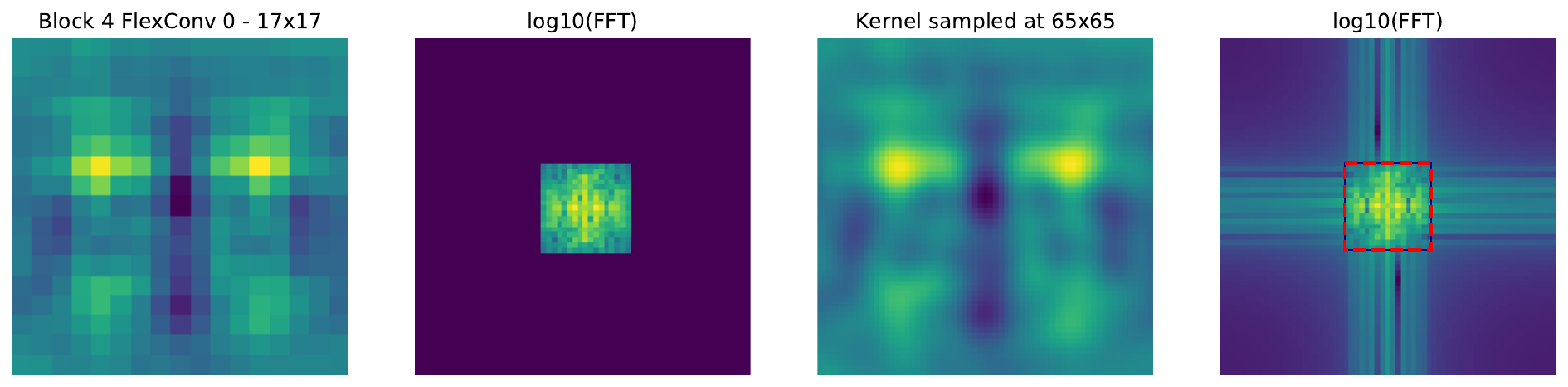}
         \label{fig:3_c10-cres-gabor-kernels-1}
         \vspace{-3mm}
         \caption{Regularizing $f^+_{\textrm{FlexConv}}$, block 4 of 7.}
     \end{subfigure}
    \caption{Example kernels from FlexNet-16 models trained (i) without regularization, (ii) with aliasing regularization of $f^+_{\textrm{MAGNet}}$, (iii) with aliasing regularization of $f^+_{\textrm{FlexConv}}$. In the columns, from left to right: (i) original kernel at $33 \times 33$, (ii) FFT of the original kernel, (iii) kernel inferred at $65 \times 65$, to find aliasing effects, (iiii) FFT of the $65 \times 65$ kernel, with the solid line showing the Nyquist frequency of the $33 \times 33$ kernel, and the red dotted line showing the maximum frequency component as computed by our analysis. For $f^+_{\textrm{FlexConv}}$ the maximum frequency matches almost exactly with the Nyquist frequency, showing that our aliasing regularization works. For $f^+_{\textrm{MAGNet}}$, the maximum frequency is slightly higher than the Nyquist frequency, as the FlexConv mask is not included in the frequency term derivation. This is reflected in the slightly worse resolution generalization results reported in Sec.~\ref{sec:3_crossresexperiments}. Furthermore, some aliasing effects are still apparent for the aliasing regularized models, as discussed in Sec.~\ref{sec:3_limitations}.
    \vspace{-2mm}}
    \label{fig:3_app-cifar10kernelfrequencies}
\end{figure}

% \section{High-frequency artefacts in unregularized FlexConvs}
% \label{sec:3_hf-artefacts}

% \begin{figure}
%     \centering
%      \begin{subfigure}[c]{0.98\hf-artefactstextwidth}
%     %  \vspace{2.5mm}
%          \centering
%          \includegraphics[width=\textwidth]{images/kernels_16x16.pdf}
%          \caption{$16 \times 16$}
%          \label{fig:3_hf-artefacts-16}
%      \end{subfigure}
%     %  \hfill
%      \begin{subfigure}[c]{0.98\textwidth}
%          \vspace{2.5mm}
%          \centering
%          \includegraphics[width=\textwidth]{images/kernels_32x32.pdf}
%         %  \vskip -0.7cm
%          \caption{$16 \times 16 \to 32 \times 32$}
%          \label{fig:3_hf-artefacts-32}
%      \end{subfigure}
%     \caption{Kernels from a FlexNet trained at $16 \times 16$ data resolution (Fig.~\ref{fig:3_hf-artefacts-16}) show high-frequency artefacts when upsampled to $32 \times 32$ (Fig.~\ref{fig:3_hf-artefacts-32}).
%     \vspace{-2mm}}
%     \label{fig:3_hf-artefacts}
% \end{figure}

% To verify the existence of high-frequency artefacts, we show the original ($17 \times 17$) and upsampled ($67 \times 67$) kernels of a FlexNets trained on a version of CIFAR-10 that was downsampled to resolution $16 \times 16$ in Fig.~\ref{fig:3_hf-artefacts}. Indeed, the upsampled kernel shows high frequency components that do not match the kernel learned during training.

\vspace{-7mm}
\section{Alias-free FlexConv regularization}
\label{sec:3_regularizingflexconv}
In this section we provide the complete derivation and analysis for our FlexConv regularization against aliasing. First, we derive the analytic maximum frequency component of a FlexConv kernel. Next, we compute the Nyquist frequency of a FlexConv kernel, and subsequently show how to combine the previous results into a regularization term to train alias-free FlexConvs.

\vspace{-7mm}
\subsection{Analyzing the frequency spectrum of FlexConv}
\label{sec:3_magnetanalysis}

In order to make FlexConv alias-free (Sec.~\ref{sec:3_crtraining}), we need to compute the maximum frequency component of the kernels generated by a MAGNet, so that we can regularize it during training. In this section we analytically derive this maximum frequency component from the parameters of the MAGNet.

Recall that MAGNets generate a kernel $\boldsymbol{\psi}(x,y)$ through of a succession of anisotropic Gabor filters and linear layers (Sec.~\ref{sec:3_magnets}, Eqs.~\ref{eq:3_mfn}--\ref{eq:3_anisotropicgaussian}):
\begin{align*}
    &\boldsymbol{\mathrm{h}}^{(1)} = \boldsymbol{\mathrm{g}}\big( [x,y]; \boldsymbol{\theta}^{(1)}\big) && \boldsymbol{\mathrm{g}}: \sR^{2} \rightarrow \sR^{\Nt_{\mathrm{hid}}}\\
    &\boldsymbol{\mathrm{h}}^{(l)} = \big(\Wm^{(l)} \boldsymbol{\mathrm{h}}^{(l-1)} + \boldsymbol{\mathrm{b}}^{(l)}\big) \cdot   \boldsymbol{\mathrm{g}}\big( [x,y] ; \boldsymbol{\theta}^{(l)}\big) && \Wm^{(l)} \in \sR^{\Nt_{\mathrm{hid}} \times \Nt_{\mathrm{hid}}}, \boldsymbol{\mathrm{b}}^{(l)} \in \sR^{\Nt_{\mathrm{hid}}} \quad \\
    &\boldsymbol{\psi}(x, y) = \Wm^{(\mathrm{L})} \boldsymbol{\mathrm{h}}^{(\mathrm{L}-1)} + \boldsymbol{\mathrm{b}}^{(\mathrm{L})} && \Wm^{(\mathrm{L})} \in \sR^{(\Nt_{\mathrm{out}} \times \Nt_{\mathrm{in}}) \times \Nt_{\mathrm{hid}}}, \boldsymbol{\mathrm{b}}^{(\mathrm{L})} \in \sR^{(\Nt_{\mathrm{out}} \times \Nt_{\mathrm{in}})}
\end{align*}
\begin{gather*}
    \boldsymbol{\mathrm{g}}\big( [x,y] ; \boldsymbol{\theta}^{(l)}\big) = \exp \bigg(-\frac{1}{2} \Big[\Big(\boldsymbol{\gamma}^{(l)}_\mathrm{X}\big(x - \boldsymbol{\mu}_\mathrm{X}^{(l)}\big)\Big)^2 \hspace{-1mm}+ \Big(\boldsymbol{\gamma}_\mathrm{Y}^{(l)}\big(y - \boldsymbol{\mu}_\mathrm{Y}^{(l)}\big)\Big)^2\Big] \bigg)\, \mathrm{Sin} \big(\Wm_\mathrm{g}^{(l)} [x, y] + \boldsymbol{\mathrm{b}}_\mathrm{g}^{(l)} \big)\\
    \boldsymbol{\theta}^{(l)} {=} \Big\{ \boldsymbol{\gamma}_{\Xt}^{(l)}\in \sR^{\Nt_{\mathrm{hid}}}, \boldsymbol{\gamma}_{\Yt}^{(l)}\in \sR^{\Nt_{\mathrm{hid}}}, \boldsymbol{\mu}_{\Xt}^{(l)}\in \sR^{\Nt_{\mathrm{hid}}}, \boldsymbol{\mu}_{\Yt}^{(l)}\in \sR^{\Nt_{\mathrm{hid}} },\Wm_\mathrm{g}^{(l)}\in \sR^{\Nt_{\mathrm{hid}} \times 2}, \boldsymbol{\mathrm{b}}_\mathrm{g}^{(l)}\in \sR^{\Nt_{\mathrm{hid}}} \Big\}
\end{gather*}

To analyse the maximum frequency component $f^+_{\textrm{MAGNet}}$, %(\boldsymbol{\theta},\Wm, \boldsymbol{h})$,
we analyse the frequency components of the Gabor filters used in MAGNet, and retain their maximum. We then plug the found frequency component into the analysis of \citet{fathony2021multiplicative} to show how the frequency responses of Gabor filters and linear layers interact in MFNs. Finally, we add the effect of the FlexConv Gaussian mask to our analysis to obtain the maximum frequency component ot the final FlexConv kernel $f^+_{\textrm{FlexConv}}$.%(\boldsymbol{\theta}, \boldsymbol{W}, \boldsymbol{h})$.

\textbf{Sine term in Gabor filters.} In a Gabor filter, the sine term is multiplied with a Gaussian envelope. The frequency (in radians) of a sine function of the form $\mathrm{Sin}(\boldsymbol{w}^T[x,y] + b)$ is given by $\boldsymbol{w}$. We divide by $2 \pi$ to convert the frequency units to Hertz, for compatibility with the rest of the analysis. For 2D inputs, the maximum frequency component of the sine function correspond to the largest frequency in the two input dimensions:
\begin{equation}
    f^+_{\mathrm{Sin}} = \max_{j} \frac{w_{j}}{2 \pi}.
\end{equation}
The sine terms in MAGNets have multiple output channels: $\mathrm{Sin} \big(\Wm_\mathrm{g} \cdot [x, y] + \boldsymbol{\mathrm{b}}_\mathrm{g}^{(l)} \big)$. Effectively, we compute the sine term independently for each channel:
\begin{equation}
    f^+_{\mathrm{Sin}, i} = \max_{j} \frac{\Wm_{\mathrm{g}, i,j}}{2 \pi}.
\end{equation}
\textbf{Gaussian term in Gabor filters.} In a Gabor filter, a Gaussian envelope modulates a sine term. Let us assume for now that the Gaussian envelope is isotropic, rather than anisotropic as in MAGNets, and has single-channel output. By applying the convolution theorem, the sine term is equivalently convolved with the Fourier transform of the Gaussian envelope in the frequency domain. Since the Fourier transform of a Gaussian envelope is another Gaussian envelope, the application of a Gaussian envelope amounts to blurring with a Gaussian kernel in the frequency domain. The size of the envelope in the Fourier domain $\sigma_{\mathrm{F}}$ can be derived from the standard deviation of the Guassian envelope in the spatial domain $\sigma_\mathrm{T}$ as follows:
\begin{equation}
    \sigma_\mathrm{T} \sigma_{\mathrm{F}} = \frac{1}{2 \pi} \Rightarrow \sigma_{\mathrm{F}} = \frac{1}{2 \pi \sigma_\mathrm{T}}.
\end{equation}
% The Gaussian blur of the frequency spectrum yields a smoothing out of impulse signals over a radius of $\frac{1}{2 \pi \sigma_T}$.
Gaussian blurs induce impulse signals to have a long tail. Consequently, we must define a cutoff point for this tail in terms of standard deviations to derive the maximum added frequency induced by the blur. We describe the cutoff point as $\sigma_{\mathrm{cut}} \in \mathbb{N}$. %in terms of standard deviations $h$, to know the maximum added frequency. 
Typical choices for $\sigma_{\mathrm{cut}}$ are known as the \textit{empirical}, or the "68-95-99.7" rule \citep{hald2007moivre}. We choose two standard deviations, i.e., $\sigma_{\mathrm{cut}}{=}2$, which covers 95\% of the mass of the Gaussian envelope.

For an isotropic Gabor filter with $\gamma {=} \sigma_\mathrm{T}^{-1}$, the maximum frequency of its Gaussian envelope $f^{+}_{\mathrm{env}}$ is given by:
%Finally, fitting with the parameterization of MAGNet using $\boldsymbol{\gamma} = \frac{1}{\sigma_T}$, and we arrive at:
\begin{equation}
    f^+_{\mathrm{env}} = \frac{\sigma_\mathrm{cut}}{2 \pi (\sigma_\mathrm{T})^{-1}} = \frac{\sigma_{\mathrm{cut}} \gamma}{2 \pi}.
\end{equation}
\textbf{Anisotropic envelopes.} Our analysis so far assumes an isotropic Gaussian envelope in the Gabor filter. However, we need to account for the anisotropic Gaussian envelopes in MAGNets. % versus the isotropic Gaussian envelopes of the MFNs. 
Anisotropic filters have not one but two $\gamma$ parameters: $\{ \gamma_{\Xt}, \gamma_{\Yt}\}$. The smallest of these will contribute most to $f^+_{\mathrm{env}}$, as it will blur the most, so it is sufficient to compute $f^+_{\mathrm{env}}$ only using the smallest of the two $\gamma$ terms:
\begin{equation}
    f^+_{\mathrm{env}}(\gamma_X, \gamma_Y) = f^+_{\mathrm{env}}(\min\{\gamma_X, \gamma_Y\}).
\end{equation}
The other assumption we made before was to work with single-channel outputs. MAGNets however use multi-channel outputs with independent Gaussian terms. The maximum frequency of multi-channel Gaussian envelopes is given by:
\begin{equation}
    f^+_{\mathrm{env}, i}(\boldsymbol{\gamma}_{\Xt}, \boldsymbol{\gamma}_{\Yt}) = f^+_{\mathrm{env}}\left(\min\{\boldsymbol{\gamma}_{\Xt, i}, \boldsymbol{\gamma}_{\Yt,i}\}\right) = \frac{\sigma_\mathrm{cut} \min\{\boldsymbol{\gamma}_{\Xt, i}, \boldsymbol{\gamma}_{\Yt,i}\}}{2 \pi}, \label{eq:3_freqgaussian}
\end{equation}
where the subscript $i$ indexes the channels of the multi-channel Gaussian envelopes.
%Note the subscript $i$, indexing over the channels of the MAGNet layer.

\begin{figure}[t]
    \centering
    \includegraphics[width=1.0\textwidth]{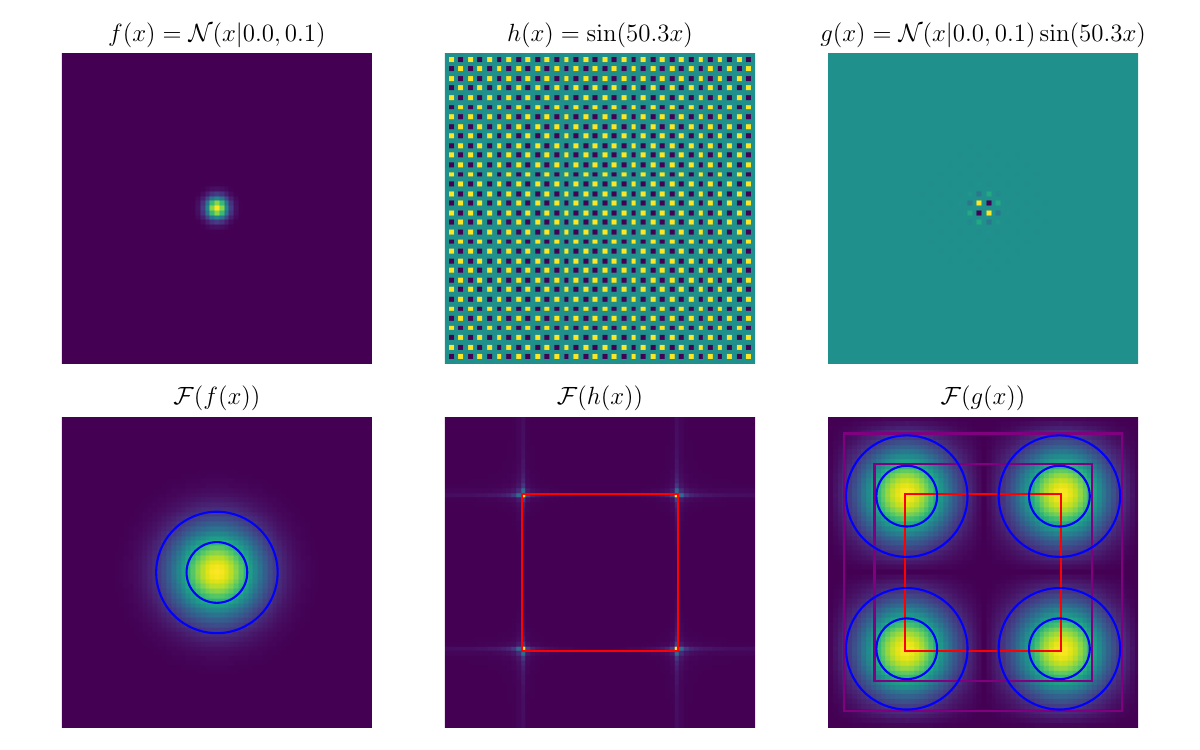}
    \vspace{-8mm}
    \caption{Decomposition of  a Gabor filter and its frequency spectrum. Top row: a decomposition of a Gabor filter (right) into its Gaussian term (left) and its sine term (center). Bottom row: frequency responses for each respective filter. The Fourier transform of a Gaussian envelope is a Gaussian envelope (blue circles show $\sigma_{\mathcal{F}}$ for $h = \{1, 2\}$). The Fourier transform of a sine pattern is a collection of symmetrical impulse signals (red box shows the Nyquist frequency). The Gaussian envelope blurs the frequency response of the sine term (purple boxes show the frequency response for $h = \{1, 2, 3\}$).}
    \label{fig:3_gabor-example}
\end{figure}

\textbf{Maximum frequency component of anisotropic Gabor filters.} Finally, the maximum frequency component of the $i$-th channel of an anisotropic Gabor filter $\boldsymbol{\mathrm{g}}$ is given by:
\begin{align}
    f^+_{\mathrm{Gabor}, i} &= f^+_{\mathrm{Sin}, i}(\Wm_\mathrm{g}) + f^+_{\mathrm{env}, i}(\boldsymbol{\gamma}_{\Xt}, \boldsymbol{\gamma}_{\Yt}) \nonumber \\
    &= \left( \max_{j} \frac{\Wm_{\mathrm{g}, i,j}}{2 \pi} \right) + \frac{\sigma_\mathrm{cut} \min\{\boldsymbol{\gamma}_{\Xt, i}, \boldsymbol{\gamma}_{\Yt,i}\}}{2 \pi}.
\end{align}
Figure~\ref{fig:3_gabor-example} illustrates the frequency spectrum of an example Gabor filter.

\textbf{Maximum frequency component of a MAGNet.} \citet{fathony2021multiplicative} characterize the expansion of each term of the isotropic Gabor layers in MFNs in the final MFN output. In Eq.~25, \citet{fathony2021multiplicative} demonstrate that the MFN representation contains a set of sine frequencies $\boldsymbol{\overline{\omega}}$ given by:
\begin{equation}
    \boldsymbol{\overline{\omega}} = \left\{ s_{\mathrm{L}} \omega_{i_{\mathrm{L}}}^{({\mathrm{L}})} + s_{\mathrm{L-1}} \omega_{i_{\mathrm{L-1}}}^{({\mathrm{L-1}})} + \cdots + s_l \omega_{i_2}^{(2)} + \omega_{i_1}^{(1)} \right\}.
\end{equation}
Here, the indexes $i_1, i_2, \cdots, i_{\mathrm{L}-1}$ range over all possible indices of each hidden unit of each layer of an MFN, and $s_2, \cdots, s_\mathrm{L} \in \{-1, +1\}$ range over all $2^{\mathrm{L}-1}$ possible binary signs. In other words, \citet{fathony2021multiplicative} demonstrate that the representation of an MFN at a particular layer contains an exponential combination of all possible positive and negative combinations of the frequencies of the sine terms in each hidden unit at each layer in the MFN up to the current layer.

The original analysis uses these terms to argue that MFNs model exponentially many terms through a linear amount of layers. For our purpose of computing the frequency response of the MAGNet kernel, we can plug our derivation of the frequencies of the Gabor filter $f_\mathrm{Gabor}$ into $\boldsymbol{\overline{\omega}}$ to compute its frequency spectrum:
\begin{equation} \label{eq:3_spectrum_magnet}
    \boldsymbol{f}^+_{\textrm{MAGNet}} = \left\{ s_\mathrm{L} f^{(\mathrm{L})}_{\mathrm{Gabor}, i_\mathrm{L}} + s_{\mathrm{L}-1} f^{(\mathrm{L}-1)}_{\mathrm{Gabor}, i_{\mathrm{L}-1}}  + \cdots + s_2 f^{(2)}_{\mathrm{Gabor}, i_2} + f^{ (1)}_{\mathrm{Gabor}, i_1} \right\}
\end{equation}
% Each term in this sum is indexed on a particular hidden channel $i_k$ of the MAGNet layer.
As stated before, we are only interested in the maximum frequency in the frequency spectrum. We can therefore simplify Eq.~\ref{eq:3_spectrum_magnet} in two ways. First, we simplify over MAGNet layers by taking the maximum value of the spectrum, which is the sum over all layers using only the positive binary signs in $s_\mathrm{L}$ (Eq.~\ref{eq:3_simplifysigns}). Next, we simplify over channel indices by retaining only the channel index that results in the highest frequency (Eq.~\ref{eq:3_simplifychannels}). The maximum frequency of a MAGNet is shown in Eq.~\ref{eq:3_magnetfreq}: 
\begin{align}
    \boldsymbol{f}^+_{\textrm{MAGNet}} &= \left\{ (+1) f^{+\hspace{0.5mm}(\mathrm{L})}_{\mathrm{Gabor}, i_\mathrm{L}} + (+1)f^{+\hspace{0.5mm}(\mathrm{L}-1)}_{\mathrm{Gabor}, i_{\mathrm{L}-1}} + \cdots + (+1) f^{+\hspace{0.5mm}(2)}_{\mathrm{Gabor}, i_2} + f^{+\hspace{0.5mm}(1)}_{\mathrm{Gabor}, i_1}\right\} \label{eq:3_simplifysigns} \\
    &= \left\{ f^{+\hspace{0.5mm}(\mathrm{L})}_{\mathrm{Gabor}, i_\mathrm{L}} + f^{+\hspace{0.5mm}(\mathrm{L}-1)}_{\mathrm{Gabor}, i_{\mathrm{L}-1}} + \cdots + f^{+\hspace{0.5mm}(2)}_{\mathrm{Gabor}, i_2} + f^{+\hspace{0.5mm}(1)}_{\mathrm{Gabor}, i_1}\right\} \nonumber\\
    f^+_{\textrm{MAGNet}} &= \max_{i_\mathrm{L}}\left(f^{+\hspace{0.5mm}(\mathrm{L})}_{\mathrm{Gabor}, i_\mathrm{L}}\right) + \max_{i_{\mathrm{L}-1}}\left(f^{+\hspace{0.5mm}(\mathrm{L}-1)}_{\mathrm{Gabor}, i_{\mathrm{L}-1}}\right) \cdots + \max_{i_2}\left(f^{+\hspace{0.5mm}(2)}_{\mathrm{Gabor}, i_2}\right) + \max_{i_1}\left(f^{+\hspace{0.5mm}(1)}_{\mathrm{Gabor}, i_1}\right) \label{eq:3_simplifychannels} \\
    &= \sum_{l=1}^{\mathrm{L}} \max_{i_l}\left(f^{+\hspace{0.5mm}(l)}_{\mathrm{Gabor}, i_l}\right) \nonumber \\
    &= \sum_{l=1}^{\mathrm{L}} \max_{i_l}\left( \left( \max_{j} \frac{\Wm^{(l)}_{\mathrm{g}, i_l,j}}{2 \pi} \right) + \frac{\sigma_\mathrm{cut} \min\{\boldsymbol{\gamma}^{(l)}_{\Xt, i_l}, \boldsymbol{\gamma}^{(l)}_{\Yt,i_l}\}}{2 \pi}\right). \label{eq:3_magnetfreq}
\end{align}
\textbf{Effect of the Gaussian mask in the frequency components of a FlexConv.} %There is a single operation remaining to be analyzed.
FlexConvs attenuate the MAGNet output with a Gaussian mask. The Gaussian mask (Eq.~\ref{eq:3_gaussianmask}) works analogously to the Gaussian envelope term in the Gabor filter: it blurs the frequency components of the generated kernel with standard deviation $\sigma_{\mathrm{F}}$. Therefore, we can reuse our derivation for the Gaussian envelope of the Gabor filter (Eq.~\ref{eq:3_freqgaussian}). The maximum frequency component of a FlexConv kernel is given by:
\begin{align}
    f^+_{\textrm{FlexConv}} &= f^+_{\textrm{MAGNet}} + f^+_{\mathrm{env}} \nonumber\\
    &= f^+_{\textrm{MAGNet}} + \frac{\sigma_{\mathrm{cut}} \min\{\sigma_{\Xt}^{-1}, \sigma_{\Yt}^{-1}\}}{2 \pi} = f^+_{\textrm{MAGNet}} + \frac{\sigma_{\mathrm{cut}}}{\max\{\sigma_\Xt, \sigma_\Yt\} 2 \pi} \nonumber\\
    &= \sum_{l=1}^{\mathrm{L}} \max_{i_l}\left( \left( \max_{j} \frac{\Wm^{(l)}_{\mathrm{g}, i_l,j}}{2 \pi} \right) + \frac{\sigma_\mathrm{cut} \min\{\boldsymbol{\gamma}^{(l)}_{\Xt, i_l}, \boldsymbol{\gamma}^{(l)}_{\Yt,i_l}\}}{2 \pi}\right) + \frac{\sigma_\mathrm{cut}}{\max\{\sigma_\Xt, \sigma_\Yt\} 2 \pi}. \label{eq:3_flexconvfreq}
\end{align}

\textbf{Visualization of regularized kernels.} Fig.~\ref{fig:3_app-cifar10kernelfrequencies} shows example kernels from FlexNets trained with aliasing regularization. The frequency domain plots confirm the accuracy of our frequency component regularization.

\vspace{-7mm}
\subsection{Regularizing the frequency response of FlexConv}
\label{sec:3_magnetreg}

\textbf{Nyquist frequency of a FlexConv kernel.} Given the sampling rate $f_\mathrm{s}$ of the kernel, we can compute its Nyquist frequency $f_{\textrm{Nyq}}$ as:% half the sampling rate:
\begin{equation}
    f_{\textrm{Nyq}} = \frac{1}{2} f_s
\end{equation}
To compute the sampling rate, we note that the kernel coordinates input to our MAGNet stretch over a $[-1, 1]^{\mathrm{D}}$ domain. For a kernel of length $k$, we therefore sample one point in every $f_s = \frac{k-1}{2}$ units. Knowing the sampling rate in terms of the kernel size allows us to express the Nyquist frequency in terms of the (pre-masked) kernel size:
\begin{equation}
    f_{\textrm{Nyq}}(k) = \frac{1}{2} \frac{k-1}{2} = \frac{k-1}{4}. \label{eq:3_flexconvnyquist}
\end{equation}
Note that the kernel size in a FlexConv is initialized to be equal to the resolution of the input, if it is odd. For even resolutions, it is equal to the input resolution plus one.

\textbf{Constructing the regularization term.} We train FlexConv with a regularization term on the frequency response of the generated kernel to ensure that aliasing effects do not distort the performance of the model when it is inferred at a higher resolution. %for example in our experiments in Sec.~\ref{sec:3_crossresexperiments}. 
This section details the implementation of the regularization function.

From the parameters of each FlexConv module, we compute $f^+_{\textrm{FlexConv}}$ according to Eq.~\ref{eq:3_flexconvfreq}. For the amount of standard deviations to use in determining $f^+_{\mathrm{env}}$ (Eq.~\ref{eq:3_freqgaussian}) we use $h = 2$. From the kernel size $k$ of the FlexConv module we compute $f_{\textrm{Nyq}}(k)$ according to Eq.~\ref{eq:3_flexconvnyquist}. We then apply an L2 regularizer over the amount that $f^+_{\textrm{FlexConv}}$ exceeds $f_{\textrm{Nyq}}(k)$:
\begin{equation}
    \mathcal{L}_{\mathrm{HF}} = ||\max\{f^+_{\textrm{FlexConv}}, f_{\textrm{Nyq}}(k)\} - f_{\textrm{Nyq}}(k)||^2. \label{eq:3_regularizeflexconv}
\end{equation}
We weight $\mathcal{L}_{\mathrm{HF}}$ by $\lambda = 0.1$ when adding it to our loss function.

\textbf{Improved implementation.} Eq.~\ref{eq:3_regularizeflexconv} contains a sum over the $\mathrm{L}$ layers of the MAGNet. In practice, we prefer to regularize each layer $l \in \mathrm{L}$ separately, so that the gradients of the regularization of different layers are not dependent on each other. We therefore implement the anti-aliasing regularization by regularizing each MAGNet layer independently, and spreading the $f^+_{\mathrm{env}}$ term from the gaussian mask uniformly over all MAGNet layers:

\begin{align}
    \mathcal{L}_{\mathrm{HF},l} &= ||\max\left\{  f^+_{\textrm{MAGNet},l} + \frac{f^+_{\mathrm{env}}}{\mathrm{L}} , \frac{f_{\textrm{Nyq}}(k)}{\mathrm{L}} \right\} - \frac{f_{\textrm{Nyq}}(k)}{\mathrm{L}} ||^2 \label{eq:3_regularizeflexconvlayerwise} \\
    &= || \max\left\{ \max_{i_l}\left(f^{+\hspace{0.5mm}(l)}_{\mathrm{Gabor}, i_l}\right) + \frac{f^+_{\mathrm{env}}}{\mathrm{L}}, \frac{f_{\textrm{Nyq}}(k)}{\mathrm{L}} \right\} - \frac{f_{\textrm{Nyq}}(k)}{\mathrm{L}} ||^2.
\end{align}

In the code, we refer to this method as the \textit{together} method, versus the \textit{summed} method of Eq.~\ref{eq:3_regularizeflexconv}. In preliminary experiments, we observed improved performance of anti-aliasing training when using the \textit{together} method. All of our experiments anti-aliasing experiments therefore use the \textit{together} setting.

% \section{Continuous Kernel Convolutions at Different Sampling Rates}
% \label{sec:3_samplingratenorm}

% In Sec.~\ref{sec:3_magnets} we show how FlexConvs can be trained on one resolution and transferred to another, simply by changing the sampling rate of the kernel function \mlp. This is possible because we represent our kernel function as a neural network \mlp\ from which we can sample at any desired resolution by changing the inputs to \mlp. 
% In this appendix, we give an intuitive explanation, as well as a formal derivation, for the requirement of a sampling rate normalization term (see Eq.~\ref{eq:3_multires}).

% As an example, let's imagine a case where we increase the resolution of a two-dimensional convolutional kernel by a factor 2 in each dimension. Each sample of the kernel function is weighted by the step size of the sample. The domain of \mlppsi\ is always $[-1, 1]$, so to achieve this, we sample with a $2 \times$ smaller step size from \mlppsi. To compensate for the extra samples, we weight all samples by by the inverse of the resolution increase: $\frac{1}{2}$. Finally, we generalize this to $\mathrm{D}$-dimensional kernels, where we have to normalize in each dimension: $\frac{1}{2}^\mathrm{D} = \frac{1}{4}$.

% \todo{TODO: Some derivation...}

% \begin{align}
%     \Big(f_{\mathrm{r}^{(2)}} * \boldsymbol{\psi}_{\mathrm{r}^{(2)}}\Big) &\approx \underbrace{\left(\frac{\mathrm{r}^{(1)}}{\mathrm{r}^{(2)}}\right)^{\mathrm{D}}}_{\text{Sampling rate norm}} \Big(f_{\mathrm{r}^{(1)}} * \boldsymbol{\psi}_{\mathrm{r}^{(1)}}\Big). \label{eq:3_multires}
% \end{align}
\vspace{-7mm}
\section{Dataset Description}
\label{sec:3_datasets}
\vspace{-5mm}
\subsection{Image Fitting Datasets}
\textbf{Kodak dataset.} The Kodak dataset \citep{kodak1991} consists of 24 natural images of size $768\times512$. This dataset is a popular benchmark used for compression and image fitting methods.% The goal of this task it to fit each of the image in this dataset with a neural network \mlppsi$: \sR \rightarrow \sR^{3}$.
\vspace{-7mm}
\subsection{Sequential Datasets}
\textbf{Sequential and Permuted MNIST.} The sequential MNIST dataset (sMNIST) \citep{le2015simple} takes the $28{\times}28$ images from the original MNIST dataset \citep{lecun1998gradient}, and presents them as a sequence of 784 pixels. The goal of this task is to perform digit classification given the representation of the last sequence element of a sequential model. Consequently, good predictions require the model to preserve dependencies up to 784 steps in the past.

The permuted MNIST dataset (pMNIST) additionally changes the order of all the sMNIST sequences by a random permutation. Consequently, models can no longer rely on local features to construct good feature representations. As a result, the classification problem becomes more difficult, and the importance of long-term dependencies more pronounced.

\textbf{Sequential and Noise-Padded CIFAR10.} The sequential CIFAR10 dataset (sCIFAR10) \citep{chang2017dilated} takes the $32{\times}32$ images from the original CIFAR10 dataset \citep{krizhevsky2009learning} and presents them as a sequence of 1,024 pixels. The goal of this task is to perform image classification given the representation of the last sequence element of a sequential model. This task is more difficult than sMNIST, as a larger memory horizon is required to solve the task and more complex structures and intra-class variations are present in the data \citep{bai2018trellis}. 

The noise-padded CIFAR10 dataset (npCIFAR10) \citep{chang2019antisymmetricrnn} flattens the images from the original CIFAR10 dataset \citep{krizhevsky2009learning} along their rows to create a sequence of length 32, and 96 channels (32 rows $\times$ 3 channels). Next, these sequences are concatenated with 968 entries of noise to form the final sequences of length 1000. As for sCIFAR10, the goal of the task is to perform image classification given the representation of the last sequence element of a sequential model. 

\textbf{CharacterTrajectories.} The CharacterTrajectories dataset is part of the UEA time series classification archive \citep{bagnall2018uea}. It consists of 2858 time series of length 182 and 3 channels representing the $x,y$ positions, and the tip force of a pen while writing Latin alphabet characters in a single stroke. The goal is to classify out of 20 classes the written character using the time series data. 

\textbf{Speech Commands.} The Speech Commands dataset \citep{warden2018speech} consists of 105,809 one-second audio recordings of 35 spoken words sampled at 16$\mathrm{kHz}$. Following \citet{kidger2020neural}, we extract 34975 recordings from ten spoken words to construct a balanced classification problem. We refer to this dataset as \textbf{SpeechCommands\_raw}, or \textbf{SC\_raw} for short. Furhtermore, we utilize the preprocessing steps of \citet{kidger2020neural} and extract mel-frequency cepstrum coefficients from the raw data. The resulting dataset, abreviated \textbf{SC}, consists of time series of length 101, and 20 channels.

\subsection{Image Benchmark Datasets}\label{appx:image_datsets}

\textbf{MNIST.} The MNIST hadwritten digits datset \citep{lecun-mnisthandwrittendigit-2010} consists of 70,000 gray-scale handwritten digits of size $28 {\times} 28$, divided into a training and test sets of 60,000 and 10,000 images, respectively. The goal of the task is to classify these digits as one of the ten possible digits $(0, 1, .. 8, 9)$.%, to be classified in ten classes, 60,000 of which are used for training. It is a subset of a database published by NIST. The images are single-channel . 

\textbf{CIFAR-10} The CIFAR-10 dataset \citep{krizhevsky2009learning} consists of 60,000 natural images from 10 classes of size $32{\times}32$, divided into training and test sets of 50,000 and 10,000 images.

\textbf{STL-10.} The STL-10 dataset \citep{coates2011analysis} is a subset of the ImageNet dataset \citep{krizhevsky2012imagenet} consisting of 5,000 natural images from 10 classes of size $96{\times}96$, divided into trainint and test sets of 4,500 and 500 images, respectively.

\textbf{ImageNet-$\mathrm{k}$.} The Imagenet-$\mathrm{k}$ \citep{ChrabaszczLH17} dataset is derived from the ImageNet dataset \cite{ILSVRC15} by downsampling all samples to a resolution $\mathrm{k} \in [64, 32, 16, 8]$. The dataset contains 1000 classes with 1,281,167 training samples and 50,000 validation samples.

\vspace{-7mm}
\section{Additional Experiments}
\label{sec:3_appx-experiments}

\begin{table}
\centering
\caption{Average PSNR for fitting of images in the Kodak dataset. Both our improved initialization scheme, as well as the inclusion of anisotropic Gabor functions lead to better reconstructions.}
\label{tab:3_kodak-results}
\vspace{-3mm}
\begin{center}
\scalebox{0.75}{
\begin{tabular}{cccc}
\toprule
    \multirow{2}{*}{\sc{Model}} & \multirow{2}{*}{\sc{\# Params}} & \sc{Improved} &  \multirow{2}{*}{\sc{PSNR}} \\
    & & \sc{Init}\\
    \midrule
    SIREN & 7.14\sc{k} & - & 25.665 \\
    MFN$_{\text{Fourier}}$ & 7.40{\sc{k}}& - & 23.276\\
    \multirow{2}{*}{MFN$_{\text{Gabor}}$} & \multirow{2}{*}{7.11{\sc{k}}} & \xmark & 25.361 \\
     & & \cmark &  \textbf{25.606} \\ 
     \midrule
    \multirow{2}{*}{MAGNet} & \multirow{2}{*}{7.36{\sc{k}}} & \xmark & 25.791 \\ 
    & & \cmark & \textbf{\underline{25.893}} \\
  \bottomrule
\end{tabular}}
\end{center}
\end{table}

\vspace{-5mm}
\subsection{Image Classification}

\begin{table}%{r}{0.48\textwidth}
\centering
\caption{Full results on CIFAR-10. We report results over three runs per setting. CIFARResNet-44 w/ CKConv is a CIFARResNet-44 where all convolutional layers are replaced with CKConvs with $k = 3$. CIFARResNet-44 w/ FlexConv is a CIFARResNet-44 where all convolutional layers are replaced with FlexConv with learned kernel size, except for the shortcut connections of the strided convolutional layers, which are pointwise convolutions. *Results are taken from the respective original works instead of reproduced. \dagger Results are from single run.}
\vspace{-3mm}
\label{tab:3_full-cifar-10}
\begin{center}
\scalebox{0.7}{
\begin{tabular}{ccc}
\toprule
    \multirow{2}{*}{\sc{Model}} & \multirow{2}{*}{\sc{Size}} & \sc{CIFAR-10} \\
    & & \sc{Acc.} \\ \midrule
    DCN-$\sigma^{ji}$ \citep{tomen2021deep} & 0.47\sc{m} & 89.7 $\pm$ 0.3* \\
    N-Jet-CIFARResNet32 \citep{pintea2021resolution} & 0.52\sc{m} & 92.3 $\pm$ 0.3* \\
    N-Jet-ALLCNN \citep{pintea2021resolution} & 1.07\sc{m} & 92.5 $\pm$ 0.1* \\ \midrule
    CIFARResNet-44 \citep{he2016deep} & 0.66\sc{m} & 92.9*\!\dagger \\
    CIFARResNet-44 \citep{he2016deep} (our reproduction) & 0.66\sc{m} & 90.9 $\pm$ 0.2 \\\midrule
    CIFARResNet-44 w/ CKConv ($k = 3$) & 2.58\sc{m} & 86.1 $\pm$ 0.9 \\
    CIFARResNet-44 w/ FlexConv & 2.58\sc{m} & 81.6 $\pm$ 0.8 \\ \midrule
    FlexNet-7 w/ conv. ($k = 3$) & 0.17\sc{m} & 89.5 $\pm$ 0.3   \\
    FlexNet-7 w/ conv. ($k = 33$) & 20.0\sc{m} & 78.0 $\pm$ 0.3  \\
    FlexNet-7 w/ N-Jet \citep{pintea2021resolution} & 0.70\sc{m} & 91.7 $\pm$ 0.1  \\ \midrule
    \midrule
    CKCNN$_{\text{SIREN}}$-3 & 0.26\sc{m} & 72.4* \\ % 72.39
    CKCNN$_{\text{Fourier}}$-3 & 0.27\sc{m} & 83.8* \\ % 83.77
    CKCNN$_{\text{Gabor}}$-3 & 0.28\sc{m} & 85.6* \\ % 85.62
    CKCNN$_{\text{MAGNet}}$-3 & 0.28\sc{m} & 86.2* \\ % 86.18
    CKCNN-7 & 0.63\sc{m} & 71.7* \\ % 71.70
    CKCNN$_{\text{Fourier}}$-7 & 0.63\sc{m} & 84.6* \\ % 84.59
    CKCNN$_{\text{Gabor}}$-7 & 0.67\sc{m} & 87.7*  \\ % 87.65
    CKCNN$_{\text{MAGNet}}$-7 & 0.67\sc{m} & 85.9* \\ % 85.91
    \midrule
    FlexNet$_{\text{SIREN}}$-7 & 0.63\sc{m} & 88.9* \\ % 88.88
    FlexNet$_{\text{Fourier}}$-7 & 0.66\sc{m} & 91.6* \\ % 91.58
    FlexNet$_{\text{Gabor}}$-7 & 0.67\sc{m} & 92.0* \\ % 91.99
    \midrule
    FlexNet-3 & 0.27\sc{m} & 90.4 $\pm$ 0.2 \\
    FlexNet-5 & 0.44\sc{m} & 91.0 $\pm$ 0.5 \\
    FlexNet-7 & 0.67\sc{m} & 92.2 $\pm$ 0.1 \\
 \bottomrule
\end{tabular}}
\end{center}
\end{table}

\begin{table}
% \begin{wraptable}{r}{0.45\textwidth}
\centering
\caption{Results on ImageNet-32. *Results are taken from the respective original works instead of reproduced. \dagger Results are from a single run.}
\label{tab:3_imagenet-32}
\vspace{-3mm}
\begin{center}
\scalebox{0.7}{
\begin{tabular}{cccc}
\toprule
    \multirow{2}{*}{\sc{Model}} & \multirow{2}{*}{\sc{Size}} &  \multicolumn{2}{c}{\sc{ImageNet-32}} \\
    & & \sc{Top-1} & \sc{Top-5} \\ \midrule
    CIFARResNet-32 & 0.53\sc{m} & 26.41 $\pm$ 0.13 & 49.37 $\pm$ 0.15 \\ 
    WRN-28-1 & 0.44\sc{m} & 32.03*\!\dagger & 57.51*\!\dagger \\ \midrule
    FlexNet-5 & 0.44\sc{m} & 24.9 $\pm$ 0.4 & 47.7 $\pm$ 0.6 \\
  \bottomrule
\end{tabular}}
\end{center}
\end{table}

\begin{table}
\centering
\caption{Results for alias-free FlexNets on CIFAR-10 and ImageNet-$\mathrm{k}$. $\Delta$ denotes difference in accuracy. 
% N-Jets and FlexConv perform comparatively bad on ImageNet-32. We train each model for 100 epochs on ImageNet-16, then evaluate on ImageNet-32. We report results over three runs per setting.
}
\label{tab:3_crossres-imagenet}
\vspace{-3mm}
\begin{center}
\scalebox{0.75}{
\begin{tabular}{cccc}
\toprule
    \multirow{2}{*}{\sc{Model}} & \multirow{2}{*}{\sc{Size}} &  \multicolumn{2}{c}{\sc{ImageNet-k Top-1}} \\
    & & $k = 16$ & $\Delta_{k = 16}$ $k = 32$ \\ \midrule % & fine-tuned \\ \midrule
    CIFARResNet-32 & 0.52\sc{m} & 16.1 $\pm$ 0.0 & -11.6 $\pm$ 0.4 \\ \midrule % \todo{RUNNING} \\ \midrule
    FlexNet-5 w/ N-Jets & 0.46\sc{m} & 15.7 $\pm$ 0.1 & -1.9 $\pm$ 0.4 \\ \midrule % & \todo{RUNNING}  \\ \midrule
    FlexNet-5 & 0.44\sc{m} & 14.9 $\pm$ 0.1 & -1.9 $\pm$ 1.7 \\ % & 24.8 $\pm$ 0.2 \\
 \bottomrule
\end{tabular}}
\end{center}
\end{table}

\textbf{CIFAR-10.} Tab.~\ref{tab:3_full-cifar-10} shows our CIFAR-10 results including more ablations.

\textbf{ImageNet-32.} Results for the ImageNet-32 experiment are shown in Table~\ref{tab:3_imagenet-32}. FlexNets are slightly worse than CIFARResNet-32 \citep{he2016deep} with slightly less parameters. However, the results reported by \citet{ChrabaszczLH17} for Wide ResNets \citep{zagoruyko2016wide} outperform FlexNets by a significant margin.

\textbf{Alias-free ImageNet-32.} We report results for alias-free FlexNets on ImageNet-$\mathrm{k}$ \citep{ChrabaszczLH17} in Table~\ref{tab:3_crossres-imagenet}, to verify the results of alias-free training at a larger scale. We find that FlexConv and N-Jet both mostly retain classification accuracy between source and target resolution, while CIFARResNet-32 degrades drastically.

\textbf{MNIST and STL-10.} We additionally report results on MNIST (Tab.~\ref{tab:3_mnist}) and STL-10 (Tab.~\ref{tab:3_stl10}. We choose these dataset for the difference in image sizes of the training data. On MNIST, though performance on MNIST is quite saturated, we are competitive with state of the art methods. On STL-10 we are significantly worse than the baseline CIFARResNet from \citep{luo2020extended}, though with significantly less parameters. We were not able to prepare a more relevant baseline for this experiment.

\begin{table}
\centering
\caption{Results on MNIST. We train each model with three different seeds and report mean and standard deviation. *Results are taken from the respective original works instead of reproduced. \dagger Results are from single run.}
\label{tab:3_mnist}
\vspace{-3mm}
\begin{center}
\scalebox{0.75}{
\begin{tabular}{ccc}
\toprule
    \multirow{2}{*}{\sc{Model}} & \multirow{2}{*}{\sc{Size}} & \sc{MNIST} \\
    & & \sc{Acc.} \\ \midrule
    % ResNet-18 \citep{luo2020extended} & 11.2\sc{m} &  \\ \midrule
    Efficient-CapsNet \citep{mazzia2021efficient} & 0.16\sc{m} & \textbf{99.8}*\!\dagger \\ % 99.84
    Network in Network \citep{lin2013network} & N/A & 99.6*\!\dagger \\ % 99.55
    VGG-5 (results from \citet{kabir2020spinalnet}) & 3.65\sc{m} & 99.7*\!\dagger \\ \midrule % 99.72
    FlexNet-16 & 0.67\sc{m} & 99.7 $\pm$ 0.0  \\
 \bottomrule
\end{tabular}}
\end{center}
\end{table}

\begin{table}
\centering
\caption{Results on STL-10. We train each model with three different seeds and report mean and standard deviation. *Results are taken from \citet{luo2020extended}. \dagger Results are from single run.}
\label{tab:3_stl10}
\vspace{-3mm}
\begin{center}
\scalebox{0.75}{
\begin{tabular}{ccc}
\toprule
    \multirow{2}{*}{\sc{Model}} & \multirow{2}{*}{\sc{Size}} & \sc{STL-10} \\
    & & \sc{Acc.} \\ \midrule
    CIFARResNet-18 & 11.2\sc{m} & 81.0*\!\dagger \\ 
    % ResNet-18 (reproduction) & 11.2\sc{m} & 64.5 $\pm$ 0.3 \\
    % ResNet-10 (reproduction) & 4.9\sc{m} & 64.3 $\pm$ 0.2 \\ \midrule
    FlexNet-16 & 0.67\sc{m} & 68.6 $\pm$ 0.7 \\
 \bottomrule
\end{tabular}}
\end{center}
\end{table}

\vspace{-7mm}
\section{Experimental Details}

\vspace{-5mm}
\subsection{FlexNet}
\label{sec:3_appx-flexnet}

\begin{figure}[t]
    \centering
    \includegraphics[width=.8\textwidth]{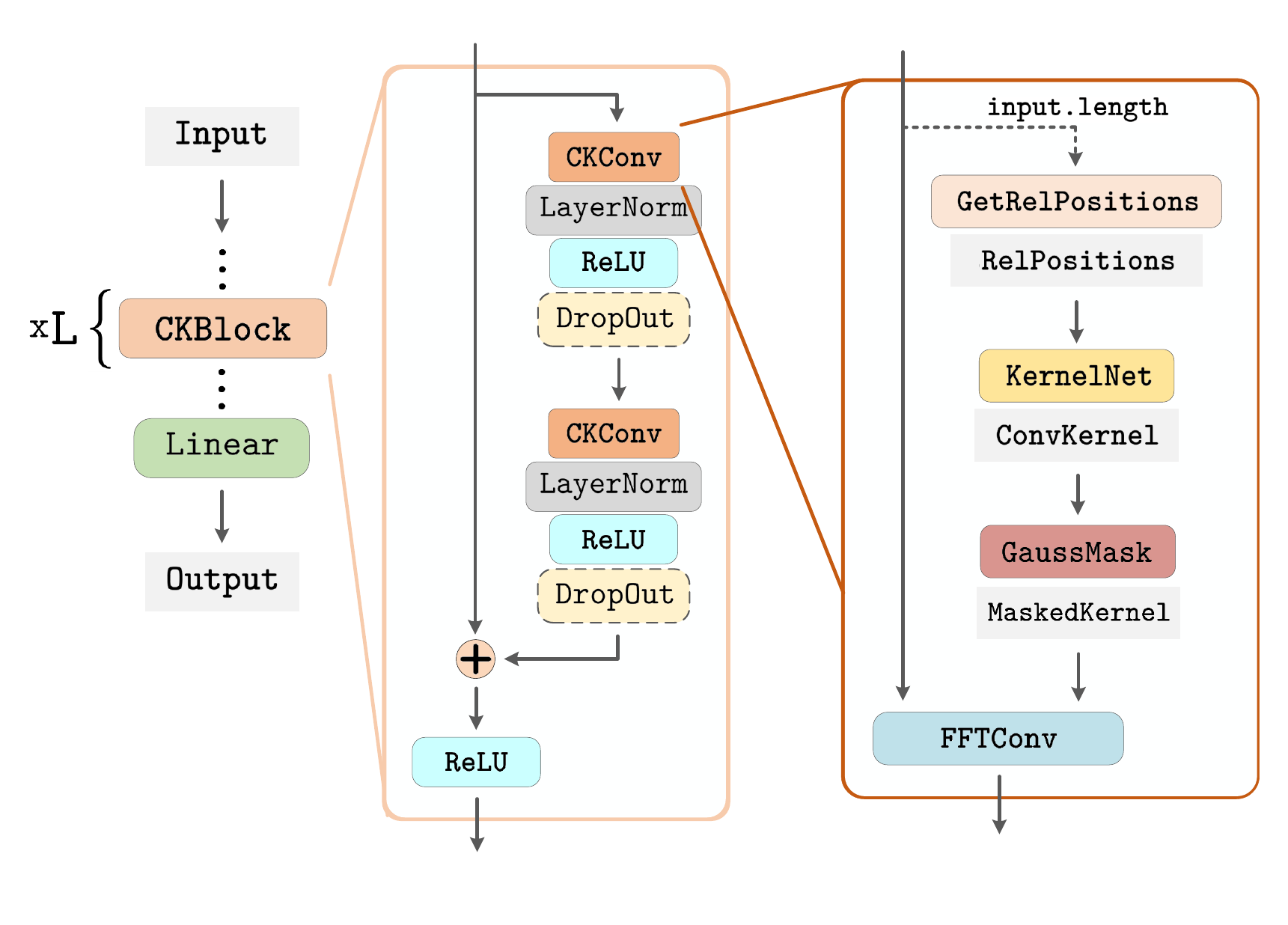}
    \vspace{-8mm}
    \caption{FlexNet architecture. FlexNet-$\mathrm{L}$ consists of $\tfrac{\mathrm{L}}{2}$ FlexBlocks, where each FlexBlock is a residual block of FlexConvs.}
    \label{fig:3_flexnet-architecture}
\end{figure}

We propose an image classification architecture named \textit{FlexNet} (Fig.~\ref{fig:3_flexnet-architecture}), consisting of a stack of FlexConv blocks followed by a global average pooling layer and a linear layer. FlexNets are named "FlexNet-$\mathrm{L}$" where $\mathrm{L}$ indicates the number of layers in the network.

\textbf{FlexBlock.} Each FlexBlock consists of two FlexConvs with BatchNorm \citep{ioffe2015batch} and dropout \citep{srivastava14a} ($d = 0.2$) as well as a residual connection. The width of a block $i$ is determined by scaling a base amount $c$ by progressively increasing factors: $c_i = [c, c \times 1.5, c \times 1.5, c \times 2.0, c \times 2.0](i)$. The default configuration of FlexNet uses $c = 22$. In FlexNet-N-Jet models, we scale $c$ to match the amount of parameters of the FlexNet in the comparison.

\textbf{FlexConv initialization.} We initialize the FlexConv mask variances small, at $\sigma^2_{\Xt}, \sigma^2_{\Yt} = 0.125$. For initializing MAGNet, we initialize the Gaussian envelopes as discussed in Sec.~\ref{sec:3_magnets}. We initialize the linear layer weights by the same Gamma distribution as used for the enveloped, modulated by a scaling factor of $25.6$. We found that this value of the scaling factor, rather than a higher one, helped in reducing the performance of alias-free models. We initialize the bias of the linear layers by $\gU(- \pi,  \pi)$.

\textbf{CIFAR-10.} In FlexNet-16 models for CIFAR-10 we use $c = 24$ to approximate the parameter count of CIFARResNets in the experiment.

\vspace{-7mm}
\subsection{Optimization}
\label{sec:3_flexnet-optimization}

We use Adam \citep{kingma2014adam} to optimize FlexNet. Unless otherwise specified, we use a learning rate of $0.01$ with a cosine annealing scheme \citep{loshchilov2017sgdr} with five warmup epochs. We use a different learning rate of $0.1\times$ the regular learning rate for the FlexConv Gaussian mask parameters. We do not use weight decay, unless otherwise specified.

\textbf{Kodak.} We overfit on each image of the dataset for 20,000 iterations. We use a learning rate of 0.01 without any learning rate scheme. We observe that SIRENs diverge with this learning rate and thus, reduce the learning rate to 0.001 for these models.

\textbf{CIFAR-10.} We train for 350 epochs with a batch size of 64. We use the data augmentation from \citet{he2016deep} when training CIFAR-10: a four pixel padding, followed by a random 32 pixel crop and a random horizontal flip.

\textbf{ImageNet-32.} We train for 350 epochs with a batch size of 2048. We use the same data augmentation as used for CIFAR-10, and a weight decay of $1\mathrm{e}{-5}$.

\textbf{Sequential and Permuted MNIST.} We train for 200 epochs with a batch size of 64 and a learning rate of 0.01. We use a weight decay of $1\mathrm{e}{-5}$.

\textbf{Sequential and Noise-Padded CIFAR-10.} For sequential CIFAR-10, we train for 200 epochs with a batch size of 64, a learning rate of 0.001 and a weight decay of $1\mathrm{e}{-5}$. For noise-padded CIFAR-10, we train for 300 epochs with a batch size of 32, a learning rate of 0.01 and no weight decay.

\textbf{Speech Commands and CharTrajectories.} We train for 300 epochs with a batch size of 32 and a learning rate of 0.001. For CharTrajectories, we use a weight decay of $1\mathrm{e}{-5}$.

\vspace{-7mm}
\subsection{Rotated Gaussian masks}
\label{sec:3_negative-steerable-gaussians}

MAGNets use anisotropic Gaussian terms in the Gabor filters, which yields improvements in descriptive power and convergence speed (Sec.~\ref{sec:3_magnets}). For the same reason, we explore making the anisotropic FlexConv Gaussian mask steerable, by including an additional vector of learnable angle parameters $\boldsymbol{\phi}^{(l)} \in \sR^{\Nt_{\mathrm{hid}}}$ that rotates the Gaussian masks. Although preliminary experiments show rotated masks lead to slight additional improvements, the computational overhead required to rotate the masks is large. Consequently, we do not consider rotated Gaussian masks in our final experiments.

% $\mathrm{Sample~at~}\mathrm{k}_1 \times \mathrm{k}_1$

%$\mathrm{Sample~at~}\mathrm{k}_2 \times \mathrm{k}_2$

%% file: backmatter/appx_5_dnarch.tex
% path to figures directory
\graphicspath{{figures/5-dnarch/}}

%=========================================================================

% \begin{savequote}[75mm]
% Nulla facilisi. In vel sem. Morbi id urna in diam dignissim feugiat. Proin molestie tortor eu velit. Aliquam erat volutpat. Nullam ultrices, diam tempus vulputate egestas, eros pede varius leo.
% \qauthor{Quoteauthor Lastname}
% \end{savequote}

\chapter{Learning Convolutional Neural Architectures by Backpropagation}
	%\label{chapter:}

%=========================================================================
\vspace{-7mm}
\section{Extended Related Work}
In this section, we position DNArch among the broad existing NAS literature, and provide a thorough comparison of DNArch with existing differentiable NAS techniques.\footnote{This section is deeply based on the excellent review of \citet{white2023neural}.}

\vspace{-7mm}
\subsection{Positioning of DNArch on the NAS literature Taxonomy}
Neural Architecture Search (NAS) techniques can be categorized based on three aspects \citep{elsken2019neural, wistuba2019survey, white2023neural}: (\textit{i}) their search space, (\textit{ii}) their search strategy, and (\textit{iii}) their performance evaluation strategy:
\begin{itemize}[topsep=0pt, leftmargin=*, itemsep=0pt]
    \item The \textit{search space} refers to the set of all architectures the NAS technique is allowed to select from. While the search space can be extremely general, incorporating domain knowledge into the design of the search spaces simplifies the search. However, this infusion of human expertise introduces human biases, which can reduce the chances of NAS techniques finding truly novel architectures.
    \item The \textit{search strategy} refers to the optimization technique used to find high-performing architectures in the search space.
    \item The \textit{performance evaluation strategy} refers to the method used to predict the performance of neural architectures in order to avoid fully training each candidate architecture. While it is possible to run a discrete search strategy by fully training and evaluating architectures chosen throughout the search, using a performance estimation can greatly increase the speed of the search.
\end{itemize}
It is important to note, though, that in this trichotomy, \textit{one-shot methods} --from which DNArch is part of-- (see below), the search strategy and the performance evaluation strategy are coupled \citep{xie2021weight}.

Within this trichonomy, DNArch can be understood as a novel search strategy. However, as we will see in the rest of this section, DNArch also has important implications for the design and flexibility of the search space, as well as the efficiency of the performance evaluation strategy.

\textbf{DNArch as a search strategy.} DNArch proposes a novel way to search among possible candidate architectures by viewing neural architectures as multi-dimensional continuous entities --with dimensions corresponding to depth, width, etc-- and using learnable parametric differentiable masks to learn their size during training. Existing search strategies can be divided into two classes: (\textit{i}) black-box optimization techniques and (\textit{ii}) one-shot techniques. Black-box optimization techniques, such as reinforcement learning \citep{zoph17,pham2018efficient,zoph2018learning}, evolutionary / genetic algorithms \citep{miller1989designing, real2019regularized, song2020efficient}, bayesian optimization \citep{swersky2014raiders, kandasamy2018neural, nguyen2021optimal} and Monte-Carlo tree search \citep{negrinho2017deeparchitect, wang2019alphax,wang2020neural}, have been widely used due to their strong performance and ease of use. However, despite their numerous advantages --simple optimization of non-differentiable objectives, simple parallelism, join optimization with other hyperparameters, etc--, these techniques require training a (vast) population of candidate architectures and often incur in immense computational costs, sometimes on the order of thousands of GPU days \citep{zoph17, real2019regularized}. \textit{One-shot} techniques avoid training each architecture from scratch by training all architectures in the search space via a single (\enquote{one-shot}) training of an over-parameterized network \citep{saxena2016convolutional,brock2017smash,liu2018darts}. Due to the extreme weight-sharing used, these methods are more scalable and efficient. 

DNArch falls within the family of one-shot techniques, as it trains a single neural architecture from which several possible sub-networks can be extracted by changing the size of the differentiable masks.

\textbf{DNArch as a one-shot model technique.} One-shot models can be divided in two main classes based on the type of over-parameterized network used to contain the space of all possible networks in the search space: (\textit{i}) a hypernetwork \citep{brock2017smash, zhang2018graph}  or (\textit{ii}) a supernetwork \citep{saxena2016convolutional, liu2018darts}. A \textit{hypernetwork} is a neural network that generates the weights of other neural networks \citep{schmidhuber1992learning, ha2017hypernetworks, romero2022ckconv}. A \textit{supernetwork}, on the other hand, is an over-parameterized architecture that contains all possible architectures in the search space as subnetworks. Currently, methods relying on supernetworks are more popular, with notorious examples being DARTS \citep{liu2018darts}, once-for-all networks \citep{cai2019once}, and slimable networks \citep{yu2018slimmable}.

Although DNArch relies on Continuous Kernel Convolution \citep{romero2022ckconv} --a type of hypernetwork-- for the generation of global kernels regardless of the resolution, length and dimensionality of the input, DNArch falls within the family of supernetworks. This is due to the fact that the differentiable masks are applied on a shared architecture to define their size. This is different from NAS techniques based on hypernetworks, where whole subnetworks are generated by the hypernetwork \citep{brock2017smash, zhang2018graph}.

\textbf{DNArch as a supernetwork technique.} The idea of a supernetwork was introduced by \citet{saxena2016convolutional} and was popularized by works such as ENAS \citep{pham2018efficient} and DARTS \citep{liu2018darts}. Supernetwork methods have become popular due to their scalability and efficiency. The reason for these properties being that \textit{a linear increase in the number of candidate operations only causes a linear increase in computational costs for training, but the number of subnetworks in the supernetwork increases exponentially} \citep{white2023neural}. Hence, an exponential number of architectures can be trained at linear cost.

While the supernetwork allows quick evaluation of all architectures, one must still define a search strategy for the selection of the final architecture. Supernetwork-based techniques can be divided in terms of the type of search strategy used: (\textit{i}) non-differentiable, and (\textit{ii}) differentiable techniques. 
While the first family uses black-box optimization algorithms to search for the best architecture \citep{pham2018efficient, bender2018understanding}, differentiable supernetwork-based methods, e.g., DARTS \citep{liu2018darts}, use of a continuous relaxation of the discrete architecture search space and rely on gradient descent to search for the best architecture in tandem with the training of the supernetwork. Through the use of gradient descent makes these methods are \textit{significantly faster} than black-box optimization methods \citep{liu2018darts, cai2019once}.

DNArch falls within the family of differentiable supernetwork-based methods. However, as discussed below, by treating the search space as a continuous space, DNArch does not experience several of the limitations present in existing differentiable supernetwork-based methods.
\vspace{-7mm}
\subsection{DNArch as a differentiable supernetwork method:\\ DARTS, follow-up works and DNArch}
\textbf{DARTS and follow-up works.} To understand the differences between DNArch and other existing differentiable supernetwork-based methods, it is important to start with a comprehensive description of DARTS \citep{liu2018darts}: the seminal work of differentiable NAS. 

At the start, each edge $(i,j)$ in the DARTS search space consists of multiple possible candidate operations $o$, each of which are associated with a continuous hyperparameter $\alpha_o^{(i, j)} \in [0,1]$. While the supernetwork is training, the edge $(i, j)$ consists of a \textit{mix of all candidate operations} weighted by each $\alpha_o^{(i, j)} \in [0,1]$. The architecture hyperparameters $\alpha$ are optimized jointly with the supernetwork weights $w$ via alternating gradient descent. After the training phase, DARTS uses a finetuning step to obtain the final discrete architecture. This is done by selecting the operation with the maximum value of $\alpha$ on each edge (the discretization step) and then re-trains it from scratch. In other words, DARTS consists of the following steps:
\begin{enumerate}[label=(\textit{\roman*}), topsep=0pt, leftmargin=*, itemsep=0pt]
\item \textit{Training.} The supernetwork is trained as a mixture of all possible components.
\item \textit{Discretization.} Once the training of supernetwork has finished, the operation with the maximum $\alpha$ value on each edge is selected to form the final architecture.
\item \textit{Fine-tuning.} The final architecture is retrained from scratch to get the final model.
\end{enumerate}
Due to its simplicity, novelty and speed, DARTS gained significant attention in the AutoML community. However, DARTS comes with several limitations, which have been addressed in several follow-up works. In the scope of DNArch, the most relevant limitations are:
\begin{itemize}[topsep=0pt, leftmargin=*, itemsep=0pt]
\item \textit{Poor test generalization.} \citet{zela2019understanding} and \citet{chen2020stabilizing} have shown that DARTS often converges to sharp local minima in the loss landscape with high validation loss curvature for the architecture hyperparameter space. As a result, running the discretization step can cause the algorithm to return an architecture with poor generalization. To alleviate this, existing works propose to make training more robust either through data augmentation and regularization, e.g., RobustDARTS \citep{zela2019understanding}, by optimizing the expected or worst-case training loss over a local neighborhood of the architecture hyperparameters, e.g., Smooth-DARTS \citep{chen2020stabilizing}, or by deriving more accurate approximations of the gradients of $\alpha$, e.g., Amended-DARTS \citep{bi2019stabilizing}, iDARTS \citep{pmlr-v139-zhang21s}.

On a different vein, GAEA \citep{li2020geometry}, XD \citep{roberts2021rethinking} and StacNAS \citep{guilin2019stacnas} propose a single-level optimization rather than the typical bi-level optimization by \textit{treating architecture hyperparameters as normal weights}, showing that this leads to better generalization.  
\item \textit{High memory consumption.} The memory required to train a supernetwork is much higher than that of a normal neural network. This is due to the fact that the size of the supernetwork \textit{scales linearly with the size of the set of candidate operations used}. To alleviate this cost, subsequent works propose to mask out all operations except for the ones corresponding to one of a few subnetworks in each training step. This has been proposed either by randomly sampling random operations with probabilities proportional to their current value of $\alpha$, e.g., ProxylessNAS \citep{cai2018proxylessnas}, by using a Gumbel-Softmax distribution over a one-hot encoding of the operation choices to sample single operations in a differentiable way, e.g., GDAS \citep{dong2019searching}, DSNAS \citep{hu2020dsnas, xie2021weight}, or by sampling a subset of the channels to be sent through the possible operations \citep{Xu2020PC-DARTS:}. 
\item \textit{Operation biases.} Several works show that DARTS tend to favor skip connections over other operations \citep{liang2019darts+, zela2019understanding, wang2021rethinking}. To avoid this, subsequent works propose to use an early stopping method based on the stability of the raking of the architecture weights, e.g., DARTS+ \citep{liang2019darts+}, separating the skip connection weights from other operation weights via auxiliary edges, e.g., DARTS- \citep{chu2020darts}, or by selecting the operations whose removal has the largest decrease of accuracy in the supernetwork, e.g., DARTS-PT \citep{wang2021rethinking}. Alternatively, DASH \citep{shen2022efficient} learns the size and dilation of convolutional kernels by modelling them as a mixture-of-operations. 
\item \textit{Discrete search space.} Although DARTS and follow-up works utilize a continuous relaxation of the discrete search space to utilize gradient descent, the nature of the search space is still \textit{discrete}. As a consequence, these methods require defining a small subset of possible operations to select from a prori, e.g., convolutions by kernels of size $3\mathrm{x}3$, $5\mathrm{x}5$, $7\mathrm{x}7$. Despite numerous advances in terms of memory consumption \citep{cai2018proxylessnas, hu2020dsnas, Xu2020PC-DARTS:}, the combinatorial nature the discrete space restrict the number of possible operations to be small for DARTS-like methods to work well. As a result, a good initial guess must be provided based on the problem at hand. However, this results in two problems: (\textit{i}) this pre-selection introduces important human biases in the search process --which prevents NAS methods from finding truly novel architectures--, and (\textit{ii}) the need to select a priori a (small) set of options poses challenges in terms of adaptation to new problems and datasets --one of the primary goals of the field of NAS--. It is important to note that in addition to the pre-selection of possible operations, well-performing methods \textit{also require a different backbone architecture depending on the problem at hand}. For instance, DASH \citep{shen2022efficient} uses a TCN backbone \citep{bai2018empirical} for time series classification, and a WRN \citep{zagoruyko2016wide} for image classification.
\end{itemize}

\vspace{-7mm}
\subsubsection{Comparison of DNArch with existing differentiable supernetwork-based methods}
DNArch exhibits several differences to DARTS and follow-up works. To see this, we briefly recap the definition of DNArch and its use in combination with CCNNs \citep{knigge2023modelling}.

DNArch proposes to view neural architectures as continuous multidimensional entities multi-dimensional continuous entities --with dimensions corresponding to depth, width, etc-- and using learnable parametric differentiable masks to learn their size during training. By combining DNArch with a CCNN backbone \citep{knigge2023modelling} --a general purpose convolutional architecture--, DNArch can be used across a broad range of tasks on data of varying lengths and dimensionalities, e.g., classification on $32{\times}32$ images, segmentation on $125{\times}125$ images, and classification on sequences of length $16000$, without the need for structural changes (Sec.~\ref{sec:5_exps}).

\textbf{DNArch operates on a continuous search space.} First of all, by viewing the neural architectures as continuous entities, DNArch \textit{operates on a continuous search space}. As a result, it does not need to consider a small subset of possible operations as in existing works, e.g., convolutional kernels of size $3\mathrm{x}3$, $5\mathrm{x}5$ and $7\mathrm{x}7$, e.g., DASH \citep{shen2022efficient}, or an small set of width multipliers $[0.25, 0.5, 0.75, 1.0]$, e.g., OFA \citep{cai2019once}. In contrast, DNArch is able to consider \textit{all possible values} within an admissible range determined by the maximum size of the supernetwork.\footnote{This supernetwork can be made as big as desired as long as the network itself fits in memory. For clarity, we emphasize that this refers to the size of the network, i.e., the number of parameters used, and not the cost of performing a forward pass.} For example, for a layer with $\mathrm{N}$ channels acting on a dataset of images of size $\mathrm{H}{\times}\mathrm{W}$, DNArch consider all convolutional sizes between $1{\times}1$ and $\mathrm{H}{\times}\mathrm{W}$ as well as all possible width values between $1$ and $\mathrm{N}$ channels. Consequently, for a single layer on CIFAR-10 with $140$ channels, this corresponds to $32{\times}32{\times}140{=}143.360$ possibilities --a much larger number of possibilities than the $3\times4{=}12$ possibilities given in the previous example--. Naturally, this number grows much larger when considering the depth as well as downsampling layers, which are also learned by DNArch in the same manner. For a CCNN$_\mathrm{(8, 280)}$ and a CCNN$_\mathrm{(12, 760)}$ backbone on CIFAR-10, DNArch is able to select from approximately $2.3{\times}10^{12}$ and $5.5{\times}10^{15}$, respectively.

In addition to the much larger search space, using a continuous search space allows DNArch to (\textit{i}) strongly reduce the human biases included in the search process of existing methods and (\textit{ii}) allows for the straightforward application of DNArch to (new) diverse problems and datasets. When looking at existing works, it is notorious that most of the existing works focus only on image classification problems and heavily rely on the design of widely-used neural architectures for the definition of the discrete search space, e.g., \citep{liu2018darts, liang2019darts+, Xu2020PC-DARTS:, hundt2019sharpdarts, guilin2019stacnas, chen2021progressive}. For example, most works either take an image classification ResNet backbone, and either learn the size of convolutional kernels among predefined choices know to work well, e.g., $3\mathrm{x}3$, $5\mathrm{x}5$, $7\mathrm{x}7$, or the width of the network based on a few width multiplier values, e.g., $[0.25, 0.5, 0.75, 1]$. Even most of the works that learn the depth of the network do so by learning the number of layers \textit{within each downsampling block} --using a very small set of options, e.g., $2,3,4$ in OFA \citep{cai2019once}--. As a result, all searchable architectures are \textit{restricted} to architectures that \textit{follow the resolution bottleneck structure of ResNets used for classification}. We emphasize that DNArch \textit{does not have such constraints}. For example, we observe that DNArch finds neural architectures that resemble \textit{concatenated U-Net-like networks for classification problems} (Tabs.~\ref{tab:5_found architectures_lra}, \ref{tab:5_found architectures_2d_ccnn4140}, \ref{tab:5_found architectures_2d}). We note that this kind of architectures are \textit{dramatically different} from standard neural designs used for classification tasks. This highlights the lack of human biases in DNArch, and its ability to find truly novel architectures.

Furthermore, it is important to note that defining a good set of options for the search space prior to the start of the NAS procedure on under-explored tasks, e.g., PDE prediction, long sequences, etc, can be very difficult. Even recent works that tackle the problem of NAS on diverse tasks, e.g., DASH \citep{shen2022efficient}, only use a very restricted number of operations dictated by informed guesses, and \textit{rely on backbone architectures that already encompass much prior information about the problem at hand}, therefore preventing finding truly novel architectures. In contrast, by relying on a continuous search space and a general-purpose backbone, DNArch is able to search thoroughly though a vast space of possible options without the need to use a previously tailored backbone architecture for the task at hand.

\textbf{DNArch does not suffer from operation biases.} DNArch operates on a continuous space, and utilizes differentiable parametric masks to learn the size of the operations. hence, similarly to DASH \citep{shen2022efficient}, DNArch learns operations as a mixture-of-operations of all operations of size smaller or equal to the current size. To visualize this, note that a kernel of size $10{\times}10$ encompasses of kernels of smaller size. Similarly, a layer if width $6$ encompasses all channels within that size, i.e., $1, 2, ..., 6$. As a result, DNArch \textit{does not suffer from operation biases by design}.

\textbf{DNArch does not require discretization / fine-tuning.} Recall that the differentiable masks define the size of the different components of the neural architecture. Since both the differentiable masks as well as the weights of the network are \textit{trained simultaneously}, the search and the tuning of the network weights \textit{happens simultaneously}. Consequently, once the training is finished, the final architecture is already ready to use, i.e., DNArch does not require the discretization and fine-tuning steps required in DARTS. As a consequence, DNArch \textit{does not suffer from poor test generalization issues by design}.

\textbf{DNArch is memory efficient.} Although the supernetwork must fit in memory, DNArch is not required to compute the response of all candidate subnetworks at every forward pass. Hence, DNArch the cost of the forward pass is constant regardless of the number of candidate operations used. As a result, DNArch \textit{does not suffer from the high-memory consumption issues of DARTS by design}.

In conclusion, DNArch encompasses several of the advantages that differentiable network-based methods have. However, thanks to its novel view of the problem of differentiable NAS, it provides several advantages and alleviates several disadvantages observed by existing works by design.

\vspace{-7mm}
\subsection{Constrained and Multi-Objective NAS}
In many settings, optimizing for a single objective, e.g., classification accuracy, is not sufficient. For instance, if a neural network must be deployed on an edge device, we may need to constrain its latency, memory usage or computational requirements. To achieve one or more objectives in addition to accuracy, the standard NAS objective is typically modified to (\textit{i}) a \textit{constrained optimization problem} \citep{bender2018understanding, cai2018proxylessnas, tan2019mnasnet} or (\textit{ii}) a \textit{multi-objective optimization problem} \citep{elsken2018efficient, hu2019efficient, guerrero2021bag}. Most relevant to DNArch is the constrained optimization setting, in which one solves the equation:
\begin{equation}
\setlength{\abovedisplayskip}{4pt}
\setlength{\belowdisplayskip}{2pt}
\min_{a \in \gA} f(a) \ \text{subject to}\ h_i(a) \leq c_i \ \text{for}\ i \in \{1, ..., k\}, 
\end{equation}
where $f(a)$ denotes the original objective function, e.g., classification accuracy, of a network $a$, and $\{h_i\}_{i \in \{1, ...,k\}}$ represents constraints as a function of the architecture. In practice, this problem is solved by transforming it into an additive unconstrained problem of the form:
\begin{equation}
\setlength{\abovedisplayskip}{4pt}
\setlength{\belowdisplayskip}{2pt}
\min_{a \in \gA} f(a) +\sum_i \lambda_i g_i(a); g_i(a){=}\max(0, h_i(a) - c_i), \label{eq:5_constrained_optim}
\end{equation}
with penalty functions $g_i$ penalizing architectures not satisfying the constrains, and hyperparameters $\lambda_i$ trading off the objective and constraints. This problem is subsequently solved using black-box optimization methods or one-shot methods. In the latter case, the penalty functions $g_i$ need to be differentiable, which is \textit{often not the case} for DARTS-like methods. Consequently, such metrics are often relaxed to continuous variables through various techniques, e.g., Gumbel softmax \citep{wu2019fbnet}.

One of the most widely-studied constrained NAS problems is \textit{hardware-aware NAS} \citep{benmeziane2021comprehensive}: a field that concerns itself with hardware efficiency constraints such as memory and latency. While simple metrics such as the number of parameters are easily computed, these often do not correlate enough with other metrics of interest. As such, several works try to approximate hardware metrics of interest either by computing hardware costs as the sum of the cost of each operation \citep{cai2018proxylessnas}, or by using a surrogate model to predict the total cost \citep{dudziak2020brp}.

\textbf{Constrained DNArch.} As shown in the main text (Sec.~\ref{sec:5_constrained_DNArch}), DNArch can also be used under constrained settings by following the method outlined in Eq.~\ref{eq:5_constrained_optim}. Importantly, by using a continuous search space in combination with continuous differentiable masks, DNArch is able to cast constraints on a continuous space based on the size of the masks on a continuum. As a result, DNArch permits the definition of penalty functions $g_i$ which are differentiable with regard to the parameters of the differentiable masks, without the tricks needed in other works, e.g., gumbel softmax \citep{wu2019fbnet}.

Although we focus on computational constraints, we note that DNArch can be easily extended to other types of constraints, as long as these constraints can be defined in terms of the mask sizes.

\vspace{-7mm}
\section{Architectures Found by DNArch}
The architectures found by DNArch in the experiments shown in Sec.~\ref{sec:5_exps}, are shown in Tabs.~\ref{tab:5_found architectures_lra},~\ref{tab:5_found architectures_2d_ccnn4140}, and \ref{tab:5_found architectures_2d}.
\begin{table}
    \centering
    \caption{Architectures found by DNArch on LRA with the target complexity of a CCNN$_4,140$.}
    \label{tab:5_found architectures_lra}
    \begin{small}
    \scalebox{0.8}{
    \begin{tabular}{ccccc}
    \toprule
     \multirow{2}{*}{\sc{Task}} &\multirow{2}{*}{\sc{Depth}} & \multirow{2}{*}{\sc{Kernel Size}} & \multirow{2}{*}{\sc{Resolution}} & \sc{Width} \\
     & & & & [$\Nin$, $\Nt_\mathrm{mid}$, $\Nout$] \\
     \midrule
     \multirow{8}{*}{\rotatebox{90}{\sc{ListOps}}} & \multirow{8}{*}{8} & 266 & 2048 & [150 189 145]\\
     & & 569 & 632 & [150 168 168] \\
     & & 1401 & 1416 & [176 186 162]\\
     & & 310 & 310 & [166 175 153]\\
     & & 213 & 213 & [154 159 163]\\
     & & 12  & 301 & [168 128 162]\\
     & & 5 & 250 & [170 158 153]\\
     & & 24 & 502 & [153 171 165]\\
     \midrule
    \multirow{8}{*}{ \rotatebox{90}{\sc{Text}}} & \multirow{8}{*}{8} &  445 & 2284 & [180 217 205]\\
     & & 691 & 2939 & [208 176 153]\\
     & & 1420 & 1420 & [152 152 120] \\
     & & 415 & 1313 & [120 120 147]\\
     & & 1467 & 1467 & [147 118 135]\\
     & & 52 & 594 & [134 173 153]\\
     & & 101 & 932 & [150 156 183]\\
     & & 149 & 1036 & [180 92 192] \\
    \midrule
    \multirow{8}{*}{\rotatebox{90}{\sc{Retrieval}}} & \multirow{8}{*}{8} & 2 & 1913 & [29 33 172] \\
    & & 136 & 2058 & [184 174 183] \\
    & & 1013 & 2363 & [205 171 161] \\
    & & 1446 & 2724 & [188 164 115] \\
    & & 7 & 2604 & [29 29 163]\\
    & & 1 & 2756 & [29 35 154] \\
    & & 6 & 3545 & [71 110 147]\\
    & & 1 & 3899 & [71 88 137] \\
    \midrule
    \multirow{8}{*}{\rotatebox{90}{\sc{Image}}} & \multirow{8}{*}{8} & 203 & 1024 & [118 155 147]\\
    & & 279 & 1024 & [146 172 164]\\
    & & 219 & 486 & [173 166 196]\\
    & & 308 & 308 & [199 197 196]\\
    & & 144 & 144 & [207 197 92]\\
    & & 8 & 125 & [106 29 75]\\
    & & 30 & 96 & [78 28 110]\\
    & & 40 & 126 & [104 51 104]\\
    \midrule
    \multirow{8}{*}{\rotatebox{90}{\sc{PathFinder}}} & \multirow{8}{*}{8} & 195 & 1024 & [109 140 171]\\
    & & 493 & 770 & [171 168 158] \\
    & & 418 & 507 & [144 183 170]\\
    & & 318 & 318 & [173 187 178]\\
    & & 236 & 236 & [182 162 160]\\
    & & 231 & 231 & [161 121 103]\\
    & & 8 & 251 & [105 47 210]\\
    & & 4 & 253 & [116 29 188]\\
    \midrule
    \multirow{5}{*}{\rotatebox{90}{\sc{Path-X}}} & \multirow{5}{*}{5}  & 2484 & 15331 & [280 174 157] \\
    & & 7204 & 7204 & [177 280 159]\\
    & & 3669 & 3772 & [167 280 98]\\
    & & 2323 & 5496 & [123 164 164]\\
    & & 513 & 4768 & [136 128 195]\\
    \bottomrule
    \end{tabular}}
    \end{small}
\end{table}

\begin{table}
    \centering
    \caption{Architectures found by DNArch on 2D datasets with the target complexity of a CCNN$_{4,140}$.}
    \label{tab:5_found architectures_2d_ccnn4140}
    \begin{small}
    \scalebox{0.8}{
    \begin{tabular}{ccccc}
    \toprule
     \multirow{2}{*}{\sc{Task}} &\multirow{2}{*}{\sc{Depth}} & \sc{Kernel Size} & \sc{Resolution} & \sc{Width} \\
     & & [\sc{y} \sc{x}] & [\sc{y} \sc{x}] & [$\Nin$, $\Nt_\mathrm{mid}$, $\Nout$] \\
     \midrule
    \multicolumn{5}{c}{\sc{Image Classification Tasks}} \\
    \midrule
    \multirow{8}{*}{\rotatebox{90}{\sc{Cifar10}}} & \multirow{8}{*}{8} & [9 7]  & [32 32] & [142 139 145]\\
    & & [12 8] & [32 32] & [145 160 157]\\
    & & [25 7] & [32 20] & [158 186 182] \\
    & & [9 10] & [9 15] & [186 208 168]\\
    & & [1 13] & [5 15] & [169 177 150] \\
    & & [1 10] & [6 11] & [151 139 156]\\
    & & [5 1] & [15 4] & [154 115 110] \\
    & & [6 5] & [11 7] & [108 41 166]\\
    \midrule
    \multirow{8}{*}{\rotatebox{90}{\sc{Cifar100}}} & \multirow{8}{*}{8} & [13 7] & [32 32] & [104 107 116] \\
    & & [6 10] & [32 32] & [114 134 134]\\
    & & [11 8] & [22 22] & [139 192 166] \\
    & & [13 7] & [16 18] & [173 201 197]\\
    & & [8 12] & [10 12] & [205 251 51]\\
    & & [1 1] & [8 9] & [62 56 157] \\
    & & [5 9] & [8 10] & [162 175 254] \\
    & & [8 7] & [9 9] & [280 280 280] \\
    \midrule
        \multicolumn{5}{c}{\sc{Dense Tasks}} \\
    \midrule
    \multirow{3}{*}{\rotatebox{90}{\makecell{\sc{Darcy}\\ \sc{Flow}}}}& \multirow{3}{*}{3} & [43 38] & [80 72] & [156 280 107] \\
    & & [22 22] & [22 22] & [180 204 78] \\
    & & [76 76] & [85 85] & [280 280 50] \\
    \midrule
    \multirow{6}{*}{\rotatebox{90}{\sc{Cosmic}}} & \multirow{6}{*}{6} & [94 111] & [128 128] & [18 110 23] \\
    & & [2 13] & [20 45] & [186 207 139] \\
    & & [129 129] & [129 129] & [126 265 100] \\
    & & [129 121] & [129 129] & [78 105 59] \\
    & & [90 89] & [129 129] & [57 201 197] \\
    & & [76 74] & [76 74] & 202 145 216] \\
    \bottomrule
    \end{tabular}}
    \end{small}
\end{table}

\begin{table}
    \centering
    \caption{Architectures found by DNArch on 2D datasets with the target complexity of a CCNN$_{6,380}$.}
    \label{tab:5_found architectures_2d}
    \begin{small}
    \scalebox{0.8}{
    \begin{tabular}{ccccc}
    \toprule
     \multirow{2}{*}{\sc{Task}} &\multirow{2}{*}{\sc{Depth}} & \sc{Kernel Size} & \sc{Resolution} & \sc{Width} \\
     & & [\sc{y} \sc{x}] & [\sc{y} \sc{x}] & [$\Nin$, $\Nt_\mathrm{mid}$, $\Nout$] \\
     \midrule
    \multicolumn{5}{c}{\sc{Image Classification Tasks}} \\
    \midrule
    \multirow{12}{*}{\rotatebox{90}{\sc{Cifar10}}} & \multirow{12}{*}{12} & [4 7]  & [32 32] & [380 328 384]\\
    & & [9 10] & [32 32] & [384 371 393]\\
    & & [12 6] & [32 32] & [392 361 391] \\
    & & [20 6] & [32 32] & [388 370 421]\\
    & & [10 11] & [23 26] & [421 417 486] \\
    & & [11 11] & [12 22]  & [496 444 479]\\
    & & [1 11] & [6 11] & [493 482 304] \\
    & & [1 6] & [5 21] & [211 78 384]\\
    & & [29 4] & [32 4] & [363 459 280] \\
    & & [18 15] & [18 15] & [277 394 67]\\
    & & [1 1] & [4 4] & [111 109 361] \\
    & & [4 3] & [21 15] & [121 374 449]\\
    \midrule
    \multirow{12}{*}{\rotatebox{90}{\sc{Cifar100}}} & \multirow{12}{*}{12} & [8 9]  & [32 32] & [343 275 354]\\
    & & [12 10] & [32 32] & [351 316 397]\\
    & & [11 10] & [32 32] & [495 355 420] \\
    & & [18 12] & [29 21] & [421 498 419]\\
    & & [11 15] & [27 24] & [432 449 407] \\
    & & [19 8] & [25 20]  & [412 419 413]\\
    & & [11 10] & [12 23] & [423 454 600] \\
    & & [8 8] & [8 9] & [709 685 416]\\
    & & [5 7] & [5 8] & [419 311 446] \\
    & & [8 4] & [8 4] & [446 433 389]\\
    & & [6 4] & [6 4] & [386 501 570] \\
    & & [8 9] & [8 9] & [568 453 655]\\
    \midrule
        \multicolumn{5}{c}{\sc{Dense Tasks}} \\
    \midrule
    \multirow{7}{*}{\rotatebox{90}{\makecell{\sc{Darcy}\\ \sc{Flow}}}}& \multirow{7}{*}{7} & [54 49] & [54 49] & [435 428 289] \\
    & & [43 47] & [70 72] & [499 393 284] \\
    & & [65 69] & [85 85] & [496 434 281] \\
    & & [67 66] & [85 85] & [323 412 275] \\
    & & [85 85] & [85 85] & [319 369 271] \\
    & & [85 85] & [85 85] & [306 379 258] \\
    & & [68 68] & [85 85] & [521 435 271] \\
    \midrule
    \multirow{12}{*}{\rotatebox{90}{\sc{Cosmic}}} & \multirow{12}{*}{12} & [35 32] & [35 33] & [146 236 272] \\
    & & [11 21] & [95 72] & [170 284 319] \\
    & & [44 24] & [128 128] & [141 339 388] \\
    & & [23 41] & [128 128] & [385 407 361] \\
    & & [28 27] & [128 128] & [351 279 356] \\
    & & [21 19] & [128 128] & [354 362 310] \\
    & & [29 24] & [128 128] & [310 351 466]] \\
    & & [18 25] & [128 128] & [396 292 183] \\
    & & [57 16] & [128 128] & [179 210 580] \\
    & & [50 11] & [127 77] & [273 250 63] \\
    & & [18 12] & [89 67] & [347 400 77] \\
    & & [22 23] & [97 79] & [171 241 79] \\
    \bottomrule
    \end{tabular}}
    \end{small}
\end{table}

% \textbf{IMDB.} Depth: [8.] Kernel: [[205.][260.][322.][372.][4097.][1832.][1390.][335.][286.][286.][286.][286.]] Spectral: [[2909.][3641.][3439.][2948.][3694.][3758.][3771.][3848.][4097.][4097.][4097.][4097.]] Channel: [[385. 395. 392.][390. 378. 393.][393. 351. 364.][367. 303. 331.][329. 330. 349.][350. 290. 365.][357. 304. 392.][378. 361. 406.][388. 388. 388.][388. 388. 388.][388. 388. 388.][388. 388. 388.]]

% \textbf{Image.} Depth: [7.] Kernel: [[57.][223.][221.][253.][924.][1025.][1025.][106.][106.][106.][106.][106.]] Spectral: [[1025.][1025.][1025.][813.][508.][309.][192.][907.][907.][907.][907.][907.]] Channel: [[192. 513. 340.][346. 391. 346.][345. 434. 391.][398. 497. 494.][497. 570. 483.][488. 571. 437.][443. 449. 484.][397. 397. 397.][397. 397. 397.][397. 397. 397.][397. 397. 397.][397. 397. 397.]]

% \textbf{ListOps.} Depth: [12.] Kernel: [[161.][634.][1174.][1377.][2049.][1544.][1350.][2049.][41.][10.][35.][39.]] Spectral: [[2049.][1733.][1085.][1142.][1298.][1342.][1353.][1361.][1311.][1424.][1561.][1362.]] Channel: [[345. 397. 390.][335. 316. 372.][371. 324. 373.][372. 349. 339.][338. 290. 289.][288. 289. 348.][348. 294. 350.][349. 351. 399.][172. 180. 396.][ 42.  63. 394.][343. 281. 391.][392. 349. 401.]]
\vspace{-7mm}
\section{Learning downsampling in the spatial Domain}\label{appx:5_downsample_no_fourier}
Differentiable Masking learns downsampling by multiplying the spectrum $\tilde{f}{=}\gF[f]$ of a signal $f$ with a differentiable mask $m(\cdot\ ; \boldsymbol{\theta})$, and cropping the output above the cutoff frequency of the mask $\omega_\mathrm{max}$ next. However, it is not strictly necessary to perform this operation in the Fourier domain. An equivalent downsampling can also be learned directly in the spatial domain.

The \textit{Fourier convolution theorem} states that the spatial convolution is equivalent to a pointwise multiplication in the Fourier domain. However, this equivalence works in both directions. That is, we can equivalently say that the pointwise multiplication on the Fourier domain is equal to a convolution on the spatial domain. Consequently, we can represent the pointwise multiplication of the spectrum of the input $\gF[f]$ and the differentiable mask $m(\cdot\ ; \boldsymbol{\theta})$ as the convolution of their inverse Fourier transforms. Formally:
\begin{align}
    \tilde{f} \cdot m(\cdot\ ; \boldsymbol{\theta}) &= \gF\left[ \gF^{-1} \left[\tilde{f} \right] * \gF^{-1}[m(\cdot\ ; \boldsymbol{\theta})] \right] \nonumber \\
   & = \gF\left[ \gF^{-1} \left[\gF[f] \right] * \gF^{-1}[m(\cdot\ ; \boldsymbol{\theta})] \right] \nonumber \\
    & =  \gF\left[f* \gF^{-1}[m(\cdot\ ; \boldsymbol{\theta})] \right] \label{eq:5_downsampling_spatial}
\end{align}
In other words, we can perform the same operation in the spatial domain by convolution the original input signal $f$ with the inverse Fourier transform of the mask $m(\cdot\ ; \boldsymbol{\theta})$.

\textbf{Defining the output resolution.} Eq.~\ref{eq:5_downsampling_spatial} defines how low-pass filtering can be performed on the spatial domain, but it does not provide information regarding the final resolution of the operation. To derive the resolution of the output, we can simply use Eqs.~\ref{eq:5_size_gauss},~\ref{eq:5_size_sigm} to analytically derive the size of the mask. Once the size of the mask is derived, we can simply take the downsampled signal --after using Eq.~\ref{eq:5_downsampling_spatial}--, and downsample it to match the size of the mask.

\vspace{-7mm}
\section{Computational complexity of masked network components}\label{appx:5_masking_other_layers}
Here, we derive the computational complexity of all layers used in the CCNN architecture with and without the use of masks. The calculation of these complexities follows the same reasoning as the pointwise linear layer provided as example in the main text.

With $\Lt$, $\Nin$ and $\Nout$ the length, number of input channels and number of output channels of a given layer, and $\mathrm{size}(m_\mathrm{res})$, $\mathrm{size}(m_\Nin)$, $\mathrm{size}(m_\Nout)$ the size of the masks along the corresponding dimensions, the complexity of the layers used in the CCNN architectures are given by:

\textbf{Pointwise linear layer:}
\begin{align*}
&\gC_{\mathrm{lin}}(f) = \Lt \cdot \Nin \cdot \Nout \\
&  \gC_{\mathrm{lin, masked}} =  \mathrm{size}(m_\mathrm{res}) \cdot \mathrm{size}(m_\Nin) \cdot \mathrm{size}(m_\Nout)
\end{align*}
\textbf{Fourier convolution:}
\begin{align*}
    &\gC_{\gF\mathrm{conv}} = \Lt  \log \left( \Lt \right) \\
    &\gC_{\gF\mathrm{conv,masked}} = \mathrm{size}(m_\mathrm{res}) \log \left(\mathrm{size}(m_\mathrm{res})\right)
\end{align*}
\textbf{Pointwise operations --GELU, DropOut, etc.--:}
\begin{align*}
    &\gC_\mathrm{pointwise} = \Lt \cdot \Nin \\
    &\gC_\mathrm{pointwise, masked} = \mathrm{size}(m_\mathrm{res})\cdot \mathrm{size}(m_\Nin)
\end{align*}
% \textbf{GELU:}
% \begin{align*}
%     &\gC_\mathrm{GELU} = \Lt \cdot \Nin \\ 
%     &\gC_\mathrm{GELU, masked} = \mathrm{size}(m_\mathrm{res})\cdot \mathrm{size}(m_\Nin)
% \end{align*}
% \textbf{DropOut:}
% \begin{align*}
%     &\gC_\mathrm{DropOut} = \Lt \cdot \Nin \\ 
%     &\gC_\mathrm{DropOut, masked} = \mathrm{size}(m_\mathrm{res})\cdot \mathrm{size}(m_\Nin)
% \end{align*}

\vspace{-7mm}
\section{Dataset descriptions}\label{appx:5_datset_description}
\vspace{-5mm}
\subsection{The Long Range Arena benchmark}\label{appx:5_long_range_arena}
The Long Range Arena \cite{tay2021long} consists of six sequence modelling tasks with sequence lenghts ranging from 1024 to over 16000. It encompasses modalities and objectives that require similarity, structural, and visiospatial reasoning. We follow the data preprocessing steps of \citet{gu2022efficiently}, which we also include here for completeness.

\textbf{ListOps.} An extended version of the dataset presented by \citet{nangia2018listops}. The task involves computing the integer result in the range zero to nine of a mathematical expression represented in prefix notation with brackets, e.g., [MAX29[MIN47]0] $\rightarrow$ 9. Characters are encoded as one-hot vectors, with 17 unique values possible (opening brackets and operators are grouped into a single token). The sequences are of unequal length. Hence, the end of shorter sequences is padded with a fixed indicator value to a maximum length of 2048. The task has 10 different classes
representing the possible integer results of the expression. It consists of 96{\sc{k}} training sequences, 2{\sc{k}}
validation sequences, and 2{\sc{k}} test sequences. No data normalization is applied.

\textbf{Text.} Based on the IMDB sentiment analysis dataset presented by \citet{maas2011learning}, the task is to classify movie reviews as having a positive or negative sentiment. The reviews are presented as a sequence of 129 unique integer tokens padded to a maximum length of 4096. The dataset contains 25{\sc{k}} training sequences and 25{\sc{k}} test sequences. No validation set is provided. No data normalization is applied.

\textbf{Retrieval.} Based on the ACL Anthology network corpus presented by \citet{radev2013acl}, the task is to classify whether two given textual citations are equivalent. To accomplish this, each citation is separately passed through an encoder, and passed to a final classifier layer. Denoting $X_1$ as the encoding for the first document and $X_2$ as the encoding for the second document, four features are created and concatenated together as:
\begin{equation*}
    X = [X_1, X_2, X_1 \times X_2, X_1 - X_2],
\end{equation*}
which are subsequently passed to a two layered MLP. The goal of the task is to evaluate how well the network can represent the text by evaluating if the two citations are equivalent or not. Characters are encoded into a one-hot vector with 97 unique values and sequences are padded to a maximum length of 4000. The dataset includes 147.086 training pairs, 18.090 validation pairs, and 17.437 test pairs. No normalization is applied.

\textbf{Image.} The Image task uses $32{\times}32$ images of the CIFAR10 dataset \cite{krizhevsky2009learning}. It views the images as sequences of length 1024 that correspond to a one-dimensional raster scan of the image. There are a total of 10 classes, 45{\sc{k}} training examples, 5{\sc{k}} validation examples and 10{\sc{k}} test examples. The RGB pixel values are converted to grayscale intensities and then normalized to have zero mean and unit variance across the entire dataset.

\textbf{PathFinder.} Based on the PathFinder challenge introduced by \citet{linsley2018learning}, the task presents a $32{\times}32$ grayscale image with an start and an end point depicted as small circles. The task is to classify whether there is a dashed line (or path) joining the start and end points while presenting the input as a one-dimension raster scan of the image, alike the {\tt Image} task. The dataset includes 160{\sc{k}} training examples, 20{\sc{k}}validation examples and 20{\sc{k}} test examples. The input data is normalized to be in the range [-1, 1].  

\textbf{Path-X.} Path-X is an \enquote{extreme} version of the {\tt PathFinder} dataset, in which the input images are of size $128{\times}128$. As a result, the input sequences are sixteen times longer with a total length of $16384$. Aside from this difference, the task is identical to the {\tt PathFinder} dataset.

\vspace{-7mm}
\subsection{Image classification datasets}
\textbf{CIFAR10 and CIFAR100.} The CIFAR10 dataset \cite{krizhevsky2009learning}
consists of 60{\sc{k}} real-world 32${\times}$32 RGB images uniformly drawn from 10 classes divided into training
and test sets of 50{\sc{k}} and 10{\sc{k}} samples, respectively. The CIFAR100 dataset \cite{krizhevsky2009learning}
is similar to the CIFAR10 dataset, with the difference that the images are uniformly drawn from
100 different classes. For validation purposes, we divide the training dataset of both CIFAR10 and CIFAR100
into training and validation sets of 45{\sc{k}} and 5{\sc{k}} samples, respectively. Both datsets are normalized to have zero mean and unit variance across the entire dataset.

\vspace{-7mm}
\subsection{NAS-Bench-360}
NAS-Bench-360 \cite{tu2022bench} is a benchmark suite to evaluate Neural Architecture Search methods beyond image classification. The benchmark is composed of ten tasks spanning a diverse array of application domains, datset sizes, problem dimensionalities, and learning objectives. In this work, we consider two tasks from the NAS-Bench-360 suite which require dense predictions: The {\tt DarcyFlow} \cite{li2020fourier} and {\tt Cosmic} \cite{zhang2020deepcr} datasets.

\textbf{DarcyFlow: Solving Partial Differential Equations.} DarcyFlow aims to solve Partial Differential Equation (PDE) by using neural networks as a replacement for traditional solvers. The input for this task is a $85{\times}85$ grid specifying the initial conditions and coordinates of a fluid, and the output is a 2D grid of the same dimensions representing the fluid state at a later time. The ground truth for this task is the result computed by a traditional solver, and the objective is to minimize the Mean Squared Error (MSE) between the predicted fluid state and the ground truth.

\textbf{Cosmic: Identifying Cosmic Ray Contamination.} The Cosmic task involves identifying and masking corruption caused by charged particles collectively referred to as \enquote{cosmic rays} on images taken from space-based facilities. It uses imaging data of local resolved galaxies collected from the Gubble Space Telescope. The input is an $128{\times}128$ image corresponding to the artifact of cosmic rays, and the output is a 2D grid of the same dimensions predicting whether each pixel in the input is an artifact of cosmic rays. We report the false-negative rate of identification results.

\vspace{-7mm}
\section{Experimental details}\label{appx:5_exp_details}
\vspace{-5mm}
\subsection{General remarks}
\textbf{Code.} Our code is written in {\tt JAX} and our experiments are conducted on {\tt TPUs} and {\tt GPUs}. As outlined in the Limitations (Sec.~\ref{sec:5_limitations}), {\tt JAX} and {\tt TPU} training prevent DNArch from performing operations that change the dimensions of arrays during training. In addition to our {\tt JAX} implementation, we release a {\tt PyTorch} implementation of DNArch that supports these operations. This implementation makes DNArch more flexible and accessible, especially in scenarios where it is crucial to keep candidate networks close to the target complexity during the course of training.

\textbf{The Continuous CNN and the CCNN residual block.} The CCNN architecture is shown in Fig.~\ref{fig:5_ccnn_architecture}. It is composed by an {\tt Encoder}, a {\tt Decoder}, and a number of CCNN residual blocks {\tt ResBlock} \cite{knigge2023modelling}. The {\tt Encoder} is defined as a sequence of [{\tt PWLinear}, {\tt BatchNorm} \cite{ioffe2015batch}, {\tt GELU} \cite{hendrycks2016gaussian}] layers. For tasks dealing with text, we additionally utilize an Embedding layer mapping each token in the vocabulary to a vector representation of length equal to that used by \citet{gu2022efficiently} (see Appx.~\ref{appx:5_long_range_arena}).
For dense prediction tasks, the {\tt Decoder} is a {\tt{PWLinear}} layer, which is preceded by {\tt GlobalAvgPooling} for global prediction tasks.

\textbf{Batch Normalization in DNArch.} As the architecture is constantly changing during the search process, we use batch-specific statistics for batch normalization instead of the global moving average. This approach was adopted after early experiments showed that using the global moving average leads to a significant discrepancy in the behavior of the validation and training curves. Specifically, we observed that while the training curves were converging to a good solution, the validation curves resembled random predictions. This issue was resolved by deactivating the global moving average in Batch Normalization layers.

\textbf{Continuous convolutional kernels \mlp$_{\psi}$.} We parameterize our convolutional kernels as a 4-layer MLP with 128 hidden units and a Fourier Encoding \citep{tancik2020fourier} of the form $\gamma(\xv)=[\cos(2\pi \omega_0 \mat{W} \xv]), \sin(2\pi \omega_0 \mat{W} \xv)]$, with $\Wm \in \sR^{\mathrm{D}\times 128}$ and $\omega_0$ a hyperparameter that acts as a prior on the frequency content of the kernels \cite{romero2022ckconv, sitzmann2020implicit}. In contrast to \citet{romero2022ckconv}, we utilize a single larger \mlp$_{\psi}$ to generate the kernels of the entire network. This allows the network \mlp$_{\psi}$ to administrate its capacity across all layers. Using different \mlp$_{\psi}$'s for each layer as \citet{romero2022ckconv} is inadequate in the learnable architectures setting as some layers can be entirely erased. With our proposed solution, the capacity of the otherwise zeroed-out \mlp$_{\psi}$ is used to generate the kernels of the remaining layers.

\textbf{Normalized relative positions.} Following \citet{romero2022ckconv, romero2022flexconv}, we normalize the coordinates going into \mlp$_{\psi}$ to lie in the space $[-1, 1]^{\Dt}$ for $\Dt$-dimensional kernels.

\textbf{Parameters and hyperparameters of the differentiable masks.} We learn some of the parameters of the masks, and leave the others constant or treat them as a hyperparameter. Specifically, for Gaussian masks, we only learn their width, i.e., $\sigma$, and fix its mean to zero. For Sigmoid masks, we learn their offset $\mu$ and treat their temperature $\tau$ as a hyperparameter. For more information regarding the values of $\tau$ used in our parameter tuning step, please refer to Appx.~\ref{appx:5_hyperparams}.

\textbf{Maximum and minimum allowable sizes for the learnable differentiable masks.} We define some minimum and maximum allowable sizes for the mask parameters, and reset them to these values after each training iteration if the updated parameter values lie outside that range. For the Gaussian mask, we constraint the minimum admissible value of $\sigma$ such that the length of the corresponding dimension never collapses to a value of $1$. This is to prevent the corresponding dimension to collapse such that it can grow afterwards if required. The minimum value depends on the resolution of the corresponding dimension, e.g., the maximum size of the convolutional kernel, and can be easily calculated with Eq.~\ref{eq:5_inv_gauss_mask}.

Note that the offset value of the Sigmoid mask $\mu$ could in principle assume any value in $\sR$. However, if not controlled, $\mu$ could become too small and mask all values along a particular dimension to zero. Similarly, if $\mu$ is too large, the gradient of the mask at all positions would become very small and it would difficult to update the mask. To avoid these situations, we define minimum and maximum values of $\mu$ such that for the lowest value, the mask at the lowest position is equal to $0.95$, and for the largest value, the mask at the highest position is equal $0.85$. These values are dependent on the value of the mask temperature $\tau$, and can be easily calculated with Eq.~\ref{eq:5_inv_sigm_mask}.

\textbf{Limiting the size of the mask to the maximum allowable ranges.} As outlined in the Limitations (Sec.~\ref{sec:5_limitations}), we must set a maximum allowable size for the width and depth of the network on {\tt JAX}. However, the maximum allowed value for the parameters of the masks (see previous paragraph) allows both masks to grow beyond the point on which the theoretical size of the masks is equal to the maximum allowable network size. For instance, for the maximum allowed parameter values of a Sigmoid mask, the last channel, i.e., the $280$-th channel, would be weighted by a factor of $0.85$. Consequently, the theoretical size of the mask as calculated by Eq.~\ref{eq:5_size_sigm} will be well beyond $280$. This value would lead to an unrealistic theoretical computational complexity that surpasses the real computational complexity the CCNN used.
\begin{table*}
    \centering
    \caption{Hyperparameters used for the experiments with the target complexity of a CCNN$_{4, 140}$.}
    \label{tab:5_hyperparams}
    \vspace{-2.5mm}
    \begin{small}
    \scalebox{0.8}{
    \begin{tabular}{lcccccccccc}
    \toprule
    \multirow{2}{*}{\sc{Dataset}} & \multirow{2}{*}{\sc{Epochs}} & \sc{Batch} & \sc{Learning} & \multirow{2}{*}{\sc{DropOut}} & \sc{Weight} & \multirow{2}{*}{$\omega_0$} & \multirow{2}{*}{$\lambda$} & \multirow{2}{*}{$\tau_\mathrm{resolution}$} & \multirow{2}{*}{$\tau_\mathrm{channel}$} & \multirow{2}{*}{$\tau_\mathrm{depth}$}\\
    & & \sc{Size} & \sc{Rate} & & \sc{Decay}\\
     \midrule
     \sc{ListOps} & 50 & 50 & 0.005 & 0.0 & 0.01 & 27.5{\sc{k}} & 5.0 & 50 & 25 & 8 \\
     \sc{Text}  & 100 & 50 & 0.02 & 0.2 & 0.01 & 19.5{\sc{k}} & 0.1 & 50 & 25 & 8 \\
     \sc{Retrieval} & 50 & 50 & 0.001 & 0.1 & 0.01 & 21.5{\sc{k}}  & 0.1 & 50 & 25 & 8 \\
     \sc{Image} & 210 & 50 & 0.02 & 0.1 & 0.001 & 12.5{\sc{k}} & 0.1 & 25 & 25 & 8 \\
     \sc{PathFinder} & 210 & 50 & 0.005 & 0.0 & 0.001 & 21.5{\sc{k}} & 0.1 & 50 & 25 & 8  \\
     \sc{Path-X} & 80 & 32 & 0.001 & 0.0 & 0.0 & 30{\sc{k}} & 0.1 & 100 & 25 & 8\\
     \midrule
     \sc{CIFAR10} & 210  & 50 & 0.01 & 0.1 & 0.01 & 21.5{\sc{k}} & 0.1 & 25 & 25 & 8 \\
     \sc{CIFAR100} & 210 & 50 & 0.01 & 0.0 & 0.01 & 6.5{\sc{k}} & 5.0 & 25 & 25 & 8\\
     \midrule
     \sc{DarcyFlow} & 310 & 8 & 0.02 & 0.0 & 0.0001 & 24.5{\sc{k}} & 0.1 & 50 & 25 & 8 \\
     \sc{Cosmic} & 310 & 8 & 0.02 & 0.3 & 0.0001 &5.5{\sc{k}} & 0.1 & 100 & 25 & 8 \\
    \bottomrule
    \end{tabular}}
    \end{small}
    \vspace{-4mm}
\end{table*}

\begin{table*}
    \centering
    \caption{Hyperparameters used for the experiments with the target complexity of a CCNN$_{6, 380}$.}
    \label{tab:5_hyperparams}
    \vspace{-2.5mm}
    \begin{small}
    \scalebox{0.8}{
    \begin{tabular}{lcccccccccc}
    \toprule
    \multirow{2}{*}{\sc{Dataset}} & \multirow{2}{*}{\sc{Epochs}} & \sc{Batch} & \sc{Learning} & \multirow{2}{*}{\sc{DropOut}} & \sc{Weight} & \multirow{2}{*}{$\omega_0$} & \multirow{2}{*}{$\lambda$} & \multirow{2}{*}{$\tau_\mathrm{resolution}$} & \multirow{2}{*}{$\tau_\mathrm{channel}$} & \multirow{2}{*}{$\tau_\mathrm{depth}$}\\
    & & \sc{Size} & \sc{Rate} & & \sc{Decay}\\
     % \midrule
     % \sc{ListOps} & 100 & 48 & 0.001 & 0.3 & 0.01 & 18.5{\sc{k}} & 5.0 & 50 & 50 & 8 \\
     % \sc{Text}  & 100 & 32 & 0.001 & 0.0 & 0.0 & 18.5{\sc{k}} & 0.1 & 25 & 50 & 16 \\
     % \sc{Retrieval} &  \\
     % \sc{Image} & 210 & 50 & 0.005 & 0.2 & 0.001 & 12{\sc{k}} & 0.1 & 25 & 50 & 16 \\
     % \sc{PathFinder} &  \\
     % \sc{Path-X} & \\
     \midrule
     \sc{CIFAR10} & 210 & 50 & 0.005 & 0.0 & 0.01 & 21.5{\sc{k}} & 0.1 & 50 & 50 & 16 \\
     \sc{CIFAR100} & 210 & 50 & 0.01 & 0.0 & 0.01 & 6.5{\sc{k}} & 0.1 & 25 & 25 & 8 \\
     \midrule
     \sc{DarcyFlow} & 310 & 12 & 0.01 & 0.2 & 0.0 & 7.5{\sc{k}} & 0.1 & 50 & 25 & 8\\
     \sc{Cosmic} & 310 & 4 & 0.01 & 0.2 & 0.01 & 5.5{\sc{k}} & 0.1 & 50 & 50 & 8\\
    \bottomrule
    \end{tabular}}
    \end{small}
\end{table*}

To overcome this issue, we limit the maximum size of the mask calculated by Eq.~\ref{eq:5_size_sigm} to be less or equal than the maximum allowable size, e.g., $\mathrm{size}(m){=}\min(\mathrm{size}(m), 280)$. It is important to note, however, that clipping the value of $\mathrm{size}$ directly would stop the gradient flow for parameter values leading to sizes larger 280. As a result, once the maximum size is reached, the mask would not be able to contract anymore. We avoid gradient flow stop by using clipping in combination with a straight-through estimator \citep{bengio2013estimating}. As a result, we are able to propagate the gradient across the clipping operation, and the resulting mask can still be modified even in the cropping operation is used.

\vspace{-7mm}
\subsection{Hyperparameters and training configurations} \label{appx:5_hyperparams}
In this section, we include more information about the found hyperparameters, the values that were considered during hyperparameter tuning, and other training settings. The final hyperparameters used are listed in Table~\ref{tab:5_hyperparams}.

\textbf{Optimizer, learning rates and learning rate schedule.} All our models are optimized with AdamW \cite{loshchilov2018decoupled} in combination with a cosine annealing learning rate scheduler \cite{loshchilov2017sgdr}, and a linear learning rate warm-up stage of 10 epochs, except for {\tt ListOps}, {\tt Retrieval} and {\tt Path-X} for which we have a warm-up stage of 5 epochs.

\textbf{Regularization.} We utilize dropout \cite{srivastava14a} --as shown in Fig.~\ref{fig:5_ccnn_architecture}-- as well as weight decay during training.

\vspace{-7mm}
\subsubsection{Hyperparameter tuning}

\textbf{Frequency prior $\omega_0$.} The possible $\omega_0$ values explored in this work are:\newline $[1, 500, 1500, 2500, ... 28500, 29500, 30000]$.

\textbf{Tuning the value of $\lambda$.} $\lambda$ plays the role of controlling the weight of the computational loss $\gL_\mathrm{comp}$ relative to the task objective loss $\gL_\mathrm{obj}$. In this work, we find two settings which require different values of $\lambda$. One, given by the tasks that converge to a low prediction values relative to perfection, i.e., {\tt ListOps} and {\tt CIFAR100}, and for which the loss $\gL_\mathrm{obj}$ remains relatively high at the end of training. The other group is given by all the other tasks, which converge to high prediction values --many even obtaining a perfect accuracy on the train set--, and for which $\gL_\mathrm{obj}$ converges to values close to zero. For the first group, we require a higher value of $\lambda$ such that the computational complexity loss $\gL_\mathrm{comp}$ remains relevant to the optimization objective. The final values of $\lambda$ used are $5.0$ and $0.1$, respectively.

\textbf{Tuning the temperature of the Sigmoid masks $\tau$.} For the resolution mask, we consider three values of $\tau$, $[25, 50, 100]$ which correspond to a minimum size of $10\%$, $5\%$ and $2.5\%$ of the corresponding dimension. For the channel mask, we consider two values $\tau\in[25, 50]$ which correspond to a minimum size of $10\%$ and $5\%$ of the corresponding dimension, but observe early during tuning that models prefer $\tau{=}25$. For the depth dimension, which is much more sparse than the channel and resolution dimensions, we consider two values $\tau\in[8, 16]$, which result on a minimum depth of $2$ and $1$ layers, respectively. We observe early during tuning that models prefer $\tau{=}8$.

\textbf{Learning rate.} The possible learning rate values explored in this work are:\newline $[0.0001, 0.0005, 0.001, 0.005, 0.01, 0.02]$.

\textbf{Dropout.} The possible dropout values explored in this work are: $[0.0, 0.1, 0.2, 0.3]$.

\textbf{Weight decay.} The possible weight decay values considered in this work are:\newline $[0.0, 0.0001, 0.001, 0.01, 0.05]$.

%% file: backmatter/appx_co_attentive.tex
% path to figures directory
\graphicspath{{figures/chapter-5/}}

%=========================================================================

% \begin{savequote}[75mm]
% Nulla facilisi. In vel sem. Morbi id urna in diam dignissim feugiat. Proin molestie tortor eu velit. Aliquam erat volutpat. Nullam    ultrices, diam tempus vulputate egestas, eros pede varius leo.
% \qauthor{Quoteauthor Lastname}
% \end{savequote}

\chapter{Focusing Equivariance on Transformations Co-Occurring in Data}
  %\label{chapter:}

%=========================================================================

\vspace{-7mm}
\section{Obtaining Co-Occurrent Attention via Equation \ref{eq:10_att_conv_local}}\label{sec:10_appendix}
\begin{figure}
    \centering
    \includegraphics[width=0.7\linewidth]{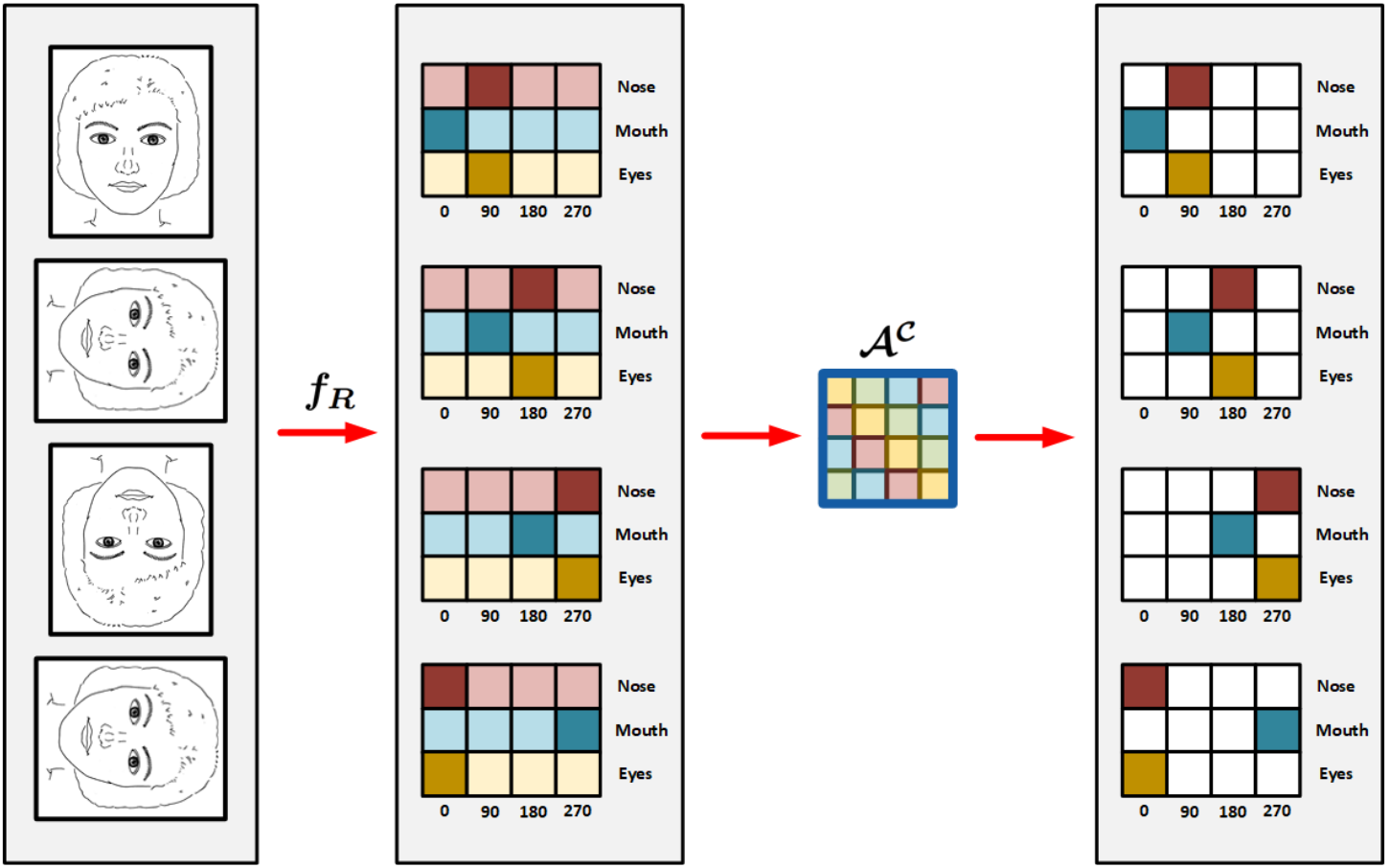}
    \caption{Synchronous movement of feature mappings and attention masks as a function of input rotation in the group $p4$ ($r_{\text{max}} = 4$).}
    \label{fig:10_my_label}
\end{figure}

In this section, we provide a meticulous description on how co-occurrent attention is obtained via the method presented in the paper. Intuitively, a direct approach to address the problem illustrated in the introduction (Section \ref{sec:10_intro}) and Figure \ref{fig:10_fig} requires an attention mechanism that acts simultaneously on $r$ and $\lambda$ (see Eq. \ref{eq:10_rot_conv_1}). However, we illustrate how the depicted problem can be simplified such that attention along $r$ is sufficient by taking advantage of the equivariance property of the network.

Let $p$ be the input of a roto-translational convolution $f_{R}: \sZ^{2} \times \Theta \times \Lambda_{0}\rightarrow \sZ^{2}  \times \Theta \times \Lambda_{1}$ as defined in Eq. \ref{eq:10_rot_conv_1}, and $\Theta$ be the set of rotations by $\theta_{r}$ degrees: $\Theta=\{\theta_{r} = r \frac{2\pi}{r_{\text{max}}}\}_{r=1}^{r_{\text{max}}}$. Let $f_{R}(p)(u) \in \sR^{r_{\text{max}} \times \Lambda_{1}}$ be the matrix consisting of the $r_{\text{max}}$ oriented responses for each $\lambda \in \Lambda_{1}$ learned representation at a certain position $u$. 
Since the vectors $f_{R}(p)(u, \lambda) \in \sR^{r_{\text{max}}}$, $\lambda \in \Lambda_{1}$ permute cyclically as a result of the rotation equviariance property of $f_{R}$, it is mandatory to ensure equivariance to cyclic permutations for each $f_{R}(p)(u, \lambda)$ during the course of the attention procedure (see Section \ref{sec:10_co_att_nns}). 
%Let $p$ be the input of a first layer roto-translational convolution $f_{R}^{(1)}: \sZ^{2} \times \Lambda^{(1)}\rightarrow \Theta \times \sZ^{2} \times \Lambda^{(2)}$ as defined in Eq. \ref{eq:10_rot_conv_1}, and $\Theta$ be the set of rotations by $\frac{\pi}{2}$ degrees: $\Theta=\{\theta_{r} = r \frac{\pi}{2}\}_{r=1}^{4}$. Moreover, let $R=\{\{f_{R}^{(1)}(p)(u)\}^4_{r=1}\}^\Lambda_{\lambda=1}$ be the matrix of dimension $4 \times \Lambda$ consisting of the $4$ oriented responses for each of the $\Lambda$ learned representations at a certain position $u$. Since the vectors $R(\lambda)=\{\{f_{R}^{(1)}(p)(u)\}^4_{r=1}\}(\lambda)$ permute cyclically as a result of the rotation equviariance property of $f^{(1)}_{R}$, it is mandatory to ensure equivariance to cyclic permutations in each $R(\lambda)$ during the course of the attention procedure (see Section \ref{sec:10_co_att_nns}). 

At first sight, one is inclined to think that there is no connection between multiple vectors $f_{R}(p)(u, \lambda)$ in $f_{R}(p)(u)$, and, therefore, in order to exploit co-occurences, one must impose additional constraints along the $\lambda$ axis. However, there is indeed an implicit restriction in $f_{R}(p)(u)$ along $\lambda$ resulting from the rotation equivariance property of the mapping $f_{R}$, which we can take advantage from to simplify the problem at hand.
Consider, for instance, the input $\theta_{i}p$, a $\theta_{i}$-rotated version of $p$. By virtue of the equivariance property of $f_{R}$, we have (locally) that $f_{R}(\theta_{i}p)=\mathcal{P}^{i}(f_{R}(p))$. Furthermore, we know that this property must hold for all the learned feature representations $f_{R}(p)(u, \lambda)$,$\forall \lambda \in \Lambda_{1}$. Resultantly, we have that:
\begin{equation}
\label{eq:10_app_att_on_mapping}
    f_{R}(\theta_{i}p)(u,r,\lambda)=\mathcal{P}^{i}(f_{R}(p)(u,r,\lambda)) \ \ , \ \ \forall\lambda\in\Lambda_{1}.
\end{equation}
In other words, if one of the learned mappings $f_{R}(p)(u, r, \lambda)$ experiences a permutation $\mathcal{P}^{i}$ along $r$, \textit{all} the learned representations $f_{R}(p)(u, r, \lambda)$, $\forall \lambda \in \Lambda_{1}$ must experience the exact same permutation $\mathcal{P}^{i}$ as well. 
Resultantly, the equivariance property of the mapping $f_{R}$ ensures that \textit{all} the $\Lambda_{1}$ learned feature representations $f_{R}(p)(u, \lambda)$ \textbf{\textit{\enquote{move synchronously}}} as a function of input rotation $\theta_{i}$.

Likewise, if we apply a cyclic equivariant attention mechanism $\mathcal{A}^{\mathcal{C}}_{\lambda}$ independently on top of each $\lambda$ learned representation $f_{R}(p)(u, \lambda)$, we obtain that the relation
\begin{equation}
\label{eq:10_app_att_on_mapping_2}
    \mathcal{A}^{\mathcal{C}}_{\lambda}(f_{R}(\theta_{i}p))(u,r,\lambda)=\mathcal{P}^{i}(\mathcal{A}^{\mathcal{C}}_{\lambda}(f_{R}(p))(u,r,\lambda)) \ \ , \ \  \forall\lambda\in\Lambda_{1},
\end{equation} 
must hold as well. Analogous to the case shown in Eq. \ref{eq:10_app_att_on_mapping} and given that $\mathcal{A}^{\mathcal{C}}_{\lambda}$ is equivariant to cyclic permutations on the domain, we obtain that \textit{all} the $\Lambda_{1}$ learned \textit{attention masks} $\mathcal{A}^{\mathcal{C}}_{\lambda}$ \textbf{\textit{\enquote{move synchronously}}} as a function of input rotation $\theta_{i}$ as well (see Fig. \ref{fig:10_my_label}). 

From Eq. \ref{eq:10_app_att_on_mapping_2} and Figure \ref{fig:10_my_label}, one can clearly see that by utilizing $\mathcal{A}_{\lambda}^{\mathcal{C}}$ independently along $r$ and taking advantage from the fact that all $\Lambda_{1}$ learned feature representations are tied with one another via $f_{R}$, one is able to prioritize learning of feature representations that co-occur together as opposed to the much looser formulation in Eq. \ref{eq:10_app_att_on_mapping}, where feedback is obtained from all orientations.

%=========================================================================

%% file: backmatter/appx_6_attgconv.tex
% path to figures directory
\graphicspath{{figures/6-attgconv/}}

%=========================================================================

% \begin{savequote}[75mm]
% Nulla facilisi. In vel sem. Morbi id urna in diam dignissim feugiat. Proin molestie tortor eu velit. Aliquam erat volutpat. Nullam ultrices, diam tempus vulputate egestas, eros pede varius leo.
% \qauthor{Quoteauthor Lastname}
% \end{savequote}

\chapter{Attentive Group Equivariant Convolutional Neural Networks}
	%\label{chapter:}

%=========================================================================
\vspace{-7mm}
\section{Generalized Visual Self-Attention} \label{appx:6_gralized_visual_self_att}
Before we derive the constraints for general visual self-attention and prove Thm.~\ref{thm1} of the main article, we first motivate our definition of group equivariant visual self-attention. In this section, we explain that our definition of attentive group convolution, as given in Eq.~\ref{eq:6_att_g_conv}, and reformulated in Eq.~\ref{eq:6_attentivegconv}, essentially describes a group equivariant linear mapping that is augmented with an additional attention function. %~\ref{eq:6_att_g_conv}
\vspace{-7mm}
\subsection{Self-attention: From Vectors to Feature Maps}
Let us first consider the general form of a linear map between respectively vector spaces (used in multi-layer perceptrons) and feature maps (used in (group) convolutional neural nets), defined as follows:
\begin{align}
\label{eq:6_linmapvectors}
\text{vectors:}
\hspace{0.4cm} 
\mathbf{x}^{out}_c &= \sum_{\tilde{c}}^{N_{\tilde{c}}} \mathbf{W}_{c,\tilde{c}} \, \mathbf{x}^{in}_{\tilde{c}},  \\
\label{eq:6_linmapfeaturemaps}
\text{feat maps:} \hspace{0.4cm} 
f^{out}_c(g) &= \sum_{\tilde{c}}^{N_{\tilde{c}}} \int\limits_G \Psi_{c,\tilde{c}}(g,\tilde{g}) f^{in}_{\tilde{c}}(\tilde{g}) {\rm d}\tilde{g}.
\end{align}
Here, the first equation describes a linear map between vectors $\mathbf{x}^{in} \in \mathbb{R}^{N_{\tilde{c}}}$ and $\mathbf{x}^{out} \in \mathbb{R}^{N_{c}}$ via matrix-vector multiplication with matrix $\mathbf{W} \in \mathbb{R}^{N_{c} \times N_{\tilde{c}}}$. The second equation describes a linear map between feature maps $f^{in} \in (\mathbb{L}_2(G))^{N_{\tilde{c}}}$ and $f^{out} \in (\mathbb{L}_2(G))^{N_{c}}$, via a two argument kernel $\Psi \in \mathbb{L}_1(G \times G)^{N_{\tilde{c}} \times N_{c}}$. The two argument kernel $\Psi$ can be seen as the continuous counterpart of the matrix $\mathbf{W}$, and matrix-vector multiplication (sum over input indices) is augmented with an integral over the input coordinates $\tilde{g}$.

Keeping this form of linear mapping, we define the self-attentive map as the regular linear map augmented with attention weights computed from the input. Consequently, we formally define the self-attentive mappings as:
\begin{align}
\label{eq:6_generalattentionvectors}
\text{vectors:} \hspace{0.6cm} 
\mathbf{x}^{out}_c &= \sum_{\tilde{c}}^{N_{\tilde{c}}} \mathbf{A}_{c,\tilde{c}}\mathbf{W}_{c,\tilde{c}} \, \mathbf{x}^{in}_{\tilde{c}}, \\
\label{eq:6_generalattentionfeaturemaps}
\text{feat maps:} \hspace{0.4cm} 
f^{out}_c(g) &= \sum_{\tilde{c}}^{N_{\tilde{c}}} \int\limits_G \alpha_{c,\tilde{c}}(g,\tilde{g}) \Psi_{c,\tilde{c}}(g,\tilde{g})
f^{in}_{\tilde{c}}(\tilde{g}) {\rm d}\tilde{g},
\end{align}
in which the attention weights are computed from the input via some operator $\mathcal{A}$, i.e., $\mathbf{A}_{c,\tilde{c}} = \mathcal{A}[\mathbf{x}^{in}]_{c,\tilde{c}}$ in the vector case and $\alpha_{c,\tilde{c}} = 
\mathcal{A}[f^{in}]_{c,\tilde{c}}$ in the case of feature maps.

\vspace{-7mm}
\subsection{Equivariant Linear Maps are Group Convolutions}
\label{sec:6_gconvderivation}
Now, since we want to preserve the spatial correspondences between the input and output feature maps, special attention should be paid to the continuous self-attentive mappings. In other words, these operators should be equivariant. By including an equivariance constraint on the linear mapping of Eq.~\ref{eq:6_linmapfeaturemaps} we obtain a group convolution (see e.g. \citet{kondor2018generalization,cohen2019general,bekkers2020bspline}). The derivation is as follows: 

Imposing the equivariance constraint $
\mathcal{L}_{{g}}[f^{in}] \underset{\text{Eq.} \ref{eq:6_linmapfeaturemaps}}{\mapsto} \mathcal{L}_{{g}}[f^{out}]$ means that for all $\overline{g},g \in G$ and all $f \in \mathbb{L}_2(G)^{N_c}$ we must guarantee that:
\begin{align*}
   \mathcal{L}_{{g}}[f^{in}] &= \mathcal{L}_{{g}}[f^{out}]\\
&\hspace{-1mm}\Leftrightarrow \\
\int_G \Psi_{c,\tilde{c}}(g,\tilde{g}) \mathcal{L}_{\overline{g}}\left[f\right](\tilde{g}) {\rm d}\tilde{g} &= \int_G \Psi_{c,\tilde{c}}(\overline{g}^{-1} g, \tilde{g}) f(\tilde{g}) {\rm d}\tilde{g} \\
&\hspace{-1mm}\Leftrightarrow \\
\int_G \Psi_{c,\tilde{c}}(g,\tilde{g}) f(\overline{g}^{-1}\tilde{g}) {\rm d}\tilde{g}& = \int_G \Psi_{c,\tilde{c}}(\overline{g}^{-1} g, \tilde{g}) f(\tilde{g}) {\rm d}\tilde{g} \\
& \hspace{-1mm}\Leftrightarrow \\
\int_G \Psi_{c,\tilde{c}}(g,\tilde{g}) f(\overline{g}^{-1}\tilde{g}) {\rm d}\tilde{g} &= \int_G \Psi_{c,\tilde{c}}(\overline{g}^{-1} g, \overline{g}^{-1}\tilde{g}) f(\overline{g}^{-1}\tilde{g}) {\rm d}\tilde{g},
\end{align*}
where the change of variables $\tilde{g}\rightarrow\overline{g}^{-1}\tilde{g}$ as well as the left-invariance of the Haar measure ( ${\rm d}(\overline{g}^{-1}\tilde{g}) = {\rm d}\tilde{g}$)) is used in the last step. Since this equality must hold for all $f\in\mathbb{L}_2(G)^{N_{\tilde{c}}}$, $\Psi$ should be left-invariant in both input arguments. In other words, we have that: $$
\forall_{\overline{g}.\in G}: \Psi(\overline{g}g,\overline{g}\tilde{g}) = \Psi(g,\tilde{g}).
$$ 
Resultantly, we can always multiply both arguments with $g^{-1}$ and obtain $\Psi(e,g^{-1} \tilde{g})$, which is effectively a single argument function $\psi(g^{-1}\tilde{g}):=\Psi(e,g^{-1}\tilde{g})$ that takes as input a relative \enquote{displacement} $g^{-1}\tilde{g}$. Consequently, under the equivariance constraint, Eq.~\ref{eq:6_linmapfeaturemaps} becomes a group convolution:
$$
f^{out}_c(g) = \sum_{\tilde{c}}^{N_{\tilde{c}}} \int\limits_G \psi_{c,\tilde{c}}(g^{-1}\tilde{g}) f^{in}_{\tilde{c}}(\tilde{g}) {\rm d}\tilde{g}. 
$$

\vspace{-7mm}
\subsection{Proof of Theorem 1}
We can apply the same type of derivation to reduce the general form of visual self-attention of Eq.~\ref{eq:6_generalattentionfeaturemaps} to our main definition of attentive group convolution:
\begin{equation}
\label{eq:6_attentivegconv}
f^{out}_c(g) = \sum_{\tilde{c}}^{N_{\tilde{c}}} \int\limits_G \alpha_{c,\tilde{c}}(g,\tilde{g})\psi_{c,\tilde{c}}(g^{-1}\tilde{g}) f^{in}_{\tilde{c}}(\tilde{g}) {\rm d}\tilde{g}.
\end{equation}
However, we cannot reduce attention map $\alpha$ to a single argument function like we did for the kernel $\Psi$ since $\alpha$ depends on the input $f^{in}$. To see this consider the following:

Without loss of generality, let $\mathfrak{A}: \mathbb{L}_2(G) \rightarrow \mathbb{L}_2(G)$ denote the attentive group convolution defined by Eq.~\ref{eq:6_attentivegconv}, with $N_c = N_{\tilde{c}} = 1$, and some $\psi$ which in the following we omit in order to simplify our derivation. Equivariance of $\mathfrak{A}$ implies that $\forall_{f \in \mathbb{L}_2(G)}$, $ \forall_{\overline{g},g \in G}:$
\begin{align*}
\mathfrak{A}\left[\mathcal{L}_{\overline{g}}\left[f\right]\right](g) &= \mathcal{L}_{\overline{g}}\left[\mathfrak{A}\left[f\right]\right](g) \\
&\hspace{-1mm}\Leftrightarrow \\
\mathfrak{A}\left[\mathcal{L}_{\overline{g}}\left[f\right]\right](g) &= \mathfrak{A}\left[f\right](\overline{g}^{-1} g) \\
&\hspace{-1mm}\Leftrightarrow \\
\int_G \mathcal{A}\left[\mathcal{L}_{\overline{g}}\left[f\right]\right](g,\tilde{g})\mathcal{L}_{\overline{g}}\left[f\right](\tilde{g}) {\rm d}\tilde{g} & = \int_G \mathcal{A}\left[f\right](\overline{g}^{-1}g,\tilde{g}) f(\tilde{g}) {\rm d}\tilde{g} \\
&\hspace{-1mm} \Leftrightarrow \\
\int_G \mathcal{A}\left[\mathcal{L}_{\overline{g}}\left[f\right]\right](g,\tilde{g})f(\overline{g}^{-1}\tilde{g}) {\rm d}\tilde{g} &= \int_G \mathcal{A}\left[f\right](\overline{g}^{-1}g,\overline{g}^{-1}\tilde{g}) f(\overline{g}^{-1}\tilde{g}) {\rm d}\tilde{g},
\end{align*}
where we once again perform the variable substitution $\tilde{g}\rightarrow \overline{g}^{-1}\tilde{g}$ at the right hand side of the last step. This must hold for all $f \in \mathbb{L}_2(G)$ and hence:
\begin{equation}
\label{eq:6_constraint}
\forall_{\overline{g}\in G}: \mathcal{A}\left[\mathcal{L}_{\overline{g}}f\right](g,\tilde{g}) = \mathcal{A}\left[f\right](\overline{g}^{-1}g,\overline{g}^{-1}\tilde{g}),
\end{equation}
which proves the constraint on $\mathcal{A}$ as given in Thm.~1 of the main article. Just as for convolutions in Sec.~\ref{sec:6_gconvderivation}, we can turn this into a single argument function as:
\begin{equation}
\label{eq:6_singleargatt}
\mathcal{A}[f](g,\tilde{g}) = \mathcal{A}\left[\mathcal{L}_{g^{-1}} f\right](e,g^{-1} \tilde{g}) =: \mathcal{A}'[\mathcal{L}_{g^{-1}} f](g^{-1}\tilde{g}),
\end{equation}
in which $\mathcal{A}'$ is an attention operator that generates a single argument attention map from an input $f$. However, this would mean that for each $g$ the input should be transformed via $\mathcal{L}_{g^{-1}}$, which does not make things easier for us. Things do get easier when we choose to attend to either the input \emph{or} the output, which we discuss next.
\begin{corollary}
\label{cor1}
Each attention operator $\mathcal{A}$ that generates an attention map $\alpha:G\times G \rightarrow [0,1]$ which is left-invariant to either one of the arguments, and thus exclusively attends either the input or output domain, satisfies the equivariance constraint of Eq.~\ref{eq:6_constraint}, iff the operator is G-equivariant, i.e., a group convolution.
\end{corollary}
\vspace{-3mm}
\begin{proof}
Left-invariant to either one of the arguments (let us now consider invariance in the first argument) means that:
\begin{align*}
\forall_{g,\tilde{g}}: \;\;  \mathcal{A}[f](g,\tilde{g}) = \mathcal{A}[f]](e,\tilde{g}),
\end{align*}
and hence, we are effectively dealing with a single argument attention map, which we define as $\mathcal{A}'[f](\tilde{g}){\coloneqq}\mathcal{A}(e,\tilde{g})$. Hence, the equivariance constraint of Eq.~\ref{eq:6_constraint} becomes:
\begin{align*}
\forall_{\overline{g}\in G}: \mathcal{A}\left[\mathcal{L}_{\overline{g}}f\right](g,\tilde{g}) &= \mathcal{A}\left[f\right](\overline{g}^{-1}g,\overline{g}^{-1}\tilde{g}) \Leftrightarrow \\
\forall_{\overline{g}\in G}: \mathcal{A}'\left[\mathcal{L}_{\overline{g}}f\right](\tilde{g}) &= \mathcal{A}'\left[f\right](\overline{g}^{-1}\tilde{g}) \Leftrightarrow \\
\forall_{\overline{g}\in G}: \mathcal{A}'\left[\mathcal{L}_{\overline{g}}f\right] &= \mathcal{L}_{\overline{g}}\left[\mathcal{A}'\right]\left[f\right].
\end{align*}
Conclusively, $\mathcal{A}'$ must be an equivariant operator.
\end{proof}
\vspace{-3mm}

The derivation of the Eq.~\ref{eq:6_constraint} together with the proof of Corollary~\ref{cor1} completes the proof of Theorem~\ref{thm1}.

\vspace{-7mm}
\subsection{Equivariance Proof of the Proposed Visual Attention}
In this section we revisit the proposed attention mechanisms and prove that they indeed satisfy Thm.~1 of the main article. Recall the general formulation of attentive group convolution given in Eq.~\ref{eq:6_attentivegconv}. Inspired by \citet{woo2018cbam}, we reduce the computation load by factorizing the attention map $\alpha$ into channel and spatial components:
$$
\alpha_{c,\tilde{c}}(g,\tilde{g}) = \alpha^{\mathcal{X}}(x, h, \tilde{h})\alpha^{C}_{c,\tilde{c}}(h,\tilde{h}),
$$
where $\alpha^{\mathcal{C}}$ attends to both input and output channels as well as input and output poses $h,\tilde{h}\in H$, and spatial attention attends to the output domain $g {=} (x,h) \in G$ for all input poses $\tilde{h} \in H$ but does not change for input spatial positions $\tilde{x} \in \mathbb{R}^{d}$. We denote the operators $\mathcal{A}^{C}$, $\mathcal{A}^{X}$ utilized to compute the attention maps as $\alpha^{\mathcal{C}} {=} \mathcal{A}^{\mathcal{C}}[f]$ and $\alpha^{\mathcal{X}} {=} \mathcal{A}^{\mathcal{X}}[f]$.

\vspace{-7mm}
\subsubsection{Channel attention}
We compute channel attention via:\footnote{In the main text, we write Eqs.~\ref{eq:6_compute_ak_2_suppl}, \ref{eq:6_spat_att} in convolution form by which only attention to $\tilde{h}$ is considered.  However, in the rot-MNIST experiments we compute attention based on $\tilde{f}$ and apply attention to both $h$ and $\tilde{h}$ via Eqs.~\ref{eq:6_compute_ak_2_suppl}, \ref{eq:6_spat_att}. %In order to do so, we slightly formulate Eq.~18 of the main article in order to obtain  Eq.~\ref{eq:6_compute_ak_2_suppl}.
}:
\begin{align}
    \mathcal{A}^{\mathcal{C}}[f](h,\tilde{h})&=\varphi^{\mathcal{C}}\left[s^{\mathcal{C}}\left[\tilde{f}[f]\right]\right](h,\tilde{h}) \nonumber\\
    &= \sigma\Big(\big[
    \mathbf{W}_{2}(h^{-1}\tilde{h})\cdot[\mathbf{W}_{1}(h^{-1}\tilde{h})\cdot s^{\mathcal{C}}_{ \text{avg}}(h,\tilde{h})]^{+} \big[
    \mathbf{W}_{2}(h^{-1}\tilde{h})\cdot[\mathbf{W}_{1}(h^{-1}\tilde{h})\cdot s^{\mathcal{C}}_{ \text{max}}(h,\tilde{h})]^{+}
    \big] \Big) \label{eq:6_compute_ak_2_suppl}
\end{align}
with 
\begin{equation}\label{eq:6_intermediate}
    \tilde{f}_{c,\tilde{c}}(x,h,\tilde{h}) := \big[f_{\tilde{c}} \star_{\mathbb{R}^{d}} \mathcal{L}_{h}[\psi_{c,\tilde{c}}]\big](x, \tilde{h})
\end{equation} the intermediary result from the convolution between the input $f$ and the $h$-transformation of the filter $\psi$, $\mathcal{L}_{h}[\psi]$ before pooling over $\tilde{c}$ and $\tilde{h}$. $s^{\mathcal{C}}_{\text{avg}}$ and $s^{\mathcal{C}}_{\text{max}}$ denote respectively average and max pooling over the ${x}$ coordinate. 

Here, we apply a slight abuse of notation with $\tilde{f}[f]$ and $s^{\mathcal{C}}[\tilde{f}]$ in order to keep track of the dependencies. In order to proof equivariance of the attention operator $\mathcal{A}^{\mathcal{C}}$ we need to proof that $\forall_{\overline{g} \in G}: \mathcal{A}^{\mathcal{C}}[\mathcal{L}_{\overline{g}}[ f]](h,\tilde{h}) = \mathcal{A}^{\mathcal{C}}[f](\overline{h}^{-1} h,\overline{h}^{-1} \tilde{h})$, with $\overline{g} = (\overline{x},\overline{h})$. To this end, we first identify the equivariance and invariance properties of the functions used in Eq.~\ref{eq:6_compute_ak_2_suppl}. 

From Eq.~\ref{eq:6_intermediate} we see that the intermediate convolution result $\tilde{f}$ is equivariant via\break $\tilde{f}[\mathcal{L}_{\overline{g}}[f]](\tilde{x},\tilde{h}) =  \tilde{f}[f](\overline{g}^{-1} x, \overline{h}^{-1} h, \overline{h}^{-1} \tilde{h})$. For the statistics operators $s^{\mathcal{C}}$ we have invariance w.r.t. translations due to the pooling over $x$, and equivariance w.r.t. parameter $\overline{h}$ via $s^{\mathcal{C}}[\tilde{f}[\mathcal{L}_{\overline{g}}{f}]](h,\tilde{h}) = s^{\mathcal{C}}[\tilde{f}[f]](\overline{h}^{-1} h,\overline{h}^{-1}\tilde{h})$. Now, we propagate the transformation on the input and compute the result of $\mathcal{A}^{\mathcal{C}}[\mathcal{L}_{\overline{g}}[f]](g,\tilde{g})$. That is, we compute the left-hand side of the constraint given in Eq.~\ref{eq:6_constraint}, where, for brevity, we omit the $s^{\mathcal{C}}_{\text{max}}$ term:
\begin{equation*}
\mathcal{A}^{\mathcal{C}}[\mathcal{L}_{\overline{g}}[f]](g,\tilde{g}) = \mathbf{W}_{2}(h^{-1}\tilde{h})\cdot[\mathbf{W}_{1}(h^{-1}\tilde{h})\cdot s^{\mathcal{C}}_{ \text{\text{avg}}}(\overline{h}^{-1}h,\overline{h}^{-1}\tilde{h})]^{+}.
\end{equation*}
The right-hand side of Eq.~\ref{eq:6_constraint} is given by:
\begin{equation*}
\mathcal{A}^{\mathcal{C}}[f](\overline{g}^{-1}g,\overline{g}^{-1}\tilde{g}) = \mathbf{W}_{2}(h^{-1}\tilde{h})\cdot[\mathbf{W}_{1}(h^{-1}\tilde{h})\cdot s^{\mathcal{C}}_{ \text{\text{avg}}}(\overline{h}^{-1}h,\overline{h}^{-1}\tilde{h})]^{+}.
\end{equation*}
and hence, Eq.~\ref{eq:6_constraint} is satisfied for all $\overline{g} \in G$. Resultantly, $\mathcal{A}^{\mathcal{C}}$ is a valid attention operator.

\vspace{-7mm}
\subsubsection{Spatial attention}

The spatial attention map $\alpha^{\mathcal{X}}$ is computed via: 
\begin{align}
\alpha^{\mathcal{X}}(g, \tilde{h}) &= \mathcal{A}^{\mathcal{X}}[f](g, \tilde{h}) \nonumber \\
&= \varphi^{\mathcal{X}}\left[s^{\mathcal{X}}\left[\tilde{f}[f]\right]\right](g,\tilde{h}) =\sigma\left([s^{\mathcal{X}} \star_{\mathbb{R}^{d}} \mathcal{L}_{h}[\psi^{\mathcal{X}}]\right)(x, \tilde{h}) \label{eq:6_spat_att}
\end{align}
where $\sigma$ is a point-wise logistic sigmoid, $\psi^{\mathcal{X}}: G \rightarrow \mathbb{R}^2$ is a group convolution filter and $s^{\mathcal{X}}[\tilde{f}]:G \times H \rightarrow \mathbb{R}^2$ is a map of averages and maximum values taken over the channel axis at each $g \in G$ in $\tilde{f}$ for each $\tilde{h} \in H$. Note that Eq.~\ref{eq:6_spat_att} corresponds to a group convolution up to the final pooling operation over $\tilde{h}$. Since the statistics operator $s^{\mathcal{X}}$ is invariant w.r.t. translations in the input and Eq.~\ref{eq:6_spat_att} corresponds to a group convolution (up to pooling over $\tilde{h}$), we have that $\mathcal{A}^{X}$ is a valid attention operator as well.% The described attention operator is equivariant, and it produces a single-argument attention map. Due to Corollary \ref{cor1}, it is thus a valid attention operator.

\vspace{-7mm}
\section{Extended Implementation Details}\label{appx:6_extended_details}
In this section we provide extended details over our implementation. For the sake of completeness and reproducibility, we summarize the exact training procedures utilized during our experiments. Moreover, we delve into some important changes performed to some network architectures during our experiments to ensure \textit{exact} equivariance, and shed light into their importance for our equivariant attention maps.

\vspace{-7mm}
\subsection{General Observations}

We utilize \texttt{PyTorch} for our implementation. Any missing parameter specification in the following sections can be safely considered to be the default value of the corresponding parameter. For batch normalization layers, we utilize eps=0.00002 similarly to \citet{cohen2016group}.

\vspace{-7mm}
\subsection{rot-MNIST}
For rotational MNIST, we utilized the same backbone network as in \citet{cohen2016group}. During training we utilize Adam \cite{kingma2014adam}, batches of size 128, weight decay of 0.0001, learning rate of 0.001, drop-out rate of 0.3 and perform training for 100 epochs. Importantly and contrarily to \citet{cohen2016group}, we consistently experience improvements when utilizing drop-out and therefore we do not exclude it for any model. 

\vspace{-7mm}
\subsection{CIFAR-10}
It is not clear from \citet{springenberg2014striving, cohen2016group} which batch size is used in their experiments. For our experiments, we always utilize batches of size 128.

\vspace{-7mm}
\subsubsection{All-CNN}
We utilize the All-CNN-C structure of \citet{springenberg2014striving}. Analogously to \citet{springenberg2014striving, cohen2016group}, we utilize stochastic gradient descent, weight decay of 0.001 and perform training for 350 epochs. We utilize a grid search on the set $\{$0.01, 0.05, 0.1, 0.25$\}$ for the learning rate and report the best obtained performance. Furthermore, we reduce the learning rate by a factor of 10 at epochs 200, 250 and 300.

\vspace{-7mm}
\subsubsection{ResNet44}
Similar to \citet{cohen2016group}, we utilize stochastic gradient descent, learning rate of 0.05 and perform training for 300 epochs. Furthermore, we reduce the learning rate by a factor of 10 at epochs 50, 100 and 150.

\vspace{-7mm}
\subsection{PCam}
During training on the PatchCamelyon dataset, we utilize Adam \cite{kingma2014adam}, batches of size 64, weight decay of 0.0001, learning rate of 0.001 and perform training for 100 epochs. Furthermore, we reduce the learning rate by a factor of 2 after 20 epochs of no improvement in the validation loss. 

\vspace{-7mm}
\section{Effects of Stride and Input Size on Equivariance}\label{sec:6_approx_equiv}
Theoretically seen, the usage of stride during pooling and during convolution is of no relevance for the equivariance properties of the corresponding mapping \cite{cohen2016group}. However, we see that in practice stride can affect equivariance for specific cases as is the case for our experiments on CIFAR-10.

\begin{figure}[t]
    \centering
    \begin{subfigure}{\textwidth}
    \centering
    \captionsetup{justification=centering}
    \includegraphics[width=0.6\textwidth]{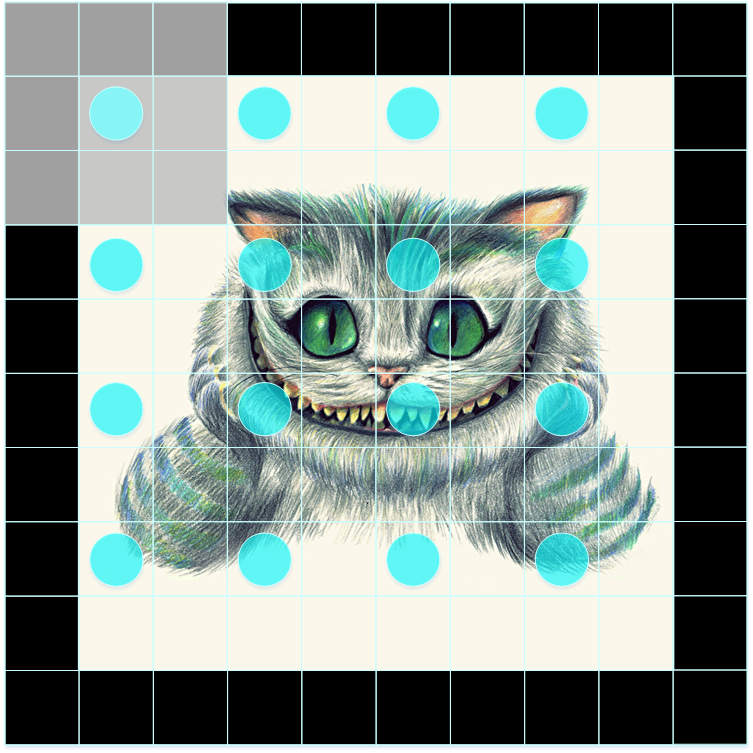}
    \caption{}\label{fig:6_norot}
    \end{subfigure}
    \begin{subfigure}{\textwidth}
    \centering
    \captionsetup{justification=centering}
        \includegraphics[width=0.6\textwidth]{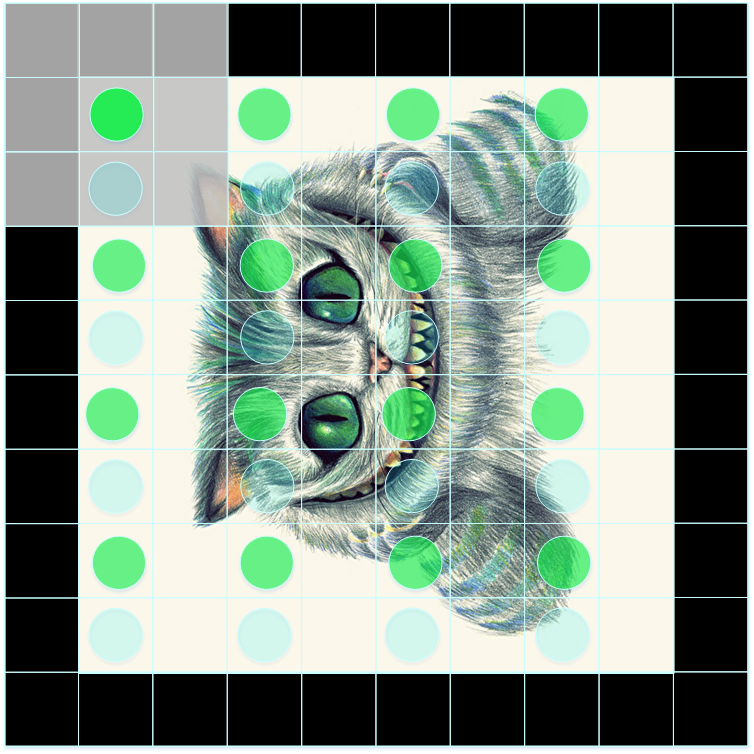}
        \caption{}\label{fig:6_yesrot}
    \end{subfigure}
    \caption{Effect of stride and input size on exact equivariance. Although group convolutions are ensured to be group equivariant, in practice, if the residual between the stride and the input size is even, as it's the case for the networks utilized in the CIFAR-10 experiments, equivariance is only approximate. This has important effects on equivariant attention maps (Fig.~\ref{fig:6_bad_examples}).}\label{fig:6_strideequiv}
\end{figure}

Consider the convolution between an input of even size and a small $3$x$3$ filter as shown in Fig.~\ref{fig:6_norot}. Via group convolutions, we can ensure that the output of the original input and a rotated one (Fig.~\ref{fig:6_yesrot}) will be exactly equal (up to the same rotation). Importantly however, note that for Fig. \ref{fig:6_strideequiv}, the local support of the filter, i.e., the input section with which the filter is convolved at a particular position, \textit{is not equivalent} for rotated versions of the input (denoted by blue circles for the non-rotated case and by green circles by for t
he rotated case). As a result, despite the group convolution itself being equivariant, the responses of both convolutions do not entirely resemble one another and, consequently, the depicted strided group convolution is \textit{not} exactly equivariant.

It is important to highlight that this behaviour is just exhibited for the special case when the residual between the used stride and the input size is even. Unfortunately, this is the case both for the ResNet44 as well as the All-CNN networks utilized in our CIFAR-10 experiments. However, as neighbouring pixels are extremely correlated with one another, the effects of this phenomenon are not of much relevance for the classification task itself. As a matter of fact, it can be interpreted as a form of data augmentation by skipping intermediary pixel values. Consequently, we can say that these networks are \textit{approximately equivariant}. 

\begin{figure*}
    \centering
    \begin{subfigure}{0.17\textwidth}
        \includegraphics[width=\textwidth]{x_0_rot.png}
    \end{subfigure}
    \hspace{1mm}
    \begin{subfigure}{0.17\textwidth}
        \includegraphics[width=\textwidth]{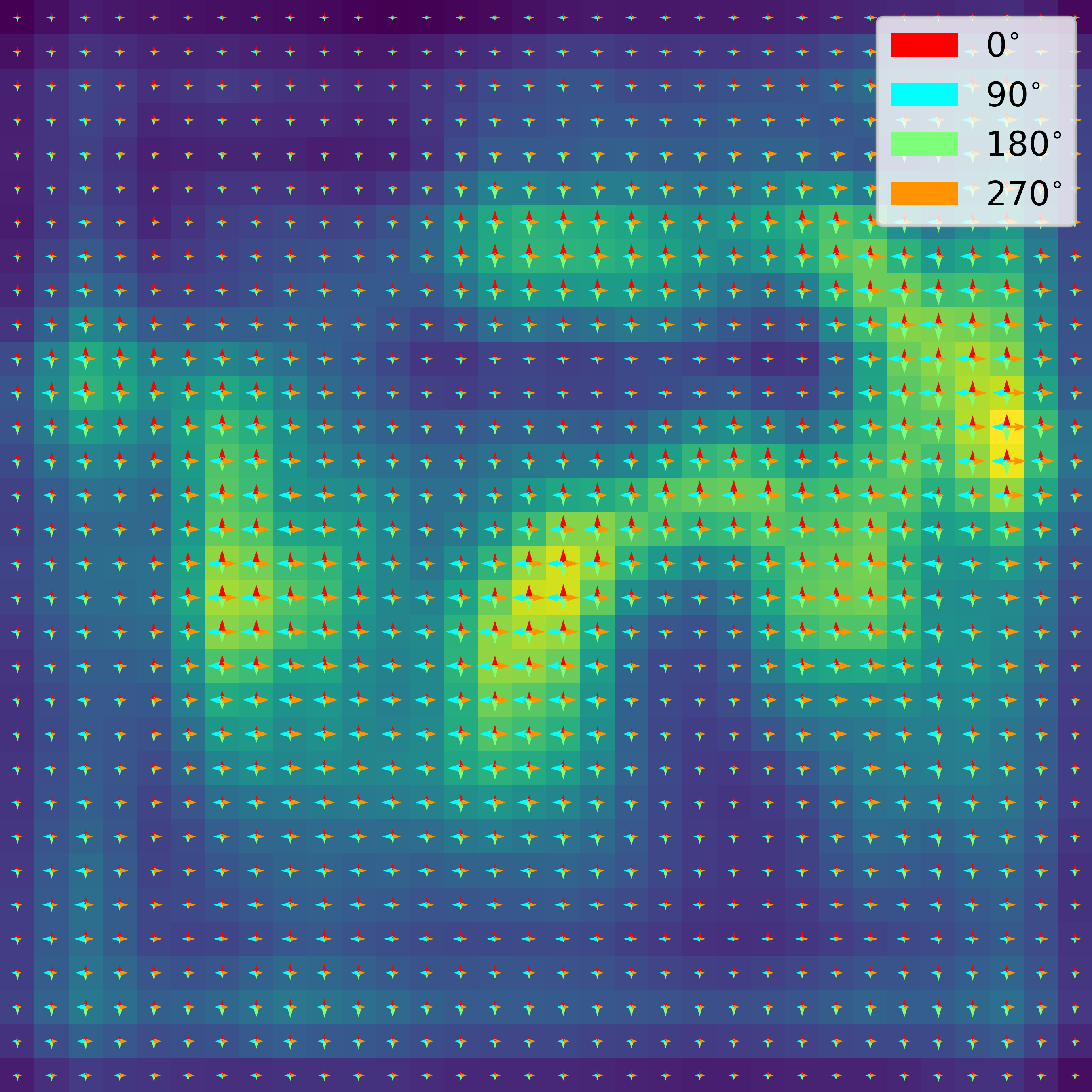}
    \end{subfigure}
    
    \vspace{6mm}
    \begin{subfigure}{0.17\textwidth}
        \includegraphics[width=\textwidth]{x_90_rot.png}
    \end{subfigure}
    \hspace{1mm}
    \begin{subfigure}{0.17\textwidth}
        \includegraphics[width=\textwidth]{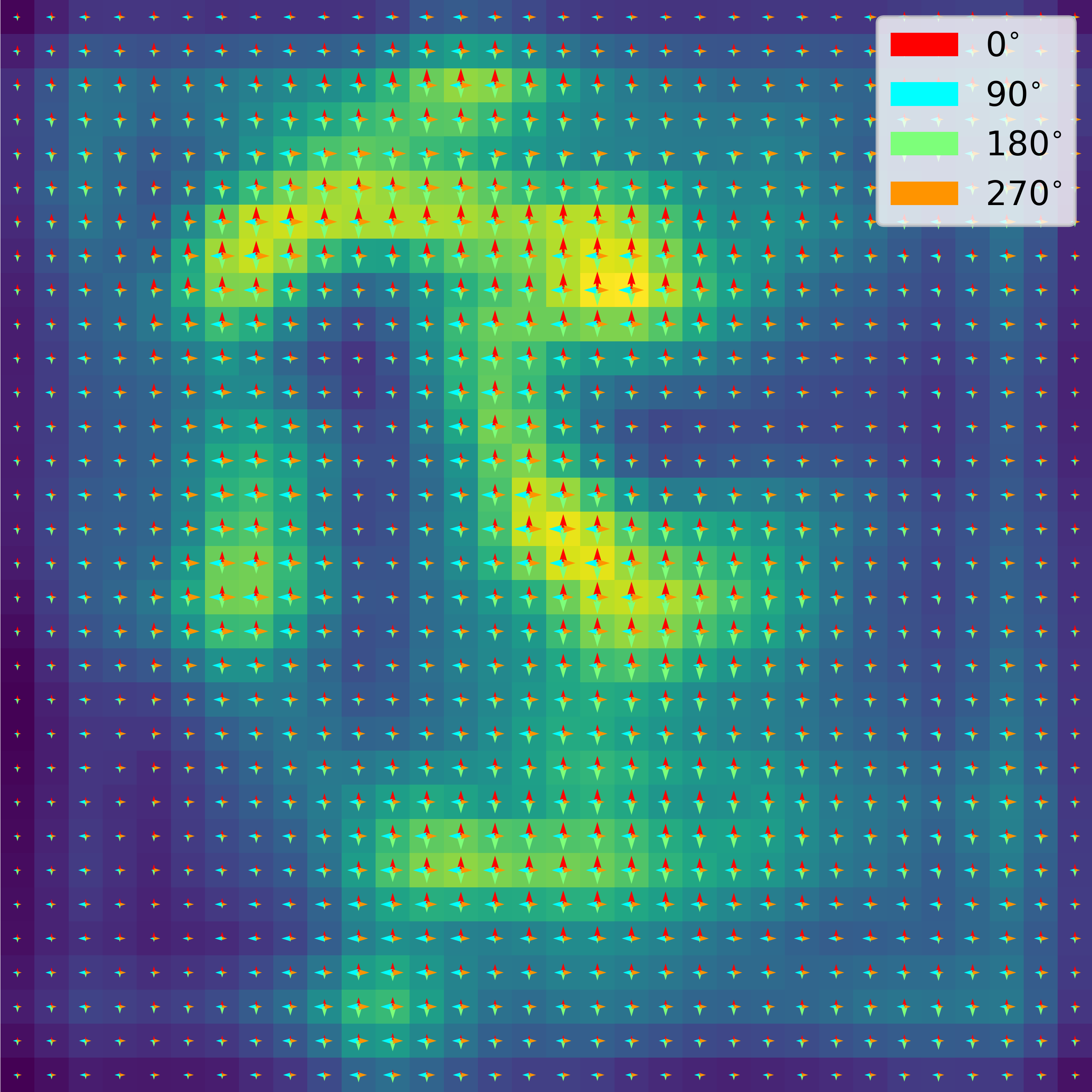}
    \end{subfigure}
    \hfill
    \begin{subfigure}{0.17\textwidth}
        \includegraphics[width=\textwidth]{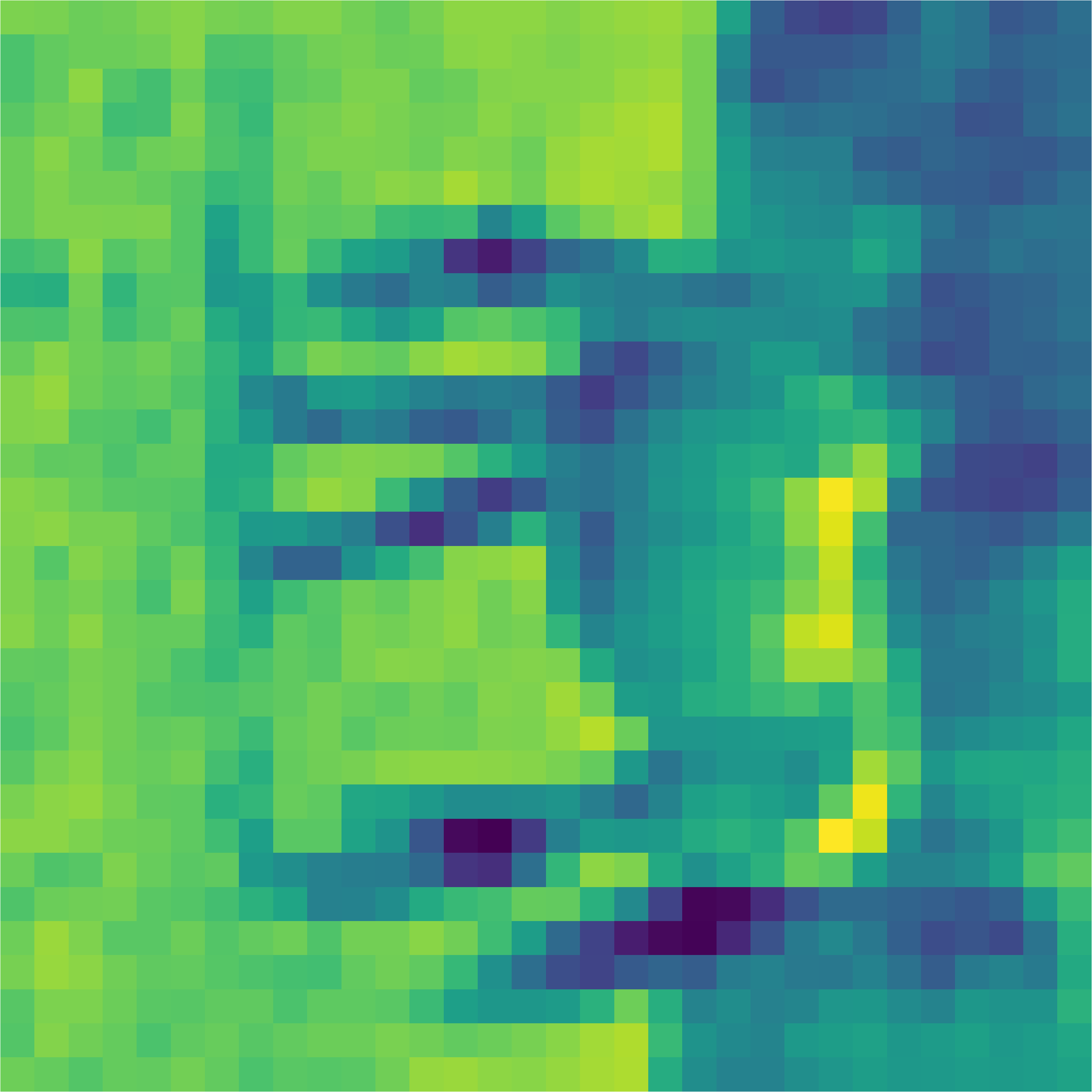}
    \end{subfigure}
    \hspace{1mm}
    \begin{subfigure}{0.17\textwidth}
        \includegraphics[width=\textwidth]{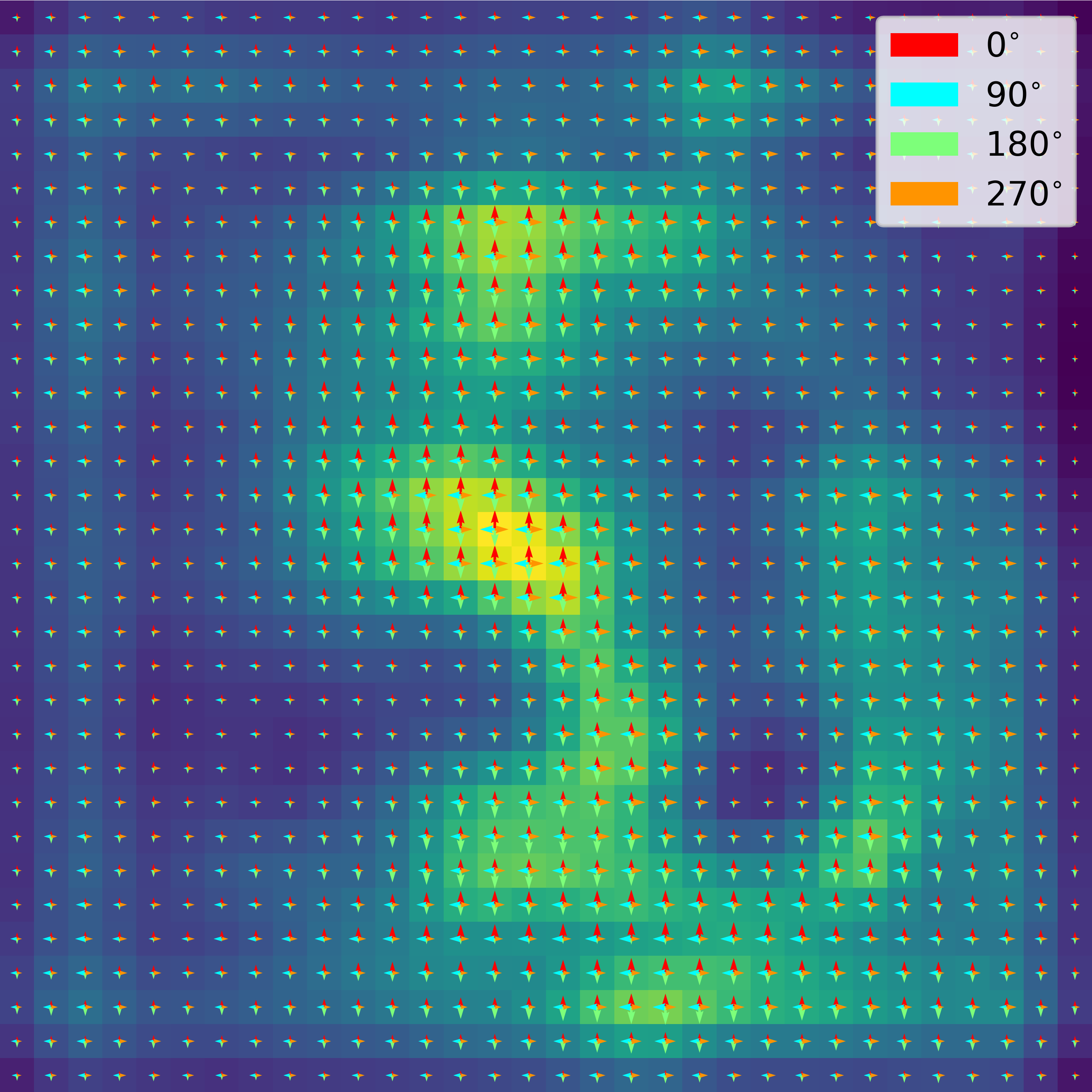}
    \end{subfigure}
    
    \vspace{6mm}
    \centering
    \begin{subfigure}{0.17\textwidth}
        \includegraphics[width=\textwidth]{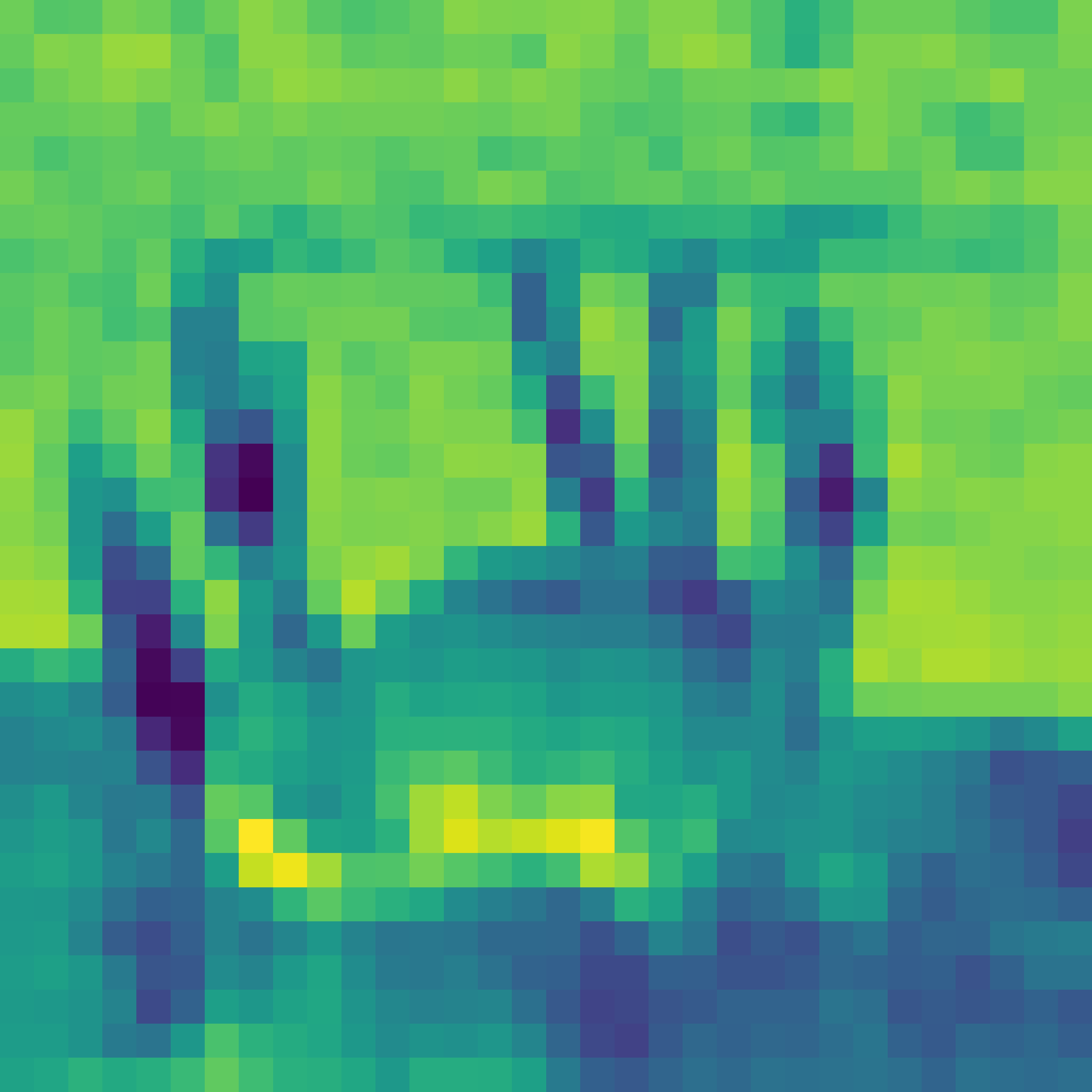}
    \end{subfigure}
    \hspace{1mm}
    \begin{subfigure}{0.17\textwidth}
        \includegraphics[width=\textwidth]{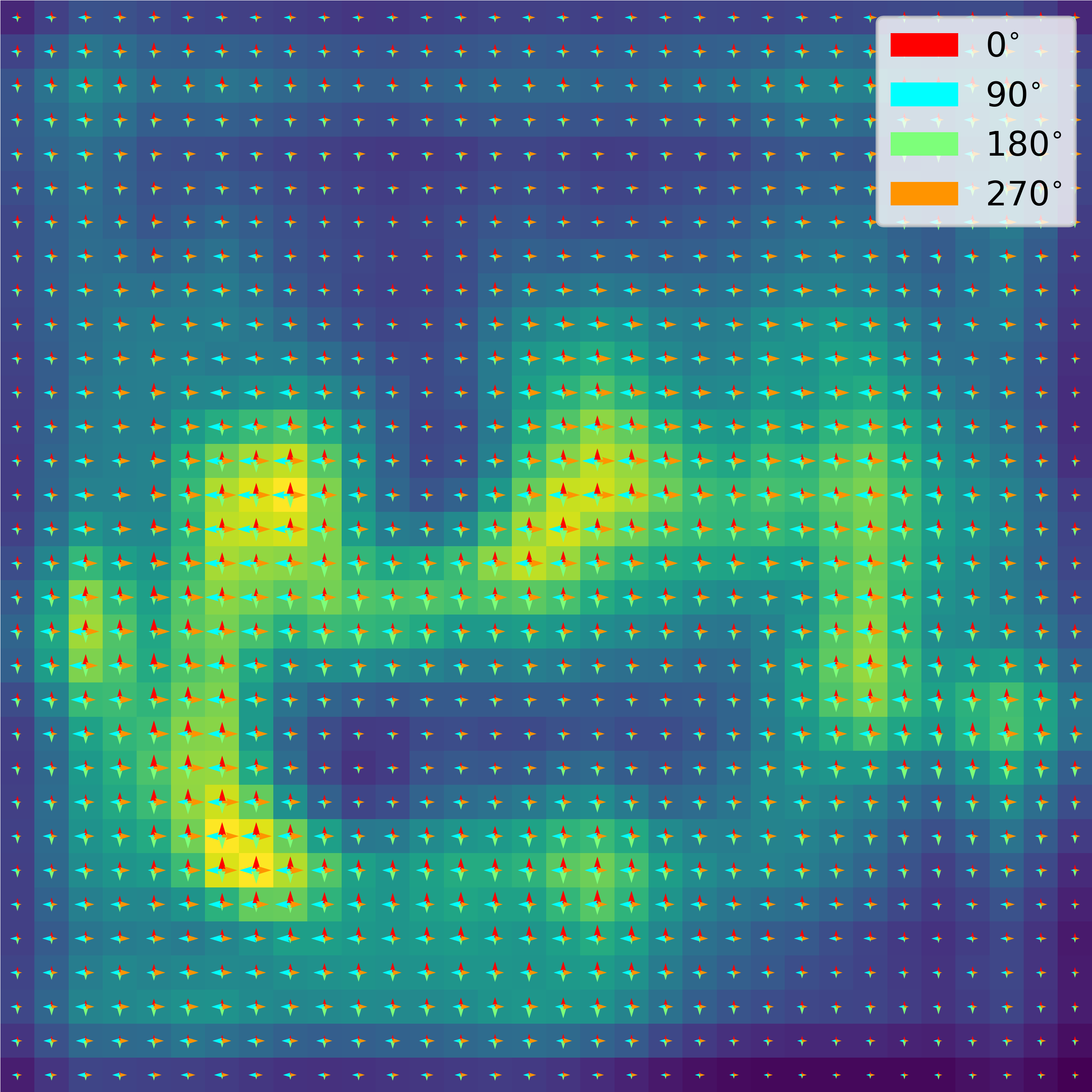}
    \end{subfigure}
    \caption{Examples of equivariant attention maps under approximate equivariance. Note that the attention map around the horse's back changes for different orientations.}\label{fig:6_bad_examples}
\end{figure*}

\begin{figure*}
    \centering
    \begin{subfigure}{0.17\textwidth}
        \includegraphics[width=\textwidth]{x_0_rot.png}
    \end{subfigure}
    \hspace{1mm}
    \begin{subfigure}{0.17\textwidth}
        \includegraphics[width=\textwidth]{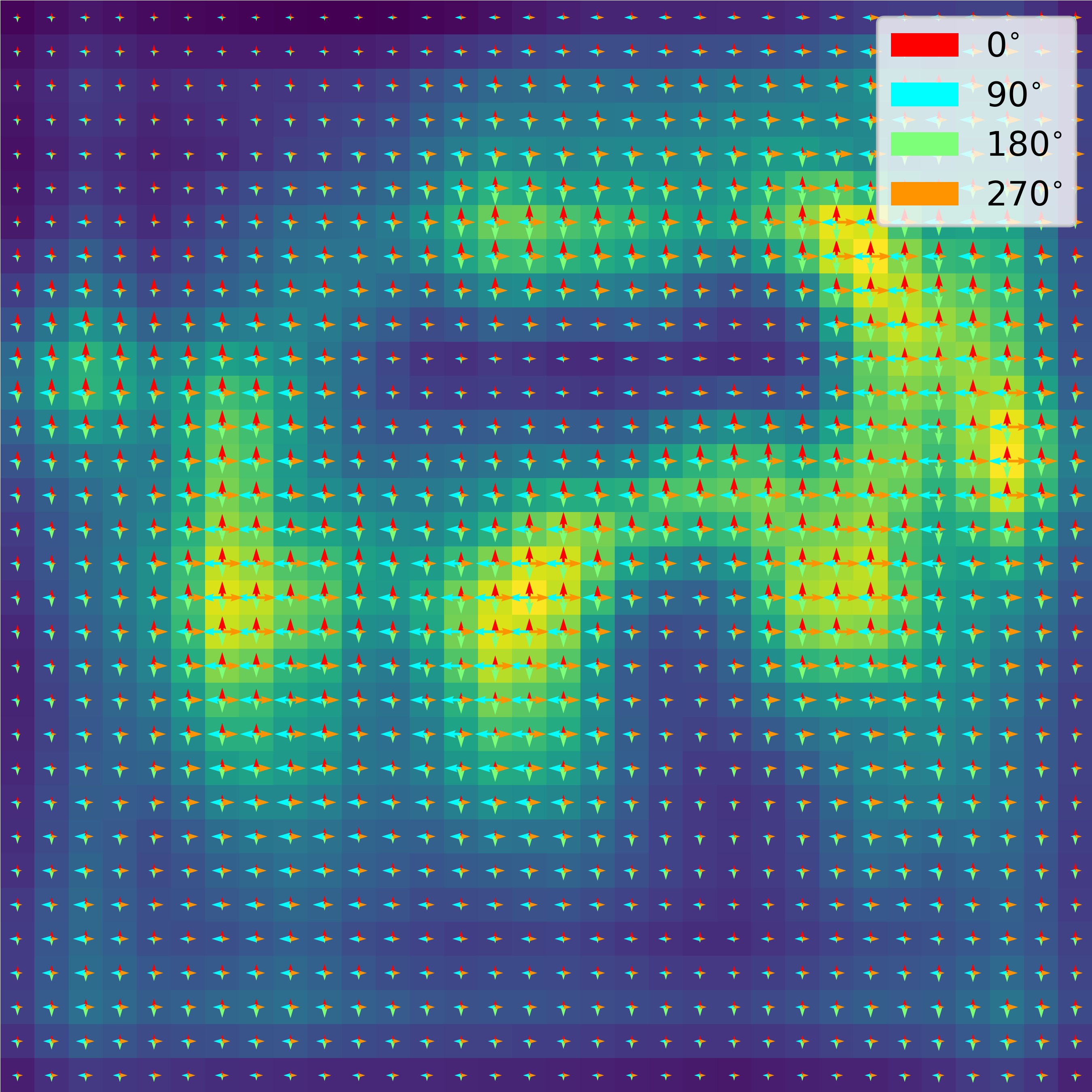}
    \end{subfigure}
    
    \vspace{6mm}
    \begin{subfigure}{0.17\textwidth}
        \includegraphics[width=\textwidth]{x_90_rot.png}
    \end{subfigure}
    \hspace{1mm}
    \begin{subfigure}{0.17\textwidth}
        \includegraphics[width=\textwidth]{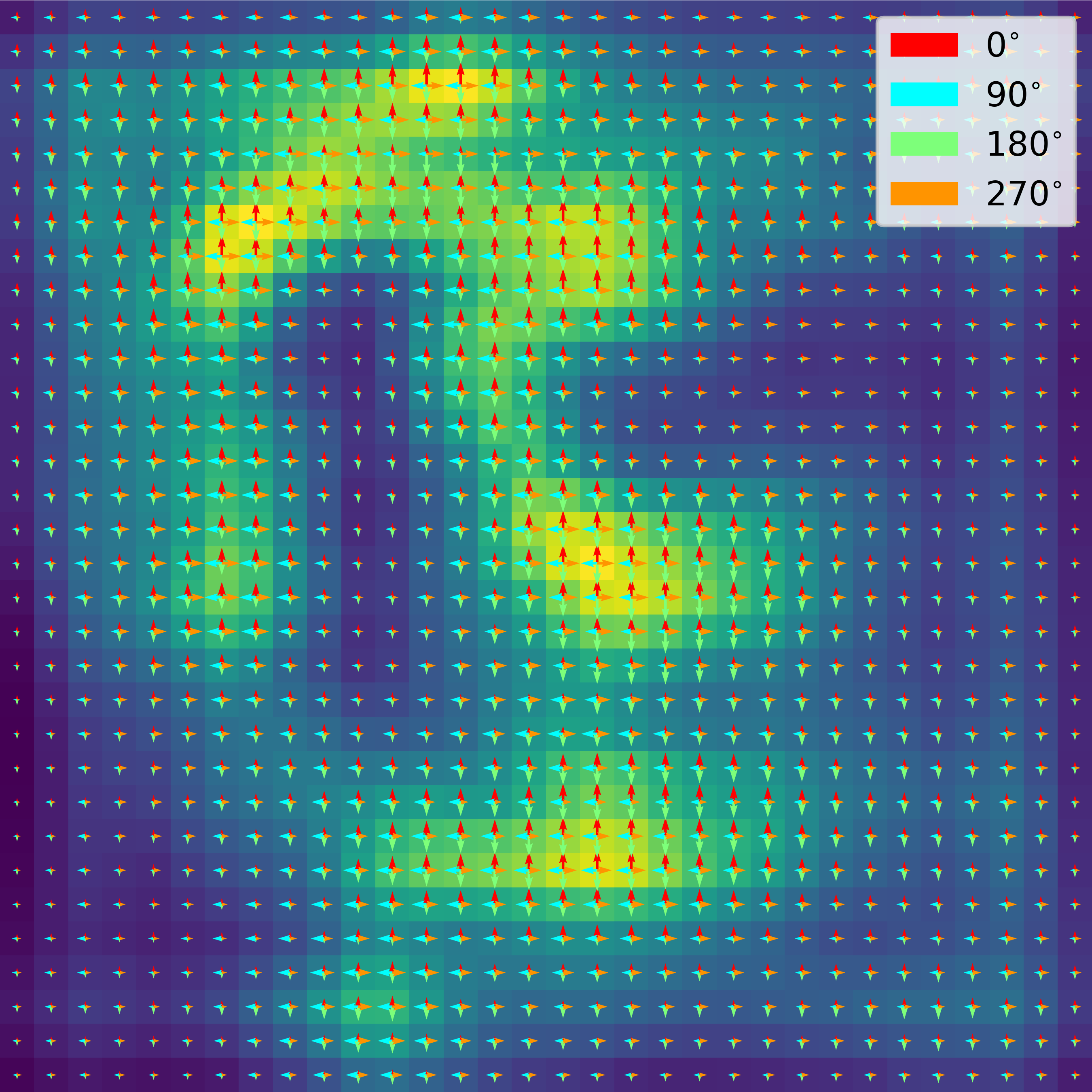}
    \end{subfigure}
    \hfill
    \begin{subfigure}{0.17\textwidth}
        \includegraphics[width=\textwidth]{x_270_rot.png}
    \end{subfigure}
    \hspace{1mm}
    \begin{subfigure}{0.17\textwidth}
        \includegraphics[width=\textwidth]{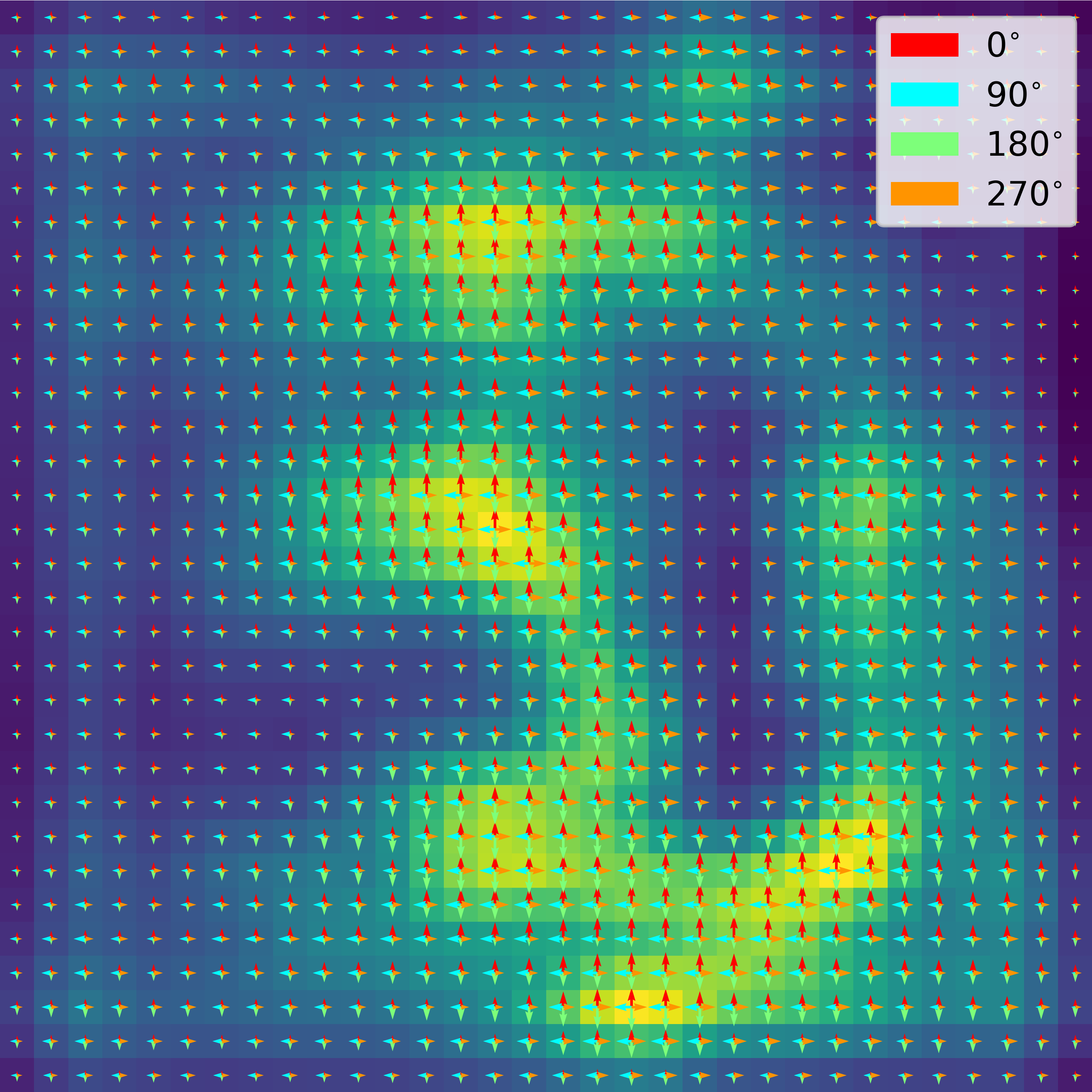}
    \end{subfigure}
    
    \vspace{6mm}
    \centering
    \begin{subfigure}{0.17\textwidth}
        \includegraphics[width=\textwidth]{x_180_rot.png}
    \end{subfigure}
    \hspace{1mm}
    \begin{subfigure}{0.17\textwidth}
        \includegraphics[width=\textwidth]{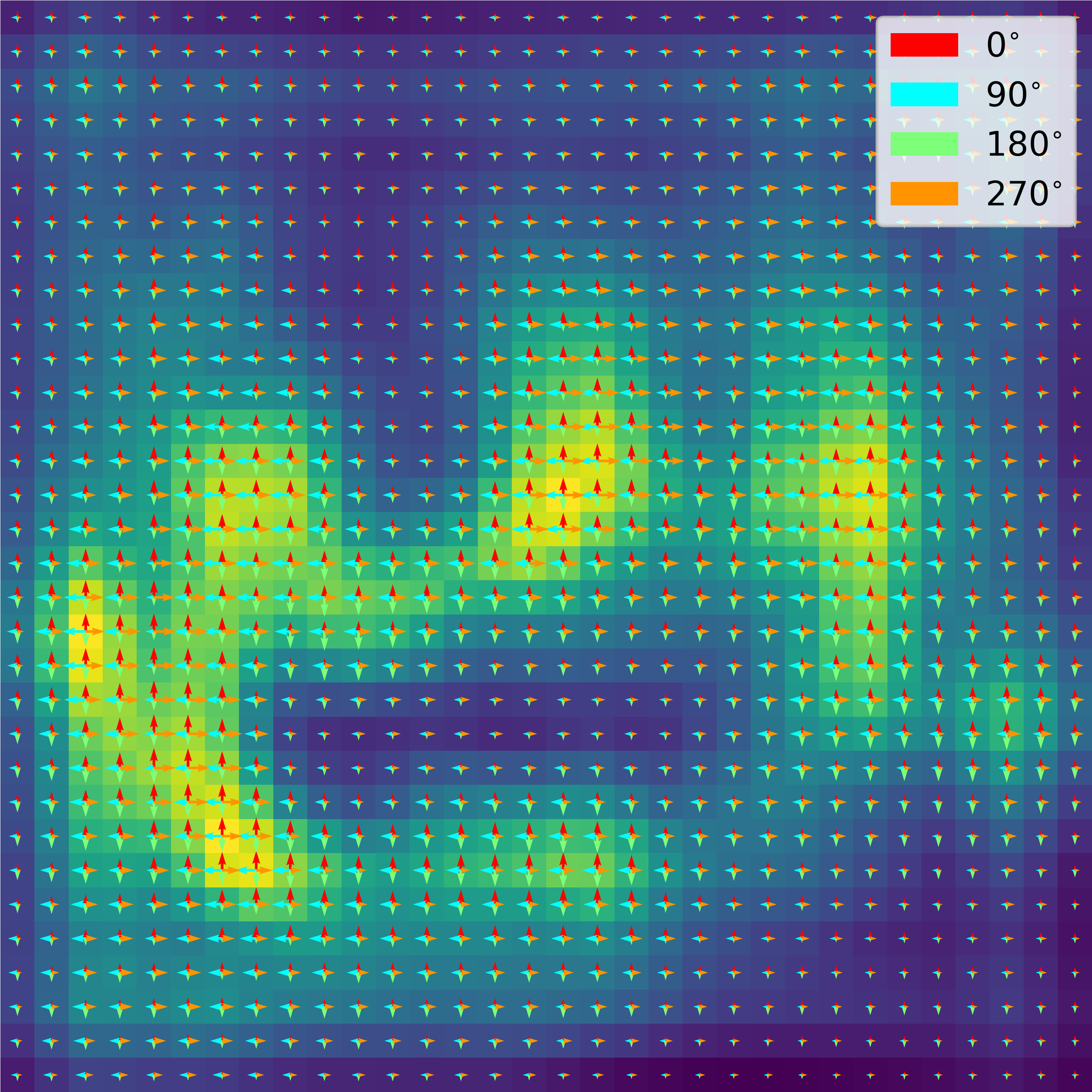}
    \end{subfigure}
    \caption{Examples of equivariant attention maps under the exact equivariance regime.}\label{fig:6_good_examples}
\end{figure*}

Importantly, this phenomenon does affect the resulting equivariant attention maps generated via attentive group convolutions as shown in Fig.~\ref{fig:6_bad_examples}. As these networks are only equivariant in an approximate manner, the generated attention maps are slightly deformed versions of one another for multiple orientations. To alleviate this problem, we replace all strided convolutions in the All-CNN and ResNet44 architectures by conventional convolutions with stride=1, followed by spatial max pooling. As a result, we can produce exactly equivariant attention maps as shown in Figs.~\ref{fig:6_examples},~\ref{fig:6_pcam_examples} and~\ref{fig:6_good_examples}.

% ==== Figure =====

% ==== Figure =====

% ==== Figure =====

%% file: backmatter/appx_7_gselfatt.tex
% path to figures directory
\graphicspath{{figures/7-gselfatt/}}

%=========================================================================

% \begin{savequote}[75mm]
% Nulla facilisi. In vel sem. Morbi id urna in diam dignissim feugiat. Proin molestie tortor eu velit. Aliquam erat volutpat. Nullam ultrices, diam tempus vulputate egestas, eros pede varius leo.
% \qauthor{Quoteauthor Lastname}
% \end{savequote}

\chapter{Group Equivariant Stand-Alone Self-Attention}
	%\label{chapter:}

%=========================================================================
\vspace{-7mm}
\section{Convolution and self-attention: A graphical comparison}

\sidecaptionvpos{figure}{t}
\begin{SCfigure}[50][h!]
    \centering
    \begin{subfigure}{0.12\textwidth}
    \centering
        \includegraphics[width=\textwidth]{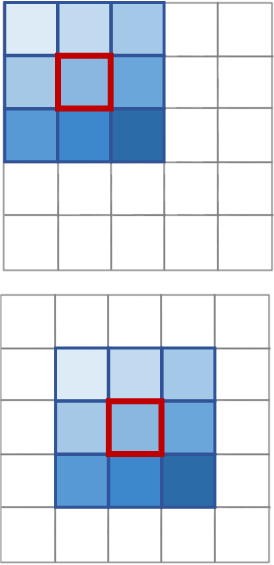}
        \caption{}
        \label{fig:7_conv_filters}
    \end{subfigure}
    \hspace{2mm}
    \begin{subfigure}{0.12\textwidth}
    \centering
        \includegraphics[width=\textwidth]{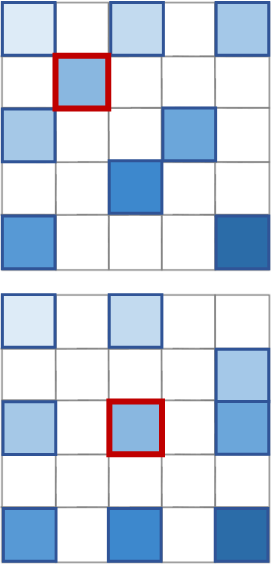}
        \caption{}
        \label{fig:7_att_filters}
    \end{subfigure}
    \hspace{2mm}
        \caption{Parameter usage in convolutional kernels (Fig.~\ref{fig:7_conv_filters}) and self-attention (Fig.~\ref{fig:7_att_filters}). Given a budget of 9 parameters, a convolutional filter ties these parameters to specific positions. Subsequently, these parameters remain static regardless of (\emph{i}) the query input position and (\emph{ii}) the input signal itself. Self-attention, on the other hand, does not tie parameters to any specific positions at all. %Contrarily, it compares the representations of all tokens falling in its receptive field. 
        As a result, provided enough heads, self-attention is more general than convolutions, as it can represent any convolutional kernel, e.g., Fig.~\ref{fig:7_conv_filters}, as well as several other functions defined on its receptive field.}
        \label{fig:7_comparison_att_conv}
\end{SCfigure}

\vspace{-7mm}
\section{Concepts From Group Theory}\label{appx:7_group_concepts}

\begin{definition}[\textbf{Group}]\label{def:group} A group is a tuple $(\gG, \cdot)$, where $\gG$ is a set and $\cdot: \gG \times \gG \rightarrow \gG$ is a binary operation on $\gG$, such that \emph{(i)} the set is closed under this operation, \emph{(ii)} the operation is associative, i.e., $(\gg_1 \cdot \gg_2) \cdot \gg_3 {=} \gg_1 \cdot (\gg_2 \cdot \gg_3)$, $\gg_1, \gg_2, \gg_3 \in \gG$, \emph{(iii)} there exists an identity element $\gr{e} \in \gG$ s.t. $\forall \gg \in \gG$ we have $\gr{e} \cdot \gg {=} \gg \cdot \gr{e} {=} \gg$, and \emph{(iv)} for each $\gg \in \gG$, there exists an inverse $\gg^{-1}$ s.t. $\gg \cdot \gg^{-1} {=} \gr{e}$.
\end{definition}
\begin{definition}[\textbf{Subgroup}]\label{def:subgroup} Let $(\gG, \cdot)$ be a group. A subset $\gH$ of $\gG$ is a subgroup of $\gG$ if $\gH$ is nonempty and closed under the group operation and inverses (i.e., $\gh_{1}, \gh_{2} \in \gH$ implies that $\gh_{1}^{-1} \in \gH$ and $\gh_{1} \cdot \gh_{2} \in \gH$). If $\gH$ is a subgroup of $\gG$ we write $\gH \leq \gG$
\end{definition}

\begin{definition}[\textbf{Semi-direct product and affine groups}]\label{def:affine_group} In practice, one is mainly interested in the analysis of data defined on $\sR^{d}$, and, consequently, in groups of the form $\gG = \sR^{d} \rtimes \gH$, resulting from the \textit{semi-direct product} ($\rtimes$) between the translation group ($\sR^{d}, +$) and an arbitrary (Lie) group $\gH$ that acts on $\sR^{d}$, e.g., rotation, scaling, mirroring, etc. This family of groups is referred to as \emph{affine groups} and their group product is defined as:
\begin{equation}\label{eq:7_affine_product}
    \gg_{1} \cdot \gg_{2} = (x_{1}, \gr{h}_{1}) \cdot (x_{2}, \gr{h}_{2}) = (x_{1} + \gr{h}_{1} \odot x_{2}, \gr{h}_{1} \cdot  \gr{h}_{2}),
\end{equation}
with $\gg = (x_{1}, \gr{h}_{1})$, $\gg_{2}=(x_{2}, \gr{h}_{2}) \in G$, $x_{1}$, $x_{2} \in \sR^{d}$ and $\gr{h}_{1}, \gr{h}_{2} \in \gH$. The operator $\odot$ denotes the \textit{action of $\gr{h} \in \gH$ on $x \in \sR^{d}$}, and it describes how a vector $x \in \sR^{d}$ is modified by elements $\gr{h} \in \gH$.
\end{definition}

\begin{definition}[\textbf{Group representation}]\label{def:group_repr} Let $\gG$ be a group and $\sL^{2}(X)$ be a space of functions defined on some vector space $X$. The \textit{(left) regular group representation} of $\gG$ is a linear transformation $\mathcal{L}: \gG \times \sL^{2}(X) \rightarrow \sL^{2}(X)$, $(\gg,f) \mapsto \mathcal{L}_{\gg}[f]\coloneqq f(\gg^{-1} \odot x)$, that shares the group structure via:
\begin{gather}
    \mathcal{L}_{\gg_1}\mathcal{L}_{\gg_2}[f] = \mathcal{L}_{\gg_1 \cdot \gg_2}[f]
\end{gather}
for any $\gg_1, \gg_2 \in G$, $f \in \mathbb{L}_{2}(X)$. That is, concatenating two such transformations, parameterized by $\gg_1$ and $\gg_2$, is equivalent to a single transformation parameterized by $\gg_1 \cdot \gg_2 \in \gG$. If the group $\gG$ is affine, the group representation $\mathcal{L}_{\gg}$ can be split as:
\begin{equation} \label{eq:7_repr_decomp}
\mathcal{L}_{g}[f] = \mathcal{L}_{x} \mathcal{L}_{\gh}[f],
\end{equation}
with $g=(x,\gh) \in \gG$, $x \in \mathbb{R}^{d}$ and $\gh \in \gH$. Intuitively, the representation of $\gG$ on a function $f$ describes how the function as a whole, i.e., $f(x), \forall x \in X$, is transformed by the effect of group elements $\gg \in \gG$.
\end{definition}

\vspace{-7mm}
\section{Actions and Representations of Groups Acting on Homogeneous Spaces for Functions Defined on Sets}\label{appx:7_homospaces}

In this section we show that the action of a group $\gG$ acting on a homogeneous space $\gX$ is well defined on sets $\gS$ gathered from $\gX$, and that it induces a group representation of functions defined on $\gS$.

Let $\gS = \{ \gi \}$ be a set and $\gX$ be a homogeneous space on which the action of $\gG$ is well-defined, i.e., $\gg x \in \gX, \forall \gg \in \gG, \ \forall x \in \gX$. Since $\gS$ has been gathered from $\gX$, there exists an injective map $x: \gS \rightarrow \gX$, that maps set elements $\gi \in \gS$ to unique elements $x_{\gi} \in \gX$. That is, there exists a map $x: \gi \mapsto x_{\gi}$ that assigns an unique value $x_{\gi} \in \gX$ to each $\gi \in \gS$ corresponding to the coordinates from which the set element has been gathered.

Since the action of $\gG$ is well defined on $\gX$, it follows that the left regular representation (Def.~\ref{def:group_repr}) $\gL_{\gg}[f^{\gX}](x_{\gi}) \coloneqq f^{\gX}(\gg^{-1} x_{\gi}) \in L_{\gY}(\gX)$ of functions $f^{\gX} \in L_{\gY}(\gX)$ exists and is well-defined. Since $x$ is injective, the left regular representation $\gL_{\gg}[f^{\gX}](x_{\gi}) = f^{\gX}(\gg^{-1} x_{\gi})$ can be expressed uniquely in terms of set indices as $ \gL_{\gg}[f^{\gX}](x_{\gi}) = f^{\gX}(\gg^{-1} x_{\gi}) = f^{\gX}(\gg^{-1} x(\gi))$. Furthermore, its inverse $x^{-1}: \gX \rightarrow \gS$, $x^{-1}: x_{\gi} \rightarrow \gi$ also exist and is well-defined. As a consequence, points $x_{\gi} \in \gX$ can be expressed uniquely in terms of set indices as $\gi = x^{-1}(x_{\gi}), \gi \in \gS$. Consequently, functions $f^{\gX} \in L_{\gY}(\gX)$ can be expressed in terms of functions $f \in L_{\gY}(\gS)$ by means of the equality $f^{\gX}(\gi) = f(x^{-1}(x_{\gi}))$. Resultantly, we see that the group representation $\gL_{\gg}[f^{\gX}](x_{\gi}) = f^{\gX}(\gg^{-1} x(\gi))$ can be described in terms of functions $f \in L_{\gY}(\gS)$ as:
\begin{equation*}
  \gL_{\gg}[f^{\gX}](x_{\gi}) = f^{\gX}(\gg^{-1} (x_{\gi})) =  f^{\gX}(\gg^{-1} x(\gi)) = f(x^{-1}(\gg^{-1} x(\gi))) = \gL_{\gg}[f](\gi),
\end{equation*}
with a corresponding group representation on $L_{\gY}(\gS)$ given by $\gL_{\gg}[f](\gi) = f(x^{-1}(\gg^{-1} x(\gi)))$, and an action of group elements $\gg \in \gG$ on set elements $\gi$ given by $\gg \gi \coloneqq x^{-1}(\gg x(\gi))$.

\vspace{-7mm}
\section{The Case of Non-Unimodular Groups: Self-Attention on the Scale-Translation Group} \label{appx:7_non_unimodular}

The lifting and group self-attention formulation provided in Sec.~\ref{sec:7_lifting_self_att} are only valid for unimodular groups. That is, for groups whose action does not change the volume of the objects they act upon, e.g., rotation, mirroring, etc. Non-unimodular groups, however, do modify the volume of the acted objects \citep{bekkers2020bspline}. The most relevant non-unimodular group for this work is the scale group $\gH = (\sR_{>0}, \times)$. To illustrate why this distinction is important, consider the following example:

Imagine we have a circle on $\sR^{2}$ of area $\pi r ^{2}$. If we rotate, mirror or translate the circle, its size is kept constant. If we increase its radius by a factor $\gh \in \sR_{>0}$, however, its size would increase by $\gh^2$. Imagine that we have an application for which we would like to recognize this circle regardless of any of these transformations by means of self-attention. For this purpose, we define a neighborhood $\gN$ for which the original circle fits perfectly. Since the size of the circle is not modified for any translated, rotated or translated versions of it, we would still be able to detect the circle regardless of these transformations. If we scale the circle by a factor of $\gh > 1$, however, the circle would fall outside of our neighborhood $\gN$ and hence, we would not be able to recognize it.

A solution to this problem is to scale our neighborhood $\gN$ in a proportional way. That is, if the circle is scaled by a factor $\gh \in \sR_{>0}$, we scale our neighborhood by the same factor $\gh$: $\gN \rightarrow \gh\gN$. Resultantly, the circle would fall within the neighborhood for any scale factor $\gh \in \sR_{>0}$. Unfortunately, there is a problem: self-attention utilizes summations over its neighborhood. Since $\sum_{\gi \in \gh \gN} \gi > \sum_{\gi \in \gN} \gi$, for $\gh > 1$, and $\sum_{\gi \in \gh \gN} \gi < \sum_{\gi \in \gN} \gi$, for $\gh < 1$, the result of the summations would still differ for different scales. Specifically, this result would always be bigger for larger versions of the neighborhood. This is problematic, as the response produced by the same circle, would still be different for different scales.

In order to handle this problem, one utilizes a normalization factor proportional to the change of size of the neighborhood considered. This ensures that the responses are equivalent for any scale $\gh \in \sR_{>0}$. As a result, we obtain that $\sum_{\gi \in \gh_{1} \gN} (\gh_{1}^{2})^{-1}\gi = \sum_{\gi \in \gh_{2} \gN} (\gh_{2}^{2})^{-1}\gi,$ $\forall \gh_{1}, \gh_{2} \in \sR_{>0}$.\footnote{The squared factor in $\gh_{1}^{2}$ and $\gh_{2}^{2}$ appears as a result that the neighborhood growth is quadratic in $\sR^{2}$.}

In the example above we have provided an intuitive description of the (left invariant) Haar measure $\du \mu(\gh)$.  As its name indicates, it is a measure defined on the group, which is invariant over all group elements $\gh \in \gH$. For several unimodular groups, the Haar measure corresponds to the Lebesgue measure as the volume of the objects the group acts upon is kept equal, i.e., $\du \mu(\gh) = \du \gh$.\footnote{This is why this subtlety is often left out in group equivariance literature.} For non-unimodular groups, however, the Haar measure requires a normalization factor proportional to the change of volume of these objects. Specifically, the Haar measure corresponds to the Lebesgue measure times a normalization factor $\gh^{d}$, where $d$ corresponds to the dimensionality of the space $\sR^{d}$ the group acts upon \citep{bekkers2020bspline, romero2020wavelet}, i.e., $\du \mu(\gh) = \tfrac{1}{\gh^{d}} \du \gh$.

In conclusion, in order to obtain group equivariance to non-unimodular groups, the lifting and group self-attention formulation provided in Eq.~\ref{eq:7_lifting_selfatt}, ~\ref{eq:7_group_selfatt} must be modified via normalization factors proportional to the group elements $\gh \in \gH$. Specifically, they are redefined as:
\begin{align}
    &\hspace{-2mm}\gr{m}^{r}_{\gG\uparrow}[f, \rho](\gi, \gh) = \varphi_{\text{out}}\Big( \bigcup_{h \in [H]} \hspace{-3mm} \sum_{\ \ \ \gj \in \gh \gN(i)} \hspace{-4.0mm} \tfrac{1}{\gh^{d}}\  \sigma\hspace{-0.5mm}_{\gj}\big(\langle \varphi^{(h)}_{\text{qry}}(f(\gi)), \varphi^{(h)}_{\text{key}}(f(\gi) + \gL_{\gh}[\rho](\gi, \gj)) \rangle \big) \varphi^{(h)}_{\text{val}}(f(\gj)) \Big)\label{eq:7_lifting_selfatt_nonunimodular}\\[-1\jot]
        &\hspace{-1.5mm}\gr{m}^{r}_{\gG}[f, \rho](\gi, \gh) = \varphi_{\text{out}}\Big( \bigcup_{h \in [H]} \sum_{\tilde{\gh} \in \gH}\hspace{-1.05cm} \sum_{\qquad \quad (\gj, \hat{\gh}) \in \gh \gN(i, \tilde{\gh})} \hspace{-0.6cm}\hspace{-5mm}\tfrac{1}{\gh^{d+1}}\sigma\hspace{-0.5mm}_{\gj, \hat{\gh}}\big(\langle \varphi_{\text{qry}}^{(h)}(f(\gi, \tilde{\gh})), \varphi_{\text{key}}^{(h)}(f(\gj, \hat{\gh}) \nonumber\\[-5\jot]
    &\hspace{7.1cm} + \gL_{\gh}[\rho]((\gi, \tilde{\gh}), (\gj, \hat{\gh})) \rangle \big) \varphi_{\text{val}}^{(h)}(f(\gj, \hat{\gh})) \Big).\label{eq:7_group_selfatt_nonunimodular}
\end{align}
The factor $d+1$ in Eq.~\ref{eq:7_group_selfatt_nonunimodular} results from the fact that the summation is now performed on the group $\gG = \sR^{d} \rtimes \gH$, an space of dimensionality $d+1$. An interesting case emerges when global neighborhoods are considered, i.e., s.t. $\gN(\gi) = \gS$, $\forall \gi \in \gS$. Since $\gh \gN(\gi) = \gN(\gi) = \gS$ for any $\gh > 1$, approximation artifacts are introduced. It is not clear if it is better to introduce normalization factors in these situations or not. An in-depth investigation of this phenomenon is left for future research.
\vspace{-7mm}
\subsection{Current empirical aspects of scale equivariant self-attention}
Self-attention suffers from quadratic memory and time complexity proportional to the size of the neighborhood considered. This constraint is particularly important for the scale group, for which these neighborhoods grow as a result of the group action. We envisage two possible solutions to this limitation left out for future research:

The most promising solution is given by incorporating recent advances in efficient self-attention in group self-attention, e.g., \citet{kitaev2020reformer, wang2020linformer, zaheer2020big, katharopoulos2020transformers, choromanski2020rethinking}. By reducing the quadratic complexity of self-attention, the current computational constraints of scale equivariant self-attention can be (strongly) reduced. The resulting architectures would be comparable to \citet{bekkers2020bspline, sosnovik2020scaleequivariant, romero2020wavelet} in terms of their functionality and the discretizations they can manage.

The second option is to draw a self-attention analogous to \citet{worrall2019deep}, where scale equivariance is implemented via dilated convolutions. One might consider an analogous to dilated convolutions via \enquote{sparse} dilations of the self-attention neighborhood. As a result, scale equivariance can be implemented while retaining an equal computational cost for all group elements. Importantly however, this strategy is viable for a dyadic set of scales only, i.e., a set of scales given by a set $\{2^j\}_{j = 0}^{j_\text{max}}$, and thus, less general that the scale-equivariant architectures listed before.

\vspace{-7mm}
\section{Experimental Details}\label{appx:7_exp_details}
\begin{figure}
    \centering
    \includegraphics[width=0.71\textwidth]{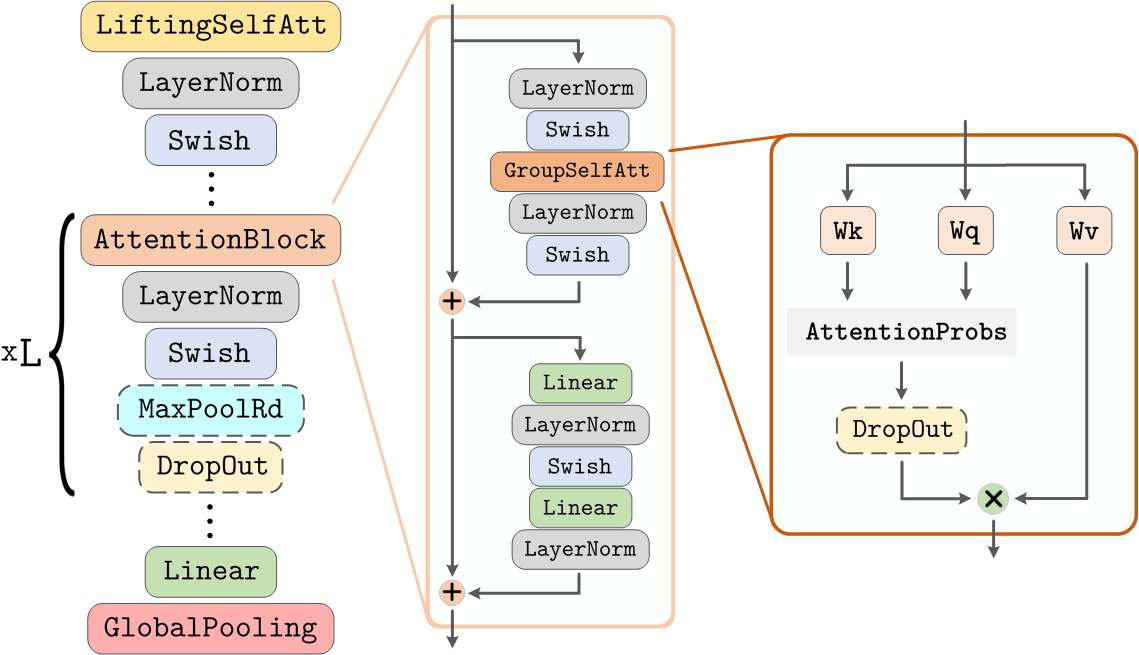}
    \vspace{-2mm}
        \caption{Graphical description of group self-attention networks. Dot-lined blocks depict optional blocks. Linear layers are applied point-wise across the feature map. Swish non-linearities \citep{ramachandran2017searching} and LayerNorm \citep{ba2016layer} are used all across the network. {\btt GlobalPooling} consists of max-pool over group elements followed by spatial mean-pool.}   \label{fig:7_net_structure}
\end{figure}

In this section we provide extended details over our implementation as well as the exact architectures and optimization schemes used in our experiments. All our models follow the structure shown in Fig.~\ref{fig:7_net_structure} and vary only in the number of blocks and channels. All self-attention operations utilize 9 heads. We utilize \texttt{PyTorch} for our implementation. Any missing specification can be safely considered to be the \texttt{PyTorch} default value

\vspace{-7mm}
\subsection{Rotated MNIST}
For rotational MNIST we use a group self-attention network composed of 5 attention blocks with 20 channels. We utilize automatic mixed precision during training to reduce memory requirements. {\btt attention\_dropout\_rate} and {\btt value\_dropout\_rate} are both set to 0.1.  We train for 300 epochs and utilize the Adam optimizer, batch size of 8, weight decay of 0.0001 and learning rate of 0.001. 
\vspace{-7mm}
\subsection{CIFAR-10}
For CIFAR-10 we use a group self-attention network composed of 6 attention blocks with 96 channels for the first two blocks and 192 channels for the rest.  {\btt attention\_dropout\_rate} and {\btt value\_dropout\_rate} are both set to 0.1. We use dropout on the input with a rate of 0.3 and additional dropout blocks of rate 0.2 followed by spatial max-pooling after the second and fourth block. We did not use automatic mixed precision training for this dataset as it made all models diverge. We perform training for 350 epochs and utilize stochastic gradient descent with a momentum of 0.9 and cosine learning rate scheduler with base learning rate 0.01 \citep{loshchilov2017sgdr}. We utilize a batch size of 24, weight decay of 0.0001 and He's initialization.
\vspace{-7mm}
\subsection{PatchCamelyon}
For PatchCamelyon we use a group self-attention network composed of 4 attention blocks with 12 channels for the first block, 24 channels for the second block, 48 channels for the third and fourth blocks and 96 channels for the last block. \texttt{attention\_dropout\_rate} and \texttt{value\_dropout\_rate} are both set to 0.1. We use an additional max-pooling block after the lifting block to reduce memory requirements. We did not use automatic mixed precision training for this dataset as it made all models diverge. We perform training for 100 epochs, utilize stochastic gradient descent with a momentum of 0.9 and cosine learning rate scheduler with base learning rate 0.01 \citep{loshchilov2017sgdr}. We utilize a batch size of 8, weight decay of 0.0001 and He's initialization.
\vspace{-7mm}
\section{Proofs} \label{appx:7_proofs}
\begin{proof}[\textbf{Proof of Proposition~\ref{claim:perm_equiv}.}]
If the self-attention formulation provided in Eqs.~\ref{eq:7_mhsa},~\ref{eq:7_full_attention_func}, is permutation equivariant, then it must hold that $\gr{m}[\gL_{\pi}[f]](\gi){=}\gL_{\pi}[\gr{m}[f]](\gi)$. Consider a permuted input signal $\gL_{\pi}[f](\gi) {=} f(\pi^{-1}(\gi))$. The self-attention operation on $\gL_{\pi}[f]$ is:
\begin{align*}
    \gr{m}\big[\gL_{\pi}[f]\big](\gi) &= \varphi_{\text{out}}\Big( \bigcup_{h \in [H]} \sum_{\gj \in \gS} \sigma\hspace{-0.5mm}_{\gj}\big(\langle \varphi^{(h)}_{\text{qry}}(\gL_{\pi}[f](\gi)), \varphi^{(h)}_{\text{key}}(\gL_{\pi}[f](\gj)) \rangle \big) \varphi^{(h)}_{\text{val}}(\gL_{\pi}[f](\gj))  \Big) \\
     =& \varphi_{\text{out}}\Big( \bigcup_{h \in [H]} \sum_{\gj \in \gS} \sigma\hspace{-0.5mm}_{\gj}\big(\langle \varphi^{(h)}_{\text{qry}}(f(\pi^{-1}(\gi))), \varphi^{(h)}_{\text{key}}(f(\pi^{-1}(\gj))) \rangle \big) \varphi^{(h)}_{\text{val}}(f(\pi^{-1}(\gj)))  \Big) \\
     =& \varphi_{\text{out}}\Big( \bigcup_{h \in [H]} \sum_{\pi(\bar{\gj}) \in \gS} \sigma\hspace{-0.5mm}_{\pi(\bar{\gj})}\big(\langle \varphi^{(h)}_{\text{qry}}(f(\bar{\gi})), \varphi^{(h)}_{\text{key}}(f(\bar{\gj})) \rangle \big) \varphi^{(h)}_{\text{val}}(f(\bar{\gj}))  \Big)\\
     =& \varphi_{\text{out}}\Big( \bigcup_{h \in [H]} \sum_{\bar{\gj} \in \gS} \sigma\hspace{-0.5mm}_{\bar{\gj}}\big(\langle \varphi^{(h)}_{\text{qry}}(f(\bar{\gi})), \varphi^{(h)}_{\text{key}}(f(\bar{\gj})) \rangle \big) \varphi^{(h)}_{\text{val}}(f(\bar{\gj}))  \Big)\\
     &=\gr{m}[f](\bar{\gi}) =\gr{m}[f](\pi^{-1}(\gi)) \\
     & = \gL_{\pi}\big[\gr{m}[f]\big](\gi).
\end{align*}
Here we have used the substitution $\bar{\gi}{=}\pi(\gi)$ and $\bar{\gj}{=}\pi(\gj)$.  Since the summation is defined over the entire set we have that $\sum_{\pi(\bar{\gj}) \in \gS}[\cdot] {=} \sum_{\bar{\gj} \in \gS}[\cdot]$.
Conclusively, we see that $\gr{m}[\gL_{\pi}[f]](\gi) = \gL_{\pi}[\gr{m}[f]](\gi)$. Hence, permutation equivariance indeed holds.
\end{proof}
\vspace{-5mm}
\begin{proof}[\textbf{Proof of Claim~\ref{claim:no_perm_equiv_no_trans_equiv}.}]
\textbf{\textit{Permutation equivariance.}} If the self-attention formulation provided in Eq.~\ref{eq:7_attention_scores_func_absolute} is permutation equivariant, then it must hold that $\gr{m}[\gL_{\pi}[f], \rho](\gi) = \gL_{\pi}[\gr{m}[f, \rho]](\gi)$. Consider a permuted input signal $\gL_{\pi}[f](\gi) {=} f(\pi^{-1}(\gi))$. The self-attention operation on $\gL_{\pi}[f]$ is given by:
\begin{align*}
    \gr{m}\big[\gL&_{\pi}[f], \rho\big](\gi)\\[-1\jot]
     &= \varphi_{\text{out}}\Big( \bigcup_{h \in [H]} \sum_{\gj \in \gN(i)}  \hspace{-1.5mm}\sigma\hspace{-0.5mm}_{\gj}\big(\langle \varphi^{(h)}_{\text{qry}}(\gL_{\pi}[f](\gi) + \rho(\gi)), \varphi^{(h)}_{\text{key}}(\gL_{\pi}[f](\gj) + \rho(\gj)) \rangle \big) \varphi^{(h)}_{\text{val}}(\gL_{\pi}[f](\gj)) \Big)
\end{align*}
As discussed in Sec.~\ref{sec:7_functional_attn}, since there exists permutations in $\sS_{N}$ able to send elements $\gj$ in $\gN(i)$ to elements $\tilde{\gj}$ outside of $\gN(i)$, it is trivial to show that Eq.~\ref{eq:7_attention_scores_func_absolute} is not equivariant to $\sS_{N}$. Consequently, in order to provide a more interesting analysis, we consider global attention here, i.e., cases where $\gN(i) = \gS$. As shown for Proposition~\ref{claim:perm_equiv}, this self-attention instantiation is permutation equivariant. Consequently, by considering this particular case, we are able to explicitly analyze the effect of introducing absolute positional encodings into the self-attention formulation. We have then that:
\begin{align*}
\gr{m}\big[\gL_{\pi}[f]&, \rho\big](\gi)\\[-2\jot]
      &= \varphi_{\text{out}}\Big( \bigcup_{h \in [H]} \sum_{\gj \in \gS}  \sigma\hspace{-0.5mm}_{\gj}\big(\langle \varphi^{(h)}_{\text{qry}}(f(\pi^{-1}(\gi)) + \rho(\gi)), \varphi^{(h)}_{\text{key}}(f(\pi^{-1}(\gj)) + \rho(\gj)) \rangle \big) \varphi^{(h)}_{\text{val}}(f(\pi^{-1}(\gj))) \Big) \\
      &= \varphi_{\text{out}}\Big( \bigcup_{h \in [H]} \sum_{\pi(\bar{\gj}) \in \gS}  \hspace{-2mm}\sigma\hspace{-0.5mm}_{\pi(\bar{\gj})}\big(\langle \varphi^{(h)}_{\text{qry}}(f(\bar{\gi}) + \rho(\pi({\bar{\gi})})), \varphi^{(h)}_{\text{key}}(f(\bar{\gj}) + \rho(\pi(\bar{\gj}))) \rangle \big) \varphi^{(h)}_{\text{val}}(f(\bar{\gj})) \Big) \\
      &= \varphi_{\text{out}}\Big( \bigcup_{h \in [H]} \sum_{\bar{\gj} \in \gS} \sigma\hspace{-0.5mm}_{\bar{\gj}}\big(\langle \varphi^{(h)}_{\text{qry}}(f(\bar{\gi}) + \rho(\pi(\bar{\gi}))), \varphi^{(h)}_{\text{key}}(f(\bar{\gj}) + \rho(\pi(\bar{\gj}))) \rangle \big) \varphi^{(h)}_{\text{val}}(f(\bar{\gj})) \Big)
\end{align*}
Here we have used the substitution $\bar{\gi}{=}\pi(\gi)$ and $\bar{\gj}{=}\pi(\gj)$.  Since the summation is defined over the entire set we have that $\sum_{\pi(\bar{\gj}) \in \gS}[\cdot] {=} \sum_{\bar{\gj} \in \gS}[\cdot]$. Since $\rho(\pi(\bar{\gi})) {\neq} \rho(\bar{\gi})$ and $\rho(\pi(\bar{\gj})) {\neq} \rho(\bar{\gj})$, we are unable to reduce the expression further towards the form of $\gr{m}[f, \rho](\bar{\gi})$. Hence, we conclude that absolute position-aware self-attention is not permutation equivariant.

\textbf{\textit{Translation equivariance.}} If the self-attention formulation provided in Eq.~\ref{eq:7_attention_scores_func_absolute}, is translation equivariant, then it must hold that $\gr{m}(\gL_{y}[f], \rho)(\gi) {=} \gL_{y}[\gr{m}(f, \rho)](\gi)$. Consider a translated input signal $\gL_{y}[f](\gi) {=} f(x^{-1}(x(\gi) - y))$. Self-attention on $\gL_{y}[f]$ is given by:
\begin{align*}
    \gr{m}\big[&\gL_{y}[f], \rho\big](\gi) \\[-1\jot]
    &= \varphi_{\text{out}}\Big( \bigcup_{h \in [H]} \sum_{\gj \in \gN(i)}  \hspace{-1.5mm}\sigma\hspace{-0.5mm}_{\gj}\big(\langle \varphi^{(h)}_{\text{qry}}(\gL_{y}[f](\gi) + \rho(\gi)),\varphi^{(h)}_{\text{key}}(\gL_{y}[f](\gj) + \rho(\gj)) \rangle \big) \varphi^{(h)}_{\text{val}}(\gL_{y}[f](\gj)) \Big)\\
    &= \varphi_{\text{out}}\Big( \bigcup_{h \in [H]} \sum_{\gj \in \gN(i)}  \hspace{-1.5mm}\sigma\hspace{-0.5mm}_{\gj}\big(\langle \varphi^{(h)}_{\text{qry}}(f(x^{-1}(x(\gi) - y)) + \rho(\gi)),\\[-4.5\jot]
    &\hspace{5cm} \varphi^{(h)}_{\text{key}}(f(x^{-1}(x(\gj) - y)) + \rho(\gj)) \rangle \big) \varphi^{(h)}_{\text{val}}(f(x^{-1}(x(\gj) - y))) \Big)\\
    &= \varphi_{\text{out}}\Big( \bigcup_{h \in [H]} \hspace{-3.1cm}\sum_{\qquad \quad \qquad \qquad \qquad x^{-1}(x(\bar{\gj}) + y)) \in \gN(x^{-1}(x(\bar{\gi}) + y))}  \hspace{-3.15cm}\sigma\hspace{-0.5mm}_{x^{-1}(x(\bar{\gj}) + y)}\big(\langle \varphi^{(h)}_{\text{qry}}(f(\bar{\gi}) + \rho(x^{-1}(x(\bar{\gi}) + y))),\\[-4.5\jot]
    &\hspace{6.9cm} \varphi^{(h)}_{\text{key}}(f(\bar{\gj}) + \rho(x^{-1}(x(\bar{\gj}) + y))) \rangle \big) \varphi^{(h)}_{\text{val}}(f(\bar{\gj})) \Big)\\
    &= \varphi_{\text{out}}\Big( \bigcup_{h \in [H]} \hspace{-3.1cm}\sum_{\qquad \qquad \quad \qquad \qquad x^{-1}(x(\bar{\gj}) + y)) \in \gN(x^{-1}(x(\bar{\gi}) + y))}  \hspace{-3.15cm}\sigma\hspace{-0.5mm}_{x^{-1}(x(\bar{\gj}) + y)}\big(\langle \varphi^{(h)}_{\text{qry}}(f(\bar{\gi}) + \rho^P(x(\bar{\gi}) + y)),\\[-4.5\jot]
    &\hspace{7.5cm} \varphi^{(h)}_{\text{key}}(f(\bar{\gj}) + \rho^P(x(\bar{\gj}) + y)) \rangle \big) \varphi^{(h)}_{\text{val}}(f(\bar{\gj})) \Big)
\end{align*}
Here, we have used the substitution $\bar{\gi} {=} x^{-1}(x(\gi) - y) \Rightarrow \gi {=} x^{-1}(x(\bar{\gi}) + y)$ and $\bar{\gj} {=} x^{-1}(x(\gj) - y) \Rightarrow \gj {=} x^{-1}(x(\bar{\gj}) + y)$. Since the area of summation remains equal to any translation $y \in \sR^{d}$, we have that: $$\sum_{x^{-1}(x(\bar{\gj}) + y) \in \gN(x^{-1}(x(\bar{\gi}) + y))}  \hspace{-1.5cm} [\cdot] \hspace{1.2cm} = \sum_{x^{-1}(x(\bar{\gj})) \in \gN(x^{-1}(x(\bar{\gi})))}  \hspace{-1.2cm} [\cdot] \hspace{1.0cm} = \sum_{\bar{\gj} \in \gN(\bar{\gi})} \hspace{-1mm} [\cdot]. $$
Hence, we can further reduce the expression above as:
\begin{align*}
 \gr{m}\big[&\gL_{y}[f], \rho\big](\gi) \\[-1\jot]
    &= \varphi_{\text{out}}\Big( \bigcup_{h \in [H]} \sum_{\bar{\gj} \in \gN(\bar{\gi})}  \hspace{-1.5mm}\sigma_{\bar{\gj}}\big(\langle \varphi^{(h)}_{\text{qry}}(f(\bar{\gi}) + \rho^P(x(\bar{\gi}) + y)), \varphi^{(h)}_{\text{key}}(f(\bar{\gj}) + \rho^P(x(\bar{\gj}) + y)) \rangle \big) \varphi^{(h)}_{\text{val}}(f(\bar{\gj})) \Big)
\end{align*}
Since, $\rho(\bar{\gi}) + y \neq \rho(\bar{\gi})$ and $\rho(\bar{\gj}) + y \neq \rho(\bar{\gj})$, we are unable to reduce the expression further towards the form of $\gr{m}(f, \rho)(\bar{\gi})$. Consequently, we conclude that the absolute positional encoding does not allow for translation equivariance either.
\end{proof}
\vspace{-3mm}
\begin{proof}[\textbf{Proof of Claim~\ref{claim:trans_equiv}.}]\label{proof:trans_equiv_rel_pos}
If the self-attention formulation provided in Eq.~\ref{eq:7_relaive_pos_att_func} is translation equivariant, then it must hold that $\gr{m}^{r}[\gL_{y}[f], \rho](\gi) {=} \gL_{y}[\gr{m}^{r}[f, \rho]](\gi)$. Consider a translated input signal $\gL_{y}[f](\gi) {=} f(x^{-1}(x(\gi) - y))$. Self-attention on $\gL_{y}[f]$ is given by:
\begin{align*}
    \gr{m}&^{r}\big[\gL_{y}[f], \rho\big](\gi)\\[-1\jot]
    &= \varphi_{\text{out}}\Big( \bigcup_{h \in [H]} \sum_{\gj \in \gN(i)} \hspace{-1.5mm}\sigma\hspace{-0.5mm}_{\gj}\big(\langle \varphi^{(h)}_{\text{qry}}(\gL_{y}[f](\gi)), \varphi^{(h)}_{\text{key}}(\gL_{y}[f](\gi) + \rho(\gi, \gj)) \rangle \big) \varphi^{(h)}_{\text{val}}(\gL_{y}[f](\gj)) \Big)\\
    &= \varphi_{\text{out}}\Big( \bigcup_{h \in [H]} \sum_{\gj \in \gN(i)} \hspace{-1.5mm}\sigma\hspace{-0.5mm}_{\gj}\big(\langle \varphi^{(h)}_{\text{qry}}(f(x^{-1}(x(\gi) - y))), \varphi^{(h)}_{\text{key}}(f(x^{-1}(x(\gj) - y))\\[-4.5\jot]
    &\hspace{8.3cm} + \rho(\gi, \gj)) \rangle \big) \varphi^{(h)}_{\text{val}}(f(x^{-1}(x(\gj) - y))) \Big) \\
    &= \varphi_{\text{out}}\Big( \bigcup_{h \in [H]}\hspace{-2.8cm}\sum_{\qquad \qquad \qquad \qquad x^{-1}(x(\bar{\gj}) + y) \in \gN(x^{-1}(x(\bar{\gi}) + y))}  \hspace{-2.95cm}\sigma\hspace{-0.5mm}_{x^{-1}(x(\bar{\gj}) + y)}\big(\langle \varphi^{(h)}_{\text{qry}}(f(\bar{\gi})), \varphi^{(h)}_{\text{key}}(f(\bar{\gj}) \\[-4.5\jot]
    &\hspace{6.4cm}+ \rho(x^{-1}(x(\bar{\gi}) + y), x^{-1}(x(\bar{\gj}) + y))) \rangle \big) \varphi^{(h)}_{\text{val}}(f(\bar{\gj})) \Big)
\end{align*}
Here, we have used the substitution $\bar{\gi} {=} x^{-1}(x(\gi) - y) \Rightarrow \gi {=} x^{-1}(x(\bar{\gi}) + y)$ and $\bar{\gj} {=} x^{-1}(x(\gj) - y) \Rightarrow \gj {=} x^{-1}(x(\bar{\gj}) + y)$. By using the definition of $\rho(\gi, \gj)$ we can further reduce the expression above as:
\begin{align*}
    \quad &= \varphi_{\text{out}}\Big( \bigcup_{h \in [H]}\hspace{-2.8cm}\sum_{\qquad \qquad \qquad \qquad x^{-1}(x(\bar{\gj}) + y) \in \gN(x^{-1}(x(\bar{\gi}) + y))}  \hspace{-2.95cm}\sigma\hspace{-0.5mm}_{x^{-1}(x(\bar{\gj}) + y)}\big(\langle \varphi^{(h)}_{\text{qry}}(f(\bar{\gi})), \varphi^{(h)}_{\text{key}}(f(\bar{\gj}) \\[-5\jot]
    &\hspace{7.3cm}+ \rho^{P}(x(\bar{\gj}) + y - (x(\bar{\gi}) + y))) \rangle \big) \varphi^{(h)}_{\text{val}}(f(\bar{\gj})) \Big) \\
    &= \varphi_{\text{out}}\Big( \bigcup_{h \in [H]}\hspace{-2.8cm}\sum_{\qquad \qquad \qquad \qquad x^{-1}(x(\bar{\gj}) + y) \in \gN(x^{-1}(x(\bar{\gi}) + y))}  \hspace{-2.95cm}\sigma\hspace{-0.5mm}_{x^{-1}(x(\bar{\gj}) + y)}\big(\langle \varphi^{(h)}_{\text{qry}}(f(\bar{\gi})), \varphi^{(h)}_{\text{key}}(f(\bar{\gj}) + \rho^{P}(x(\bar{\gj}) - x(\bar{\gi}))) \rangle \big) \varphi^{(h)}_{\text{val}}(f(\bar{\gj})) \Big)\\
    &= \varphi_{\text{out}}\Big( \bigcup_{h \in [H]}\hspace{-2.8cm}\sum_{\qquad \qquad \qquad \qquad x^{-1}(x(\bar{\gj}) + y) \in \gN(x^{-1}(x(\bar{\gi}) + y))}  \hspace{-2.95cm}\sigma\hspace{-0.5mm}_{x^{-1}(x(\bar{\gj}) + y)}\big(\langle \varphi^{(h)}_{\text{qry}}(f(\bar{\gi})), \varphi^{(h)}_{\text{key}}(f(\bar{\gj}) + \rho(\bar{\gi}, \bar{\gj})) \rangle \big) \varphi^{(h)}_{\text{val}}(f(\bar{\gj})) \Big)
\end{align*}
Since the area of the summation remains equal to any translation $y \in \sR^{d}$, we have that:
$$\sum_{x^{-1}(x(\bar{\gj}) + y) \in \gN(x^{-1}(x(\bar{\gi}) + y))}  \hspace{-1.5cm} [\cdot] \hspace{1.2cm} = \sum_{x^{-1}(x(\bar{\gj})) \in \gN(x^{-1}(x(\bar{\gi})))}  \hspace{-1.2cm} [\cdot] \hspace{1.0cm} = \sum_{\bar{\gj} \in \gN(\bar{\gi})} \hspace{-1mm} [\cdot].  $$ Resultantly, we can further reduce the expression above as:
\begin{align*}
    \gr{m}^{r}\big[\gL_{y}[f], \rho\big](\gi) &= \varphi_{\text{out}}\Big( \bigcup_{h \in [H]}\sum_{\bar{\gj}\in \gN(\bar{\gi})}  \hspace{-1.5mm}\sigma\hspace{-0.5mm}_{\bar{\gj}}\big(\langle \varphi^{(h)}_{\text{qry}}(f(\bar{\gi})), \varphi^{(h)}_{\text{key}}(f(\bar{\gj}) + \rho(\bar{\gi},\bar{\gj})) \rangle \big) \varphi^{(h)}_{\text{val}}(f(\bar{\gj})) \Big)\\
    &= \gr{m}^{r}[f, \rho](\bar{\gi}) = \gr{m}^{r}[f, \rho](x^{-1}(x(\gi) - y))\\
    &= \gL_{y}\big[\gr{m}^{r}[f, \rho]\big](\gi)
\end{align*}
We see that indeed $\gr{m}^{r}[\gL_{y}[f], \rho](\gi) {=} \gL_{y}[\gr{m}^{r}[f, \rho]](\gi)$. Hence, the relative positional encoding allows for translation equivariance. We emphasize that this is a consequence of the fact that $\rho(x^{-1}(x(\bar{\gi}) + y), x^{-1}(x(\bar{\gj}) + y))) {=} \rho(\bar{\gi}, \bar{\gj}), \forall y \in \sR^{d}$. That is, it comes from the fact that relative positional encoding is invariant to the action of the translation group.
\end{proof}
\vspace{-3mm}
\begin{proof}[\textbf{Proof of Claim \ref{claim:lifting_gequiv}.}]
If the lifting self-attention formulation provided in Eq.~\ref{eq:7_relaive_pos_att_func} is $\gG$-equivariant, then it must hold that $\gr{m}^{r}_{\gG\uparrow}[\gL_{\gg}[f], \rho](\gi, \gh) {=} \gL_{\gg}[\gr{m}^{r}_{\gG\uparrow}[f, \rho]](\gi, \gh)$. Consider a $\gg$-transformed input signal $\gL_{\gg}[f](\gi) {=} \gL_{y}\gL_{\tilde{\gh}}[f](\gi) {=} f(x^{-1}(\tilde{\gh}^{-1} (x(\gi) - y)))$, $\gg {=} (y, \tilde{\gh})$, $y \in \sR^{d}$, $\tilde{\gh} \in \gH$. The lifting group self-attention operation on $\gL_{\gg}[f]$ is given by:
\begin{align*}
    \gr{m}&^{r}_{\gG\uparrow}\big[\gL_{\gr{y}}\gL_{\tilde{\gh}}[f], \rho\big](\gi, \gh) \\[-1.5\jot]
    &=\varphi_{\text{out}}\Big( \bigcup_{h \in [H]} \sum_{\gj \in \gN(i)} \hspace{-2mm}\sigma\hspace{-0.5mm}_{\gj}\big(\langle \varphi^{(h)}_{\text{qry}}(\gL_{y}\gL_{\tilde{\gh}}[f](\gi)), \varphi^{(h)}_{\text{key}}(\gL_{y}\gL_{\tilde{\gh}}[f](\gi) \\[-6\jot]
    &\hspace{8.3cm}+ \gL_{\gh}[\rho](\gi, \gj)) \rangle \big) \varphi^{(h)}_{\text{val}}(\gL_{y}\gL_{\tilde{\gh}}[f] (\gj)) \Big)\\
    &=\varphi_{\text{out}}\Big( \bigcup_{h \in [H]} \sum_{\gj \in \gN(i)} \hspace{-2mm}\sigma\hspace{-0.5mm}_{\gj}\big(\langle \varphi^{(h)}_{\text{qry}}(f(x^{-1}(\tilde{\gh}^{-1} (x(\gi) - y)))), \varphi^{(h)}_{\text{key}}(f(x^{-1}(\tilde{\gh}^{-1} (x(\gj) - y))) \\[-4\jot]
    &\hspace{6.7cm} + \gL_{\gh}[\rho](\gi, \gj)) \rangle \big) \varphi^{(h)}_{\text{val}}(f(x^{-1}(\tilde{\gh}^{-1} (x(\gj) - y)))) \Big)\\
    &=\varphi_{\text{out}}\Big( \bigcup_{h \in [H]} \hspace{-3.1cm}\sum_{\qquad \qquad \quad \qquad \qquad x^{-1}(\tilde{\gh}x(\bar{\gj}) + y) \in \gN(x^{-1}(\tilde{\gh}x(\bar{\gi}) + y))} \hspace{-3.25cm}\sigma\hspace{-0.5mm}_{x^{-1}(\tilde{\gh}x(\bar{\gj}) + y)}\big(\langle \varphi^{(h)}_{\text{qry}}(f(\bar{\gi})), \varphi^{(h)}_{\text{key}}(f(\bar{\gj})\\[-3\jot]
    &\hspace{6.0cm} + \gL_{\gh}[\rho](x^{-1}(\tilde{\gh}x(\bar{\gi}) + y), x^{-1}(\tilde{\gh}x(\bar{\gj}) + y))) \rangle \big) \varphi^{(h)}_{\text{val}}(f(\bar{\gj})) \Big)
\end{align*}
Here we have used the substitution $\bar{\gi} {=} x^{-1}(\tilde{\gh}^{-1} (x(\gi) - y)) \Rightarrow \gi {=} x^{-1}(\tilde{\gh}x(\bar{\gi}) + y)$ and $\bar{\gj} {=} x^{-1}(\tilde{\gh}^{-1} (x(\gj) - y)) \Rightarrow \gj {=} x^{-1}(\tilde{\gh}x(\bar{\gj}) + y)$. By using the definition of $\rho(\gi, \gj)$ we can further reduce the expression above as:
\begin{align*}
    \quad &=\varphi_{\text{out}}\Big( \bigcup_{h \in [H]} \hspace{-3.1cm}\sum_{\qquad \qquad \quad \qquad \qquad x^{-1}(\tilde{\gh}x(\bar{\gj}) + y) \in \gN(x^{-1}(\tilde{\gh}x(\bar{\gi}) + y))} \hspace{-3.25cm}\sigma\hspace{-0.5mm}_{x^{-1}(\tilde{\gh}x(\bar{\gj}) + y)}\big(\langle \varphi^{(h)}_{\text{qry}}(f(\bar{\gi})), \varphi^{(h)}_{\text{key}}(f(\bar{\gj}) \\[-5\jot]
    &\hspace{6.0cm}+ \rho^{P}(\gh^{-1}(\tilde{\gh}x(\bar{\gj}) + y) - \gh^{-1}(\tilde{\gh}x(\bar{\gi}) + y)) \rangle \big) \varphi^{(h)}_{\text{val}}(f(\bar{\gj})) \Big) \\
    &= \varphi_{\text{out}}\Big( \bigcup_{h \in [H]} \hspace{-3.1cm}\sum_{\qquad \qquad \quad \qquad \qquad x^{-1}(\tilde{\gh}x(\bar{\gj}) + y) \in \gN(x^{-1}(\tilde{\gh}x(\bar{\gi}) + y))} \hspace{-3.25cm}\sigma\hspace{-0.5mm}_{x^{-1}(\tilde{\gh}x(\bar{\gj}) + y)}\big(\langle \varphi^{(h)}_{\text{qry}}(f(\bar{\gi})), \varphi^{(h)}_{\text{key}}(f(\bar{\gj}) \\[-5.5\jot]
    &\hspace{7.5cm} + \rho^{P}(\gh^{-1}\tilde{\gh}(x(\bar{\gj}) - x(\bar{\gi})))) \rangle \big) \varphi^{(h)}_{\text{val}}(f(\bar{\gj})) \Big)\\
    &= \varphi_{\text{out}}\Big( \bigcup_{h \in [H]} \hspace{-3.1cm}\sum_{\qquad \qquad \quad \qquad \qquad x^{-1}(\tilde{\gh}x(\bar{\gj}) + y) \in \gN(x^{-1}(\tilde{\gh}x(\bar{\gi}) + y))} \hspace{-3.25cm}\sigma\hspace{-0.5mm}_{x^{-1}(\tilde{\gh}x(\bar{\gj}) + y)}\big(\langle \varphi^{(h)}_{\text{qry}}(f(\bar{\gi})), \varphi^{(h)}_{\text{key}}(f(\bar{\gj}) + \gL_{\tilde{\gh}^{-1}\gh}[\rho](\bar{\gi},\bar{\gj})) \rangle \big) \varphi^{(h)}_{\text{val}}(f(\bar{\gj})) \Big)
\end{align*}
Since, for unimodular groups, the area of summation remains equal for any $\gg \in \gG$, we have that:
$$\sum_{x^{-1}(\tilde{\gh}x(\bar{\gj}) + y) \in \gN(x^{-1}(\tilde{\gh}x(\bar{\gi}) + y))}  \hspace{-1.5cm} [\cdot] \hspace{1.2cm} = \sum_{x^{-1}(\tilde{\gh}x(\bar{\gj})) \in \gN(x^{-1}(\tilde{\gh}x(\bar{\gi})))}  \hspace{-1.2cm} [\cdot] \hspace{1.0cm} = \sum_{x^{-1}(x(\bar{\gj})) \in \gN(x^{-1}(x(\bar{\gi})))}  \hspace{-1.05cm} [\cdot] \hspace{0.95cm} = \sum_{\bar{\gj} \in \gN(\bar{\gi})} \hspace{-1mm} [\cdot]. $$ Resultantly, we can further reduce the expression above as:
\vspace{-2mm}
\begin{align*}
    \gr{m}^{r}_{\gG\uparrow}\big[\gL_{\gr{y}}\gL_{\tilde{\gh}}[f], \rho&\big](\gi, \gh) \\[-1\jot]
    &= \varphi_{\text{out}}\Big( \bigcup_{h \in [H]} \sum_{\bar{\gj} \in \gN(\bar{\gi})} \hspace{-1.5mm}\sigma\hspace{-0.5mm}_{\bar{\gj}}\big(\langle \varphi^{(h)}_{\text{qry}}(f(\bar{\gi})), \varphi^{(h)}_{\text{key}}(f(\bar{\gj}) + \gL_{\tilde{\gh}^{-1}\gh}[\rho](\bar{\gi},\bar{\gj})) \rangle \big) \varphi^{(h)}_{\text{val}}(f(\bar{\gj})) \Big)\\
    &=\gr{m}^{r}_{\gG\uparrow}[f, \rho](\bar{\gi}, \tilde{\gh}^{-1}\gh) = \gr{m}^{r}_{\gG\uparrow}[f, \rho](x^{-1}(\tilde{\gh}^{-1} (x(\gi) - y)), \tilde{\gh}^{-1}\gh) \\
    &= \gL_{\gr{y}}\gL_{\tilde{\gh}}\big[\gr{m}^{r}_{\gG\uparrow}[f, \rho]\big](\gi, \gh).
\end{align*}
We see indeed that $\gr{m}^{r}_{\gG\uparrow}[\gL_{\gr{y}}\gL_{\tilde{\gh}}[f], \rho](\gi, \gh) {=} \gL_{\gr{y}}\gL_{\tilde{\gh}}[\gr{m}^{r}_{\gG\uparrow}[f, \rho]](\gi, \gh)$. Consequently, we conclude that the lifting group self-attention operation is group equivariant. We emphasize once more that this is a consequence of the fact that $\gL_{g}[\rho](\gi, \gj) {=} \rho(\gi, \gj), \ \forall \gg \in \gG$. In other words, it comes from the fact that the positional encoding used is invariant to the action of elements $\gg \in \gG$.
\end{proof}
\vspace{-5mm}
\begin{proof}[\textbf{Proof of Claim~\ref{claim:g_equivariance_g_attention}.}]
If the group self-attention formulation provided in Eq.~\ref{eq:7_relaive_pos_att_func} is $\gG$-equivariant, then it must hold that $\gr{m}^{r}_{\gG}[\gL_{\gg}[f], \rho](\gi, \gh) {=} \gL_{\gg}[\gr{m}^{r}_{\gG}[f, \rho]](\gi, \gh)$. Consider a $\gg$-transformed input signal $\gL_{\gg}[f](\gi, \tilde{h}) {=} \gL_{y}\gL_{\bar{\gh}}[f](\gi, \tilde{\gh}) = f(\rho^{-1}(\bar{\gh}^{-1}(\rho(\gi) - y)), \bar{\gh}\tilde{\gh})$, $\gg {=} (y, \bar{\gh})$, $y \in \sR^{d}$, $\bar{\gh} \in \gH$. The group self-attention operation on $\gL_{\gg}[f]$ is given by:
\begin{align*}
    \gr{m}&^{r}_{\gG}\big[\gL_{y}\gL_{\bar{\gh}}[f], \rho\big](\gi, \gh) \\[-1.5\jot]
    &= \varphi_{\text{out}}\Big( \bigcup_{h \in [H]} \sum_{\tilde{\gh} \in \gH}\hspace{-1.05cm} \sum_{\qquad \quad (\gj, \hat{\gh}) \in \gN(i, \tilde{\gh})} \hspace{-0.6cm}\hspace{-5mm}\sigma\hspace{-0.5mm}_{\gj, \hat{\gh}}\big(\langle \varphi_{\text{qry}}^{(h)}(\gL_{y}\gL_{\bar{\gh}}[f](\gi, \tilde{\gh})), \varphi_{\text{key}}^{(h)}(\gL_{y}\gL_{\bar{\gh}}[f](\gj, \hat{\gh})\\[-5\jot]
    &\hspace{6.7cm} + \gL_{\gh}[\rho]((\gi, \tilde{\gh}), (\gj, \hat{\gh})) \rangle \big) \varphi_{\text{val}}^{(h)}(\gL_{y}\gL_{\bar{\gh}}[f](\gj, \hat{\gh})) \Big) \\
    &= \varphi_{\text{out}}\Big( \bigcup_{h \in [H]} \sum_{\tilde{\gh} \in \gH}\hspace{-1.05cm} \sum_{\qquad \quad (\gj, \hat{\gh}) \in \gN(i, \tilde{\gh})} \hspace{-0.6cm}\hspace{-5mm}\sigma\hspace{-0.5mm}_{\gj, \hat{\gh}}\big(\langle \varphi_{\text{qry}}^{(h)}(f(x^{-1}(\bar{\gh}^{-1}(x(\gi) - y)), \bar{\gh}^{-1}\tilde{\gh})), \varphi_{\text{key}}^{(h)}(f(x^{-1}(\bar{\gh}^{-1}(x(\gj) - y)), \bar{\gh}^{-1}\hat{\gh}) \nonumber\\[-5\jot]
    &\hspace{5.5cm} + \gL_{\gh}[\rho]((\gi, \tilde{\gh}), (\gj, \hat{\gh})) \rangle \big) \varphi_{\text{val}}^{(h)}(f(x^{-1}(\bar{\gh}^{-1}(x(\gj) - y)), \bar{\gh}^{-1}\hat{\gh})) \Big)\\
    &= \varphi_{\text{out}}\Big( \bigcup_{h \in [H]} \sum_{\bar{\gh}\tilde{\gh}' \in \gH} \hspace{-5cm}\sum_{\qquad\qquad\qquad\qquad\qquad\qquad \quad (x^{-1}(\bar{\gh}x(\bar{\gj}) + y), \bar{\gh}\hat{\gh}') \in \gN(x^{-1}(\bar{\gh}x(\bar{\gi}) + y), \bar{\gh}\tilde{\gh}')} \hspace{-4.8cm}\sigma\hspace{-0.5mm}_{x^{-1}(\bar{\gh}x(\bar{\gj}) + y), \bar{\gh}\hat{\gh}'}\big(\langle \varphi_{\text{qry}}^{(h)}(f(\bar{\gi}, \tilde{\gh}')), \varphi_{\text{key}}^{(h)}(f(\bar{\gj}, \hat{\gh}')  + \gL_{\gh}[\rho]((x^{-1}(\bar{\gh}x(\bar{\gi}) + y), \bar{\gh}\tilde{\gh}'),\nonumber\\[-5\jot]
    &\hspace{9.4cm} (x^{-1}(\bar{\gh}x(\bar{\gj}) + y), \bar{\gh}\hat{\gh}')) \rangle \big) \varphi_{\text{val}}^{(h)}(f(\bar{\gj}, \hat{\gh}')) \Big)
\end{align*}
Here we have used the substitutions $\bar{\gi} {=} x^{-1}(\bar{\gh}^{-1} (x(\gi) - y)) \Rightarrow \gi {=} x^{-1}(\bar{\gh}x(\bar{\gi}) + y))$, $\tilde{\gh}' {=} \bar{\gh}^{-1}\tilde{\gh}$, and $\bar{\gj} {=} x^{-1}(\bar{\gh}^{-1} (x(\gj) - y)) \Rightarrow \gi {=} x^{-1}(\bar{\gh}x(\bar{\gi}) + y))$, $\hat{\gh}' {=} \bar{\gh}^{-1}\hat{\gh}$. By using the definition of $\rho((\gi, \tilde{\gh}), (\gj, \hat{\gh}))$ we can further reduce the expression above as:
\begin{align*}
& =\varphi_{\text{out}}\Big( \bigcup_{h \in [H]} \sum_{\bar{\gh}\tilde{\gh}' \in \gH} \hspace{-5cm}\sum_{\qquad\qquad\qquad\qquad\qquad\qquad \quad (x^{-1}(\bar{\gh}x(\bar{\gj}) + y), \bar{\gh}\hat{\gh}') \in \gN(x^{-1}(\bar{\gh}x(\bar{\gi}) + y), \bar{\gh}\tilde{\gh}')} \hspace{-4.8cm}\sigma\hspace{-0.5mm}_{x^{-1}(\bar{\gh}x(\bar{\gj}) + y), \bar{\gh}\hat{\gh}'}\big(\langle \varphi_{\text{qry}}^{(h)}(f(\bar{\gi}, \tilde{\gh}')), \varphi_{\text{key}}^{(h)}(f(\bar{\gj}, \hat{\gh}') \\[-2\jot] &\hspace{5.65cm} + \rho^P( \gh^{-1}(\bar{\gh}x(\bar{\gj}) + y - (\bar{\gh}x(\bar{\gi}) + y)), \gh^{-1}\bar{\gh}\tilde{\gh}'^{-1}\hat{\gh}')) \rangle \big) \varphi_{\text{val}}^{(h)}(f(\bar{\gj}, \hat{\gh}')) \Big)\\
& = \varphi_{\text{out}}\Big( \bigcup_{h \in [H]} \sum_{\bar{\gh}\tilde{\gh}' \in \gH} \hspace{-5cm}\sum_{\qquad\qquad\qquad\qquad\qquad\qquad \quad (x^{-1}(\bar{\gh}x(\bar{\gj}) + y), \bar{\gh}\hat{\gh}') \in \gN(x^{-1}(\bar{\gh}x(\bar{\gi}) + y), \bar{\gh}\tilde{\gh}')} \hspace{-4.8cm}\sigma\hspace{-0.5mm}_{x^{-1}(\bar{\gh}x(\bar{\gj}) + y), \bar{\gh}\hat{\gh}'}\big(\langle \varphi_{\text{qry}}^{(h)}(f(\bar{\gi}, \tilde{\gh}')), \varphi_{\text{key}}^{(h)}(f(\bar{\gj}, \hat{\gh}')\\[-2\jot] &\hspace{7.65cm}  + \rho^P( \gh^{-1}\bar{\gh}(x(\bar{\gj})- x(\bar{\gi}), \tilde{\gh}'^{-1}\hat{\gh}'))) \rangle \big)  \varphi_{\text{val}}^{(h)}(f(\bar{\gj}, \hat{\gh}')) \Big)\\
& = \varphi_{\text{out}}\Big( \bigcup_{h \in [H]} \sum_{\bar{\gh}\tilde{\gh}' \in \gH} \hspace{-5cm}\sum_{\qquad\qquad\qquad\qquad\qquad\qquad \quad (x^{-1}(\bar{\gh}x(\bar{\gj}) + y), \bar{\gh}\hat{\gh}') \in \gN(x^{-1}(\bar{\gh}x(\bar{\gi}) + y), \bar{\gh}\tilde{\gh}')} \hspace{-4.8cm}\sigma\hspace{-0.5mm}_{x^{-1}(\bar{\gh}x(\bar{\gj}) + y), \bar{\gh}\hat{\gh}'}\big(\langle \varphi_{\text{qry}}^{(h)}(f(\bar{\gi}, \tilde{\gh}')), \varphi_{\text{key}}^{(h)}(f(\bar{\gj}, \hat{\gh}')  \\[-2\jot] &\hspace{7.65cm} + \gL_{\bar{\gh}^{-1}\gh}[\rho]((\bar{\gi}, \tilde{\gh}'),(\bar{\gj}, \hat{\gh}'))) \rangle \big)  \varphi_{\text{val}}^{(h)}(f(\bar{\gj}, \hat{\gh}')) \Big)
\end{align*}
Furthermore, since for unimodular groups the area of summation remains equal for any transformation $\gg \in \gG$, we have that:
\begin{align*}
    \sum_{(x^{-1}(\bar{\gh}x(\bar{\gj}) + y), \bar{\gh}\hat{\gh}') \in \gN(x^{-1}(\bar{\gh}x(\bar{\gi}) + y), \bar{\gh}\tilde{\gh}')}  \hspace{-2cm} [\cdot] \hspace{1.5cm} &= \hspace{-1mm} \sum_{(x^{-1}(\bar{\gh}x(\bar{\gj})), \bar{\gh}\hat{\gh}') \in \gN(x^{-1}(\bar{\gh}x(\bar{\gi})), \bar{\gh}\tilde{\gh}')}  \hspace{-1.7cm} [\cdot] \hspace{1.2cm} \\
    & = \hspace{-1mm} \sum_{(x^{-1}(x(\bar{\gj})), \hat{\gh}') \in \gN(x^{-1}(x(\bar{\gi})), \tilde{\gh}')}  \hspace{-1.45cm} [\cdot] \hspace{1cm} = \hspace{-1mm}\sum_{(\bar{\gj}, \hat{\gh}') \in \gN(\bar{\gi}, \tilde{\gh}')} \hspace{-0.5cm} [\cdot].
\end{align*}
Additionally, we have that $ \sum_{\bar{\gh}\tilde{\gh}' \in \gH}[\cdot] = \sum_{\tilde{\gh}' \in \gH}[\cdot]$. Resultantly, we can further reduce the expression above as:
\begin{align*}
\gr{m}^{r}_{\gG}\big[\gL_{\gr{y}}\gL_{\bar{\gh}}[f], \rho\big](\gi, \gh) &= \varphi_{\text{out}}\Big( \bigcup_{h \in [H]}\sum_{\tilde{\gh}' \in \gH} \sum_{ (\bar{\gj}, \hat{\gh}') \in \gN(\bar{\gi}, \tilde{\gh}')} \hspace{-0.65cm}\sigma\hspace{-0.5mm}_{\bar{\gj}, \hat{\gh}'}\big(\langle \varphi_{\text{qry}}^{(h)}(f(\bar{\gi}, \tilde{\gh}')), \varphi_{\text{key}}^{(h)}(f(\bar{\gj}, \hat{\gh}')  + \nonumber\\[-5\jot]
&\hspace{4.5cm}\gL_{\bar{\gh}^{-1}\gh}[\rho]((\bar{\gi}, \tilde{\gh}'),(\bar{\gj}, \hat{\gh}'))) \rangle \big)  \varphi_{\text{val}}^{(h)}(f(\bar{\gj}, \hat{\gh}')) \Big)\\
&=\gr{m}^{r}_{\gG}[f, \rho](\bar{\gi}, \bar{h}^{-1}\gh) = \gr{m}^{r}_{\gG}[f, \rho](x^{-1}(\bar{\gh}^{-1} (x(\gi) - y)), \bar{h}^{-1}\gh) \\
&= \gL_{y}\gL_{\bar{h}}\big[\gr{m}^{r}_{\gG}[f, \rho]\big](\gi, \gh).
\end{align*}
We see that indeed $\gr{m}^{r}_{\gG}[\gL_{y}\gL_{\bar{h}}[f], \rho](\gi, \gh) {=} \gL_{y}\gL_{\bar{h}}[\gr{m}^{r}_{\gG}[f, \rho]](\gi, \gh)$. Hence, we can conclude that the group self-attention operation is group equivariant. We emphasize once more that this is a consequence of the fact that $\gL_{\gg}[\rho]((\gi, \tilde{\gh}), (\gj, \hat{\gh})) {=} \rho((\gi, \tilde{\gh}), (\gj, \hat{\gh})), \ \forall \gg \in \gG$. That is, it comes from the fact that the positional encoding used is invariant to the action of elements $\gg \in \gG$.
\end{proof}

%% file: backmatter/appx_8_waveletnets.tex
% path to figures directory
\graphicspath{{figures/8-waveletnets/}}

%=========================================================================

% \begin{savequote}[75mm]
% Nulla facilisi. In vel sem. Morbi id urna in diam dignissim feugiat. Proin molestie tortor eu velit. Aliquam erat volutpat. Nullam ultrices, diam tempus vulputate egestas, eros pede varius leo.
% \qauthor{Quoteauthor Lastname}
% \end{savequote}

\chapter{Scale-Translation Equivariant Learning From Raw Time-Series}

\vspace{-7mm}
\section{Group and group action}\label{appx:group_and_action}
\textbf{Group.} A group is an ordered pair $(\gG, \cdot)$ where $\gG$ is a set and $\cdot: \gG \times \gG \rightarrow \gG$ is a binary operation on $\gG$, such that \emph{(i)} the set is closed under this operation, \emph{(ii)} the operation is associative, i.e., $(g_1 \cdot g_2) \cdot g_3 = g_1 \cdot (g_2 \cdot g_3)$, $g_1, g_2, g_3 \in \gG$, \emph{(iii)} there exists an identity element $e \in \gG$ such that $\forall g \in \gG$ we have $e \cdot g = g \cdot e = g$, and \emph{(iv)} for each $g \in \gG$, there exists an inverse $g^{-1}$ such that $g \cdot g^{-1} = e$.

\textbf{Subgroup.} Given a group $(\gG, \cdot)$, we say that a subset $\gH$ is a subgroup of $\gG$ if the tuple $(\gH, \cdot)$ also complies to the group axioms. For example, the set of rotations by $90^\circ$, $\gH{=}\{0^\circ, 90^\circ, 180^\circ, 270^\circ\}$, is a subgroup of the continuous rotation group as it also complies to the group axioms.

\textbf{Group action.}
Let $\gG$ be a group and $\gX$ be a set. The (left) group action of $\gG$ on $\gX$ is a function 
\begin{equation}
\setlength{\abovedisplayskip}{4pt}
\setlength{\belowdisplayskip}{4pt}
\gA: \gG \times \gX \rightarrow \gX, \ \ \ \gA_{g}: x \rightarrow x',
\end{equation}
such that for any $g_1$, $g_2 \in \gG$, $\gA_{g_2 g_1} {=} \gA_{g_2} \circ \gA_{g_1}$.  
In other words, the action of $\gG$ on $\gX$ describes how the elements in the set $x \in \gX$ are transformed by elements $g \in \gG$. For brevity, $\gA_{g}(x)$ is written as $gx$.

\vspace{-7mm}
\section{Equivariance properties of time-frequency transforms}\label{appx:equiv_gral}
\vspace{-5mm}
\subsection{The Fourier transform}\label{appx:equiv_fourtransf}
The Fourier transform represents a function with finite energy $f \in \ltwo(\mathbb{R})$ as a sum of complex sinusoidal waves $\eu^{\iu \omega t} {=} \cos{\omega t} + \iu \sin{\omega t}$:
\begin{equation}
\setlength{\abovedisplayskip}{0pt}
\setlength{\belowdisplayskip}{4pt}
    f(t) = \frac{1}{2\pi}\int_{- \infty}^{\infty}\hat{f}(\omega)\ \eu^{\iu \omega t} \, \du \omega\nonumber,
\end{equation}
where, $\hat{f}(\omega)$ depicts the amplitude of each component $\eu^{\iu \omega t}$ in $f$. The \textit{Fourier transform} $\mathcal{F}$ is defined as:
\begin{equation}
\setlength{\abovedisplayskip}{4pt}
\setlength{\belowdisplayskip}{4pt}
    \mathcal{F}[f](\omega) = \hat{f}(\omega) = \langle f, \eu^{\iu \omega t} \rangle =  \int_{- \infty}^{\infty}f(t)\ \eu^{- \iu \omega t} \, \du t. \nonumber
\end{equation}
In other words, the Fourier transform encodes $f$ into a time-frequency dictionary $\mathcal{D}{=}\{\eu^{i\omega t}\}_{\omega \in \mathbb{R}}$. 

\textbf{Input translation.} Let $\mathcal{L}_{t_{0}}[f](t){=}f(t-t_0)$ be a translated version of $f$. Its Fourier transform is given by:
  \begin{align}
  \setlength{\abovedisplayskip}{4pt}
\setlength{\belowdisplayskip}{4pt}
    \mathcal{F}[\mathcal{L}_{t_{0}}[f]](\omega) &= \int_{- \infty}^{\infty}f(t -t_0)\ \eu^{- \iu \omega t} \, \du t \ \Big|\ \tilde{t}{=}t -t_0\text{; } \du \tilde{t} {=} \du t  \nonumber \\
    &=\int_{- \infty}^{\infty}f(\tilde{t})\ \eu^{- \iu \omega (\tilde{t} + t_0)} \, \du \tilde{t}  = \eu^{- \iu \omega t_0} \int_{- \infty}^{\infty}f(\tilde{t})\ \eu^{- \iu \omega \tilde{t}} \, \du \tilde{t} = \eu^{- \iu \omega t_0}  \mathcal{F}[f](\omega) \label{eq:four_transequiv}
  \end{align}
In other words, a translation of $t_0$ corresponds to a phase modulation of $\eu^{- \iu \omega t_0}$ in the frequency domain. 
% In order to prove the translation equivariance property of the Fourier transformation, it is sufficient to show that the a the corresponding atom that reacts to the shifted input is a member of $\mathcal{D}$. Let $f(t - \bar{t})$ be a translated version of $f$. The corresponding Fourier transformation is:
% \begin{equation}
%     \int_{- \infty}^{\infty}f(t - \bar{t})\ \eu^{- i \omega t} \, \du t  \stackrel{\hat{t} = t - \bar{t}}{=} \int_{- \infty}^{\infty}f(t)\ \eu^{- i \omega (\hat{t} + \bar{t})} \, \du \hat{t} 
% \end{equation}
% And hence, a translation in the input domain is equivalent to a change in phase to the input domain. Note that the phase is indeed defined in a domain $[0, 2\pi)$. However, since the main assumption of the Fourier transformation is that the signals being analyzed are periodic. It is indeed a translation equivariant mapping.

\textbf{Input scaling.} Let $\mathcal{L}_{s_{0}}[f](t) {=} f(s_0^{-1} t)$, $s_{0} \in \mathbb{R}_{>0}$, be a scaled version of $f$. Its Fourier transform equals:
  \begin{align}
  \setlength{\abovedisplayskip}{4pt}
\setlength{\belowdisplayskip}{4pt}
    \mathcal{F}[\mathcal{L}_{s_{0}}[f]](\omega) &= \int_{- \infty}^{\infty}f(s_0^{-1} t)\ \eu^{- \iu \omega t} \, \du t\ \Big|\ \tilde{t}{=}s_0^{-1} t\text{; } \du \tilde{t} {=} s_0^{-1} \du t  \nonumber\\
    &=\int_{- \infty}^{\infty}f(\tilde{t})\ \eu^{- \iu \omega (s_{0}\tilde{t})} \, \du (s_{0}\tilde{t}) = s_{0} \int_{- \infty}^{\infty}f(\tilde{t})\ \eu^{- \iu (s_{0}\omega) \tilde{t}} \, \du \tilde{t} \nonumber\\
    &=  s_{0} \mathcal{F}[f](s_{0}\omega) = s_{0} \mathcal{L}_{s_{0}^{-1}}[\mathcal{F}[f]](\omega) \label{eq:four_scaleequiv}
  \end{align}
In other words,  we observe that a dilation on the time domain produces a compression in the Fourier domain times the inverse of the dilation. 

\textbf{Simultaneous input translation and scaling.} Following the same derivation procedure, we can show the behavior of the Fourier transform to simultaneous translations and dilations of the input:
\begin{equation}
\setlength{\abovedisplayskip}{4pt}
\setlength{\belowdisplayskip}{4pt}
    \mathcal{F}[\mathcal{L}_{s_0}\mathcal{L}_{t_0}[f]](\omega) = s_{0}\eu^{- \iu \omega t_0} \mathcal{F}[f](s_{0}\omega)=\eu^{- \iu \omega t_0} s_{0} \mathcal{L}_{s_{0}^{-1}}[\mathcal{F}[f]](\omega).
\end{equation}
This corresponds to the superposition of the previously exhibited behaviours.

\textbf{Effect of input transformations on the spectral density.} The spectral density of a function $f \in \ltwo(\mathbb{R})$ is given by $\mathopen|\mathcal{F}[f](\omega)\mathopen|^{2}$. Input translations and dilations produce the following transformations:
\begin{align}
\setlength{\abovedisplayskip}{4pt}
\setlength{\belowdisplayskip}{4pt}
\mathopen|\mathcal{F}[\mathcal{L}_{t_0}[f]](\omega)\mathclose|^{2}&=\mathopen|\mathcal{F}[f](\omega)\mathclose|^{2} \label{eq:four_transequiv_2}\\
\mathopen|\mathcal{F}[\mathcal{L}_{s_0}[f]](\omega)\mathclose|^{2}&=\mathopen|s_{0}\mathclose|^{2}\mathopen|\mathcal{L}_{s_0^{-1}}[\mathcal{F}[f]](\omega)\mathclose|^{2} \label{eq:four_scaleequiv_2}
\end{align}
 \textbf{Equivariance and invariance properties of the Fourier transform.} From Eq.~\ref{eq:four_transequiv} we can see that the Fourier transform is translation equivariant as it encodes translations of the input as a phase modulation of the output. In addition, it is also scale equivariant (Eq.~\ref{eq:four_scaleequiv}), as it encodes dilations of the input as a modulation of the frequency components~in~the~output. We can prove that the Fourier transform is dilation and translation equivariant by showing that the output transformations $\eu^{- \iu \omega t_0}$ and $s_{0} \mathcal{L}_{s_{0}^{-1}}$ are group representations of the translation and scaling group in the Fourier space.

\begin{tcolorbox}[enhanced, frame hidden, sharp corners, before skip=5pt, after skip=10pt]
\textbf{Group representation.} Let $\gG$ be a group and $f$ be a function on a given functional space $ L_{\gV}(\gX)$. The (left) regular representation of $\gG$ is a linear transformation $\gL: \gG \times L_{\gV}(\gX) \rightarrow L_{\gV}(\gX)$ which extends group actions to functions on $ L_{\gV}(\gX)$ by:
\begin{equation}
\setlength{\abovedisplayskip}{4pt}
\setlength{\belowdisplayskip}{4pt}
\gL_{g}: f \rightarrow f', \ \ f'(\gA_{g}(x)) = f(x) \Leftrightarrow f'(x) = f(g^{-1}x),  \nonumber
\end{equation}
such that for any  $g_1$, $g_2 \in \gG$, $\gL_{g_2 g_1} {=} \gL_{g_2} \circ \gL_{g_1}$. In other words, the group representation describes how a function on a functional space $f \in L_{\gV}(\gX)$ is modified by the effect of group elements $g \in \gG$. 
\end{tcolorbox}

We can show that the combination of input translations $t_0, t_1 \in \mathbb{R}$ or dilations $s_0, s_1 \in \mathbb{R}_{>0}$ produces a transformation on the Fourier domain that preserves the group structure. In other words, that the transformations previously outlined are group representations. Specifically, for $\mathcal{L}_{t_1}[\mathcal{L}_{t_0}[f]]$ and $\mathcal{L}_{s_1}[\mathcal{L}_{s_0}[f]]$ it holds:
\begin{align*}
\setlength{\abovedisplayskip}{4pt}
\setlength{\belowdisplayskip}{4pt}
 \mathcal{F}\big[\mathcal{L}_{t_1}[\mathcal{L}_{t_0}[f]]\big](\omega) &= \eu^{- \iu \omega t_1}\eu^{- \iu \omega t_0}\mathcal{F}[f](\omega)= \eu^{- \iu \omega (t_1 + t_0)}\mathcal{F}[f](\omega) = \mathcal{L}^\mathrm{Fourier}_{t_1 + t_0}[\mathcal{F}[f]](\omega) \\
 \mathcal{F}\big[\mathcal{L}_{s_1}[\mathcal{L}_{s_0}[f]]\big](\omega) &= s_{1} \mathcal{L}_{s_{1}^{-1}}\big[ s_{0} \mathcal{L}_{s_{0}^{-1}}[\mathcal{F}[f]]\big](\omega)= (s_{0}s_{1}) \mathcal{F}[f](s_1 s_0 \omega) = \mathcal{L}^\mathrm{Fourier}_{s_1 s_0}[\mathcal{F}[f]](\omega)
\end{align*}
with $\mathcal{L}^\mathrm{Fourier}_{t}[\mathcal{F}[f]](\omega) {=} \eu^{- \iu \omega t} \mathcal{F}[f](\omega) $ the representation of the Fourier transform for the translation group, and $\mathcal{L}^\mathrm{Fourier}_{s}[\mathcal{F}[f]](\omega) {=} s \mathcal{F}[f](s \omega)$ the representation of the Fourier transform for the dilation group.

Unfortunately, the resulting group representations rapidly become cumbersome specially in the presence of several input components. In addition, although the calculation of the spectral density leaves the scale equivariance property of the transformation unaffected, Eq.~\ref{eq:four_transequiv_2} shows that it reduces translation equivariance of the Fourier transform to \textit{translation invariance}. This is why the Fourier transform is commonly considered not to carry positional information.% It is important to note that this lost of information destroys the invertibility property of the Fourier transform.
\vspace{-7mm}
\subsection{The short-time Fourier transform} \label{appx:equiv_wind_four}
The short-time Fourier transform of a signal $f \in \ltwo(\mathbb{R})$ is given by:
\begin{equation}
\setlength{\abovedisplayskip}{4pt}
\setlength{\belowdisplayskip}{4pt}
    \mathcal{S}[f](t, \omega) = \int_{-\infty}^{+\infty}f(\tau) w(\tau - t)\  {\rm e}^{-\iu \omega \tau} \, {\rm d}\tau. \nonumber
\end{equation}
In other words, it encodes the input $f$ into a time-frequency dictionary $\mathcal{D}{=}\{\phi_{t, \omega}\}$, $\phi_{t, \omega} {=}  w(\tau - t)\  {\rm e}^{-\iu \omega \tau}$.

 \textbf{Input translation.} Let $\mathcal{L}_{t_{0}}[f](\tau){=}f(\tau{-}t_0)$ be a translated version of $f$. Its short-time Fourier transform is given by:
  \begin{align}
    \mathcal{S}[\mathcal{L}_{t_{0}}[f]](t, \omega) &= \int_{- \infty}^{\infty}\hspace{-3mm}f(\tau -t_0) w(\tau-t) \ \eu^{- \iu \omega \tau} \, \du \tau \ \Big| \ \tilde{t}{=}\tau -t_0\text{; } \du \tilde{t} {=} \du \tau  \nonumber \\
    &=\int_{- \infty}^{\infty}\hspace{-3mm}f(\tilde{t})w(\tilde{t}+t_{0}-t) \ \eu^{- \iu \xi (\tilde{t} + t_0)} \, \du \tilde{t} = \eu^{- \iu \xi t_0} \int_{- \infty}^{\infty}\hspace{-3mm}f(\tilde{t})w(\tau-(t-t_{0})) \ \eu^{- \iu \omega \tilde{t}} \, \du \tilde{t} \nonumber\\
    &= \eu^{- \iu \omega t_0}  \mathcal{S}[f](t-t_{0}, \omega) = \eu^{- \iu \omega t_0}  \mathcal{L}_{t_{0}}[\mathcal{S}[f]](t, \omega) \label{eq:win_four_transequiv}
  \end{align}
In other words, a translation by $t_{0}$ in the time domain, corresponds to a shift by $t_{0}$ on the time axis of the short-time Fourier transform, and an additional phase modulation of $\eu^{- \iu \xi t_0}$ similar to that of the Fourier transform (Eq.~\ref{eq:four_transequiv}).
\begin{figure*}
    \centering
    \includegraphics[width=0.9
    \textwidth]{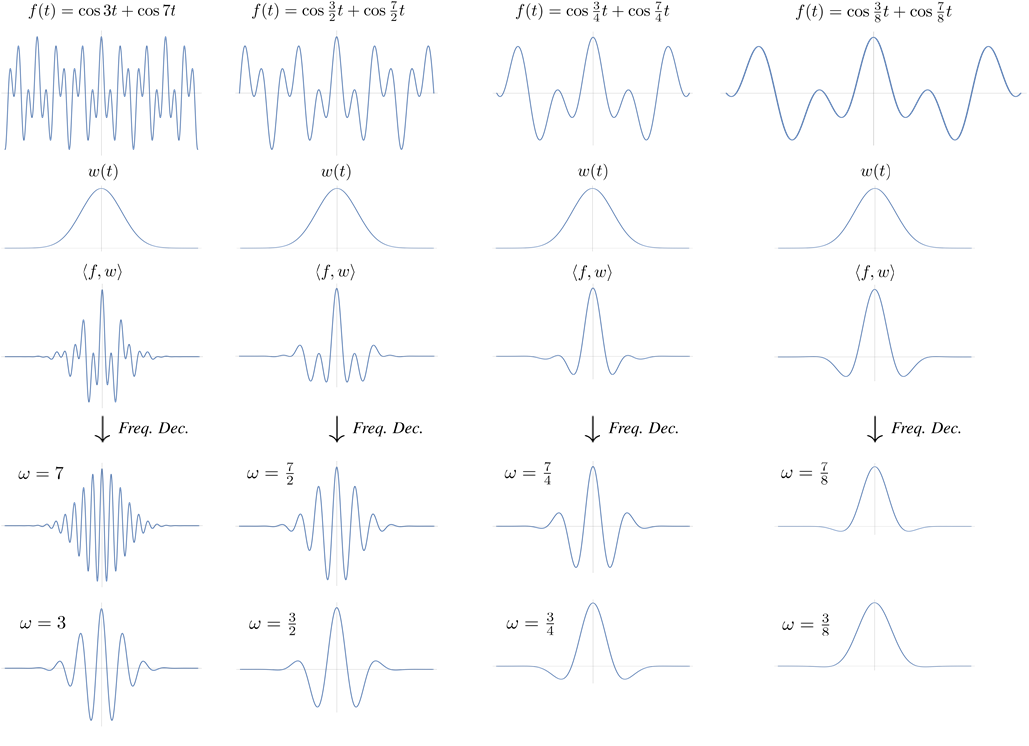}
    \vspace{-5mm}
    \caption{Scale equivariance of the short-time Fourier transform. Consider a function $f(t) {=} \cos{\omega_{1} t} + \cos{\omega_{2}t}$ composed of two frequencies $\omega_{1} {=} 3$ and $\omega_{2} {=} 7$, and a window function $w(t)$, with which the short-time Fourier transform is calculated. For relatively high frequencies (left column), the dot-product of $f$ and $w$, $\langle f, w \rangle$, is able to capture sufficient spectral information from $f$ to correctly extract the frequencies $\omega_{1}, \omega_{2}$ from it. However, for dilated versions of the same signal $f$ (right columns) obtained by reducing the frequency of the spectral components $\omega_{1}, \omega_{2}$ of $f$, the capacity of the dot-product $\langle f, w \rangle$ to capture the spectral information in the input gradually degrades and, eventually, is entirely lost. Consequently, scale equivariance holds (approximately) for scales for which \textit{all} of the spectral components of the signal $f$ lie within the range of the window $w$.
    \vspace{-4mm}}
    \label{fig:equiv_in_STFT}
\end{figure*}

 \textbf{Input scaling.} Let $\mathcal{L}_{s_{0}}[f](\tau) {=} f(s_0^{-1} \tau)$, $s_{0} \in \mathbb{R}_{>0}$, be a scaled version of $f$. Its short-time Fourier transform is given by:
  \begin{align}
    \mathcal{S}[\mathcal{L}_{s_{0}}[f]](t, \omega) &= \int_{- \infty}^{\infty}\hspace{-1.5mm}f(s_0^{-1}t) w(\tau-t) \ \eu^{- \iu \omega \tau} \, \du \tau \ \Big| \ \tilde{t}{=}s_0^{-1}\tau\text{; } \du \tilde{t} {=} s_0^{-1} \du \tau  \nonumber\\
    &=\int_{- \infty}^{\infty}\hspace{-1.5mm}f(\tilde{t})w(s_{0}\tilde{t}-t)\ \eu^{- \iu \omega (s_{0}\tilde{t})} \, \du (s_{0}\tilde{t}) = s_{0} \int_{- \infty}^{\infty}\hspace{-1.5mm}f(\tilde{t})w(s_{0}\tilde{t}-t)\ \eu^{- \iu (s_{0}\omega) \tilde{t}} \, \du \tilde{t} \nonumber \ \Big|\ t{=}s_0^{-1}s_{0} t\\
    &= s_{0} \int_{- \infty}^{\infty}\hspace{-1.5mm}f(\tilde{t})w(s_{0}(\tilde{t}-s_{0}^{-1}t))\ \eu^{- \iu (s_{0}\omega) \tilde{t}} \, \du \tilde{t}\nonumber \ \Big|\ w(s \tau) \approx w(\tau)\\
    &\approx  s_{0} \int_{- \infty}^{\infty}\hspace{-1.5mm}f(\tilde{t})w(\tilde{t}-s_{0}^{-1}t)\ \eu^{- \iu (s_{0}\omega) \tilde{t}} \, \du \tilde{t} \approx s_{0}\ \mathcal{S}[f](s_{0}^{-1}t, s_{0}\omega) \label{eq:win_four_scaleequiv}
  \end{align}
In other words, a dilation in the time domain produces a compression in the frequency domain analogous to the Fourier transform (Eq.~\ref{eq:four_scaleequiv}). However, it is important to note that we rely on the approximate $w(x){\approx} w(s x)$ to arrive to the final expression. Nevertheless, it is important to note that this approximate \textit{does not generally holds in practice}. This approximation implies that the window function $w$ is invariant to scaling, which holds only for increasing window sizes, i.e., when the short-term Fourier transform starts to approximate the (global) Fourier transform.
%Since $w$ has local support, this property holds as long as the size of the frequency components modulated by $s_{0}$ can be \enquote{contained} within the window $w$. To visualize this, consider a window of spatial support $\mathfrak{s}_{w}$. Resultantly, only frequencies whose bandwidth $\mathfrak{s}_{f} \leq \mathfrak{s}_{w}$ can be captured by the Fourier transform within the window $w$. Resultantly, $w$ is not linear, nor equivariant to scalings $s_{0} \in \mathbb{R}^{+}$ for which $\mathfrak{s}_{f} > \mathfrak{s}_{w}$ (Fig. X). 

\textbf{Simultaneous input translation and scaling.} Following the same derivation procedure, we can show the behavior of the short-time Fourier transform to simultaneous translations and scaling. We have that:
\begin{equation}
\setlength{\abovedisplayskip}{4pt}
\setlength{\belowdisplayskip}{4pt}
    \mathcal{S}[\mathcal{L}_{s_0}\mathcal{L}_{t_0}[f]](t, \omega) = s_{0}\eu^{- \iu \omega t_0} \mathcal{S}[f](s_{0}^{-1}(t - t_{0}), s_{0}\omega) %= s_{0}^{2}\eu^{- \iu \xi t_0} \langle f, w_{s_{0}^{-1}(u - t_{0}), s_{0}\xi}\rangle.
\end{equation}

 \textbf{Effect of input transformations on the spectrogram.} The spectrogram of a function $f \in \ltwo(\mathbb{R})$ is given by $\mathopen|\mathcal{S}[f](t, \omega)\mathopen|^{2}$. Input translations and dilations produce the following transformations:
\begin{align}
\mathopen|\mathcal{F}[\mathcal{L}_{t_0}[f]](t, \omega)\mathclose|^{2}&=\mathopen|\mathcal{L}_{t_{0}}[\mathcal{S}[f]](t, \omega)\mathclose|^{2} \label{eq:win_four_transequiv_2}\\
\mathopen|\mathcal{F}[\mathcal{L}_{s_0}[f]](t, \omega)\mathclose|^{2}&=\mathopen|s_{0}\mathclose|^{2}\mathopen|\mathcal{S}[f](s_{0}^{-1}t, s_{0}\omega)\mathclose|^{2} \label{eq:win_four_scaleequiv_2}
\end{align}

 \textbf{Equivariance and invariance properties of the short-time Fourier transform.} The short-time Fourier transform is \textit{approximately} translation and scale equivariance in a manner similar to that of the Fourier transform. In contrast to the Fourier transform, however, it decomposes input translations into a translation $t - t_{0}$ and a phase shift $\eu^{-i\omega t_{0}}$ in the output (Eq.~\ref{eq:win_four_transequiv}). This decomposition can be interpreted as a rough estimate $t - t_{0}$ signalizing the position in which the window $w$ is localized, and a fine grained localization within that window given by the phase shift $\eu^{-i\omega t_{0}}$ indicating the relative position of the pattern within the window $\mathcal{L}_{(t-t_{0})}[w](\tau)$. 

Equivariance to dilations is analogous to the Fourier transform up to the fact that time and frequency are now jointly described. However, since the window itself does not scale with the sampled frequency --as is the case in wavelet transforms--, exact equivariance is not obtained. Note that equivariance to dilations is only approximate, and is restricted to the set of scales that can be detected with the width of the window used (see Fig.~\ref{fig:equiv_in_STFT} for a visual explanation). Since this is not generally the case, the short-time Fourier transform is \textit{not scale equivariant}. 

The calculation of the spectrogram leaves the scale equivariance property of the transformation unaffected and is equivalent in a join manner to the scale equivariance property of the Fourier transform (Eq.~\ref{eq:win_four_scaleequiv}). Differently however, Eq.~~\ref{eq:win_four_transequiv_2} shows that translation equivariance is partially preserved and only information about the phase shift within the window is lost. This is why the short-time Fourier transform is said to carry positional information, i.e., to be (approximately) translation equivariant.

\vspace{-7mm}
\subsection{The Wavelet Transform}\label{appx:equiv_wavelet}
The wavelet transform of a signal $f \in \ltwo(\mathbb{R})$ is given by:
\begin{equation}
\setlength{\abovedisplayskip}{4pt}
\setlength{\belowdisplayskip}{4pt}
    \mathcal{W}[f](t, s) = \langle f, \psi_{t,s}\rangle = \int_{-\infty}^{+\infty}f(\tau) \  \frac{1}{\sqrt{s}} \psi^{*}\hspace{-1mm}\left( \frac{\tau-t}{s} \right) \, {\rm d}\tau,\nonumber
\end{equation}
and is equivalent to encoding $f$ into a time-frequency dictionary $\mathcal{D}=\{\psi_{t, s}\}_{u \in \sR, s \in \sR_{>0}}$, $\psi_{t, s}(\tau){=}\frac{1}{\sqrt{s}} \psi^{*}( \frac{\tau-t}{s})$.

\textbf{Input translation.} Let $\mathcal{L}_{t_{0}}[f](\tau){=}f(\tau-t_0)$ be a translated version of $f$. Its wavelet transform is given by:
\begin{align}
\setlength{\abovedisplayskip}{4pt}
\setlength{\belowdisplayskip}{4pt}
\mathcal{W}[\mathcal{L}_{t_{0}}[f]](t, s) &= \int_{- \infty}^{\infty}\hspace{-2.5mm}f(\tau -t_0) \sqrt{s}^{-1} \psi^{*}\hspace{-1mm}\left( s^{-1}(\tau -t) \right) \, \du \tau \ \Big| \ \tilde{t}{=}\tau -t_0\text{; } \du \tilde{t} {=} \du \tau  \nonumber \\
&=\int_{- \infty}^{\infty}\hspace{-2.5mm}f(\tilde{t})\sqrt{s}^{-1} \psi^{*}\hspace{-1mm}\left( s^{-1}(\tilde{t} + t_{0} -t) \right) \, \du \tilde{t}=\int_{- \infty}^{\infty}\hspace{-2.5mm}f(\tilde{t})\sqrt{s}^{-1} \psi^{*}\hspace{-1mm}\left( s^{-1}(\tilde{t} - (t - t_{0})) \right) \, \du \tilde{t} \nonumber\\
&= \mathcal{W}[f](t-t_{0}, s) = \mathcal{L}_{t_{0}}\mathcal{W}[f](t, s) \label{eq:wavelet_transequiv}
\end{align}
In other words, a translation of the input produces an equivalent translation in the wavelet domain.

 \textbf{Input scaling.} Let $\mathcal{L}_{s_{0}}[f](t) {=} f(s_{0}^{-1} t)$ be a scaled version of $f$. The corresponding wavelet transform is:
\begin{align}
\mathcal{W}[\mathcal{L}_{s_{0}}[f]](t, s) &= \int_{- \infty}^{\infty}\hspace{-2.5mm}f(s_{0}^{-1}t) \sqrt{s}^{-1} \psi^{*}\hspace{-1mm}\left( s^{-1}(\tau -t) \right) \, \du \tau \ \Big| \ \tilde{t}{=}s_0^{-1} \tau\text{; } \du \tilde{t} {=} s_0^{-1} \du \tau  \nonumber \\
&=\int_{- \infty}^{\infty}\hspace{-2.5mm}f(\tilde{t})\sqrt{s}^{-1} \psi^{*}\hspace{-1mm}\left( s^{-1}(s_{0}\tilde{t} -t) \right) \, \du (s_{0}\tilde{t}) \ \Big| \ t{=}s_0^{-1}s_{0} t \nonumber\\
&=\int_{- \infty}^{\infty}\hspace{-2.5mm}f(\tilde{t})\sqrt{s}^{-1}s_{0} \psi^{*}\hspace{-1mm}\left( s^{-1}s_{0}(\tilde{t} -s_{0}^{-1}t) \right) \, \du \tilde{t} \ \Big| \ s_0{=}\sqrt{s_{0}^{-1}s_{0}^{-1}}^{-1} \nonumber\\
&=\sqrt{s_{0}}\int_{- \infty}^{\infty}\hspace{-2mm}f(\tilde{t})\sqrt{s_{0}^{-1}s}^{-1} \psi^{*}\hspace{-1mm}\left( \big(s_{0}^{-1}s\big)^{-1}(\tilde{t} -s_{0}^{-1}t) \right) \, \du \tilde{t} \nonumber\\
&= \sqrt{s_{0}}\ \mathcal{W}[f](s_{0}^{-1}t, s_{0}^{-1}s) = \sqrt{s_{0}}\ \mathcal{L}_{s_{0}}\mathcal{W}[f](t,s)\label{eq:wavelet_scaleequiv}
\end{align}
In other words, a dilation $s_0$ in the input domain produces an \textit{equivalent} dilation in the wavelet domain on both components $(t,s)$, multiplied by a factor $\sqrt{s_{0}}$. That is, the wavelet transform is translation equivariant.

\textbf{Simultaneous input translation and scaling.} Following the same procedure, we can show the behavior of the wavelet transform to simultaneous translations and dilations of the input:
\begin{equation}
\setlength{\abovedisplayskip}{4pt}
\setlength{\belowdisplayskip}{4pt}
    \mathcal{W}[f(s_{0}^{-1}(\tau - t_{0})](t,s) = \sqrt{s_{0}}\ \mathcal{W}[f](s_{0}^{-1}(t-t_{0}),s_{0}^{-1}s) = \sqrt{s_{0}}\ \mathcal{L}_{t_{0}}\mathcal{L}_{s_{0}}\mathcal{W}[f](t,s)
\end{equation}
We observe that the Wavelet transform is the only time-frequency transform that respects equivariance with equivalent group representations in the input and output.

 \textbf{Effect of input transformations on the scalogram.} The scalogram of a function $f \in \ltwo(\mathbb{R})$ is given by $\mathopen|\mathcal{W}[f](u, s)\mathopen|^{2}$. Input translations and dilations produce the following transformations on the scalogram:
\begin{align}
    \mathopen| \mathcal{W}[\mathcal{L}_{t_{0}}[f]](u,s)\mathclose|^{2}  &= \mathopen|\mathcal{L}_{t_{0}}[\mathcal{W}[f]] (u,s)\mathclose|^{2} \label{eq:wavelet_transequiv_2}\\ 
    \mathopen| \mathcal{W}[\mathcal{L}_{s_{0}}[f]](u,s)\mathclose|^{2}  &= \mathopen| \mathcal{L}_{s_{0}}[\mathcal{W}[f]](u,s)\mathclose|^{2} \label{eq:wavelet_scaleequiv_2}
\end{align}
In other words the scalogram is exactly equivariant to both translations and dilations. 

 \textbf{Equivariance and invariance properties of the wavelet transform.} From Eq.~\ref{eq:wavelet_transequiv}, we can see that the wavelet transform is \textit{exactly equivariant to translations} and the group representation of the output space equals that of the input space. Furthermore, translation equivariance is preserved in the scalogram as well (Eq.~\ref{eq:wavelet_transequiv_2}). Similarly, scale equivariance is preserved on the wavelet transform up to a multiplicative factor (Eq.~\ref{eq:wavelet_scaleequiv}). However, the scalogram preserves both translation and dilation equivariance exactly (Eq.~\ref{eq:wavelet_scaleequiv_2}). 

We emphasize that the group representation on the output space resembles that of the input space. This behavior leads to much more straightforward group representations than that exhibited by the Fourier transform and the short-time Fourier transform. Additionally, exact scale equivariance is only obtained on the scalogram (Eq.~\ref{eq:wavelet_scaleequiv_2}), whilst for the wavelet transform it is retained up to multiplicative factor (Eq.~\ref{eq:wavelet_scaleequiv}). This behavior elucidates the fact that time-frequency transforms have been optimized for energy density representations rather than for the time-frequency representations themselves.

\vspace{-7mm}
\section{Experimental details}
\label{appx:exp_details}
Whenever possible, we use existing code for the baselines of our wavelet networks as a starting point for the general infrastructure of our model. Specifically, we utilize the PyTorch implementation provided in \url{https://github.com/philipperemy/very-deep-convnets-raw-waveforms} and \url{https://github.com/kyungyunlee/sampleCNN-pytorch} as baseline for the US8K experiments and the MTAT experiments \citet{lee2017sample}, respectively. By doing so, we aim to preserve the reproducibility of the experiments in the baseline papers during our own experiments, as some important training factors are not specified in the baseline papers, e.g., the learning rate used in \citet{dai2017very}. Unfortunately, \citet{abdoli2019end} do not provide code and we were forced to interpret some of the ambiguities in the paper, e.g., the pooling type utilized in the pooling layers and the loss metric used. 

Any omitted parameters can safely be considered to be the default values in \texttt{PyTorch 1.5.0}. Our experiments are carried out in a Nvidia TITAN RTX GPU. 

\vspace{-7mm}
\subsection{UrbanSound8K} \label{appxx:details_us8k}
\textbf{W$n$-Nets.} We use a sampling rate of $22.05$\si{\kilo\hertz} as opposed to the $8$\si{\kilo\hertz} used in \cite{dai2017very}. An early study that indicated that some classes were indistinguishable for the human ear at this sampling rate.\footnote{See %\url{https://github.com/dwromero/wavelet_networks/experiments/UrbanSound8K/data\_analysis.ipynb}.
\textit{[link removed for the sake of the double-blind review process].}
} We zero-pad signals shorter than 4 seconds so that all input signals have a constant length of $80200$ samples. Following the implementation of \citet{dai2017very}, we utilize the Adam optimizer \citep{kingma2014adam} with \texttt{lr=1e-2} and \texttt{weight\_decay=1e-4}, and perform training on the official first 9 folds and test on the $10^{th}$ fold. We noticed that reducing the learning rate from \texttt{1e-2} to \texttt{1e-3} increased the performance of our W-Nets. The reported results of the W-Net variants are obtained with this learning rate.

We utilize batches of size 16 and perform training for 400 epochs. The learning rate is reduced by half after 20 epochs of no improvement in validation loss. The W$n$-nets used are specified in Table~\ref{tab:mnets}. See \citet[Tab.~1]{dai2017very} for comparison.
\begin{table}
\centering
\vspace{-2mm}
    \begin{small}
    \begin{sc}
    \scalebox{0.82}{
    \begin{tabular}{ccccc}
    \toprule
        W3-Net & W5-Net & W11-Net & W18-Net & W34-Net \\
         (0.219m) & (0.558m) & (1.806m) & (3.759m) & (4.021m)\\
        \midrule
        \multicolumn{5}{c}{Input: 80200x 1 time-domain waveform} \\
        \midrule
        \multicolumn{5}{c}{\textit{Lifting Layer ( 9 scales)}} \vspace{1mm} \\
        $[79/4, 150]$ & $[79/4, 74]$ & $[79/4, 51]$ & $[79/4, 57]$ & $[79/4, 45]$ \\
        \midrule
        \multicolumn{5}{c}{Maxpool: 4x1 (output: 80200x 9x 1)} \\
        \midrule
        \multirow{2}[1]{*}{$[3, 150, 3]$} & \multirow{2}[1]{*}{$[3, 74, 3]$} & \multirow{2}[1]{*}{$[3, 51, 3] \times 2$} & \multirow{2}[1]{*}{$[3, 57, 3] \times 4$}& \multirow{2}[1]{*}{$\begin{bmatrix}
        3, 45 \\
         3, 45 \end{bmatrix} \times 3 $} \\
         \vspace{1mm}\\
         \midrule
         \multicolumn{5}{c}{Maxpool: 4x1 (output: 80200x 7x 1)} \\
         \midrule
        & \multirow{2}[1]{*}{$[3, 148, 3]$} & \multirow{2}[1]{*}{$[3, 102, 3] \times 2$} & \multirow{2}[1]{*}{$[3, 114, 3] \times 4$}& \multirow{2}[1]{*}{$\begin{bmatrix}
        3, 90 \\
         3, 90 \end{bmatrix} \times 4 $} \\
         \vspace{1mm}\\
         \cmidrule{2-5}
        & \multicolumn{4}{c}{Maxpool: 4x1 (output: 80200x 5x 1)} \\
        \cmidrule{2-5}
        & \multirow{2}[1]{*}{$[3, 296, 3]$} & \multirow{2}[1]{*}{$[3, 204, 3] \times 3$} & \multirow{2}[1]{*}{$[3, 228, 3] \times 4$}& \multirow{2}[1]{*}{$\begin{bmatrix}
        3, 180 \\
         3, 180 \end{bmatrix} \times 6 $} \\
         \vspace{1mm}\\
          \cmidrule{2-5}
        & \multicolumn{4}{c}{Maxpool: 4x1 (output: 80200x 3x 1)} \\
        \cmidrule{2-5}
         &  & \multirow{2}[1]{*}{$[3, 408, 3] \times 2$} & \multirow{2}[1]{*}{$[3, 456, 3] \times 4$}& \multirow{2}[1]{*}{$\begin{bmatrix}
        3, 360 \\
         3, 360 \end{bmatrix} \times 3 $} \\
         \vspace{1mm}\\
         \midrule
         \multicolumn{5}{c}{Global average pooling (output: 1 x n)} \\
         \midrule
         \multicolumn{5}{c}{Softmax $[1 10]$ (output: 1 x n)} \\
         \bottomrule
    \end{tabular}}
    \end{sc}
    \end{small}
    \caption{W$n$-networks. W3-Net (0.219M) denotes a 3-layer network with 0.219M parameters. $[79/4, 150, 3]$ denotes a group convolutional layer with a nominal kernel size of 79 samples, 150 filters and 3 scales, with a stride of $4$. Stride is omitted for stride 1 (e.g., $[3, 150, 3]$ has stride 1). Each convolutional layer uses batch normalization right after the convolution, after which ReLU is applied. Following the findings of Romero et al. \cite[Appx.~C]{romero2020attentive} on the influence of stride in the equivariance of the network, we replace strided convolutions by normal convolutions, followed by spatial pooling. $[ \dots ] \times k$ denotes $k$ stacked layers and double layers in brackets denote residual blocks as defined in \cite[Fig.~1b]{dai2017very}. In each of the levels of convolutional layers and residual blocks, the first convolution of the first block has scale 3 and the remaining convolutional layers at that level has scale 1.}
    \label{tab:mnets}
\end{table}

\textbf{W-1DCNN.} Following \citet{abdoli2019end}, we utilize a sampling rate of $16$\si{\kilo\hertz} during our experiments. We zero-pad signals shorter than 4 seconds so that all input signals have a constant length of $64000$ samples. Following the experimental description of the paper, we utilize the AdaDelta optimizer \citep{zeiler2012adadelta} with \texttt{lr=1.0} and perform training in a 10-fold cross validation setting as described in Sec.~\ref{sec:experiments}. We use batches of size 100 and perform training for 100 epochs. We utilize the $50999$-1DCNN variant of \citet{abdoli2019end}, as it is the variant that requires the less human engineering.\footnote{The remaining architectures partition the input signal into overlapping windows after which the predictions of each windows are summarized via a voting mechanism. Consequently, one could argue that the $50999$-1DCNN is the only variant that truly receives the raw waveform signal. Nevertheless it is not clear from the paper how the input signal of 64000 samples is reduced to 50999 samples, which is the input dimension of the raw signal for this architecture type.}

Unfortunately, we were not able to replicate the results reported in \cite{abdoli2019end} (83$\pm$1.3\%) in our experiments. Our replication of \cite{abdoli2019end} lead to a 10-cross fold accuracy of 62$\pm$1.3\%, which is 21 accuracy points less relative to the results reported. We experiment with our interpretation of the mean squared logarithmic error (MSLE) loss defined in \cite[Eq.~4]{abdoli2019end}. However, we find that the conventional cross-entropy loss leads to better results. Consequently, all our reported results are based on training with this loss.\footnote{The MSLE loss in \citet[Eq.~4]{abdoli2019end} is defined as $\tfrac{1}{N}\sum_{i=1}^{N} \log \tfrac{p_{i} + 1}{a_{i} + 1}^{2}$, where $p_{i}$, $a_{i}$ and $N$ are the predicted class, the actual class, and the number of samples respectively. Note, however, that obtaining the predicted class $p_{i}$, i.e., $p_{i} {=} \text{argmax}_{o} f(x_{i})$, where $f(x_{i}) \in \mathbb{R}^{O}$ is the output of the network for a classification problem with $O$ classes and input $x_{i}$, is a non-differentiable function. Consequently, it is not possible to train the network based on the formulation provided there. In order to train our model with this loss, we re-formulate the MSLE loss as $\tfrac{1}{N}\sum_{i=1}^{N}\sum_{o=1}^{O} \log \tfrac{p_{i,o} + 1}{a_{i,o} + 1}^{2}$, where $\{a_{i,o}\}_{o=1}^{O}$ is a one-hot encoded version of the label $a_{i}$. That is, we measure the difference between the one-hot encoded label and the output.} 

The description of the Wavelet $50999$-1DCNN \cite{abdoli2019end} is provided in Table~\ref{tab:1dcnn} (see \cite[Tab.~1]{abdoli2019end} for comparison). 
\begin{table}
\vspace{-3mm}
    \begin{small}
    \begin{sc}
    \scalebox{0.82}{
    \begin{tabular}{c}
    \toprule
        W-1DCNN (0.549m) \\
        \midrule
        Input: 64000x 1\\
        \midrule
        \textit{Lifting Layer ( 9 scales)} \vspace{1mm} \\
        $[63/2, 12]$ \\
        \midrule
        Maxpool: 8x1 \\
        \midrule
        $[31/2, 24, 3]$ \\
         \midrule
         Maxpool: 8x1\\
         \midrule
        $[15/2, 48, 3]$ \\
        $[7/2, 96, 3]$ \\
         $[3/2, 408, 3]$\\
         \midrule
          Maxpool: 5x1\\
          \midrule
          Flatten $196 \times 6 \rightarrow 1152 $ \\
          FC: $[1152, 96]$ \\
          FC: $[96, 48]$ \\
          FC: $[48, 10]$ \\
          \midrule
          Softmax \\
         \bottomrule
    \end{tabular}}
    \end{sc}
    \end{small}
    \caption{Wavelet network variant of the $50999$-1DCNN \citep{abdoli2019end}. $[31/2, 24, 3]$ denotes a group convolutional layer with a nominal kernel size of 31 samples, 24 filters and 3 scales, with a stride of $2$. FC: $[96,48]$ denotes a fully-connected layer with 96 input channels and 48 output channels. Each convolutional layer uses batch normalization right after the convolution followed by ReLU. All fully connected layers expect for the last one use dropout of 0.25 and ReLU. Following the findings of \citet[Appx.~C]{romero2020attentive} on the influence of stride in the equivariance of the network, we replace strided convolutions with normal convolutions followed by spatial pooling. We note that the input size of our network is (presumably) different from that in \cite{abdoli2019end}. Consequently, the last pooling layer utilizes a region of 5, in contrast to 4 as used in \cite{abdoli2019end}. However, as it is not clear how the input dimension is reduced from 64000 to 50999 in \cite{abdoli2019end} and we stick to their original sampling procedure. We interpret their poling layers as max-pooling ones.}
    \label{tab:1dcnn}
\end{table}

\vspace{-7mm}
\subsection{MagnaTagATune}
\textbf{W$3^{9}$-Network.} For the experiments in the MTAT dataset, we utilize the PyTorch code provided by Lee et al. \cite{lee2017sample}. We use the data and tag preprocessing used in \cite{lee2017sample}. We utilize the SGD optimizer with \texttt{lr=1e-2}, \texttt{weight\_decay=1e-6} and \texttt{nesterov=True}. We use batches of size 23 and perform training for 100 epochs. The learning rate is reduced by 5 after 3 epochs of no improvement in the validation loss. Early stopping is used if the learning rate drops under \texttt{1e-7}.

We were unable to replicate the per-class AUC results reported in \cite{lee2017sample}. Our experiments indicated a per-class AUC of 0.893 instead of the 0.905 reported in \cite{lee2017sample}. Details of the W$3^{9}$-Net used are given in Table~\ref{tab:39net} (see \cite[Tab.~1]{lee2017sample} for comparison).

\begin{table}
\vspace{-3mm}
    \begin{small}
    \begin{sc}
    \scalebox{0.82}{
    \begin{tabular}{c}
    \toprule
        W-$3^{9}$ Net (2.404m) \\
        \midrule
        Input: 59049x 1\\
        \midrule
        \textit{Lifting Layer ( 9 scales)} \vspace{1mm} \\
        $[3/3, 90]$ \\
        \midrule
        $[3/1, 90, 3]$, MP:3x1 \\
        $[3/1, 90, 1]$, MP:3x1 \\
        $[3/1, 180, 1]$, MP:3x1 \\
        $[3/1, 180, 3]$, MP:3x1 \\
        $[3/1, 180, 1]$, MP:3x1 \\
        $[3/1, 180, 1]$, MP:3x1 \\
        $[3/1, 180, 3]$, MP:3x1 \\
        $[3/1, 180, 1]$, MP:3x1 \\
        $[3/1, 360, 1]$, MP:3x1 \\
        $[3/1, 360, 3]$ \\
        \midrule
        FC: $[360, 50]$ \\
        \midrule
          Sigmoid \\
         \bottomrule
    \end{tabular}}
    \end{sc}
    \end{small}
    \caption{W$3^{9}$-network. $[3/1, 90, 3]$ denotes a group convolutional layer with a nominal kernel size of 3 samples, 90 filters and 3 scales, with a stride of 1. MP:3x1 denotes a max-pooling layer of size 3. FC: $[360,50]$ denotes a fully-connected layer with 360 input channels and 50 output channels. Each convolutional layer uses batch normalization after the convolution followed by ReLU. Dropout of 0.5 is used after the 6$^{\text{th}}$ and 11$^{\text{th}}$ layer. Following the findings of \citet[Appx.~C]{romero2020attentive} on the influence of stride in the network equivariances, we replace strided convolutions by normal convolutions followed by spatial pooling.}
    \label{tab:39net}
\end{table}

%% file: backmatter/appx_9_partialgcnn.tex
% path to figures directory
\graphicspath{{figures/9-partialgcnn/}}

%=========================================================================

% \begin{savequote}[75mm]
% Nulla facilisi. In vel sem. Morbi id urna in diam dignissim feugiat. Proin molestie tortor eu velit. Aliquam erat volutpat. Nullam ultrices, diam tempus vulputate egestas, eros pede varius leo.
% \qauthor{Quoteauthor Lastname}
% \end{savequote}

\chapter{Learning Equivariances and Partial Equivariances from Data}

% ------------------
\vspace{-7mm}
\section{Groups, subgroups, group actions and other group theoretical concepts}
\label{appx:9_background}
\vspace{-2mm}
\textbf{Groups.} Group theory is the mathematical language that describes symmetries. The core mathematical object is that of a \textit{group}, and defines what it means for something to exhibit symmetries. Specifically, a group is a tuple $(\gG, \cdot)$ consisting of a set of transformations $\gG$, and a binary operation $\cdot$, that exhibit the following properties: (i) closure, i.e., $g_1 \times g_2 = g_3 \in \gG, \forall g_1, g_2 \in \gG$ (ii) associativity, i.e., $g_1 \cdot (g_2 \cdot g_3) {=} (g_1 \cdot g_2) \cdot g_3$ for all $g_1, g_2, g_3 \in \gG$,  (iii) the existence of an identity element $e \in \gG$, such that $g \cdot e {=} e \cdot g = g$, and (iv) the existence of an inverse $g^{-1} \in \gG$~for~all~$g~\in~\gG$. 

\textbf{Subgroups.} Given a group $\left(\gG, \cdot\right)$, we say that a subset $\gH$ of the group $\gG$, is a \textit{subgroup} of $\gG$ if this subset also complies to the group axioms under the binary operation $\cdot$. For instance, the set of rotations by $90^\circ$, $\gH{=}\{0^\circ, 90^\circ, 180^\circ, 270^\circ\}$, is a subgroup of the rotation group $\mathrm{SO(2)}$, because it also complies to the closure, associativity, identity and inverse group axioms. 

\textbf{Group action.} One can define the \textit{action of the group} $\gG$ on a set $\gX$. This action describes how group elements $g \in \gG$ modify the set $\gX$ when the transformation is applied. For instance, the action of elements in the group of planar rotations $\theta \in \textrm{SO(2)}$ on an image $x \in \gX$ --written $\theta x$--, depicts how the image $x$ changes when the rotation $\theta$ is applied.

\textbf{Lie groups.} A group whose elements form a smooth manifold is referred to as a \textit{Lie group}. Since $\gG$ is not necessarily a vector space, we cannot add or subtract group elements --the only operation defined on the group is the binary operation $\cdot$ --. However, if the group is a Lie group, one can link the group $\gG$ to a vector space --tangent space at the identity $T_{e}(\gG)$--, called the \textit{Lie algebra}. Consequently, one can readily expand group elements on the Lie algebra using a basis $A{=}\sum_k a^{k}e_{k}$ and use these components for calculations. As neural networks work on vector spaces --by means of sums and products--, it is desirable to define convolutional kernels on the Lie algebra as $\psi{=}\mathtt{MLP}: \mathfrak{g}\rightarrow \sR^{\mathrm{N_{in} \times N_{out}}}$, where $\mathrm{N_{in}}$ and $\mathrm{N_{out}}$ depict the input and output channels of a convolutional kernel, respectively \citep{finzi2020generalizing}. 

\textbf{Relevant groups for computer vision applications.} In this work, we consider computer vision applications and thus, are mainly interested in groups that have direct effect on these applications. These groups compose the translation group $\mathrm{T(2)}$, the rotation group $\mathrm{SO(2)}$, the group of rotations and reflections $\mathrm{O(2)}$ and combinations thereof.\footnote{The names $\mathrm{SO(2)}$, $\mathrm{O(2)}$ are derived from their formal names: Special Orthogonal and Orthogonal group.} The actions of these groups can intuitively be understood as the translation, the rotation, and the rotation and reflection of 2D functions, respectively.

These groups can be combined by means of the \textit{semi-direct product} $(\rtimes)$ to construct groups that represent combined symmetries. For instance, the 2D roto-translation group $\mathrm{SE(2)}{=}\mathrm{T(2)}\rtimes\mathrm{SO(2)}$ encompasses symmetries described by both translations and rotations on 2D. Similarly, we can construct a group that describes 2D symmetries given by rotations, translations and reflections $\mathrm{E(2)}{=}\mathrm{T(2)}\rtimes\mathrm{O(2)}$.\footnote{The names $\mathrm{SE(2)}$, $\mathrm{E(2)}$ are derived from their formal names: Special Euclidean and Euclidean group.} Considering equivariance to these groups allows us to construct neural networks that respect the combined symmetries described by them. 
%There exist two mappings that go back and forth between a group and its Lie algebra. The exponential map $\mathrm{exp}: \mathfrak{g} \rightarrow \gG$ maps elements from the Lie algebra to the Lie group, converting infinitesimal transformations to group elements. Oftentimes, the image of the exponential map covers the entire group, and an inverse map: the logarithm map $\mathrm{log}: \gG \rightarrow \mathfrak{g}$, can be defined. Luckily, the $\mathrm{exp}$ and $\mathrm{log}$ maps can be computed in closed form for several important groups such as the translation group $\mathrm{T(n)}$, the 2D and 3D rotation groups $\mathrm{SO(2)}$, $\mathrm{SO(3)}$, and the translation and rotation group $\mathrm{SE(n)}$.
%\textbf{Convolutional kernels on the Lie algebra.} Neural networks are defined on vector spaces. Consequently, it can be cumbersome to define continuous convolutional kernels on the group elements directly. To solve this issue, \citet{finzi2020generalizing} defines continuous convolutional kernels directly on the Lie algebra, which maps elements on the Lie algebra to the value of the convolutional kernel in the group.
\vspace{-7mm}
% \section{An illustrative explanation of partial group equivariance}
% \vspace{-2mm}
\section{(Approximate) equivariance in partial group convolutions}\label{appx:9_equiv_error}
\vspace{-5mm}
\subsection{Partial group convolutions from the group $\gG$ to a subset $\gS$}\label{appx:9_from_g_to_s}
The partial group convolution from signals on $\gG$ to signals on a subset $\gS$ can be interpreted as a group convolution for which the output signal outside of $\gS$ is set to zero. Consequently, we can calculate the equivariance difference $\eerror$ in the feature representation, by calculating the difference on the subset $\gS$ of a group convolution with a group-transformed input $(\gL_{w}f * \psi)$ and a group convolution with a canonical input proceeded by the same transformation on $\gS$, i.e., $\gL_w(f * \psi)$.

The equivariance difference $\eerror^{\mathrm{out}}$ resulting from the effect of considering a subset $\gS$ in the output domain of the operation is given by:

\begin{align*}
    \eerror^{\mathrm{out}} &= \left\| \int_{\gS}\gL_w(\psi * f)(u)\, \du \mu_{\gG}(u) - \int_{\gS}(\psi * \gL_{w}f )(u)\, \du \mu_{\gG}(u) \right\|^{2}_{2}\\
    &= \left\| \int_{w^{-1}\gS}(\psi * f)(w^{-1}u)\, \du \mu_{\gG}(u) - \int_{\gS}(\psi * f )(w^{-1}u)\, \du \mu_{\gG}(u) \right\|^{2}_{2}\\
    &= \left\| \int_{\gS}(\psi * f)(u)\, \du \mu_{\gG}(u) - \int_{w\gS}(\psi * f )(u)\, \du \mu_{\gG}(u) \right\|^{2}_{2}\\
    &= \left\| \int^{s_{\max}}_{s_{\min}} (\psi * f)(u)\, \du \mu_{\gG}(u) - \int^{ws_{\max}}_{ws_{\min}}(\psi * f )(u)\, \du \mu_{\gG}(u) \right\|^{2}_{2}\\
    &= \bigg\| \left(\int^{s_{\max}}_{ws_{\max}} (\psi * f)(u)\, \du \mu_{\gG}(u) + \int^{ws_{\max}}_{s_{\min}} (\psi * f)(u)\, \du \mu_{\gG}(u) \right) - \\[-2\jot]
    &\hspace{3cm} \left(\int^{ws_{\max}}_{s_{\min}}(\psi * f )(u)\, \du \mu_{\gG}(u) + \int^{s_{\min}}_{ws_{\min}}(\psi * f )(u)\, \du \mu_{\gG}(u) \right) \bigg\|^{2}_{2}\\
    &= \left\| \int^{s_{\max}}_{ws_{\max}} (\psi * f)(u)\, \du \mu_{\gG}(u) - \int^{s_{\min}}_{ws_{\min}}(\psi * f )(u)\, \du \mu_{\gG}(u) \right\|^{2}_{2}
\end{align*}
From the first line to the second we take advantage of the equivariance property of the group convolution: $(\gL_{w}f * \psi)(u){=}\gL_{w}(f * \psi)(u)$, and account for the fact that only the region within $\gS$ is visible at the output. We use the change of variables $u=w^{-1}u$ from the second to third line, and specify the boundaries of $\gS$, $(s_{\max}, s_{\min} )$ from the third to the fourth line. In the fifth line we separate the integration over $\gS$ as a sum of two integrals which depict the same range. In the last line, we cancel out the overlapping parts of the two integrals to come to the final result.

In conclusion, the equivariance difference induced by a subset $\gS^{(2)}$ on the domain of the output $\eerror^{\mathrm{out}}$ is given by the difference between the part of the representation that leaves the subset $\gS$, and the part that comes to replace it instead. This behaviour is illustrated in Figure~\ref{fig:9_illustrated_partialequiv}.

\vspace{-7mm}
\subsection{Partial group convolutions from a subset $\gS^{(1)}$  to a subset $\gS^{(2)}$}\label{appx:9_from_s1_to_s2}
To isolate the effect of having a group subset as domain of the input signal $f$, we first consider the domain of the output to be the group, i.e., $\gS^{(2)}{=}\gG$. The equivariance difference in this case is given by the difference across the entire output representation of the group convolution calculated on an input subset $\gS^{(1)}$ with a canonical input $f$, and with a group transformed input $\gL_w f$.

The equivariance difference $\eerror^{\mathrm{in}}$ resulting from the effect of considering a subset $\gS^{(1)}$ in the input domain of the operation is given by:
\begin{align*}
    \eerror^{\mathrm{in}} &=  \left\|\int_{\gG}\int_{\gS}\psi(v^{-1}u)f(v)\, \du \mu_{\gG}(v) \du \mu_{\gG}(u) - \int_{\gG}\int_{\gS}\psi(v^{-1}u)f(w^{-1}v)\, \du \mu_{\gG}(v) \du \mu_{\gG}(u)\right\|^{2}_{2}\\
    &=  \left\|\int_{\gG}\left[\int_{\gS}\psi(v^{-1}u)f(v)\, \du \mu_{\gG}(v) - \int_{\gS}\psi(v^{-1}u)f(w^{-1}v)\, \du \mu_{\gG}(v) \right]\du \mu_{\gG}(u)\right\|^{2}_{2}\\
    &=  \left\|\int_{\gG}\int_{\gS}\psi(v^{-1}u) \left[f(v) - f(w^{-1}v)\right] \, \du \mu_{\gG}(v) \du \mu_{\gG}(u)\right\|^{2}_{2}\\
\end{align*}
In other words, the equivariance difference induced by a subset $\gS^{(1)}$ on the domain of the input $\eerror^{\mathrm{in}}$ is given by the difference in $\gS^{(1)}$ between the input $f$, and the part that comes to replace it when the input is modified by a group transformation $w$. This behavior is illustrated in Figure~\ref{fig:9_subset_input}.

\vspace{-7mm}
\section{Equivariance property of Monte-Carlo approximations}\label{appx:9_monte_carlo}
Consider the Monte-Carlo approximation shown in the main paper:
\begin{equation*}
    (\psi \hat{*} f)(u_i) = \sum\nolimits_{j} \psi(v_j^{-1}u_{i})f(v_j) \Bar{\mu}_{\gG}(v_j).
\end{equation*}
For a transformed version of the $\gL_w f$, we can show that the Monte-Carlo approximation of the group convolution is equivariant in expectation. The proof follows the same steps than \citet{finzi2020generalizing} except that the last step of the proof follows a different reason resulting from the fact that input and output elements can be sampled from different probability distributions. 

For a transformed version of the $\gL_w f$, we have that:
\begin{align*}
    (\psi \ \hat{*}\  \gL_w f)(u_i) &= \sum\nolimits_{j} \psi(v_j^{-1}u_{i})f(w^{-1}v_j) \Bar{\mu}_{\gG}(v_j)\\
   &= \sum\nolimits_{j} \psi(\tilde{v}_j^{-1}w^{-1}u_{i})f(\tilde{v}_j) \Bar{\mu}_{\gG}(\tilde{v}_j)\\
   &\stackrel{d}{=} (\psi\ \hat{*}\ f)(w ^{-1} u_i) = \gL_{w}(\psi\ \hat{*}\ f)(u_i)
\end{align*}
From the first to the second line, we use the change of variables $\Tilde{v}_j=wv_j$  and the fact that, group elements in the input domain are sampled from the Haar measure for which it holds that $\Bar{\mu}_{\gG}(v_j){=}\Bar{\mu}_{\gG}(\Tilde{v}_j)$. However, from the second to the third line, \textit{we must also assume that this holds for the output domain}. That is, that the probability of drawing $w^{-1}u_j$ is equal to that of drawing $u_j$. We emphasize that this is of particular importance in the partial equivariance setting as this might not be the case in general.
\vspace{-7mm}
\section{Experimental details}\label{appx:9_exp_details_sec}
\vspace{-5mm}
\subsection{Dataset description}

\textbf{Dataset availability and licensing.} We note that all the datasets used in this paper are publicly available. MNIST is available online under Creative Commons Attribution-Share Alike 3.0 license. CIFAR-10 and CIFAR-100 are available online under MIT license. PatchCamelyon is available online under MIT license.

\textbf{Rotated MNIST.} The rotated MNIST dataset \citep{larochelle2007empirical} contains 62,000 gray-scale 28x28 handwritten digits extracted from the MNIST dataset \citep{lecun1998gradient} uniformly rotated on the circle. The dataset is split into training, validation and test sets of 10,000, 2,000, and 50,000 images, respectively. 

\textbf{CIFAR-10 and CIFAR-100.} The CIFAR-10 dataset \citep{krizhevsky2009learning} consists of 60,000 real-world 32x32 RGB images uniformly drawn from 10 classes divided into training and test sets of 50,000 and 10,000 samples respectively. The CIFAR100 dataset \citep{krizhevsky2009learning} is similar to the CIFAR0 dataset, with the difference that images are uniformly drawn from 100 different classes. For validation purposes, we divide the training dataset of the CIFAR-10 and CIFAR-100 datasets into training and validation sets of 45,000 and 5,000 samples.

\textbf{PatchCamelyon.} The PatchCamelyon dataset \citep{veeling2018rotation} consists of 327,000 RGB image patches of tumorous and non-tumorous braset tissues extracted from the Camelyon16 dataset \citep{bejnordi2017diagnostic}, where each patch was labelled as tumorous if the central region of 32x32 pixels contained at least one tomorous pixel as givel by the original annotation in \citet{bejnordi2017diagnostic}. The dataset is divided into train, validation and test sets of 262,144, 32,768 and 32,768 images, respectively.
\vspace{-7mm}
\subsection{General remarks}

\textbf{Hardware.} Our code is written in {\btt PyTorch}. Our experiments were performed on NVIDIA TITAN RTX and V100 GPUs, depending on their availability and the size of the datasets.

\textbf{Network specifications.} For almost all the experiments in this paper --except those using the 13-layer CNN of \citet{laine2016temporal}--, we use the architecture shown in Fig.~\ref{fig:9_network_arch} with an initial lifting convolutional layer followed by $2$ ResBlocks with full, partial or regular convolutional layers for Regular G-CNNs, Partial G-CNNs and conventional ($\mathrm{T(2)}$) CNNs. All datasets use a network with $32$ feature maps in the hidden layers, Batch Normalization and ReLU.

For MNIST6-M and MNIST6-180, max-pooling is performed after each of the Residual Blocks. In the case of rotMNIST, max-pooling is performed after the lifting convolutional layer and the first group convolutional layer. For CIFAR-10 and CIFAR-100, we use max-pooling after each of the residual blocks. Finally, for PatchCamelyon, we apply max-pooling after the lifting convolution as well as both residual blocks. At the end of the network, a global max-pooling layer is used to create invariant features used for classification. These networks have approximately 460{\sc{k}} parameters.

\textbf{The continuous group convolutional kernels.} The convolutional kernels of Partial G-CNNs are parameterized as 3-layer SIRENs with 32 hidden units. For the experiments in the main text, we use $\omega_0{=}10.0$. We compare these to other conventional nonlinearities in Appx.~\ref{appx:9_extra_results} (Tab.~\ref{tab:9_kernel_comparison}). In the case of (partial) group equivariant 13-layer CNNs, the convolutional kernels are constructed as a 3-layer SIREN with 8 hidden units.
\vspace{-7mm}
\subsection{Hyperparameters and training details}\label{appx:9_exp_details}

To facilitate replicating our experiments, we provide the list of commands used for our experiments in \href{https://github.com/merlresearch/partial-gcnn/EXPERIMENTS.md}{\texttt{github.com/merlresearch/partial-gcnn/EXPERIMENTS.md}}

\textbf{Optimization and learning rate schedulers.} Networks on MNIST6-180, MNIST6-M, rotMNIST, CIFAR-10 and CIFAR-100 are trained for $300$ epochs and networks on PatchCamelyon are trained for $30$ epochs. Furthermore, we utilize a cosine annealing scheduler and combine it with a linear learning rate warm-up for $5$ epochs.

\textbf{Learning schedulers for the probability distributions $\mathrm{p}(u)$.} In order to improve the stability of learning the probability distributions on the groups, we utilize a learning rate scheduler similar to that of the main network, i.e., learning rate warm-up for 5 epochs followed by a cosine annealing scheduler, but with a lower base learning rate. Specifically, we use a base learning rate for all probability distributions $\mathrm{p}(u)$ of $1e{-}4$.

\textbf{Hyperparameters.} We note that all hyperparameters were chosen based on the best performance of the fully equivariant G-CNNs on the validation datasets. The found hyperparameters are subsequently used for the training of our Partial G-CNNs.

We use a batch size of 64 for all networks. In the case of CIFAR-10, CIFAR-100 and PatchCamelyon datasets, we also use a weight decay of $1e{-}4$.

\textbf{13-layer CNNs.} Additionally, in the case of 13-layer CNNs we use a dropout rate of 0.3 and train for 200 epochs with batches of size 128. These settings are used on rotMNIST, CIFAR10 and CIFAR100.
\vspace{-7mm}
\section{Additional Experiments}\label{appx:9_extra_results}

\textbf{Classification results on PatchCamelyon.} Table \ref{tab:9_pcam_results} shows the results obtained for G-CNNs and Partial G-CNNs on the PatchCamelyon dataset \citep{veeling2018rotation}. Partial G-CNNs match the performance of G-CNNs in this full equivariant setting. Similar to the rotMNIST case (Fig.~\ref{fig:9_learned_subsets}), the learned probability distributions over the group elements for PatchCamelyon are consistent with Regular G-CNNs.

\begin{table}
\caption{Image recognition accuracy on PatchCam dataset.}
\label{tab:9_pcam_results}
\vspace{-3mm}
\scalebox{0.8}{
    \centering
    \begin{tabular}{cccc}
    \toprule
    \sc{\shortstack{Base\\ Group}} & \sc{\shortstack{No.\\ Elements}} & \sc{\shortstack{Partial\\ Equiv.}}  & \sc{\shortstack{Classification Accuracy \\ on PatchCam (\%)}} \\
         
         \toprule
           $\mathrm{T(2)}$ & 1 & - & 67.59\\
         \midrule
         
         \multirow{4}{*}{$\mathrm{SE(2)}$} & \multirow{2}{*}{8} & \xmark & \textbf{89.87}\\
        & & \cmark & 89.07\\
        \cmidrule{2-4}
        
         & \multirow{2}{*}{16}  & \xmark & 89.71 \\
        & & \cmark & \underline{\textbf{90.31}}\\
        \midrule
        
        % \cmidrule{3-8}
         \multirow{2}{*}{$\mathrm{E(2)}$}& \multirow{2}{*}{16}  & \xmark & \textbf{89.77} \\
        & & \cmark & 88.13 \\
        \bottomrule
    \end{tabular}
    }
\end{table}

\textbf{Convolution kernels as implicit neural representations.} Next, we validate that SIRENs are better suited to parameterize group convolutional kernels than other alternatives. Tab.~\ref{tab:9_kernel_comparison} shows that $\mathrm{SE(2)}$-CNNs with SIREN kernels consistently outperform $\mathrm{SE(2)}$-CNNs with other parameterizations by a large margin on all the image benchmarks considered. \siren\ kernels consistently lead to better accuracy than other existing kernel parameterizations.% This result suggests that SIRENs are indeed better suited to represent continuous group convolutional kernels.

% \textbf{Evaluation of existing kernel parameterizations.} We observe that \siren\ kernels consistently lead to better accuracy than other existing kernel parameterizations. Our results are shown in Tab.~\ref{tab:9_kernel_comparison}.

\begin{table}
\caption{Comparison of kernel parameterizations.}
\label{tab:9_kernel_comparison}
\vspace{-3mm}
% \begin{small}
\scalebox{0.8}{
    \centering
    \begin{tabular}{cccccc}
    \toprule
         \multirow{2}{*}{\sc{Model}} & \multirow{2}{*}{\sc{\shortstack{No.\\ Elements}}} & \multirow{2}{*}{\sc{\shortstack{Kernel\\ Type}}}  & \multicolumn{3}{c}{\sc{Classification Accuracy (\%)}} \\
         \cmidrule{4-6}
         &   &  & \sc{RotMNIST} & \sc{CIFAR-10} & \sc{CIFAR-100} \\
         \toprule
         \multirow{12}{*}{$\mathrm{SE(2)}$-CNN}  & \multirow{4}{*}{4} & \relu & 96.49 & 59.95 & 28.01 \\ 
         &  & \leakyrelu & 94.47 & 56.19 & 27.36\\
         &  & \swish & 94.41 & 66.12 & 34.20  \\
         & & \siren & \textbf{99.10 }& \textbf{83.73} & \textbf{52.35} \\
          \cmidrule{2-6}
         &  \multirow{4}{*}{8} & \relu & 97.73 & 68.29 & 37.81  \\ 
         &  & \leakyrelu & 97.65 & 68.94 & 36.30 \\
         & & \swish & 97.72 & 69.20 & 34.10 \\
         &  & \siren & \textbf{99.17} & \textbf{86.08} & \underline{\textbf{55.55}}  \\
          \cmidrule{2-6}
      & \multirow{4}{*}{16} & \relu & 98.49 &  66.84 & 37.72 \\ 
          & & \leakyrelu & 98.53 & 68.01 & 38.29 \\
          & & \swish & 98.55 & 65.99 & 37.72\\
          & & \siren & \underline{\textbf{99.24}} & \underline{\textbf{86.68}} & \textbf{51.51} \\
        \bottomrule
    \end{tabular}
    }
\end{table}

\textbf{Enforcing monotonic decreasing group subsets over depth.}
Once a Partial G-CNN becomes partial equivariant at some depth, the network is, in general, unable to become fully equivariant at subsequent layers.\footnote{An exception to this rule is when the a layer goes back to the original input space, i.e., $\gS^{(2)}{=}\gX$, and the immediately subsequent layer goes back to the full group. This case is equivalent to performing a projection along a group axis, and going back to the full group afterwards, i.e., a lifting convolution.}  As a consequence, using fully equivariant layers after a partially equivariant layer does not restore full equivariance. 

Based on this observation, one could argue that it is beneficial to impose a monotonically decreasing size to the learned group subsets in order to prevent the at first sight meaningless situation in which the network goes back to larger group subsets. This can be encouraged with an additional \textit{monotonic equivariance loss} term in the training loss, which penalizes bigger subsets at subsequent layers:
\begin{equation}
    L_{\text{mon.~equiv}} = \sum_{l=1}^{L-1}(\gamma_l - \max(\gamma_{l+1}, \gamma_l)).
\end{equation}
Here, $\gamma_l$ represents the limit of the subset learned at the $l$-th layer.

Interestingly, we find that due to the reasons explained in Sec.~\ref{sec:9_discussion} imposing a monotonic decrease on the learned subsets leads to slightly worse performance than an unconstrained model (see Tabs.~\ref{tab:9_monotonic},~\ref{tab:9_vision_tasks}).

\begin{table}
\caption{Results with penalty term to encourage monotonicity in the subset sizes}
\label{tab:9_monotonic}
\vspace{-3mm}
\scalebox{0.8}{
\centering
\begin{tabular}{ccccc}
\toprule
\sc{Group} & \sc{No. Elements} & \sc{rotMNIST} & \sc{CIFAR10} & \sc{CIFAR100} \\
\toprule
$\mathrm{SE(2)}$ & 16 &  99.15 & 87.02 & 57.11 \\
$\mathrm{E(2)}$ & 16 & 98.41 & 89.00 & 58.85 \\
\bottomrule
\end{tabular}
}
\end{table}
% \vspace{-7mm}
% \section{Broader social impact}
% \vspace{-2mm}
% This work is fundamental and mathematical in nature. We believe it does not pose any immediate harm to society. However, the exact applications of these ideas could have negative impact and thus, care should be taken when using these ideas in machine learning. One motivation of this paper is to make deep networks more robust to nuisance factors and can hopefully be safer than earlier works.

%% file: backmatter/siks_dissertatiereeks.tex
% \begin{savequote}[75mm]
% Nulla facilisi. In vel sem. Morbi id urna in diam dignissim feugiat. Proin molestie tortor eu velit. Aliquam erat volutpat. Nullam    ultrices, diam tempus vulputate egestas, eros pede varius leo.
% \qauthor{Quoteauthor Lastname}
% \end{savequote}

\chapter{SIKS Dissertatiereeks}

\begin{xltabular}{\linewidth}{@{} l @{\hspace{0.5em}} l @{\hspace{1em}} X @{}}

% \toprule
2016
	&	 01	&	 Syed Saiden Abbas (RUN), Recognition of Shapes by Humans and Machines\\
	&	 02	&	 Michiel Christiaan Meulendijk (UU), Optimizing medication reviews through decision support: prescribing a better pill to swallow\\
	&	 03	&	 Maya Sappelli (RUN), Knowledge Work in Context: User Centered Knowledge Worker Support\\
	&	 04	&	 Laurens Rietveld (VUA), Publishing and Consuming Linked Data\\
	&	 05	&	 Evgeny Sherkhonov (UvA), Expanded Acyclic Queries: Containment and an Application in Explaining Missing Answers\\
	&	 06	&	 Michel Wilson (TUD), Robust scheduling in an uncertain environment\\
	&	 07	&	 Jeroen de Man (VUA), Measuring and modeling negative emotions for virtual training\\
	&	 08	&	 Matje van de Camp (TiU), A Link to the Past: Constructing Historical Social Networks from Unstructured Data\\
	&	 09	&	 Archana Nottamkandath (VUA), Trusting Crowdsourced Information on Cultural Artefacts\\
	&	 10	&	 George Karafotias (VUA), Parameter Control for Evolutionary Algorithms\\
	&	 11	&	 Anne Schuth (UvA), Search Engines that Learn from Their Users\\
	&	 12	&	 Max Knobbout (UU), Logics for Modelling and Verifying Normative Multi-Agent Systems\\
	&	 13	&	 Nana Baah Gyan (VUA), The Web, Speech Technologies and Rural Development in West Africa - An ICT4D Approach\\
	&	 14	&	 Ravi Khadka (UU), Revisiting Legacy Software System Modernization\\
	&	 15	&	 Steffen Michels (RUN), Hybrid Probabilistic Logics - Theoretical Aspects, Algorithms and Experiments\\
	&	 16	&	 Guangliang Li (UvA), Socially Intelligent Autonomous Agents that Learn from Human Reward\\
	&	 17	&	 Berend Weel (VUA), Towards Embodied Evolution of Robot Organisms\\
	&	 18	&	 Albert Mero\~{n}o Pe\~{n}uela (VUA), Refining Statistical Data on the Web\\
	&	 19	&	 Julia Efremova (TU/e), Mining Social Structures from Genealogical Data\\
	&	 20	&	 Daan Odijk (UvA), Context \& Semantics in News \& Web Search\\
	&	 21	&	 Alejandro Moreno C\'{e}lleri (UT), From Traditional to Interactive Playspaces: Automatic Analysis of Player Behavior in the Interactive Tag Playground\\
	&	 22	&	 Grace Lewis (VUA), Software Architecture Strategies for Cyber-Foraging Systems\\
	&	 23	&	 Fei Cai (UvA), Query Auto Completion in Information Retrieval\\
	&	 24	&	 Brend Wanders (UT), Repurposing and Probabilistic Integration of Data; An Iterative and data model independent approach\\
	&	 25	&	 Julia Kiseleva (TU/e), Using Contextual Information to Understand Searching and Browsing Behavior\\
	&	 26	&	 Dilhan Thilakarathne (VUA), In or Out of Control: Exploring Computational Models to Study the Role of Human Awareness and Control in Behavioural Choices, with Applications in Aviation and Energy Management Domains\\
	&	 27	&	 Wen Li (TUD), Understanding Geo-spatial Information on Social Media\\
	&	 28	&	 Mingxin Zhang (TUD), Large-scale Agent-based Social Simulation - A study on epidemic prediction and control\\
	&	 29	&	 Nicolas H\"{o}ning (TUD), Peak reduction in decentralised electricity systems - Markets and prices for flexible planning\\
	&	 30	&	 Ruud Mattheij (TiU), The Eyes Have It\\
	&	 31	&	 Mohammad Khelghati (UT), Deep web content monitoring\\
	&	 32	&	 Eelco Vriezekolk (UT), Assessing Telecommunication Service Availability Risks for Crisis Organisations\\
	&	 33	&	 Peter Bloem (UvA), Single Sample Statistics, exercises in learning from just one example\\
	&	 34	&	 Dennis Schunselaar (TU/e), Configurable Process Trees: Elicitation, Analysis, and Enactment\\
	&	 35	&	 Zhaochun Ren (UvA), Monitoring Social Media: Summarization, Classification and Recommendation\\
	&	 36	&	 Daphne Karreman (UT), Beyond R2D2: The design of nonverbal interaction behavior optimized for robot-specific morphologies\\
	&	 37	&	 Giovanni Sileno (UvA), Aligning Law and Action - a conceptual and computational inquiry\\
	&	 38	&	 Andrea Minuto (UT), Materials that Matter - Smart Materials meet Art \& Interaction Design\\
	&	 39	&	 Merijn Bruijnes (UT), Believable Suspect Agents; Response and Interpersonal Style Selection for an Artificial Suspect\\
	&	 40	&	 Christian Detweiler (TUD), Accounting for Values in Design\\
	&	 41	&	 Thomas King (TUD), Governing Governance: A Formal Framework for Analysing Institutional Design and Enactment Governance\\
	&	 42	&	 Spyros Martzoukos (UvA), Combinatorial and Compositional Aspects of Bilingual Aligned Corpora\\
	&	 43	&	 Saskia Koldijk (RUN), Context-Aware Support for Stress Self-Management: From Theory to Practice\\
	&	 44	&	 Thibault Sellam (UvA), Automatic Assistants for Database Exploration\\
	&	 45	&	 Bram van de Laar (UT), Experiencing Brain-Computer Interface Control\\
	&	 46	&	 Jorge Gallego Perez (UT), Robots to Make you Happy\\
	&	 47	&	 Christina Weber (UL), Real-time foresight - Preparedness for dynamic innovation networks\\
	&	 48	&	 Tanja Buttler (TUD), Collecting Lessons Learned\\
	&	 49	&	 Gleb Polevoy (TUD), Participation and Interaction in Projects. A Game-Theoretic Analysis\\
	&	 50	&	 Yan Wang (TiU), The Bridge of Dreams: Towards a Method for Operational Performance Alignment in IT-enabled Service Supply Chains\\
	
\midrule
2017
	&	 01	&	 Jan-Jaap Oerlemans (UL), Investigating Cybercrime\\
	&	 02	&	 Sjoerd Timmer (UU), Designing and Understanding Forensic Bayesian Networks using Argumentation\\
	&	 03	&	 Dani\"{e}l Harold Telgen (UU), Grid Manufacturing; A Cyber-Physical Approach with Autonomous Products and Reconfigurable Manufacturing Machines\\
	&	 04	&	 Mrunal Gawade (CWI), Multi-core Parallelism in a Column-store\\
	&	 05	&	 Mahdieh Shadi (UvA), Collaboration Behavior\\
	&	 06	&	 Damir Vandic (EUR), Intelligent Information Systems for Web Product Search\\
	&	 07	&	 Roel Bertens (UU), Insight in Information: from Abstract to Anomaly\\
	&	 08	& 	 Rob Konijn (VUA), Detecting Interesting Differences:Data Mining in Health Insurance Data using Outlier Detection and Subgroup Discovery\\
	&	 09	&	 Dong Nguyen (UT), Text as Social and Cultural Data: A Computational Perspective on Variation in Text\\
	&	 10	&	 Robby van Delden (UT), (Steering) Interactive Play Behavior\\
	&	 11	&	 Florian Kunneman (RUN), Modelling patterns of time and emotion in Twitter \#anticipointment\\
	&	 12	&	 Sander Leemans (TU/e), Robust Process Mining with Guarantees\\
 	&	 13	& 	 Gijs Huisman (UT), Social Touch Technology - Extending the reach of social touch through haptic technology\\
 	&	 14	&	 Shoshannah Tekofsky (TiU), You Are Who You Play You Are: Modelling Player Traits from Video Game Behavior\\
	&	 15	&	 Peter Berck (RUN),  Memory-Based Text Correction\\
	&	 16	&	 Aleksandr Chuklin (UvA), Understanding and Modeling Users of Modern Search Engines\\
	&	 17	&	 Daniel Dimov (UL), Crowdsourced Online Dispute Resolution\\
	&	 18	&	 Ridho Reinanda (UvA), Entity Associations for Search\\
	&	 19	& 	 Jeroen Vuurens (UT), Proximity of Terms, Texts and Semantic Vectors in Information Retrieval\\
	&	 20	&	 Mohammadbashir Sedighi (TUD), Fostering Engagement in Knowledge Sharing: The Role of Perceived Benefits, Costs and Visibility\\
	&	 21	&	 Jeroen Linssen (UT), Meta Matters in Interactive Storytelling and Serious Gaming (A Play on Worlds)\\
	&	 22	&	 Sara Magliacane (VUA), Logics for causal inference under uncertainty\\
	&	 23	&	 David Graus (UvA), Entities of Interest --- Discovery in Digital Traces\\
	&	 24	&	 Chang Wang (TUD), Use of Affordances for Efficient Robot Learning\\
	&	 25	&	 Veruska Zamborlini (VUA), Knowledge Representation for Clinical Guidelines, with applications to Multimorbidity Analysis and Literature Search\\
	&	 26	&	 Merel Jung (UT), Socially intelligent robots that understand and respond to human touch\\
	&	 27	&	 Michiel Joosse (UT), Investigating Positioning and Gaze Behaviors of Social Robots: People's Preferences, Perceptions and Behaviors\\
	&	 28	&	 John Klein (VUA), Architecture Practices for Complex Contexts\\
	&	 29	&	 Adel Alhuraibi (TiU), From IT-BusinessStrategic Alignment to Performance: A Moderated Mediation Model of Social Innovation, and Enterprise Governance of    IT"\\
	&	 30	&	 Wilma Latuny (TiU), The Power of Facial Expressions\\
	&	 31	&	 Ben Ruijl (UL), Advances in computational methods for QFT calculations\\
	&	 32	& 	 Thaer Samar (RUN), Access to and Retrievability of Content in Web Archives\\
	&	 33	&	 Brigit van Loggem (OU), Towards a Design Rationale for Software Documentation: A Model of Computer-Mediated Activity\\
	&	 34	&	 Maren Scheffel (OU), The Evaluation Framework for Learning Analytics \\
	&	 35	&	 Martine de Vos (VUA), Interpreting natural science spreadsheets \\
	&	 36	&	 Yuanhao Guo (UL), Shape Analysis for Phenotype Characterisation from High-throughput Imaging \\
	&	 37	&	 Alejandro Montes Garcia (TU/e), WiBAF: A Within Browser Adaptation Framework that Enables Control over Privacy \\
	&	 38	&	 Alex Kayal (TUD), Normative Social Applications \\
	&	 39	&	 Sara Ahmadi (RUN), Exploiting properties of the human auditory system and compressive sensing methods to increase   noise robustness in ASR \\
	&	 40	&	 Altaf Hussain Abro (VUA), Steer your Mind: Computational Exploration of Human Control in Relation to Emotions, Desires and Social Support For applications in human-aware support systems \\
	&	 41	&	 Adnan Manzoor (VUA), Minding a Healthy Lifestyle: An Exploration of Mental Processes and a Smart Environment to Provide Support for a Healthy Lifestyle\\
	&	 42	&	 Elena Sokolova (RUN), Causal discovery from mixed and missing data with applications on ADHD  datasets\\
	&	 43	&	 Maaike de Boer (RUN), Semantic Mapping in Video Retrieval\\
	&	 44	&	 Garm Lucassen (UU), Understanding User Stories - Computational Linguistics in Agile Requirements Engineering\\
	&	 45	&	 Bas Testerink	(UU), Decentralized Runtime Norm Enforcement\\
	&	 46	&	 Jan Schneider	(OU), Sensor-based Learning Support\\
	&	 47	&	 Jie Yang (TUD), Crowd Knowledge Creation Acceleration\\
	&	 48	&	 Angel Suarez (OU), Collaborative inquiry-based learning\\

\midrule
2018
	&	 01	&	 Han van der Aa (VUA), Comparing and Aligning Process Representations \\
	&	 02	&	 Felix Mannhardt (TU/e), Multi-perspective Process Mining \\
	&	 03	&	 Steven Bosems (UT), Causal Models For Well-Being: Knowledge Modeling, Model-Driven Development of Context-Aware Applications, and Behavior Prediction\\
	&	 04	&	 Jordan Janeiro (TUD), Flexible Coordination Support for Diagnosis Teams in Data-Centric Engineering Tasks \\
	&	 05	&	 Hugo Huurdeman (UvA), Supporting the Complex Dynamics of the Information Seeking Process \\
	&	 06	&	 Dan Ionita (UT), Model-Driven Information Security Risk Assessment of Socio-Technical Systems \\
	&	 07	&	 Jieting Luo (UU), A formal account of opportunism in multi-agent systems \\
	&	 08	&	 Rick Smetsers (RUN), Advances in Model Learning for Software Systems \\
	&	 09	&	 Xu Xie	(TUD), Data Assimilation in Discrete Event Simulations \\
	&	 10	&	 Julienka Mollee (VUA), Moving forward: supporting physical activity behavior change through intelligent technology \\
	&	 11	&	 Mahdi Sargolzaei (UvA), Enabling Framework for Service-oriented Collaborative Networks \\
	&	 12	&	 Xixi Lu (TU/e), Using behavioral context in process mining \\
	&	 13	&	 Seyed Amin Tabatabaei (VUA), Computing a Sustainable Future \\
	&	 14	&	 Bart Joosten (TiU), Detecting Social Signals with Spatiotemporal Gabor Filters \\
	&	 15	&	 Naser Davarzani (UM), Biomarker discovery in heart failure \\
	&	 16	&	 Jaebok Kim (UT), Automatic recognition of engagement and emotion in a group of children \\
	&	 17	&	 Jianpeng Zhang (TU/e), On Graph Sample Clustering \\
	&	 18	& 	 Henriette Nakad (UL), De Notaris en Private Rechtspraak \\
	&	 19	&	 Minh Duc Pham (VUA), Emergent relational schemas for RDF \\
	&	 20	&	 Manxia Liu (RUN), Time and Bayesian Networks \\
	&	 21	&	 Aad Slootmaker (OU), EMERGO: a generic platform for authoring and playing scenario-based serious games \\
	&	 22	&	 Eric Fernandes de Mello Ara\'{u}jo (VUA), Contagious: Modeling the Spread of Behaviours, Perceptions and Emotions in Social Networks \\
	&	 23	&	 Kim Schouten (EUR), Semantics-driven Aspect-Based Sentiment Analysis \\
	&	 24	&	 Jered Vroon (UT), Responsive Social Positioning Behaviour for Semi-Autonomous Telepresence Robots \\
	&	 25	&	 Riste Gligorov (VUA), Serious Games in Audio-Visual Collections \\
	&	 26	& 	 Roelof Anne Jelle de Vries (UT),Theory-Based and Tailor-Made: Motivational Messages for Behavior Change Technology \\
	&	 27	&	 Maikel Leemans (TU/e), Hierarchical Process Mining for Scalable Software Analysis \\
	&	 28	&	 Christian Willemse (UT), Social Touch Technologies: How they feel and how they make you feel \\
	&	 29	&	 Yu Gu (TiU), Emotion Recognition from Mandarin Speech \\
	&	 30	&	 Wouter Beek (VUA),  The "K" in "semantic web" stands for "knowledge": scaling semantics to the web \\
	
\midrule
2019
	&	 01	&	 Rob van Eijk (UL),Web privacy measurement in real-time bidding systems. A graph-based approach to RTB system classification \\
	&	 02	&	 Emmanuelle Beauxis Aussalet (CWI, UU), Statistics and Visualizations for Assessing Class Size Uncertainty \\
	&	 03	&	 Eduardo Gonzalez Lopez de Murillas (TU/e), Process Mining on Databases: Extracting Event Data from Real Life Data Sources \\
	&	 04	&	 Ridho Rahmadi (RUN), Finding stable causal structures from clinical data \\
	& 	 05	&	 Sebastiaan van Zelst (TU/e), Process Mining with Streaming Data \\
	&	 06	& 	 Chris Dijkshoorn (VUA), Nichesourcing for Improving Access to Linked Cultural Heritage Datasets \\
	&	 07	&	 Soude Fazeli (TUD), Recommender Systems in Social Learning Platforms \\
	& 	 08	&	 Frits de Nijs (TUD), Resource-constrained Multi-agent Markov Decision Processes \\
	&	 09	&	 Fahimeh Alizadeh Moghaddam (UvA), Self-adaptation for energy efficiency in software systems \\
	&	 10	&	 Qing Chuan Ye (EUR), Multi-objective Optimization Methods for Allocation and Prediction \\
	&	 11	&	 Yue Zhao (TUD), Learning Analytics Technology to Understand Learner Behavioral Engagement in MOOCs \\
	&	 12	&	 Jacqueline Heinerman (VUA), Better Together \\
	&	 13	&	 Guanliang Chen (TUD), MOOC Analytics: Learner Modeling and Content Generation \\
	&	 14	&	 Daniel Davis (TUD), Large-Scale Learning Analytics: Modeling Learner Behavior \& Improving Learning Outcomes in Massive Open Online Courses \\
	&	 15	&	 Erwin Walraven (TUD), Planning under Uncertainty in Constrained and Partially Observable Environments \\
	&	 16	&	 Guangming Li (TU/e), Process Mining based on Object-Centric Behavioral Constraint (OCBC) Models \\
	&	 17	&	 Ali Hurriyetoglu (RUN),Extracting actionable information from microtexts \\
	&	 18	&	 Gerard Wagenaar (UU), Artefacts in Agile Team Communication \\
	&	 19	&	 Vincent Koeman (TUD), Tools for Developing Cognitive Agents \\
	&	 20	&	 Chide Groenouwe (UU), Fostering technically augmented human collective intelligence \\
	&	 21	&	 Cong Liu (TU/e), Software Data Analytics: Architectural Model Discovery and Design Pattern Detection \\
	&	 22	&	 Martin van den Berg (VUA),Improving IT Decisions with Enterprise Architecture \\
	&	 23	&	 Qin Liu (TUD), Intelligent Control Systems: Learning, Interpreting, Verification\\
	&	 24	&	 Anca Dumitrache (VUA),  Truth in Disagreement - Crowdsourcing Labeled Data for Natural Language Processing\\
	&	 25	&	 Emiel van Miltenburg (VUA), Pragmatic factors in (automatic) image description \\
	&	 26	&	 Prince Singh (UT), An Integration Platform for Synchromodal Transport \\
	&	 27	&	 Alessandra Antonaci (OU), The Gamification Design Process applied to (Massive) Open Online Courses\\
	&	 28	&	 Esther Kuindersma (UL), Cleared for take-off: Game-based learning to prepare airline pilots for critical situations \\
	&	 29	&	 Daniel Formolo (VUA), Using virtual agents for simulation and training of social skills in safety-critical circumstances \\
	&	 30	&	 Vahid Yazdanpanah (UT), Multiagent Industrial Symbiosis Systems \\
	&	 31	&	 Milan Jelisavcic (VUA), Alive and Kicking: Baby Steps in Robotics \\
	&	 32	&	 Chiara Sironi (UM), Monte-Carlo Tree Search for Artificial General Intelligence in Games \\
	&	 33	&	 Anil Yaman (TU/e), Evolution of Biologically Inspired Learning in Artificial Neural Networks \\
	&	 34	&	 Negar Ahmadi (TU/e), EEG Microstate and Functional Brain Network Features for Classification of Epilepsy and PNES \\
	&	 35	&	 Lisa Facey-Shaw (OU), Gamification with digital badges in learning programming \\
	&	 36	&	 Kevin Ackermans (OU), Designing Video-Enhanced Rubrics to Master Complex Skills \\
	&	 37	&	 Jian Fang (TUD), Database Acceleration on FPGAs \\
	&	 38	&	 Akos Kadar (OU), Learning visually grounded and multilingual representations \\

\midrule
2020
	&	 01	&	 Armon Toubman (UL), Calculated Moves: Generating Air Combat Behaviour \\
	&	 02	&	 Marcos de Paula Bueno (UL), Unraveling Temporal Processes using Probabilistic Graphical Models \\
	&	 03	&	 Mostafa Deghani (UvA), Learning with Imperfect Supervision for Language Understanding \\
	&	 04	&	 Maarten van Gompel (RUN), Context as Linguistic Bridges \\
	&	 05	&	 Yulong Pei (TU/e), On local and global structure mining \\
	&	 06	&	 Preethu Rose Anish (UT), Stimulation Architectural Thinking during Requirements Elicitation - An Approach and Tool Support \\
	&	 07	&	 Wim van der Vegt (OU), Towards a software architecture for reusable game components \\
	&	 08	&	 Ali Mirsoleimani (UL),Structured Parallel Programming for Monte Carlo Tree Search \\
	&	 09	&	 Myriam Traub (UU), Measuring Tool Bias and Improving Data Quality for Digital Humanities Research \\
	&	 10	&	 Alifah Syamsiyah (TU/e), In-database Preprocessing for Process Mining \\
	&	 11	&	 Sepideh Mesbah (TUD), Semantic-Enhanced Training Data AugmentationMethods for Long-Tail Entity Recognition Models \\
	&	 12	&	 Ward van Breda (VUA), Predictive Modeling in E-Mental Health: Exploring Applicability in Personalised Depression Treatment \\
	&	 13	&	 Marco Virgolin (CWI), Design and Application of Gene-pool Optimal Mixing Evolutionary Algorithms for Genetic Programming \\
	&	 14	&	 Mark Raasveldt (CWI/UL), Integrating Analytics with Relational Databases \\
	&	 15	&	 Konstantinos Georgiadis (OU),  Smart CAT: Machine Learning for Configurable Assessments in Serious Games \\
	&	 16	&	 Ilona Wilmont (RUN), Cognitive Aspects of Conceptual Modelling \\
	&	 17	&	 Daniele Di Mitri (OU), The Multimodal Tutor: Adaptive Feedback from Multimodal Experiences \\
  	&	 18	&	 Georgios Methenitis (TUD), Agent Interactions \& Mechanisms in Markets with Uncertainties: Electricity Markets in Renewable Energy Systems \\
	&	 19	&	 Guido van Capelleveen (UT), Industrial Symbiosis Recommender Systems \\
	&	 20	&	 Albert Hankel (VUA), Embedding Green ICT Maturity in Organisations \\
	&	 21	&	 Karine da Silva Miras de Araujo (VUA), Where is the robot?: Life as it could be \\
	&	 22	&	 Maryam Masoud Khamis (RUN), Understanding complex systems implementation through a modeling approach: the case of e-government in Zanzibar \\
	&	 23	&	 Rianne Conijn (UT), The Keys to Writing: A writing analytics approach to studying writing processes using keystroke logging \\
	&	 24	&	 Lenin da N\'{o}brega Medeiros (VUA/RUN), How are you feeling, human? Towards emotionally supportive chatbots \\
	&	 25	&	 Xin Du (TU/e), The Uncertainty in Exceptional Model Mining \\
	&	 26	&	 Krzysztof Leszek Sadowski (UU), GAMBIT: Genetic Algorithm for Model-Based mixed-Integer opTimization \\
	&	 27	&	 Ekaterina Muravyeva (TUD), Personal data and informed consent in an educational context \\
	&	 28	&	 Bibeg Limbu (TUD), Multimodal interaction for deliberate practice: Training complex skills with augmented reality \\
	&	 29	&	 Ioan Gabriel Bucur (RUN), Being Bayesian about Causal Inference \\
	&	 30	&	 Bob Zadok Blok (UL), Creatief, Creatiever, Creatiefst \\
	&	 31	&	 Gongjin Lan (VUA), Learning better -- From Baby to Better \\
	&	 32	& 	 Jason Rhuggenaath (TU/e), Revenue management in online markets: pricing and online advertising \\
	&	 33	& 	 Rick Gilsing (TU/e), Supporting service-dominant business model evaluation in the context of business model innovation \\
	&	 34	&	 Anna Bon (UM), Intervention or Collaboration? Redesigning Information and Communication Technologies for Development \\
	&	 35	&	 Siamak Farshidi (UU), Multi-Criteria Decision-Making in Software Production \\

\midrule
2021
	&	 01	&	 Francisco Xavier Dos Santos Fonseca (TUD),Location-based Games for Social Interaction in Public Space \\
	&	 02	&	 Rijk Mercuur (TUD), Simulating Human Routines: Integrating Social Practice Theory in Agent-Based Models \\
	&	 03	&	 Seyyed Hadi Hashemi (UvA), Modeling Users Interacting with Smart Devices \\
	&	 04	&	 Ioana Jivet (OU), The Dashboard That Loved Me: Designing adaptive learning analytics for self-regulated learning \\
	&	 05	&	 Davide Dell'Anna (UU), Data-Driven Supervision of Autonomous Systems \\
	&	 06	&	 Daniel Davison (UT), "Hey robot, what do you think?" How children learn with a social robot \\
	&	 07	&	 Armel Lefebvre (UU), Research data management for open science \\
	&	 08	&	 Nardie Fanchamps (OU), The Influence of Sense-Reason-Act Programming on Computational Thinking \\
	&	 09	&	 Cristina Zaga (UT), The Design of Robothings. Non-Anthropomorphic and Non-Verbal Robots to Promote Children's Collaboration Through Play \\
	&	 10	&	 Quinten Meertens (UvA), Misclassification Bias in Statistical Learning \\
	&	 11	&	 Anne van Rossum (UL), Nonparametric Bayesian Methods in Robotic Vision \\
	&	 12	&	 Lei Pi (UL), External Knowledge Absorption in Chinese SMEs \\
	&	 13	&	 Bob R. Schadenberg (UT), Robots for Autistic Children: Understanding and Facilitating Predictability for Engagement in Learning \\
	&	 14	&	 Negin Samaeemofrad (UL), Business Incubators: The Impact of Their Support \\
	&	 15	& 	 Onat Ege Adali (TU/e), Transformation of Value Propositions into Resource Re-Configurations through the Business Services Paradigm  \\
	&	 16	&	 Esam A. H. Ghaleb (UM), Bimodal emotion recognition from audio-visual cues \\
	&	 17	&	 Dario Dotti (UM), Human Behavior Understanding  from motion and bodily cues using deep neural networks \\
	&	 18	&	 Remi Wieten (UU), Bridging the Gap Between Informal Sense-Making Tools and Formal Systems - Facilitating the Construction of Bayesian Networks and Argumentation Frameworks \\
	&	 19	&	 Roberto Verdecchia (VUA), Architectural Technical Debt: Identification and Management \\
	&	 20	&	 Masoud Mansoury (TU/e), Understanding and Mitigating Multi-Sided Exposure Bias in Recommender Systems \\
	&	 21	&	 Pedro Thiago Timb\'{o} Holanda (CWI), Progressive Indexes \\
	&	 22	&	 Sihang Qiu (TUD), Conversational Crowdsourcing \\
	&	 23	&	 Hugo Manuel Proen\c{c}a (UL), Robust rules for prediction and description \\
	&	 24	&	 Kaijie Zhu (TU/e), On Efficient Temporal Subgraph Query Processing \\
	&	 25	&	 Eoin Martino Grua (VUA), The Future of E-Health is Mobile: Combining AI and Self-Adaptation to Create Adaptive E-Health Mobile Applications \\
	&	 26	& 	 Benno Kruit (CWI/VUA), Reading the Grid: Extending Knowledge Bases from Human-readable Tables \\
	&	 27	&	 Jelte van Waterschoot (UT), Personalized and Personal Conversations: Designing Agents Who Want to Connect With You \\
	&	 28	&	 Christoph Selig (UL), Understanding the Heterogeneity of Corporate Entrepreneurship Programs \\

\midrule
2022
	&	 01	&	Judith van Stegeren (UT), Flavor text generation for role-playing video games \\
	&	 02	&	Paulo da Costa (TU/e), Data-driven Prognostics and Logistics Optimisation: A Deep Learning Journey \\
	&	 03	&	Ali el Hassouni (VUA), A Model A Day Keeps The Doctor Away: Reinforcement Learning For Personalized Healthcare \\
	&	 04	&	\"{U}nal Aksu (UU), A Cross-Organizational Process Mining Framework \\
	&	 05	&	Shiwei Liu (TU/e), Sparse Neural Network Training with In-Time Over-Parameterization \\
	&	 06	& 	Reza Refaei Afshar (TU/e), Machine Learning for Ad Publishers in Real Time Bidding \\
	&	 07	&	Sambit Praharaj (OU), Measuring the Unmeasurable? Towards Automatic Co-located Collaboration Analytics \\
	&	 08	&	Maikel L. van Eck (TU/e), Process Mining for Smart Product Design \\
	&	 09	&	Oana Andreea Inel (VUA), Understanding Events: A Diversity-driven Human-Machine Approach \\
	&	 10	&	Felipe Moraes Gomes (TUD), Examining the Effectiveness of Collaborative Search Engines \\
	&	 11	&	Mirjam de Haas (UT), Staying engaged in child-robot interaction, a quantitative approach to studying preschoolers' engagement with robots and tasks during second-language tutoring \\
	&	 12	&	Guanyi Chen (UU),  Computational Generation of Chinese Noun Phrases \\
	&	 13	&	Xander Wilcke (VUA), Machine Learning on Multimodal Knowledge Graphs: Opportunities, Challenges, and Methods for Learning on Real-World Heterogeneous and Spatially-Oriented Knowledge \\
	&	 14	&	Michiel Overeem (UU), Evolution of Low-Code Platforms \\
	&	 15	&	Jelmer Jan Koorn (UU), Work in Process: Unearthing Meaning using Process Mining \\
	&	 16	&	Pieter Gijsbers (TU/e), Systems for AutoML Research \\
	&	 17	&	Laura van der Lubbe (VUA), Empowering vulnerable people with serious games and gamification \\
	&	 18	&	Paris Mavromoustakos Blom (TiU), Player Affect Modelling and Video Game Personalisation \\
	&	 19	&	Bilge Yigit Ozkan (UU), Cybersecurity Maturity Assessment and Standardisation \\
	&	 20	&	Fakhra Jabeen (VUA), Dark Side of the Digital Media - Computational Analysis of Negative Human Behaviors on Social Media \\
	&	 21	&	Seethu Mariyam Christopher (UM), Intelligent Toys for Physical and Cognitive Assessments \\
	&	 22	&	Alexandra Sierra Rativa (TiU), Virtual Character Design and its potential to foster Empathy, Immersion, and Collaboration Skills in Video Games and Virtual Reality Simulations \\
	&	 23	&	Ilir Kola (TUD), Enabling Social Situation Awareness in Support Agents \\
	&	 24	&	Samaneh Heidari (UU), Agents with Social Norms and Values - A framework for agent based social simulations with social norms and personal values \\
	&	 25	&	Anna L.D. Latour (UL), Optimal decision-making under constraints and uncertainty \\
	&	 26	&	Anne Dirkson (UL), Knowledge Discovery from Patient Forums: Gaining novel medical insights from patient experiences \\
	&	 27	&	Christos Athanasiadis (UM), Emotion-aware cross-modal domain adaptation in video sequences \\
	&	 28	&	Onuralp Ulusoy (UU), Privacy in Collaborative Systems \\
	&	 29	&	Jan Kolkmeier (UT), From Head Transform to Mind Transplant: Social Interactions in Mixed Reality \\
	&	 30	&	Dean De Leo (CWI), Analysis of Dynamic Graphs on Sparse Arrays \\
	&	 31	&	Konstantinos Traganos (TU/e), Tackling Complexity in Smart Manufacturing with Advanced Manufacturing Process Management \\
	&	 32	&	Cezara Pastrav (UU), Social simulation for socio-ecological systems \\
	&	 33	&	Brinn Hekkelman (CWI/TUD), Fair Mechanisms for Smart Grid Congestion Management \\
	&	 34	&	Nimat Ullah (VUA), Mind Your Behaviour: Computational Modelling of Emotion \& Desire Regulation for Behaviour Change \\
	&	 35	&	Mike E.U. Ligthart (VUA), Shaping the Child-Robot Relationship: Interaction Design Patterns for a Sustainable Interaction \\

\midrule
2023
	&	 01	&	Bojan Simoski (VUA), Untangling the Puzzle of Digital Health Interventions \\
	&	 02	&	Mariana Rachel Dias da Silva (TiU), Grounded or in flight? What our bodies can tell us about the whereabouts of our thoughts \\
	&	 03	&	Shabnam Najafian (TUD), User Modeling for Privacy-preserving Explanations in Group Recommendations \\
	&	 04	&	Gineke Wiggers (UL), The Relevance of Impact: bibliometric-enhanced legal information retrieval \\
	&	 05	&	Anton Bouter (CWI), Optimal Mixing Evolutionary Algorithms for Large-Scale Real-Valued Optimization, Including Real-World Medical Applications \\
	&	 06	&	António Pereira Barata (UL), Reliable and Fair Machine Learning for Risk Assessment \\
	&	 07	&	Tianjin Huang (TU/e), The Roles of Adversarial Examples on Trustworthiness of Deep Learning \\
	&	 08	&	Lu Yin (TU/e), Knowledge Elicitation using Psychometric Learning \\
	&	 09	&	Xu Wang (VUA), Scientific Dataset Recommendation with Semantic Techniques \\
	&	 10	&	Dennis J.N.J. Soemers (UM), Learning State-Action Features for General Game Playing \\
	&	 11	&	Fawad Taj (VUA), Towards Motivating Machines: Computational Modeling of the Mechanism of Actions for Effective Digital Health Behavior Change Applications \\
	&	 12	&	Tessel Bogaard (VUA), Using Metadata to Understand Search Behavior in Digital Libraries \\
	&	 13	&	Injy Sarhan (UU), Open Information Extraction for Knowledge Representation \\
	&	 14	&	Selma Čaušević (TUD), Energy resilience through self-organization \\
	&	 15	&	Alvaro Henrique Chaim Correia (TU/e), Insights on Learning Tractable Probabilistic Graphical Models \\
	&	 16	&	Peter Blomsma (TiU), Building Embodied Conversational Agents: Observations on human nonverbal behaviour as a resource for the development of artificial characters \\
	&	 17	&	Meike Nauta (UT), Explainable AI and Interpretable Computer Vision – From Oversight to Insight \\
	&	 18	&	Gustavo Penha (TUD), Designing and Diagnosing Models for Conversational Search and Recommendation \\
	&	 19	&	George Aalbers (TiU), Digital Traces of the Mind: Using Smartphones to Capture Signals of Well-Being in Individuals \\
	&	 20	&	Arkadiy Dushatskiy (TUD), Expensive Optimization with Model-Based Evolutionary Algorithms applied to Medical Image Segmentation using Deep Learning \\
	&	 21	&	Gerrit Jan de Bruin (UL), Network Analysis Methods for Smart Inspection in the Transport Domain \\
	&	 22	&	Alireza Shojaifar (UU), Volitional Cybersecurity \\
	&	 23	&	Theo Theunissen (UU), Documentation in Continuous Software Development \\
	&	 24	&	Agathe Balayn (TUD), Practices Towards Hazardous Failure Diagnosis in Machine Learning \\
	&	 25	&	Jurian Baas (UU), Entity Resolution on Historical Knowledge Graphs \\
	&	 26	&	Loek Tonnaer (TU/e), Linearly Symmetry-Based Disentangled Representations and their Out-of-Distribution Behaviour \\
	&	 27	&	Ghada Sokar (TU/e), Learning Continually Under Changing Data Distributions \\
	&	 28	&	Floris den Hengst (VUA), Learning to Behave: Reinforcement Learning in Human Contexts \\
	&	 29	&	Tim Draws (TUD), Understanding Viewpoint Biases in Web Search Results \\

\midrule
2024
	&	 01	&	Daphne Miedema (TU/e), On Learning SQL: Disentangling concepts in data systems education \\
	&	 02	&	Emile van Krieken (VUA), Optimisation in Neurosymbolic Learning Systems \\
	&	 03	&	Feri Wijayanto (RUN), Automated Model Selection for Rasch and Mediation Analysis \\
	&	 04	&	Mike Huisman (UL), Understanding Deep Meta-Learning \\
	&	 05	&	Yiyong Gou (UM), Aerial Robotic Operations: Multi-environment Cooperative Inspection \& Construction Crack Autonomous Repair \\
	&	 06	&	Azqa Nadeem (TUD), Understanding Adversary Behavior via XAI: Leveraging Sequence Clustering to Extract Threat Intelligence \\
	&	 07	&	Parisa Shayan (TiU), Modeling User Behavior in Learning Management Systems \\
	&	 08	&	Xin Zhou (UvA), From Empowering to Motivating: Enhancing Policy Enforcement through Process Design and Incentive Implementation \\
	&	 09	&	Giso Dal (UT), Probabilistic Inference Using Partitioned Bayesian Networks \\
	&	 10	&	Cristina-Iulia Bucur (VUA), Linkflows: Towards Genuine Semantic Publishing in Science \\
	&	 11	&	withdrawn \\
	&	 12	&	Peide Zhu (TUD), Towards Robust Automatic Question Generation For Learning \\
	&	 13	&	Enrico Liscio (TUD), Context-Specific Value Inference via Hybrid Intelligence \\
	&	 14	&	Larissa Capobianco Shimomura (TU/e), On Graph Generating Dependencies and their Applications in Data Profiling \\
	&	 15	&	Ting Liu (VUA), A Gut Feeling: Biomedical Knowledge Graphs for Interrelating the Gut Microbiome and Mental Health \\
	&	 16	&	Arthur Barbosa Câmara (TUD), Designing Search-as-Learning Systems \\
	&	 17	&	Razieh Alidoosti (VUA), Ethics-aware Software Architecture Design \\
	&	 18	&	Laurens Stoop (UU), Data Driven Understanding of Energy-Meteorological Variability and its Impact on Energy System Operations \\
	&	 19	&	Azadeh Mozafari Mehr (TU/e), Multi-perspective Conformance Checking: Identifying and Understanding Patterns of Anomalous Behavior\\
	&	 20	&	Ritsart Anne Plantenga (UL), Omgang met Regels \\
	&	 21	&	Federica Vinella (UU), Crowdsourcing User-Centered Teams \\
	&	 22	&	Zeynep Ozturk Yurt (TU/e), Beyond Routine: Extending BPM for Knowledge-Intensive Processes with Controllable Dynamic Contexts \\
	&	 23	&	Jie Luo (VUA), Lamarck’s Revenge: Inheritance of Learned Traits Improves Robot Evolution \\
	&	 24	&	Nirmal Roy (TUD), Exploring the effects of interactive interfaces on user search behaviour \\
	&	 25	&	Alisa Rieger (TUD), Striving for Responsible Opinion Formation in Web Search on Debated Topics \\
	&	 26	&	Tim Gubner (CWI), Adaptively Generating Heterogeneous Execution Strategies using the VOILA Framework \\
	&	 27	&	Lincen Yang (UL), Information-theoretic Partition-based Models for Interpretable Machine Learning \\
	&	 28	&	Leon Helwerda (UL), Grip on Software: Understanding development progress of Scrum sprints and backlogs \\

% \bottomrule
\end{xltabular}